\pdfoutput=1
\documentclass[11pt,openany]{book}
\usepackage{acl}

\usepackage{titlesec}
\titleformat{\chapter}[display]
  {\normalfont\huge\bfseries} % change \huge to any size you prefer
  {\chaptertitlename\ \thechapter}
  {20pt}
  {\Large} % Title text size

\makeatletter
\renewcommand{\maketitle}{%
    \begin{titlepage}
        \centering
        {\Huge\bfseries\@title\par} % Set the title size here
        \vspace{2cm}
        {\Large\@author\par}
        \vspace{1cm}
        {\Large\@date\par}
    \end{titlepage}
}
\makeatother

\usepackage{fancyhdr}
\usepackage{lipsum}  % for generating dummy text

\fancypagestyle{plain}{
    \fancyhf{}
    \fancyfoot[C]{\thepage}
    \fancyhead[LE,RO]{\thepage}
    \fancyhead[LO]{\nouppercase{\rightmark}}
    \fancyhead[RE]{\nouppercase{\leftmark}}
    
}

\fancypagestyle{plain}{
    \fancyhf{}  % Clear all header and footer fields
    \fancyfoot[C]{\thepage}  % Only the page number centered in the footer
}

\usepackage{float}
\usepackage{avm}

\usepackage{times}
\usepackage{latexsym}
\usepackage[T4,T2A,T1]{fontenc}
\newcommand{\AS}[1]{{\fontencoding{T4}\fontfamily{fcr}\selectfont#1}}
\usepackage[russian,english]{babel}
\usepackage{substitutefont}
\substitutefont{T2A}{\familydefault}{Tempora-TLF}

\usepackage{CJKutf8}

\usepackage[utf8]{inputenc}

\usepackage{microtype}

\usepackage{inconsolata}

\usepackage{times}
\usepackage{latexsym}
\usepackage{tabularx}
\usepackage{graphicx}
\usepackage{todonotes}
\usepackage{makecell}
\usepackage{amsmath}
\usepackage{amssymb}
\usepackage{covington}
\usepackage{subcaption}
\usepackage{tipa}

\usepackage{hyperref}
\usepackage{booktabs}
\usepackage{bookmark}
\usepackage{float}
\usepackage{covington}	%For glossing—We recommend covington.
\usepackage{qtree}	%For drawing trees—we recommend qtree.
\usepackage{tree-dvips}
\usepackage{longtable}
\usepackage{diagbox}
\usepackage{xr}
\usepackage{stackengine}
\usepackage{rotating}
\usepackage{multirow}

\usepackage[list=true,listformat=simple]{subcaption}
\makeatletter
\renewcommand*\l@subfigure{\@dottedtocline{1}{1.5em}{2.5em}}
\makeatother

\newcommand{\posscitet}[1]{\citeauthor{#1}'s (\citeyear{#1})}
\newcommand{\possciteauthor}[1]{\citeauthor{#1}'s}
\newcommand{\pgposscitet}[2]{\citeauthor{#1}'s (\citeyear{#1}:~#2)}

\newcommand{\pgcitealt}[2]{\citealt{#1}:~#2}
\newcommand{\seccitealt}[2]{\citealt{#1}:~$\S$#2}

\newcommand{\pgcitet}[2]{\citeauthor{#1} (\citeyear{#1}:~#2)}
\newcommand{\seccitet}[2]{\citeauthor{#1} (\citeyear{#1}:~$\S$#2)}

\title{A quantitative and typological study of Early Slavic participle clauses and their competition}

\author{Nilo Pedrazzini \\
  University of Oxford (Oxford, United Kingdom)\\
  \texttt{nilo.pedrazzini@ling-phil.ox.ac.uk}}

\begin{document}
\maketitle
\begin{abstract}
This thesis investigates the semantic and pragmatic properties of Early Slavic participle constructions (conjunct participles and dative absolutes) to understand the principles motivating their selection over one another and over their main finite competitor (\textit{jegda}-clauses). The issue is tackled by adopting two broadly different approaches, which inform the division of the thesis into two parts. 

The first part of the thesis uses detailed linguistic annotation on Early Slavic corpora at the morphosyntactic, dependency, information-structural, and lexical levels to obtain indirect evidence for different potential functions of participle clauses and their main finite competitor. The goal of this part of the thesis is to understand the roles of compositionality and default discourse reasoning as explanations for the distribution of participle constructions and \textit{jegda}-clauses in the Early Slavic corpus. The investigation shows that the competition between conjunct participles, absolute constructions, and \textit{jegda}-clauses occurs at the level of discourse organization, where the main determining factor in their distribution is the distinction between \textit{background} and \textit{foreground} content of an (elementary or complex) discourse unit. The analysis also shows that the major common denominator between the three constructions is that all of them can function as frame-setting devices (i.e. background clauses), albeit to very different extents. In fact, conjunct participles are more typically associated with the foreground constituent of a discourse unit, whereas dative absolutes and \textit{jegda}-clauses are typically associated with the background content. 

The second part of the thesis uses massively parallel data, including Old Church Slavonic and Ancient Greek, and analyses typological variation in how languages express the semantic space of English \textit{when}, whose scope encompasses that of Early Slavic participle constructions and \textit{jegda}-clauses. To do so, probabilistic semantic maps are generated and statistical methods (including Kriging, Gaussian Mixture Modelling, precision and recall analysis) are used to induce cross-linguistically salient dimensions from the parallel corpus and to study conceptual variation within the semantic space of the hypothetical concept \textsc{when}.  Clear typological correspondences and differences with Early Slavic from linguistic phenomena in other languages are then exploited to corroborate and refine observations made on the core semantic-pragmatic properties of participle constructions and \textit{jegda}-clauses on the basis of annotated Early Slavic data. 

The analysis shows that `null’ constructions (juxtaposed clauses such as participles and converbs, or independent clauses) consistently cluster in particular regions of the semantic map cross-linguistically, which clearly indicates that participle clauses are not equally viable as alternatives to any use of \textsc{when}, but carry particular meanings that make them less suitable for some of its functions. The investigation helped identify genealogically and areally unrelated languages that seem typologically very similar to Old Church Slavonic in the way they divide the semantic space of \textsc{when} between overtly subordinated and `null’ constructions. Comparison with these languages reveals great similarities between the functions of Early Slavic participle constructions and of linguistic phenomena in some of these languages (particularly clause chaining, bridging, insubordination, and switch reference). Crucially, new clear correspondences are found between these phenomena and `non-canonical’ usages of participle constructions (i.e. coreferential dative absolutes, syntactically independent absolutes and conjunct participles, and participle constructions with no apparent matrix clause), which had often been written off as `aberrations’ by previous literature on Early Slavic. 

\end{abstract}

\noindent \textbf{Citation Note:} This manuscript is a version of the author's PhD thesis (\citealt{pedrazziniphd}), formatted for readability and broader dissemination. Please cite this work as: Pedrazzini, Nilo. 2023. \textit{A quantitative and typological study of Early Slavic participle clauses and their competition}. Ph.D. thesis, University of Oxford. DOI: \href{http://dx.doi.org/10.5287/ora-8gv0b4qyo}{10.5287/ora-8gv0b4qyo}.

\tableofcontents

\chapter*{List of Abbreviations}

Glossing and abbreviations follow the Leipzig Glossing Rules, with the following additions.\\ 

\begin{tabular}{ll}
\textsc{a} &  transitive subject function; Actor macrorole; Actor\\
\textsc{aor} &  aorist\\
\textsc{coll} &  collective\\
\textsc{ds} &  different subject\\
\textsc{exist} &  existential\\
\textsc{impf} &  imperfect\\
\textsc{narr} &  narrative\\
\textsc{nsbj} &  non-subject\\
\textsc{pred} &  predicate marker\\
\textsc{ptc} &  discourse particle\\
\textsc{remp} &  remote past tense\\
\textsc{result} &  resultative\\
\textsc{seq} &  sequence\\
\textsc{sim} &  simultaneous\\
\textsc{ss} &  same subject\\
\textsc{sup} &  supine\\
\textsc{todp} &  today's past tense\\
\textsc{vis} &  visible, speaker’s area\\
\textsc{u} &  undergoer macrorole\\
\textsc{univ} &  universal\\
\end{tabular}

\clearpage
\chapter*{Introduction}
\markboth{Introduction}{}
\addcontentsline{toc}{chapter}{Introduction}

\section{The problem}

Participial clauses are employed widely in Early Slavic and are in evident competition with finite clauses introduced by the generic temporal subordinator \textit{jegda} `when' (\ref{parallelocsexx1})-(\ref{parallelocsexx2}).\footnote{Citations under the examples use italics for the (conventional) title of the text and regular font for the name of the manuscript. Citations of the Early Slavonic New Testament are all from the Old Church Slavonic Codex Marianus in PROIEL, so the manuscript name is omitted throughout the thesis.  Citations of the Greek New Testament are instead all from \posscitet{tischen} version in PROIEL. Translations of biblical passages are mostly adapted from the King James Version and the New King James Version. Translations of all the examples from the \textit{Life of Mary of Egypt} from the Bdinski Sbornik are by \citealt{coelhomstthesis}. Translations of other examples, both Early Slavic and other languages, are my own unless otherwise noted.}

\begin{example}
    \begin{itemize}
        \item[a.] 
        \gll \textbf{viděv} že ḟeklu celova ju
        see.\textsc{ptcp.pfv.m.nom.sg} \textsc{ptc} Thecla.\textsc{acc} kiss.\textsc{aor.3.sg} \textsc{3.sg.f.acc}
        \glt ‘When he saw Thecla, he kissed her' (\textit{Life of Thecla}, Bdinski Sbornik f. 50v)
        \glend
        \item[b.] 
        \gll \textbf{Egda} že se \textbf{vidě} Zosima, too v\foreignlanguage{russian}{ь} veštši strach v\foreignlanguage{russian}{ь}pade
        when \textsc{ptc} this.\textsc{n.acc.sg} see.\textsc{aor.3.sg} Zosima.\textsc{nom} then in greater.\textsc{acc} terror.\textsc{acc} fall.\textsc{aor.3.sg}
        \glt ‘When Zosima saw this, he fell in a greater terror' (\textit{Life of Mary of Egypt}, Bdinski Sbornik ff. 169r-169v)
        \glend
        \label{parallelocsexx1}
    \end{itemize}
\end{example}

\begin{example}
\begin{itemize}
    \item[a.]
    \gll tože takože i ot boljar\foreignlanguage{russian}{ъ} m\foreignlanguage{russian}{ъ}nozi \textbf{prichodęšte} povědachut\foreignlanguage{russian}{ь} jemu gněv\foreignlanguage{russian}{ъ} knęž\foreignlanguage{russian}{ь} na togo sušt\foreignlanguage{russian}{ь} i molęchut\foreignlanguage{russian}{ь} i ne suprotiviti sę jemu
    also so {\sc ptc} from Boyars.{\sc gen.pl} many.{\sc nom.pl} come.{\sc ptcp.ipfv.nom.pl} speak.{\sc impf.3.pl} {\sc 3.sg.m.dat} anger.{\sc acc.sg} prince's.{\sc acc.sg} on that.{\sc gen.sg} be.{\sc ptcp.ipfv.acc.sg} and beg.{\sc impf.3.pl} {\sc ptc} {\sc neg} oppose {\sc refl} {\sc 3.sg.m.dat}
    \glt ‘And many of the Boyars, when they came to him, would speak of the prince's anger and begged not to oppose the prince.’ (\textit{Life of Feodosij Pečerskij}, Uspenskij Sbornik f. 58v)
    \glend
    \item[b.]
    \gll i \textbf{jegda} že sii \textbf{prichožaachu} k\foreignlanguage{russian}{ъ} njemu to že si i tako po bž\foreignlanguage{russian}{ь}stv\foreignlanguage{russian}{ь}něm\foreignlanguage{russian}{ь} tom\foreignlanguage{russian}{ь} učenii prěd\foreignlanguage{russian}{ъ}stavljaaše těm\foreignlanguage{russian}{ъ} trępezu ot braš\foreignlanguage{russian}{ь}n\foreignlanguage{russian}{ъ} těch\foreignlanguage{russian}{ъ} manastyr\foreignlanguage{russian}{ь}skyich\foreignlanguage{russian}{ъ}
    and when {\sc ptc} this.{\sc nom.pl} come.{\sc impf.3.pl} to {\sc 3.sg.m.dat} then {\sc ptc} this.{\sc nom.sg} {\sc ptc} so after divine.{\sc loc.sg} that.{\sc loc.sg} teaching.{\sc loc.sg} treat.{\sc impf.3.sg} that.{\sc dat.pl} {dining table.{\sc acc.sg}} from food.{\sc gen.pl} that.{\sc gen.pl} monastery's.{\sc gen.pl}
    \glt ‘And when someone came to Theodosius, in the same way, after the spiritual conversation, he would treat those who came to dinner with the supplies of the monastery.’ (\textit{Life of Feodosij Pečerskij}, Uspenskij Sbornik f. 46b)
    \glend
\end{itemize}
\label{parallelocsexx2}
\end{example}

The focus of previous scholarship on Early Slavic participle clauses has been overwhelmingly on their origin and relationship with their Ancient Greek counterparts (e.g. \citealt{birnbaum1958a, r1958a, r1961a, r1963a, ve1961a, ve1997a, skupskij1993a}), which show very similar functions, rather than on the motivations governing their distribution from the synchronic perspective in the Early Slavic corpus. Several works (e.g. \citealt{trost, r1961a, andersen, berent, worth1994a, corin1995a,collins2004a,collins2011a}) have also been dedicated to the specific functions of Early Slavic absolute constructions, a type of participle clause structurally akin to English absolutes (e.g. \textit{all things considered} or \textit{weather permitting}), but which receives special dative marking, both on the participle and on its subject, if overt. Absolute constructions are regularly found in distributional overlap with \textit{jegda}-clauses, as in (\ref{egdacompdas}). As (\ref{xadvcompdas}) shows, particularly sentence-initially, competition can also be observed between absolute constructions and agreeing participle clauses (`conjunct participles').

\begin{example}
    \begin{itemize}
        \item[a.] 
        \gll i \textbf{koncęjuštju} \textbf{sę} lět tomu. vygnaša žiroslava is posadnic\foreignlanguage{russian}{ь}stva. i daša zavidu nerevinicju
        {and} {finish.{\sc ptcp.ipfv.n.dat.sg}} {\sc{refl}} {year.{\sc dat.sg}} {that.{\sc dat.sg}} {drive out.{\sc aor.3.pl}} {Žiroslav.{\sc gen.sg}} {from} {posadnichestvo.{\sc gen.sg}} {and} {give.{\sc aor.3.pl}} {Zavid.{\sc dat.sg}} {Nerenevic.{\sc dat.sg}}
        \glt `And towards the end of that year [lit. when that year was ending], they drove Žiroslav out of the \textit{posadnichestvo} and gave it to Zavid Nerenevic' (\textit{Novgorod First Chronicle}, Synodal Manuscript f. 40r) %270183
        \glend
        \item[b.] 
        \gll \textbf{egda} \textbf{končaša} \textbf{sę} dn\foreignlanguage{russian}{ь}e {.e. desęt\foreignlanguage{russian}{ь}nii}. snide dch\foreignlanguage{russian}{ъ} styi na aply
        {when} {finish.{\sc aor.3.pl}} {\sc{refl}} {day.{\sc gen.pl}} {fiftieth.{\sc nom.pl}} {descend.{\sc aor.3.sg}} {spirit.{\sc nom.sg}} {holy.{\sc nom.sg}} {on} {apostle.{\sc acc.pl}}
        \glt `When fifty days had passed, the Holy Spirit descended on the apostles' (\textit{Primary Chronicle}, Codex Laurentianus f. 35v) %127295
        \glend
        \label{egdacompdas}
    \end{itemize}
\end{example}

\begin{example}
    \begin{itemize}
        \item[a.] 
        \gll i \textbf{prišedšju} emu v korsun\foreignlanguage{russian}{ь} uvidě jako is korsunę bliz\foreignlanguage{russian}{ь} ust\foreignlanguage{russian}{ь}e dněpr\foreignlanguage{russian}{ь}skoe
        {and} {arrive.{\sc ptcp.pfv.m.dat.sg}} {\sc 3.sg.m.dat} {in} {Korsun.{\sc acc}} {learn.{\sc aor.3.sg}} {that} {from} {Korsun.{\sc gen}} {close.{\sc nom}} {mouth.{\sc nom}} {Dniepr's.{\sc nom}} 
        \glt ‘When he arrived in Korsun, he learned that the mouth of the Dnieper was not far from Korsun' (\textit{Primary Chronicle}, Codex Laurentianus f. 3v)%262264
        \glend
        \item[b.] 
        \gll i \textbf{prišed} v\foreignlanguage{russian}{ъ} vizantiju vidě na tom\foreignlanguage{russian}{ъ} městě z gor\foreignlanguage{russian}{ъ} i glušic\foreignlanguage{russian}{ь} morskych\foreignlanguage{russian}{ъ} mnogo
        {and} {arrive.{\sc ptcp.pfv.m.nom.sg}} {in} {Byzantium.{\sc acc}} {see.{\sc aor.3.sg}} {in} {that.{\sc loc.sg}} {place.{\sc loc.sg}} {seven} {hill.{\sc gen.pl}} {and} {bay.{\sc gen.pl}} {sea.{\sc gen.pl}} {much.{\sc acc.sg}}
        \glt ‘And when he arrived in Byzantium, he saw seven hills and many sea gulfs' (\textit{The Tale of the Fall of Constantinople} f. 283r) 
        \glend
        \label{xadvcompdas}
    \end{itemize}
\end{example}

To pinpoint the factors explaining the distribution of competing constructions, one needs to draw from several levels of analysis at once. Consider (\ref{englishfirstexx}a)-(\ref{englishfirstexx}d).

\begin{example}
    \begin{itemize}
        \item[a.] Having finished his thesis, he cried with relief
        \item[b.] His thesis (being) finished, he cried with relief
        \item[c.] When he finished his thesis, he cried with relief
        \item[d.] He finished his thesis and cried with relief
    \end{itemize}
    \label{englishfirstexx}
\end{example}

To various extents, all these sentences relate two clauses in semantically underspecified ways. The nonfiniteness of the participle clause in (\ref{englishfirstexx}a) and (\ref{englishfirstexx}b) indicates dependency on the following (matrix) clause. In the absence of an explicit connective, however, their semantic relation must be pragmatically inferred, often similarly to syntactically coordinated independent clauses (\ref{englishfirstexx}d). In (\ref{englishfirstexx}c) the first finite clause is introduced by the temporal subordinating conjunction \textit{when}. Yet, the precise temporal relation between the \textit{when}-clause and the matrix clause is not established by the subjunction itself, since, unlike more explicit subjunctions, such as \textit{after} or \textit{before}, \textit{when} is temporally underspecified and compatible with different core temporal interpretations (e.g. anteriority, simultaneity, inclusion). Whichever the surface realization, however, all these examples intuitively convey that the \textit{crying} event is an effect of the \textit{thesis-finishing} event, and that the second situation occurs at some point (sooner or later) after the first situation. \\
\indent On one level, we may consider the compositional semantics of participle clauses at the clause- and sentence-level and understand how far compositionality can go in explaining functional differences between constructions. As we will see, the core temporal semantic properties of Early Slavic participle clauses, for example, can be accounted for, to a large extent, by the compositional semantics of the tense-aspect system of Early Slavic, specifically how the aspect of participles interacts with their matrix clause. On another level, the use of different participle clauses or \textit{jegda}-clauses entails the strategic selection of one construction over another at the level of discourse structure. This involves temporal relations \textit{across} sentences and the broader level of text organization, in which information structure and the rhetorical relations between \textit{discourse} units play a crucial role.\\
\indent The goal of this thesis is to investigate the semantic and pragmatic properties of Early Slavic conjunct participles and absolute constructions to understand what principles motivate their selection over one another and over \textit{jegda}-clauses at the synchronic level. The issue is tackled broadly using two different approaches. First, it uses detailed linguistic annotation on Early Slavic corpora at the morphosyntactic, dependency, information-structural, and lexical level, as well as parallel Ancient Greek data (when available) aligned at the token level, to obtain indirect evidence for different potential functions of participle clauses. This approach is broadly concerned with the aforementioned tension between compositionality and default discourse reasoning in accounting for the distribution of participle constructions and \textit{jegda}-clauses in the Early Slavic corpus. Second, it uses `massively parallel' data (\citealt{mayer-cysouw}) and leverages typological variation in how languages express the semantic space covered by these constructions to assess the extent to which the division of labour observed among participle constructions and \textit{jegda}-clauses is reflected cross-linguistically. Clear correspondences and differences with Early Slavic are then exploited to make generalizations about the core semantic-pragmatic properties of different competing temporal constructions in Early Slavic.\\
\indent In the rest of this introduction, I first provide a preliminary formal definition of conjunct participles and absolute constructions (Section \ref{definitions}) and introduce the main theoretical frameworks to which this thesis will make consistent reference to. In particular, Section \ref{frameworks} summarizes \posscitet{baryhaug2011} compositional-semantic account of the temporal semantics and discourse properties of conjunct participles in Ancient Greek, on which the first three chapters of the thesis will be based, while Section \ref{framesetterssec} touches on \posscitet{charolles05} \textit{framing adverbials}, \posscitet{fabricius-hansen2012b} \textit{fronted adjuncts}, and \posscitet{chafe1976a} and \posscitet{krifka2007a} \textit{frame setters}. Section \ref{sdrtintro} introduces Segmented Discourse Representation Theory, a formal framework for discourse representation which will be used as a tool to reason about and formalize rhetorical relations between clauses beyond compositional-semantic accounts such as \posscitet{baryhaug2011}. Section \ref{whatswithwhen} gives an overview of some of the main issues in the interpretation of \textit{when}-clauses, on the basis of previous extensive research on event structure, temporal relations, and the anaphoric properties of tense, while Section \ref{tokenbased} elaborates on the need for a gradient approach to the study of competition and introduces the concept of token-based typology, which is central to the methodology used in the second part of the thesis. Finally, Section \ref{corpus} presents the Early Slavic corpus used in this work and Section \ref{thesisoverview} summarizes the structure of the thesis.

\section{Background}\label{background}
\subsection{Defining conjunct participles and dative absolutes}\label{definitions}
Under a Lexical-Functional Grammar (LFG) framework, the participle constructions under analyses are classified as \textit{adjuncts}, namely grammatical functions that are not subcategorizable (i.e. required as arguments) by a predicate (\pgcitealt{dalrymple2019}{12}).\\
\indent We can distinguish two main types of participial adjuncts in Early Slavic. In so-called (\textit{co-})\textit{predicative}\footnote{The term \textit{copredicative} participles is used by \pgcitet{haspelmathconverbs}{17–20}, rather than only \textit{predicative}, used, for example, by \citet{breakingdowneckhoff}, \citet{baryhaug2011} and \citet{haug2011a}.} or \textit{conjunct} participles, the subject is coreferential with an argument of the matrix VP, most typically the subject, which determines the case of the participle form itself. In other words, conjunct participles are \textit{functionally controlled} by an argument of the matrix clause. Control constructions are defined in LFG as those in which `either syntactic or lexical constraints require coreference between an argument of the matrix clause [controller] and an argument of a subordinate or modifying adjunct clause [controllee]’ (\pgcitealt{dalrymple2019}{543}). In functional (as opposed to ana\-phoric) control, the case of the controllee is determined by the grammatical function (e.g. \textsc{sub}ject, \textsc{obj}ect, etc.) of the controller. \\
\indent (\ref{xadvintro1})-(\ref{xadvintro3}) are typical examples of conjunct participles.

\begin{example}
\gll az\foreignlanguage{russian}{ъ} \textbf{prišed\foreignlanguage{russian}{ъ}} iscěljǫ i 
\textsc{1.sg.nom} come.\textsc{ptcp.pfv.m.nom.sg} heal.\textsc{prs.1.sg} \textsc{3.sg.m.acc}
\glt ‘I will come and heal him’ (Matthew 8:7)
\glend
\label{xadvintro1}
\end{example}

\begin{example}
\gll \textbf{vykradše} pervoe svętoslava po tom\foreignlanguage{russian}{ъ} ubiša kytana i družinu jego vsju izbiša
kidnap.{\sc ptcp.pfv.m.nom.pl} first Svyatoslav.{\sc gen} after that.{\sc loc.sg} kill.{\sc aor.3.pl} Kytan.{\sc gen} and retinue.{\sc acc.sg} {\sc 3.sg.m.gen} entire.{\sc acc.sg} thrash.{\sc aor.3.pl}
\glt ‘After kidnapping Svyatoslav first, they killed Kytan and thrashed his entire retinue’ (\textit{Primary Chronicle}, Codex Laurentianus f. 75d)
\glend
\label{xadvintro2}
\end{example}

\begin{example}
\gll i \textbf{padši} kolěnoma i čelom\foreignlanguage{russian}{ь} {\textbf{tl\foreignlanguage{russian}{ь}kjuštii} {\normalfont [sic]}} na zemli prěd ikonu styje bce i sia slovesa načech glati
and fall.{\sc ptcp.pfv.f.nom.sg} knee.{\sc inst.du} and brow.{\sc inst.sg} touch.{\sc ptcp.ipfv.f.nom.sg} on ground.{\sc loc.sg} before icon.{\sc acc.sg} holy.{\sc gen.sg} {Mother of God}.{\sc gen.sg} and this.{\sc acc.pl} word.{\sc acc.pl} start.{\sc aor.1.sg} speak.{\sc inf} 
\glt ‘Falling on my knees and touching the ground before the icon of the holy Mother of God with my brow, I started speaking the following words' (\textit{Life of Mary of Egypt}, Bdinski Sbornik f. 177r).
\glend
\label{xadvintro3}
\end{example}

Conjunct participles are thus adjuncts with an external argument, an `open slot' which needs to be filled---in LFG terms, they are \textsc{xadj}uncts of verb-headed (as opposed to noun-headed, as in attributive uses) $f$-structures (as in (\ref{xadjfstr}), from example (\ref{xadvintro1})), or verb-headed \textit{open adjuncts} in other frameworks (e.g. \citealt{bigeventsbook}). 

\begin{example}
\label{xadjfstr}
\centering
\begin{avm}
\[  \sc pred  & `\it icěliti \< \sc subj,obj \>' \\
    \sc asp &  \sc pfv  \\
    \sc tense &  \sc prs  \\
    \sc subj  &  \@1 \[ \sc pred & `\it az\foreignlanguage{russian}{ъ}'\]  \\
    \sc obj   &  \[ \sc pred & `\it i' \] \\
    \sc xadj  & \[  \sc pred & `\it priti \< \sc subj \>'  \\
                    \sc asp &  \sc pfv  \\
                    \sc subj   &  \@1\] \]
\end{avm}
\end{example}

\textit{Absolute constructions}, on the other hand, are not functionally controlled by any argument of the matrix VP. They thus have an internal subject which needs not be coreferential with any constituent in the superordinate clause. As in several other early Indo-European languages (e.g. Ancient Greek, Latin, Gothic, Sanskrit), both the participle and its subject generally appear in oblique cases, dative in Early Slavic, which is why we often refer to this construction as \textit{dative absolute}. (\ref{absintro1})-(\ref{absintro3}) are typical examples of dative absolutes from the same sources as (\ref{xadvintro1})-(\ref{xadvintro3}).

\begin{example}
\gll i \textbf{s\foreignlanguage{russian}{ъ}chodęštem\foreignlanguage{russian}{ъ}} im\foreignlanguage{russian}{ъ} s\foreignlanguage{russian}{ъ} gory zapovědě im\foreignlanguage{russian}{ъ} is\foreignlanguage{russian}{ъ} glę
and come.down.\textsc{ptcp.ipfv.m.dat.pl} \textsc{3.pl.dat} from mountain.\textsc{gen.sg} command.\textsc{aor.3.sg} \textsc{3.pl.dat} Jesus.\textsc{nom} say.\textsc{ptcp.prs.m.nom.sg} 
\glt ‘As they came down from the mountain, Jesus commanded them saying' (Matthew 17:9)
\glend
\label{absintro1}
\end{example}

\begin{example}
\gll i \textbf{umnoživšem\foreignlanguage{russian}{ъ}} sę člvkom\foreignlanguage{russian}{ъ} na zemli pomysliša sozdati stolp\foreignlanguage{russian}{ъ} do nbse v\foreignlanguage{russian}{ъ} dni nektana i faleka
and multiply.{\sc ptcp.pfv.m.dat.pl} \textsc{refl} person.{\sc dat.m.pl} on earth.{\sc loc.sg} plan.{\sc aor.3.pl} create.{\sc inf} pillar.{\sc acc.sg} to heaven.{\sc gen.sg} in day.{\sc acc.pl} Nectan.{\sc gen.sg} and Peleg.{\sc gen.sg} 
\glt ‘And when people had multiplied on earth, they planned to build a pillar to heaven in the days of Nectan and Peleg.’ (The \textit{Primary Chronicle}, Codex Laurentianus f. 2v) %123324)
\glend
\label{absintro2}
\end{example}

\begin{example}
\gll lětu \textbf{skončavšu} se i priide paki zosima v\foreignlanguage{russian}{ь} pustynju
year.\textsc{dat.n.sg} end.\textsc{ptcp.pfv.n.dat.sg} \textsc{refl} \textsc{ptc} come.\textsc{aor.3.sg} again Zosima.\textsc{nom} in desert.\textsc{acc.sg}
\glt ‘When the year was ending Zosima came again to the desert’ (\textit{Life of Mary of Egypt}, Bdinski Sbornik f. 187r)
\glend
\label{absintro3}
\end{example}

If conjunct participles are open adjuncts, then absolutes can be considered \textit{closed} adjuncts. In LFG terms, they are \textsc{adj}uncts of verb-headed \textit{f}-structures, as represented in (\ref{absfstr}) from example (\ref{absintro1}).

\begin{example}
\label{absfstr}
\centering
\begin{avm}
\[  \sc pred  & `\it zapověděti \< \sc subj, obj, obl$_{\textsc{goal}}$ \>'\\
    \sc asp &  \sc pfv  \\
    \sc tense &  \sc pst  \\
    \sc subj  &  \[ \sc pred & `\it Isus\foreignlanguage{russian}{ъ}'\]  \\
    \sc obl$_{\textsc{goal}}$ &  \[ \sc pred & `\it i*'$_{\@1}$ \] \\
    \sc obj &  \[ ...\] \\
    \sc adj  & \[  \sc pred & `\it s\foreignlanguage{russian}{ъ}choditi \< \sc subj,obl$_{\textsc{source}}$ \>'  \\
                    \sc asp &  \sc ipfv  \\
                    \sc subj   &  \[ \sc pred & `\it i*'$_{\@1}$ \] \\
                    \sc obl$_{\textsc{source}}$ & \[ \sc pred & `\it s\foreignlanguage{russian}{ъ} \< \sc obj \>'\\
                                                     \sc obj & \[ \sc pred & `\it gora' \]
                                                        \]\\
                    \sc xadj &  \[ \sc pred & `\it glagolati \< \sc subj\>' \\
                                   \sc asp & \sc ipfv \\
                                   \sc subj & \@1 \] \] \]
\end{avm}
\end{example}

Traditionally, subject non-coreferentiality (or \textit{switch reference)} between the absolute construction and the matrix clause has been considered obligatory, but, as it has often been pointed out, this is by no means always the case in the Early Slavic sources (cf. \citealt{worth1994a,collins2004a,collins2011a}), similarly to what we also independently observe in Ancient Greek (cf. \citealt{fuller}). (\ref{abssamesubj}) and (\ref{abssamesubj2}) are examples of subject co-reference in Early Slavic dative absolutes.

\begin{example}
\gll i byst\foreignlanguage{russian}{ъ} \textbf{idǫštem\foreignlanguage{russian}{ъ}} im\foreignlanguage{russian}{ь}. ištistišę sę
and happen.\textsc{aor.3.sg} go.\textsc{ptcp.ipfv.dat.pl} \textsc{3.pl.dat} cleanse.\textsc{aor.3.pl} \textsc{refl}
\glt ‘And it came to pass that, as they went, they were cleansed’ (Luke 17:14)
\glend
\label{abssamesubj}
\end{example}

\begin{example}
\gll i \textbf{všedšim} v gorod\foreignlanguage{russian}{ъ} utěšista volodimercě
{and} {enter.{\sc ptcp.pfv.dat.pl}} {in} {city.{\sc acc.sg}} {console.{\sc aor.3.du}} {people of Vladimir.{\sc acc.pl}}
\glt ‘And when they entered the city, they consoled the people of Vladimir’ (\textit{Suzdal Chronicle}, Codex Laurentianus f. 126r) %275330
\glend
\label{abssamesubj2}
\end{example}

Such cases of subject co-referentiality have sometimes been called `nonstandard' (\citealt{worth1994a}) or `non-canonical' (\citealt{collins2011a}), if not outright `aberrations' or signs of `decay' (e.g. \pgcitealt{ve1961a}{49}; \pgcitealt{ve1996}{190}; \pgcitealt{corin1995a}{268}). There are other frequently observed instances of `non-canonical' absolutes (as well as conjunct participles), most notably those which are overtly coordinated to their matrix or which do not have a clear matrix clause they attach to. Conjoined absolutes have been commented on particularly in the literature on Old East Slavic (cf., for example, \citealt{necasek, ve1961a, borkovskijkuznecov, alekseev87, gebert, remneva, corin1995a}), but it is in fact well attested in South Slavic recensions of Church Slavonic as well (cf. \citealt{grkovic07} and \citealt{collins2011a}). As argued by \citet{collins2011a}, several `non-canonical' absolutes should be interpreted by taking into account the broader discourse context rather than the inter-clausal or inter-sentential level. As I will show in the second part of this thesis, these occurrences are not at odds with the core functions of dative absolutes as discourse devices. In fact, most, if not all, the non-canonical usages of absolutes previously observed in the literature are phenomena frequently observed to co-occur cross-linguistically on constructions with similar functions as Early Slavic absolutes.

\subsection{Conjunct participles according to Bary \& Haug 2011}\label{frameworks}
According to \cite{baryhaug2011}, conjunct participles in Ancient Greek fulfil one of three main discourse functions (labelled as \textsc{Frames}, \textsc{Independent Rhemes}, \textsc{Elaborations}), each involving a specific temporal relation to the matrix clause and certain discourse effects that can be modelled at the compositional level. As the corpus evidence in \cite{haug2012a} shows, each function is more likely to surface in specific syntactic, semantic and information-structural configurations. \posscitet{baryhaug2011} analysis is based on Ancient Greek, but we can easily exemplify the concept, in a nutshell, using Early Slavic participles, leaving the semantic formalization aside for simplicity. Since both Old Church Slavonic and Ancient Greek participles have been shown to express aspect, rather than relative tense (\citealt{eckhoff2015b}; \citealt{kamphuis2020a}), I will talk about the `aspect' of participles and refer to what is traditionally called ‘past participles’ and ‘present participles’ as `perfective participles and `imperfective participles' instead, which is reflected in the glossing criteria.

\begin{itemize}
\item \textsc{Frames} set the stage for and provide temporal anchoring for the matrix event.
    \begin{itemize}
    \item They are strictly anaphoric/presuppositional: they depend on the context for their own temporal reference, thus referring to previously mentioned or easily inferable situations. In (\ref{frameex}), for example, there is no explicit mention of Mstislav leaving for Pereyaslav, so the motion event is contextually inferred and used as anchor for the matrix event.
    \item Because they are presuppositional, they tend to occur sentence-initially.
    \item They are not modally dependent on the matrix verb, so that if the matrix verb is an imperative, for example, the participle is not interpreted as part of the command, as (\ref{frameex3}) shows.
    \item When perfective, they always induce \textit{narrative progression}, whereby the matrix eventuality is interpreted as taking place (just) after the participle event. When imperfective, the adjunct event includes the matrix event, as in (\ref{frameex2}).
    \item Their dominant verbal aspect within a corpus is likely to vary with the genre: ‘both while \textit{x}-ing and after \textit{x}-ing are possible ways of linking to a previously mentioned or accessible event’ (\pgcitealt{haug2012a}{311-312}).
    \item They are more likely to constitute `predictable' (\citealt{haug2012a}) predications in the discourse in which they appear, which may be expected to be reflected as overall lower lexical variation (\pgcitealt{haug2012a}{312}).
    \end{itemize}
    
    \begin{example}
    \gll i \textbf{prišed\foreignlanguage{russian}{ъ}} mstislav\foreignlanguage{russian}{ъ} v perejaslavl\foreignlanguage{russian}{ь}. poima ženu svoju ide v lučesk\foreignlanguage{russian}{ъ}.
    {and} {arrive.{\sc ptcp.pfv.m.nom.sg}} {Mstislav.{\sc m.nom}} {in} {Pereyaslav.{\sc acc}} {take.{\sc ptcp.ipfv.m.nom.sg}} {wife.{\sc acc.sg}} {his.{\sc f.acc.sg}} {go.{\sc aor.3.sg}} {to} {Luchesk.{\sc acc}}
    \glt ‘And when Mstislav arrived in Pereyaslav, he went to Luchesk, taking his wife with him' (\textit{Suzdal Chronicle}, Codex Laurentianus f. 115r) %274685
    \glend 
    \label{frameex}
    \end{example}

    \begin{example}
    \gll ty že \textbf{postę} sę pomaži glavǫ svojǫ i lice tvoe umyi
    {\sc 2.sg.nom} {\textsc{ptc}} {fast.{\sc ptcp.ipfv.m.nom.sg}} {\sc refl} {anoint.{\sc imp.2.sg}} {head.{\sc acc.sg}} {your.{\sc acc.sg}} {and} {face.{\sc acc.sg}} {your.{\sc acc.sg}} {wash.{\sc imp.2.sg}}
    \glt ‘But when you fast, put oil on your head and wash your face' (Matthew 6:17) %38430
    \glend 
    \label{frameex3}
    \end{example}

    \begin{example}
    \gll \textbf{ischodęšte} že obrětǫ čka kẏriněiska
    {go out.{\sc ptcp.ipfv.m.nom.pl}} {\sc ptc} {receive.{\sc aor.3.pl}} {person.{\sc gen.sg}} {of Cyrene.{\sc gen.sg}}
    \glt ‘As they were going out, they met a man from Cyrene' (Matthew 27:32) %51188
    \glend 
    \label{frameex2}
    \end{example}

\item \textsc{Independent Rhemes} are discourse-coordinated, therefore in\-for\-ma\-tion-structurally very similar to independent clauses.

    \begin{itemize}
        \item Like \textsc{frames}, they relate to the preceding context for their temporal reference, but unlike \textsc{frames}, they generally introduce new events.
        \item They also provide the temporal reference for the next event in line and can be stacked up similarly to a series of independent clauses, as in (\ref{indrh1}).
        \item Unlike \textsc{frames}, they are modally dependent on the matrix verb. They are, for example, interpreted as part of the command if the matrix is an imperative, as in (\ref{indrh2}).
        \item Like \textsc{frames}, they also always produce narrative progression when perfective. 
        \item They are predominantly perfective and precede the matrix verb.
        \item They are expected to encode new information: this could surface as overall more lexical variation among participles with this function. 
    \end{itemize}

    \begin{example}
        \gll i abie \textbf{tek\foreignlanguage{russian}{ъ}} edin\foreignlanguage{russian}{ъ} ot\foreignlanguage{russian}{ъ} nich\foreignlanguage{russian}{ъ}. i \textbf{priem\foreignlanguage{russian}{ъ}} gǫbǫ. \textbf{ispl\foreignlanguage{russian}{ь}n\foreignlanguage{russian}{ь}} oc\foreignlanguage{russian}{ь}ta. i \textbf{v\foreignlanguage{russian}{ь}znez\foreignlanguage{russian}{ъ}} na tr\foreignlanguage{russian}{ь}st\foreignlanguage{russian}{ь}. napaěše i
        and immediately run.\textsc{ptcp.pfv.m.nom.sg} one.\textsc{m.nom.sg} from \textsc{3.pl.gen} and take.\textsc{ptcp.pfv.m.nom.sg} sponge.\textsc{acc.sg} fill.\textsc{ptcp.pfv.m.nom.sg} vinegar.\textsc{gen.sg} and put.\textsc{ptcp.pfv.m.nom.sg} on reed.\textsc{acc.sg} give.to.drink.\textsc{impf.3.sg} \textsc{3.m.sg.acc}
        \glt ‘Immediately one of them ran and took a sponge, filled it with sour wine and put it on a reed, and offered it to him to drink’ (Matthew 27:48)
        \glend
        \label{indrh1}
\end{example}

\begin{example}
\gll i ně li ti tamo povoica. a \textbf{kr\foreignlanguage{russian}{ь}voši} prisoli
{and} {not be.{\sc 3.sg.prs.act}} {if} {\sc 2.sg.dat} {there} {veil.{\sc sg.gen}} {then} {buy.{\sc ptcp.pfv.f.nom.sg}} {send.{\sc imp.2.sg}}
\glt ‘And if you don’t have a veil there, then buy one and send it.’\hfill(Birch bark letter 731; translation by \citealt{schaeken2018}) %255134
\glend
\label{indrh2}
\end{example}

\item \textsc{Elaborations} add granularity to the semantics of the main event and typically express manner or means/instrument.

    \begin{itemize}
        \item They are more likely to be imperfectives since their function is to provide more information on the matrix event by describing `“concomitant circumstances” that are cotemporal with the main event’ (\pgcitealt{haug2012a}{312}).
        \item They are more likely to encode new information.
        \item They are more likely to follow the matrix clause, because they are temporally dependent on the main verb and because they are non-topical and non-focused.
        \item Like \textsc{independent rhemes} they are modally dependent on the matrix verb, as (\ref{elab1}) shows.
    \end{itemize}
    
    \begin{example}
    \gll ašte ty esi ch\foreignlanguage{russian}{ъ}. r\foreignlanguage{russian}{ъ}ci nam\foreignlanguage{russian}{ъ} ne obinǫję sę
    if \textsc{2.sg.nom} be.\textsc{pres.2.sg} Christ.\textsc{nom}  say.\textsc{imp.2.sg} \textsc{1.pl.dat} \textsc{neg} conceal.\textsc{ptcp.ipfv.m.nom.sg} \textsc{refl}
    \glt ‘If You are the Christ, tell us plainly' (John 10:24)
    \glend
    \label{elab1}
    \end{example}
\end{itemize}

\posscitet{baryhaug2011} framework is mostly built on the analysis of conjunct participles,\footnote{The authors only include one example containing an absolute participle, which is treated as a \textsc{Frame} (see \pgcitealt{baryhaug2011}{8}).} which begs the question of whether absolute constructions also fulfil all three functions with some frequency. The intuition is that absolute constructions should typically show properties of \textsc{frames}, given the main characterization of dative absolutes in the literature as ‘backgrounding’ or ‘stage-setting’ devices (\pgcitealt{worth1994a}{30}; \pgcitealt{corin1995a}{259}; \pgcitealt{collins2011a}{113}).\\
\indent As already mentioned, earlier scholarship on Early Slavic adjunct participle clauses has overwhelmingly focussed on their origin and relationship with Ancient Greek conjunct participles and genitive absolutes. More recently, \citet{sakharova2007,sakharova2010a} analysed the pragmatic motivations for the differences in the distribution of participle constructions and finite independent clauses for a number of verbs in Old East Slavic \textit{Novgorod First Chronicle}. Sakharova suggests that the usage of participle constructions (dative absolutes and conjunct participles alike) mostly depends on their `secondary status' (\foreignlanguage{russian}{второстепенности}) in the discourse (\pgcitealt{sakharova2010a}{88}), which in most cases has to do with discourse properties typical of `backgrounded' units. Background and foreground, according to Sakharova, can be defined on the basis of `what information is to be considered more or less relevant to the rhetorical goal':\footnote{`[...] \foreignlanguage{russian}{\textit{какую информацию считать менее соответствующей риторической цели}'} (\pgcitealt{sakharova2010a}{89}).} \textit{foreground} refers to `unexpected, non-sterotypical' situations, whereas \textit{background} refers to predictable information about `stereotyped sequences of situations'.\footnote{`\foreignlanguage{russian}{`\textit{Ожидаемой и, следовательно, менее выделенной прагматически, оказывается информация о стереотипных последовательностях ситуаций}.}' (\pgcitealt{sakharova2010a}{89}).}

This characterization of background and foreground unit bears, in fact, great similarities with the properties of \textsc{frames} and \textsc{independent rhemes}, respectively, as described by \citet{baryhaug2011} and outlined above. \\

\subsection{Frame setters and fronted adjuncts}\label{framesetterssec}
Much of the discussion in Chapters 1-4 revolves around the possible framing function of participle constructions and \textit{jegda}-clauses, and the configurations in which their interpretation as \textsc{frames} is, in fact, available. As we will see already at the beginning of Chapter 1, the sentence-initial position is where conjunct participles may be ambiguous between a \textsc{frame} and an \textsc{independent rheme} function, and it is also where the vast majority of dative absolutes in the Gospels are found, which, I will argue, is in itself indicative of their typical function as \textsc{frames}. As argued by \citet{baryhaug2011}, \textsc{frames} are generally found sentence-initially, since they are presuppositional and because they set the stage and provide temporal anchoring for the matrix event. In this, they are similar to \posscitet{charolles05} \textit{framing adverbials}, \posscitet{fabricius-hansen2012b} \textit{fronted adjuncts}, and \posscitet{chafe1976a} and \posscitet{krifka2007a} \textit{frame setters}, all of which are most naturally fronted because of their discourse-cohesive function of providing `the frame in which the following expression should be interpreted' (\pgcitealt{krifka2008}{269}), and because `they are topical (i.e. "anchored" in the preceding context)' (\pgcitealt{charolles05}{13}) and locate `the matrix event relative to an eventuality that [...] has already been explicitly established in or is inferable from the preceding context' (\pgcitealt{fabricius-hansen2012b}{54}). It is important to keep in mind some of the properties in common to these notions and \posscitet{baryhaug2011} \textsc{frames} since they will be particularly useful to account for a group of absolute constructions that, outside of the dataset used in Chapter 1, occur in post-matrix position but whose function is still compatible with \posscitet{baryhaug2011} \textsc{frames}, despite these being overwhelmingly more frequent in pre-matrix position, as already mentioned.\\
\indent Frame setters, according to \pgcitet{chafe1976a}{50}, are used `to limit the applicability of the main predication to a certain restricted domain'. \citet{krifka2007a} built on \posscitet{chafe1976a} definition by adding that, where a frame setter is not explicit, the domain or precise evaluation of the main predicate is unspecified (hence presumably inferred). Explicit frame setters specify this domain by choosing one out of a set of possible frames, thereby stating that the main predicate holds within that frame---which is why, according to Krifka, frame setters always contain a separate focus, which is meant as `the presence of alternatives that are relevant for the interpretation of linguistic expressions' (\pgcitealt{krifka2007a}{247}). Krifka's treatment of frame setters (like that of co-eventive adjuncts in \citealt{fabricius-hansen2012b}, as well as that of participle adjuncts in \citealt{baryhaug2011}) is based on the assumption shared by several theories of information structure and non-linear prosody (e.g. \citealt{truckenbrodt95,truckenbrodt07,buring2006,fery2007,fery2008}) that sentences present a double bipartition into \textit{background}-\textit{focus}\footnote{Note that this is not the same as \textit{background} as used so far, which is more in line with the usage of the term in formal theories of discourse representation such as SDRT.} and \textit{topic}-\textit{comment} (also called \textit{theme}-\textit{rheme}). While \textit{focus} always refers to the presence of alternatives, \textit{topic} can refer to \textit{aboutness} topic, in broad terms corresponding to `the entity that a speaker identifies, about which then information, the comment, is given' (\pgcitealt{krifka2007a}{265}), but also to frame setters, sometimes also referred to as \textit{frame topics} (e.g. \pgcitealt{fabricius-hansen2012b}{41}) or \textit{frame setting topics} (e.g. \citealt{fery2007}; \pgcitealt{Song2017}{25}). Both aboutness and frame-setting topics refer to entities that are already established or inferrable from the preceding discourse---`individuals in the case of aboutness topics, time and places in the case of frame topics' (e.g. \pgcitealt{fabricius-hansen2012b}{41}). While sentences always have explicit foci, topics are not necessarily expressed (as in [\textit{the HOUSE is on fire}]$_{Comment}$; \pgcitealt{krifka2007a}{267}).\footnote{The example given by \citet{krifka2007a} leverages prosody (as the all-capitals on \textit{house} indicates), to make certain arguments stronger---something which we cannot obviously rely on for a dead language.}\\
\indent The relation between \textit{topic}-\textit{comment} and \textit{background}-\textit{focus} is much more complex than the mere equivalences \textit{topic} $=$ \textit{background} and \textit{comment} $=$ \textit{focus}. Crucially, a topic can contain a focus. Though not pursuing the issue in depth, \citet{fabricius-hansen2012b} suggest that closed adjuncts (i.e. absolutes), because of their internal subject, may have a topic of their own serving as a contrastive partial topic of a common topic (i.e. the referent introduced by the subject of the matrix). A contrastive topic, in \posscitet{krifka2007a} treatment, is `an aboutness topic that contains a focus, which is doing what focus always does, namely indicating an alternative' (\pgcitealt{krifka2007a}{267}), which in the case of contrastive topic is an alternative aboutness topic, as in (\ref{contrtopic}).

\begin{example}
(From \pgcitealt{krifka2007a}{268})
    \begin{itemize}
        \item[A:] Does your sister speak Portuguese?
        \item[B:] [My [BROther]$_{Focus}$]$_{Topic}$ [[DOES]$_{Focus}$]$_{Comment}$
    \end{itemize}
    \label{contrtopic}
\end{example}.

\citet{fabricius-hansen2012b} argue that clause-final restrictive adjuncts (which, they note, often receive either mildly or strongly contrastive readings) semantically correspond to frame-setting fronted adjuncts, so that the former must be syntactically analysed on a par with fronted adjuncts, namely adjoined high up in the sentence---`above rather than below the subject of the matrix clause' (p.54). In (\ref{clausefinalcontr}a), for instance, the closed adjunct \textit{with the leaves on the trees} may receive the restrictive interpretation which is instead explicitly triggered in its paraphrasis in (\ref{clausefinalcontr}b). Similarly, \pgcitet{krifka2007a}{270} compared contrastive topics and frame setters, saying that what they have in common `is that they express that, for the communicative needs at the current point of discourse, the current contribution only gives a limited or incomplete answer'. This is one of the information structural differences between (\ref{clausefinalcontr}e), which has no adjunct, and the others in (\ref{clausefinalcontr}). The domain of (\ref{clausefinalcontr}e) has to be inferred from the context, since there is no frame-setter providing one out of a set of alternatives within which the eventuality holds. 

\begin{example}
(a. and b. from \pgcitealt{fabricius-hansen2012b}{48})
\begin{itemize}
\item[a.] The cave entrance is easy to miss \textit{with the leaves on the trees}
\item[b.] The cave entrance is easy to miss \textit{when the leaves are on the trees}
\item[c.] \textit{With the leaves on the trees} the cave entrance is easy to miss
\item[d.] \textit{When the leaves are on the trees} the cave entrance is easy to miss
\item[e.] The cave entrance is easy to miss
\label{clausefinalcontr}
\end{itemize}
\end{example}

Crucially, in all of (\ref{clausefinalcontr}a)-(\ref{clausefinalcontr}d) the adjunct is very much \textsc{frame}-like from the temporal perspective. The (mildly) contrastive reading possible in at least (\ref{clausefinalcontr}a) and (\ref{clausefinalcontr}b) does not change our observations about the type of semantic contribution these make compared to that in (\ref{clausefinalcontr}c) and (\ref{clausefinalcontr}d). \\

\subsection{Segmented Discourse Representation Theory}\label{sdrtintro}
\posscitet{baryhaug2011} framework intentionally leaves aside semantic effects that derive from default discourse reasoning (cf \pgcitealt{baryhaug2011}{36-37}) and only formally models the temporal dimension. Their formalization is thus not fully specified as far as the rhetorical links introduced by participle clauses are concerned. However, \textsc{frames}, \textsc{independent rhemes} and \textsc{elaborations} \textit{are} primarily associated with specific discourse functions, as we have seen. \textsc{Frames}, as \pgcitet{baryhaug2011}{13} say, are similar to \pgposscitet{krifka2008}{269} \textit{stage topics} or \textit{frame setters}, which provide “the frame in which the following expression should be interpreted”. \textsc{Independent rhemes} are like independent clauses from the discourse perspective. \textsc{Elaborations} provide more information (elaborate) on the matrix event, similarly to how the \textit{Elaboration} relation is described in formal theories of discourse representation like Segmented Discourse Representation Theory (SDRT; \citealt{asherlasca2003}), on which \posscitet{baryhaug2011} discourse analysis is loosely based. \\
\indent SDRT is a theory of discourse interpretation which, like classic theories in dynamic semantics (\citealt{groenendijkstokhof, kamp1993a, kamp2011a}), works under the assumption that ‘the interpretation of an utterance is made relative to the interpretation of the preceding utterance’ (\pgcitealt{asherlasca2003}{39}) so that every incoming sentence updates the discourse as it unfolds. SDRT models the semantics-pragmatics interface using rhetorical relations (such as the aforementioned \textit{Elaboration} relation), which are logical links computed on the basis of commonsense reasoning (in the sense of \citealt{hobbs1993a}) with non-linguistic information (e.g. domain knowledge) (\pgcitealt{asherlasca2003}{39}), which allows accounting for semantically underspecified links between clauses. This captures what was mentioned above, namely that, much like a sequence of two main clauses, the precise semantic relation between participle clauses and their matrix clause (e.g. non-temporal effects such as causality, condition, instrument) must be pragmatically inferred.  \\
\indent An important aspect of SDRT is that its so-called Glue Logic (i.e. the set of inference rules applied to pragmatically enrich the `underspecified logical forms' derived by the syntax and the compositional semantics) is non-monotonic. Namely, it captures \textit{defeasible} inferences, meaning that the discourse representation structure can be modified if other cues emerge later in the discourse licensing different rhetorical relations than previously inferred. One of \posscitet{baryhaug2011} reasons for avoiding a fully-fledged SDRT formalization of discourse is that, in SDRT, aspect (which SDRT broadly captures as an \textit{eventive} versus \textit{stative} distinction) is only used as a non-monotonic cue for certain rhetorical relations, whereas, as \citet{baryhaug2011} showed, aspect can be represented to a large extent in the compositional semantics of Ancient Greek participles. In SDRT, for example, a sequence of two eventive discourse units may (non-monotonically) trigger a \textit{Narration} relation (\ref{eq:2}), but not a \textit{Background} relation, which instead requires a stative and an eventive unit (\ref{eq:1}).\footnote{?(\textit{$\alpha$,$\beta$}) simply indicates an underspecified link between the discourse units $\alpha$ and $\beta$.}

\begin{example}
\textit{Event}($e_{\beta}$) $\land$ \textit{Event}($e_{\alpha}$) $\land$ ?(\textit{$\alpha$,$\beta$}) $>$ \textsc{Narration}($\alpha$,$\beta$)
 \label{eq:2}
\end{example}

\begin{example}
  \textit{Event}($e_{\alpha}$) $\land$ \textit{State}($e_{\beta}$) $\land$ ?(\textit{$\alpha$,$\beta$}) $>$ \textsc{Background}($\alpha$,$\beta$)\\
\textit{State}($e_{\beta}$) $\land$ \textit{Event}($e_{\alpha}$) $\land$ ?(\textit{$\alpha$,$\beta$}) $>$ \textsc{Background}($\alpha$,$\beta$)
 \label{eq:1}
\end{example}

Unlike SDRT, \citet{baryhaug2011} also do not make specific assumptions regarding the form a discourse structure should take, whereas in SDRT discourse structures are represented as graphs, which aptly captures the core concepts of defeasible inference and discourse update. An SDRT discourse representation structure is thus incrementally and hierarchically built and updated with each incoming discourse unit. To take a classic SDRT example, consider (\ref{exsdrt}) from \pgcitet{asherlasca2003}{8}.

\begin{example}
    \begin{itemize}
        \item[a.] John had a great evening last night. 
        \item[b.] He had a great meal.
        \item[c.] He ate salmon.
        \item[d.] He devoured cheese.
        \item[e.] He then won a dancing competition.
    \end{itemize}
    \label{exsdrt}
\end{example}

\begin{figure}[!h]
\centering
\includegraphics[width=1\linewidth]{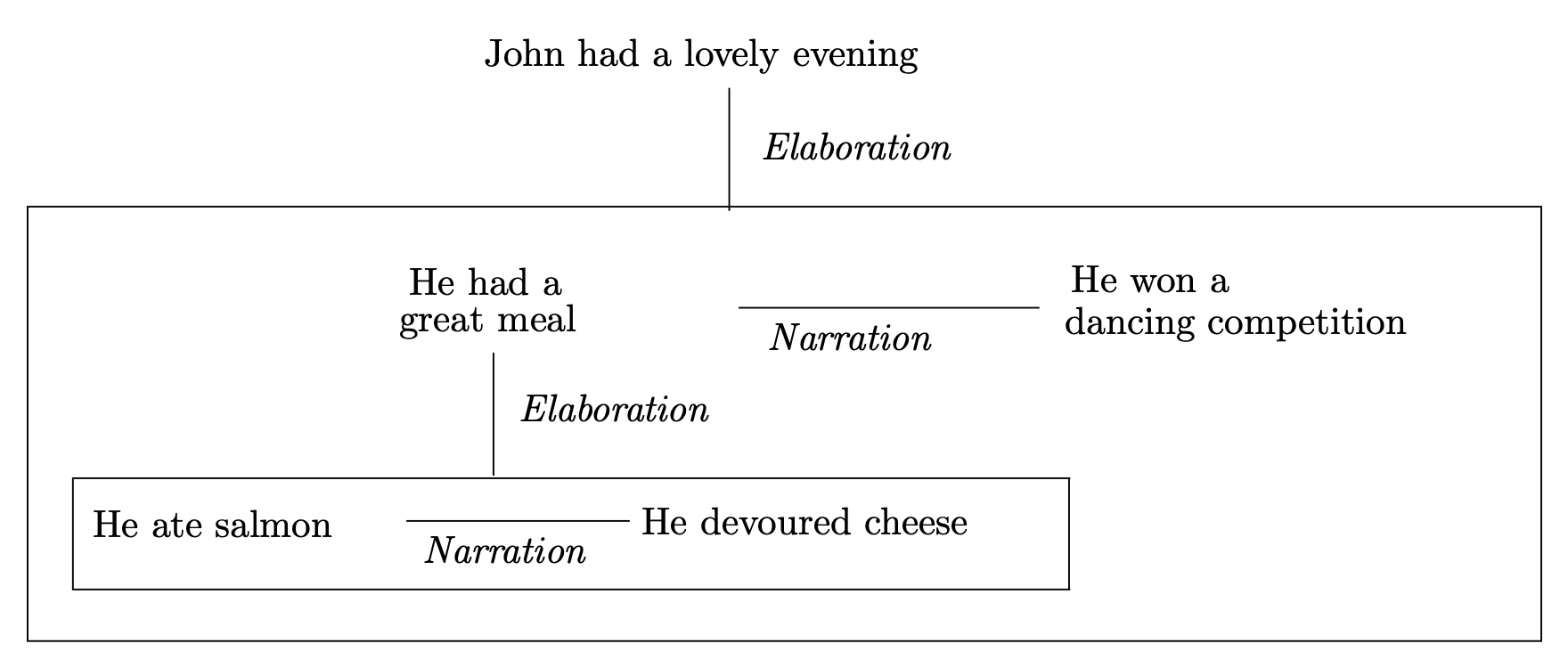}
\caption[Discourse representation structure of a complex discourse from \pgcitet{asherlasca2003}{9}]{Discourse structure for (\ref{exsdrt}), from \pgcitet{asherlasca2003}{9}.}
\label{sdrsex}
\end{figure}

A possible structure for (\ref{exsdrt}) could be built by inferring that (\ref{exsdrt}b) adds granularity to (\ref{exsdrt}a), thus attaching to it via the SDRT relation \textit{Elaboration}. Both (\ref{exsdrt}c) and (\ref{exsdrt}d), then, jointly elaborate on (\ref{exsdrt}b). (\ref{exsdrt}c) and (\ref{exsdrt}d) are therefore linked by the rhetorical relation \textit{Narration} and together form a complex discourse unit attaching to (\ref{exsdrt}b) via \textit{Elaboration}. Finally, (\ref{exsdrt}e) continues the elaboration of (\ref{exsdrt}a) in the same way as (\ref{exsdrt}b) does. (\ref{exsdrt}d) is therefore linked to (\ref{exsdrt}b) via \textit{Narration} and forms a complex discourse unit with it, so that they now \textit{jointly} elaborate on (\ref{exsdrt}a). Figure (\ref{sdrsex}) shows the discourse structure in a simple graph form.

As we will see in the thesis, both the non-monotonic logic and the graph representation of discourse structure of SDRT are particularly useful to formally capture the discourse functions of participle constructions and \textit{jegda}-clauses in more complex discourses.  When several adjuncts co-occur in the same discourse, each adjunct can be interpreted as triggering a particular rhetorical relation at the clause-level, but may then be combined with subsequent units in the discourse structure to form complex discourse units that may attach to the subsequent discourse via a yet different relation. 

\section{What's with \textit{when}?}\label{whatswithwhen}
There are several factors that heavily interact with tense-aspect in the interpretation of finite temporal clauses and their temporal relation to the matrix. These have been investigated specifically with regard to \textit{when}-clauses since at least \citet{steedman82} and include, most importantly, the distinction between \textit{events} and \textit{states} (DRT \& SDRT), between \textit{durative} and \textit{non-durative} situations (\citealt{heinamaki74}), between \textit{point} and \textit{interval} (\citealt{ritchie}), between \textit{achievements}, \textit{accomplishments}, \textit{activities} and \textit{states} (\citealt{vendler}), and, similarly, \posscitet{comrieaspect} higher-level categorization of these distinctions into three main Aktionsarten, namely \textsc{telic}, \textsc{atelic} and \textsc{stative}. The precise differences between some of these categories are often marred by terminological discussions continuing since the sudden plethora of works, often very influential, published in the 1970s on tense and aspect (among others, \citealt{verkuyl72,dowty72,dowty77,dowty79,heinamaki74,bennett75,comrieaspect,steedman77,ritchie,isard74,smith78}) and in the 1980s on event structure, temporal relations, and the anaphoric properties of tense (e.g. \citealt{steedman82,partee84,comrietense,nerbonne86,passonneau87,webber83,webber87,webber88,dowty86,hinrichs86,moenssteedman87}, all drawing from \citealt{reichenbach47}).\\
\indent In common to many theories is that initial (i.e. pre-matrix) temporal adverbials, such as \textit{when}-clauses, provide the reference time within which the situation described by the main clause holds temporally, a temporal frame that establishes how the main eventuality is to be interpreted. The main issue has long been explaining how different types of temporal referents introduced by an adverbial (e.g. a punctual event versus a period of time) interact with the temporal referent introduced by the matrix clause, which depends on the temporal adverbial for its interpretation. \\
\indent It has long been observed (e.g. \citealt{kamprohrer,partee84,hinrichs86,sandstrom,bonomi97,s2012a}) that when either the temporal adverbial clause or the matrix clause (or both) describe a stative/imperfective eventuality there is a temporal symmetry between the two clauses, whereby if the role is switched the temporal interpretation is preserved, as in (\ref{symmetry}) from \pgcitet{s2012a}{1422}.

\begin{example}
    \begin{itemize}
        \item[a.]
        \gll My s Iroj gotovili dokumenty, kogda pozvonil Borja.
        we with Ira prepared.\textsc{ipfv} documents when called.\textsc{pfv} Boris
        \glt ‘Irina and I were preparing the documents when Boris called.’
        \glend
        \item[b.]
        \gll Kogda my s Iroj gotovili dokumenty, pozvonil Borja.
        when we with Ira prepared.\textsc{ipfv} documents called.\textsc{pfv} Boris
        \glt ‘When Irina and I were preparing the documents Boris called.’
        \glend
    \end{itemize}
    \label{symmetry}
\end{example}

As pointed out by \pgcitet{s2012a}{1422}, (\ref{symmetry}a) and (\ref{symmetry}b) may have information-structural differences or may involve different discourse relations, but the temporal relation is unchanged. When both eventualities are eventive/perfective, the symmetry does not hold, as (\ref{asymmetry}) shows.

\begin{example}
    \begin{itemize}
        \item[a.] When she died she left a massive doll collection.\\
        (\pgcitet{s2012a}{1423})
        \item[b.] She died when she left a massive doll collection.
    \end{itemize}
    \label{asymmetry}
\end{example}

\cite{sandstrom} argues that the temporal referent introduced by a \textit{when}-clause in English must be a bounded referent: either the \textit{phase} of a state (`when I was eight, I moved to Oslo'), a result state (`when they had written the letter, they sighed with relief'), or an individual event (`when he saw him, he smiled at him'). Even so, however, several other factors are at play in determining whether a \textit{when}-clause can be employed or not. Compare the pairs in 

\begin{example}  \label{wheneng1}
\begin{itemize}
    \item[a.] She reached him, and said something in his ear.
    \item[b.] When she reached him, she said something in his ear.
\end{itemize}
\end{example}

\begin{example}  \label{wheneng2}
\begin{itemize}
    \item[a.] They chose violence and attacked innocent people.
    \item[b.] ?When they chose violence, they attacked innocent people.
\end{itemize}
\end{example}

\begin{example}  \label{wheneng3}
\begin{itemize}
\item[a.] He burped loudly and put the glass back on the counter.
     \item[b.] ?When he burped loudly he put the glass back on the counter.
\end{itemize}
\end{example}

\begin{example}  \label{wheneng4}
\begin{itemize}
   \item[a.] ?Sue killed her husband and put poison in his whiskey. \label{ex4a}
     \item[b.] When Sue killed her husband, she put poison in his whiskey. \label{ex4b}
\end{itemize}
\end{example}

\begin{example}   \label{wheneng5}
\begin{itemize}
   \item[a.] ?When they built the fifth bridge, a man drowned in the river. \label{ex5a}
    \item[b.] When they were building the fifth bridge, a man drowned in the river.
     \label{ex5b}
\end{itemize}
\end{example}

These examples (from \citealt{sandstrom}, except (\ref{wheneng5}) from \citealt{ritchie} and (\ref{wheneng2}) from me) show that sequences of two event-type predicates can be felicitously rendered either as a sequence of two main clauses, as \textit{when}-structures, or both, even though the sentences in each pair (\ref{wheneng1})-(\ref{wheneng5}) are in fact structurally identical. Parallel structures like (\ref{wheneng2}), if deemed equally acceptable, can hardly be considered to convey the same temporal relation between the events predicated of---(\ref{wheneng2}a) suggests a temporal sequence, where they first chose violence and, as a result of it, they attacked. (\ref{wheneng2}b) instead suggests that \textit{every time} they chose violence they attacked innocent people (with the presupposition that they sometimes do not choose violence). In other instances, as in (\ref{wheneng4}), the main clause is a `constitutive part' (\citealt{moens}) or `subevent' (\citealt{sandstrom}) of the event referent introduced by the \textit{when}-clause, which ultimately shows how mere temporal inclusion of a main-clause event in a given temporal referent cannot be felicitously conveyed by an eventive \textit{when}-clause (as in (\ref{wheneng5}a), a reading that demands the progressive instead (\ref{wheneng5}b).\\
\indent Even without wandering far from English in genealogical terms, it is clear that these (sometimes very subtle) constraints are not identical cross-linguistically. The scope of each \textit{when}-equivalent varies widely across different languages, which may favour other constructions in situations where \textit{when}-clauses are perhaps preferred or more commonly found in English. This is not necessarily an observation about the lack of particular translational equivalents in either English or any of the other languages, but rather about (also potentially very subtle) differences in the way each language categorizes meanings, and, therefore, in the division of labour among competing constructions. Any \textit{when}-clause in an English version of the Bible may correspond to different constructions in other translations:

\begin{example}
Acts 27:38 \label{parallelsrandom}
\begin{itemize}

\item[a.]And \textbf{when they had eaten enough}, they lightened the ship\\(English Standard Version)

\item[b.]
\gll \textit{En} \textit{\textbf{nadat}} \textit{\textbf{hulle}} \textit{\textbf{met}} \textit{\textbf{voedsel}} \textit{\textbf{versadig}} \textit{\textbf{was}}, \textit{het} \textit{hulle} \textit{die} \textit{skip} \textit{ligter} \textit{gemaak}
and after they with food filled were had they the ship lighter made
\glt (Afrikaans)\footnote{1933 [1953] South African Bible Society translation.}
\glend
     
\item[c.]
\gll Quando si furono saziati di cibo, alleggerirono la nave
when \textsc{refl} had filled of food, they-lightened the ship
\glt (Italian)\footnote{1998 [1948-53] \textit{Bibbia di Gerusalemme}.}
\glend

\item[d.]
\gll \textit{I} \textit{\textbf{nasitivši}} \textit{\textbf{se}} \textit{\textbf{jela}}, \textit{olakšaše} \textit{ladju}
and having-filled \textsc{refl} with-food they-lightened ship
\glt (Serbian)\footnote{1868 Daničić-Karadžić translation.}
\glend

\item[e.]
\gll \textit{Luego}, \textit{\textbf{satisfechos}} \textit{\textbf{de}} \textit{\textbf{la}} \textit{\textbf{comida}}, \textit{aligeraban} \textit{la} \textit{nave}
later satisfied of the meal they-lightened the ship
\glt (Spanish)\footnote{2015 \textit{Reina Valera Actualizada} translation.}
\glend

\item[f.]
\gll \textit{\textbf{După}} \textit{\textbf{ce}} \textit{\textbf{s'au}} \textit{\textbf{săturat}}, \textit{au} \textit{uşurat} \textit{corabia}
After that self-have filled, they-have lightened ship
\glt (Romanian)\footnote{1921 \textit{Biblia Cornilescu} translation.}
\glend

\end{itemize}
\end{example}

(\ref{parallelsrandom}) is only one of the possible scenarios, where the English version has a \textit{when}-clause and the other languages show a range of different possible constructions, instead of the \textit{when}-equivalents that do exist in each of those languages. Several other scenarios are obviously possible and are, in fact, widely attested, including those where all the languages in (\ref{parallelsrandom}) use \textit{when}-equivalents or where all but English use a \textit{when}-equivalent. Even among Indo-European languages, the scope of \textit{when}-equivalents can however be quite different from the one of English \textit{when}-clauses. In Ancient Greek, where non-finite temporal adverbial clauses are much more extensively employed than in English, we find that there is no direct way of employing \textit{when}-equivalents (\textit{hóte}/\textit{hótan}) to express simple temporal inclusion in the past, for which imperfective participle clauses are generally employed instead, as in (\ref{greekptcpengwhen}b) where the English \textit{when}-clause corresponds to a genitive absolute in Greek (and similarly a dative absolute in Old Church Slavonic (\ref{greekptcpengwhen}c)).

\begin{example}
Matthew 28:11
\begin{itemize}
\item[a.]Now \textbf{when they were going}, behold, some of the watch came into the city, and shewed unto the chief priests all the things that were done. \\
(King's James Version)
\item[b.]
\gll \textit{\textbf{Poreuomenōn}} \textit{de} \textit{\textbf{autōn}} \textit{idou} \textit{tines} \textit{tēs} \textit{koustōdías} \textit{elthontes} \textit{eis} \textit{tēn} \textit{polin} \textit{apēngeilan} \textit{tois} \textit{archiereusin} \textit{hapanta} \textit{ta} \textit{genomena}
{go.{\sc ptcp.ipfv.gen.pl}} {\sc ptc} {\sc 3.pl.gen} {behold} {some} {the.{\sc gen.sg}} {watch.{\sc gen.sg}} {come.{\sc ptcp.pfv.nom.pl}} {in} {the.acc} {city.acc} {announce.{\sc aor.3.pl}} {the.{\sc dat.pl}} {chief priest.{\sc dat.pl}} {all.{\sc n.acc.pl}} {the.{\sc n.acc.pl}} {happen.{\sc ptcp.pfv.n.acc.pl}}
\glt
\glend
\item[c.]
\gll \textbf{idǫštama} že \textbf{ima}. se edini ot\foreignlanguage{russian}{ъ} kustodiję prišed\foreignlanguage{russian}{ъ}še v\foreignlanguage{russian}{ъ} grad\foreignlanguage{russian}{ъ}. v\foreignlanguage{russian}{ь}zvěstišę archiereom\foreignlanguage{russian}{ъ} v\foreignlanguage{russian}{ь}sě byv\foreignlanguage{russian}{ъ}šaa.
{go.{\sc ptcp.ipfv.dat.du}} {\sc ptc} {\sc 3.du.dat} {behold} {some.{\sc nom.pl}} {from} {watch.{\sc gen.sg}} {come.{\sc ptcp.pfv.nom.pl}} {in} {city.{\sc acc.sg}} {announce.{\sc aor.3.pl}} {chief priest.{\sc dat.pl}} {all.{\sc n.acc.pl}} {happen.{\sc ptcp.pfv.n.acc.pl}}
\glt
\glend
\end{itemize}
\label{greekptcpengwhen}
\end{example}

Based on our knowledge of Ancient Greek aspectual semantics, we might expect to find \textit{when}-equivalents in the imperfect to enable mere temporal inclusion of an eventive main clause, similarly to how the progressive in English enables simple temporal inclusion in the \textit{when}-clause, as in (\ref{wheneng5}). Corpus data, however, indicates that such a configuration (i.e. an imperfect \textit{when}-equivalent followed by an eventive--aorist--main clause) is exceedingly rare in the Ancient Greek corpus\footnote{Only 1 occurrence of pre-matrix \textit{hóte}-clause with an imperfect verb followed by an aorist main verb was found in PROIEL, and none of \textit{hótan}. The verb in the imperfect is a form of \textit{eimí} `to be, to exist' (\textit{kaì hóte ēn dekaétēs ho paīs, prēgma es hautòn toiónde genómenon exéphēné min} `Now when the boy was ten years old, the truth about him was revealed in some such way as this', Herodotus 1:114). There are 3 occurrences of post-matrix imperfect \textit{hóte}-clauses, 2 of which are forms of \textit{eimí} and 1 of the verb \textit{záō} `to live'. In such cases, we could argue that the order of the eventualities involved is aorist-imperfect, not imperfect-aorist, which is somewhat less rare, although still relatively uncommon (13 occurrences).}. Both imperfective participles and imperfect \textit{jegda}-clauses with an aorist main clause are instead found in Early Slavic, as the dative absolute in (\ref{greekptcpengwhen}c) and the \textit{jegda}-clause in (\ref{greekptcpengwhen2}c) show. Once again, Greek does not use a \textit{hōte}- or \textit{hōtan}-clause, but this time a nominalized accusative with infinitive (\ref{greekptcpengwhen2}c).

\begin{example}
Luke 11:27 %40681
\begin{itemize}
\item[a.]And it came to pass, as he was saying these things a woman from the crowd called out
\item[b.]
\gll {Egeneto} {de} {\textbf{en}} {\textbf{t\^oi}} {\textbf{legein}} {auton} {tauta} {eparasa} {tis} {ph\^on\^en} {gun\^e} {ek} {tou} {okhlou} {eipen} {aut\^oi}
{happen.{\sc aor.3.sg}} {\sc ptc} {in} {the.{\sc m.dat.sg}} {say.{\sc inf.prs}} {\sc 3.sg.m.acc} {this.{\sc n.acc.pl}} {raise.{\sc ptcp.pfv.f.nom.sg}} {certain.{\sc m.nom.sg}} {voice.{\sc f.acc.sg}} {woman.{\sc f.nom.sg}} {from} {the.{\sc gen.sg}} {crowd.{\sc gen.sg}} {say.{\sc aor.3.sg}} {\sc 3.sg.m.dat}
\glt
\glend
\item[c.]
\gll byst\foreignlanguage{russian}{ъ} že \textbf{egda} \textbf{glaaše} se. v\foreignlanguage{russian}{ъ}zdvig\foreignlanguage{russian}{ъ}ši glas\foreignlanguage{russian}{ъ} edina žena ot\foreignlanguage{russian}{ъ} naroda reče emu
{happen.{\sc aor.3.sg}} {\sc ptc} {when} {say.{\sc impf.3.sg}} {this.{\sc n.acc.sg}} {raise.{\sc ptcp.pfv.f.nom.sg}} {voice.{\sc acc}} {some.{\sc f.nom.sg}} {woman.{\sc f.nom.sg}} {from} {crowd.{\sc gen.sg}} {say.{\sc aor.3.sg}} {\sc 3.sg.m.dat}
\glt
\glend
\end{itemize}
\label{greekptcpengwhen2}
\end{example}

It is clear that the competition between participle adjuncts and \textit{jegda}-clauses cannot simply be captured in terms of discrete, categorical variables, but it should be modelled as a continuum allowing a degree of overlap and should aim to reveal broader patterns in a probabilistic, rather than a fully deterministic way. This is what token-based typology can help us achieve.

\section{A token-based typological approach to the study of language-internal variation}\label{tokenbased}
In our exploration of the functions of participle clauses and \textit{jegda}-clauses we do not have access, \textit{a priori}, to a (set of) comparative concept(s) (\citealt{haspelmath2010}) that can be leveraged for language comparison. We have simply observed that there is overlap in the usage of certain language-particular constructions, whose individual functional remit must still be defined---which is, in fact, the very goal of this thesis. As we have seen, there are good reasons to expect that the semantics of Early Slavic participial adjuncts, like their Ancient Greek counterparts, can be defined compositionally to a good extent, particularly with respect to the temporal-aspectual dimension. However, simply by virtue of being juxtaposed forms, there are, especially non-temporal, aspects of their semantics which are left underspecified. This is, as we saw, why \posscitet{baryhaug2011} framework is partly integrated with SDRT as a discourse formalization tool, particularly for dealing with more complex discourse structures. \\
\indent The closest we get to a comparative concept encompassing the constructions under study is what \citet{cristofarowals} refers to as \textsc{when}-clauses\footnote{Small caps is mine. \citet{cristofarowals} uses inverted commas (`when') to refer to the concept \textsc{when} to distinguish it from the English token \textit{when}.}, defined in functional terms as a construction `encoding a temporal relation between two events, such that there is a temporal overlap between the two' and in which `the exact extent of the overlap is unspecified and subject to variation'. Cristofaro's definition encompasses both \textit{deranked} verb forms (\citealt{Stassen-1985,Croft-1990,Cristofaro-1998,cristofarosubordination}), that is, verb forms lacking marking of one or more tense, aspect, or mood distinctions compared to independent clauses in the same language (e.g. participles in Early Slavic and Ancient Greek) and \textit{balanced} ones, which include not only finite clauses introduced by a subordinator (e.g. \textit{jegda}-clauses or English \textit{when}-clauses), but also juxtaposed independent clauses whose temporal relation needs to be pragmatically inferred. \\
\indent We have seen in Section \ref{whatswithwhen} that English \textit{when}-clauses, as well as \textit{jegda}-clauses, can express more than just overlap. Cristofaro's \textsc{when}-clauses may either reflect a misnomer, for lack of better generic alternatives, or their definition may refer to a more general sense of \textit{overlap} as \textit{co-temporality} in the sense of, for example, \citet{behrens2012a}. Whichever the case, the scope of English \textit{when}-clauses \textit{does} observably span both that of \textit{jegda}-clauses and that of conjunct participles and absolute constructions, as can be evinced by the counterparts to each of these constructions in different English versions of the Bible, that is, all three OCS constructions can correspond to a \textit{when}-clause in English, as the examples in this Introduction already suggested. As a preliminary umbrella term for the functional continuum in which conjunct participles, absolute constructions and \textit{jegda}-clauses are set, we may therefore stick to `\textsc{when}-clauses'.  \\
\indent With these assumptions in mind, we can use a token-based typological approach (\citealt{levshina19,levshina21}) to explore the semantic ground covered by English \textit{when} and \textit{induce} cross-linguistically common semantic dimensions from parallel corpora available for Old Church Slavonic. \textit{Token-based typology} `makes generalizations and classifies languages using the tokens of specific linguistic units or structures observed in language use, as approximated by corpora' (\pgcitealt{levshina19}{534}), allowing us to capture the gradience and overlap between linguistic phenomena like \textsc{when}-clauses and the language-internal variation which is inherent to the very concept of competition. This is in contrast with traditional \textit{type-based} approaches which compare languages based on pre-defined categorical (e.g. \textit{balanced} versus \textit{deranked}; SV versus VS) or ordinal variables (e.g. number of basic colour categories), as the discrete categories in the World Atlas of Language Structures (WALS; \citealt{wals}) indicate.\\
\indent Like previous experiments in token-based typology that aimed to induce cross-lin\-guis\-ti\-cal\-ly salient dimensions from a parallel corpus (e.g. \citealt{walchli-cysouw2012, hartcysouwhaspl, levshina15}) probabilistic semantic maps are employed in this thesis as a data-driven method to study conceptual variation within the semantic space of the hypothetical concept \textsc{when}. As we will see, the analysis based on this method provides us with clearer functional boundaries between the Early Slavic constructions under analysis, supporting and complementing the findings of the corpus-based and temporal-semantic analysis of the first part of the thesis, while also setting Early Slavic \textsc{when}-clauses in typological perspective.

\section{The Early Slavic corpus}\label{corpus}
Most of the Early Slavic data used in this thesis is extracted from the TOROT\footnote{Tromsø Old Church Slavonic and Old Russian Treebanks (\citealt{eckhoff2015a}). The version used in this thesis corresponds to release $20200116$, which can be obtained from \url{http://torottreebank.github.io}.} treebank. Table \ref{torotbreakdown} shows a breakdown of the texts from which dative absolutes, conjunct participles and \textit{jegda}-clauses were extracted, with an indication of the (high-level) variety of Early Slavic which each text can be ascribed to and the total number of manually annotated tokens.\\
\indent All TOROT data is lemmatized, morphologically analysed and syntactically annotated, and all layers of annotation have either been manually post-corrected or fully manually annotated.\footnote{The Old Church Slavonic \textit{Codex Zographensis} in TOROT was excluded from the dataset because it mostly overlaps in content with the Codex Marianus, which has more detailed annotation. The modern Russian portion of the TOROT treebank is also excluded.} \\
\indent As an offspring of the Old Church Slavonic portion of the PROIEL\footnote{Pragmatic Resources in Old Indo-European Languages (\citealt{proiel}), \url{https://proiel.github.io}.} treebank, TOROT uses the PROIEL annotation scheme, which is modelled on dependency grammar (DG). Like DG, it uses overt elements in the sentences as nodes in a syntactic representation, rather than adding phrasal nodes as in phrase structure grammar. The PROIEL scheme, however, deviates from classic DG schemes (e.g. the one adopted by the Prague Dependency Treebank) in some important ways, most notably by allowing empty nodes (e.g. in cases such as asyndetic coordination and verb ellipsis) and by explicitly signalling secondary dependency (whenever there is a nonfinite construction whose subject is coreferent with an element of the matrix verb, as is the case with conjunct participles) (\citealt{haug-etal-2009-computational}). \\
\indent The \textit{Codex Marianus}, an Old Church Slavonic tetraevangelion from the PROIEL treebanks, also contains information-structural annotation, including givenness status (e.g. \textsc{new}, \textsc{old}, \textsc{accessible}) and anaphoric links from anaphors to their antecedents, as well as additional layers of semantic annotation, such as animacy of nominal referents and Aktionsart at the level of verb lemma. Furthermore, it is aligned to the Ancient Greek version of the New Testament in PROIEL\footnote{The Greek New Testament version in PROIEL is \citet{tischen}, which is not the source text of the Codex Marianus, as is clear when comparing some of the mismatches between the Ancient Greek and the Old Church Slavonic version. In the course of the relevant chapters, I will point out whether potential mismatches between the two languages according to the PROIEL version correspond to a match when compared to other Ancient Greek versions.}. Because of its alignment with the Ancient Greek version and its more detailed annotation, the analysis will be carried out separately on the \textit{Codex Marianus} and on the rest of the TOROT treebank. I will refer to the former as a \textit{deeply annotated} treebank, to distinguish it from the rest of the texts from the TOROT treebank, which contain only up to dependency annotation, namely the minimum requirement for a corpus to be classified as a treebank, and which I will refer to as \textit{standard} treebanks.\\
\indent Note that East Slavic is somewhat overrepresented in the standard treebanks, especially as far as non-translated or `original' Early Slavic texts are concerned, but generally also in the time span covered. The East Slavic subcorpus covers the whole period from Old East Slavic (11th-late 14th century) to Early Middle Russian (15th century) and Late Middle Russian (16-late 17th century), with a total of over 200,000 tokens, whereas the South Slavic one (i.e. South Slavic texts excluding the Old Church Slavonic subcorpus and all classified under the general label of \textit{Church Slavonic}) are represented by only around 13,000 tokens. \\
\indent Although not meant to actually compensate for this imbalance, a smaller group of South Slavic texts is used in a brief case study at the end of Chapter 2. The case study summarizes the findings from \citet{pedrazzinijhs}, who looked at the extent to which corpora with considerably fewer and shallower levels of annotation can be exploited to investigate a discourse-driven syntactic phenomenon, particularly as a means of corroborating results emerging from treebanks with deeper, curated annotation. The small South Slavic corpus was experimentally fully automatically annotated, including lemmatization, morphological analysis, and dependency parsing.\footnote{The simple script and dictionary used to lemmatize is available at \url{https://doi.org/10.6084/m9.figshare.24166254}. Part-of-speech and morphological tags were added using \posscitet{scherrer2018a} pre-modern Slavic CLSTM tagger \citet{scherrer2019a}, while for dependency parsing I used \posscitet{oldslavnet} dependency parser, OldSlavNet (see also \citealt{pedrazzinichr}.} A breakdown of these texts can be found in Table \ref{stratbreakdown}. The case study from this dataset, which was not manually post-corrected and hence contains many errors, will be presented separately from both deeply-annotated and standard treebanks and will be referred to as \textit{strategically annotated} treebanks.\footnote{All strategically annotated texts were normalized by bringing down superscript letters, removing diacritics, substituting non-Unicode characters with their Unicode counterparts, among other preprocessing steps, all of which can be found in the preprocessing Python scripts in the project repository (\url{https://doi.org/10.6084/m9.figshare.24166254}). \textit{Yagičev Zlatoust}, \textit{Manasses Chronicles}, the \textit{Sbornici} of Vladislav Grammarian, and the Zografski Sbornik are from the open-source digital editions of the \textit{Cyrillomethodiana} project (\url{https://histdict.uni-sofia.bg/}). The Bdinski Sbornik is from the Obdurodon project (\url{http://bdinski.obdurodon.org/}. \textit{Hilandar Typikon} and \textit{Karyes Typikon} are from the \textit{Monumenta Serbica} project (\url{http://monumentaserbica.branatomic.com/}). }

\subsection{Identification of the constructions in the corpora}\label{identifconstr}
\begin{figure}[!h]
\centering
\begin{subfigure}{0.23\textwidth}
\includegraphics[width=1\linewidth]{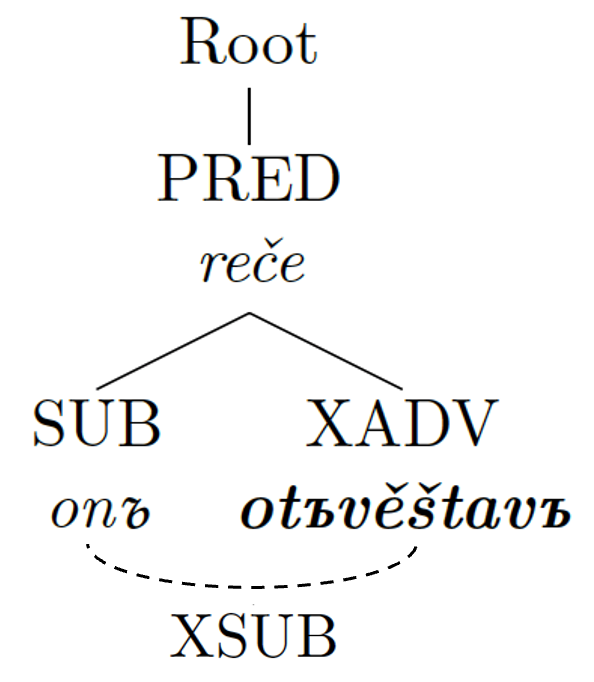}
\caption[]{}
\label{overtslashxadv}
\end{subfigure}\hspace{0.2\textwidth}
\begin{subfigure}{0.20\textwidth}
\includegraphics[width=1\linewidth]{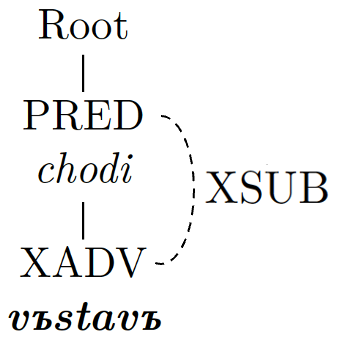} 
\caption[]{}
\label{nullslashxadv}
\end{subfigure}
\caption[Slash notation (secondary dependencies) in PROIEL/TOROT]{The slash notation in PROIEL/TOROT. (a) is an example of an overt-subject conjunct participle; (b) is an example of a null-subject conjunct participle.}
\label{conjunctstree}
\end{figure}

Absolute constructions and conjunct participles are not labelled as such in the annotated treebanks. To identify and extract potential occurrences, a combination of morphosyntactic tags and dependency relations were used. As already mentioned, deeply annotated and standard treebanks from TOROT use the PROIEL dependency tagset, which includes separate labels for clauses with an external argument (e.g. \textsc{xadv} and \textsc{xcomp}), whereas the strategically annotated treebanks included in the dataset were syntactically annotated with OldSlavNet, which uses the Universal Dependency tagset. \\
\indent From the TOROT treebanks, conjunct participles were identified by looking for participle forms in any tense-aspect, excluding resultative ones (i.e. \textit{l}-participles), with an \textsc{xadv} relation. The referent of the subject of conjunct participle was identified via the `slash' notation available for all clauses with an external argument to indicate coreference relations within one dependency tree. Most conjunct participles share the subject with their matrix clause, so the \textsc{xadv} node will also be connected to the matrix subject node, if the subject is overt, as in Figure \ref{overtslashxadv}, or to the matrix verb node itself, if the subject is null, via the relation \textsc{xsub}, to indicate argument sharing, as Figure \ref{nullslashxadv} shows.

\begin{figure}[!h]
\centering
\begin{subfigure}{0.50\textwidth}
\caption[]{}
\includegraphics[width=1\linewidth]{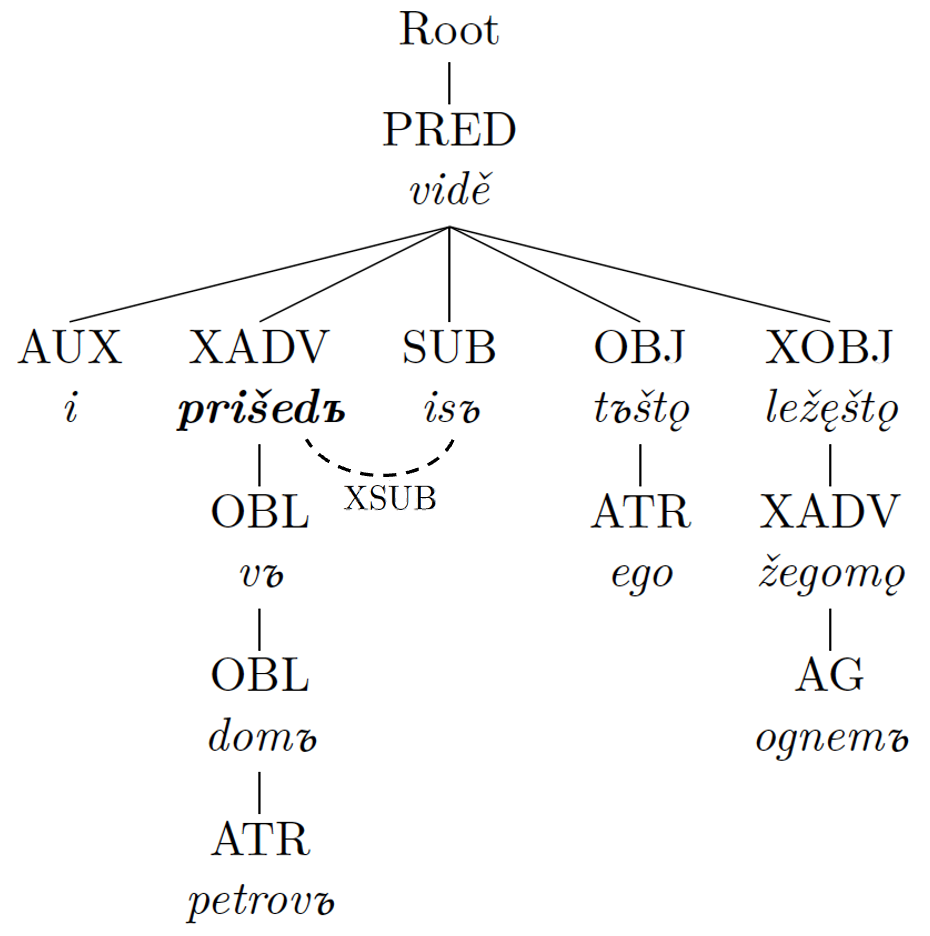}
\label{conservativetree}
\end{subfigure}
\begin{subfigure}{0.50\textwidth}
\caption[]{}
\includegraphics[width=1\linewidth]{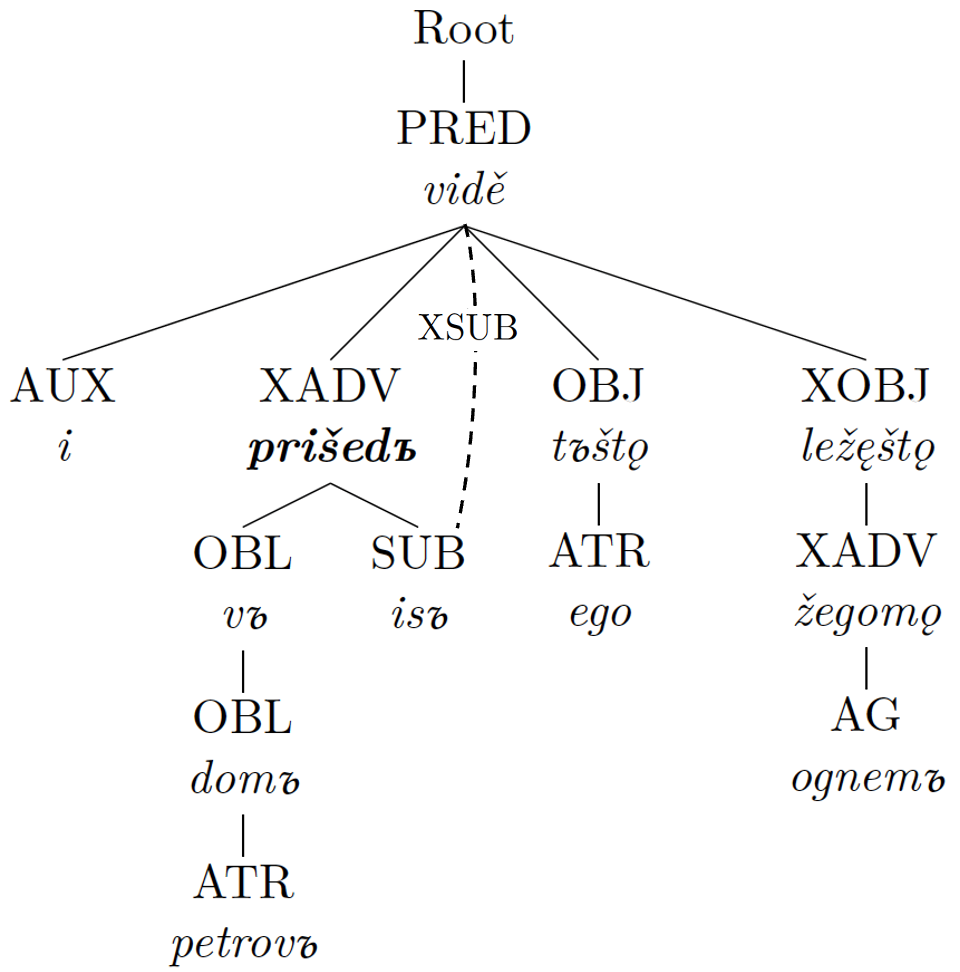} 
\label{linearizedtree}
\end{subfigure}
\caption[PROIEL analysis versus linearization analysis of an XSUB]{(a) PROIEL analysis of (\ref{linearex}) (b) linearization analysis of (\ref{linearex}), in which the subject of the conjunct participle, shared with the matrix clause, is analyzed as a dependent of the participle clause.}
\label{linearizedtrees}
\end{figure}

The annotation convention in PROIEL is to always analyze the shared arguments in control structures as dependents of the matrix clause. However, following the analysis of control in Greek and Latin participial adjuncts in \citet{haug2011a,haug2017a}, we can consider \textit{some} conjunct participle configurations to be able to head a subject, specifically the leftmost \textsc{frame} or \textsc{independent rheme} conjunct participle linearly preceding the matrix clause. Such conjunct participles can be analyzed as having a control relation of \textit{equality} (rather than \textit{subsumption}; cf. \pgcitealt{dalrymple2019}{553-556}) to the matrix clause, albeit with an additional \textit{linearization} constraint on the overt realization of shared arguments. Namely, if overt, the shared argument is always found in the leftmost position relative to the matrix clause (`in the leftmost legitimate host'; \pgcitealt{haug2017a}{152}). Throughout the thesis, therefore, only the leftmost pre-matrix conjunct participles are considered as potentially able to head their subject and, therefore, to be strictly referred to as \textit{overt}- or \textit{null}-subject. Those whose \textsc{xsub} slash points to an overt argument (i.e. a node other than the matrix verb itself) will be considered as overt-subject only if the overt argument is found adjacent to or between material belonging to participle clause, which turned out to be the case in the vast majority of cases.\footnote{In only 10 cases, the shared overt subject is found after the matrix verb instead.} In some instances, the dependence of the shared argument on the conjunct participle is evidenced by the fact that the shared argument is intermingled with, material that clearly belongs to the participle clause, as in (\ref{linearex}), where \textit{is\foreignlanguage{russian}{ъ}} `Jesus', the shared argument, occures between the conjunct participle \textit{prišed\foreignlanguage{russian}{ъ}} `having come' and another argument of the participle clause, i.e. \textit{v\foreignlanguage{russian}{ъ} dom\foreignlanguage{russian}{ъ} petrov\foreignlanguage{russian}{ъ}}. Figures \ref{linearizedtrees} show the difference between the conservative dependency analysis of (\ref{linearex}) in PROIEL and its linearization analysis.

\begin{example}
\gll i {\normalfont [}\textbf{prišed\foreignlanguage{russian}{ъ}} is\foreignlanguage{russian}{ъ} v\foreignlanguage{russian}{ъ} dom\foreignlanguage{russian}{ъ} petrov\foreignlanguage{russian}{ъ}{\normalfont ]} vidě t\foreignlanguage{russian}{ъ}štǫ ego ležęštǫ ognem\foreignlanguage{russian}{ъ} žegomǫ
{and} {come.{\sc ptcp.pfv.m.nom.sg}} {Jesus.{\sc nom}} {in} {house.{\sc acc}} {Peter's.{\sc acc}} {see.{\sc aor.3.sg}} {mother-in-law.{\sc acc}} {\sc 3.sg.m.gen} {lie.{\sc ptcp.ipfv.f.acc}} {fire.{\sc inst}} {burning.{\sc ptcp.ipfv.pas.f.acc}}
\glt ‘When Jesus came into Peter’s house, he saw Peter’s mother-in-law lying in bed with a fever' (Matthew 8:14) %38512
\glend
\label{linearex}
\end{example}

In some instances, both analyses may be possible, since the overtly-realized shared argument is adjacent to both the sentence-initial conjunct participle and the matrix verb, as in (\ref{ambiglinear}). 

\begin{example}
\gll i \textbf{prišed\foreignlanguage{russian}{ъ}še} oučenici ego v\foreignlanguage{russian}{ъ}zboudišę i
{and} {come.{\sc ptcp.pfv.nom.pl}} {disciple.{\sc nom.pl}} {\sc 3.sg.m.gen} {wake.{\sc aor.3.pl}} {\sc 3.sg.m.acc}
\glt ‘The disciples went and woke him' (Matthew 8:25) %50732
\glend
\label{ambiglinear}
\end{example}

\begin{figure}[!h]
\centering
\includegraphics[width=0.5\textwidth]{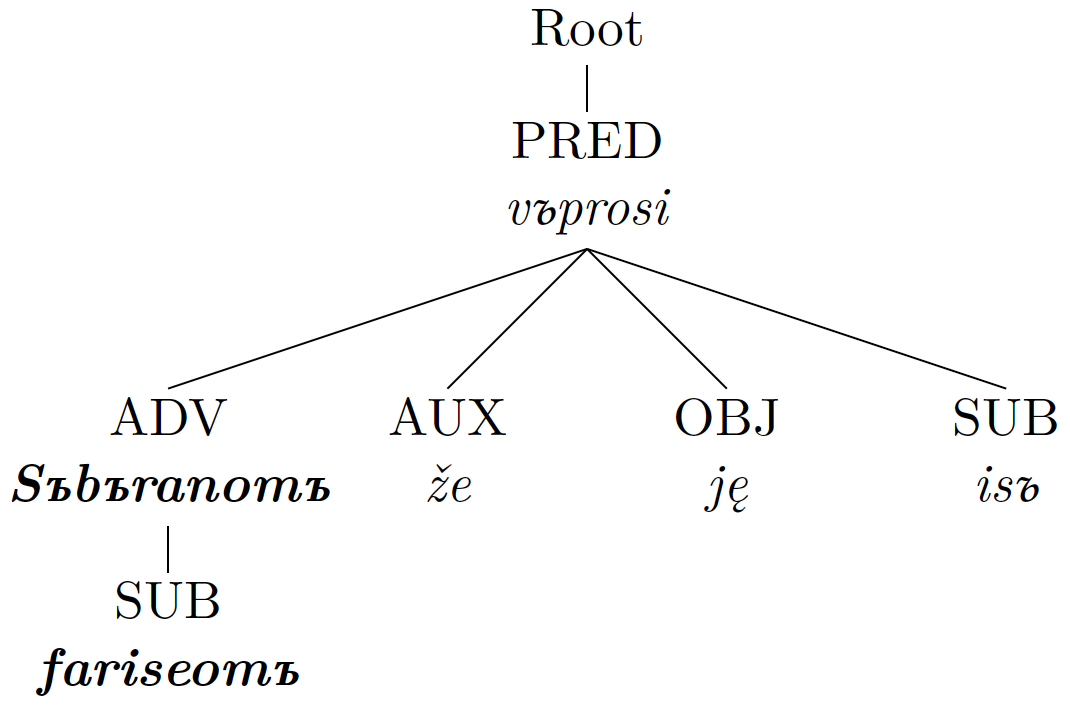}
\caption{Dependency tree with a dative absolute and its subject}
\label{simpledatree}
\end{figure}

Throughout the analysis, cases like (\ref{ambiglinear}) will also be considered as overt-subject conjunct participles, with the caveat in mind that an alternative analysis may be possible.\\
\indent Dative absolutes were identified by looking for participle forms in the dative and in any tense-aspect, again excluding resultative ones, with an \textsc{adv} relation. The presence of an overt subject was not specified but identified, if present, by looking for tokens in the dative with a \textsc{sub} relation and dependent on the participle itself, as in Figure \ref{simpledatree}.

Note that, without specifying overt subjects as a requirement, this query also returns forms that meet the criteria but that are not dative absolutes. Given its relatively small size, the dataset of potential absolutes from deeply annotated treebanks (Codex Marianus) was fully manually corrected to filter out any wrongly extracted occurrences. From standard treebanks, most potential post-matrix null-subject dative participles were cases of nominalized participles, as in (\ref{notda}), and were therefore all excluded without a full manual check.

\begin{example}
    \gll i naprasnijem\foreignlanguage{russian}{ъ} prěloženija čudo s\foreignlanguage{russian}{ъ}tvoril\foreignlanguage{russian}{ъ} \textbf{z\foreignlanguage{russian}{ъ}ręštiim\foreignlanguage{russian}{ъ}}
    {and} {suddenness.{\sc inst.sg}} {transformation.{\sc gen.sg}} {miracle.{\sc acc.sg}} {make.{\sc ptcp.result.m.nom.sg}} {watch.{\sc ptcp.ipfv.m.dat.pl}}
    \glt `And by the suddenness of the transformation he has performed a miracle for the witnesses' (\textit{Encomion on the 40 Martyrs of Sebasteia}, Codex Suprasliensis 6: 412) %85444
    \glend
    \label{notda}
    \end{example} 

\begin{figure}[!h]
\centering
\includegraphics[width=0.5\textwidth]{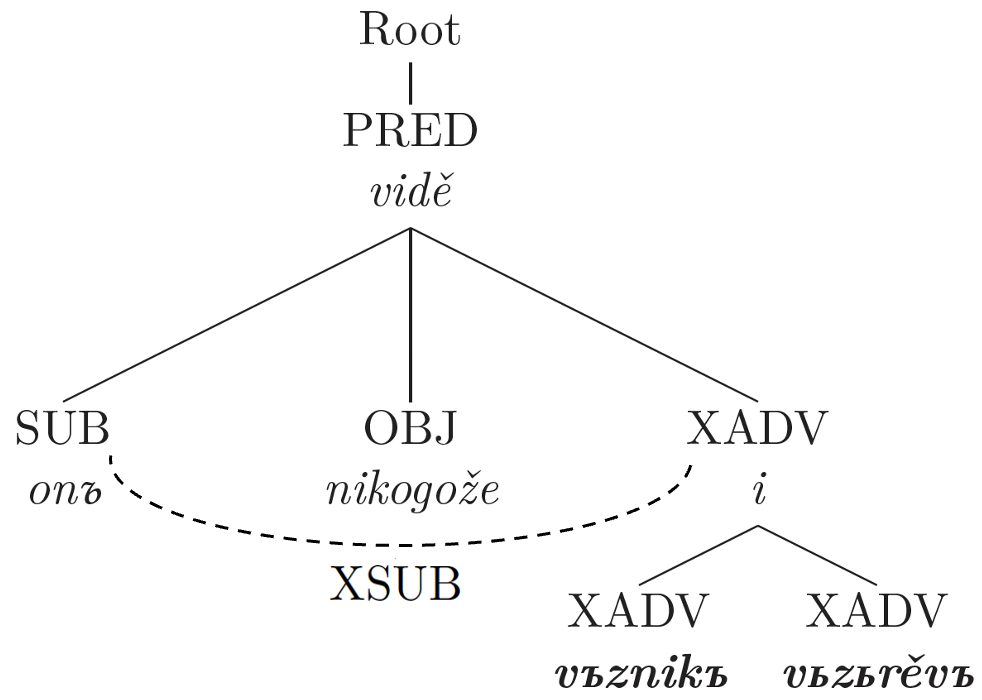}
\caption{Dependency tree containing two coordinated conjunct participles}
\label{multipleptcptree}
\end{figure}

Potential null-subject dative absolutes preceding the matrix clause were instead manually checked and wrongly extracted occurrences (around 30\% of potential pre-matrix null-subject dative absolutes) were filtered out.
Only if the lemma was \textit{byti} `be', all occurrences were included regardless of position since a manual check indicated that these are mostly temporal expressions of the kind \textit{pozdě byv\foreignlanguage{russian}{ъ}šu} `once it got late', a very common usage of absolutes throughout the corpora.\\
\indent If the syntactic subject of a participle clause was made of several coordinated constituents (e.g. `Jesus and the disciples'), then the one linearly occurring first (e.g. `Jesus' from `Jesus and the disciples') was extracted for the purpose of analysing the properties of subjects.\\ 
\indent The matrix verb of both conjunct participles and dative absolutes was considered to be the direct head of the participle itself if the head node was a verb, as in Figures \ref{conjunctstree}-\ref{simpledatree}. When the direct head of the participle is not a verb, it is normally due to one of two structures. If the direct head is a conjunction, generally multiple (typically two) coordinated participles are dominated by the same matrix verb, in which case PROIEL assigns the same relation tag as the participles themselves to the conjunction node (i.e. \textsc{xadv} or \textsc{adv}), as in Figure \ref{multipleptcptree}. In this case, the head of the conjunction node typically corresponds to the matrix verb itself, bearing the relation tag \textsc{pred}, in which case the \textsc{pred} node is identified as the `true head' of the participle clause.

\begin{figure}[!h]
\centering
\includegraphics[width=0.6\textwidth]{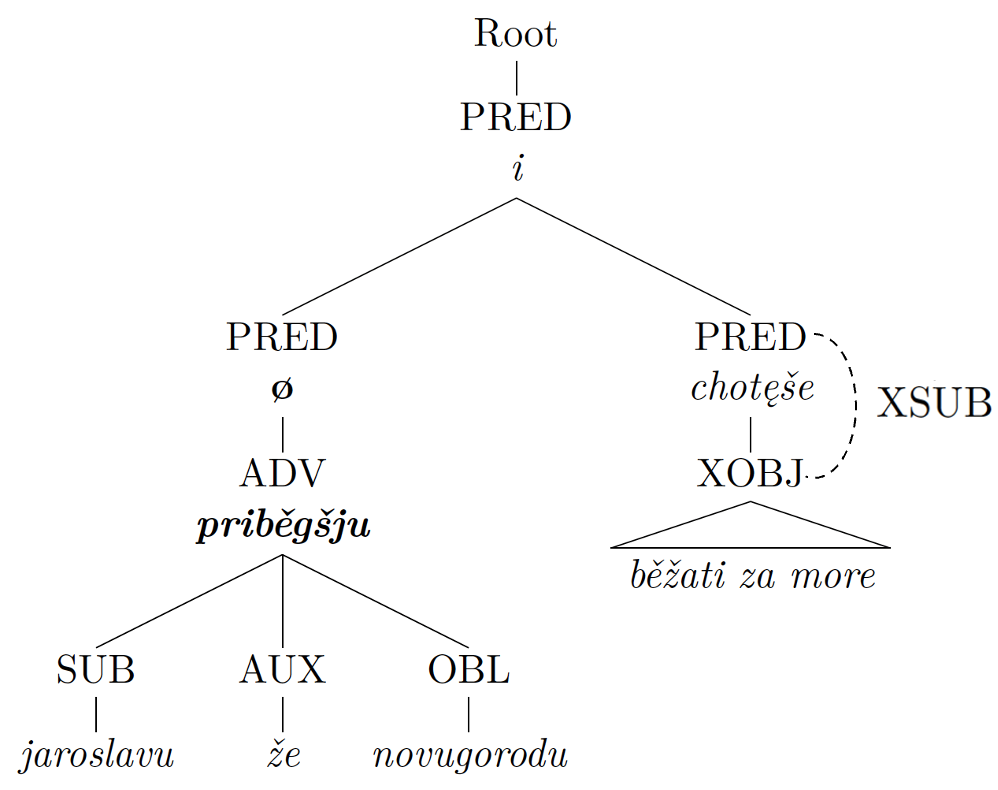}
\caption{Dependency tree containing a dative absolute coordinated to the main clause}
\label{emptynodetree}
\end{figure}

If the direct head is an empty node, it generally indicates syntactic independence of some kind, most often overt coordination to the finite clause, which falls into the set of `non-canonical' participle constructions mentioned above.  This is because, in order to avoid direct coordination of participle constructions with finite clauses, the annotation convention in PROIEL (for the sake of retrievability) is to treat such examples as elliptical constructions. During their annotation, an empty verbal node is therefore added above the participle clause to stand in for the matrix clause, as Figure \ref{emptynodetree} shows.

\begin{figure}[!h]
\centering
\includegraphics[width=0.45\textwidth]{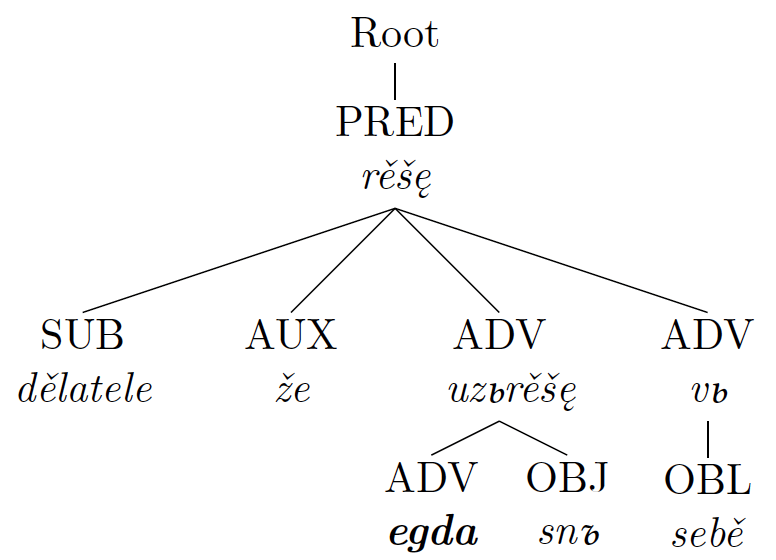}
\caption{Dependency tree containing a \textit{jegda}-clause.}
\label{egdanormaltree}
\end{figure}

\begin{figure}[!h]
\centering
\includegraphics[width=0.7\textwidth]{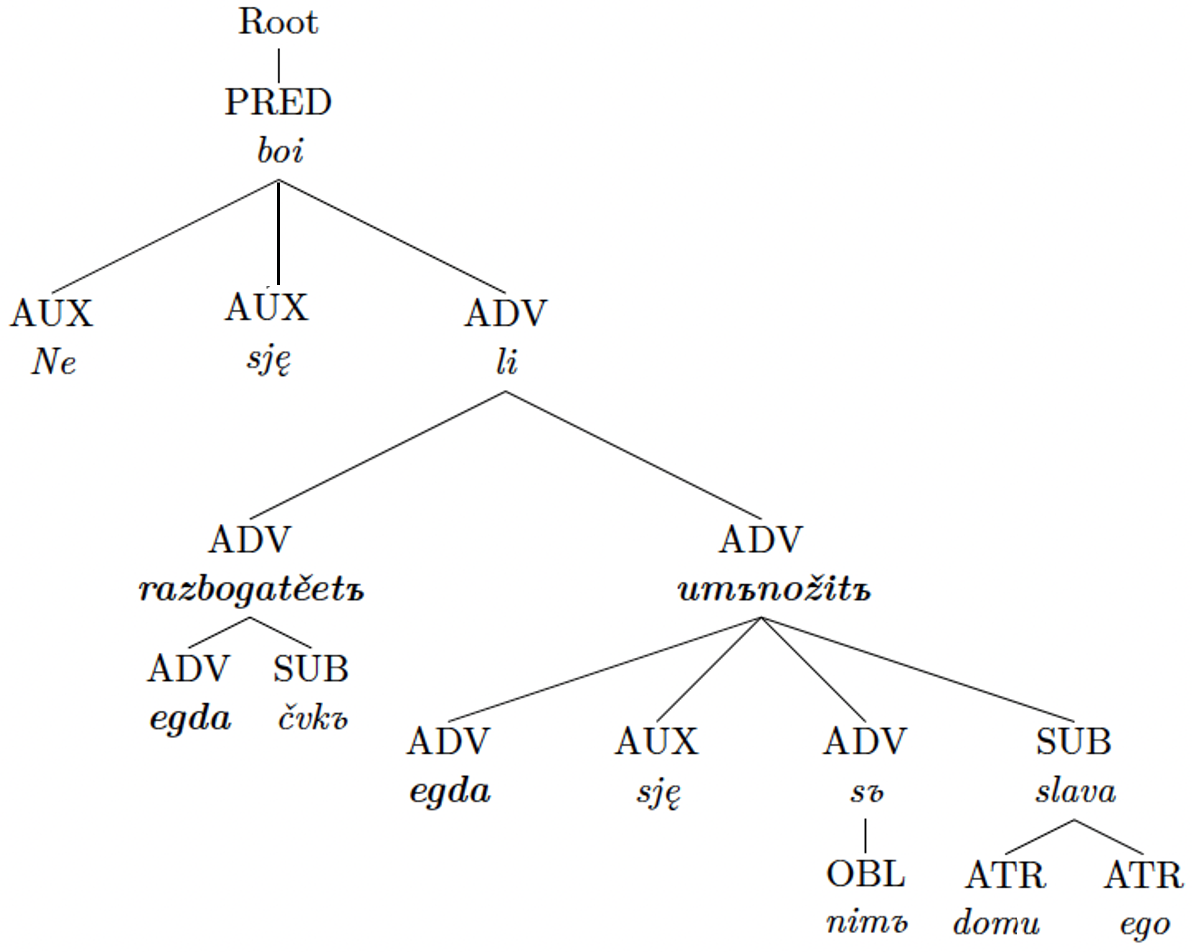}
\caption{Dependency tree containing two coordinated \textit{jegda}-clauses}
\label{multipleegdastree}
\end{figure}

In such cases, for the purpose of the analysis, the matrix verb is identified with the empty node itself, which makes `non-canonical' occurrences of participle constructions (i.e. conjunct participles and dative absolutes that are overtly coordinated to a following independent clause) more easily retrievable.\\
\indent From OldSlavNet-annotated treebanks, dative absolutes were identified by looking for participles in the dative with an \textsc{advcl} dependency relation, conjunct participles by looking for participles in the nominative, also with an \textsc{advcl} dependency relation. Both these simple queries also retrieve several forms that are neither conjunct participles nor dative absolutes. In particular, adverbials with an external subject (\textsc{xadv} in PROIEL) and secondary dependency relations (e.g. \textsc{xsub} in PROIEL) are not currently well-captured by state-of-the-art Early Slavic dependency parsers. This means that we are unable to extract the subjects of conjunct participles using fully automated methods, which naturally limits our analysis. To test whether fully automated methods can be used to formulate hypotheses or confirm previous results, the dataset extracted with these queries was not fully post-corrected. However, in addition to the quantitative analysis of strategically annotated treebanks, a case study on one particular strategically annotated text is carried out in Chapter 2, its occurrences checked and closely analysed, thus allowing us not only to evaluate the usefulness of the automatic annotation but also to compare the results of a distant-reading analysis to the observations made by close-reading the occurrences.\\
\indent \textit{Jegda}-clauses were first identified by looking for all occurrences of the lemma \textit{jegda} and then extracting the verb dominating the subjunction to represent the \textit{jegda}-clause itself. \textit{Jegda}-clauses used as explicit relative clauses (as in `the time \textit{when}...'), were excluded from the dataset by filtering out any \textit{jegda}-clause bearing an \textsc{atr} (attribute) or \textsc{apos} (apposition) relation. The same method used to identify the `true matrix' clause of a \textit{jegda}-clause was used. As a rule of thumb, the node immediately dominating the \textit{jegda}-clause verb was taken to be the matrix verb, as in Figure \ref{egdanormaltree}, unless it was a conjunction, in which case the node dominating the conjunction was extracted instead, as in Figure \ref{multipleegdastree}.

As a first approximation, overt subjects were identified by looking for a \textsc{sub} node headed by the \textit{jegda}-clause verb, unless it was a conjunction, indicating multiple coordinated subjects, in which case the first conjunct was extracted as the subject for the purpose of the analysis in Chapter 3. This method, however, is only possible for \textit{jegda}-clauses with a different subject than the one of the matrix clause. This is because, when co-referential, the subject is analysed as a daughter node of the matrix verb, without indicating secondary dependency relations such as the \textsc{xsub} one used with conjunct participles. As we will see in Chapter 3, in such cases, we are only able to establish such secondary dependency relation for deeply annotated treebanks, by leveraging the anaphoric link annotation in PROIEL, which also includes null constituents. Namely, if the overt subject of the matrix clause is anaphorically linked to the null subject of the \textit{jegda}-clause, then we can identify the referent of the subject of the \textit{jegda}-clause itself.

\begin{table}
\centering
\begin{tabular}{cccc}
    \toprule
\textbf{variety}  &  \textbf{title}  &  \textbf{tokens} \\ 
    \midrule
OCS  &  Codex Suprasliensis &  99,150  \\ 
 &  Codex Marianus  &  57,556 \\
   &  Psalterium Sinaiticum  &  30,488  \\  
  &  Euchologium Sinaiticum  &  717 \\ 
   &  Kiev Missal   &  352  \\
CS    &  \textit{Vita Constantini}  &   8,925   \\ 
  &  \textit{Vita Methodii}  &   3,014  \\ 
&  \textit{Life of Petka Tarnovska} (Novaković)  &  859  \\ 
  &  The Dobrejsho Gospel  &   366  \\ 
OES    &  \textit{Primary Chronicle}, Codex Laurentianus  &   55,088  \\ 
  &  \textit{Suzdal Chronicle}, Codex Laurentianus  &   23,343  \\ 
  &  Uspenskij Sbornik  &  24,734  \\ 
  &  \textit{Novgorod First Chronicle}, Synodal Manuscript  &   17,419  \\
  &  \textit{Russkaja pravda}  & 3,903  \\ 
  &  \textit{Primary Chronicle}, Codex Hypatianus  &   3,512   \\ 
  &  \textit{The Tale of Igor’s Campaign}  &  2,778 \\ 
  &  \textit{Zadonshchina}  &   2,343  \\ 
  &  1229 Smolensk-Riga-Gotland Treaty &    1,391 \\ 
  &  The \textit{Kiev Chronicle}, Codex Hypatianus  &   535  \\  
  &  Statute of Prince Vladimir  &  476  \\ 
   &  Colophon to Mstislav’s Gospel book  & 258  \\
&  Colophon to the Ostromir Codex  &  193   \\ 
  &  Mstislav’s letter  &  157   \\ 
  &  Varlaam’s donation charter to Xutyn  &  140 \\ 
EMRus  &  \textit{Life of Sergij of Radonezh}  &   19,966  \\ 
 &  \textit{The Tale of the Fall of Constantinople}  &  9,116  \\
 &  Afanasij Nikitin’s \textit{Journey Beyond Three Seas}  &  6,498  \\
 &  \textit{The Tale of Dracula}  &  2,446 \\
  &  Missive from Prince Ivan of Pskov  &   337  \\  
LMRus  &  \textit{Domostroj}  &   22,622   \\ 
  &  \textit{The Life of Avvakum}, Pustozerskij Sbornik &  22,168   \\ 
  & \textit{The Taking of Pskov}  &  2,290  \\ 
  &  Materials for the history of the schism  &  1,818  \\ 
  &  Vesti-Kuranty  &  1,114  \\ 
  &  Correspondence of Peter the Great  &   95  \\ 
ONov  &  Birch bark letters  &  1,864  \\ 
    \bottomrule
  \end{tabular}
  \caption[Main corpus breakdown]{Main corpus breakdown, with an indication of the language variety and the number of annotated tokens. OCS = Old Church Slavonic, OES = Old East Slavic, EMRus = Early Middle Russian, LMRus = Late Middle Russian, ONov = Old Novgorodian.}
    \label{torotbreakdown}
\end{table}

\begin{table}
\centering
\begin{tabular}{cccc}
    \toprule
\textbf{variety}  &  \textbf{title}   &  \textbf{tokens} \\ 
    \midrule
Middle Bulgarian & \textit{Yagičev Zlatoust} &  64,225 \\
 & \textit{Manasses Chronicles} & 41,794 \\
  & \textit{Sbornici} of Vladislav Grammarian & 32,040 \\
   & Bdinski Sbornik & 18,764 \\
 & Zografski Sbornik & 10,315 \\
Old Serbian & Saint Sava's \textit{Hilandar Typikon} & 11,982 \\
 & Saint Sava's \textit{Karyes Typikon} &  1,008 \\
    \bottomrule
  \end{tabular}
  \caption{Breakdown of strategically annotated texts}
    \label{stratbreakdown}
\end{table}

\section{Thesis overview}\label{thesisoverview}
The thesis is structured in two parts, differing in their scope and methodology. \\
\indent \textbf{Part I} contains four chapters and looks into the distribution of participial adjuncts and \textit{jegda-}clauses in Early Slavic, presenting a quantitative and descriptive corpus analysis and giving an account of their discourse functions and temporal semantics. \\
\indent More specifically, \textbf{Chapters 1} and \textbf{2} focus on absolute constructions and conjunct participles and use corpus data to understand the extent to which we can apply previous accounts of the discourse functions of Ancient Greek conjunct participles to Early Slavic participle constructions, including dative absolutes. These first two chapters follow the same overall structure, with Chapter 1 looking at deeply annotated treebanks and Chapter 2 at standard treebanks. Both
chapters show that, much like in Ancient Greek, adjunct participle clauses in Early Slavic largely function as `grammaticalized discourse relations' (\textsc{frames}, \textsc{independent rhemes}, and \textsc{elaborations}; \citealt{baryhaug2011}) and that their relation to the matrix clause can partly be inferred compositionally through a combination of aspect, position in the sentence, lexical variation, and information-structural properties. The specific treatment of absolute constructions as a subordination strategy in competition with conjunct participles led to the observation that they can be considered as specialized frame-setters and topic-shifters, unlike conjunct participles, whose interpretation is heavily dependent on other variables, such as tense-aspect and position in the sentence, whereby pre-matrix (especially perfective) participles are more likely to be \textsc{independent rhemes} and post-matrix (especially imperfective) participles are more likely to be \textsc{elaborations}.\\
\indent \textbf{Chapter 3} focuses on \textit{jegda}-clauses. The analysis indicates that there is overall more evident competition with absolutes (which are typically \textsc{frames}) than with conjunct participles. This confirms the intuition that \textit{jegda}, like English \textit{when} in SDRT, can be considered as a trigger for the rhetorical relation \textit{Background}, which is by and large compatible with \posscitet{baryhaug2011} treatment of \textsc{frames}. However, unlike participles, the aspect of \textit{jegda}-clauses cannot always be straightforwardly established on the basis of morphology alone, due to the high frequency of present-tense forms, so that their temporal relations cannot be fully established compositionally in the same way as with participles under \posscitet{baryhaug2011} framework. \textbf{Chapter 4}, therefore, provides a tentative account of the typical functions of \textit{jegda}-clauses and their temporal relation to the matrix clause by drawing from previous (broadly neo-Kleinian) accounts of the temporal semantics of temporal adverbials. It also formalizes Early Slavic examples containing both \textit{jegda}- and participle clauses on the basis of the SDRT account of the rhetorical relation \textit{Background} and \textit{Narration}. \\
\indent \textbf{Part II} contains two chapters and looks at adjunct participle clauses and finite temporal subordinates from a typological perspective. \\
\indent \textbf{Chapter 5} takes a temporary step back from Early Slavic and generates and analyses a probabilistic semantic map of \textsc{when}-clauses using a massively parallel Bible corpus, to investigate whether the division of labour between participle clauses and finite temporal subordinates in Early Slavic and Ancient Greek corresponds to well-attested cross-linguistic patterns. It demonstrates that `null' constructions (juxtaposed clauses such as participles, converbs, or independent clauses) cluster in particular regions of the semantic map cross-linguistically, clearly indicating that participle clauses are not equally viable as alternatives to any use of \textsc{when}, but carry particular meanings that make them less suitable for some of its functions. It also shows that the distinction found in Ancient Greek between \textit{hóte} and \textit{hótan}, not found in Early Slavic (which only has \textit{jegda}), corresponds to a widely attested cross-linguistic pattern, pointing to a general typological difference between the two languages. \textbf{Chapter 6} addresses the issue of understanding the functions of null constructions in the semantic map of \textsc{when}, started in the previous chapter, not only at the level of their competition with \textit{when}-counterparts but also with respect to the competition between different null constructions, which the semantic map of \textsc{when} could not fully address. It provides a comparison between, on the one hand, conjunct participles and absolutes, and, on the other hand, cross-linguistic phenomena known to have similar functions to them, specifically clause chaining, bridging, insubordination, and switch reference.\\
\indent The clear parallels to these phenomena not only provide independent evidence for the observations made in previous chapters on the division of labour between participle constructions, but they are also able to motivate several of the uses of `non-canonical' participles constructions, particularly dative absolutes, which had otherwise often been written off as `aberrations' by the earlier scholarship on Early Slavic.

% TODO
% - highlight some terms and add subheaders or, better paragraphs

\chapter{Participle constructions in deeply-annotated treebanks}

\section{Introduction}
In this chapter and the next, I provide a descriptive and quantitative analysis of conjunct participles and dative absolutes as they appear in the Early Slavic corpus described in the Introduction.\footnote{This chapter and the next expand on material published in \citet{pedrazzinijhs}.} I present statistics about their distribution and analyze the constructions across variables which can indirectly capture their discourse functions. The main framework for this analysis is \citet{baryhaug2011} and the corpus evidence for their treatment of conjunct participles in Ancient Greek in \citet{haug2012a}. Similarly to the latter, the main variables I attempt to capture as a possible indication of the discourse functions of participle constructions are: 
\begin{itemize}
    \item[a.] the relative order of participle and matrix verb;
    \item[b.] aspect distribution;
    \item[c.] the properties of the subjects involved;
    \item[d.] the lexical variation among participles across different sentence configurations (i.e. pre- and post-matrix participle clauses, different positions of overt subjects relative to the participle).
\end{itemize}
As outlined in the Introduction, \posscitet{baryhaug2011} framework is mostly built on conjunct participles. In this chapter, however, I apply the same criteria to the analysis of both conjunct participles and absolute constructions, with the goal of gauging whether:
\begin{itemize}
    \item[1.] the functions of Early Slavic conjunct participles can be modelled in the same way as in Ancient Greek, namely within a main tripartite distinction between \textsc{frames}, \textsc{independent rhemes} and \textsc{elaborations}, by capturing these via indirect corpus evidence as defined above;
    \item[2.] the functions of absolute constructions can also be modelled as those of conjunct participles, namely as either \textsc{frames}, \textsc{independent rhemes} and \textsc{elaborations}, or whether their overall behaviour cannot be predicted based on the same criteria.
\end{itemize}

The analysis will be structured following the annotation depths available for each of the Early Slavic subcorpora and divided across the current and the next chapter. This chapter exploits data from a \textit{deeply-annotated} treebank, the Old Church Slavonic Codex Marianus, currently the only text in the corpus aligned at the token level with the Greek source text and the only treebank for which information-structural annotation is also available, alongside several other annotation layers on large portions of the text, including Aktionsart and rhetorical relations. Comparison with the Greek parallel is made throughout the section by providing the Greek counterpart and commenting on differences, where relevant. The next chapter will deal with treebanks without information-structural or verb-semantic annotation, which were referred to as \textit{standard treebanks} in the Introduction and which contain mixed data from Old Church Slavonic, different varieties of Church Slavonic, and Old East Slavic. It will also look at \textit{strategically-annotated} treebanks, consisting of fully automatically annotated Middle Bulgarian and Old Serbian texts. This structure also serves, incidentally, as a proof of concept showing the advantages or disadvantages of building small, but deeply-annotated treebanks for historical languages with very high diatopic and diachronic variation (such as Early Slavic), as opposed to larger corpora with shallower annotation, as well as the extent to which methodology and results may differ when using historical treebanks with different annotation depths\\
\indent Section \ref{deep-order} of this chapter comments on the relative order of participle constructions and matrix clause, and the distribution of aspect among participle constructions in different sentence configurations; Section \ref{deep-subj} analyses the properties of the subjects in participle constructions; Section \ref{deep-variation} looks into the lexical variation among participle constructions in different sentence configurations; Section \ref{greekocscompa} compares Old Church Slavonic to the Ancient Greek parallel, looking at correspondences and mismatches between the two version in the choice of construction; finally, Section \ref{rhetrelsec} takes a look at the rhetorical relation annotation available for the Gospel of Luke.

\section{Position in the sentence and aspect distribution}\label{deep-order}
As the frequencies in Table \ref{tab:tab1} show, there is a very strong tendency for dative absolutes to occur to the left of the main verb, while that is somewhat less the case for conjunct participles.

\begin{table}[!h]
\centering
\begin{tabular}{cccc}
\hline
& \textbf{pre-matrix} & \textbf{post-matrix} & \textit{tot.}\\
\hline
\textbf{absolute} & 92.9\% (171)    & 7.1\% (13)  &  184 \\
\textbf{conjunct} & 66.6\% (1044) & 33.4\% (523) & 1567 \\
\hline
\end{tabular}
\caption[Relative order of participle and matrix clause in the Codex Marianus]{Relative order of participle and matrix clause in the Codex Marianus}
\label{tab:tab1}
\end{table}

The neat predominance of pre-matrix dative absolutes already suggests that the position of dative absolutes in the sentence is connected, in itself, to their functions and supports the widespread view that the unifying function of dative absolutes is a ‘backgrounding’ or ‘stage-setting’ one (\pgcitealt{worth1994a}{30}; \pgcitealt{corin1995a}{259}; \pgcitealt{collins2011a}{113}). As argued by \citet{baryhaug2011} and \citet{haug2012a}, framing participles should always be found in the leftmost position since they set the stage and provide temporal anchoring for the whole sentence. \textsc{frames} can, in fact, be treated as fronted adjuncts, that is, as information-structurally marked material in topicalized position (\pgcitealt{haug2012a}{307}). Like \textit{frame setters} in \pgcitet{chafe1976a}{51} and \pgcitet{krifka2007a}{45-48}, the concept of \textsc{Frame} in \citet{baryhaug2011} is tightly connected to that of aboutness (or sentence) topic. As \citet{fabricius-hansen2012b} explain, frame setters and aboutness topics are similar in that `they both refer to entities [...] that are already established in, or easily inferable from, the preceding discourse’. \textsc{frames} typically refer to a preceding event, as in (\ref{ex8}), where the event providing the topic time for the whole sentence is Jesus casting out a demon. In other instances, they refer to bridging events linking to other events by means of `motion from one scene to another or perception of some previous action’ (\pgcitealt{haug2012a}{299}), as in (\ref{ex9}), where the temporal anchor for the whole sentence is set by Jesus and the disciples moving – with the presupposition that they have set off from a previous setting. 

\begin{example}
\begin{itemize}
\item[a.]
\gll i \textbf{izg\foreignlanguage{russian}{ъ}nanu} běsu progla němy 
and cast.out.\textsc{ptcp.pfv.pass.m.dat.sg} demon.\textsc{m.dat.sg} speak.\textsc{aor.3.sg} mute.\textsc{m.nom.sg}
\glt
\glend
\item[b.]
\gll kai \textbf{ekblēthentos} tou daimoniou elalēsen ho kōphos
and cast.out.\textsc{ptcp.pfv.pass.m.gen.sg} the.\textsc{m.gen.sg} demon.\textsc{m.gen.sg} speak.\textsc{aor.3.sg} the.\textsc{m.nom.sg} mute.\textsc{m.nom.sg}
\glt ‘And when the demon was cast out, the mute spoke’ (Matthew 9.33, 38591, 15005)
\glend
\label{ex8}
\end{itemize}
\end{example}

\begin{example}
\begin{itemize}
\item[a.]
\gll i \textbf{s\foreignlanguage{russian}{ъ}chodęštem\foreignlanguage{russian}{ъ}} im\foreignlanguage{russian}{ъ} s\foreignlanguage{russian}{ъ} gory zapovědě im\foreignlanguage{russian}{ъ} is\foreignlanguage{russian}{ъ} glę 
and come.down.\textsc{ptcp.ipfv.dat.pl} he.\textsc{dat.pl} from mountain.\textsc{gen.sg} command.\textsc{aor.3.sg} he.\textsc{dat.pl} Jesus.\textsc{nom} say.\textsc{ptcp.ipfv.m.nom.sg}
\glt
\glend
\item[b.] 
\gll kai \textbf{katabainontōn} autōn ek tou orous eneteilato autois ho Iēsous legōn 
and come.down.\textsc{ptcp.ipfv.gen.pl} he.\textsc{gen.pl} from the.\textsc{n.gen.sg} mountain.\textsc{n.gen.sg} instruct.\textsc{aor.3.sg} he.\textsc{dat.pl} the Jesus.\textsc{nom} say.\textsc{ptcp.ipfv.m.nom.sg}
\glt ‘And as they came down from the mountain, Jesus commanded them, saying’ 
(Matthew 17:9)
\glend
\label{ex9}
\end{itemize}
\end{example}

Similar observations can, however, be made for several sentence-initial conjunct participles, as the one in (\ref{xadvframepfv}), which refers to an event in the earlier discourse (the \textit{Miraculous Catch of Fish} in Luke 5.1-7), or the one in (\ref{xadvframeipfv}), which anchors the topic time for the instruction contained in the matrix by referring to the time frame within which its interpretation holds (‘when praying’).

\begin{example}
\begin{itemize}
\item[a.]
\gll \textbf{viděv\foreignlanguage{russian}{ъ}} že simon\foreignlanguage{russian}{ъ} petr\foreignlanguage{russian}{ъ}. pripade k\foreignlanguage{russian}{ъ} kolěnoma isvama glę
see.\textsc{ptcp.pfv.m.nom.sg} \textsc{ptc} Simon.\textsc{nom} Peter.\textsc{nom} fall.down.\textsc{prs.3.sg} to knee.\textsc{dat.du} Jesus.\textsc{adj.dat.du} say.\textsc{ptcp.ipfv.m.nom.sg}
\glt
\glend
\item[b.]
\gll \textbf{idōn} de Simōn Petros prosepesen tois gonasin Iēsou legōn 
see.\textsc{ptcp.pfv.m.nom.sg} \textsc{ptc} Simon.\textsc{nom} Peter.\textsc{nom} fall.\textsc{prs.3.sg} the.\textsc{dat.pl} knee.\textsc{dat.pl} Jesus.\textsc{gen} say.\textsc{ptcp.ipfv.m.nom.sg}
\glt ‘When Simon Peter saw (that), he fell down at Jesus’ knees’ (Luke 5:8)
\glend
\label{xadvframepfv}
\end{itemize}
\end{example}

\begin{example}
\begin{itemize}
\item[a.]
\gll \textbf{molęšte} že sę ne licho glte. ěkože i języč\foreignlanguage{russian}{ъ}nici 
pray.\textsc{ptcp.ipfv.nom.pl} \textsc{ptc} \textsc{refl} \textsc{neg} vainly speak.\textsc{imp.2.pl} as also heathen.\textsc{nom.pl}
\glt
\glend
\item[b.]
\gll \textbf{proseuchomenoi} de mē battalogēsēte hōsper hoi ethnikoi
pray.\textsc{ptcp.ipfv.nom.pl} \textsc{ptc} \textsc{neg} speak.vainly.\textsc{sbjv.aor.2.pl} as the heathen.\textsc{nom.pl}
\glt ‘And when you pray, do not use vain repetitions as the heathen do’ (Matthew 6:7)
\glend
\label{xadvframeipfv}
\end{itemize}
\end{example}

Although \textsc{frames} should generally be looked for to the left of the matrix verb, not all sentence-initial participles necessarily serve that function. Among dative absolutes, a prototypical \textsc{frame} function seems likely because those in pre-matrix position \textit{far} outnumber those following the matrix. Among conjunct participles, a clear pattern instead emerges when we cross their position in the sentence with the frequency of imperfective and perfective forms depending on the position of the participle relative to the matrix.

\begin{table}[!h]
\centering
\begin{tabular}{ccc}
\hline
& \textbf{imperfectives} & \textbf{perfectives} \\
\hline
\textbf{pre-matrix} & 59.1\% (101) &	40.9\% (70) \\
\textbf{post-matrix} & 53.8\% (7) &	46.2\% (6) \\
\textbf{tot.} & 58.7\% (108) &	41.3\% (76) \\
\hline
\end{tabular}
\caption[Dative absolutes in the Codex Marianus: aspect distribution by position relative to the matrix clause]{Dative absolutes in the Codex Marianus: aspect distribution by position relative to the matrix clause (row percentage)}
\label{deepdasasp}
\end{table}

\begin{table}[!h]
\centering
\begin{tabular}{ccc}
\hline
& \textbf{imperfectives} & \textbf{perfectives} \\
\hline
\textbf{pre-matrix} & 13.1\% (121) &	86.9\% (806)     \\
\textbf{post-matrix} & 80.3\% (216) &	19.7\% (53) \\
\textbf{tot.} & 28.2\% (337) &	71.8\% (859) \\
\hline
\end{tabular}
\caption[Conjunct participles in the Codex Marianus: aspect distribution by position relative to the matrix clause]{Conjunct participles in the Codex Marianus: aspect distribution by position relative to the matrix clause (row percentage)}
\label{deepxadvasp}
\end{table}

As Table \ref{deepdasasp} shows, among dative absolutes, imperfectives are somewhat more frequent. Post-matrix absolutes are too few to say anything about possible associations between aspect and position. However, considering only pre-matrix occurrences, a binomial test (2-tailed) suggests that there is a significantly higher probability for imperfectives than perfectives to occur in pre-matrix position ($p=0.01$). This is, naturally, an incomplete picture based on few occurrences\footnote{The binomial test can most intuitively seen as a test for the probability of successes. In the case of dative absolutes in this dataset, out of 171 attempts (i.e. the total number of occurrences), 101 were successful, 70 were not. However, the relatively small number of overall occurrences means that, in this case, only 5 occurrences would have changed the test result from significant to non-significant, hence why one should be wary of it for now.} and what really is noticeable is the difference in number between pre- and post-matrix occurrences. \\
\indent As Table \ref{deepxadvasp} shows, among conjunct participles, the association between aspect and position is statistically significant, with a strong effect size ($\phi=0.62$, Cramér's $V$): $\chi^{2}(1) = 462.54$, $p<0.01$, with much higher odds for perfectives to occur in pre-matrix than post-matrix position and for imperfectives to occur in post-matrix than pre-matrix position (odds ratio: 27.15). This is in line with \posscitet{haug2012a} Ancient Greek data, showing that participles to the right of the matrix are more likely to be \textsc{elaborations} while those to the left are more typically \textsc{independent rhemes}. \textsc{Elaborations} normally result in a `complex rheme', where the event described by the elaborating participle adds granularity to the lexical meaning of the matrix verb and often receives a manner or means interpretation, as the example in (\ref{elabascomplexrheme}). \textsc{Independent rhemes} are instead similar to independent clauses from the discourse perspective, as in (\ref{indrehemeindclause}).

\begin{example}
\begin{itemize}
\item[a.]
\gll i v\foreignlanguage{russian}{ъ}str\foreignlanguage{russian}{ъ}zaachǫ učenici ego klasy. i ěděachǫ \textbf{istirajǫšte} rǫkama
and pluck.\textsc{impf.3.pl} disciple.\textsc{nom.pl} \textsc{3.sg.gen} head.of.grain.\textsc{acc.pl} and eat.\textsc{impf.3.pl} rub.\textsc{ptcp.ipfv.nom.pl} hand.\textsc{ins.du}
\glt
\glend
\item[b.]
\gll kai etillon hoi mathētai autou tous stachyas kai ēsthion \textbf{psōchontes} tais chersin
and pluck.\textsc{impf.3.pl} the disciple.\textsc{nom.pl} \textsc{3.sg.gen} the.\textsc{acc.pl} head.of.grain.\textsc{acc.pl} and eat.\textsc{impf.3.pl} rub.\textsc{ptcp.ipfv.nom.pl} the.\textsc{f.dat.pl} hand.\textsc{f.dat.pl}
\glt ‘And his disciples plucked the heads of grain and ate them, rubbing them in their hands’ (Luke 6:1)
\glend
\label{elabascomplexrheme}
\end{itemize}
\end{example}

\begin{example}
\begin{itemize}
\item[a.]
\gll az\foreignlanguage{russian}{ъ} \textbf{prišed\foreignlanguage{russian}{ъ}} iscěljǫ i
{\sc 1.sg.nom} {come.{\sc ptcp.pfv.m.nom.sg}} {heal.{\sc prs.1.sg}} {\sc 3.sg.m.acc}
\glt
\glend
\item[b.]
\gll {egō} {elthōn} {therapeusō} {auton}
{\sc 1.sg.nom} {come.{\sc ptcp.pfv.m.nom.sg}} {heal.{\sc fut.1.sg}} {\sc 3.sg.m.acc}
\glt ‘I will come and heal him’ (Matthew 8:7)  %50710
\glend
\label{indrehemeindclause}
\end{itemize}
\end{example}

As pointed out by \pgcitet{haug2012a}{311}, while the perfective is generally the dominant aspect in narrative contexts, this is particularly true of the simple narrative style of the New Testament. The preponderance of perfective \textsc{independent rhemes} is thus as unsurprising as that of perfective main verbs. Stacked perfective \textsc{independent rhemes} always induce narrative progression (\pgcitealt{baryhaug2011}{15-16}), as clearly transpires from (\ref{narrprogr}), for example. Although much less frequent, even when imperfective, they are equivalent to main verbs from the information-structural perspective, which is often reflected in their English translation, as shown in (\ref{imperfindrheme}).

\begin{example}
\begin{itemize}
\item[a.]
\gll i abie \textbf{tek\foreignlanguage{russian}{ъ}} edin\foreignlanguage{russian}{ъ} ot\foreignlanguage{russian}{ъ} nich\foreignlanguage{russian}{ъ}. i \textbf{priem\foreignlanguage{russian}{ъ}} gǫbǫ. \textbf{ispl\foreignlanguage{russian}{ь}n\foreignlanguage{russian}{ь}} oc\foreignlanguage{russian}{ь}ta. i \textbf{v\foreignlanguage{russian}{ь}znez\foreignlanguage{russian}{ъ}} na tr\foreignlanguage{russian}{ь}st\foreignlanguage{russian}{ь}. napaěše i
and immediately run.\textsc{ptcp.pfv.m.nom.sg} one.\textsc{m.nom.sg} from \textsc{3.pl.gen} and take.\textsc{ptcp.pfv.m.nom.sg} sponge.\textsc{acc.sg} fill.\textsc{ptcp.pfv.m.nom.sg} vinegar.\textsc{gen.sg} and put.\textsc{ptcp.pfv.m.nom.sg} on reed.\textsc{acc.sg} give.to.drink.\textsc{impf.3.sg} \textsc{3.sg.acc}
\glt
\glend
\item[b.]
\gll kai eutheōs \textbf{dramōn} heis ex autōn kai \textbf{labōn} spongon \textbf{plēsas} te oxous kai \textbf{peritheis} kalamō epotizen auton
and immediately run.\textsc{ptcp.pfv.m.nom.sg} one.\textsc{nom.sg} from \textsc{3.pl.gen} and take.\textsc{ptcp.pfv.m.nom.sg} sponge.\textsc{acc.sg} fill.\textsc{ptcp.pfv.m.nom.sg} with vinegar.\textsc{gen.sg} and put.\textsc{ptcp.pfv.m.nom.sg} reed.\textsc{dat.sg} give.to.drink.\textsc{impf.3.sg} \textsc{3.sg.acc}
\glt ‘Immediately one of them ran and took a sponge, filled it with sour wine and put it on a reed, and offered it to him to drink’ (Matthew 27:48)
\glend
\label{narrprogr}
\end{itemize}
\end{example}

\begin{example}
\begin{itemize}
\item[a.]
\gll n\foreignlanguage{russian}{ъ} i psi \textbf{prichodęšte} oblizaachǫ gnoi ego
but also dog.\textsc{nom.pl} come.\textsc{ptcp.ipfv.nom.pl} lick.\textsc{impf.3.pl} wound.\textsc{acc.pl} \textsc{3.sg.gen}
\glt
\glend
\item[b.]
\gll alla kai hoi kynes \textbf{erchomenoi} epeleichon ta helkē autou
but also the dog.\textsc{nom.pl} come.\textsc{ptcp.ipfv.nom.pl} lick.\textsc{impf.3.pl} the.\textsc{acc.pl} wound.\textsc{acc.pl} \textsc{3.sg.gen}
\glt ‘Even the dogs would come and lick his sores’ (Luke 16:21)
\glend
\label{imperfindrheme}
\end{itemize}
\end{example}

Given the significant association among conjunct participles between the pre-matrix position and the perfective aspect, it seems likely that the \textit{typical} function of pre-matrix conjunct participles is as \textsc{independent rhemes}. But as \ref{xadvframepfv} already showed, the sentence-initial position can be ambiguous between a \textsc{frame} and an \textsc{independent rheme} interpretation, as \citet{baryhaug2011} already argued for Ancient Greek conjunct participles. In (\ref{xadvframeindrhambig}), the ambiguity of the sentence-initial conjunct participle between a frame and an independent rheme function is, to some extent, also reflected in the different English translations.

\begin{example}
\begin{itemize}
\item[a.]
\gll i \textbf{postav\foreignlanguage{russian}{ь}še} jǫ po srědě. glašę emu.
{and} {place.{\sc ptcp.pfv.nom.pl}} {\sc 3.sg.f.acc} {in} {midst.{\sc dat}} {say.{\sc aor.3.pl}} {\sc 3.sg.m.dat}
\glt
\glend
\item[b.]
\gll {kai} {\textbf{stēsantes}} {autēn} {en} {mesōi} {legousin} {autōi}
{and} {set up.{\sc ptcp.pfv.nom.pl}} {\sc 3.sg.f.acc} {in} {midst.{\sc dat}} {say.{\sc prs.3.pl}} {\sc 3.sg.m.dat}
\glt `They made her stand before the group and said to Jesus' (New International Version)\\
‘When they had set her in the midst, they say unto him' (King James Bible)\\
`Placing her in the midst, they said to him' (English Standard Version)\\
(John 8:3–4) %42113
\glend
\label{xadvframeindrhambig}
\end{itemize}
\end{example}

\indent As we will see, subject realization and the position of overt subjects relative to the participle (specifically, the VS configuration) have some part in their interpretation as \textsc{frames}. Overall, however, sentence-initial position conjunct participles will inevitably retain a certain degree of ambiguity, particularly with a null subject.

\subsection{A remark about \textit{byst\foreignlanguage{russian}{ъ}}-clauses}\label{bystsec}
Before looking at other variables, it is worthwhile to comment briefly on a small, but coherent group of occurrences of dative absolutes in a construction introduced by \textit{byst\foreignlanguage{russian}{ъ}} ‘it came to pass’, the usage of which in Old Church Slavonic can clearly be linked to a relatively widespread and well-studied construction attested in several pre-modern Indo-European languages, consisting, in its basic form, of

\begin{example}
    $ \textsc{it came to pass} \quad [... ... ...]_{\alpha} \quad [... ... ...]_{\beta}. $ 
\end{example}

where $\alpha$ is filled by an adverbial clause and $\beta$ by the main predication. Works on discourse marking (e.g. \pgcitealt{traugott2002a}{52}; \pgcitealt{brinton1996a}{336}) have treated ‘happen, come to pass’ in such contexts on a par with discourse connectives such as \textit{in fact}, \textit{well}, \textit{then} or \textit{so}, which hardly contribute to the truth-conditional meaning of a proposition, but rather mark the speaker’s stance towards the sequential relationship between discourse units. In particular, the discourse functions of Old English \textit{gelamp}-clauses (the counterpart to Slavonic \textit{byst\foreignlanguage{russian}{ъ}}) and the syntactic status of its constituents, which have been thoroughly analyzed by \seccitet{brinton1996a}{5-6}, strongly support the intuition that Old Church Slavonic \textit{byst\foreignlanguage{russian}{ъ}}, like its Greek counterpart \textit{egeneto} (and analogous forms in Gospel translations in other historical languages, such as Gothic \textit{warþ} or Latin \textit{factum est}), has a clear formulaic function. \pgcitet{brinton1996a}{134}\footnote{Note that \citet{brinton1996a} adduces plenty of support from previous literature. In particular, \citeauthor{traugott1982a}’s extensive work on grammaticalization provides further solidity to Brinton’s claims. Traugott (e.g. \citeyear{traugott1982a}; [with König] \citeyear{traugott1991a}; [with Dasher] \citeyear{traugott2002a}; \citeyear{traugott2003a}; [with Hopper] \citeyear{hopper2003a}; \citeyear{traugott2004a}; \citeyear{traugott2008a}) has dealt with a large variety of utterance-to-discourse-marker grammaticalization processes, which have helped tackle the status of ‘happen’ in Old English \textit{gelamp}-constructions in \citet{brinton1996a}. For an in-depth survey of the most influential studies on discourse markers, including Traugott’s seminal works, see \citet{maschler2015a}.} argues that these are typically made of an episode-initiating, backgrounded clause which gives the time, place or circumstances for a (linearly) following discourse-foregrounded clause, which introduces the narrative proper. \pgcitet{brinton1996a}{142} states that ‘the discourse strategy [in \textit{gelamp}-clauses] can be better understood as framing, as well as focusing, where the main clause [‘it came to pass’] and adverbial provide a frame which directs the reader’s focus to the events denoted in the complement clause’. Similar considerations are made by \pgcitet{hogeterp2018a}{334-336} on \textit{egeneto} formulas in New Testament Greek, where \textit{egeneto} constructions are considered to be ‘important linguistic markers of narrative turns’, with the ‘structural function to introduce a new literary section or pericope'.\\
\indent It is interesting to note that several of the mismatches between Old Church Slavonic and the Greek original (i.e. where Old Church Slavonic uses an absolute construction and Greek does not) occur in \textit{byst\foreignlanguage{russian}{ъ}}-constructions. This suggests that the particular meaning of the $\alpha$ slot in this construction may correspond particularly well to the overall function of dative absolutes in Old Church Slavonic. (\ref{ex4})-(\ref{ex7}) are some examples of dative absolutes in \textit{byst\foreignlanguage{russian}{ъ}}-constructions.

\begin{example}
\begin{itemize}
\item[a.]
\gll i byst\foreignlanguage{russian}{ъ} \textbf{idǫštem\foreignlanguage{russian}{ъ}} im\foreignlanguage{russian}{ь}. ištistišę sę
and happen.\textsc{aor.3.sg} go.\textsc{ptcp.ipfv.dat.pl} \textsc{3.pl.dat} cleanse.\textsc{aor.3.pl} \textsc{refl}
\glt
\glend
\item[b.]
\gll kai egeneto \textbf{en} \textbf{tōi} \textbf{hupagein} autous ekatharisthēsan
and happen.\textsc{aor.3.sg} in the.\textsc{n.dat.sg} go.\textsc{prs.inf} \textsc{3.pl.acc} cleanse.\textsc{aor.pass.3.pl} 
\glt ‘And it came to pass that, as they went, they were cleansed’ (Luke 17:14)
\glend
\label{ex4}
\end{itemize}
\end{example}

\begin{example}
\begin{itemize}
\item[a.]
\gll byst\foreignlanguage{russian}{ъ} že běsu \textbf{izg\foreignlanguage{russian}{ъ}nanu} progla němy
happen.\textsc{aor.3.sg} \textsc{ptc} demon.\textsc{m.dat.sg} cast.out.\textsc{ptcp.pfv.pass.m.dat.sg} speak.\textsc{aor.3.sg} mute.\textsc{m.nom.sg}
\glt
\glend
\item[b.]
\gll egeneto de tou daimoniou \textbf{exelthontos} elalēsen ho kōphos
happen.\textsc{aor.3.sg} \textsc{ptc} the.\textsc{m.gen.sg} demon.\textsc{m.gen.sg} cast.out.\textsc{ptcp.pfv.pass.m.gen.sg} speak.\textsc{aor.3.sg} the.\textsc{m.nom.sg} mute.\textsc{m.nom.sg}
\glt ‘It came to pass that, as the demon had been cast out, the mute man spoke’ (Luke 11:14)
\glend
\label{ex5}
\end{itemize}
\end{example}

\begin{example}
\begin{itemize}
\item[a.]
\gll byst\foreignlanguage{russian}{ъ} že \textbf{chodęštem\foreignlanguage{russian}{ъ}} im\foreignlanguage{russian}{ъ} i t\foreignlanguage{russian}{ъ} v\foreignlanguage{russian}{ъ}nide is\foreignlanguage{russian}{ъ} v\foreignlanguage{russian}{ъ} ves\foreignlanguage{russian}{ь} edinǫ
happen.\textsc{aor.3.sg} \textsc{ptc} walk.\textsc{ptcp.ipfv.dat.pl} he.\textsc{dat.pl} and that.\textsc{m.nom.sg} enter.\textsc{aor.3.sg} Jesus.\textsc{nom} in village.\textsc{f.acc.sg} certain.\textsc{f.acc.sg}
\glt
\glend
\item[b.]
\gll egeneto de \textbf{en} \textbf{tōi} \textbf{poreuesthai} autous kai autos eisēlthen eis kōmēn tina
happen.\textsc{aor.3.sg} \textsc{ptc} in the.\textsc{n.dat.sg} proceed.\textsc{prs.inf} he.\textsc{acc.pl} and he.\textsc{nom.sg} enter.\textsc{aor.3.sg} into village.\textsc{f.acc.sg} certain.\textsc{f.acc.sg}
\glt ‘It came to pass that, as they were walking, Jesus entered a certain village’ (Luke 10:38)
\glend
\label{ex6}
\end{itemize}
\end{example}

\begin{example}
\begin{itemize}
\item[a.]
\gll i byst\foreignlanguage{russian}{ъ} \textbf{idǫštju} emu v\foreignlanguage{russian}{ъ} im\foreignlanguage{russian}{ъ}. i t\foreignlanguage{russian}{ъ} prochoždaaše meždju samariejǫ i galilěejǫ
and happen.\textsc{aor.3.sg} go.\textsc{ptcp.ipfv.m.dat.sg} he.\textsc{dat.sg} to Jerusalem.\textsc{acc} and that.\textsc{m.nom.sg} walk.through.\textsc{impf.3.sg} between Samaria.\textsc{f.ins.sg} and Galilee.\textsc{f.ins.sg}
\glt 
\glend
\item[b.]
\gll kai egeneto \textbf{en} \textbf{tōi} \textbf{poreuesthai} eis Hierousalēm kai autos diērcheto dia meson Samarias kai Galilaias.
and	happen.\textsc{aor.3.sg} in the.\textsc{n.dat.sg} proceed.\textsc{prs.inf} to Jerusalem.\textsc{acc} and he.\textsc{nom.sg} walk.through.\textsc{impf.3.sg} through midst.\textsc{n.acc.sg} Samaria.\textsc{gen.sg} and Galilee.\textsc{gen.sg}
\glt ‘And it came to pass, as he went to Jerusalem, that he passed through the midst of Samaria and Galilee’ (Luke 17:11)
\glend
\label{ex7}
\end{itemize}
\end{example}

While (\ref{ex5}a) translates a Greek genitive absolute (\ref{ex5}b), the dative absolutes in (\ref{ex4}a) and (\ref{ex6}a)-(\ref{ex7}a) correspond to a nominalized accusative with infinitive introduced by \textit{en} ‘in, during’. The second conjunction in (\ref{ex6})-(\ref{ex7}), apparently introducing the main clause, can in fact be seen as a calque from the Greek, which, in turn, is likely a calque from Hebrew: the construction (\textit{kai}) \textit{egeneto} (\textit{de}) ‘(and) it came to pass that’ itself, opening a narrative section, has been recognized as a syntactic Semitism characteristic of Luke’s Gospel (\seccitealt{hogeterp2018a}{6.1}; \pgcitealt{janse2007a}{657}). \pgcitet{hogeterp2018a}{326} specifically single out the configuration (\textit{kai}) \textit{egeneto} \textit{de} + temporal expression + \textit{kai} +  finite verb (i.e. with a second conjunction seemingly introducing the main clause) as the one standing out the most for its ‘Hebraistic quality’ among the possible \textit{egeneto} formulas, where the second \textit{kai} would reflect the temporal use of Biblical Hebrew \textit{waw}\footnote{Biblical Hebrew \textit{waw} has been described as an ‘all-purpose connector’ (\citealt{steiner2000a}), which ‘places propositions or clauses one after another, without indicating the hierarchical relation between them’ and ‘frequently joins logically subordinate clauses to a main clause’ (\pgcitealt{waltke1990a}{649}).} in constructions of the type ‘when … then’ (\pgcitealt{waltke1990a}{436}). This is claimed by the authors by comparing the occurrences of the same construction in Septuagint Greek with their counterparts in the Masoretic Text, as well as with the usage of \textit{en toi} + infinitive followed by \textit{kai} in Ancient Greek, which appears to be largely at odds with the one in Koine Greek\footnote{\pgcitet{hogeterp2018a}{436} find that ‘in the relatively few cases where \textit{en tōi} with infinitive has a temporal significance in the voluminous literary Greek works of Thucydides, Polybius, Diodorus Siculus, and Longus, it is never syndetically followed by an apodotic \textit{kai} in their works’.}.\\
\indent Whichever syntactic analyses for (\ref{ex4})-(\ref{ex7}) one decides to endorse, the function of the dative absolute in \textit{byst\foreignlanguage{russian}{ъ}}-clauses is thus consistent throughout – namely, to provide the setting, or frame, for the eventuality expressed by the following finite clause. 

\subsection{Summary}
In this section, we observed that the vast majority of dative absolutes occur before the matrix clause, suggesting that the position of absolutes in the sentence is, in itself, likely to be associated with their primary function. This can be identified as a stage-setting or backgrounding one, as previous studies on Early Slavic dative absolutes had already observed, or, under \posscitet{baryhaug2011} terminology, as \textsc{frame} participles. Conjunct participles, similarly to their Ancient Greek counterparts (as per the corpus evidence in \citealt{haug2012a}), showed a significant association between the position of the participle relative to the matrix clause and the aspect of the participle. Pre-matrix conjunct participles are much more likely to be perfective, whereas post-matrix conjunct participles are much more likely to be imperfective. This is in line with \posscitet{baryhaug2011} analysis, whereby conjunct participles as \textsc{independent rhemes} are typically pre-matrix and perfective, whereas as \textsc{elaborations} they are typically imperfective and generally occur after the matrix verb. We have also seen that, sentence-initially, conjunct participles, even when perfective, can also function as \textsc{frames}, either clearly so, as in (\ref{xadvframepfv}), or as a possible interpretation in ambiguous cases such as (\ref{xadvframeindrhambig}), where they could also be interpreted as \textsc{independent rhemes}.\\
\indent Finally, I made a digression on \textit{byst\foreignlanguage{russian}{ъ}}-clauses (`it came to pass that...), a construction occurring rather frequently in the Gospels which has widely studied counterparts in other pre-modern Indo-European languages. The construction is made of functionally relatively fixed slots, and we have seen that dative absolutes often occur in what has been analysed as the episode-initiating, backgrounded predication which precedes a following foregrounded, main eventuality. This observation independently supports the analysis of dative absolutes as prototypical \textsc{frames}, particularly since their usage in \textit{byst\foreignlanguage{russian}{ъ}}-constructions occurs independently of the Greek sources.

% between dative absolutes as a whole and post-matrix conjunct participles, which are thus most typically expected to be \textsc{elaborations}. Prototypical \textsc{frames} have, as it were, a diametrically opposite status to that of prototypical \textsc{elaborations}, both in terms of surface realization (the latter more likely post-matrix, the former most likely sentence-initial) and of information status (\textsc{elaborations} adding new information, \textsc{frames} topicalizing a mentioned or accessible event). On the other hand, pre-matrix conjunct and absolute participles are still, to some extent, potentially competing \textsc{frames}. 

\section{Subjects}\label{deep-subj}
Examples like (\ref{exboh1}) and (\ref{exboh2}) are discursively very similar. The main difference appears to be syntactic: the subject of the conjunct participle in (\ref{exboh2}) is co-indexed with that of the matrix clause, whereas the subject of the dative absolutes in (\ref{exboh1}) is not.

\begin{example}
\begin{itemize}
\item[a.]
\gll i \textbf{v\foreignlanguage{russian}{ъ}šed\foreignlanguage{russian}{ъ}šju} emu v\foreignlanguage{russian}{ъ} dom\foreignlanguage{russian}{ъ}. učenici ego v\foreignlanguage{russian}{ъ}prašachǫ i edinogo. ěko
{and} {enter.{\sc ptcp.pfv.m.dat.sg}} {\sc 3.sg.m.dat} {in} {house.{\sc acc.sg}} {disciple.{\sc nom.pl}} {\sc 3.sg.m.gen} {ask.{\sc impf.3.pl}} {\sc 3.sg.m.acc} {one.{\sc m.gen.sg}} {that}
\glt
\glend
\item[b.] 
\gll {kai} {\textbf{eiselthontos}} {autou} {eis} {oikon} {hoi} {mathētai} {autou} {kat'} {idian} {epērōtōn} {auton}
{and} {enter.{\sc ptcp.pfv.m.gen.sg}} {\sc 3.sg.m.gen} {in} {house.{\sc acc.sg}} {the.{\sc m.nom.pl}} {disciple.{\sc m.nom.pl}} {\sc 3.sg.m.gen} {in} {private.{\sc acc}} {ask.{\sc impf.3.pl}} {\sc 3.sg.m.acc}
\glt `And when he had entered the house, his disciples began to ask him privately' (Mark 9:28) %36851
\glend
\label{exboh1}
\end{itemize}
\end{example}

\begin{example}
\begin{itemize}
\item[a.]
\gll i \textbf{v\foreignlanguage{russian}{ъ}šed\foreignlanguage{russian}{ъ}} v\foreignlanguage{russian}{ъ} crkv\foreignlanguage{russian}{ъ}. načęt\foreignlanguage{russian}{ъ} izgoniti prodajǫštęję v\foreignlanguage{russian}{ь} nei
{and} {enter.{\sc ptcp.pfv.m.nom.sg}} {in} {temple.{\sc acc.sg}} {start.{\sc aor.3.sg}} {drive out.{\sc inf}} {sell.{\sc ptcp.ipfv.acc.pl}} {in} {\sc 3.sg.f.loc}
\glt
\glend
\item[b.] 
\gll {Kai} \textbf{{eiselthōn}} {eis} {to} {hieron} {ērksato} {ekballein} {tous} {pōlountas}
{and} {enter.{\sc ptcp.pfv.m.nom.sg}} {in} {the.{\sc acc.sg}} {temple.{\sc acc.sg}} {rule.{\sc aor.3.sg}} {drive out.{\sc inf.prs}} {the.{\sc m.acc.pl}} {sell.{\sc ptcp.ipfv.acc.pl}}
\glt `When Jesus entered the temple, he began to drive out those who were selling there' (Luke 19:45) %41204
\glend
\label{exboh2}
\end{itemize}
\end{example}

It has often been observed, however, that in both Early Slavic (\citealt{collins2004a, collins2011a}) and Ancient Greek (\citealt{fuller}; \citealt{haug2011a}) subject co-indexation between absolutes and matrix clauses is, in fact, attested in the Gospels (\ref{chugrcsamesubj}), as well as in Classical Greek (\ref{grcsamesubj}) and Early Slavic texts (\ref{chusamesubj}), despite traditional descriptive grammars treating them as `exceptions'.

\begin{example}
\gll i byst\foreignlanguage{russian}{ъ} \textbf{idǫštem\foreignlanguage{russian}{ъ}} im\foreignlanguage{russian}{ь}. ištistišę sę
and happen.\textsc{aor.3.sg} go.\textsc{ptcp.ipfv.dat.pl} he.\textsc{dat.pl} cleanse.\textsc{aor.3.pl} \textsc{refl}
\glt ‘And it came to pass that, as they went, they were cleansed’ (Luke 17:14)
\glend
\label{chugrcsamesubj}
\end{example}

\begin{example}
\gll {Kroise} {\textbf{anartēmenou}} {seu} {andros} {basileos} {khrēsta} {erga} {kai} {epea} {poieein} {aiteo} {dosin} {hēntina} {bouleai} {toi} {genesthai} {parautika}
{Croesus.{\sc m.voc.sg}} {to be prepared.{\sc ptcp.prf.pass.m.gen.sg}} {yourself.{\sc 2.m.gen.sg}} {man.{\sc m.gen.sg}} {king.{\sc m.gen.sg}} {useful.{\sc pl.n.acc}} {work.{\sc pl.n.acc}} {and} {word.{\sc pl.n.acc}} {make.{\sc prs.inf}} {ask.{\sc 2.sg.prs.imp.mid}} {giving.{\sc sg.f.acc}} {which.{\sc sg.f.acc}} {wish.{\sc 2.sg.prs.mid}} {you.{\sc 2.sg.dat}} {happen.{\sc pfv.pst.inf.mid}} {directly}
\glt `Croesus, now that you, a king, are determined to act and to speak with integrity, ask me directly for whatever favor you like.' (Herodotus, \textit{Histories} 1.90.1; translation by \citealt{herodotus}.)
\glend
\label{grcsamesubj}
\end{example}

\begin{example}
\gll vam\foreignlanguage{russian}{ь} zloslavnom \textbf{suštem} ot kudu imate pokazati jako sergie on\foreignlanguage{russian}{ь} pravoslavn\foreignlanguage{russian}{ь} bě i cha radi postrada
you.\textsc{dat.pl} heretic.\textsc{dat.pl} be.\textsc{ptcp.ipfv.dat.pl} from where will.\textsc{prs.2.pl} show.\textsc{inf} that Sergius.\textsc{nom} that.\textsc{nom.m.sg} orthodox.\textsc{nom.m.sg} be.\textsc{impf.2.sg} and Christ.\textsc{gen} for suffer.\textsc{aor.3.sg}
\glt `Since you are heretics, how will you show that Sergius was an Orthodox and suffered for Christ?' (Euthymius of Tarnovo, \textit{Life of Hilarion of Meglin}, Sbornici of Vladislav Grammarian 171r)
\glend
\label{chusamesubj}
\end{example}

Several partial explanations have been given as to why the authors or translators would not use an agreeing participle where syntactically possible: a change in the subject's referent's semantic role between adverbial and main clause (e.g. \citealt{collins2004a,collins2011a}), or of the `underlying subject' even if not of the `formal subject' (\citealt{levinsohn2000discourse}); the possibility that an absolute `brings out the participial member with more prominence and force' than a conjunct participle (\citealt{winer}); the goal of making the participial clause more prominent (\citealt{goodwin}); or even the scribal late realization that they could have used an agreeing participle instead (\citealt{winer}). While the syntactic factor plays a role, there are broader information-structural and discourse factors involved, which neither morphosyntactic nor semantic-role explanations can single-handedly capture. It may therefore be useful to look into the properties of the subjects of dative absolutes and conjunct participles beyond subject co-referentiality. 

\subsection{Realization and position}\label{subjpositionsecdeep}
As discussed in the Introduction, the annotation convention in PROIEL is to always analyze the argument shared by conjunct participles and their matrix as dependents of the matrix clause. As \citet{haug2011a,haug2017a} argues, however, conjunct participles in some configurations can be considered heads of the shared argument, specifically the leftmost \textsc{frame} or \textsc{independent rheme} conjunct participle linearly preceding the matrix clause. Only the leftmost pre-matrix conjunct participles, therefore, were here considered as potentially able to head their subject and, therefore, to be strictly referred to as \textit{overt}- or \textit{null}-subject. Dative absolutes, on the other hand, have their own internal subject, regardless of configuration. A fair comparison between dative absolutes and conjunct participles, then, may be made between the constructions in sentence-initial position, which is also where the clearest functional overlap occurs, since they can both work as \textsc{frames}. Table \ref{nullvsovertxadvdas} shows the frequency of overt and potential null subjects in sentence-initial conjunct participles and dative absolutes.

\begin{table}[!h]
\centering
\begin{tabular}{cccc}
\hline
& \textbf{overt} & \textbf{null}\\
\hline
\textbf{conjunct (sentence-initial)} & 60.5\% (455) & 39.5\% (297) \\
\textbf{absolute (sentence-initial)} & 85.4\% (123) & 14.6\% (21)\\
\hline
\end{tabular}
\caption{Frequency of overt and null subjects among sentence-initial conjunct participles and dative absolutes in the Codex Marianus}
\label{nullvsovertxadvdas}
\end{table}

The vast majority of sentence-initial dative absolutes have an overt subject. All the 21 dative absolute occurrences classified as `null' in Table \ref{nullvsovertxadvdas} are, in fact, expressions of the type \textit{pozdě byv\foreignlanguage{russian}{ъ}šu} `when it got late', \textit{byv\foreignlanguage{russian}{ъ}šju že d\foreignlanguage{russian}{ь}ni} `when the day came' or \textit{večerou byv\foreignlanguage{russian}{ъ}šju} ‘when the evening came’. In PROIEL/TOROT, these are all analyzed as impersonal temporal constructions consisting of a copula and a predicative adverb or noun, the latter agreeing with the participle. Although one could argue for a finer-grained distinction between personal and impersonal temporal constructions among these occurrences,\footnote{There is arguably a difference between, for example, \textit{pozdě byv\foreignlanguage{russian}{ъ}šu} `when it got late' and \textit{byv\foreignlanguage{russian}{ъ}šju že d\foreignlanguage{russian}{ь}ni} `when the day came'. In the latter, \textit{d\foreignlanguage{russian}{ь}ni} `day' should probably be analysed as the subject of the absolute. Still, manually disambiguating personal from impersonal constructions would not change the overall observation that overt subjects are much more frequent than null subjects among sentence-initial dative absolutes.} from the discourse perspective, these can all still be analyzed as generic frame-setters. \\
\indent Among sentence-initial conjunct participles, there are as many as 39.5\% potential null subjects. Intuitively, this reflects the fact that conjunct participles in sentence-initial position (which can be either \textsc{frames} or \textsc{independent rhemes}) are functionally more diverse than dative absolutes (which are instead typically \textsc{frames}). \\
\indent As \pgcitet{haug2012a}{316} observes, `sentences that shift the subject are more likely to have a frame to anchor it to the previous discourse, whereas this is not as necessary in continuous sequences of events with the same agent'. \textsc{Frames} are particularly useful to reinstate previously mentioned referents that had no longer been in focus in the previous discourse, making them very likely to overtly realize their subject but also, specifically, to have the subject following the participle itself (\pgcitealt{haug2012a}{316-320}). Following \citet{haug2012a}, then, overt-subject conjunct participles in the VS configuration should be more likely to function as \textsc{frames} than \textsc{independent rhemes}. It is not difficult to find examples where this is clearly the case, as in the VS and SV conjunct participles in (\ref{config-a}) and (\ref{config-b}), functioning as a \textsc{frame} and an \textsc{independent rheme}, respectively (as the English translation also suggests).

\begin{example}
\begin{itemize}
\item[a.]
\gll \textbf{Slyšav\foreignlanguage{russian}{ъ}} že junoša slovo otide skr\foreignlanguage{russian}{ъ}bę
{hear.{\sc ptcp.pfv.m.nom.sg}} {\sc ptc} {joung man.{\sc nom.sg}} {word.{\sc acc.sg}} {go away.{\sc aor.3.sg}} {mourn.{\sc ptcp.ipfv.m.nom.sg}}
\glt
\glend
\item[b.] 
\gll {\textbf{akousas}} {de} {ho} {neaniskos} {apēlthen} {lupoumenos}
{hear.{\sc ptcp.pfv.m.nom.sg}} {\sc ptc} {the.{\sc m.nom.sg}} {young man.{\sc m.nom.sg}} {depart.{\sc aor.3.sg}} {offend.{\sc ptcp.ipfv.pas.m.nom.sg}}
\glt But when the young man heard this, he went away sad' (Matthew 19:22) %39171
\glend
\label{config-a}
\end{itemize}
\end{example}

\begin{example}
\begin{itemize}
\item[a.]
\gll i se prokažen\foreignlanguage{russian}{ъ} \textbf{pristǫp\foreignlanguage{russian}{ь}} klaněše sę emu glę
and \textsc{ptc} leper.{\sc m.nom.sg} move-forward.{\sc ptcp.pfv.m.nom.sg} worship.{\sc impf.3.sg} \textsc{refl} him.{\sc dat} say.{\sc ptcp.ipfv.m.nom.sg}
\glt
\glend
\item[b.] 
\gll {kai} {idou} {lepros} {\textbf{proselthōn}} {prosekunei} {autōi} {legōn}
and \textsc{ptc} leper.{\sc m.nom.sg} move-forward.{\sc ptcp.pfv.m.nom.sg} worship.{\sc impf.3.sg} him.{\sc dat} say.{\sc ptcp.ipfv.m.nom.sg}
\glt ‘A man with leprosy came and knelt before him and said' \hfill (Matthew 8:2)%47740
\glend
\label{config-b}
\end{itemize}
\end{example}

As Table \ref{svversusvs} shows, the VS configuration is much more frequent than the SV one among sentence-initial dative absolutes, once again supporting their interpretation as typical \textsc{frames}. Among sentence-initial conjunct participles, the split between the two configurations is instead more even.

\begin{table}[!h]
\centering
\begin{tabular}{ccc}
\hline
& \textbf{SV} & \textbf{VS}\\
\hline
\textbf{conjunct (sentence-initial)} & 45.7\% (208) & 54.3\% (247)\\
\textbf{absolute (sentence-initial)} & 21.9\% (27) & 78.1\% (96)\\
\hline
\end{tabular}
\caption{Position of overt subjects relative to sentence-initial conjunct participles and dative absolutes in the Codex Marianus}
\label{svversusvs}
\end{table}

\begin{figure}[!h]
\centering
\includegraphics[width=0.8\textwidth]{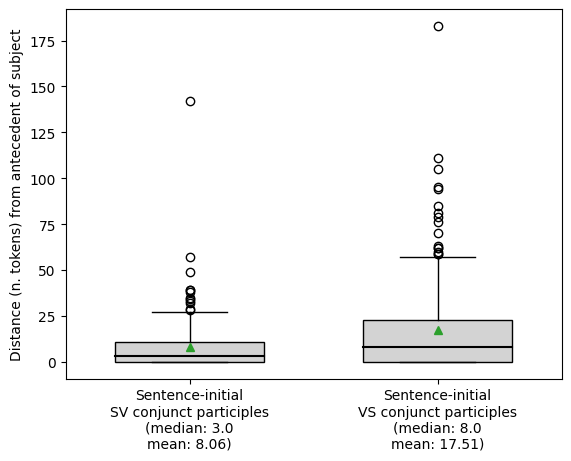}
\caption[Average anaphoric distance between the subject and immediate antecedent in sentence-initial SV and VS conjunct participle]{\label{distancevssvxadv}Average anaphoric distance between the subject and immediate antecedent in sentence-initial SV and VS conjunct participle. Test results: Welch’s $t$-test: -5.06, $p$-value $<0.01$; one-tailed Mann–Whitney $U$-test: 17526.5, $p$-value $<0.01$.}
\end{figure}

If the VS configuration is associated with the reinstatement of older referents (which framing participles help contextualize in the new discourse), the expectation is that, more than other functions, \textsc{frames} should involve subject referents whose immediately previous mention in the discourse is generally more distant in the preceding discourse. We can look at the anaphoric distance between the (overt or null) subject of a participle construction and its antecedent using the anaphoric-link annotation from nominal referents to their antecedents in the Greek New Testament in PROIEL. This level of annotation also includes links between null arguments, so that we can check when the last mention of a referent was even if either the anaphor or the antecedent is prodropped.\\
\indent As Figure \ref{distancevssvxadv} shows, sentence-initial conjunct participles with an overt subject following the participle itself have, on average, a significantly more distant antecedent (in number of tokens) than those whose subject precedes the participle.

\begin{figure}[!h]
\centering
\includegraphics[width=0.8\textwidth]{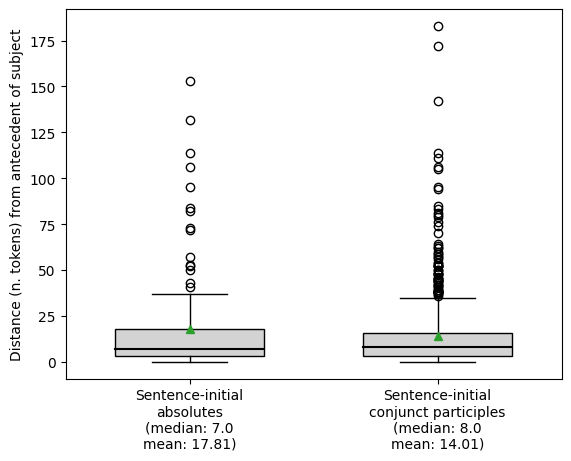}
\caption[Average anaphoric distance between the subject and immediate antecedent in sentence-initial dative absolutes and conjunct participle (null and overt)]{\label{distancedasxadv}Average anaphoric distance between the subject and immediate antecedent in sentence-initial dative absolutes and conjunct participle, including both null and overt subjects. Test results: Welch’s $t$-test: 1.44, $p$-value $=0.07$; one-tailed Mann–Whitney $U$-test: 42747, $p$-value $=0.49$.}
\end{figure}

This matches the observation that sentence-initial conjunct participles with their subject following the participle itself are more likely to function as \textsc{frames} than those with their subject preceding the participle, since the VS configuration is associated with the reinstatement of old or inactive referents.\\
\indent When we compare sentence-initial dative absolutes and conjunct participles, including both null- and overt-subject, we find that the difference is not significant (despite dative absolutes having a higher mean), as Figure \ref{distancedasxadv} shows. 

\begin{figure}[!h]
\centering
\includegraphics[width=0.8\textwidth]{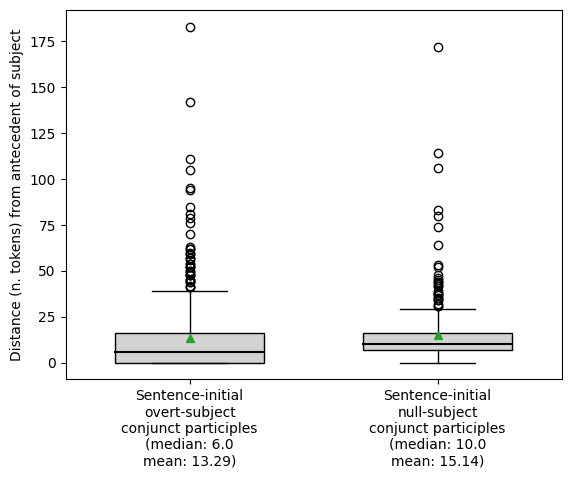}
\caption[Average anaphoric distance between the subject and immediate antecedent in overt-subject and null-subject conjunct participles in sentence-initial position]{\label{xadvovertnulldistance}Average anaphoric distance between the subject and immediate antecedent in overt-subject and null-subject conjunct participles in sentence-initial position. Test results: Welch’s $t$-test: -1.26, $p$-value $=0.2$; one-tailed Mann–Whitney $U$-test: 43138.5, $p$-value $<0.01$.}
\end{figure}

This result is somewhat surprising, since sentence-initial conjunct participles often have a null subject, whereas dative absolutes do not, and null subjects may be expected to occur, overall, with highly salient referents (e.g. `Jesus' in the New Testament) and therefore to be able to have longer and more distant anaphoric relations without incurring in ambiguity in the resolution of the anaphora. This is what we find if we compare the average distance of the antecedent of overt-subject and null-subject conjunct participles in sentence-initial position (Figure \ref{xadvovertnulldistance}), with null subjects having, on average, a more distant antecedent than overt subjects.

\begin{figure}[!h]
\centering
\includegraphics[width=0.8\textwidth]{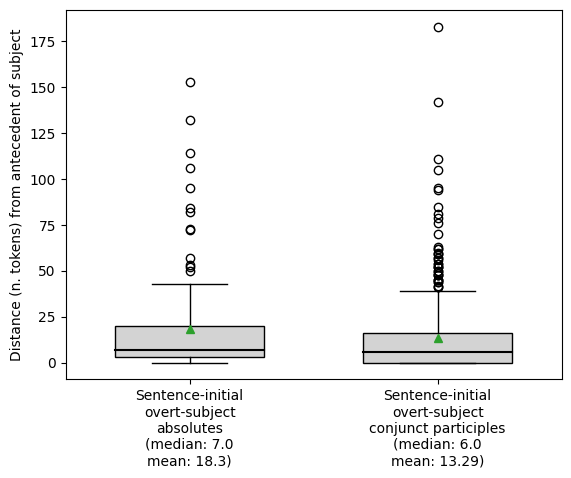}
\caption[Average anaphoric distance between the subject and immediate antecedent in sentence-initial dative absolutes and conjunct participle (overt subjects)]{\label{distancedasxadvovert}Average anaphoric distance between the subject and immediate antecedent in sentence-initial dative absolutes and conjunct participle, overt subjects only. Test results: Welch’s $t$-test: 1.8, $p$-value $=0.03$; one-tailed Mann–Whitney $U$-test: 28299.5, $p$-value $=0.01$.}
\end{figure}

This overall property of null subjects would be expected to skew the result `in favour' of the subjects of conjunct participles in Figure \ref{distancedasxadv}, but that was not the case. In itself, this suggests that the reinstatement of older referents as topics for a new discourse is a consistent and more evident function of dative absolutes, which, as we have seen, typically work as \textsc{frames}. As further confirmation, if we compare only overt subjects in sentence-initial dative absolutes and conjunct participles, the former have, on average, significantly more distant antecedents than the latter, as Figure \ref{distancedasxadvovert} shows.

\begin{figure}[!h]
\begin{subfigure}{0.50\textwidth}
\includegraphics[width=0.9\linewidth, height=6cm]{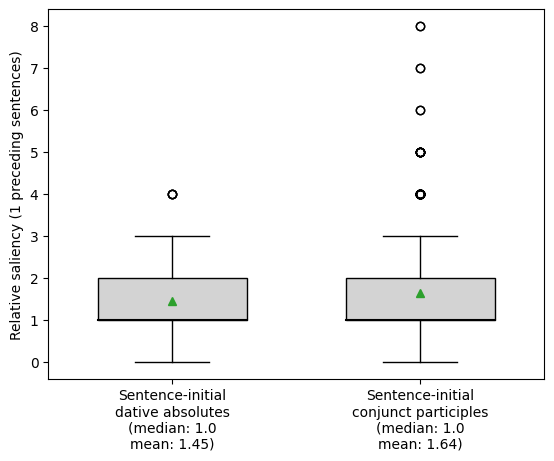} 
\caption[]{}
\label{sent-initi_xadv_das_sal1}
\end{subfigure}
\begin{subfigure}{0.50\textwidth}
\includegraphics[width=0.9\linewidth, height=6cm]{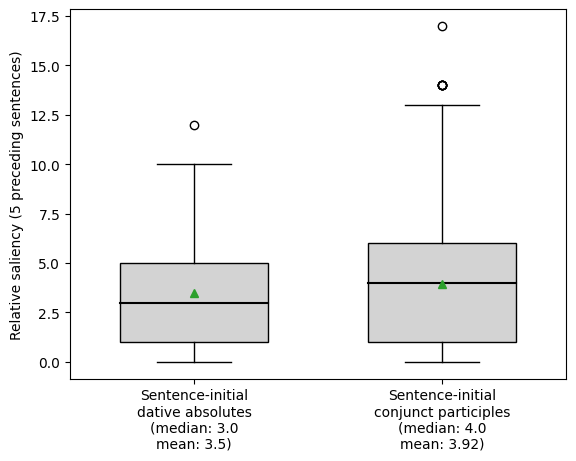}
\caption[]{}
\label{sent-initi_xadv_das_sal5}
\end{subfigure}
\begin{subfigure}{0.50\textwidth}
\includegraphics[width=0.9\linewidth, height=6cm]{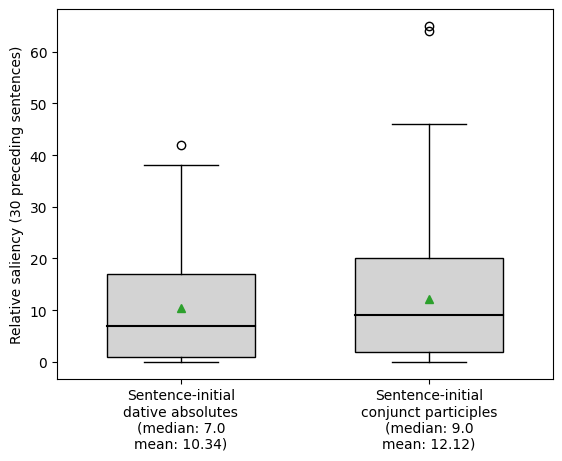}
\caption[]{}
\label{sent-initi_xadv_das_sal30}
\end{subfigure}
\begin{subfigure}{0.50\textwidth}
\includegraphics[width=0.9\linewidth, height=6cm]{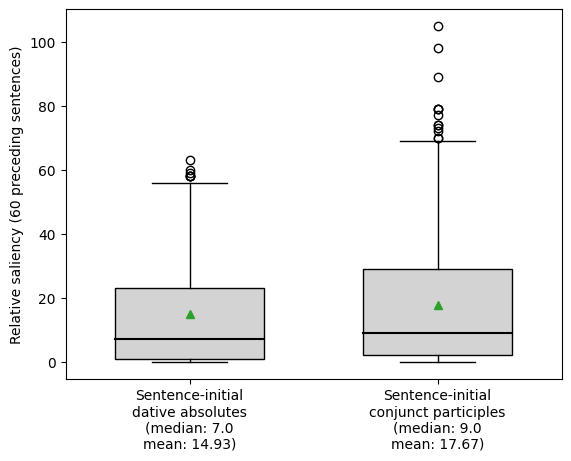}
\caption[]{}
\label{sent-initi_xadv_das_sal60}
\end{subfigure}
\caption[Pick-up rates of subject referents in the previous discourse: sentence-initial dative absolutes and conjunct participles (overt and null subjects)]{Pick-up rates of subject referents in the previous discourse: sentence-initial dative absolutes and conjunct participles (both overt and null subjects), based on the preceding 1, 5, 30, and 60 sentences. The green triangle represents the mean. Test results: (a) Welch’s $t$-test: -1.82, $p$-value $=0.07$; one-tailed Mann–Whitney $U$-test: 40503.5, $p$-value $=0.19$. (b) Welch’s $t$-test: -1.42, $p=0.15$; one-tailed Mann–Whitney $U$-test: 39996, $p=0.14$. (c) Welch’s $t$-test: -1.69, $p=0.09$; one-tailed Mann–Whitney $U$-test: 39996, $p=0.11$. (d) Welch’s $t$-test: -1.5, $p=0.13$; one-tailed Mann–Whitney $U$-test: 39910, $p=0.13$.}
\label{sent-initi_xadv_das_sal}
\end{figure}

\begin{figure}[!h]
\begin{subfigure}{0.50\textwidth}
\includegraphics[width=0.9\linewidth, height=6cm]{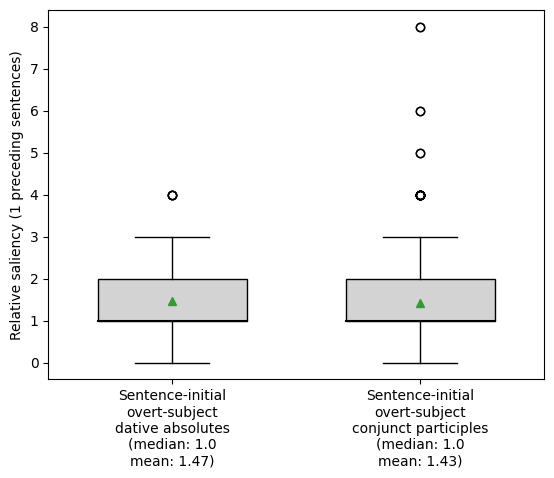} 
\caption[]{}
\label{overt_sent-initi_xadv_das_sal1}
\end{subfigure}
\begin{subfigure}{0.50\textwidth}
\includegraphics[width=0.9\linewidth, height=6cm]{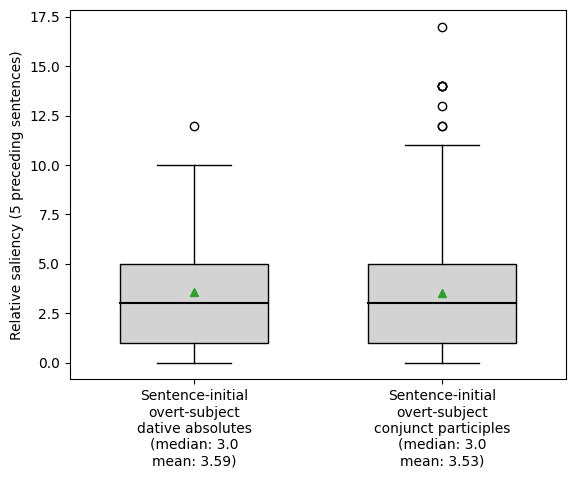}
\caption[]{}
\label{overt_sent-initi_xadv_das_sal5}
\end{subfigure}
\begin{subfigure}{0.50\textwidth}
\includegraphics[width=0.9\linewidth, height=6cm]{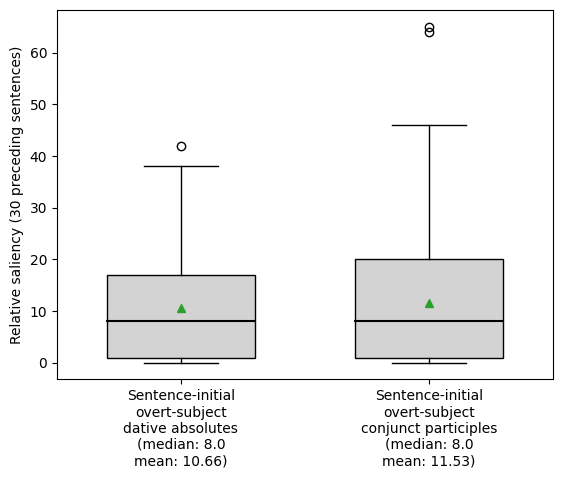}
\caption[]{}
\label{overt_sent-initi_xadv_das_sal30}
\end{subfigure}
\begin{subfigure}{0.50\textwidth}
\includegraphics[width=0.9\linewidth, height=6cm]{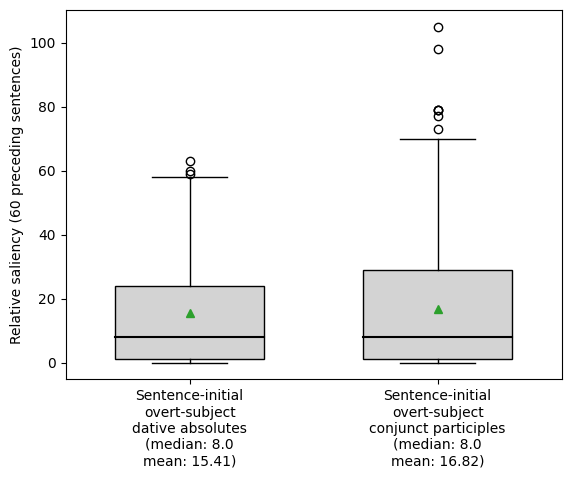}
\caption[]{}
\label{overt_sent-initi_xadv_das_sal60}
\end{subfigure}
\caption[Pick-up rates of subject referents in the previous discourse: sentence-initial dative absolutes and conjunct participles (overt subjects)]{Pick-up rates of subject referents in the previous discourse: sentence-initial dative absolutes and conjunct participles (overt subjects only), based on the preceding 1, 5, 30, and 60 sentences. The green triangle represents the mean. Test results: (a) Welch’s $t$-test: 0.33, $p$-value $=0.73$; one-tailed Mann–Whitney $U$-test: 26498, $p$-value $=0.42$. (b) Welch’s $t$-test: 0.18, $p=0.85$; one-tailed Mann–Whitney $U$-test: 25941.5, $p=0.68$. (c) Welch’s $t$-test: -0.77, $p=0.44$; one-tailed Mann–Whitney $U$-test: 25091, $p=0.87$. (d) Welch’s $t$-test: -0.72, $p=0.47$; one-tailed Mann–Whitney $U$-test: 25184, $p=0.92$.}
\label{overt_sent-initi_xadv_das_sal}
\end{figure}

Yet, if we look at the pick-up \textit{rates} of the subject referents of sentence-initial conjunct participles and dative absolutes in the previous discourse, we see that there is no significant difference between the two constructions, whether we consider both null and overt subjects or overt subjects only. Namely, without considering the \textit{distance} of previous sentences (in number of tokens), the two constructions are very similar when it comes to the number of times the referent of their subject is mentioned in the previous discourse. Figures \ref{sent-initi_xadv_das_sal1}-\ref{sent-initi_xadv_das_sal60} compare the average pick-up rate of the subject referent of sentence-initial dative absolutes and conjunct participles (both overt and null subjects) at a different number of preceding sentences (1, 5, 30 and 60). Figures \ref{overt_sent-initi_xadv_das_sal1}-\ref{overt_sent-initi_xadv_das_sal60} compare the average pick-up rate of the subject referent of sentence-initial dative absolutes and conjunct participles, but only considering overt subjects. 
 
Conjunct participles and dative absolutes, then, seem to be quite similar when it comes to the overall saliency of their subjects, considered only as the number of mentions in the previous discourse. When crossed with these results, however, the frequency of different parts of speech among their subjects suggests a somewhat more complex picture.

\subsection{Parts of speech}
As Table \ref{DASpostab} shows, most of the subjects of dative absolutes belong to categories typically encoding old or accessible referents.

\begin{table}[!h]
\centering
\begin{tabular}{cc}
\hline
\textbf{part of speech}        & \textbf{frequency} \\
\hline
\textbf{personal pronouns}  & 53.3\% (98) \\
\textbf{common nouns}    & 22.3\% (41) \\
\textbf{proper nouns}   & 5.9\% (11)  \\
\textbf{demonstrative pronouns}   & 2.2\% (4)   \\
\textbf{indefinite pronouns} & 2.2\% (4)   \\
\textit{null}               & 14.1\% (26) \\
\hline
\end{tabular}
\caption{Dative absolutes in the Codex Marianus: subject parts of speech}
\label{DASpostab}
\end{table}

Only indefinite pronouns and common nouns can potentially introduce new referents: these together represent only 24.5\% of all subjects, without, however, considering that several common nouns will also likely have old or accessible referents (e.g. \textit{narod\foreignlanguage{russian}{ъ}} `crowd, multitude', \textit{běs\foreignlanguage{russian}{ъ}} `demon', \textit{gospod\foreignlanguage{russian}{ь}} `Lord', \textit{sl\foreignlanguage{russian}{ъ}n\foreignlanguage{russian}{ь}ce} `sun'). 
\\ \indent As Table \ref{DASlemmastab} shows, as many as 51.1\% of all subjects are a third-personal pronoun \textit{*i}.\footnote{The asterisk (here and below) is due to the fact the third-person personal pronoun is not attested in Old Church Slavonic in the nominative singular. Instead, demonstratives are generally used in the nominative (\textit{t\foreignlanguage{russian}{ъ}} `that one', \textit{on\foreignlanguage{russian}{ъ}} `that one there' and \textit{s\foreignlanguage{russian}{ь}} `this one').}

\begin{table}[!h]
\centering
\begin{tabular}{ll}
\hline
\textbf{lemma}        & \textbf{frequency} \\
\hline
\textit{*i} ‘he, they’      & 51.1\% (94) \\
\textit{narod\foreignlanguage{russian}{ъ}} ‘crowd, multitude’    & 3.3\% (6)  \\
\textit{isus\foreignlanguage{russian}{ъ}} ‘Jesus’     & 2.7\% (5)  \\
\textit{sl\foreignlanguage{russian}{ъ}n\foreignlanguage{russian}{ь}ce} ‘sun’      & 2.2\% (4)  \\
\textit{v\foreignlanguage{russian}{ь}s\foreignlanguage{russian}{ь}} ‘all’         & 1.6\% (3)  \\
\textit{t\foreignlanguage{russian}{ъ}} ‘that (one)’    & 1.6\% (3)  \\
\textit{běs\foreignlanguage{russian}{ъ}} ‘demon’       & 1.1\% (2)  \\
\textit{gospod\foreignlanguage{russian}{ь}} ‘Lord’     & 1.1\% (2)  \\
\textit{ljudije} ‘people’       & 1.1\% (2)  \\
\textit{člověk\foreignlanguage{russian}{ъ}} ‘person’       & 1.1\% (2)  \\
\hline
\end{tabular}
\caption{Dative absolutes in the Codex Marianus: ten most-frequent subject lemmas}
\label{DASlemmastab}
\end{table}

Common nouns are instead the most frequent part of speech among the subjects of conjunct participles (Table \ref{xadvsubjpos}), followed not too far behind by proper nouns. 

\begin{table}[!h]
\centering
\begin{tabular}{cc}
\hline
\textbf{part of speech}        & \textbf{frequency} \\
\hline
\textbf{common nouns}  & 20.1\% (315) \\
\textbf{proper nouns}  & 17.2\% (270) \\
\textbf{demonstrative pronouns}  & 8.1\% (127) \\
\textbf{indefinite pronouns}  & 2.2\% (35) \\
\textbf{adjectives}  & 1.7\% (26) \\
\textbf{verbs}  & 1.7\% (26) \\
\textbf{personal pronouns}  & 1.4\% (22) \\
\textbf{cardinal numerals}  & 1.2\% (19) \\
\textbf{relative pronouns}  & 1\% (15) \\
\textbf{ordinal numerals}  & 0.4\% (6) \\
\textbf{interrogative pronouns}  & 0.2\% (4) \\
\textit{null}              & 44.8\% (703) \\
\hline
\end{tabular}
\caption{Conjunct participles in the Codex Marianus: subject parts of speech}
\label{xadvsubjpos}
\end{table}

Demonstratives are also very frequent among the subjects of conjunct participles. Note that, although there is no etymological form attested for the nominative third-person personal pronoun (*\textit{i}), demonstrative forms such as \textit{on\foreignlanguage{russian}{ъ}} `that one (there) and \textit{t\foreignlanguage{russian}{ъ}} `that one', particularly the former, are often used instead. Demonstrative pronouns and personal pronouns, then, should perhaps be counted together in Table \ref{xadvsubjpos} for a fair comparison with dative absolutes (where the third person personal pronoun occurs in the dative). 

\begin{table}[!h]
\centering
\begin{tabular}{ll}
\hline
\textbf{lemma}        & \textbf{frequency} \\
\hline
\textit{isus\foreignlanguage{russian}{ъ}} ‘Jesus’ &   12.1\%  (190) \\
\textit{on\foreignlanguage{russian}{ъ} }`that one (there)' & 6.8\% (106) \\
\textit{učenik\foreignlanguage{russian}{ъ}} `disciple' & 2.48\% (39) \\
\textit{jedin\foreignlanguage{russian}{ъ}} `someone' & 2.1\% (33) \\
\textit{žena} `woman, wife' & 1.9\% (30) \\
\textit{farisei} `Pharisee' & 1.46\% (23) \\
\textit{petr\foreignlanguage{russian}{ъ}} `Peter' & 1.08\% (17) \\
\textit{narod\foreignlanguage{russian}{ъ}} `crowd, multitude' & 1.08\% (17) \\
\textit{archierei} `bishop' & 1.02\% (16) \\
\textit{člověk\foreignlanguage{russian}{ъ}} `person' & 0.89\% (14) \\
\hline
\end{tabular}
\caption{Conjunct participles in the Codex Marianus: ten most-frequent subject lemmas}
\label{conjsubjlemmas}
\end{table}

After \textit{isus\foreignlanguage{russian}{ъ}} `Jesus', the demonstrative \textit{on\foreignlanguage{russian}{ъ}} `that one (there)' is the most common lemma among the subjects of conjunct participles. As Table \ref{mostcommoncommnouns} shows, among the top 10 common nouns occurring as the subject of conjunct participles, there are only human referents. Among the subjects of dative absolutes, it is instead relative frequent to find inanimate referents, such as \textit{sl\foreignlanguage{russian}{ъ}n\foreignlanguage{russian}{ь}ce} ‘sun’, \textit{pečal\foreignlanguage{russian}{ь}} `suffering, grief', \textit{čas\foreignlanguage{russian}{ъ}} `time, moment, hour', \textit{dv\foreignlanguage{russian}{ь}r\foreignlanguage{russian}{ь}} `door', and \textit{větr\foreignlanguage{russian}{ъ}} `wind'.

\begin{table}[!h]
\centering
\begin{tabular}{p{4cm}|p{7cm}}
\hline
\textbf{construction}   & \textbf{most frequent common nouns} \\
\hline
\textbf{dative absolutes}  & \textit{narod\foreignlanguage{russian}{ъ}} `crowd, multitude', \textit{sl\foreignlanguage{russian}{ъ}n\foreignlanguage{russian}{ь}ce} ‘sun’, \textit{gospod\foreignlanguage{russian}{ь}} ‘Lord’, \textit{d\foreignlanguage{russian}{ъ}šti} `daughter', \textit{ljudije} ‘people’, \textit{člověk\foreignlanguage{russian}{ъ}} ‘person’, \textit{čas\foreignlanguage{russian}{ъ}} `time, moment, hour', \textit{pečal\foreignlanguage{russian}{ь}} `suffering, grief', běs\foreignlanguage{russian}{ъ} `demon', \textit{dv\foreignlanguage{russian}{ь}r\foreignlanguage{russian}{ь}} `door', \textit{větr\foreignlanguage{russian}{ъ}} `wind'\\
\hline
\textbf{conjunct participles}  & \textit{učenik\foreignlanguage{russian}{ъ}} `disciple', \textit{žena} `woman, wife', \textit{farisei} `Pharisee', \textit{narod\foreignlanguage{russian}{ъ}} `crowd, multitude', \textit{archierei} `bishop', \textit{člověk\foreignlanguage{russian}{ъ}}, \textit{s\foreignlanguage{russian}{ъ}t\foreignlanguage{russian}{ь}nik\foreignlanguage{russian}{ъ}} `centurion', \textit{rab\foreignlanguage{russian}{ъ}} `servant, slave', \textit{angel\foreignlanguage{russian}{ъ}} `angel', \textit{cěsar\foreignlanguage{russian}{ь}} `emperor, king' \\
\hline
\end{tabular}
\caption{Dative absolutes and conjunct participles in the Codex Marianus: most frequent common nouns among subjects lemmas}
\label{mostcommoncommnouns}
\end{table}

As typically framing adverbials, this is not surprising, since inanimate referents are often part of the stage in which a new discourse is set (e.g. `when the door opened', `as the wind blow', `when the sun rose', etc.).\\
\indent If sentence-initial conjunct participles in the VS configuration are more likely to be interpreted as \textsc{frames} than those in the SV one, which may be more likely to function as \textsc{independent rhemes}, we may also expect some differences in the parts of speech most commonly occurring among the subjects in the two configurations. Tables \ref{vsxadvssubposdeep} and \ref{svxadvssubposdeep} show the frequency of different parts of speech among sentence-initial conjunct participles in the VS and in the SV configuration, respectively.

\begin{table}[!h]
\centering
\begin{tabular}{cc}
\hline
\textbf{part of speech}        & \textbf{frequency} \\
\hline
\textbf{proper nouns}  & 50.5\% (138) \\
\textbf{common nouns}  & 35.5\% (97) \\
\textbf{verbs}  & 4\% (11) \\
\textbf{demonstrative pronouns}  & 2.9\% (8) \\
\textbf{indefinite pronouns}  & 2.6\% (7) \\
\textbf{cardinal numerals}  & 2.2\% (6) \\
\textbf{personal pronouns}  & 0.7\% (2) \\
\textbf{adjectives}  & 0.7\% (2) \\
\textbf{relative pronouns}  & 0.4\% (1) \\
\textbf{ordinal numerals}  & 0.4\% (1) \\
\hline
\end{tabular}
\caption{Pre-matrix VS conjunct participles in the Codex Marianus: subject parts of speech}
\label{vsxadvssubposdeep}
\end{table}

\begin{table}[!h]
\centering
\begin{tabular}{cc}
\hline
\textbf{part of speech}        & \textbf{frequency} \\
\hline
\textbf{demonstrative pronouns}  & 30.1\% (90) \\
\textbf{common nouns}  & 28.4\% (85) \\
\textbf{proper nouns}  & 20.7\% (62) \\
\textbf{personal pronouns}  & 5\% (15) \\
\textbf{adjectives}  & 3.7\% (11) \\
\textbf{indefinite pronouns}  & 3.7\% (11) \\
\textbf{relative pronouns}  & 3.3\% (10) \\
\textbf{cardinal numerals}  & 2\% (6) \\
\textbf{interrogative pronouns}  & 1.3\% (4) \\
\textbf{verbs}  & 1\% (3) \\
\textbf{ordinal numerals}  & 0.7\% (2) \\
\hline
\end{tabular}
\caption{Pre-matrix SV conjunct participles in the Codex Marianus: subject parts of speech}
\label{svxadvssubposdeep}
\end{table}

We observe, in fact, neat differences between the two configurations, most evidently as far as the relative frequency of proper nouns and demonstrative pronouns is concerned. Demonstratives are much more frequent in the SV than in the VS configuration (90 occurrences as opposed to 8, respectively), whereas proper nouns are more common in the latter than in the former (138 in the VS as opposed to 62 in the SV configuration). Most of the personal pronouns are also found in the SV one (15, as opposed to 2 in the VS one), which seems to support what we just observed above, namely that there is a case to be made for demonstrative pronouns and personal pronouns to be counted together in the analysis. These figures, overall, support what we have already observed, namely that the VS configuration in conjunct participles is much more likely to indicate a framing function than the SV one, as \citet{haug2012a} had already argued regarding conjunct participles in Ancient Greek. In order to resolve the anaphora in a clause with a pronoun or a demonstrative, the antecedent should not, for ease of processing, be too far in the surrounding discourse from the anaphor itself, except with very salient referents. This is reflected both in the smaller average distance of the antecedent in sentence-initial conjunct participles, as we saw above, and in the inherently anaphoric parts of speech (demonstratives and personal pronouns), which are much more common in the SV than in the VS configuration. Among sentence-initial VS conjunct participles, which are more likely to be used as \textsc{frames}, on the other hand, proper nouns (which are naturally referentially more explicit than demonstratives and personal pronouns) are much more common than in the SV configuration. This also agrees with the observation that the average distance of the immediate antecedent of sentence-initial SV conjunct participles is significantly greater than that in the VS configuration. \\
\indent These observations, however, seem at odds with the fact that the subjects of dative absolutes in the Codex Marianus have, on average, similarly distant antecedents as sentence-initial conjunct participles but \textit{also} have inherently anaphoric parts of speech occurring overwhelmingly more frequently among their subjects than more explicit ones (e.g. proper nouns, as with VS conjunct participles). 

\begin{figure}[!h]
\centering
\includegraphics[width=0.8\textwidth]{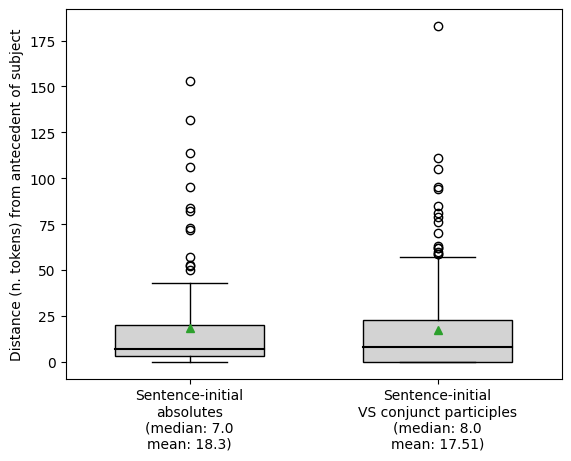}
\caption[Average anaphoric distance between subject and immediate antecedent in sentence-initial dative absolutes and VS conjunct participle (overt subjects only)]{\label{distance-vsxadv-das}Average anaphoric distance between subject and immediate antecedent in sentence-initial dative absolutes and VS conjunct participle, overt subjects only. Test results: Welch’s $t$-test: 0.25, $p$-value $=0.39$; one-tailed Mann–Whitney $U$-test: 14003.5, $p$-value $=0.46$.}
\end{figure}

We may expect the average distance to the antecedent of the subjects of dative absolutes to be at least shorter than those of sentence-initial VS conjunct participles, among which, as we have seen, referentially more explicit parts of speech are instead more common. As Figure \ref{distance-vsxadv-das} shows, however, that is not the case. Just like the other configurations (cf. Figures \ref{distancedasxadv} and \ref{distancedasxadvovert}), there is no significant difference between sentence-initial dative absolutes and sentence-initial VS conjunct participles in terms of average distance from their subject to its immediate antecedent.

This difference between the subjects of dative absolutes and the subjects of VS conjunct participles, which are otherwise functionally very similar to dative absolutes, suggests that dative absolutes occur overwhelmingly more frequently with highly discourse-prominent referents (in the sense of \citealt{himmelmann2015a} and \citealt{heusinger2019a}, among others), which a measure of saliency as mere number of previous mentions in the discourse may not be enough to capture. As we will see in more detail in the next chapter with a much larger and more diverse dataset, dative absolutes generally seem to attract fewer, but particularly prominent, referents than conjunct participles as a whole. In the Gospels, the majority of third-person singular personal pronouns are likely to refer to `Jesus' and the majority of third-person plurals to the `Apostles', which are highly salient and prominent referents throughout the New Testament. As \pgcitet{heusinger2019a}{123} argues, more prominent discourse referents ‘show a higher forward-looking potential in that they are referred to more often and they more frequently constitute the topic of the next utterance’, providing ‘a better accessibility for subsequent anaphoric expressions’, which may explain why, despite the longer and more distant anaphoric chains observed in the Gospels, inherently anaphoric parts of speech are overwhelmingly more frequent than others among the subjects of dative absolutes.

\subsection{Summary}
This section analyzed some of the properties of the subjects of conjunct participles and dative absolutes in sentence-initial position, which is where both constructions can function as \textsc{frames} and where conjunct participles can be ambiguous between a \textsc{frame} and an \textsc{independent rheme} interpretation. The vast majority of sentence-initial dative absolutes, as expected, were found to have an overt subject, whereas sentence-initial conjunct participles showed a more even split, although overt subjects were found to be somewhat more frequent (60.5\% overt subjects versus 39.5\% null subjects). \\
\indent Zooming in on overt-subject conjunct participles and dative absolutes, we saw that the position of the subject relative to the participles is predominantly post-verbal in dative absolutes, whereas in conjunct participles the SV and VS configurations are found with relatively even frequency (45.7\% SV and 54.3\% VS). \\
\indent Looking at the anaphoric properties of overt subjects in sentence-initial conjunct participles, we observed a significant association between the position of the subject relative to the participle and the average anaphoric distance of the subject to its immediate antecedent, whereby subjects in the VS configuration have, on average, significantly more distant antecedents than subjects in the SV one. This had already been observed by \citet{haug2012a} regarding Ancient Greek conjunct participles and can be associated with the reinstatement of old or inactive referents. This is a function that can be generally ascribed to \textsc{frames}, which are adjuncts in topicalized position and, as such, are more likely to be used to explicitly topicalize eventualities (and their participants) when there is a shift of some kind in the discourse (of subject, agent, place, time, etc.). \\
\indent A comparison between sentence-initial dative absolutes and sentence-initial conjunct participles indicated that the subjects of the two constructions have overall similarly distant antecedents and that they are similarly activated in the previous discourse. Relative \textit{saliency} was here simply calculated as the number of mentions (pick-up rates) of a subject referent at different sentence windows (1, 5, 30, and 60 preceding sentences). While no significant differences between the relative saliency of subjects in sentence-initial dative absolutes and conjunct participles were detected, the frequency of different parts of speech among their subjects suggested that there may be differences between their subjects which a simple measure of saliency is not able to capture. Among VS conjunct participles, which are more likely to work as \textsc{frames}, referentially more explicit parts of speech (e.g. proper nouns) are the most frequent. Because VS participle constructions help reinstate older referents, the predominance of such parts of speech among VS conjunct participles was explained with the fact that they may help facilitate anaphora resolution. However, the vast majority of subjects among dative absolutes are personal pronouns (i.e. inherently anaphoric parts of speech), despite also working typically as \textsc{frames} and having their subject following the participles. A possibility was advanced that absolute constructions generally attract few, but highly prominent subject referents (e.g. Jesus and the Apostles, as far as the New Testament is concerned). However, only a larger and more diverse dataset (Chapter 2) may verify whether this is the case.

\section{Lexical variation among participles}\label{deep-variation}
Lexical variation among participles in different syntactic configurations can also indirectly point to important functional differences. Table \ref{mostfreqlemmadeepptcp} shows the 10 most frequent lemmas among dative absolutes and conjunct participles.

\begin{longtable}{|p{3cm}|p{5cm}|p{5cm}|}
\hline
\textbf{subcorpus} & \textbf{absolutes} & \textbf{conjuncts}\\
\hline
\textbf{total}	&	\textit{byti} `be', \textit{glagolati} `speak, say', \textit{v\foreignlanguage{russian}{ъ}niti} `go in', \textit{iti} `go', \textit{is\foreignlanguage{russian}{ъ}choditi} `exit, go out', \textit{v\foreignlanguage{russian}{ъ}zležati} `lay down', \textit{jasti} `eat', \textit{priti} `come, arrive', \textit{iziti} `go down', \textit{iměti} `have'	&	\textit{viděti} `see', \textit{priti} `come, arrive', \textit{slyšati} `hear', \textit{iti} `go', \textit{pristǫpiti} `ascend, approach', \textit{iziti} `go down', \textit{prijęti} `take, receive', \textit{v\foreignlanguage{russian}{ъ}stati} `stand up', \textit{byti} `be', \textit{v\foreignlanguage{russian}{ъ}z\foreignlanguage{russian}{ь}rěti} `look up at' \\   
\hline
\textbf{left}	&	\textit{byti} `be', \textit{glagolati} `speak, say', \textit{v\foreignlanguage{russian}{ъ}niti} `go in', \textit{iti} `go', \textit{is\foreignlanguage{russian}{ъ}choditi} `exit, go out', \textit{v\foreignlanguage{russian}{ъ}zležati} `lay down', \textit{jasti} `eat', \textit{iziti} `go down', \textit{priti} `come, arrive', \textit{iměti} `have'	&	\textit{viděti} `see', \textit{priti} `come, arrive', \textit{slyšati} `hear', \textit{iti} `go', \textit{pristǫpiti} `ascend, approach', \textit{iziti} `go down', \textit{prijęti} `take, receive', \textit{v\foreignlanguage{russian}{ъ}stati} `stand up', \textit{v\foreignlanguage{russian}{ъ}z\foreignlanguage{russian}{ь}rěti} `look up at', \textit{byti} `be' \\ 
\hline
\textbf{right}	&	\textit{byti} `be', \textit{pristǫpiti} `ascend, approach', \textit{v\foreignlanguage{russian}{ъ}sijati} `begin to shine', \textit{s\foreignlanguage{russian}{ъ}pati} `sleep', \textit{vlasti} `rule', \textit{v\foreignlanguage{russian}{ъ}niti} `go in', \textit{zatvoriti} `close', \textit{v\foreignlanguage{russian}{ъ}zdati} `give (in homage)', \textit{pospěš\foreignlanguage{russian}{ь}stvovati} `assist, promote', \textit{utvr\foreignlanguage{russian}{ь}ždati} `establish, build'	&	\textit{učiti} `learn, teach', \textit{byti} `be', \textit{iměti} `have', \textit{iskušati} `tempt, try', \textit{viděti} `see', \textit{iskati} `seek', \textit{rešti} `say', \textit{věděti} `know', \textit{moliti sę} `pray', \textit{propovědati} `preach' \\ 
\hline
\textbf{SV}	&	\textit{glagolati} `speak, say', \textit{byti} `be', \textit{zatvoriti} `close', \textit{is\foreignlanguage{russian}{ъ}choditi} `exit, go out', \textit{četvr\foreignlanguage{russian}{ь}tovlast\foreignlanguage{russian}{ь}stvovati} `serve as a tetrarch', \textit{v\foreignlanguage{russian}{ъ}sijati} `begin to shine', \textit{čuditi} `marvel', \textit{plęsati} `dance', \textit{v\foreignlanguage{russian}{ъ}zležati} `lay down', \textit{ugoditi} `please, satisfy'	&	\textit{viděti} `see', \textit{slyšati} `hear', \textit{byti} `be', \textit{priti} `come, arrive', \textit{učiti} `learn, teach', \textit{iměti} `have', \textit{iti} `go', \textit{věděti} `know', \textit{prijęti} `take, receive', \textit{iskušati} `tempt, try' \\ 
\hline
\textbf{VS}	&	\textit{byti} `be', \textit{v\foreignlanguage{russian}{ъ}niti} `go in', \textit{glagolati} `speak, say', \textit{iti} `go', \textit{iziti} `go down', \textit{priti} `come, arrive', \textit{v\foreignlanguage{russian}{ъ}zležati} `lay down', \textit{is\foreignlanguage{russian}{ъ}choditi} `exit, go out', \textit{jasti} `eat', \textit{s\foreignlanguage{russian}{ъ}choditi} `go down'	&	\textit{viděti} `see', \textit{iti} `go', \textit{priti} `come, arrive', \textit{pristǫpiti} `ascend, approach', \textit{slyšati} `hear', \textit{iziti} `go down', \textit{v\foreignlanguage{russian}{ъ}stati} `stand up', \textit{prijęti} `take, receive', \textit{v\foreignlanguage{russian}{ъ}niti} `go in', \textit{v\foreignlanguage{russian}{ъ}z\foreignlanguage{russian}{ь}rěti} `look up at' \\
\hline
\caption{Dative absolutes and conjunct participles in the Codex Marianus: ten most frequent lemmas, overall, by position of the participle relative to the matrix clause, and by position of the subject relative to the participle}
\label{mostfreqlemmadeepptcp}
\end{longtable}

Motion verbs, such as \textit{priti} `come, arrive', \textit{v\foreignlanguage{russian}{ъ}niti} `go in', and simply \textit{iti} `go' are common among both conjunct participles and dative absolutes across all configurations. \textit{Byti} `be' is the top lemma among dative absolutes in virtually all configurations, except the SV one, which, however, is an infrequent configuration among absolutes in the first place. Among dative absolutes we find several verbs denoting activities, such as \textit{jasti} `eat', \textit{s\foreignlanguage{russian}{ъ}pati} `sleep', and \textit{plęsati} `dance', which one can easily envision as framing situations within which a main eventuality occurs, as in (\ref{activity1})-(\ref{activity2}).

\begin{example}
\begin{itemize}
\item[a.]
\gll \textbf{s\foreignlanguage{russian}{ъ}pęštem\foreignlanguage{russian}{ъ}} že člvkom\foreignlanguage{russian}{ъ}. pride vrag\foreignlanguage{russian}{ъ} ego. i v\foreignlanguage{russian}{ь}sě plěvel\foreignlanguage{russian}{ъ} po srědě pšenicę i otide.
{sleep.{\sc ptcp.ipfv.m.dat.sg}} {\sc ptc} {person.{\sc pl.m.dat}} {come.{\sc aor.3.sg}} {enemy.{\sc m.nom.sg}} {\sc 3.sg.m.gen} {and} {sow.{\sc aor.3.sg}} {weed.{\sc m.acc.sg}} {among} {wheat.{\sc sg.f.dat}} {and.{\sc sg.f.gen}} {and} {leave.{\sc aor.3.sg}}
\glt
\glend
\item[b.]
\gll {en} {de} {tōi} {\textbf{katheudein}} {tous} {anthrōpous} {ēlthen} {autou} {ho} {ekhthros} {kai} {epespeiren} {zizania} {ana} {meson} {tou} {sitou} {kai} {apēlthen}
{in} {\sc ptc} {the.{\sc n.dat.sg}} {sleep.{\sc prs.inf}} {the.{\sc pl.m.acc}} {man.{\sc pl.m.acc}} {come.{\sc aor.3.sg}} {\sc 3.sg.m.gen} {the.{\sc m.nom.sg}} {enemy.{\sc m.nom.sg}} {and} {sow.{\sc aor.3.sg}} {weed.{\sc pl.n.acc}} {in} {among.{\sc n.acc.sg}} {the.{\sc m.gen.sg}} {wheat.{\sc m.gen.sg}} {and} {depart.{\sc aor.3.sg}}
\glt `But while his men were sleeping, his enemy came and sowed weeds among the wheat, and left' (Matthew 13:25) %15221
\glend
\label{activity1}
\end{itemize}
\end{example}

\begin{example}
\begin{itemize}
\item[a.]
\gll \textbf{Ědǫštem\foreignlanguage{russian}{ъ}} že im\foreignlanguage{russian}{ъ} priim\foreignlanguage{russian}{ь} is chlěb\foreignlanguage{russian}{ъ} i blgsšt\foreignlanguage{russian}{ь} prělomi
{eat.{\sc ptpc.ipfv.m.dat}} {\sc ptc} {\sc 3.pl.m.dat} {take.{\sc ptpc.pfv.m.nom.sg}} {Jesus.{\sc m.nom.sg}} {bread.{\sc m.acc.sg}} {and} {bless.{\sc ptpc.pfv.m.nom.sg}} {break.{\sc aor.3.sg}}
\glt
\glend
\item[b.]
\gll {\textbf{Esthiontōn}} {de} {autōn} {labōn} {ho} {Iēsous} {arton} {kai} {eulogēsas} {eklasen}
{eat.{\sc ptpc.ipfv.m.gen}} {\sc ptc} {him.{\sc m.gen.pl}} {take.{\sc ptcp.pfv.m.nom.sg}} {the.{\sc m.nom.sg}} {Jesus.{\sc m.nom.sg}} {bread.{\sc m.acc.sg}} {and} {bless.{\sc ptcp.pfv.m.nom.sg}} {break.{\sc aor.3.sg}}
\glt `And as they were eating, Jesus took bread, blessed and broke it' (Matthew 26:26)
\glend
\label{activity2}
\end{itemize}
\end{example}

Among conjunct participles, besides the pool of frequent lemmas in common with dative absolutes, we find several telic verbs, such as \textit{v\foreignlanguage{russian}{ъ}zęti} `take', \textit{prijęti} ‘take, receive’,
\textit{slyšati} `hear' and \textit{v\foreignlanguage{russian}{ъ}z\foreignlanguage{russian}{ь}rěti} `look up at'. We can more easily envision these as part of serial events, namely as \textsc{independent rhemes}, as in (\ref{telic1})-(\ref{telic2}) (see also \textit{priim\foreignlanguage{russian}{ь}} in (\ref{activity2})).

\begin{example}
\begin{itemize}
\item[a.]
\gll t\foreignlanguage{russian}{ъ} \textbf{v\foreignlanguage{russian}{ъ}zem\foreignlanguage{russian}{ъ}} dažd\foreignlanguage{russian}{ъ} im\foreignlanguage{russian}{ъ} za mę i za sę.
{that.{\sc m.acc.sg}} {take.{\sc ptpc.pfv.m.nom.sg}} {give.{\sc 2.sg.prs.imp}} {him.{\sc m.dat.pl}} {for} {\sc 1.sg.acc} {and} {for} {self.{\sc acc.sg}}
\glt
\glend
\item[b.]
\gll {ekeinon} {\textbf{labōn}} {dos} {autois} {anti} {emou} {kai} {sou}
{that.{\sc m.acc.sg}} {take.{\sc ptcp.pfv.m.nom.sg}} {give.{\sc imp.pfv.pst.2.sg}} {\sc 3.pl.m.dat} {for} {\sc 1.sg.gen} {and} {yourself.{\sc gen.sg}}
\glt `Take that and give it to them for me and you' (Matthew 17:27)
\glend
\label{telic1}
\end{itemize}
\end{example}

\begin{example}
\begin{itemize}
\item[a.]
\gll is že \textbf{v\foreignlanguage{russian}{ъ}z\foreignlanguage{russian}{ь}rěv\foreignlanguage{russian}{ъ}} na n\foreignlanguage{russian}{ь} v\foreignlanguage{russian}{ъ}zljubi i
{Jesus.{\sc m.nom.sg}} {\sc ptc} {look at.{\sc ptpc.pfv.m.nom.sg}} {on} {\sc 3.sg.m.acc} {love.{\sc aor.3.sg}} {\sc 3.sg.m.acc}
\glt
\glend
\item[b.]
\gll {ho} {de} {Iēsous} {\textbf{emblepsas}} {autōi} {ēgapēsen} {auton}
{the.{\sc m.nom.sg}} {\sc ptc} {Jesus.{\sc m.nom.sg}} {look at.{\sc ptcp.pfv.m.nom.sg}} {\sc 3.sg.m.dat} {love.{\sc aor.3.sg}} {\sc 3.sg.m.acc}
\glt `Jesus looked at him and loved him' (Mark 10:21)
\glend
\label{telic2}
\end{itemize}
\end{example}

A large portion of the Greek New Testament in PROIEL includes annotation on the lexical aspect (Aktionsart) of verbs, including the categories \textsc{activity}, \textsc{semelfactive}, \textsc{state}, and \textsc{telic}. This level of annotation is experimental (\pgcitealt{proiel}{41}) and seems to broadly follow \posscitet{vendler} terminology and \posscitet{comrieaspect} categorization into three higher-level Aktionsarten, namely \textsc{telic}, \textsc{atelic} and \textsc{stative}, with the first two further divided into two subcategories. \textsc{Accomplishments} and \textsc{achievements} fall into the \textsc{telic} category, while \textsc{semelfactives} and \textsc{activities} fall into the \textsc{atelic} one. The PROIEL annotation thus only attempted to further distinguish \textsc{atelic} verbs and kept \textsc{accomplishments} and \textsc{achievements} subsumed under \textsc{telic}. \\
\indent We can cross the observations just made about the type of lemmas occurring among the constructions with the Aktionsart annotation in PROIEL on the Greek token corresponding to the participles. Since the main ambiguity is between pre-matrix constructions, we shall narrow down the extraction of tags to pre-matrix occurrences only. Tables \ref{aktionsartprexadv} and \ref{aktionsartpredas} show the Aktionsart tags corresponding to pre-matrix conjunct participles and pre-matrix dative absolutes, respectively.\footnote{Note that, despite the wide coverage, the annotation is not complete, so some occurrences do not figure here.}

\begin{table}[!h]
\centering
\begin{tabular}{cc|c|c|c|c|c|}
\cline{3-7}
 & & \multicolumn{4}{c|}{\textbf{matrix}} & \multirow{2}*{\textbf{\textit{tot.}}}\\
\cline{3-6}
  & & \textsc{act} & \textsc{sem}   & \textsc{st}  & \textsc{tel} & \\
\hline
\multicolumn{1}{|l|}{\multirow{4}{*}{\rotatebox[origin=c]{90}{\textbf{conjunct}}}} & \textsc{act} & 0 & 0 & 0 & 3.58\% (21) & 3.58\% (21) \\
\multicolumn{1}{|l|}{} & \textsc{sem} & 0 & 0 & 0 & 0.68\% (4) & 0.68\% (4)\\
\multicolumn{1}{|l|}{} & \textsc{st} & 0.51\% (3) & 0 & 1.7\% (10) & 12.61\% (74)& 14.82\% (87) \\
\multicolumn{1}{|l|}{}  & \textsc{tel} & 6.64\% (39) & 1.53\% (9) & 2.39\% (14) & 70.36\% (413) & 80.92\% (475)\\
\hline
\end{tabular}
\caption[Aktionsarten of pre-matrix conjunct participles and their matrix clause in the Codex Marianus]{Aktionsarten of pre-matrix conjunct participles and their matrix clause in the Codex Marianus. \textsc{act} = activity, \textsc{sem} = semelfactive, \textsc{st} = state, \textsc{tel} = telic.}
\label{aktionsartprexadv}
\end{table}

\begin{table}[!h]
\centering
\begin{tabular}{cc|c|c|c|c|c|}
\cline{3-7}
 & & \multicolumn{4}{c|}{\textbf{matrix}} & \multirow{2}*{\textbf{\textit{tot.}}}\\
\cline{3-6}
  & & \textsc{act} & \textsc{sem}   & \textsc{st}  & \textsc{tel} & \\
\hline
\multicolumn{1}{|l|}{\multirow{4}{*}{\rotatebox[origin=c]{90}{\textbf{absolute}}}} & \textsc{act} & 0 & 	0.58\% (1) & 	1.17\% (2) & 	18.71\% (32) & 20.47\% (35)\\
\multicolumn{1}{|l|}{} & \textsc{sem} & 0 &  0 &  0 & 	0.58\% (1) & 0.58\% (1)\\
\multicolumn{1}{|l|}{} & \textsc{st} & 0 &  0 &  0 & 	15.2\% (26) & 15.2\% (26)\\
\multicolumn{1}{|l|}{}  & \textsc{tel} & 5.26\% (9) & 	0 & 	5.26\% (9) & 	53.22\% (91) & 63.74\% (109)\\
\hline
\end{tabular}
\caption[Aktionsarten of pre-matrix dative absolutes and their matrix clause in the Codex Marianus]{Aktionsarten of pre-matrix dative absolutes and their matrix clause in the Codex Marianus. \textsc{act} = activity, \textsc{sem} = semelfactive, \textsc{st} = state, \textsc{tel} = telic.}
\label{aktionsartpredas}
\end{table}

Telic verbs are the most frequent among both constructions, although more prominently so among conjunct participles. As anticipated, what is most striking is the high frequency of activity verbs among dative absolutes. Their total raw frequency is, in fact, higher than the one of activity verbs among conjunct participles, despite the difference in overall frequency between the two constructions. \\
\indent As \pgcitet{haug2012a}{320} argues on the basis of Ancient Greek, \textsc{frames} are assumed to be more ‘predictable’, since they are ‘presupposed (ana\-phoric or accommodated) and not explicitly asserted’, so that we expect to find less lexical variation among sentence-initial participles than in other positions. \textsc{Elaborations} and \textsc{independent rhemes} are instead expected to present more variation as they typically introduce new information. Since the post-matrix position is likely to be predominantly occupied by \textsc{elaborations}, we can expect post-matrix conjunct participles to show a very high degree of lexical variation. Pre-matrix conjunct participles are instead often ambiguous between \textsc{frames} and \textsc{independent rhemes}, so we expect a lower degree of lexical variation than among post-matrix participles. To measure the overall lexical richness among participle lemmas in each participle construction, I use two metrics: the percentage of participle occurrences belonging to the 10 most frequent lemmas for a given configuration or subcorpus and a moving-average type-token ratio (MATTR). Standard type-token ratio (TTR) is a simple measure of lexical richness consisting of a text's vocabulary size divided by the text size and can be used to look at the proportion of total number of tokens and and unique lemmas. Given the difference in dataset sizes between conjunct participles and dative absolutes, I use a variation of TTR (MATTR; \citealt{mattr}) consisting of the average between the TTRs of a moving window of \textit{n} occurrences, so that the final score is not affected by sample size. MATTR is, of course, normally calculated over a window of adjacent tokens in running text. I instead treat subsequent occurrences of each participle construction as the adjacent tokens and calculate MATTR over the resulting list. For example, assuming a text has 10 occurrences of dative absolutes, we take the lemma of each occurrence and place them in an ordered list as follows:

\begin{verbatim}
[`byti', `byti', `priiti', `byti', `chotěti', 
`chotěti', `glagolati', `priiti', `byti', `chotěti']
\end{verbatim}

We then treat the ordered list as running text and calculate MATTR over the predefined window size. For example, if our window was of 2 tokens, we would calculate the TTR of the list above for the first two occurrences, then for the second and third occurrence, then for the third and fourth, and so on. We then average all the TTRs obtained in this way to get the MATTR score. The closer the MATTR to 1, the higher the degree of lexical richness, thus, in our case, variation among participle lemmas. A window of 40 occurrences was chosen after manually evaluating between windows of \verb|{20,40,60}| occurrences and observing that 40 was the smallest number of occurrences that allowed us to look at the lexical variation in most subsamples of the corpus without losing information about differences in variation. Too great a window size would result in not being able to look at variation in potentially meaningful subsamples (e.g. the SV configuration of dative absolutes) where the number of occurrences is below the window size, while too small a window size would result in not capturing variation at all.\\
\indent In Table \ref{lexvarptcpdeep}, the first column for each of the constructions indicates how many occurrences belong to the 10 most-frequent lemmas in that configuration (i.e. whether we are dealing with few high-frequency lemmas or several low-frequency lemmas), the second the MATTR score.\footnote{The numbers in Table \ref{lexvarptcpdeep} do not include \textit{glagolati} ‘speak, say’ (normally imperfective: \textit{gl}[\textit{agol}]\textit{ję}) for post-matrix conjunct participles and \textit{ot\foreignlanguage{russian}{ъ}věštati} ‘answer’ (normally per\-fective: \textit{ot\foreignlanguage{russian}{ъ}\-věštav\foreignlanguage{russian}{ъ}}) for pre-matrix conjunct participles, since these would likely skew the frequencies. As noted by \pgcitet{haug2012a}{288} on the respective Greek forms (\textit{legōn} `say' and \textit{apokritheis} `answer'), these work roughly like quotative particles, a usage sometimes considered a Semiticism (ibid.). \textit{Glagolati} 'talk, say' in absolute constructions should arguably not be considered on a par with the same lemma in conjunct participles, since in dative absolutes it is generally used as an intransitive verb (‘talking’ rather than ‘saying’).}

\begin{table}[!h]
\centering
\begin{tabular}{|c|c|c|c|c|}
\hline
\multirow{2}*{\textbf{Subsample}} & \multicolumn{2}{|c|}{\textbf{Absolutes}} & \multicolumn{2}{|c|}{\textbf{Conjuncts}}\\
\cline{2-5}
 & \textbf{10MFL} & \textbf{MATTR} & \textbf{10MFL}   & \textbf{MATTR}\\
\hline
\textbf{total}	&	50.54\%	&	0.64	&	36.79\%	&	0.72\\ 
\hline
\textbf{SV}	&	\textit{NA}	&	\textit{NA}	& 47.34 & 0.70\\ 
\textbf{VS} & \textbf{45.83} & \textbf{0.64} &\textbf{ 64.71} & \textbf{0.50}\\ 
\hline
\textbf{left}	&	51.46\%	&	0.64	&	44.12\%	&	0.67\\ 
\textbf{right}	&	\textit{NA}	&	\textit{NA}	&	30.11\%	&	0.81\\
\hline
\end{tabular}
\caption[Lexical variation among participles, overall, by position relative to the matrix clause, and by position of the subject relative to the participle]{Lexical variation among participles, overall, by position relative to the matrix clause, and by position of the subject relative to the participle. \textit{10MFL} = 10 most frequent lemmas; \textit{MATTR} = moving-average type-token ratio; \textit{left} and \textit{right} = pre-matrix and post matrix, respectively. Note that the closer the MATTR is to 1 (and the lower the percentage for the 10MFL), the greater the lexical variation.}
\label{lexvarptcpdeep}
\end{table}

As before, post-matrix dative absolutes, as well as sentence-initial SV occurrences, are too few to make a comparison, which will have to await the next, much larger dataset. However, as predicted, dative absolutes show less total lexical variation than conjunct participles, except for the VS configuration, where it is lower among conjunct participles. This also supports the observation that VS conjunct participles are more likely to function as \textsc{frames} than SV ones, which are much more likely \textsc{independent rhemes}. Much like in Greek (see the data in \citealt{haug2012a}), Old Church Slavonic conjunct participles to the right of the main verb show an overall higher degree of lexical variation than those to the left, as the difference in MATTR and 10MFL in the table indicate. Only a much larger dataset, as the one used in the next chapter, however, may confirm whether this is, in fact, a general pattern.

\section{Greek-Old Church Slavonic mismatches}\label{greekocscompa}
It is hard, from a philological standpoint, to judge the extent to which the different participle clauses should be considered `native' Slavonic constructions. The evident pervasiveness of the phenomena intuitively suggests that both dative absolutes and conjunct participles may have, at most, expanded their scope under the influence of Greek, rather than to represent mechanical translation devices. This has already been claimed in the literature \citep[e.g.][]{birnbaum1958a, r1958a, macrobert1986, corin1995a} and finds, in fact, a certain degree of evidence in the treebank data. Old Church Slavonic does follow the Greek to a large degree, translating most genitive absolutes into dative absolutes and most conjunct participles into conjunct participles. However, it also displays limited but consistent signs of independence in the use of the constructions as either \textsc{frames}, \textsc{independent rhemes} or \textsc{elaborations}.

\subsection{Conjunct participles}
Tables \ref{xadvmismatches_1} and \ref{xadvmismatches_2} show the number of matches and mismatches between conjunct participles in Old Church Slavonic and Greek.

\begin{table}[!h]
\centering
\begin{tabular}{ll}
\hline
\textbf{Conjunct participle}          & 1424\\
\textbf{Finite main clause}          & 46 \\
\textbf{Dative absolute}         & 4   \\
\textbf{Finite subordinate}         & 2   \\
\textbf{Infinitive}         &   2   \\
\textbf{Other/NA}         &   33   \\
\hline
\end{tabular}
\caption{Old Church Slavonic constructions corresponding to a conjunct participle in Ancient Greek}
\label{xadvmismatches_1}
\end{table}

\begin{table}[!h]
\centering
\begin{tabular}{ll}
\hline
\textbf{Conjunct participle}          & 1424\\
\textbf{Finite main clause}          & 34 \\
\textbf{Genitive absolute}         & 1   \\
\textbf{Other/NA}         &   112   \\
\hline
\end{tabular}
\caption{Ancient Greek constructions corresponding to a conjunct participle in Old Church Slavonic}
\label{xadvmismatches_2}
\end{table}

Given the much greater frequency of conjunct participles compared to dative absolutes, and since a close-reading comparison between each occurrence of the construction in the Codex Marianus and Tischendorf's New Testament is beyond the scope of this chapter, I will highlight the most interesting mismatches between the two languages.\\
\indent We can identify two main groups of mismatches which more clearly show the independence of Old Church Slavonic (i.e. clear signs of productivity regardless of Greek) in the usage of conjunct participles as \textsc{independent rhemes} and \textsc{elaborations}. The first concerns the translation of Greek finite main verbs into conjunct participles. Of the 34 occurrences (once again, according to the PROIEL version), 24 correspond to pre-matrix, perfective conjunct participles, clearly functioning as \textsc{independent rhemes}. 10 correspond instead to post-matrix, imperfective conjunct participles, most of which are, however, examples of \textit{glę} `saying' following a main verb such as \textit{reče} `said' (where the participle functions roughly as a quotative particle; cf footnote 10), whereas Greek only has a finite verb, such as \textit{efē} `said'. \textsc{Independent rhemes} are particularly evident when Old Church Slavonic translates two Greek imperatives into a participle-imperative sequence as in (\ref{xadvimperative}).

\begin{example}
\begin{itemize}
\item[a.]
\gll \textbf{v\foreignlanguage{russian}{ъ}stav\foreignlanguage{russian}{ъ}} chodi
rise.\textsc{ptcp.pfv.m.nom.sg} walk.\textsc{imp.2.sg}
\glt
\glend
\item[b.]
\gll \textbf{egeire} kai peripatei
rise.\textsc{prs.imp.2.sg} and walk.\textsc{prs.imp.2.sg}
\glt ‘Rise and walk’ (Matthew 9:5)
\glend
\label{xadvimperative}
\end{itemize}
\end{example}

The second group of interesting mismatches are those in which Old Church Slavonic employs expressions involving a participle to freely render a Greek construction. These are listed under `Other/NA' in Table \ref{xadvmismatches_2}, and they are mostly clear examples of \textsc{elaborations}, as in (\ref{fixedelab}).

\begin{example}
\begin{itemize}
\item[a.]
\gll t\foreignlanguage{russian}{ъ}gda reče im\foreignlanguage{russian}{ъ} is \textbf{ne} \textbf{obinuję} \textbf{sę}
then say.\textsc{aor.3.sg} \textsc{3.pl.dat} Jesus.\textsc{nom} \textsc{neg} conceal.\textsc{ptcp.ipfv.m.nom.sg} \textsc{refl} 
\glt
\glend
\item[b.]
\gll tote oun eipen autois ho Iēsous \textbf{parrēsiai}
then therefore say.\textsc{aor.3.sg} \textsc{3.pl.dat} the Jesus.\textsc{nom} plainly
\glt ‘Then said Jesus unto them plainly' (John 11:14)
\glend
\label{fixedelab}
\end{itemize}
\end{example}

\subsection{Absolute constructions}

\begin{table}[!h]
\centering
\begin{tabular}{ll}
\hline
\textbf{Dative absolute}          & 153 \\
\textbf{Finite subordinate}         & 3  \\
\textbf{Conjunct participle}                       & 3   \\
\textbf{Finite main clause}         & 2  \\
\textbf{Prepositional phrase}         & 1  \\
\hline
\end{tabular}
\caption{Old Church Slavonic constructions corresponding to a genitive absolute in Ancient Greek}
\label{grcgaocsnoda}
\end{table}

Table \ref{grcgaocsnoda} shows which constructions correspond to a genitive absolute in the Greek New Testament version in PROIEL.\footnote{The text in PROIEL, based on Tischendorf's eighth edition of the Greek New Testament, is not the closest to the Old Church Slavonic translation. The Byzantine text-type (or `Majority text'), on the other hand, seems to be one of the closest (several apparent mismatches have an exact parallel in it). Also note that the number of genitive absolutes is only taken from verses in the Greek New Testament which have a parallel in the Codex Marianus. The latter has gaps in Matthew 5, Mark 16, Luke 2, Luke 24, John 1-2, John 18, and John 20.}\\ 
\indent 10 genitive absolutes do not appear to be translated into a dative absolute. Of these, the two main clauses in Old Church Slavonic (Luke 23.45, Matthew 17.26) are contradicted by versions of the Greek New Testament other than Tischendorf's also having a main clause, whereas one of the conjunct participles (John 21.11) has the dubious mixed-case form \textit{tolikou sǫšt\foreignlanguage{russian}{ь}} ‘(there) being so many’, which however corresponds to a regular dative absolute in the \textit{Codex Zographensis} (\textit{tolikou sǫštju}).\\
\indent Of the potentially genuine discrepancies, three are Greek genitive absolutes corresponding to Old Church Slavonic finite \textit{egda}-clauses (\ref{ga-egda1}-\ref{ga-egda2}), two of which occur in the same sentence; one is a Greek absolute rendered with a prepositional phrase in Old Church Slavonic, perhaps due to the difficulty in finding an exact match to Greek \textit{mesoō} ‘to be in the middle’ (\ref{ga-pp}):

\begin{example}
\begin{itemize}
\item[a.]
\gll i \textbf{egda} \textbf{pride} v\foreignlanguage{russian}{ъ} crkv\foreignlanguage{russian}{ъ} pristǫpišę k\foreignlanguage{russian}{ъ} nemu učęštju archierei i star\foreignlanguage{russian}{ь}ci ljud\foreignlanguage{russian}{ь}stii glšte
and when come.\textsc{aor.3.sg} in temple.\textsc{acc} approach.\textsc{aor.3.pl} towards \textsc{3.sg.dat} teach.\textsc{ptcp.ipfv.m.dat.sg} chief.priest.\textsc{nom.pl} and elder.\textsc{nom.pl} people.\textsc{adj.nom.pl} say.\textsc{ptcp.ipfv.m.nom.pl}
\glt
\glend
\item[b.]
\gll Kai \textbf{elthontos} autou eis to hieron prosēlthon autōi didaskonti hoi archiereis kai hoi presbyteroi tou laou legontes
and come.\textsc{ptcp.aor.m.gen.sg} he.\textsc{gen} in the temple.\textsc{acc} approach.\textsc{aor.3.pl} \textsc{3.sg.dat} teach.\textsc{ptcp.ipfv.m.dat.sg} the chief.priest.\textsc{nom.pl} and the elder.\textsc{nom.pl} the.\textsc{gen.sg} people.\textsc{gen.sg} say.\textsc{ptcp.ipfv.m.nom.pl}
\glt ‘And when he came into the temple, the chief priests and the elders of the people confronted him as he was teaching, saying' (Matthew 21:23)
\glend
\label{ga-egda1}
\end{itemize}
\end{example}

\begin{example}
\begin{itemize}
\item[a.]
\gll da ne \textbf{egda} \textbf{položit\foreignlanguage{russian}{ъ}} osnovaniě. i ne možet\foreignlanguage{russian}{ъ} s\foreignlanguage{russian}{ъ}vr\foreignlanguage{russian}{ъ}šiti. v\foreignlanguage{russian}{ь}si vidęštei nač\foreignlanguage{russian}{ъ}nǫt\foreignlanguage{russian}{ъ} rǫgati sę emu ěko
so-that \textsc{neg} when lay.\textsc{prs.3.sg} foundation.\textsc{acc} and \textsc{neg} be-able.\textsc{prs.3.sg} finish.\textsc{inf} all.\textsc{nom.pl} see.\textsc{ptcp.ipfv.nom.pl} begin.\textsc{prs.3.pl} mock.\textsc{inf} \textsc{refl} he.\textsc{dat} that
\glt
\glend
\item[b.]
\gll hina mēpote \textbf{thentos} autou themelion kai mē ischyontos ektelesai pantes hoi heōrountes arxōntai autōi empaizein legontes hoti
thus not-ever lay.\textsc{ptcp.aor.m.gen.sg} he.\textsc{gen.sg} foundation.\textsc{acc} and \textsc{neg} be-able.\textsc{ptcp.ipfv.m.gen.sg} finish.\textsc{aor.inf} all.\textsc{nom.pl} the see.\textsc{ptcp.ipfv.m.nom.pl} begin.\textsc{aor.sbjv.mid} he.\textsc{dat} mock.\textsc{prs.inf} say.\textsc{ptcp.ipfv.nom.pl} that
\glt ‘… lest, after he has laid the foundation, and is not able to finish, all who see it begin to mock him’ (Luke 14:29-30)		
\glend
\label{ga-egda2}
\end{itemize}
\end{example}

\begin{example}
\begin{itemize}
\item[a.]
\gll Abie že \textbf{v\foreignlanguage{russian}{ъ}} \textbf{prěpolovlenie} prazd\foreignlanguage{russian}{ь}nika v\foreignlanguage{russian}{ь}zide is v\foreignlanguage{russian}{ъ} crk\foreignlanguage{russian}{ъ} i učaaše
now \textsc{ptc} in middle.\textsc{acc} feast.\textsc{gen} go-up.\textsc{aor.3.sg} Jesus.\textsc{nom} in temple.\textsc{acc} and teach.\textsc{impf.3.sg}
\glt
\glend
\item[b.]
\gll Ēdē de tēs heortēs \textbf{mesousēs} anebē Iēsous eis to hieron kai edidaske
now \textsc{ptc} the feast.\textsc{gen} be-in-middle.\textsc{ptcp.ipfv.f.gen.sg} go-up.\textsc{aor.3.sg} Jesus.\textsc{nom} in the temple.\textsc{acc} and teach.\textsc{impf.3.sg}
\glt ‘Now about the midst of the feast Jesus went up into the temple, and taught’ 
(John 7:14)
\label{ga-pp}
\glend
\end{itemize}
\end{example}

Two coordinated genitive absolutes correspond to two conjunct participles in the dative, agreeing with an oblique argument of the matrix (\ref{xadvdat}).

\begin{example}
\begin{itemize}
\item[a.]
\gll i vy podob\foreignlanguage{russian}{ь}ni člkom\foreignlanguage{russian}{ь} čajǫštem\foreignlanguage{russian}{ь} ga svoego kogda v\foreignlanguage{russian}{ъ}zvratit\foreignlanguage{russian}{ъ} sę ot\foreignlanguage{russian}{ъ} brak\foreignlanguage{russian}{ъ}. da \textbf{prišed\foreignlanguage{russian}{ъ}šju} i \textbf{tl\foreignlanguage{russian}{ъ}knǫv\foreignlanguage{russian}{ъ}šju} abie otvr\foreignlanguage{russian}{ъ}zǫt\foreignlanguage{russian}{ъ} sę emu
{and} {\sc 2.pl.nom} {similar.{\sc m.nom.pl}} {man.{\sc dat.pl}} {wait.{\sc ptcp.ipfv.dat.pl}} {master.{\sc m.gen.sg}} {own.{\sc m.gen.sg}} {when} {come back.{\sc prs.3.sg}} {\sc refl} {from} {wedding feast.{\sc gen.pl}} {so that} {come.{\sc ptcp.pfv.m.dat.sg}} {and} {knock.{\sc ptcp.pfv.m.dat.sg}} {immediately} {open.{\sc prs.3.pl}} {\sc refl} {\sc 3.sg.m.dat}
\glt
\glend
\item[b.]
\gll {kai} {humeis} {homoioi} {anthrōpois} {prosdekhomenois} {ton} {kurion} {heautōn} {pote} {analusēi} {ek} {tōn} {gamōn} {hina} {\textbf{elthontos}} {kai} {\textbf{krousantos}} {eutheōs} {anoiksōsin} {autōi}
{and} {\sc 2.pl.nom} {same.{\sc m.nom.pl}} {man.{\sc dat.pl}} {receive.{\sc ptcp.ipfv.dat.pl}} {the.{\sc m.acc.sg}} {master.{\sc m.acc.sg}} {self.{\sc 3.pl.m.gen}} {when} {return.{\sc sbjv.aor.3.sg}} {from} {the.{\sc m.gen.pl}} {wedding feast.{\sc m.gen.pl}} {so that} {come.{\sc ptcp.pfv.m.gen.sg}} {and} {strike.{\sc ptcp.pfv.m.gen.sg}} {immediately} {open.{\sc sbjv.aor.3.pl}} {\sc 3.sg.m.dat}
\glt `and be like men who are waiting for their master to come home from the wedding feast, so that they may open the door to him at once \textit{when he comes and knocks} [lit. to him coming and knocking].' (Luke 12:36) %40777 
\label{xadvdat}
\glend
\end{itemize}
\end{example}

More interesting are the mismatches in the opposite direction:

\begin{table}[!h]
\centering
\begin{tabular}{ll}
\hline
\textbf{Genitive absolute}          & 153 \\
\textbf{Accusative with infinitive}                        & 16  \\
\textbf{`Agreeing absolutes'} & 9   \\
\textbf{Finite subordinate}         & 2   \\
\textbf{Conjunct participle}                       & 1   \\
\textbf{Finite main clause}         & 1  \\
\textit{Other}/\textit{NA}         & 2  \\
\hline
\end{tabular}
\caption{Ancient Greek constructions corresponding to a dative absolute in Old Church Slavonic}
\label{danoga}
\end{table}

One dative absolute (Mark 1.42) does not have a Greek parallel at all in Tischendorf's edition (other versions however have a genitive absolute), whereas one (Matthew 28.1) corresponds to a Greek participle in the dative, which can be read (in both languages) as an impersonal temporal expression (Old Church Slavonic \textit{svitajǫšti}, Greek \textit{tēi epiphōskousēi}, `being dawn').
\\ \indent As many as 16 dative absolutes render Greek accusatives with infinitive. Most accusatives with infinitive translated into dative absolutes are found in nominalized \textit{en tōi} + infinitive constructions (14 occurrences), as exemplified in (\ref{aciabs}). The majority of these occur in \textit{egeneto}-clauses (9 examples), the discourse structure of which was discussed in Section \ref{bystsec}, arguing that the temporal adverbial clause regularly following \textit{egeneto}/\textit{byst\foreignlanguage{russian}{ъ}} (and similar formulas in other languages) can be generally considered as a framing expression introducing a narrative turn.

\begin{example}
\begin{itemize}
\item[a.]
\gll i uboěšę že sę. \textbf{v\foreignlanguage{russian}{ъ}šed\foreignlanguage{russian}{ъ}šema} ima oněma v\foreignlanguage{russian}{ъ} oblak\foreignlanguage{russian}{ъ}
and fear.\textsc{aor.3.pl} \textsc{ptc} \textsc{refl} enter.\textsc{ptcp.pfv.dat.du} \textsc{3.du.dat} that.\textsc{dat.du} in cloud.\textsc{acc.sg}
\glt
\glend
\item[b.]
\gll	ephobēthēsan de \textbf{en} \textbf{tōi} \textbf{eiselthein} autous eis tēn nephelēn
fear.\textsc{aor.3.pl} \textsc{ptc} in the.\textsc{n.dat.sg} enter.\textsc{aor.inf} \textsc{3.pl.acc} in the.\textsc{f.acc.sg} cloud.\textsc{f.acc.sg}
\glt ‘And they were afraid, as they entered the cloud’ (Luke 9:34)
\glend
\label{aciabs}
\end{itemize}
\end{example}

Of the remaining accusatives with infinitive, two are subject complement clauses of \textit{egeneto}/\textit{ginetai} (\ref{aciabs2}-\ref{aciabs3}), which are rendered as \textit{byst\foreignlanguage{russian}{ъ}}-clauses syntactically and functionally equivalent to the ones translating the \textit{en tōi} + accusative with infinitive construction. 

\begin{example}
\begin{itemize}
\item[a.]
\gll i byst\foreignlanguage{russian}{ъ} \textbf{v\foreignlanguage{russian}{ъ}zležęštju} emu v\foreignlanguage{russian}{ъ} domu ego. i m\foreignlanguage{russian}{ъ}nodzi mytare i grěš\foreignlanguage{russian}{ъ}nici v\foreignlanguage{russian}{ъ}zležachǫ s\foreignlanguage{russian}{ъ}	ism\foreignlanguage{russian}{ь} i s\foreignlanguage{russian}{ъ} učeniky ego	
and happen.\textsc{aor.3.sg} dine.\textsc{ptcp.ipfv.m.dat.sg} \textsc{3.sg.dat} in house.\textsc{loc.sg} \textsc{3.sg.gen} and many.\textsc{nom.pl} tax.collector.\textsc{nom.pl} and sinner.\textsc{nom.pl} dine.\textsc{impf.3.pl} with Jesus.\textsc{ins} and with disciple.\textsc{ins.pl} \textsc{3.sg.gen}
\glt
\glend
\item[b.]
\gll kai ginetai \textbf{katakeisthai} auton en tēi oikia autou, kai polloi telōnai kai hamartōloi synanekeinto tōi Iēsou kai tois mathētais autou
and happen.\textsc{prs.3.sg} dine.\textsc{prs.inf} \textsc{3.sg.acc=} in the house.\textsc{dat.sg} \textsc{3.sg.gen} and many tax.collector.\textsc{nom.pl} and sinner.\textsc{nom.pl} dine.\textsc{impf.3.pl} the.\textsc{dat.sg} Jesus.\textsc{dat} and the.\textsc{dat.pl} disciple.\textsc{dat.pl} \textsc{3.sg.gen}
\glt ‘And it came to pass, as he was dining in his house, that many tax collectors and sinners also sat together with Jesus and his disciples’ (Mark 2:15)
\glend
\label{aciabs2}
\end{itemize}
\end{example}

\begin{example}
\begin{itemize}
\item[a.]
\gll i byst\foreignlanguage{russian}{ъ} mimo \textbf{chodęštju} emu isu. v\foreignlanguage{russian}{ъ} soboty skvozě sěniě. i načęšę učenici ego pǫt\foreignlanguage{russian}{ь} tvoriti v\foreignlanguage{russian}{ъ}str\foreignlanguage{russian}{ъ}gajǫšte klasy
and happen.\textsc{aor.3.sg} close-by walk.\textsc{ptcp.ipfv.m.dat.sg} \textsc{3.sg.dat} Jesus.\textsc{dat} on Sabbath through cornfield.\textsc{gen.pl} and start.\textsc{aor.3.pl} disciple.\textsc{nom.pl} \textsc{3.sg.gen} way.\textsc{acc.sg} make.\textsc{inf} pluck.\textsc{ptcp.ipfv.nom.pl} corn.\textsc{acc.pl}
\glt
\glend
\item[b.]
\gll Kai egeneto auton en tois sabbasin \textbf{paraporeuesthai} dia tōn sporimōn, kai hoi mathētai autou ērxanto hodon poiein tillontes tous stachuas
and happen.\textsc{aor.3.sg} \textsc{3.sg.acc} in the.\textsc{dat.pl} Sabbath.\textsc{dat.pl} walk.\textsc{prs.inf} through the.\textsc{gen.pl} cornfield.\textsc{gen.pl} and the.\textsc{nom.pl} disciple.\textsc{nom.pl} \textsc{3.sg.gen} start.\textsc{aor.3.pl} way.\textsc{acc.sg} make.\textsc{prs.inf} pluck.\textsc{ptcp.ipfv.nom.pl} the.\textsc{acc.pl} corn.\textsc{acc.pl}
\glt ‘And it came to pass that he went through the cornfields on the Sabbath day and his disciples began to pluck the ears of corn as they went’ (Mark 2:23)
\glend
\label{aciabs3}
\end{itemize}
\end{example}

One dative absolute translates a nominalized \textit{dia to} + accusative with infinitive ‘because of the \textit{x}-ing of \textit{y}’ (where \textit{x} is the infinitive and \textit{y} the agent), whose translation into Old Church Slavonic is coincidentally also one of the very few examples of potential null-subject dative absolutes (\ref{nullda5}). 

\begin{example}
\begin{itemize}
\item[a.]
\gll se izide sějęi da sěet\foreignlanguage{russian}{ъ}. i \textbf{sějǫštumu}. ova ubo padǫ pri pǫti. 
behold go.out.\textsc{aor.3.sg} sow.\textsc{ptcp.ipfv.m.nom.sg} to sow.\textsc{sup} and sow.\textsc{ptcp.ipfv.m.dat.sg} some.\textsc{nom.pl} then fall.\textsc{aor.3.pl} along way.\textsc{loc.sg}
\glt
\glend
\item[b.]
\gll idou exēlthen ho speirōn tou speirein. kai \textbf{en} \textbf{tōi} \textbf{speirein} auton ha men epesen para tēn hodon
behold go.out.\textsc{aor.3.sg} the sow.\textsc{ptcp.ipfv.m.nom.sg} the.\textsc{n.gen.sg} sow.\textsc{prs.inf} and in the.\textsc{n.dat.sg} sow.\textsc{prs.inf} \textsc{3.sg.acc} some.\textsc{n.nom.pl} indeed fall.\textsc{aor.3.pl} along the.\textsc{f.acc.sg} road.\textsc{f.acc.sg}
\glt ‘Behold, a sower went out to sow. And as he sowed, some seed fell by the wayside; and the birds came and devoured them’ (Matthew 13:4)
\glend
\label{nullda5}
\end{itemize}
\end{example}

The choice of dative absolute in this example may also have to do with the clause-bridging function of the adverbial in this sentence, a function which, as we will see in the next chapters (when looking at standard Early Slavic treebanks), is typical of dative absolutes outside of the Codex Marianus. The bridging function is clear from the repetition of the predicate in the immediately preceding sentence (`sow'), with a change from foreground (a sower [went out to sow]$_{foreground}$) to background for a new foregrounded clause ([as he sowed]$_{background}$ [...]$_{foreground}$), involving a shift in perspective or setting, in this case a change in event participants of the foregrounded event. Outside the Codex Marianus, as we will see, it is not uncommon to see dative absolutes used to bridge two discourse segments where shifts of different kinds occur (e.g. shift in location or timeframe).\footnote{Old Church Slavonic adjectives (attributive and no\-mi\-na\-lized) and participles are inflected in either the `long' or the `short' form (sometimes referred to as `weak' and `strong' respectively, from the Germanic tradition). It is interesting that the dative absolute in this example appears in the `long form', which is notably rare among dative absolutes (\pgcitealt{lindberg2013a}{37}). The difference is generally one of definiteness: old or infer\-able nominal referents are marked on the adjective through the long form, whereas adjectives whose referent cannot be retrieved via context or world knowledge normally appear in the short form. In (\ref{nullda5}), the expected short form would have been *\textit{sějǫštu}. The choice of the long form here is likely, in fact, to be connected with the pro-dropping of the subject.}\\
\indent 9 dative absolutes correspond to a pre-matrix participle construction with an agreeing subject in the dative, as in (\ref{backcontrda1}-\ref{backcontrda2}), or accusative, as in (\ref{backcontrda3}) (`agreeing absolutes' in Table \ref{danoga}). In PROIEL, these are analysed as absolute constructions, even though the clause agrees in case with an argument of the matrix clause, because the argument is found inside the participle clause and is then repeated in the matrix clause. Of the dative absolutes translating this type of participle construction, one (Matthew 14.6) matches a genitive absolute in New Testament versions other than Tischendorf's. The remaining 8 examples do not stand out with respect to the typical properties of dative absolutes observed so far: all except one (\ref{backcontrda2}) involve a verb of movement, very common among dative absolutes; they all have a third-person pronoun (6 examples) or \textit{Jesus} (1 example) as their subject, which always follows the participle; they are all sentence-initial. They can thus be identified as typical \textsc{frames}, bridging two events by means of motion from one scene to another.

\begin{example}
\begin{itemize}
\item[a.]
\gll poslěd\foreignlanguage{russian}{ь} že \textbf{v\foreignlanguage{russian}{ь}zležęštem\foreignlanguage{russian}{ъ}} im\foreignlanguage{russian}{ъ}. edinuemu-na-desęte ěvi sę
after \textsc{ptc} sit.\textsc{ptcp.ipfv.dat.pl} \textsc{3.pl.dat}  eleven.\textsc{dat} appear.\textsc{aor.3.sg}  \textsc{refl}
\glt
\glend
\item[b.]
\gll hysteron \textbf{anakeimenois} autois tois hendeka ephanerōthē
after sit.\textsc{ptcp.ipfv.dat.pl} \textsc{3.pl.dat} the.\textsc{dat.pl} eleven.\textsc{dat.pl} appear.\textsc{aor.3.sg}
\glt ‘Afterward he appeared unto the eleven as they sat at meat’ (Mark 16:14)
\glend
\label{backcontrda1}
\end{itemize}
\end{example}

\begin{example}
\begin{itemize}
\item[a.]
\gll i \textbf{v\foreignlanguage{russian}{ь}lěz\foreignlanguage{russian}{ъ}šu} emu v\foreignlanguage{russian}{ъ} korab\foreignlanguage{russian}{ь}. po nem\foreignlanguage{russian}{ь} idǫ učenici ego
and enter.\textsc{ptcp.pfv.m.dat.sg} \textsc{3.sg.dat} in boat.\textsc{acc.sg} after \textsc{3.sg.loc} go.\textsc{aor.3.pl} disciple.\textsc{nom.pl} \textsc{3.sg.gen}
\glt
\glend
\item[b.]
\gll kai \textbf{embanti} autōi eis to ploion ēkolouthēsan autōi hoi mathētai autou
and go.\textsc{ptcp.ipfv.dat.sg} \textsc{3.sg.dat} in the boat.\textsc{acc.sg} follow.\textsc{aor.3.pl} \textsc{3.sg.dat} the disciple.\textsc{nom.pl} \textsc{3.sg.gen}
\glt ‘When he got into a boat, his disciples followed him’ (Matthew 8:23)
\glend
\label{backcontrda2}
\end{itemize}
\end{example}

\begin{example}
\begin{itemize}
\item[a.]
\gll \textbf{iš\foreignlanguage{russian}{ь}d\foreignlanguage{russian}{ъ}šu} že emu v\foreignlanguage{russian}{ъ} vrata. uz\foreignlanguage{russian}{ь}rě drugaě. i gla im\foreignlanguage{russian}{ъ}
go.out.\textsc{ptcp.pfv.m.dat.sg} \textsc{ptc} \textsc{3.sg.dat} in porch.\textsc{acc.pl} see.\textsc{aor.3.sg} another.\textsc{f.nom.sg} and say.\textsc{aor.3.sg} \textsc{3.sg.dat}
\glt
\glend
\item[b.]
\gll \textbf{exelthonta} de auton eis ton pulōna, eiden auton allē kai legei tois ekei
go.out.\textsc{ptcp.pfv.m.acc.sg} \textsc{ptc} \textsc{3.sg.acc} in the porch.\textsc{acc.sg} see.\textsc{aor.3.sg} \textsc{3.sg.acc} another.\textsc{f.nom.sg} and say.\textsc{aor.3.sg} that.\textsc{dat.pl} there
\glt ‘When he had gone out into the porch, another [woman] saw him and told them' (Matthew 26:71)
\glend
\label{backcontrda3}
\end{itemize}
\end{example}

Of the remaining dative absolutes from Table \ref{danoga}, one of the occurrences seemingly translating a conjunct participle is possibly scribal error (\ref{nullda1}), since both Greek and the \textit{Codex Zographensis} have the `expected' conjunct participle (\textit{legontes} and \textit{gl}[\textit{agol}]\textit{jǫšte}).

\begin{example}
\begin{itemize}
\item[a.]
\gll v\foreignlanguage{russian}{ъ}prosišę že i \textbf{gl}[\textbf{agol}]\textbf{jǫštju}. učitelju. kogda ubo si bǫdǫt\foreignlanguage{russian}{ъ}.
ask.\textsc{impf.3.pl} \textsc{ptc} \textsc{3.sg.acc} say.\textsc{ptcp.ipfv.m.dat.sg} master.\textsc{voc.sg} when then this.\textsc{nom.pl} be.\textsc{fut.3.pl}
\glt
\glend
\item[b.]
\gll epērōtēsan de auton \textbf{legontes}· didaskale, pote oun tauta estai;
ask.\textsc{impf.3.pl} \textsc{ptc} \textsc{3.sg.acc} say.\textsc{ptcp.ipfv.nom.pl} master.\textsc{voc.sg} when then this.\textsc{nom.pl} be.\textsc{fut.3.pl}
\glt ‘And they asked him, saying, Master, but when shall these things be?’ 
(Luke 21:7)
\glend
\label{nullda1}
\end{itemize}
\end{example}

However, note that this specific usage of dative absolutes, namely as introducers of direct speech, is relatively common outside of the Codex Marianus, regardless of referential identity between the subjects of the matrix and of the participle. As we will see in the final chapter, it is, in fact, a rather common usage of switch-reference markers cross-linguistically.\\
\indent The remaining apparent mismatches (two corresponding to finite temporal subordinates, Mark 4.6 and John 2.3, and one to a finite main clause, Mark 6.21) are contradicted by genitive absolutes in New Testament editions other than Tischendorf’s.

% \begin{table}[!h]
% \centering
% \begin{tabular}{ll}
% \hline
% \textbf{Conjunct participle}          & 1464 + 5 +27\\
% \textbf{Finite main clause}          & 46 \\
% \textbf{Dative absolute}         & 4   \\
% \textbf{Finite relative clause}          & 2 \\
% \textbf{Finite subordinate}         & 2   \\
% \textbf{Infinitive}         &   2   \\
% \textbf{Prepositional phrase}         &   2   \\
% \hline
% \end{tabular}
% \caption{Old Church Slavonic constructions corresponding to a conjunct participle in Greek}
% \label{xadvmismatches_1}
% \end{table}
% 1435

% Comment on DAs --> what type of XADV does it translate (same subj? backward control?)

% \begin{table}[!h]
% \centering
% \begin{tabular}{ll}
% \hline
% \textbf{Conjunct participle}          & 1464 + 5 +27\\
% \textbf{Finite main clause}          & 34 \\
% \textbf{Prepositional phrase}          & 6 \\
% \textbf{Finite relative clause}          & 2 \\
% \textbf{Genitive absolute}          & 1 \\
% \textbf{Adjective}          & 1 \\
% \textit{Other}/\textit{NA}         & 71  \\
% \hline
% \end{tabular}
% \caption{Greek constructions corresponding to a conjunct participle in Old Church Slavonic}
% \label{xadvmismatches_2}
% \end{table}

% other      71
% pred       31
% atr        20
% adv         9
% xobj        6
% apos        5
% sub         2
% voc         1
% obl         1
% parpred     1

\section{Remarks on the discourse-relation annotation on the Gospel of Luke}\label{rhetrelsec}
Another experimental layer of annotation in the PROIEL treebanks regards discourse relations combining three main frameworks for the annotation of rhetorical relations, namely the Penn Discourse Treebank 2.0 (\citealt{prasad2008penn}), the RST Discourse Corpus (\citealt{carlson2001discourse}), and the DISCOR project (\citealt{reeseetal}).\footnote{Details on this experiment were kindly provided by Hanne Eckhoff in a personal communication. The annotation is available in the official releases of the Greek New Testament in the PROIEL treebanks.} The annotation is currently available for the whole Gospel of Luke from the Greek New Testament. A rhetorical relation holds between two discourse units, identified as those roughly corresponding to any verbal clause except those with non-predicative dependency relations such as subject, object, and nominal arguments. As in \citet{prasad2008penn}, the PROIEL annotation has a hierarchical structure, with five higher-level relations, each having two possible further levels of granularity describing the type of discourse relation holding between any two given discourse units. \\
\indent Table \ref{xadvrhetrel} and \ref{dasrhetrel} shows the frequency of discourse relations introduced by conjunct participles and dative absolutes.

\begin{table}[!h]
\centering
\begin{tabular}{lll}
\hline
\textbf{discourse relation tag} & \textit{\textbf{n}} & \textbf{\%}\\
\hline
TEMPORAL\textunderscore Narration\textunderscore same story & 133 & 31.29 \\ 
TEMPORAL\textunderscore Synchronous\textunderscore circumstance & 71 & 16.70 \\ 
TEMPORAL\textunderscore Asynchronous\textunderscore precedence & 67 & 15.76 \\ 
TEMPORAL\textunderscore Coincidence\textunderscore means & 44 & 10.35 \\ 
TEMPORAL\textunderscore Continuation\textunderscore  & 16 & 3.76 \\ 
TEMPORAL\textunderscore Synchronous\textunderscore overlap & 16 & 3.76 \\ 
CONTINGENCY\textunderscore Cause\textunderscore reason & 14 & 3.29 \\ 
TEMPORAL\textunderscore Coincidence\textunderscore manner & 10 & 2.35 \\ 
ATTRIBUTION\textunderscore Content\textunderscore speech content & 8 & 1.88 \\ 
COMPARISON\textunderscore Concession\textunderscore  & 7 & 1.64 \\ 
EXPANSION\textunderscore Elaboration\textunderscore characterization & 6 & 1.41 \\ 
CONTINGENCY\textunderscore Condition\textunderscore hypothetical & 5 & 1.17 \\ 
COMPARISON\textunderscore Juxtaposition\textunderscore contrast & 5 & 1.17 \\ 
TEMPORAL\textunderscore Narration\textunderscore new story & 5 & 1.17 \\ 
CONTINGENCY\textunderscore Purpose\textunderscore  & 3 & 0.70 \\ 
CONTINGENCY\textunderscore Result\textunderscore real result & 3 & 0.70 \\ 
CONTIGENCY\textunderscore Cause\textunderscore reason & 2 & 0.47 \\ 
COMPARISON\textunderscore Juxtaposition\textunderscore parallel & 2 & 0.47 \\ 
EXPANSION\textunderscore Restatement\textunderscore specification & 2 & 0.47 \\ 
ATTRIBUTION\textunderscore Content\textunderscore  & 1 & 0.23 \\ 
ATTRIBUTION\textunderscore Content\textunderscore Non-verbal content & 1 & 0.23 \\ 
CONTINGENCY\textunderscore Condition\textunderscore general & 1 & 0.23 \\ 
EXPANSION\textunderscore Alternative\textunderscore chosen alternative & 1 & 0.23 \\ 
CONTINGENCY\textunderscore Result\textunderscore pragmatic result & 1 & 0.23 \\ 
COMPARISON\textunderscore Sameness\textunderscore  & 1 & 0.23 \\
\hline
\end{tabular}
\caption{Frequency of discourse relations tags among conjunct participles in the Codex Marianus}
\label{xadvrhetrel}
\end{table}

\begin{table}[!h]
\centering
\begin{tabular}{lll}
\hline
\textbf{discourse relation tag} & \textit{\textbf{n}} & \textbf{\%}\\
\hline
TEM\-PO\-RAL\textunderscore Synchronous\textunderscore overlap & 33 & 55.93 \\
TEM\-PO\-RAL\textunderscore Asynchronous\textunderscore precedence & 15 & 25.42 \\
TEM\-PO\-RAL\textunderscore Continuation\textunderscore  & 5 & 8.47 \\
CONTINGENCY\textunderscore Cause\textunderscore reason & 2 & 3.38 \\
TEM\-PO\-RAL\textunderscore Synchronous\textunderscore circumstance & 1 & 1.69 \\
TEM\-PO\-RAL\textunderscore Narration\textunderscore new story & 1 & 1.69 \\
TEM\-PO\-RAL\textunderscore Narration\textunderscore same story & 1 & 1.69 \\
TEM\-PO\-RAL\textunderscore Coincidence\textunderscore means & 1 & 1.69 \\
\hline
\end{tabular}
\caption{Frequency of discourse relations tags among dative absolutes in the Codex Marianus}
\label{dasrhetrel}
\end{table}

We can immediately see that the top discourse relation introduced by conjunct participles in the Gospel of Luke is \textit{TEM\-PO\-RAL\textunderscore Nar\-ration\textunderscore same story}, which most typically holds between two main clauses belonging to the same line of narration or subject matter, but also in cases such as (\ref{tempnarrsamesto}). The next most frequent relation is \textit{TEM\-PO\-RAL\textunderscore Syn\-chro\-nous\textunderscore cir\-cum\-stance}, which is introduced by clauses such as \textit{rubbing} in \textit{plucked the heads of grain and ate them, rubbing them} in (\ref{elabascomplexrheme}) above, repeated here as (\ref{elabascomplexrheme-repeated}).

\begin{example}
\begin{itemize}
\item[a.]
\gll on\foreignlanguage{russian}{ъ} že \textbf{prošed\foreignlanguage{russian}{ъ}} po srědě ich\foreignlanguage{russian}{ъ} iděaše
{that.{\sc m.nom.sg}} {\sc ptc} {walk through.{\sc ptpc.pfv.m.nom.sg}} {in} {middle.{\sc dat}} {\sc 3.pl.m.gen} {go.{\sc impf.3.sg}}
\glt
\glend
\item[b.]
\gll {autos} {de} {\textbf{dielthōn}} {dia} {mesou} {autōn} {eporeueto}
{\sc 3.sg.m.nom} {\sc ptc} {go through.{\sc ptcp.pfv.m.nom.sg}} {in} {middle.{\sc n.gen.sg}} {\sc 3.pl.m.gen} {go.{\sc impf.mid.3.sg}}
\glt `But he walked right through the crowd and went on his way' (Luke 4:30) %20353
\glend
\label{tempnarrsamesto}
\end{itemize}
\end{example}

\begin{example}
\begin{itemize}
\item[a.]
\gll i v\foreignlanguage{russian}{ъ}str\foreignlanguage{russian}{ъ}zaachǫ učenici ego klasy. i ěděachǫ \textbf{istirajǫšte} rǫkama
and pluck.\textsc{impf.3.pl} disciple.\textsc{nom.pl} \textsc{3.sg.gen} head.of.grain.\textsc{acc.pl} and eat.\textsc{impf.3.pl} rub.\textsc{ptcp.ipfv.nom.pl} hand.\textsc{ins.du}
\glt
\glend
\item[b.]
\gll kai etillon hoi mathētai autou tous stachyas kai ēsthion \textbf{psōchontes} tais chersin
and pluck.\textsc{impf.3.pl} the disciple.\textsc{nom.pl} \textsc{3.sg.gen} the.\textsc{acc.pl} head.of.grain.\textsc{acc.pl} and eat.\textsc{impf.3.pl} rub.\textsc{ptcp.ipfv.nom.pl} the.\textsc{f.dat.pl} hand.\textsc{f.dat.pl}
\glt ‘And his disciples plucked the heads of grain and ate them, rubbing them in their hands’ (Luke 6:1)
\glend
\label{elabascomplexrheme-repeated}
\end{itemize}
\end{example}

These most clearly correspond to \textsc{elaboration} participles, whereas the former are most likely \textsc{independent rhemes}.\\
\indent Focussing on the tags occurring more than ten times, next among the discourse relations introduced by conjunct participles are \textit{TEM\-PO\-RAL\textunderscore A\-syn\-chro\-nous-precedence}, \textit{TEM\-PO\-RAL-Co\-in\-ci\-dence\textunderscore means}, \textit{TEM\-PO\-RAL\textunderscore Con\-ti\-nu\-a\-tion\textunderscore }, \textit{TEM\-PO\-RAL\textunderscore Syn\-chron\-ous\-\textunderscore o\-ver\-lap}, \textit{CON\-TIN\-GEN\-CY\textunderscore Cause\textunderscore rea\-son}, \textit{TEM\-PO\-RAL\textunderscore Co\-in\-ci\-dence\textunderscore man\-ner}.  \\
\indent \textit{TEM\-PO\-RAL\textunderscore A\-syn\-chro\-nous\textunderscore pre\-ce\-dence} and \textit{TEM\-PO\-RAL\textunderscore Syn\-chro\-nous\textunderscore overlap} can most likely be expected to be introduced by framing expressions, the former when referring back to some time before the event time of the situation being described by the current topic time (i.e. roughly equivalent to those introduced by `after'-clauses or `when'-clauses), the latter when the time of the framing eventuality overlaps (and generally \textit{includes}) with that of the event time of the main eventuality (i.e. roughly equivalent to those introduced by `while'-clauses), as in (\ref{tempasynchpre}) and (\ref{tempsynchover}), respectively.

\begin{example}
\begin{itemize}
\item[a.]
\gll i \textbf{v\foreignlanguage{russian}{ъ}zvrašt\foreignlanguage{russian}{ь}še} sę apli povědašę emou eliko s\foreignlanguage{russian}{ъ}tvorišę
{and} {return.{\sc ptcp.pfv.m.nom.pl}} {\sc refl} {apostle.{\sc pl.m.nom}} {tell.{\sc aor.3.pl}} {\sc 3.sg.m.dat} {everything} {make.{\sc aor.3.pl}}
\glt 
\glend
\item[b.]
\gll {kai} {\textbf{hupostrepsantes}} {hoi} {apostoloi} {diēgēsanto} {autōi} {hosa} {epoiēsan}
{and} {turn back.{\sc ptcp.pfv.m.nom.pl}} {the.{\sc pl.m.nom}} {apostle.{\sc pl.m.nom}} {tell.{\sc aor.3.pl.mid}} {\sc 3.sg.m.dat} {everything.{\sc n.acc.pl}} {make.{\sc aor.3.pl}}
\glt `And the apostles, when they had returned, told Him all that they had done' (Luke 9:10)
\glend
\label{tempasynchpre}
\end{itemize}
\end{example}

\begin{example}
\begin{itemize}
\item[a.]
\gll \textbf{s\foreignlanguage{russian}{ъ}vr\foreignlanguage{russian}{ъ}šen\foreignlanguage{russian}{ъ}} že v\foreignlanguage{russian}{ъ}sěk\foreignlanguage{russian}{ъ} bǫdet\foreignlanguage{russian}{ъ} ěkože oučitel\foreignlanguage{russian}{ь} ego
{prepare.{\sc ptcp.pfv.pass.m.nom.sg}} {\sc ptc} {anyone.{\sc m.nom.sg}} {be.{\sc fut.3.sg}} {like} {teacher.{\sc m.nom.sg}} {\sc 3.sg.m.gen}
\glt
\glend
\item[b.]
\gll {\textbf{katērtismenos}} {de} {pas} {estai} {hōs} {ho} {didaskalos} {autou}
{prepare.{\sc ptcp.prf.pass.m.nom.sg}} {\sc ptc} {everyone.{\sc m.nom.sg}} {be.{\sc fut.3.sg.mid}} {like} {the.{\sc m.nom.sg}} {teacher.{\sc m.nom.sg}} {\sc 3.sg.m.gen}
\glt `But anyone, when they have been fully trained, will be like their teacher' (Luke 6:40)
\glend 
\label{tempsynchover}
\end{itemize}
\end{example}            

\textit{TEM\-PO\-RAL\textunderscore Co\-in\-ci\-dence\textunderscore means} and \textit{TEM\-PO\-RAL\textunderscore Co\-in\-ci\-dence\textunderscore man\-ner} are other two typical functions of \textsc{elaborations}, which can often be read as manner or means adverbs, as in (\ref{meansmannerelab}).

\begin{example}
\begin{itemize}
\item[a.]
\gll mariě že s\foreignlanguage{russian}{ъ}bljudaaše v\foreignlanguage{russian}{ь}sę gly siję \textbf{s\foreignlanguage{russian}{ъ}lagajǫšti} v\foreignlanguage{russian}{ъ} srdci svoem\foreignlanguage{russian}{ь}
{Mary.{\sc nom}} {\sc ptc} {treasure.{\sc impf.3.sg}} {all.{\sc m.acc.pl}} {word.{\sc m.acc.pl}} {this.{\sc acc.pl}} {ponder.{\sc ptcp.ipfv.f.nom.sg}} {in} {heart.{\sc n.loc.sg}} {own.{\sc loc.sg}}
\glt
\glend
\item[b.]
\gll {hē} {de} {Maria} {panta} {sunetērei} {ta} {rhe\^mata} {tauta} {\textbf{sunballousa}} {en} {tēi} {kardiai} {autēs}
{the.{\sc sg.f.nom}} {\sc ptc} {Mary.{\sc sg.f.nom}} {all.{\sc pl.n.acc}} {protect.{\sc impf.3.sg}} {the.{\sc pl.n.acc}} {thing.{\sc pl.n.acc}} {this.{\sc pl.n.acc}} {consider.{\sc ptcp.ipfv.f.nom.sg}} {in} {the.{\sc sg.f.dat}} {heart.{\sc sg.f.dat}} {him.{\sc 3.sg.f.gen}}
\glt `But Mary treasured all these things, pondering them in her heart' (Luke 2:19)
\glend 
\label{meansmannerelab}
\end{itemize}
\end{example}

\textit{CON\-TIN\-GEN\-CY\textunderscore Cause\textunderscore rea\-son} is a discourse function which we can equally expect to be compatible with \textsc{elaborations} and \textsc{frames}, though it is perhaps more typical for a participle to receive a causal reading when it follows the matrix clause, where \textsc{elaborations} are more common. (\ref{causepre}) and (\ref{causepost}) are examples of this tag on pre- and post-matrix conjunct participles respectively.

\begin{example}
\begin{itemize}
\item[a.]
\gll i ne \textbf{obrět\foreignlanguage{russian}{ъ}še} kǫdǫ v\foreignlanguage{russian}{ъ}nesti i narodom\foreignlanguage{russian}{ъ}. v\foreignlanguage{russian}{ъ}zlěz\foreignlanguage{russian}{ъ}še na chram\foreignlanguage{russian}{ъ}. skvozě skǫdel\foreignlanguage{russian}{ь} niz\foreignlanguage{russian}{ъ}věsišę i.
{and} {\sc neg} {find.{\sc ptcp.pfv.m.nom.pl}} {where} {bring in.{\sc prs.inf}} {him.{\sc 3.m.acc.sg}} {crowd.{\sc m.inst.sg}} {climb.{\sc ptcp.pfv.m.nom.pl}} {on} {building.{\sc m.acc.sg}} {through} {tiles.{\sc sg.f.acc}} {tiling.{\sc aor.3.pl}} {let down.{\sc 3.m.acc.sg}}
\glt 
\glend
\item[b.]
\gll {kai} {mē} {\textbf{heurontes}} {poias} {eisenegkōsin} {auton} {dia} {ton} {okhlon} {anabantes} {epi} {to} {dōma} {dia} {tōn} {keramōn} {kathēkan} {auton}
{and} {not} {find.{\sc ptcp.pfv.m.nom.pl}} {what way.{\sc sg.f.gen}} {bring in .{\sc aor.3.pl.sbjv}} {him.{\sc 3.m.acc.sg}} {because of} {the.{\sc m.acc.sg}} {crowd.{\sc m.acc.sg}} {go up.{\sc ptcp.pfv.m.nom.pl}} {on} {the.{\sc n.acc.sg}} {house.{\sc n.acc.sg}} {through} {the.{\sc pl.m.gen}} {tiling.{\sc pl.m.gen}} {let down.{\sc aor.3.pl}} {him.{\sc 3.m.acc.sg}}
\glt `And when they could not find by what way they might bring him in because of the crowd, they went upon the housetop, and let him down through the tiling' (Luke 5:19)
\glend 
\label{causepre}
\end{itemize}
\end{example}

\begin{example}
\begin{itemize}
\item[a.]
\gll i rǫgaachǫ sę emu \textbf{vědǫšte} ěko um\foreignlanguage{russian}{ь}rět\foreignlanguage{russian}{ъ}
{and} {laugh.{\sc impf.3.pl}} {\sc refl} {\sc 3.sg.m.dat} {know.{\sc ptpc.ipfv.m.nom}} {that} {die.{\sc aor.3.sg}}
\glt 
\glend
\item[b.]
\gll {kai} {kategelōn} {autou} {\textbf{eidotes}} {hoti} {apethanen}
{and} {laugh at.{\sc impf.3.pl}} {\sc 3.sg.m.gen} {know.{\sc ptcp.prf.m.nom.pl}} {that} {die.{\sc aor.3.sg}}
\glt `And they laughed at Him, knowing that she was dead' (Luke 8:53)
\glend 
\label{causepost}
\end{itemize}
\end{example}

\textit{TEM\-PO\-RAL\textunderscore Con\-ti\-nu\-a\-tion\textunderscore } is likely to mostly occur between two elaborating participles headed by the same matrix clause. If some equivalence be made between this tag and the SDRT description of the discourse relation \textit{Continuation}, then that seems to be clearly the case. According to SDRT, \textit{Continuation} is a discourse relation

\begin{quote}
    whose sole semantic content is to mark that its terms bear the same discourse relation to a dominating constituent. This implies that coordinated constituents of a sub-structure must behave in a homogeneous fashion with respect to a dominating constituent. (\pgcitealt{ashervieusubcoorddisc}{595})
\end{quote}

Most of the examples are, in fact, similar to (\ref{contin}), where \textit{TEM\-PO\-RAL\textunderscore Con\-ti\-nu\-a\-tion\textunderscore } holds between the coordinated elaborating participles (bracketed in the example) that follow the matrix clause.

\begin{example}
\begin{itemize}
\item[a.]
\gll Pride sn\foreignlanguage{russian}{ъ} člvčsky. {\normalfont[}\textbf{ědy} i \textbf{piję}{\normalfont]}
{come.{\sc aor.3.sg}} {son.{\sc nom.sg}} {man's.{\sc nom.m.sg}} {eat.{\sc ptcp.ipfv.m.nom.sg}} {and} {drink.{\sc ptcp.ipfv.m.nom.sg}}
\glt 
\glend
\item[b.]
\gll {elēluthen} {ho} {huios} {tou} {anthrōpou} {\normalfont[}{\textbf{esthiōn}} {kai} {\textbf{pinōn}}{\normalfont]}
{come.{\sc aor.3.sg}} {the.{\sc nom.m.sg}} {son.{\sc nom.m.sg}} {the.{\sc gen.m.sg}} {man.{\sc gen.m.sg}} {eat.{\sc ptcp.ipfv.m.nom.sg}} {and} {drink.{\sc ptcp.ipfv.m.nom.sg}}
\glt `The Son of Man came eating and drinking' (Luke 7:34)
\glend 
\label{contin}
\end{itemize}
\end{example}

If we look at the frequency of relations among pre- (Table \ref{prexadvrels}) and post-matrix (Table \ref{postxadvrels}) conjunct participles, we can see that our intuitions are largely confirmed. 

\begin{table}[!h]
\centering
\begin{tabular}{lll}
\hline
\textbf{discourse relation tag} & \textit{\textbf{n}} & \textbf{\%}\\
\hline
TEMPORAL\textunderscore Narration\textunderscore same story & 129 & 45.58 \\ 
TEMPORAL\textunderscore Asynchronous\textunderscore precedence & 66 & 23.32 \\ 
TEMPORAL\textunderscore Coincidence\textunderscore means & 11 & 3.88 \\ 
TEMPORAL\textunderscore Synchronous\textunderscore overlap & 10 & 3.53 \\ 
CONTINGENCY\textunderscore Cause\textunderscore reason & 9 & 3.18 \\ 
ATTRIBUTION\textunderscore Content\textunderscore speech content & 8 & 2.82 \\ 
TEMPORAL\textunderscore Synchronous\textunderscore circumstance & 8 & 2.82 \\ 
COMPARISON\textunderscore Concession\textunderscore  & 5 & 1.76 \\ 
TEMPORAL\textunderscore Narration\textunderscore new story & 5 & 1.76 \\ 
TEMPORAL\textunderscore Coincidence\textunderscore manner & 5 & 1.76 \\ 
CONTINGENCY\textunderscore Condition\textunderscore hypothetical & 5 & 1.76 \\ 
COMPARISON\textunderscore Juxtaposition\textunderscore contrast & 4 & 1.41 \\ 
CONTINGENCY\textunderscore Result\textunderscore real result & 3 & 1.06 \\ 
TEMPORAL\textunderscore Continuation\textunderscore  & 3 & 1.06 \\ 
CONTIGENCY\textunderscore Cause\textunderscore reason & 2 & 0.70 \\ 
COMPARISON\textunderscore Juxtaposition\textunderscore parallel & 2 & 0.70 \\ 
EXPANSION\textunderscore Elaboration\textunderscore characterization & 2 & 0.70 \\ 
ATTRIBUTION\textunderscore Content\textunderscore  & 1 & 0.35 \\ 
CONTINGENCY\textunderscore Purpose\textunderscore  & 1 & 0.35 \\ 
CONTINGENCY\textunderscore Condition\textunderscore general & 1 & 0.35 \\ 
EXPANSION\textunderscore Alternative\textunderscore chosen alternative & 1 & 0.35 \\ 
CONTINGENCY\textunderscore Result\textunderscore pragmatic result & 1 & 0.35 \\ 
COMPARISON\textunderscore Sameness\textunderscore  & 1 & 0.35 \\
\hline
\end{tabular}
\caption{Frequency of discourse relation tags among pre-matrix conjunct participles in the Codex Marianus}
\label{prexadvrels}
\end{table} 

\begin{table}[!h]
\centering
\begin{tabular}{lll}
\hline
\textbf{discourse relation tag} & \textit{\textbf{n}} & \textbf{\%}\\
\hline
TEMPORAL\textunderscore Synchronous\textunderscore circumstance & 63 & 44.36 \\ 
TEMPORAL\textunderscore Coincidence\textunderscore means & 33 & 23.23 \\ 
TEMPORAL\textunderscore Continuation\textunderscore  & 13 & 9.15 \\ 
TEMPORAL\textunderscore Synchronous\textunderscore overlap & 6 & 4.22 \\ 
TEMPORAL\textunderscore Coincidence\textunderscore manner & 5 & 3.52 \\ 
CONTINGENCY\textunderscore Cause\textunderscore reason & 5 & 3.52 \\ 
TEMPORAL\textunderscore Narration\textunderscore same story & 4 & 2.81 \\ 
EXPANSION\textunderscore Elaboration\textunderscore characterization & 4 & 2.81 \\ 
COMPARISON\textunderscore Concession\textunderscore  & 2 & 1.40 \\ 
CONTINGENCY\textunderscore Purpose\textunderscore  & 2 & 1.40 \\ 
EXPANSION\textunderscore Restatement\textunderscore specification & 2 & 1.40 \\ 
ATTRIBUTION\textunderscore Content\textunderscore Non-verbal content & 1 & 0.70 \\ 
TEMPORAL\textunderscore Asynchronous\textunderscore precedence & 1 & 0.70 \\ 
COMPARISON\textunderscore Juxtaposition\textunderscore contrast & 1 & 0.70\\
\hline
\end{tabular}
\caption{Frequency of discourse relation tags among post-matrix conjunct participles in the Codex Marianus}
\label{postxadvrels}
\end{table} 

We can see that the vast majority of the tag \textit{TEM\-PO\-RAL\textunderscore Nar\-ration\textunderscore same story} occurs in pre-matrix position, where we expect most \textsc{independent rhemes}. The same is true for \textit{TEM\-PO\-RAL\textunderscore A\-syn\-chro\-nous\textunderscore pre\-ce\-dence} and, to a lesser degree, for TEM\-PO\-RAL\textunderscore Syn\-chro\-nous\textunderscore overlap. These we can expect to occur more frequently with pre-matrix conjunct participles functioning as \textsc{frames}, particularly when the participle is translatable as an \textit{after}-clause (namely, when it introduces the \textit{TEM\-PO\-RAL\textunderscore A\-syn\-chro\-nous\textunderscore pre\-ce\-dence relation}). Among post-matrix conjunct participles, on the other hand, \textit{TEM\-PO\-RAL\textunderscore Syn\-chro\-nous\textunderscore cir\-cum\-stance} is by far the most common and overwhelmingly more frequent than in pre-matrix position. \textit{TEM\-PO\-RAL\textunderscore Co\-in\-ci\-dence\textunderscore means} and \textit{TEM\-PO\-RAL\textunderscore Con\-ti\-nu\-a\-tion\textunderscore }, other two relations which were strongly expected to occur most frequently with \textsc{elaborations} are also much more common among post-matrix participles. \\
\indent Finally, looking back at Table \ref{dasrhetrel}, we can see that the relations \textit{TEM\-PO\-RAL\textunderscore Syn\-chro\-nous\textunderscore o\-ver\-lap} and \textit{TEM\-PO\-RAL\textunderscore A\-syn\-chro\-nous\textunderscore pre\-ce\-dence} are by far the most common among dative absolutes, which fully agrees with their typical functions as \textsc{frames}.

\section{Summary}
Dative absolutes in the Codex Marianus were found to align strongly with the properties of \posscitet{baryhaug2011} \textsc{frames} regardless of syntactic configuration, which supports the observations made in previous literature (e.g. by \pgcitealt{worth1994a}{30}; \pgcitealt{corin1995a}{259}; \pgcitealt{collins2011a}{113}) that the unifying function of dative absolutes is a ‘backgrounding’ or ‘stage-setting’ one. The typical framing function of absolutes was immediately supported by their neat predominance in pre-matrix position, indicating that their placement in the sentence is in itself connected to their discourse function. Following \posscitet{baryhaug2011} account of framing participles, and similarly of \posscitet{chafe1976a} and \posscitet{krifka2007a} frame-setters, \textsc{frames} are, in fact, typically found in the leftmost position in the sentences, since they can be modeled as fronted adjuncts in topicalized position and they set the stage for the whole sentence. \\
\indent Conjunct participles, on the other hand, were found to clearly fulfil different discourse functions depending on syntactic configuration, aspect of the participle, and position of the subject (when overt). Similarly to \posscitet{baryhaug2011} and \posscitet{haug2012a} analysis of Ancient Greek conjunct participles, a highly significant association was found between the perfective aspect and pre-matrix position imperfective aspect and post-matrix position. This was taken as indirect evidence that pre-matrix conjunct participles in Early Slavic are more likely to function as \textsc{independent rhemes}, which are similar to independent clauses from the discourse perspective, whereas post-matrix conjunct participles are more likely to function as \textsc{elaborations}, which are temporally dependent on the matrix verb (i.e. they never introduce new temporal referents and they never induce narrative progression) and typically express manner or means/instrument. \\
\indent In sentence-initial position, even when perfectives, conjunct participles can be ambiguous between a \textsc{frame} and an \textsc{independent rheme} interpretation. We examined the anaphoric properties of sentence-initial conjunct participles and observed a significant connection between the subject's position relative to the participle and the average distance between the subject and its immediate antecedent. Subjects in the VS configuration tend to have, on average, antecedents that are significantly farther away compared to subjects in the SV configuration, which agrees with findings from \citet{haug2012a} on Ancient Greek data. The VS configuration in sentence-initial conjunct participles was attributed to the reintroduction of old or inactive referents, a function which can be associated with the use of \textsc{frames} to explicitly signal a shift in the event participants or their information status from the previous discourse.\\
\indent The subjects of dative absolutes, albeit very similar to sentence-initial VS conjunct participles in terms of average distance of the immediate antecedents and overall referential activation in the previous discourse, were found to be represented predominantly by inherently anaphoric parts of speech, whereas among VS conjunct participles more explicit ones (e.g. proper nouns) were found to be the most common. A possibility was advanced that dative absolutes predominantly attract few, very prominent discourse referents (e.g. Jesus or the Apostles, in the case of the Gospels), whereas framing conjunct participles may be frequently used with any salient referent. \\
\indent Annotation on Aktionsart on the Greek New Testament also allowed us to check which Aktionsarten more commonly occur among the parallels to dative absolutes and conjunct participles. Telic verbs are the most common among both constructions, although more prominently so among conjunct participles, whereas activity verbs are markedly more common among dative absolutes. This is also in line with the typical functions of the constructions observed through other variables: activity verbs are intuitively more easily used in framing constructions, whereas telic verbs more readily so for series of events with narrative progression.\\
\indent The analysis of discourse-relation annotation on the Gospel of Luke largely confirmed these patterns, showing that rhetorical relations intuitively more compatible with \textsc{independent rhemes} occur more frequently among pre-matrix conjunct participles, those more compatible with \textsc{frames} equally so among dative absolutes and pre-matrix conjunct participles, while those more compatible with \textsc{elaborations} are most frequent among post-matrix conjunct participles.\\
\indent Finally, Old Church Slavonic-Greek parallel data allowed us to single out potential mismatches between the two languages and to corroborate or refine observations made on the basis of largely overlapping patterns. Mismatches between absolute constructions have highlighted two main groups of independently used dative absolutes, those rendering nominalized accusatives with infinitive and those translating `agreeing absolutes' (i.e. pre-matrix participles constructions with a separate subject, like absolutes, but with the whole close agreeing with a matrix argument, like conjunct participles), in both cases clearly functioning as framing participles. Among conjunct participles, two main groups of mismatches can be most clearly identified, namely Greek main verbs which get translated into conjunct participles (of which those corresponding to a Greek imperative are a particularly clear sign of independent Slavic usage) and free rendition of Greek phrases other than participles into a (particularly elaborating) conjunct participle.
\\ \indent The next chapter replicates the analysis of dative absolutes and conjunct participles carried out so far on a much larger and diverse dataset, leveraging these findings as a `blueprint' to obtain further insights into the constructions' functions across different texts and varieties.

\chapter{Participle constructions in standard treebanks}

\section{Introduction}
This chapter looks into the properties of conjunct participles and dative absolutes in a much larger dataset containing morphosyntactic and dependency tagging, but no information-structural annotation, such as givenness status and anaphoric links from nominal referents to their antecedents. It is a relatively diverse subcorpus, in terms of geographic and historical varieties represented. The first and main part of the analysis uses all the treebanks from the overview in the Introduction, except for those referred to as \textit{strategically annotated} (i.e. fully automatically annotated) and the Codex Marianus, the latter used in the previous chapter. The second part of the analysis looks more briefly at the constructions in strategically annotated treebanks, which contain texts written in South Slavic varieties not well represented in the first part of the analysis.  \\
\indent The analysis in this chapter follows the same method adopted in Chapter 1 but for the parts concerning information status. The overarching goal is to compare the overall behaviour of the constructions in the Codex Marianus, representing one of the oldest and most deeply annotated Old Church Slavonic text---with aligned Greek parallels---, with that observed in texts of different Early Slavic varieties and genres, including non-religious or non-liturgical texts (most notably, original Slavic texts). 

\section{Dative absolutes and conjunct participles in standard treebanks}
% Dative absolutes were extracted in the same way as with the dataset used in Chapter 1, except for how null-subject dative absolutes were identified. To narrow down the occurrences of adverbial participles in the dative to potential dative absolutes, only potential null-subject dative participles preceding the matrix clause were included, unless the lemma was \textit{byti} `be', in which case they were included regardless of position, since a manual check indicated they are mostly temporal expressions of the kind \textit{pozdě byv\foreignlanguage{russian}{ъ}šu} `once it got late', which were also commented on in the previous chapter. The potential pre-matrix null-subject dative absolutes (41 occurrences, excluding those with \textit{byti} `be' as lemma) were manually checked and 28 of these were confirmed as actual absolutes. Most post-matrix null-subject dative participles were instead cases of nominalized participles, as in (\ref{notda}), and were therefore all excluded.

% \begin{example}
% \gll i naprasnijem\foreignlanguage{russian}{ъ} prěloženija čudo s\foreignlanguage{russian}{ъ}tvoril\foreignlanguage{russian}{ъ} \textbf{z\foreignlanguage{russian}{ъ}ręštiim\foreignlanguage{russian}{ъ}}
% {and} {suddenness.{\sc sg.n.inst}} {transformation.{\sc sg.n.gen}} {miracle.{\sc sg.n.acc}} {make.{\sc sg.result.ptcp.act.m.nom.strong}} {watch.{\sc ptcp.ipfv.m.dat.pl}}
% \glt `And by the suddenness of the transformation he has performed a miracle for the witnesses' (Codex Suprasliensis, \textit{Encomion on the 40 Martyrs of Sebasteia}) %85444
% \glend
% \label{notda}
% \end{example}

Following the extraction criteria outlined in Section \ref{identifconstr} of the Introduction, the final dataset resulted in 1,534 dative absolutes and 8,232 conjunct participles.\footnote{Query performed in December 2022.} As in Chapter 1, the analysis is divided into groups of variables expected to capture the relevant properties of the constructions. Section \ref{positionshallowsec} looks at the position of participle constructions relative to the matrix clause and at the distribution of imperfective and perfective occurrences, Section \ref{subjectsshallowsec} at the properties of realized subjects, and Section \ref{lexicalvarptcpshallowsec} at the lexical variation among the constructions in different subcorpora.\footnote{Note that `(Old) Church Slavonic' will be used to refer together to Old Church Slavonic and later Church Slavonic texts from this dataset (i.e. excluding the \textit{Codex Marianus}) unless otherwise specified.}

\subsection{Position and aspect distribution
}\label{positionshallowsec}
As Table \ref{ordershallow} shows, dative absolutes are more frequent in pre-matrix position (67.4\%), which is what we would expect given the clear framing function that emerged from the analysis in Chapter 1. However, as many as 13.9\% of the occurrences follow the matrix. This indicates that, outside the limited narrative style represented by the Gospels, the matrix-dative absolute configuration is not as uncommon as the first dataset suggested.\\
\indent The overall position of conjunct participles, as in the Gospels, does not in itself serve as a strong indication of particular functions. The pre-matrix position is somewhat more frequent (59.9\%), which is also to be expected, given the double function of conjunct participles in that position as \textsc{frames} and \textsc{independent rhemes}. As many as 40\% are instead found in post-matrix position. Based on the previous Chapter, these can be expected to mainly be \textsc{elaborations}, which, as we will see, seems to normally be the case.\\
\indent No clear matrix clause was identified for 18.7\% of dative absolutes and for 0.1\% of conjunct participles (\textit{NA} in Table \ref{ordershallow}) using the extraction method outlined in Section \ref{identifconstr} of the Introduction.

\begin{table}[!h]
\centering
\begin{tabular}{cccc}
\hline
& \textbf{pre-matrix} & \textbf{post-matrix} & \textit{NA} \\
\hline
\textbf{absolute} & 67.4\% (1034)    & 13.9\% (213) & 18.7\% (287)  \\
\textbf{conjunct} & 59.9\% (4932) & 40\% (3293) & 0.1\% (7)\\
\hline
\end{tabular}
\caption{Relative order of participle and matrix clause in standard treebanks}
\label{ordershallow}
\end{table}

Most of the occurrences of dative absolutes counted under \textit{NA} are connected to the matrix clause by a coordinating conjunction (\textit{i} ‘and’ or \textit{a} ‘but, whereas’), in which case, as explained in the Introduction, the annotation convention in PROIEL/TOROT is to add an empty verb node to stand in for the matrix clause. In such cases, purely for the purpose of the analysis in this chapter, the matrix verb is identified with the empty node itself, in order to make `non-canonical' occurrences of participle constructions more easily retrievable.\footnote{That is, these occurrences are by no means necessarily to be analysed as `true' elliptic constructions.} The remainder of the occurrences for which no information on the main verb was immediately available are absolutes lacking a clearly identifiable matrix clause, to which they could be attached in the same sentence. In many of these examples, they introduce reported speech, a usage examined by \pgcitet{collins2011a}{113-122} in his broader analysis of ‘syntactically independent’ dative absolutes in Early Slavic. As Collins notes, when the wider discourse is taken into consideration, it generally becomes clear that such occurrences are dependent on ‘larger-than-clause units’ (\pgcitealt{collins2011a}{116-118}). \\
\indent At the clause level, the dative absolutes \textit{prišedšju} `having come' and \textit{rekšju} `having told' in (\ref{ex18}a), for instance, appear to function as \textsc{independent rhemes}, and so do the conjunct participles in the following segment (\ref{ex18}b) (\textit{prišed\foreignlanguage{russian}{ъ}} `having come', \textit{razględavša} `having looked around', \textit{znamenavša} `having marked'). This analysis seems to be reinforced by the second-position focus marker/emphatic particle \textit{že} at the beginning of each clause in (\ref{ex18}a)-(\ref{ex18}c) (cf. \citealt{dimitrovazhe}). 

\begin{example}
\begin{itemize}
\item[a.]
\gll \textbf{prišedšju} že igumenu ko mně. i \textbf{rekšju} mi poidevě v pečeru k feodos\foreignlanguage{russian}{ь}jevi.
come.\textsc{ptcp.pfv.m.dat.sg} \textsc{ptc} prior.\textsc{dat} to \textsc{1.sg.dat} and say.\textsc{ptcp.pfv.m.dat.sg} \textsc{1.sg.dat} go.\textsc{imp.1.du} to crypt.\textsc{acc} to Theodosius.\textsc{dat} 
\glt
\glend
\item[b.]
\gll az\foreignlanguage{russian}{ъ} že \textbf{prišed}\foreignlanguage{russian}{ъ} i so igumenom\foreignlanguage{russian}{ъ}. ne svěduštju nikomuže. \textbf{razględavša} kudě kopati. i \textbf{znamenavša} město kdě kopati. kromě oust\foreignlanguage{russian}{ь}ja. 
\textsc{1.sg.nom} \textsc{ptc} come.\textsc{ptcp.pfv.m.nom.sg} and with prior.\textsc{ins} \textsc{neg} witness.\textsc{ptcp.ipfv.m.dat.sg} nobody.\textsc{dat} look-around.\textsc{ptcp.pfv.nom.du} where dig.\textsc{inf} and mark.\textsc{ptcp.pfv.nom.du} place.\textsc{acc} where dig.\textsc{inf} beyond entrance.\textsc{gen} 
\glt
\glend
\item[c.]
\gll \textbf{reč} že ko mně igumen\foreignlanguage{russian}{ъ}.
\textsc{ptc} say.\textsc{aor.3.sg} to \textsc{1.sg.dat} prior.\textsc{nom}
\glt ‘When the prior came to me and told me “let us go to the crypt of Theodosius”, I came, and, without anyone seeing, we looked about where to dig, and made a sign on the spot outside the entrance where we should excavate. The prior then said to me’ (\textit{Primary Chronicle}, Codex Laurentianus f. 70a)
\glend
\label{ex18}
\end{itemize}
\end{example}

The subsequent discourse (\ref{ex18}b)-(\ref{ex18}c) reveals that (\ref{ex18}a) is in fact part of a series of non-finite clauses leading up to a finite main clause (\textit{reč} `(he) said') (\ref{ex18}c). The order of the different types of clauses in (\ref{ex18}) can be motivated at the level of discourse organisation, where the dative absolutes in (\ref{ex18}a) still function as \textsc{frames} beyond the clause-level. As argued by \pgcitet{collins2011a}{126}, ‘the relation between the absolute and the unit (in some cases, larger than a clause) to which it is most closely linked in semantic terms is not always subordination in the syntactic sense’, and its function can be seen as ‘signalling that the proposition that it expresses is secondary in its discourse context’. \\
\indent The number of \textit{NA}s in Table \ref{ordershallow} for conjunct participles is negligible, compared to that of dative absolutes. 6 of the 7 occurrences are examples in which the main verb is implied from the preceding sentence, as in (\ref{exxadvna}), where the implied main verb is \textit{s\foreignlanguage{russian}{ъ}pověda} `told', which is in fact the main verb of the preceding sentence.

\begin{example}
\gll to že \textbf{uslyšav\foreignlanguage{russian}{ъ}} bod\foreignlanguage{russian}{ъ}chnovenyi oc\foreignlanguage{russian}{ь} naš\foreignlanguage{russian}{ь} ḟeodosii \textbf{osklabiv\foreignlanguage{russian}{ъ}} sja lic\foreignlanguage{russian}{ь}m\foreignlanguage{russian}{ь} i malo \textbf{prosl\foreignlanguage{russian}{ь}ziv\foreignlanguage{russian}{ъ}} sja <s\foreignlanguage{russian}{ъ}pověda> tomu
{that.{\sc m.acc.sg}} {\sc ptc} {listen.{\sc ptcp.pfv.m.nom.sg}} {filled.with.the.Holy.Spirit.{\sc m.nom.sg}} {father.{\sc m.nom.sg}} {our.{\sc nom.m}} {Feodosij.{\sc nom.m}} {smile.{\sc ptcp.pfv.m.nom.sg}} {\sc refl} {face.{\sc inst.sg}} {and} {little} {start.crying.{\sc ptcp.pfv.m.nom.sg}} {\sc refl} <{tell.{\sc aor.3.sg}}> {that.{\sc dat.m.sg}}
\glt ‘Then after listening, our father Feodosij, filled with the Holy Spirit, smiling with his face and tearing up a little <told> him’ (\textit{Life of Feodosij Pečerskij}, Uspenskij Sbornik f. 46b)
\glend
\label{exxadvna}
\end{example}

One occurrence is instead an overtly subordinated participle in a predicative complement clause introduced by \textit{jako} `that' (\ref{xadvcomp}).

\begin{example}
\gll t\foreignlanguage{russian}{ъ}gda že oc\foreignlanguage{russian}{ь} naš\foreignlanguage{russian}{ь} ḟeodosii nap\foreignlanguage{russian}{ъ}lniv\foreignlanguage{russian}{ъ} sę stgo dúcha načat\foreignlanguage{russian}{ъ} togo obličati jako neprav\foreignlanguage{russian}{ь}d\foreignlanguage{russian}{ь}no \textbf{s\foreignlanguage{russian}{ъ}tvoriv\foreignlanguage{russian}{ъ}ša} i ne po zakonu \textbf{sěd\foreignlanguage{russian}{ъ}ša} na stolě tom\foreignlanguage{russian}{ь} i jako ocę si i brat starěišago \textbf{prog\foreignlanguage{russian}{ь}nav\foreignlanguage{russian}{ъ}ša}
{then} {\sc ptc} {father.{\sc m.nom.sg}} {our.{\sc 1.m.nom.sg}} {Feodosij.{\sc m.nom.sg}} {fill.{\sc ptcp.pfv.m.nom.sg}} {\sc refl} {holy.{\sc sg.m.gen}} {spirit.{\sc sg.m.gen}} {begin.{\sc aor.3.sg}} {that.{\sc sg.m.gen}} {reproach.{\sc inf}} {that} {unfairly.{\sc sg.n.acc.strong}} {act.{\sc ptcp.pfv.m.gen.sg}} {and} {\sc neg} {according to} {law.{\sc m.dat.sg}} {sit.{\sc ptcp.pfv.m.gen.sg}} {on} {throne.{\sc sg.m.loc}} {that.{\sc sg.m.loc}} {and} {that} {father.{\sc sg.m.gen}} {this.{\sc 3.m.dat.sg}} {and} {brother.{\sc sg.m.gen}} {oldest.{\sc sg.m.gen.comp}} {banish.{\sc ptcp.pfv.m.gen.sg}}
\glt ‘At the same time, our father Feodosij, filled with the Holy Spirit, began to reproach the prince that he had acted unfairly, that he had not sat on the throne lawfully, and that he had expelled his father and older brother’ (\textit{Life of Feodosij Pečerskij}, Uspenskij Sbornik f. 58b)
\glend
\label{xadvcomp}
\end{example}

The usage of a conjunct participle introduced by \textit{jako} in contexts such as the one in (\ref{xadvcomp}) is not really at odds with the function of conjunct participles in their typical configuration as \textsc{independent rhemes} or \textsc{frames} (pre-matrix, juxtaposed, mostly perfective). The participles in (\ref{xadvcomp}) describe a series of events occurring one after the other in a chain-like fashion and presuppose that they collectively lead to some main situation (the current one in which Feodosij reproaches the prince).\\
\indent A closer look at potential post-matrix dative absolutes reveals that a few of these occurrences are either pre-matrix from the discourse perspective (e.g. \textit{byst\foreignlanguage{russian}{ъ}}-clauses in (\ref{ex19}); cf. Section \ref{bystsec} of Chapter 1) or augmented with overt subordinating conjunctions, such as \textit{jako} ‘as, since’ or \textit{ašte li} ‘if’,\footnote{The occurrence of absolute constructions introduced by a connective is obviously not unheard of (see in particular \citealt{stump-a}, and \citealt{koenig1991a} on English \textit{with}-augmented absolutes).} which, as the syntactic head of the participle, were extracted by the query as the potential matrix of the construction (thus resulting as post-matrix from the raw dataset). Some absolutes introduced by a subordinating conjunction do indeed follow the matrix (\ref{ex20}), whereas others precede it (\ref{ex21}):

\begin{example}
\gll i byst\foreignlanguage{russian}{ъ} \textbf{besědujǫštema} ima i \textbf{s\foreignlanguage{russian}{ъ}v\foreignlanguage{russian}{ъ}prašajǫštema} sę čto byv\foreignlanguage{russian}{ь}šee i č\foreignlanguage{russian}{ъ}to bǫdǫšteje i sam\foreignlanguage{russian}{ъ} sps\foreignlanguage{russian}{ъ} približ\foreignlanguage{russian}{ь} sę iděše s nima
and happen.\textsc{aor.3.sg} converse.\textsc{ptcp.ipfv.dat.du} \textsc{3.du.m.dat} and discuss.\textsc{ptcp.ipfv.dat.du} \textsc{refl} what be.\textsc{ptcp.pfv.n.acc.sg} and what be.\textsc{ptcp.fut.n.acc.sg} and self.\textsc{nom} saviour.\textsc{nom} approach.\textsc{ptcp.pfv.m.nom.sg} \textsc{refl} go.\textsc{impf.3.sg} with \textsc{3.du.inst}
\glt ‘And it came to pass that, as they were conversing and discussing what had been and what would be, the Saviour himself approached them and started walking with them’ (John Chrysostom, \textit{Homily on Christ’s resurrection}, Codex Suprasliensis f. 103v)
\glend
\label{ex19}
\end{example}

\begin{example}
\gll i plakati sę nača popad\foreignlanguage{russian}{ь}ja \textbf{jako} \textbf{mertvu} \textbf{suštju} onomu
and cry.\textsc{inf} \textsc{refl} begin.\textsc{aor.3.sg} priest’s-wife.\textsc{nom} as dead.\textsc{dat} be.\textsc{ptcp.ipfv.m.dat.sg} that.\textsc{m.dat}
\glt ‘And the priest’s wife started to weep, as if he were dead’ (\textit{Primary Chronicle}, Codex Laurentianus f. 88b)
\glend
\label{ex20}
\end{example}

\begin{example}
\gll \textbf{ašte} li sice. \textbf{izvolšu} bgu. ... da budet volę gnę.
if \textsc{q} thus desire.\textsc{ptcp.pfv.m.dat.sg} god.\textsc{dat} \textsc{...} \textsc{vol} be.\textsc{fut.3.sg} will.\textsc{f.nom.sg} lord.\textsc{adj.f.nom.sg}
\glt ‘It thus God desires ... may Lord’s will be done’ (\textit{Life of Sergij of Radonezh} f. 59v)
\glend
\label{ex21}
\end{example}

Overall, however, \textit{byst\foreignlanguage{russian}{ъ}}-clauses and augmented absolutes are very few compared to the total number of occurrences (15 \textit{byst\foreignlanguage{russian}{ъ}}-clauses and 8 augmented absolutes), so that the vast majority of post-matrix absolutes in Table \ref{ordershallow} should be considered genuine occurrences following their matrix clause.\\
\indent As in the Gospels, by crossing aspect with position in the sentence, we obtain clearer functional patterns, as Table \ref{xadv-aspect-position} shows.

\begin{table}[!h]
\centering
\begin{tabular}{ccc}
\hline
& \textbf{imperfectives} & \textbf{perfectives} \\
\hline
\textbf{pre-matrix} & 45.8\% (474) & 54.2\% (560)\\
\textbf{post-matrix}  & 76.1\% (162) & 23.9\% (51)\\
\textit{NA}  & 50.9\% (146) & 49.1\% (141)\\
\textbf{tot.} & 51\% (782) & 49\% (752) \\
\hline
\end{tabular}
\caption[Dative absolutes in standard treebanks: aspect distribution by position relative to the matrix clause]{Dative absolutes in standard treebanks: aspect distribution by position relative to the matrix clause (row percentage)}
\label{abs-aspect-position}
\end{table}

\begin{table}[!h]
\centering
\begin{tabular}{ccc}
\hline
& \textbf{imperfectives} & \textbf{perfectives} \\
\hline
\textbf{pre-matrix} & 27\% (1334) & 73\% (3598)\\
\textbf{post-matrix}  & 76.5\% (2518) & 23.5\% (775)\\
\textit{NA}  & 57.1\% (4) & 42.9\% (3)\\
\textbf{tot.} & 46.8\% (3856) & 53.2\% (4376) \\
\hline
\end{tabular}
\caption[Conjunct participles in standard treebanks: aspect distribution by position relative to the matrix clause]{Conjunct participles in standard treebanks: aspect distribution by position relative to the matrix clause (row percentage).}
\label{xadv-aspect-position}
\end{table}

Just like in the Gospels, the association between aspect and position is statistically significant among conjunct participles, with moderate to strong effect size ($\phi=0.48$, Cramér's $V$): $\chi^{2}(1) = 1924.52$, $p<0.01$, with much higher odds for perfectives to occur in pre-matrix than post-matrix position and for imperfectives to occur in post-matrix than pre-matrix position (odds ratio: 8.69). This indicates that their usage as \textsc{independent rhemes}, as in (\ref{indrheme}), and \textsc{elaborations}, as in (\ref{elab}), is overall very similar to that found in the Codex Marianus. 

\begin{example}
\gll drakula že \textbf{vostav\foreignlanguage{russian}{ъ}} \textbf{vzem} meč\foreignlanguage{russian}{ь} svoi iskoči s polaty
{Dracula.{\sc m.nom.sg}} {\sc ptc} {stand.{\sc ptcp.pfv.m.nom.sg}} {take.{\sc ptcp.pfv.m.nom.sg}} {sword.{\sc sg.m.acc}} {his.{\sc acc.sg}} {jump out.{\sc aor.3.sg}} {from} {palace.{\sc f.gen.sg}}
\glt ‘Then Dracula stood up, took his sword and jumped out of the palace’ (\textit{The Tale of Dracula}, f. 215v)
\glend
\label{indrheme}
\end{example}

\begin{example}
\gll ękove brate ebi \textbf{ležę} {ebechoto aesovo}
{Jakov.{\sc voc}} {brother.{\sc voc.sg}} {screw.{\sc imp.2.sg}} {lie down.{\sc ptcp.ipfv.m.nom.sg}} {horny ballser.{\sc voc.sg}}
\glt ‘Brother Jakov, do your screwing lying down, you horny ballser!' (Birchbark letters St. R. 35, translation by \citealt{schaeken2018}) %%255189
\glend 
\label{elab}
\end{example}

Among dative absolutes, perfectives and imperfectives are overall almost equally frequent (49\% and 51\% respectively). Among pre-matrix occurrences, perfectives are somewhat more frequent, whereas among post-matrix ones, imperfectives predominate. Also in this case there is a statistically significant association between aspect and position in the sentence, but with a much lower effect size ($\phi=0.25$, Cramér's $V$): $\chi^{2}(1) = 76.88$, $p<0.01$, with higher odds for perfectives to occur in pre-matrix than post-matrix position and for imperfectives to occur in post-matrix than pre-matrix position (odds ratio: 4.07). However, because of the very different numbers of pre- and post-matrix absolutes (1034 versus 213), the raw number of imperfective absolutes is still higher among pre-matrix than post-matrix absolutes, which means that a test of independence is perhaps not the most suitable in this case. Among dative absolutes to the left of the matrix, we can see that the proportion of the two aspects is relatively even, in line with the pattern found in the Gospels. This was explained in Section \ref{frameworks} of the Introduction, following \pgcitet{haug2012a}{311-312}, with the fact that the dominant aspect of framing participles is expected to vary with genre – unlike conjunct participles, whose aspect was shown to strongly correlate with their position. The difference in number between imperfectives and perfectives in pre-matrix position is, in fact, not statistically significant in itself.\footnote{Binomial test, $p=0.99$} Among dative absolutes to the right of the matrix, however, imperfectives are clearly more frequent. Recall that for the Gospels no generalizations about post-matrix absolutes were possible, because of the very low number of occurrences in that position compared to pre-matrix absolutes. This time we observe a significantly higher number of imperfective than perfective post-matrix dative absolutes.\footnote{Binomial test, $p<0.01$.} \textit{Semantically}, it appears that three main groups of post-matrix dative absolutes could be identified, exemplified by (\ref{postdastype1})-(\ref{postdastype3}). 

\begin{example}
\gll on\foreignlanguage{russian}{ъ} že šed\foreignlanguage{russian}{ъ} sěde na stolě černigově jaroslavu \textbf{suštju} nověgorodě togda
{that.{\sc m.nom.sg}} {\sc ptc} {go.{\sc ptcp.pfv.m.nom.sg}} {sit.{\sc aor.3.sg}} {on} {throne.{\sc sg.m.loc}} {Chernigov.{\sc sg.m.loc}} {Yaroslav.{\sc m.dat.sg}} {be.{\sc ptcp.ipfv.m.dat.sg}} {Novgorod.{\sc sg.m.loc}} {then}
\glt ‘He thus departed thence and established himself upon the throne of Chernigov, while Yaroslav was at Novgorod.’ (\textit{Primary Chronicle}, Codex Laurentianus f. 50b)
\glend
\label{postdastype1}
\end{example}

\begin{example}
\gll i chodi po manastyrem\foreignlanguage{russian}{ъ}. i ne v\foreignlanguage{russian}{ъ}zljubi bu ni \textbf{chotęštju}
{and} {go.{\sc aor.3.sg}} {about} {monastery.{\sc m.dat.pl}} {and} {\sc neg} {love.{\sc aor.3.sg}} {God.{\sc m.dat.sg}} {\sc neg} {want.{\sc ptcp.ipfv.m.dat.sg}} 
\glt ‘He went about the monasteries and liked none of them, since God did not so will’ (\textit{Primary Chronicle}, Codex Laurentianus f. 53a)
\glend
\label{postdastype2}
\end{example}

\begin{example}
\gll i oněm\foreignlanguage{russian}{ъ} že stav\foreignlanguage{russian}{ъ}šem\foreignlanguage{russian}{ъ} na nošt\foreignlanguage{russian}{ь}něm\foreignlanguage{russian}{ь} stanovišti blaženyi že ne doida jako i z\foreignlanguage{russian}{ь}rěimo ich\foreignlanguage{russian}{ъ} tu že opočivaaše jedinomu bogu \textbf{s\foreignlanguage{russian}{ъ}bljudajuštju} i
{and} {that.{\sc m.dat.pl}} {\sc ptc} {stay.{\sc ptcp.pfv.m.dat.pl}} {in} {night.{\sc sg.n.loc}} {camp.{\sc sg.n.loc}} {blessed.{\sc m.nom.sg}} {\sc ptc} {\sc neg} {arrive.{\sc ptcp.ipfv.m.nom.sg}} {as} {\sc ptc} {distance.{\sc sg.n.acc}} {them.{\sc gen}} {here} {\sc ptc} {rest.{\sc impf.3.sg}} {alone.{\sc m.dat.sg}} {God.{\sc m.dat.sg}} {watch over.{\sc ptcp.ipfv.m.dat.sg}} {\sc 3.sg.m.acc}
\glt ‘And when they stayed overnight, the blessed, stopping to see them from afar, spent the night here, and God alone guarded him.' (\textit{Life of Feodosij Pečerskij}, Uspenskij sbornik f. 31a) %278033
\glend
\label{postdastype3}
\end{example}

The first, and most common, exemplified by (\ref{postdastype1}), involves absolutes receiving a temporal interpretation and adding the circumstances in which the main eventuality occurs, specifically with respect to an event participant outside of the foreground. As we will see, temporally, these should still be considered on a par with what we have called \textsc{frames} so far. The second group, exemplified by (\ref{postdastype2}) involves those receiving a causal interpretation. \textit{Temporally}, these are also \textsc{frame}-like. The reason why neither of these examples can be considered an \textsc{elaboration} is that \textsc{elaborations} are temporally dependent on the matrix clause, whose runtime they inherit, and enforce a reading whereby the runtime of the matrix may \textit{include} or, more often, is equivalent to, that of the participle. Both \textsc{frames} and \textsc{elaborations}, when imperfective, indicate that the participle event goes on for the whole runtime of the matrix event, but the opposite, namely that the main event holds for the whole runtime of the participle event, is only true for \textsc{elaborations} (\pgcitealt{haug2012a}{296}). Note the difference between (\ref{inclusion}a), a \textsc{frame}, and (\ref{inclusion}b), an \textsc{elaboration}.

\begin{example}
\begin{itemize}
    \item[a.] \textit{Pondering over what had happened}, he finally realized what went wrong
    \item[b.] He walked through the park \textit{pondering over what had happened}
\end{itemize}
\label{inclusion}
\end{example}

The states introduced by the predicates `be in Novgorod' in (\ref{postdastype1}) and `(not) will' in (\ref{postdastype2}) do not necessarily only last as long as the main eventuality does. In both cases, in fact, a reading in which the event time of the matrix clause is \textit{included} in that of the absolute seems more likely. \\
\indent The third example, however, is quite different from the other two and is more compatible with \textsc{elaborations}. \textit{S\foreignlanguage{russian}{ъ}bljudajuštju} `guarding' in (\ref{postdastype3}) clearly uses the matrix event time as its reference time. The event time of \textit{spent the night here} lasts for the whole event time of \textit{God alone guarding him}, but also, crucially, the whole event time of \textit{God alone guarding him} lasts for the whole event time of \textit{spent the night here}. Intuitively, the event time of \textit{God alone guarding him} does not necessarily cover the preceding discourse within the same sentence, namely \textit{stopping to see from afar}, but rather forms a complex predicate only with the matrix clause. \\
\indent In some examples, the \textsc{elaboration} function of the dative absolute is unmistakable, as in (\ref{clearelabda1}).

\begin{example}
\gll potom\foreignlanguage{russian}{ъ} v\foreignlanguage{russian}{ъ}zmǫštajet\foreignlanguage{russian}{ъ} sę i v\foreignlanguage{russian}{ъ}skypit\foreignlanguage{russian}{ъ} zǫbom\foreignlanguage{russian}{ъ} \textbf{zyb\foreignlanguage{russian}{ь}jǫštem\foreignlanguage{russian}{ъ}} sę žilam\foreignlanguage{russian}{ъ} \textbf{s\foreignlanguage{russian}{ъ}gr\foreignlanguage{russian}{ъ}ždajǫštem\foreignlanguage{russian}{ъ}} sę i v\foreignlanguage{russian}{ь}semu tělu nevolejǫ \textbf{s\foreignlanguage{russian}{ъ}lęcajǫštu} sę
{afterwards} {shake.{\sc prs.3.sg}} {\sc refl} {and} {boil.{\sc prs.3.sg}} {tooth.{\sc m.dat.pl}} {become loose.{\sc ptcp.ipfv.m.dat.pl.strong}} {\sc refl} {vein.{\sc pl.f.dat}} {become contracted.{\sc ptcp.ipfv.f.dat.pl}} {\sc refl} {and} {whole.{\sc sg.n.dat}} {body.{\sc sg.n.dat}} {discomfort.{\sc f.inst.sg}} {twist.{\sc ptcp.ipfv.n.dat.sg}} {\sc refl}
\glt ‘Afterwards, he was shaking and boiling up, the teeth going loose, the veins becoming contracted, the whole body twisting in discomfort’ (\textit{Encomion on the 40 Martyrs of Sebasteia}, Codex Suprasliensis 6)
\glend
\label{clearelabda1}
\end{example}

In this example, the absolutes clearly add granularity to the main verbs, with the temporal effect that the running time of the matrix and of the participles is identical.\\
\indent A small group of post-matrix imperfective absolutes may receive a contrastive or adversative reading---a semantic effect which strikes a clear chord, cross-linguistically, with the double function of temporal connectives (expressing simultaneity) as adversative conjunctions (cf. English \textit{while}, as in \textit{he went out with friends, while I stayed at home}), particularly in postponed position (cf. \citealt{maurigiacalonementre} and \citealt{musimentre}). (\ref{contrastelab1}) and (\ref{contrastelab2}) are two such examples. 

\begin{example}
\gll i povelě prinesti svěštę goręštę. i prižagati lice paule. propovědniku \textbf{v\foreignlanguage{russian}{ь}pijǫštu} i glagolǫštu
{and} {command.{\sc aor.3.sg}} {bring.{\sc inf}} {candle.{\sc pl.f.acc}} {burn.{\sc ptcp.ipfv.f.acc.pl}} {and} {burn.{\sc inf}} {face.{\sc sg.n.acc}} {Paul's.{\sc sg.n.acc.strong}} {preacher.{\sc m.dat.sg}} {cry out.{\sc ptcp.ipfv.m.dat.sg}} {and} {say.{\sc ptcp.ipfv.m.dat.sg}}
\glt ‘And he commanded to bring burning candles and burn Paul’s face, while the preacher was crying out, saying’ (\textit{Vita of Paul and Juliana}, Codex Suprasliensis f. 7r)
\glend
\label{contrastelab1}
\end{example}

\begin{example}
\gll togo že styi priim\foreignlanguage{russian}{ъ} prěpodob\foreignlanguage{russian}{ь}nyima svoima rǫkama pojemlę maslo ot\foreignlanguage{russian}{ъ} kandila mazaaše jemu v\foreignlanguage{russian}{ь}se tělo \textbf{kažuštu} bolęštuumu i \textbf{naznamenujǫštu} v\foreignlanguage{russian}{ъ} koem\foreignlanguage{russian}{ъ} městě pl\foreignlanguage{russian}{ъ}zaaše dch\foreignlanguage{russian}{ъ} oky zmija v\foreignlanguage{russian}{ь}gněždaę sę v\foreignlanguage{russian}{ь} nem\foreignlanguage{russian}{ь}
{that.{\sc sg.m.gen}} {\sc ptc} {saint.{\sc m.nom.sg}} {take.{\sc ptcp.pfv.m.nom.sg}} {holy.{\sc f.inst.du}} {his.{\sc 3.f.inst.du}} {hand.{\sc f.inst.du}} {take.{\sc ptcp.ipfv.m.nom.sg}} {oil.{\sc sg.n.acc}} {from} {vigil lamp.{\sc sg.n.gen}} {smear.{\sc impf.3.sg}} {\sc 3.sg.m.dat} {whole.{\sc sg.n.acc}} {body.{\sc sg.n.acc}} {show.{\sc ptcp.ipfv.m.dat.sg}} {be sick.{\sc ptcp.ipfv.m.dat.sg}} {and} {point.{\sc ptcp.ipfv.m.dat.sg}} {in} {which.{\sc sg.n.loc}} {place.{\sc sg.n.loc}} {creep.{\sc impf.3.sg}} {soul.{\sc m.nom.sg}} {like} {serpent.{\sc f.nom.sg}} {nest.{\sc ptcp.ipfv.m.nom.sg}} {\sc refl} {in} {\sc 3.sg.m.loc}
\glt ‘After taking him with his holy hands, the saint took oil from the vigil lamp and started smearing his whole body, while the sick was showing and pointing in which place the soul was creeping like a serpent nesting itself in him’ (\textit{Vita of St. Aninas the Wonderworker}, Codex Suprasliensis f. 150r)
\glend
\label{contrastelab2}
\end{example}

Similarly, some examples receive a conditional interpretation, as in (\ref{contrastelab3}).

\begin{example}
\gll n\foreignlanguage{russian}{ъ} i tako ne priimǫt\foreignlanguage{russian}{ъ} blagoděti židove m\foreignlanguage{russian}{ь}ně ne \textbf{sǫštu} tu
{but} {even} {so} {\sc neg} {receive.{\sc prs.3.pl}} {grace.{\sc gen.sg}} {Jew.{\sc m.nom.pl.}} {\sc 1.sg.m.dat} {\sc neg} {be.{\sc ptcp.ipfv.m.dat.sg}} {here}
\glt ‘Jews will not receive grace while I am not here’ (John Chrysostom, \textit{Homily on the resurrection of Lazarus after four days}, Codex Suprasliensis f. 23v)
\glend
\label{contrastelab3}
\end{example}

Both cases such as (\ref{postdastype1}), and those such as (\ref{contrastelab1})-(\ref{contrastelab2}) and (\ref{contrastelab3}) would be considered by \citet{fabricius-hansen2012b} as temporally and semantically equivalent to fronted, frame-setting adjuncts, even though they occur in post-matrix position. As discussed in Section \ref{framesetterssec} of the Introduction, they can be interpreted as \textit{restrictive} adverbial adjuncts, since, unlike typical \textsc{independent rhemes} and \textsc{elaborations}, they restrict the domain of the matrix VP by introducing a \textit{state} within which the main eventuality holds. If we consider the similar semantic contribution of (mildly or strongly contrastive) post-matrix restrictive adjuncts and frame-setters, we can see how dative absolutes may have a fairly consistent function regardless of position in the sentence and semantic effects (such as constrastivity and causality). \\
\indent The interpretation of imperfective dative absolutes as \textsc{elaborations} in post-matrix position is, however, clearly possible (as (\ref{clearelabda1}) showed), and may to some extent be compositionally driven, given the correlation between imperfective aspect and post-matrix position. Table \ref{elabvsframes} shows the number of imperfective post-matrix dative absolutes which \textit{could} be interpreted as \textsc{elaborations} on the basis of temporal relations between the matrix and the absolutes, versus those that cannot. Those with \textit{byti} in the aorist as the matrix verb are excluded from the count, as they are all \textit{byst\foreignlanguage{russian}{ъ}}-clauses, where the absolute is pre-matrix from the discourse perspective. Non-\textsc{elaborations} largely have a \textsc{frame}-like temporal semantics (i.e. they do not depend on the matrix clause temporally and the event time they introduce includes that of the matrix clause) and can be mostly interpreted as clause-final restrictive adjuncts. These, as already discussed, can be semantically equivalent to fronted adjuncts, so they are referred to as \textsc{frames}, even if they are not fronted.\footnote{The numbers are based on my own annotations, without the opinion of a second annotator, as is standard practice when annotating a corpus, particularly of a dead language, so one should be cautious when evaluating them. The dataset with the contexts tagged as \textsc{elaboration} or \textsc{frame} can be found in the repository: \url{https://doi.org/10.6084/m9.figshare.24166254}.}

\begin{table}[!h]
\centering
\begin{tabular}{cc}
\hline
\textbf{function}  & \textbf{\textit{n}}\\
\hline
\textsc{elaboration} & 16.2\% (23)\\
\textsc{frame}  & 83.8\% (119)\\
\hline
\end{tabular}
\caption[\textsc{Elaborations} versus \textsc{frame} imperfective absolutes in post-matrix position]{\textsc{Elaborations} versus \textsc{frame} imperfective absolutes in post-matrix position}
\label{elabvsframes}
\end{table}

Overall, post-matrix dative absolutes functioning as \textsc{frames} are much more common than those serving as clear \textsc{elaborations}. Many \textsc{elaboration} dative absolutes involve verbs such as \textit{pomagati} `help', \textit{pospěš\foreignlanguage{russian}{ь}estvovati} `contribute, help', \textit{s\foreignlanguage{russian}{ъ}bljudati} `guard, watch over', \textit{ukrěpljati} `give strength', with their subject being invariably \textit{bog\foreignlanguage{russian}{ъ}} `God'. These are rather clear examples of \textsc{elaborations} receiving a \textit{manner}, \textit{comitative} or \textit{instrument} reading (e.g. `with the help of God'), as in (\ref{elabwiththehelp}), which is quite typical of elaborating adverbial clauses as a whole (cf. \citealt{fabricius-hansen2012b}).

\begin{example}
\gll i to rek\foreignlanguage{russian}{ъ} paki po pervoe dr\foreignlanguage{russian}{ъ}žaše sę dobroe ustroenie. \textbf{bogu} \textbf{pomagajuštu} \textbf{emu} na blagoe proizvolenie
{and} {that.{\sc n.acc.sg}} {say.{\sc ptcp.pfv.m.nom.sg}} {more} {on} {righteous.{\sc n.acc.sg}} {stay.{\sc impf.3.sg}} {\sc refl} {good.{\sc n.acc.sg}} {behaviour.{\sc sg.n.acc}} {god.{\sc sg.m.dat}} {help.{\sc ptcp.ipfv.m.dat.sg}} {\sc 3.sg.m.dat} {on} {good.{\sc n.acc.sg}} {intention.{\sc n.acc.sg}}
\glt `After saying this, he stayed on the righteous path more than before, with God helping him in his good intentions' (\textit{Life of Sergij of Radonezh} f. 46v)
\glend
\label{elabwiththehelp}
\end{example}

As Table \ref{elabdasfreqvarieties} shows, examples of dative absolutes as \textsc{elaborations} are relatively evenly distributed across varieties if we consider the proportions of \textsc{frame}-like and \textsc{elaboration} dative absolutes. In raw terms, \textit{post-matrix} absolutes may seem to be more frequent in non-translated texts and East Slavic varieties, but these are, as already mentioned, over-represented in this dataset to start with.

\begin{longtable}{cccc}
\hline
 & \textsc{elaboration} 	 & \textsc{frame} \\
 \hline
\textbf{East Slavic} &  15 & 73\\
\textbf{South Slavic} &  8 & 46\\
\hline
\textbf{Old East Slavic} & 8 & 64\\
\textbf{Early Middle Russian} &  6 & 9\\
\textbf{Late Middle Russian} &  1 & 0\\
\textbf{Old Church Slavonic} &  8 & 46\\
\textbf{Later Church Slavonic} & 0 & 0 \\
\hline
\textbf{translations} &  8	& 46\\
\textbf{originals} &  15 & 73\\
 \hline
 {Codex Suprasliensis}	 & 8	 & 45\\
 \textit{Domostroj}  &	1	 & 0\\
 \textit{Life of Sergij of Radonezh}  &	6	 & 9\\
 \textit{Psalterium Sinaiticum}  &	0 & 	1\\
 \textit{First Novgorod Chronicle} (Synodal)  &	0	 & 3\\
 \textit{Primary Chronicle} (Hypthianus)  &	1 & 	0\\
 \textit{Primary Chronicle} (Laurentianus)  &	1	 & 25\\
 \textit{Suzdal Chronicle} (Laurentianus)  & 4	 & 14\\	
 {Uspenskij Sbornik} &  2	 & 22\\
\hline
\caption[Frequency of \textsc{elaborations} and \textsc{frame} imperfective absolutes by variety, text (or manuscript), and by translated versus original Early Slavic texts]{Frequency of \textsc{elaborations} and \textsc{frame} imperfective absolutes by variety, text (or manuscript), and by translated versus original Early Slavic texts. Only manuscripts in which at least one \textsc{elaboration} is found are indicated. Varieties are listed regardless}
\label{elabdasfreqvarieties}
\end{longtable}

\subsubsection{Summary}
As in deeply annotated treebanks, the analysis of conjunct participles indicated a strong association between their position relative to their matrix clause and their aspect, pointing to their overall typical functions (already established in the previous chapter) as \textsc{independent rhemes} in pre-matrix position, where they are much more likely to be perfective, and as \textsc{elaborations} in post-matrix position, where they are much more likely to be imperfective. Post-matrix dative absolutes were found to be less frequent (13.9\%) than those in pre-matrix position, but not as uncommon as post-matrix absolutes in the Codex Marianus (where they only constituted 7.1\% of the occurrences). Given the higher number of post-matrix absolutes than those extracted from the first dataset, we were able to look for potential associations between position in the sentence and other variables. We found some (significant, but not as strong as with conjunct participles) association between the perfective aspect and the pre-matrix position, and between the imperfective and the post-matrix position. We therefore looked more closely into post-matrix occurrences and observed that three main groups of post-matrix absolutes could be distinguished semantically, each with potential differences in their most likely temporal interpretation. 

The first and most common group is made of absolutes receiving a primary temporal interpretation and clearly functioning as \textsc{frames} despite being found in post-matrix position (e.g. \textit{He thus departed thence and established himself upon the throne of Chernihiv, \textbf{while Yaroslav was at Novgorod}}). Several of these occurrences may also receive a (mildly or strongly) contrastive or conditional reading (e.g. \textit{And he commanded to bring burning candles and burn Paul’s face, \textbf{while the preacher was crying out}, saying} or \textit{The Jews will not receive Grace \textbf{if I’m not here}}). These can be interpreted as clause-final restrictive adjuncts, which, as \citet{fabricius-hansen2012b} explain and as we mentioned in Section \ref{framesetterssec} of the Introduction, often receive contrastive readings and correspond semantically to frame-setting fronted adjuncts, in that they restrict the domain of the matrix clause by introducing a \textit{state} within which the main eventuality holds. 

The second group are absolutes receiving a causal interpretation (e.g. \textit{He went about the monasteries and liked none of them, \textbf{since God did not so will}}), which can also be considered similar to \textsc{frames} from the \textit{temporal} perspective, since their runtime may \textit{include} the runtime of the matrix clause, which may be considered incompatible with an \textsc{elaboration} interpretation, since with \textsc{elaborations} the runtime of the matrix clause and of the participle clause are identical. 

The third and final main group of post-matrix absolutes are those that can actually be interpreted as \textsc{elaborations}. In such cases, we may get the same temporal identity we observe with elaborating conjunct participles (e.g. \textit{Afterwards, he was shaking and boiling up, the teeth \textbf{going} loose, the veins \textbf{becoming} contracted, the whole body \textbf{twisting} in discomfort}). A manual annotation of post-matrix occurrences indicated that about 16\% of post-matrix absolutes \textit{can} receive an \textsc{elaboration} interpretation, but as many as 83.8\% of them are still compatible with the temporal semantics of \textsc{frames} despite being in post-matrix position. Furthermore, several elaborating dative absolutes are represented by almost-fixed expressions of the type `with God helping \textsc{person}' and occurring with verbs such as \textit{s\foreignlanguage{russian}{ъ}bljudati} `guard, watch over', \textit{pomagati} `help', \textit{ukrěpljati} `give strength', with their subject being invariably \textit{bog\foreignlanguage{russian}{ъ}} `God' (e.g. \textit{And when they stayed overnight, the blessed, stopping to see them from afar, spent the night here, with \textbf{God alone guarding him}}).

\subsection{Properties of subjects}\label{subjectsshallowsec}
Table \ref{nullvsovertxadvdasshallow} shows the frequency of overt and null subjects among sentence-initial conjunct participles and dative absolutes.

\begin{table}[!h]
\centering
\begin{tabular}{cccc}
\hline
& \textbf{overt} & \textbf{null}\\
\hline
\textbf{conjunct (sentence-initial)} & 50.8\% (1564) & 49.2\% (1512) \\
\textbf{absolute (sentence-initial)} & 94.4\% (656) & 5.6\% (39)\\
\hline
\end{tabular}
\caption[Overt and null subjects in sentence-initial conjunct participles and dative absolutes in standard treebanks]{Overt and null subjects in sentence-initial conjunct participles and dative absolutes in standard treebanks (row percentage)}
\label{nullvsovertxadvdasshallow}
\end{table}

As in the previous dataset, the vast majority of sentence-initial dative absolutes have an overt subject, while among sentence-initial conjunct participles overt and null subjects are virtually equally frequent. \\
\indent When we look at the position of the overt subject relatively sentence-initial conjunct participles and dative absolutes (Table \ref{svversusvsshallow}), we see that the VS configuration is not predominant as in the Codex Marianus: in fact, it is slightly less frequent than the SV one. 

\begin{table}[!h]
\centering
\begin{tabular}{ccc}
\hline
& \textbf{SV} & \textbf{VS}\\
\hline
\textbf{conjunct (sentence-initial)} & 65.2\% (1020) & 34.8\% (544)\\
\textbf{absolute (sentence-initial)} & 52.4\% (344) & 47.6\% (312)\\
\hline
\end{tabular}
\caption[Position of overt subjects relative to sentence-initial conjunct participles and dative absolutes in standard treebanks]{Position of overt subjects relative to sentence-initial conjunct participles and dative absolutes in standard treebanks (row percentage)}
\label{svversusvsshallow}
\end{table}

As Table \ref{dasubpos} shows, over half of the subjects of dative absolutes are represented by parts of speech which are either inherently anaphoric (personal and demonstrative pronouns) or likely to encode old or accessible referents (proper nouns).

\begin{longtable}{cc}
\hline
\textbf{part of speech}        & \textbf{frequency (\%)} \\
\hline
\textit{null}    & \textless 4.4  \\
common nouns           & 38.5             \\
personal pronouns     & 25               \\
proper nouns          & 15.1            \\
demonstrative pronouns & 9.9             \\
adjectives          & 3.1              \\
indefinite pronouns   & 2.1             \\
verbal nouns        & 0.8            \\
cardinal numerals       & 0.6          \\
ordinal numerals       & 0.2            \\
relative pronouns     & 0.1 \\
quantifier          & 0.1          \\
interrogative pronouns & \textless 0.1    \\
possessive pronouns    & \textless 0.1   \\
\hline
\caption{Dative absolutes in standard treebanks: subject parts of speech}
\label{dasubpos}
\end{longtable}

In comparison to the Gospels, subject parts of speech are much more varied (13 parts of speech, against 5 in the Gospels). However, many of these occur relatively rarely, and the only four standing out are common nouns, personal pronouns, proper nouns, and demonstratives. The relative order of the most frequent parts of speech in the two syntactic positions, namely pre- (Table \ref{predassubpos}) and post-matrix (Table \ref{postdassubpos}), is the same as for dative absolutes as a whole. Common nouns have a higher frequency among post-matrix absolutes than among pre-matrix ones, while the opposite is true for demonstratives. This could be due to the fact that pre-matrix participles are more likely to have their subject referring back to the immediately preceding sentence, whereas in post-matrix position, more explicit reference to the subject may be used to avoid anaphoric ambiguity (e.g. with the matrix subject).

\begin{longtable}{cc}
\hline
\textbf{part of speech}        & \textbf{frequency (\%)} \\
\hline
\textit{null}    & 4.9   \\
common nouns           & 36.0            \\
personal pronouns     & 26.0             \\
proper nouns          & 15.7           \\
demonstrative pronouns & 10.4             \\
adjectives          & 3.5            \\
indefinite pronouns   & 1.6              \\
verbal nouns        & 0.8            \\
cardinal numerals       & 0.3              \\
relative pronouns     & 0.2    \\
quantifier          & 0.2         \\
ordinal numerals       & 0.2            \\
interrogative pronouns & 0.1    \\
possessive pronouns    & 0.1   \\
\hline
\caption{Pre-matrix dative absolutes in standard treebanks: subject parts of speech}
\label{predassubpos}
\end{longtable}

\begin{table}[!h]
\centering
\begin{tabular}{cc}
\hline
\textbf{part of speech}        & \textbf{frequency (\%)} \\
\hline
\textit{null}        & 2.8          \\
common nouns           & 51.6            \\
personal pronouns     & 19.7            \\
proper nouns          & 11.7        \\
demonstrative pronouns & 5.2           \\
indefinite pronouns   & 4.2            \\
cardinal numerals       & 1.9              \\
adjectives          & 1.9             \\
verbal nouns        & 1          \\
\hline
\end{tabular}
\caption{Post-matrix dative absolutes in standard treebanks: subject parts of speech}
\label{postdassubpos}
\end{table}

Among the conjunct participles which were considered to be able to their own subject (i.e. the leftmost conjunct participle preceding the matrix clause), around half (50.7\%) were found to be null, which is not surprising considering the high percentage of null-subject conjunct participles that had been found in the previous dataset (44.8\%). 
Subject parts of speech among overt-subject conjunct participles are even more varied than among dative absolutes (18 parts of speech in total), but most of the labour is done by common nouns, proper nouns, and demonstratives.

\begin{longtable}{cc}
\hline
\textbf{part of speech}        & \textbf{frequency (\%)} \\
\hline
null & 50.7 \\
common nouns & 19.8 \\ 
proper nouns & 11.2 \\ 
demonstrative pronouns & 7.0 \\ 
personal pronouns & 3.8 \\ 
adjectives & 3.7 \\ 
indefinite pronouns & 1.4 \\ 
relative pronouns & 0.8 \\ 
verbal nouns & 0.5 \\ 
cardinal numerals & 0.5 \\ 
possessive pronouns & 0.1 \\ 
quantifier & \textless 0.1 \\ 
interrogative pronouns & \textless 0.1 \\ 
personal reflexive pronoun & \textless 0.1 \\
interrogative adverb & \textless 0.1 \\ 
possessive reflexive pronoun & \textless 0.1 \\ 
ordinal numerals & \textless 0.1 \\ 
relative adverb & \textless 0.1 \\ 
\hline
\caption[Subject parts of speech among conjunct participles that may head an overt subject in standard treebanks]{Subject parts of speech among conjunct participles that may head an overt subject in standard treebanks (i.e. leftmost conjunct participles preceding the matrix clause)}
\label{tab:tab14}
\end{longtable}

When we consider the position of the subject relative to overt-subject conjunct participles, the results are overall in line with the observations made in Chapter 1. First, inherently anaphoric parts of speech (personal pronouns and demonstratives) are much more common among SV (Table \ref{svxadvssubposshallow}) than VS conjunct participles (Table \ref{vsxadvssubposshallow}). 

\begin{table}[!h]
\centering
\begin{tabular}{cc}
\hline
\textbf{part of speech}        & \textbf{frequency} \\
\hline
common nouns   &  32.9 \\
proper nouns   &  21.9\\
demonstrative pronouns   &  19.6\\
personal pronouns   &  10.2 \\
adjectives   &   6.8 \\
indefinite pronouns   &   3.5 \\
relative pronouns   &   2.6 \\
cardinal numerals   &   1\\
verbal nouns   &   0.7 \\
possessive pronouns   &   0.38 \\
interrogative pronouns   &   0.2 \\
interrogative adverbs   &   0.1 \\
\hline
\end{tabular}
\caption{Pre-matrix SV conjunct participles in standard treebanks: subject parts of speech}
\label{svxadvssubposshallow}
\end{table}

\begin{table}[!h]
\centering
\begin{tabular}{cc}
\hline
\textbf{part of speech}        & \textbf{frequency} \\
\hline
common nouns    & 54.1 \\
proper nouns    & 24.6 \\
adjectives   &  8.6 \\
demonstrative pronouns    &  3.6 \\
personal pronouns    &  3.2 \\
indefinite pronouns    &  1.7 \\
verbal nouns    &  1.7 \\
cardinal numerals    &  1.3 \\
quantifiers    &  0.5 \\
personal reflexive pronouns    &  0.1 \\
possessive reflexive pronouns    &  0.1 \\
ordinal numerals    &  0.1 \\
possessive pronouns     & 0.1\\
\hline
\end{tabular}
\caption{Pre-matrix VS conjunct participles in standard treebanks: subject parts of speech}
\label{vsxadvssubposshallow}
\end{table}

In Chapter 1, we explained this difference with the observation that the VS configuration in conjunct participles is much more likely than the SV one to be used when reinstating older or inactive referents in the discourse, a function which (following \citealt{haug2012a}) we associated with \textsc{frames}. This would explain why referentially more explicit parts of speech, such as proper nouns or common nouns, tend to be more frequent among VS conjunct participles, whereas inherently anaphoric parts of speech, such as demonstratives and personal pronouns, are much more common among SV conjunct participles. The latter are more likely to be used when there is a topic continuation from the immediately preceding discourse when anaphoric relations are thus easier to resolve. However, we also observed that among dative absolutes, which typically function as \textsc{frames} like VS conjunct participles, the vast majority of subjects belonged to inherently anaphoric parts of speech (predominantly personal pronouns), which was tentatively explained with the fact that dative absolutes may tend to attract only very highly prominent referents as their subject. In the Gospels, the VS configuration was found to be by far the most frequent one among dative absolutes, to the extent we were not able to look at differences with the SV configuration, since occurrences of the latter were far too few. In this dataset, however, we see that the predominance of personal pronouns among VS dative absolutes was not an idiosyncrasy of the previous dataset, since this is what we also find in our standard treebanks, as Table \ref{vsdassubposshallow} shows. But we also see that among SV dative absolutes demonstratives are clearly much more frequent in the VS configuration, as Table \ref{svdassubposshallow} shows.

\begin{table}[!h]
\centering
\begin{tabular}{cc}
\hline
\textbf{part of speech}        & \textbf{frequency} \\
\hline
common nouns   & 41.7\\
proper nouns   & 19.1\\
demonstrative pronouns   & 17.5\\
personal pronouns  &  14.8\\
adjectives   &  3.3\\
indefinite pronouns   &  1.7\\
verbal nouns   &  0.5\\
relative pronouns   &  0.3\\
ordinal numerals   &  0.3\\
cardinal numerals   &  0.2\\
interrogative pronouns   &  0.2\\
quantifiers   &  0.2\\
possessive pronouns   &  0.2\\
\hline
\end{tabular}
\caption{Pre-matrix SV dative absolutes in standard treebanks: subject parts of speech}
\label{svdassubposshallow}
\end{table}

\begin{table}[!h]
\centering
\begin{tabular}{cc}
\hline
\textbf{part of speech}        & \textbf{frequency} \\
\hline
personal pronouns   &  45.4\\
common nouns   &  32\\
proper nouns  &   12.6\\
adjectives  &   4.2\\
indefinite pronouns    &  1.7\\
demonstrative pronouns   &   1.5\\
verbal nouns  &   1.2\\
cardinal pronouns   &   0.5\\
quantifiers   &   0.2\\
\hline
\end{tabular}
\caption{Pre-matrix VS dative absolutes in standard treebanks: subject parts of speech}
\label{vsdassubposshallow}
\end{table}

The constant between sentence-initial conjunct participles and dative absolutes is that demonstratives are clearly more frequent in the SV configuration, which we can easily explain with regular principles of anaphora resolution. What is peculiar, then, is the predominance of personal pronouns among VS absolutes. A possibility, already tentatively offered in the previous chapter, is that absolutes (VS absolutes, specifically, which were the vast majority there) attract very few, but very highly prominent subjects in the discourse. The lack of information-structural annotation on standard treebanks prevents us from looking at variables such as the average distance of the subject antecedents or referential activation in the previous discourse. However, we can look at the lexical variation among the subjects of sentence-initial participle constructions.\\ 
\indent Similarly to what we did in the previous Chapter for participle lemmas, Table \ref{lexvarsubshallow} shows two different metrics capturing lexical diversity among overt subjects: the percentage of subjects belonging to the 10 most frequent lemmas (10MFL) and the (moving-average) type-token ratio (MATTR) among subject lemmas (with a moving window of 40 tokens). These two metrics are calculated for the subjects of dative absolutes and conjunct participles over different subcorpora. `(O)CS' refers to (Old) Church Slavonic, (O)ES to Old East Slavic and Middle Russian texts together.  When providing the metrics for the whole dataset, `non-N' shows the numbers obtained from the data without normalizing the (O)CS and OES/MRus spellings, whereas `N' indicates that the (O)CS and OES/MRus lemma variants corresponding to the same lemma were normalized and merged into one. This is needed because some differences in orthographic conventions between the (O)CS and the OES/MRus subcorpora in the TOROT treebank would result in listing as two separate lemmas some lexical items which should be regarded as one, as in \textit{t\foreignlanguage{russian}{ъ}} `that one' according to the (O)CS convention and \textit{tyi} according to the OES/MRus one, for example. Naturally, further normalization of the two subcorpora can only result in less overall variation than no normalization, but it is still useful to be aware of the extent to which our observations may be affected.\footnote{The main changes made are: 1) \textit{je} and word-initial \textit{a} were changed to \textit{e} and \textit{ja} respectively, 2) Cyrillic characters for nasal vowels were changed to the respective east Slavic outcome (\textit{ę}/\textit{ję} > \textit{ja}, \textit{ǫ}/\textit{jǫ} > \textit{ju}), 3) \textit{ě} was changed to \textit{e}, 4) multiple variants for \textit{y} were merged (and all set to <\foreignlanguage{russian}{ы}>), 5) word-final \textit{ii} and \textit{yi} in OES texts were changed to \textit{\foreignlanguage{russian}{ь}} and  \textit{\foreignlanguage{russian}{ъ}} respectively, 6) T\textit{\foreignlanguage{russian}{ъ}}RT T\textit{\foreignlanguage{russian}{ь}}RT > TR\textit{\foreignlanguage{russian}{ъ}}T TR\textit{\foreignlanguage{russian}{ь}}T. The normalization script can be found in the data depository at \url{https://doi.org/10.6084/m9.figshare.24166254}.} The remaining rows show the lexical variation in the individual subcorpora, including different subject positions.

\begin{table}[!h]
\centering
\begin{tabular}{|c|c|c|c|c|}
\hline
\multirow{2}*{\textbf{\small Subsample}} & \multicolumn{2}{|c|}{\textbf{Absolutes}} & \multicolumn{2}{|c|}{\textbf{Conjuncts}}\\
\cline{2-5}
 & \textbf{\small 10MFL} & \textbf{\small MATTR} & \textbf{\small 10MFL}   & \textbf{\small MATTR}\\
\hline
\textbf{\small (O)CS + (O)ES + Mar} (non-N)  &   46.34\%  &   0.54  &   27.11\%  &   0.58\\
\textbf{\small (O)CS + (O)ES + Mar}(N)  &   49.85\%  &   0.53  &   29.09\%  &   0.58\\
\textbf{\small (O)CS + (O)ES} (non-N)  &   46.34\%  &   0.54  &   27.11\%  &   0.58\\
\textbf{\small (O)CS + (O)ES} (N)  &   49.85\%  &   0.53  &   29.09\%  &   0.58\\
\textbf{\small (O)CS}  &   56.58\%  &   0.54  &   29.16\%  &   0.61\\
\textbf{\small (O)ES}  &   47.73\%  &   0.53  &   36.82\%  &   0.55\\
\textbf{\small (O)CS + (O)ES + Mar} \textbf{\small (SV)}  &   48.55\%  &   0.57  &   35.69\%  &   0.58\\
\textbf{\small (O)CS + (O)ES + Mar} \textbf{\small (VS)}  &   58.97\%  &   0.48  &   22.98\%  &   0.64\\
\textbf{\small (O)CS + (O)ES (SV)}  &   48.55\%  &   0.57  &   35.69\%  &   0.58\\
\textbf{\small (O)CS + (O)ES (VS)}  &   58.97\%  &   0.48  &   22.98\%  &   0.64\\
\textbf{\small (O)CS (SV)}  &   63.31\%  &   0.55  &   36.32\%  &   0.65\\
\textbf{\small (O)CS (VS)}  &   61.21\%  &   0.51  &   30.23\%  &   0.60\\
\textbf{\small (O)ES (SV) } &   43.90\%  &   0.58  &   41.75\%  &   0.53\\
\textbf{\small (O)ES (VS)}  &   68.03\%  &   0.42  &   31.69\%  &   0.69\\
\hline
\end{tabular}
\caption[Lexical variation among subjects in standard treebanks, by subcorpus and subject position]{Lexical variation among subjects in standard treebanks, by subcorpus and subject position. \textit{10MFL} = 10 most frequent lemmas; \textit{MATTR} = moving-average type-token ratio; \textit{Mar} = Codex Marianus; \textit{(O)ES} = Old East Slavic and Middle Russian; \textit{N} = normalized; \textit{non-N} = non-normalized. Note that the closer the MATTR is to 1 (and the lower the percentage for the 10MFL), the greater the lexical variation. }
\label{lexvarsubshallow}
\end{table}

\begin{figure}[!h]
\begin{subfigure}{0.50\textwidth}
\includegraphics[width=0.9\linewidth]{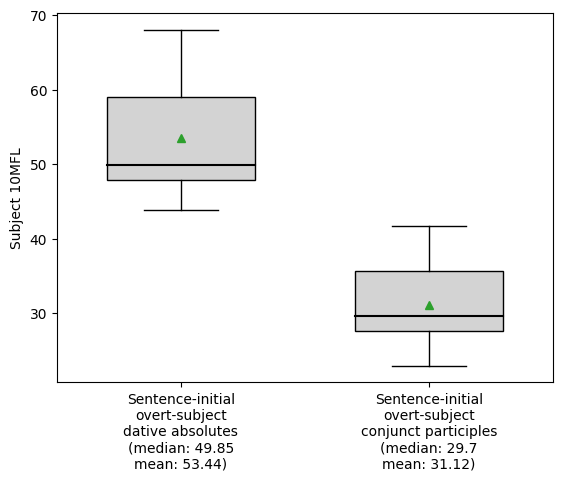}
\caption[]{}
\label{10mflsubjall}
\end{subfigure}
\begin{subfigure}{0.50\textwidth}
\includegraphics[width=0.9\linewidth]{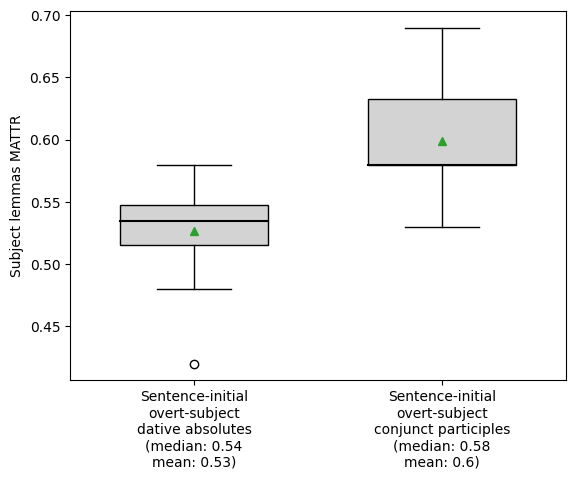}
\caption[]{}
\label{mattrssubjall}
\end{subfigure}
\caption[Lexical variation among subject lemmas in sentence-initial dative absolutes and conjunct participles]{Lexical variation among subject lemmas in sentence-initial dative absolutes and conjunct participles. (a) Average percentage of subjects belonging to the 10 most frequent lemmas (10MFL). Welch’s $t$-test: 8.94, $p$-value $<0.01$; one-tailed Mann–Whitney $U$-test: 196, $p$-value $<0.01$. (b) Average moving-average type-token ratio (MATTR). Welch’s $t$-test: -4.51, $p$-value $<0.01$; one-tailed Mann–Whitney $U$-test: 15, $p$-value $<0.01$.}
\end{figure}

As we can see by comparing the scores in Table \ref{lexvarsubshallow} before and after normalization of mixed (O)CS and OES/MRus datasets, the further normalization only yielded a slight decrease in variation according to the 10MFL value, whereas the MATTR remained the same regardless of normalization. The first clear observation is that lexical variation is overall greater among conjunct participles than dative absolutes, as Figures \ref{10mflsubjall} and \ref{mattrssubjall} show. The difference in both average MATTR and percentage of subjects belonging to the ten most frequent lemmas is statistically significant with a $p$-value < 0.01 obtained from both a Welch's $t$-test and a Mann–Whitney $U$-test. 

\begin{figure}[!h]
\begin{subfigure}{0.50\textwidth}
\includegraphics[width=0.9\linewidth]{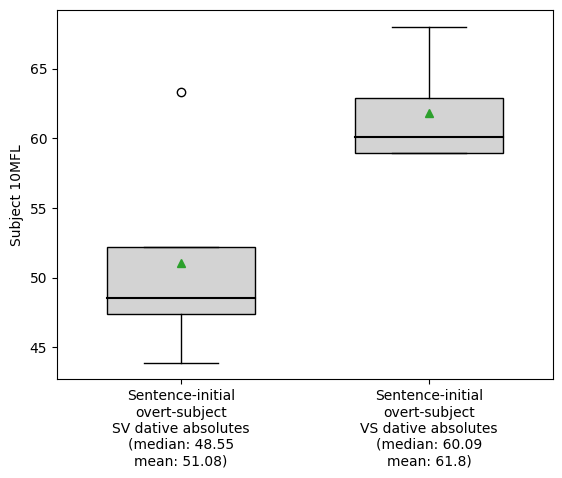}
\caption[]{}
\label{dassub_10mflvssv}
\end{subfigure}
\begin{subfigure}{0.50\textwidth}
\includegraphics[width=0.9\linewidth]{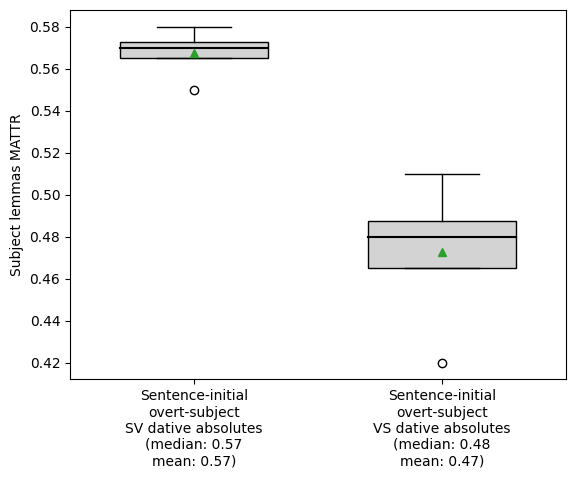}
\caption[]{}
\label{dasmattrs_vssv}
\end{subfigure}
\caption[Lexical variation among subject lemmas in sentence-initial dative absolutes in the SV and VS configuration]{Lexical variation among subject lemmas in sentence-initial dative absolutes in the SV and VS configuration. (a) Average percentage of subjects belonging to the 10 most frequent lemmas (10MFL). Welch’s $t$-test: -2.26, $p$-value $=0.04$; one-tailed Mann–Whitney $U$-test: 3, $p$-value $=0.09$. (b) Average moving-average type-token ratio (MATTR). Welch’s $t$-test: 4.77, $p$-value $<0.01$; one-tailed Mann–Whitney $U$-test: 16, $p$-value $=0.01$.}
\label{das_vssv}
\end{figure}

\begin{figure}[!h]
\begin{subfigure}{0.50\textwidth}
\includegraphics[width=0.9\linewidth]{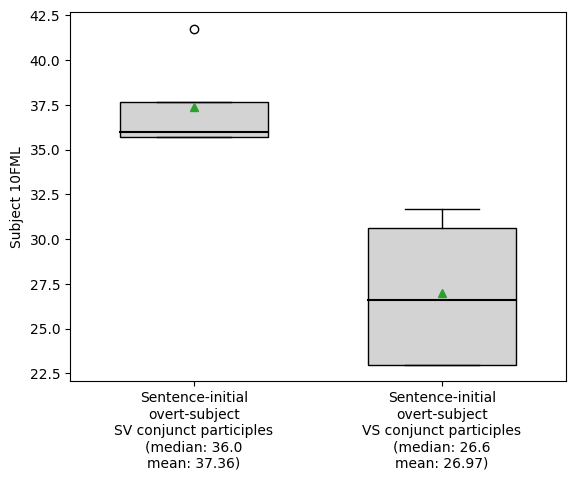}
\caption[]{}
\label{xadvs_svvs_10mfl}
\end{subfigure}
\begin{subfigure}{0.50\textwidth}
\includegraphics[width=0.9\linewidth]{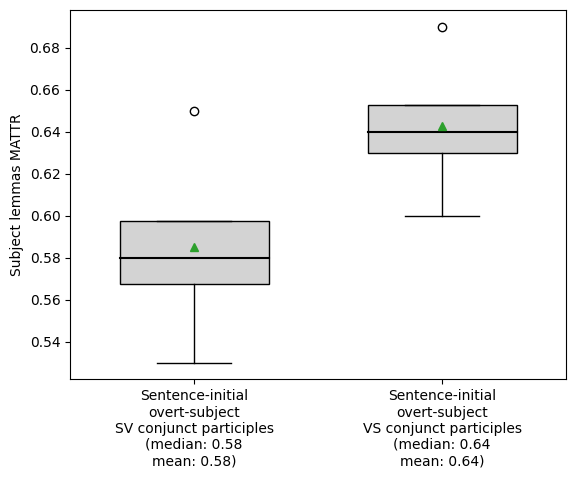}
\caption[]{}
\label{xadvs_svvs_mattrs}
\end{subfigure}
\caption[Lexical variation among subject lemmas in sentence-initial conjunct participles in the SV and VS configuration]{Lexical variation among subject lemmas in sentence-initial conjunct participles in the SV and VS configuration. (a) Average percentage of subjects belonging to the 10 most frequent lemmas (10MFL). Welch’s $t$-test: 3.78, $p<0.01$; one-tailed Mann–Whitney $U$-test: 16, $p=0.01$. (b) Average moving-average type-token ratio (MATTR). Welch’s $t$-test: -1.86, $p=0.05$; one-tailed Mann–Whitney $U$-test: 3, $p=0.09$.}
\label{xadvs_svvs}
\end{figure}

When we compare the lexical variation of subjects in sentence-initial SV and VS conjunct participles (Figure \ref{xadvs_svvs}), we see that the VS configuration shows significantly more variation than the SV configuration, whereas the opposite is true of sentence-initial dative absolutes (Figure \ref{das_vssv}).

The much lower overall lexical variation among the subjects of sentence-initial dative absolutes and the much lower lexical variation among those in the VS configuration (contrary to conjunct participles) seems to support the explanation offered in Chapter 1 that dative absolutes, particularly those in the VS configuration, tend to attract few but very prominent referents as their subjects. \\
\indent Given the heterogeneity of the corpus in terms of genres, and geographic and historical varieties, it may be useful to directly check whether there are clear differences between the most frequent subjects of sentence-initial dative absolutes and conjunct participles in the (O)CS and in the OES/MRus subcorpora among. Table \ref{lemmastopsubjshallow} shows the most frequent subject lemmas by subcorpus and position in the sentence relative to the matrix clause.

\begin{longtable}{|p{3cm}|p{5cm}|p{5cm}|}
\hline
\textbf{Subcorpus}  & \textbf{Absolutes} & \textbf{Conjuncts}\\
\hline
\textbf{(O)CS + (O)ES + Mar. (N)} & \textit{*i} `he', \textit{on\foreignlanguage{russian}{ъ}} `that one (there)', \textit{t\foreignlanguage{russian}{ъ}} `that one', \textit{s\foreignlanguage{russian}{ь}} `this one', \textit{d\foreignlanguage{russian}{ь}n\foreignlanguage{russian}{ь}} `day', \textit{bog\foreignlanguage{russian}{ъ}} `God', \textit{leto} `year', \textit{svęt\foreignlanguage{russian}{ъ}} `saint', \textit{az\foreignlanguage{russian}{ъ} }`I',  \textit{Iz\foreignlanguage{russian}{ъ}jaslav\foreignlanguage{russian}{ъ}} `Izjaslav' & \textit{on\foreignlanguage{russian}{ъ}} `that one (there)', \textit{az\foreignlanguage{russian}{ъ} } `I', \textit{cěsar\foreignlanguage{russian}{ь}} `emperor', \textit{k\foreignlanguage{russian}{ъ}nędz\foreignlanguage{russian}{ь}} `king', \textit{blažen\foreignlanguage{russian}{ъ}} `(the) blessed (one)', \textit{Iz\foreignlanguage{russian}{ъ}jaslav\foreignlanguage{russian}{ъ}} `Izjaslav', \textit{my} `we', \textit{svęt\foreignlanguage{russian}{ъ}} `saint', \textit{sam\foreignlanguage{russian}{ъ}} `self', Volodimer\foreignlanguage{russian}{ъ} `Volodymyr'\\
\hline
\textbf{(O)CS + (O)ES (N)} & \textit{*i} `he', \textit{on\foreignlanguage{russian}{ъ}} `that one (there)', \textit{t\foreignlanguage{russian}{ъ}} `that one', \textit{s\foreignlanguage{russian}{ь}} `this one', \textit{d\foreignlanguage{russian}{ь}n\foreignlanguage{russian}{ь}} `day', \textit{bog\foreignlanguage{russian}{ъ}} `God', \textit{leto} `year', \textit{svęt\foreignlanguage{russian}{ъ}} `saint', \textit{az\foreignlanguage{russian}{ъ} }`I',  \textit{Iz\foreignlanguage{russian}{ъ}jaslav\foreignlanguage{russian}{ъ}} `Izjaslav' & \textit{on\foreignlanguage{russian}{ъ}} `that one (there)', \textit{az\foreignlanguage{russian}{ъ}} `I', \textit{cěsar\foreignlanguage{russian}{ь}} `emperor', \textit{k\foreignlanguage{russian}{ъ}nędz\foreignlanguage{russian}{ь}} `king', \textit{blažen\foreignlanguage{russian}{ъ}} `(the) blessed (one)', \textit{Iz\foreignlanguage{russian}{ъ}jaslav\foreignlanguage{russian}{ъ}} `Izjaslav', \textit{my} `we', \textit{svęt\foreignlanguage{russian}{ъ}} `saint', \textit{sam\foreignlanguage{russian}{ъ}} `self', Volodiměr\foreignlanguage{russian}{ъ} `Volodymyr'\\
\hline
\textbf{(O)ES} & \textit{*i} `he', \textit{on\foreignlanguage{russian}{ъ}} `that one (there)', \textit{bog\foreignlanguage{russian}{ъ}} `God', \textit{d\foreignlanguage{russian}{ь}n\foreignlanguage{russian}{ь}} `day', \textit{Iz\foreignlanguage{russian}{ъ}jaslav\foreignlanguage{russian}{ъ}} `Izjaslav', \textit{Feodosii} `Feodosij', \textit{az\foreignlanguage{russian}{ъ}} `I', \textit{k\foreignlanguage{russian}{ъ}nędz\foreignlanguage{russian}{ь}} `king' & \textit{on\foreignlanguage{russian}{ъ}} `that one (there)', \textit{Iz\foreignlanguage{russian}{ъ}jaslav\foreignlanguage{russian}{ъ}} `Izjaslav', \textit{az\foreignlanguage{russian}{ъ}} `I', \textit{k\foreignlanguage{russian}{ъ}nędz\foreignlanguage{russian}{ь}} `king', Volodiměr\foreignlanguage{russian}{ъ} `Volodymyr', \textit{blažen\foreignlanguage{russian}{ъ}} `(the) blessed (one)', \textit{cěsar\foreignlanguage{russian}{ь}} `emperor', \textit{my} `we', \textit{sam\foreignlanguage{russian}{ъ}} `self', \textit{polov\foreignlanguage{russian}{ь}čin\foreignlanguage{russian}{ъ}} `Cuman'\\
\hline
\textbf{(O)CS} & \textit{*i} `he', \textit{t\foreignlanguage{russian}{ъ}} `that one', \textit{s\foreignlanguage{russian}{ь}} `this one', \textit{svęt\foreignlanguage{russian}{ъ}} `saint', \textit{on\foreignlanguage{russian}{ъ}} `that one (there)', \textit{leto} `year', \textit{narod\foreignlanguage{russian}{ъ}} `crowd', \textit{d\foreignlanguage{russian}{ь}n\foreignlanguage{russian}{ь}} `day', \textit{ty} `you',\textit{ čas\foreignlanguage{russian}{ъ}} `hour' & \textit{on\foreignlanguage{russian}{ъ}} `that one (there)',\textit{svęt\foreignlanguage{russian}{ъ}} `saint', \textit{cěsar\foreignlanguage{russian}{ь}} `emperor',\textit{vojevoda} `commander', \textit{k\foreignlanguage{russian}{ъ}nędz\foreignlanguage{russian}{ь}} `king', \textit{episkup\foreignlanguage{russian}{ъ}} `bishop', \textit{an\foreignlanguage{russian}{ъ}tipat\foreignlanguage{russian}{ъ}} `proconsul', \textit{az\foreignlanguage{russian}{ъ}} `I', t\foreignlanguage{russian}{ъ} `that one', \textit{blažen\foreignlanguage{russian}{ъ}} `(the) blessed (one)'\\
\hline
\textbf{(O)CS + (O)ES (SV)} & \textit{*i} `he', \textit{on\foreignlanguage{russian}{ъ}} `that one (there)', \textit{t\foreignlanguage{russian}{ъ}} `that one', \textit{s\foreignlanguage{russian}{ь}} `this one', \textit{bog\foreignlanguage{russian}{ъ}} `God', \textit{d\foreignlanguage{russian}{ь}n\foreignlanguage{russian}{ь}} `day', \textit{leto} `year',  \textit{Iz\foreignlanguage{russian}{ъ}jaslav\foreignlanguage{russian}{ъ}} `Izjaslav', \textit{Feodosii} `Feodosij', \textit{sl\foreignlanguage{russian}{ъ}n\foreignlanguage{russian}{ь}ce} `sun' & \textit{on\foreignlanguage{russian}{ъ}} `that one (there)', az\foreignlanguage{russian}{ъ} `I',  \textit{Iz\foreignlanguage{russian}{ъ}jaslav\foreignlanguage{russian}{ъ}} `Izjaslav', \textit{cěsar\foreignlanguage{russian}{ь}} `emperor', \textit{my} `we', \textit{blažen\foreignlanguage{russian}{ъ}} `(the) blessed (one)', \textit{s\foreignlanguage{russian}{ь}} `this one', \textit{sam\foreignlanguage{russian}{ъ}} `self', \textit{ty} `you', \textit{polov\foreignlanguage{russian}{ь}čin\foreignlanguage{russian}{ъ}} `Cuman' \\
\hline
\textbf{(O)CS + (O)ES (VS)} & \textit{*i} `he', az\foreignlanguage{russian}{ъ} `I', \textit{svęt\foreignlanguage{russian}{ъ}} `saint', \textit{d\foreignlanguage{russian}{ь}n\foreignlanguage{russian}{ь}} `day', \textit{leto} `year', \textit{cěsar\foreignlanguage{russian}{ь}} `emperor', \textit{ot\foreignlanguage{russian}{ь}c\foreignlanguage{russian}{ь}} `father', \textit{ty} `you', \textit{prazd\foreignlanguage{russian}{ь}nik\foreignlanguage{russian}{ъ}} `celebration, festivity', \textit{narod\foreignlanguage{russian}{ъ}} `crowd' & \textit{k\foreignlanguage{russian}{ъ}nędz\foreignlanguage{russian}{ь}} `king', \textit{cěsar\foreignlanguage{russian}{ь}} `emperor', \textit{blažen\foreignlanguage{russian}{ъ}} `(the) blessed (one)', \textit{svęt\foreignlanguage{russian}{ъ}} `saint', an\foreignlanguage{russian}{ъ}tipat\foreignlanguage{russian}{ъ} `proconsul', \textit{episkup\foreignlanguage{russian}{ъ}} `bishop', star\foreignlanguage{russian}{ь}c\foreignlanguage{russian}{ь} `old man'\\
\hline
\textbf{(O)CS (SV)} & \textit{*i} `he', \textit{on\foreignlanguage{russian}{ъ}} `that one (there)', \textit{t\foreignlanguage{russian}{ъ}} `that one', \textit{s\foreignlanguage{russian}{ь}} `this one', \textit{leto} `year',  \textit{d\foreignlanguage{russian}{ь}n\foreignlanguage{russian}{ь}} `day', \textit{narod\foreignlanguage{russian}{ъ}} `crowd', čas\foreignlanguage{russian}{ъ} `hour', \textit{bog\foreignlanguage{russian}{ъ}} `God', \textit{sluga} `servant', \textit{vrěmę} `time' & \textit{on\foreignlanguage{russian}{ъ}} `that one (there)', az\foreignlanguage{russian}{ъ} `I', \textit{cěsar\foreignlanguage{russian}{ь}} `emperor', \textit{s\foreignlanguage{russian}{ь}} `this one', \textit{svęt\foreignlanguage{russian}{ъ}} `saint', \textit{ty} `you', \textit{jedin\foreignlanguage{russian}{ъ}} `one, a certain one', \textit{t\foreignlanguage{russian}{ъ}} `that one', \textit{vojevoda} `commander', \textit{v\foreignlanguage{russian}{ь}s\foreignlanguage{russian}{ь}} `all'\\
\hline
\textbf{(O)CS (VS)} & \textit{*i} `he', \textit{svęt\foreignlanguage{russian}{ъ}} `saint', \textit{ot\foreignlanguage{russian}{ь}c\foreignlanguage{russian}{ь}} `father', \textit{ty} `you', \textit{narod\foreignlanguage{russian}{ъ}} `crowd', az\foreignlanguage{russian}{ъ} `I', \textit{Pionii} `Pionos', \textit{Iona} `Jonah', \textit{v\foreignlanguage{russian}{ь}s\foreignlanguage{russian}{ь}} `all', \textit{vrěmę} `time', an\foreignlanguage{russian}{ъ}tipat\foreignlanguage{russian}{ъ} `proconsul' & \textit{k\foreignlanguage{russian}{ъ}nędz\foreignlanguage{russian}{ь}} `king', \textit{svęt\foreignlanguage{russian}{ъ}} `saint', an\foreignlanguage{russian}{ъ}tipat\foreignlanguage{russian}{ъ} `proconsul', \textit{episkup\foreignlanguage{russian}{ъ}} `bishop', \textit{vojevoda} `commander', \textit{cěsar\foreignlanguage{russian}{ь}} `emperor', \textit{sluga} `servant', \textit{farisei} `Pharisee', \textit{mǫž\foreignlanguage{russian}{ь}} `man', star\foreignlanguage{russian}{ь}c\foreignlanguage{russian}{ь} `old man'\\
\hline
\textbf{(O)ES (SV)} & \textit{*i} `he', \textit{on\foreignlanguage{russian}{ъ}} `that one (there)', \textit{bog\foreignlanguage{russian}{ъ}} `God', \textit{Iz\foreignlanguage{russian}{ъ}jaslav\foreignlanguage{russian}{ъ}} `Izjaslav', \textit{Feodosii} `Feodosij', \textit{d\foreignlanguage{russian}{ь}n\foreignlanguage{russian}{ь}} `day', \textit{sii} `this one', M\foreignlanguage{russian}{ь}stislav\foreignlanguage{russian}{ъ} `Mstislav', \textit{k\foreignlanguage{russian}{ъ}njaz\foreignlanguage{russian}{ь}} `king', \textit{Gjurgi} `Georgij' & \textit{on\foreignlanguage{russian}{ъ}} `that one (there)', az\foreignlanguage{russian}{ъ} `I', \textit{Iz\foreignlanguage{russian}{ъ}jaslav\foreignlanguage{russian}{ъ}} `Izjaslav', \textit{my} `we', \textit{polov\foreignlanguage{russian}{ь}čin\foreignlanguage{russian}{ъ}} `Cuman', \textit{sam\foreignlanguage{russian}{ъ}} `self', Volodiměr\foreignlanguage{russian}{ъ} `Volodymyr', \textit{blažen\foreignlanguage{russian}{ъ}} `(the) blessed (one)', \textit{k\foreignlanguage{russian}{ъ}njaz\foreignlanguage{russian}{ь}} `king', \textit{cěsar\foreignlanguage{russian}{ь}} `emperor'\\
\hline
\textbf{(O)ES (VS)} & \textit{*i} `he', az\foreignlanguage{russian}{ъ} `I', \textit{d\foreignlanguage{russian}{ь}n\foreignlanguage{russian}{ь}} `day', \textit{cěsar\foreignlanguage{russian}{ь}} `emperor', \textit{leto} `year', \textit{prazd\foreignlanguage{russian}{ь}nik\foreignlanguage{russian}{ъ}} `celebration, festivity', \textit{Volodiměr\foreignlanguage{russian}{ъ}} `Volodymyr', \textit{pěšii} `pedestrian', p\foreignlanguage{russian}{ъ}lk\foreignlanguage{russian}{ъ} `troop', \textit{k\foreignlanguage{russian}{ъ}njaz\foreignlanguage{russian}{ь}} `king', \textit{bratija} `brothers, brotherhood' & \textit{k\foreignlanguage{russian}{ъ}njaz\foreignlanguage{russian}{ь}} `king', \textit{cěsar\foreignlanguage{russian}{ь}} `emperor', \textit{Volodiměr\foreignlanguage{russian}{ъ}} `Volodymyr', \textit{blažen\foreignlanguage{russian}{ъ}} `(the) blessed (one)', \textit{Iz\foreignlanguage{russian}{ъ}jaslav\foreignlanguage{russian}{ъ}} `Izjaslav', \textit{ljudie} `people', \textit{Davyd\foreignlanguage{russian}{ъ}} `David', \textit{nov\foreignlanguage{russian}{ъ}gorod\foreignlanguage{russian}{ь}c\foreignlanguage{russian}{ь}} `Novgorodian', \textit{on\foreignlanguage{russian}{ъ}} `that one (there)', \textit{bog\foreignlanguage{russian}{ъ}} `God', Jaroslav\foreignlanguage{russian}{ъ} `Jaroslav'\\
\hline
\caption[Sentence-initial dative absolutes and conjunct participles: ten most frequent subject lemmas by subcorpus and position of the subject relative to the participle]{Sentence-initial dative absolutes and conjunct participles: ten most frequent subject lemmas by subcorpus and position of the subject relative to the participle. Lemmas for the mixed (O)CS/(O)ES subcorpus are given on the basis of the normalized data. \textit{Mar.} = Codex Marianus; \textit{(O)ES} = Old East Slavic and Middle Russian.}
\label{lemmastopsubjshallow}
\end{longtable}

The pool of subject lemmas appearing among the 10 most frequent ones is very similar between absolute constructions and conjunct participles. In both cases, we see that the most frequent common nouns clearly or are likely to refer to old or accessible referents (e.g. \textit{bog\foreignlanguage{russian}{ъ}} `God', \textit{k\foreignlanguage{russian}{ъ}nędz\foreignlanguage{russian}{ь}} `king', \textit{bratija} `community (of monks)', \textit{gospod\foreignlanguage{russian}{ь}} `Lord'). Other lemmas are likely to refer to prominent subjects in a limited part of the subcorpus. Most occurrences of \textit{svęt\foreignlanguage{russian}{ъ}} ‘saint’, for instance, are found in hagiographical passages contained in the Codex Suprasliensis, where they might be expected introduce a new event related to the life of a given saint, as in the dative absolute in (\ref{saintda}) and the conjunct participle in (\ref{saintxadv}).

\begin{example}
\gll \textbf{v\foreignlanguage{russian}{ъ}š\foreignlanguage{russian}{ъ}d\foreignlanguage{russian}{ъ}šu} že \textbf{stuumu} i \textbf{stav\foreignlanguage{russian}{ъ}šu} na sǫdišti rěšę k\foreignlanguage{russian}{ъ} nemu vl\foreignlanguage{russian}{ь}svi
enter.\textsc{ptcp.pfv.m.dat.sg} \textsc{ptc} saint.\textsc{dat} and stand.\textsc{ptcp.pfv.m.dat.sg} on seat.\textsc{loc} say.\textsc{aor.3.pl} to \textsc{3.sg.dat} mage.\textsc{nom.pl}
\glt ‘When the Saint came in and stood at the seat the mages told him' (\textit{Vita of Jonah and Barachesios}, Codex Suprasliensis f. 13r) %23, 138584
\glend
\label{saintda}
\end{example}

\begin{example}
\gll i \textbf{raždeg\foreignlanguage{russian}{ъ}} sę \textbf{styi} douchom\foreignlanguage{russian}{ь} teče k\foreignlanguage{russian}{ъ} koumirou
{and} {kindle.{\sc ptcp.pfv.m.nom.sg}} {\sc refl} {saint.{\sc m.nom.sg}} {spirit.{\sc m.inst.sg}} {run.{\sc aor.3.sg}} {to} {idol.{\sc m.dat.sg}}
\glt `And kindled by the Spirit the saint ran towards the idol’ (\textit{Vita of Konon of Isauria}, Codex Suprasliensis f. 17r)
\glend
\label{saintxadv}
\end{example}

As in the Codex Marianus, one clear observation from the subjects of dative absolutes as opposed to conjunct participles is that nouns denoting times (e.g. \textit{lěto} `year' or \textit{d\foreignlanguage{russian}{ь}n\foreignlanguage{russian}{ь}} `day') and inanimate objects (e.g. \textit{dv\foreignlanguage{russian}{ь}r\foreignlanguage{russian}{ь}} `door' or \textit{sl\foreignlanguage{russian}{ъ}n\foreignlanguage{russian}{ь}ce} `Sun') are quite common, which once again points to the typical framing nature of absolute constructions, since these nouns are generally used to describe the background to a main eventuality (as in `when the day ended' or `after shutting the doors'). 
% Some of the most frequent subjects of dative absolutes, particularly post-matrix ones, are clearly those found in semi-fixed expressions or formulae that are quite common as framing expressions, e.g. \textit{nik\foreignlanguage{russian}{ъ}tože} `nobody' in clauses meaning `without anybody knowing' or `while everybody was unaware', as in (\ref{vedetifixed1}) and (\ref{vedetifixed2}) below.

\subsubsection{Summary}
In this section, the properties of subjects were analysed, focussing on dative absolutes and conjunct participles in sentence-initial position, since that is where we observe the clearest functional overlap (since both can function as \textsc{frames}) and since that is where conjunct participles were argued to be able to syntactically head a subject. We saw clear differences between the two constructions. Among sentence-initial dative absolutes, overt subjects are overwhelmingly more frequent, with only 5.6\% (39) of absolutes having a null subject. Among sentence-initial conjunct participles, overt and null subjects are almost equally frequent (50.8\% overt versus 49.2\% null). Among both overt-subject conjunct participles and dative absolutes, there are some differences between the SV and VS configurations. The subjects of conjunct participles display very similar properties to those observed in the Codex Marianus. Namely, besides common nouns, which are frequent in both configurations (albeit more so among the VS one), among the subjects of SV conjunct participles, anaphoric parts of speech (demonstratives and personal pronouns) are much more common than among VS conjunct participles. This was explained by the fact that VS conjunct participles are much more likely to be used when reintroducing older or inactive referents in the discourse, which may call for referentially more explicit parts of speech to be used. Among sentence-initial dative absolutes, demonstratives, as with conjunct participles, are much more frequent in the SV configuration, whereas personal pronouns are overwhelmingly more present in the VS one. Tentatively, this was explained by the fact that VS absolutes attract very few but highly discourse-prominent subjects. The analysis of lexical variation among subjects showed that the subjects of sentence-initial dative absolutes display overall significantly smaller lexical variation than conjunct participles but also that subjects in VS dative absolutes are much less lexically varied than those in the SV configuration. \\
\subsection{Lexical variation among participles}\label{lexicalvarptcpshallowsec}

The pattern observed regarding lexical variation among the subjects of conjunct participles and dative absolutes can, to some extent, also be observed in the lexical variation among the participles themselves. Table \ref{lexvarptcp} shows the number of participles belonging to the ten most frequent lemmas (10MFL) and the MATTR for all participle lemmas at a moving window of 40 tokens, divided into subcorpora and position in the sentence relative to the matrix clause.

\begin{longtable}{|c|c|c|c|c|}
\hline
\multirow{2}*{\textbf{\small Subsample}} & \multicolumn{2}{|c|}{\textbf{Absolutes}} & \multicolumn{2}{|c|}{\textbf{Conjuncts}}\\
\cline{2-5}
 & \textbf{\small 10MFL} & \textbf{\small MATTR} & \textbf{\small 10MFL}   & \textbf{\small MATTR}\\
\hline
\textbf{\footnotesize (O)CS + (O)ES + Mar (non-N)}	&	36.15\%	&	0.68	&	23.20\%	&	0.78\\ 
\textbf{\footnotesize (O)CS + (O)ES + Mar (N)}	&	38.53\%	&	0.68	&	26.78\%	&	0.78\\ 
\textbf{\small (O)CS + (O)ES (non-N)} &	36.05\%	&	0.68	&	23.07\%	&	0.79\\ 
\textbf{\small (O)CS + (O)ES (N)}	&	38.46\%	&	0.68	&	26.38\%	&	0.79\\ 
\textbf{\small (O)CS + Mar}	&	42.74\%	&	0.66	&	27.41\%	&	0.78\\ 
\textbf{\small (O)CS}	&	43.31\%	&	0.66	&	26.92\%	&	0.80\\ 
\textbf{\small (O)ES}	&	37.93\%	&	0.69	&	26.38\%	&	0.79\\ 
\textbf{\small Mar}	&	50.54\%	&	0.64	&	36.79\%	&	0.72\\ 
\hline
\textbf{\small (O)CS + (O)ES + Mar (SV)} & 40.97\% & 0.70 & 33.11\% & 0.76\\ 
\textbf{\small (O)CS + (O)ES + Mar (VS)}& \textbf{37.99\%} & \textbf{0.67} & \textbf{53.55\%}& \textbf{0.55}\\ 
\textbf{\small (O)CS + (O)ES (SV)}& 39.24\% & 0.70 & 32.41\% & 0.77\\ 
\textbf{\small (O)CS + (O)ES (VS)} &\textbf{39.74\%} & \textbf{0.68} & \textbf{54.51\%} & \textbf{0.57}\\ 
\textbf{\small (O)CS (SV)} & 54.68\% & 0.62 & 36.04\% & 0.75\\ 
\textbf{\small (O)CS (VS)} & \textbf{38.79\%} & \textbf{0.74} & \textbf{50.17\%} & \textbf{0.61}\\ 
\textbf{\small (O)ES (SV)} & 35.12\% & 0.76 & 31.86\% & 0.77\\ 
\textbf{\small (O)ES (VS)} & \textbf{52.38\%} & \textbf{0.61} & \textbf{65.27\%} & \textbf{0.51}\\ 
\textbf{\small Mar (SV)} & \textit{NA} & \textit{NA} & 47.34\% & 0.70\\ 
\textbf{\small Mar (VS)} & \textbf{45.83\%} & \textbf{0.64} &\textbf{ 64.71\%} & \textbf{0.50}\\ 
\hline
\textbf{\footnotesize (O)CS + (O)ES + Mar (left, N)}	&	39.83\%	&	0.68	&	28.42\%	&	0.77\\ 
\textbf{\footnotesize (O)CS + (O)ES + Mar (right, N)} &	40.71\%	&	0.71	&	32.00\%	&	0.74\\ 
\textbf{\small (O)CS + (O)ES (left, N)}&	39.36\%	&	0.68	&	26.88\%	&	0.79\\ 
\textbf{\small (O)CS + (O)ES (right, N)}	&	41.31\%	&	0.71	&	32.95\%	&	0.74\\ 
\textbf{\small (O)CS (left)}	&	44.22\%	&	0.67	&	27.36\%	&	0.79\\ 
\textbf{\small (O)CS (right)}	&	45.78\%	&	0.72	&	33.76\%	&	0.74\\ 
\textbf{\small (O)ES (left)}	&	39.47\%	&	0.70	&	27.06\%	&	0.78\\ 
\textbf{\small (O)ES (right)}	&	46.92\%	&	0.68	&	32.84\%	&	0.73\\ 
\textbf{\small Mar (left)}	&	51.46\%	&	0.64	&	44.12\%	&	0.67\\ 
\textbf{\small Mar (right)}	&	\textit{NA}	&	\textit{NA}	&	30.11\%	&	0.81\\
\hline
\caption[Lexical variation among participles by subcorpus]{Lexical variation among participles by subcorpus. \textit{10MFL} = ten most frequent lemmas; \textit{MATTR} = moving-average type-token ratio; \textit{Mar} = Codex Marianus; \textit{left} and \textit{right} = pre-matrix and post matrix, respectively; \textit{N} = normalized; \textit{non-N} = non-normalized. Note that the closer the MATTR is to 1 (and the lower the percentage for the 10MFL), the greater the lexical variation. Bold = configurations where conjunct participles show a lower degree of lexical variation than dative absolutes.}
\label{lexvarptcp}
\end{longtable}

\begin{figure}[!h]
\begin{subfigure}{0.50\textwidth}
\includegraphics[width=0.9\linewidth]{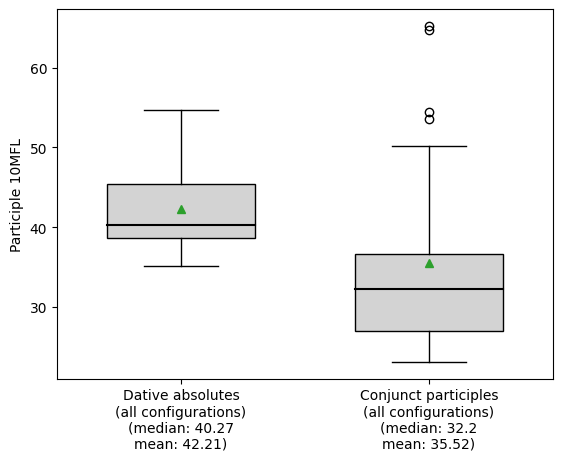}
\caption[]{}
\label{10mfl_ptcp}
\end{subfigure}
\begin{subfigure}{0.50\textwidth}
\includegraphics[width=0.9\linewidth]{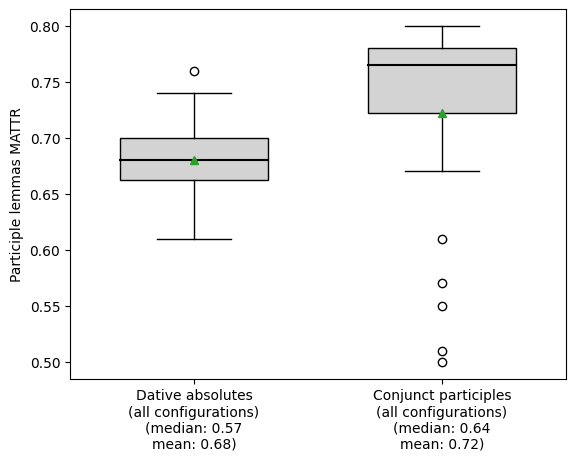}
\caption[]{}
\label{mattrs_ptcp}
\end{subfigure}
\caption[Lexical variation among participles, any configuration]{Lexical variation among participles, any configuration. (a) Average percentage of participles belonging to the ten most frequent lemmas (10MFL). Welch’s $t$-test: 2.56, $p$-value $=0.01$; one-tailed Mann–Whitney $U$-test: 530, $p$-value $<0.01$. (b) Average moving-average type-token ratio (MATTR) among lemmas. Welch’s $t$-test: -2.16, $p$-value $=0.03$; one-tailed Mann–Whitney $U$-test: 158, $p$-value $<0.01$.}
\label{lexvarptcpshall}
\end{figure}

Lexical variation among conjunct participles is almost always consistently greater than among dative absolutes across subcorpora and positions in the sentence. This is not surprising, considering what we have observed regarding the typical framing character of dative absolutes, as opposed to the varied functions of conjunct participles. As argued in Chapter 1 (following \pgcitet{haug2012a}{320}), \textsc{frames} are assumed to be more ‘predictable’, since they are presupposed (ana\-phoric or accommodated), whereas \textsc{elaborations} and \textsc{independent rhemes} typically introduce new information, so we do expect to find less lexical variation among \textsc{frames} than participle constructions with other functions. As Figures \ref{10mfl_ptcp} and \ref{mattrs_ptcp} show, both mean and median 10MFL/MATTR for conjunct participles are well beyond the midspread of the values for dative absolutes (they are in fact well beyond the whiskers of the plot for the latter, meaning that their means/medians are beyond the maximum or minimum observed values for dative absolutes). The difference is significant according to all tests ($p$-value $<0.01$).

\begin{figure}[!h]
\begin{subfigure}{0.50\textwidth}
\includegraphics[width=0.9\linewidth]{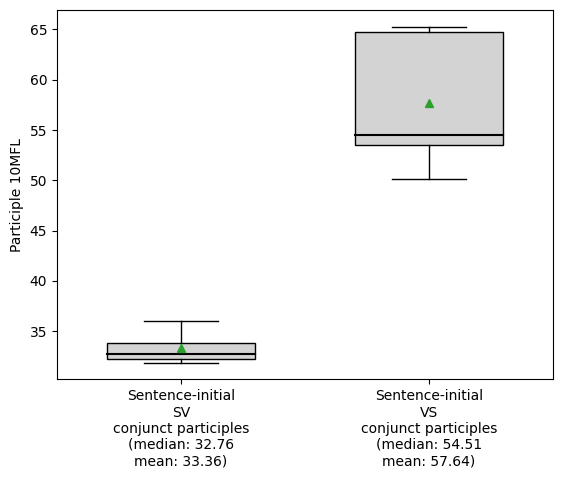}
\caption[]{}
\label{XADVS_lexvar_SVVS_10MFL}
\end{subfigure}
\begin{subfigure}{0.50\textwidth}
\includegraphics[width=0.9\linewidth]{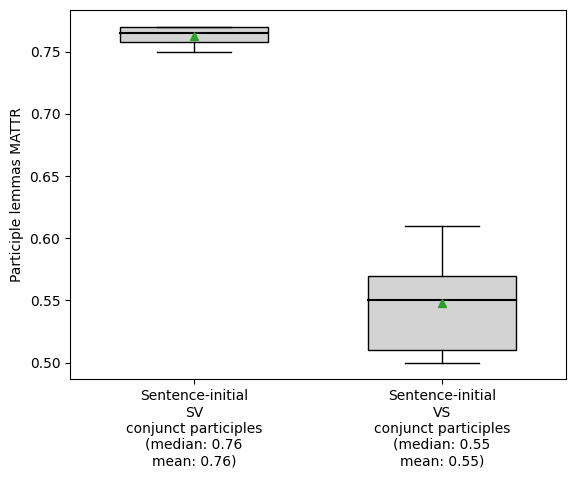}
\caption[]{}
\label{XADVS_lexvar_SVVS_MATTR}
\end{subfigure}
\caption[Lexical variation among participles in sentence-initial conjunct participles (SV versus VS)]{Lexical variation among participles in sentence-initial participles, SV versus VS configuration. (a) Average percentage of participles belonging to the 10 most frequent lemmas (10MFL). Welch’s $t$-test: -7.53, $p$-value $<0.01$; one-tailed Mann–Whitney $U$-test: 0, $p$-value $<0.01$. (b) Average moving-average type-token ratio (MATTR) among lemmas. Welch’s $t$-test: 10.38, $p$-value $<0.01$; one-tailed Mann–Whitney $U$-test: 20, $p$-value $<0.01$.}
\label{XADVS_lexvar_SVVS}
\end{figure}

As in the Gospels, the level of lexical variation among conjunct participles is very different depending on the position of the subject relative to the participle, with the conjunct participles in the VS configuration showing much less lexical variation according to both tests, which could, again, be associated with the clearer \textsc{frame} function of sentence-initial conjunct participles. The difference is highly significant according to both tests (Figure \ref{XADVS_lexvar_SVVS}).

\begin{figure}[!h]
\begin{subfigure}{0.50\textwidth}
\includegraphics[width=0.9\linewidth]{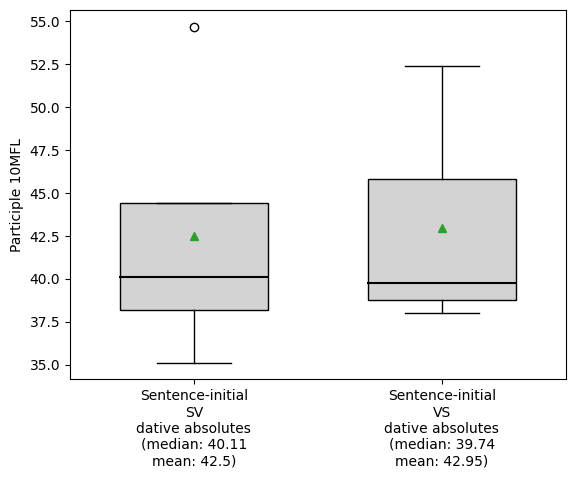}
\caption[]{}
\label{DAS_lexvar_SVVS_10MFL}
\end{subfigure}
\begin{subfigure}{0.50\textwidth}
\includegraphics[width=0.9\linewidth]{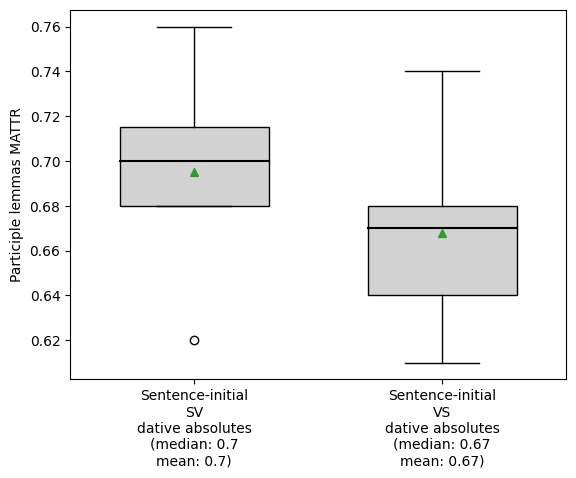}
\caption[]{}
\label{DAS_lexvar_SVVS_MATTR}
\end{subfigure}
\caption[Lexical variation among participles in sentence-initial dative absolutes (SV versus VS)]{Lexical variation among participles in sentence-initial dative absolutes (SV versus VS). (a) Average percentage of participles belonging to the 10 most frequent lemmas (10MFL). Welch’s $t$-test: -0.08, $p$-value $=0.93$; one-tailed Mann–Whitney $U$-test: 10, $p$-value $1$. (b) Average moving-average type-token ratio (MATTR) among lemmas. Welch’s $t$-test: 0.74, $p$-value $=0.48$; one-tailed Mann–Whitney $U$-test: 14, $p$-value $=0.19$.}
\label{DAS_lexvar_SVVS}
\end{figure}

A different situation is observed instead among sentence-initial dative absolutes: 
there is no significant difference in lexical variation between dative absolutes in the SV and in the VS configuration, as Figure \ref{DAS_lexvar_SVVS} shows.

Overall, this result is also not surprising in itself. As we have already observed when looking at the properties of subjects above and in the previous Chapter (Section \ref{deep-subj}), the principles driving the choice of VS and SV configuration in sentence-initial dative absolutes and conjunct participles are clearly not identical. In light of all the results obtained so far, it is clear that sentence-initial dative absolutes (and, in fact, \textit{most} participles in any configuration) have relatively consistent functions throughout.
% Judging from the type of lexical variation that the two scores are likely to capture, it would seem that, when we consider the most frequent participle lemmas among the totality of occurrences of dative absolutes, there is no difference between dative absolutes in the SV and VS configurations, which is what the 10MFL score is more likely to capture. On the other hand, when we consider a window of occurrences at a time (using the MATTR score), we also partly capture groups of dative absolutes that occur closer to one another in specific texts (although, of course, the window stretches \textit{across} texts as well). In this case, we observe significantly more lexical variation among VS dative absolutes. 
In Table \ref{lexvarptcp}, rows in bold indicate when variation is smaller among conjunct participles than dative absolutes: in all cases, these are VS configurations. Given the consistently small lexical variation among dative absolutes and the clear correlation between subject position and lexical variation among conjunct participles, it seems evident that, even beyond the Gospels, VS conjunct participles are very likely to function as \textsc{frames} and SV ones more as either \textsc{independent rhemes} or \textsc{elaborations}, depending on the position relative to the matrix clause, whereas dative absolutes tend to have a consistently small lexical variation throughout, reflecting their typical framing function. \\
\indent Table \ref{mostfreqlemmaptcp} show the most frequent lemmas among dative absolutes and conjunct participles, divided by subcorpus.

\begin{longtable}{|p{3cm}|p{5cm}|p{5cm}|}
\hline
\textbf{Subcorpus} & \textbf{Absolutes} & \textbf{Conjuncts}\\
\hline
\textbf{(O)CS + (O)ES + Mar. (N)}	&	\textit{byti} `be', \textit{priti} `come, arrive', \textit{glagolati} `speak, say', \textit{iti} `go', \textit{chotěti} `want', \textit{rešti} `say', \textit{sěděti} `sit', \textit{moliti sę} `pray', \textit{minǫti} `pass', \textit{v\foreignlanguage{russian}{ъ}niti} `go in'	&	\textit{glagolati} `speak, say', \textit{viděti} `see', \textit{byti} `be', \textit{slyšati} `hear', \textit{priti} `come, arrive', \textit{rešti} `say', \textit{iti} `go', \textit{chotěti} `want', \textit{prijęti} `take, receive', \textit{v\foreignlanguage{russian}{ъ}zęti} `take' \\ 
\hline
\textbf{(O)CS + (O)ES (N)}	&	\textit{byti} `be', \textit{priti} `come, arrive', \textit{glagolati} `speak, say', \textit{chotěti} `want', \textit{rešti} `say', \textit{iti} `go', \textit{sěděti} `sit', \textit{moliti sę} `pray', \textit{minǫti} `pass', \textit{stojati} ‘stand’	&	\textit{glagolati} `speak, say', \textit{byti} `be', \textit{viděti} `see', \textit{slyšati} `hear', \textit{rešti} `say', \textit{priti} `come, arrive', \textit{chotěti} `want', \textit{iti} `go', \textit{v\foreignlanguage{russian}{ъ}zęti} `take', \textit{prijęti} `take, receive' \\ 
\hline
\textbf{(O)CS + Mar.}	&	\textit{byti} `be', \textit{glagolati} `speak, say', \textit{priti} `come, arrive', \textit{rešti} `say', \textit{v\foreignlanguage{russian}{ъ}niti} `go in', \textit{iti} `go', \textit{minǫti} `pass', \textit{chotěti} `want', \textit{moliti sę} `pray', \textit{doiti} `arrive'	&	\textit{glagolati} `speak, say', \textit{viděti} `see', \textit{byti} `be', \textit{slyšati} `hear', \textit{priti} `come, arrive', \textit{iti} `go', \textit{prijęti} `take, receive', \textit{chotěti} `want', \textit{iměti} `have', \textit{s\foreignlanguage{russian}{ъ}tvoriti} `make' \\ 
\hline
\textbf{(O)CS}	&	\textit{byti} `be', \textit{priti} `come, arrive', \textit{glagolati} `speak, say', \textit{rešti} `say', \textit{chotěti} `want', \textit{doiti} `arrive', \textit{minǫti} `pass', \textit{moliti sę} `pray', \textit{iti} `go', \textit{kon\foreignlanguage{russian}{ь}čati} `finish'	&	\textit{glagolati} `speak, say', \textit{byti} `be', \textit{viděti} `see', \textit{slyšati} `hear', \textit{chotěti} `want', \textit{iti} `go', \textit{priti} `come, arrive', \textit{prijęti} `take, receive', \textit{v\foreignlanguage{russian}{ъ}zęti} `take', \textit{s\foreignlanguage{russian}{ъ}tvoriti} `make' \\ 
\hline
\textbf{(O)ES}	&	\textit{byti} `be', \textit{priiti} `come, arrive', \textit{chotěti} `want', \textit{iti} `go', \textit{sěděti} `sit', \textit{prispěti} `rush, happen', \textit{stojati} ‘stand’, \textit{žiti} `live', \textit{rešti} `say', \textit{nastati} `begin'	&	\textit{glagolati} `speak, say', \textit{rešti} `say', \textit{slyšati} `hear', \textit{viděti} `see', \textit{priiti} `come, arrive', \textit{byti} `be', \textit{chotěti} `want', \textit{iti} `go', \textit{v\foreignlanguage{russian}{ъ}zęti} `take', \textit{iměti} `have' \\ 
\hline
\textbf{Mar.}	&	\textit{byti} `be', \textit{glagolati} `speak, say', \textit{v\foreignlanguage{russian}{ъ}niti} `go in', \textit{iti} `go', \textit{is\foreignlanguage{russian}{ъ}choditi} `exit, go out', \textit{v\foreignlanguage{russian}{ъ}zležati} `lay down', \textit{jasti} `eat', \textit{priti} `come, arrive', \textit{iziti} `go down', \textit{iměti} `have'	&	\textit{viděti} `see', \textit{priti} `come, arrive', \textit{slyšati} `hear', \textit{iti} `go', \textit{pristǫpiti} `ascend, approach', \textit{iziti} `go down', \textit{prijęti} `take, receive', \textit{v\foreignlanguage{russian}{ъ}stati} `stand up', \textit{byti} `be', \textit{v\foreignlanguage{russian}{ъ}z\foreignlanguage{russian}{ь}rěti} `look up at' \\ 
\hline
\textbf{(O)CS + (O)ES + Mar. (SV)}	&	\textit{byti} `be', \textit{glagolati} `speak, say', \textit{priti} `come, arrive', \textit{rešti} `say', \textit{chotěti} `want', \textit{moliti sę} `pray', \textit{minǫti} `pass', \textit{iti} `go', \textit{pomagati} `help', \textit{sěděti} `sit'	&	\textit{glagolati} `speak, say', \textit{viděti} `see', \textit{byti} `be', \textit{slyšati} `hear', \textit{rešti} `say', \textit{chotěti} `want', \textit{priti} `come, arrive', \textit{iti} `go', \textit{iměti} `have', \textit{prijęti} `take, receive' \\ 
\hline
\textbf{(O)CS + (O)ES + Mar. (VS)}	&	\textit{byti} `be', \textit{priti} `come, arrive', \textit{iti} `go', \textit{sěděti} `sit', \textit{glagolati} `speak, say', \textit{chotěti} `want', \textit{v\foreignlanguage{russian}{ъ}niti} `go in', \textit{cěsar\foreignlanguage{russian}{ь}stvovati} `rule, serve as emperor', \textit{doiti} `arrive', \textit{umrěti} `die'	&	\textit{slyšati} `hear', \textit{viděti} `see', \textit{priti} `come, arrive', \textit{iti} `go', \textit{pristǫpiti} `ascend, approach', \textit{v\foreignlanguage{russian}{ъ}stati} `stand up', \textit{byti} `be', \textit{prijęti} `take, receive', \textit{iziti} `go down', \textit{ot\foreignlanguage{russian}{ъ}věštati} `reply' \\ 
\hline
\textbf{(O)CS + (O)ES (SV)}	&	\textit{byti} `be', \textit{priti} `come, arrive', \textit{glagolati} `speak, say', \textit{rešti} `say', \textit{chotěti} `want', \textit{iti} `go', \textit{moliti sę} `pray', \textit{minǫti} `pass', \textit{pomagati} `help', \textit{sěděti} `sit'	&	\textit{glagolati} `speak, say', \textit{byti} `be', \textit{viděti} `see', \textit{rešti} `say', \textit{slyšati} `hear', \textit{chotěti} `want', \textit{iti} `go', \textit{priti} `come, arrive', \textit{iměti} `have', \textit{v\foreignlanguage{russian}{ъ}zęti} `take' \\ 
\hline
\textbf{(O)CS + (O)ES (VS)}	&	\textit{byti} `be', \textit{priti} `come, arrive', \textit{iti} `go', \textit{sěděti} `sit', \textit{chotěti} `want', \textit{glagolati} `speak, say', \textit{cěsar\foreignlanguage{russian}{ь}stvovati} `rule, serve as emperor', \textit{doiti} `arrive', \textit{umrěti} `die', \textit{moliti sę} `pray'	&	\textit{slyšati} `hear', \textit{viděti} `see', \textit{priti} `come, arrive', \textit{ot\foreignlanguage{russian}{ъ}věštati} `reply', \textit{byti} `be', \textit{razgněvati} `anger', \textit{iti} `go', \textit{uvěděti} `discover', \textit{v\foreignlanguage{russian}{ъ}zęti} `take', \textit{prijęti} `take, receive' \\ 
\hline
\textbf{(O)CS (SV)}	&	\textit{byti} `be', \textit{glagolati} `speak, say', \textit{rešti} `say', \textit{priti} `come, arrive', \textit{chotěti} `want', \textit{minǫti} `pass', \textit{umrěti} `die', \textit{moliti sę} `pray', \textit{stati} `become', \textit{kon\foreignlanguage{russian}{ь}čati} `finish'	&	\textit{glagolati} `speak, say', \textit{byti} `be', \textit{viděti} `see', \textit{slyšati} `hear', \textit{chotěti} `want', \textit{iměti} `have', \textit{prijęti} `take, receive', \textit{v\foreignlanguage{russian}{ъ}zęti} `take', \textit{priti} `come, arrive', \textit{iti} `go' \\ 
\hline
\textbf{(O)CS (VS)}	&	\textit{byti} `be', \textit{priti} `come, arrive', \textit{glagolati} `speak, say', \textit{doiti} `arrive', \textit{cěsar\foreignlanguage{russian}{ь}stvovati} `rule, serve as emperor', \textit{rešti} `say', \textit{iti} `go', \textit{moliti sę} `pray', \textit{v\foreignlanguage{russian}{ъ}niti} `go in', \textit{ležati} `lie, recline'	&	\textit{slyšati} `hear', \textit{viděti} `see', \textit{razgněvati} `anger', \textit{ot\foreignlanguage{russian}{ъ}věštati} `reply', \textit{byti} `be', \textit{priti} `come, arrive', \textit{iti} `go', \textit{v\foreignlanguage{russian}{ъ}zęti} `take', \textit{izlěsti} `crawl, climb out', \textit{uvěděti} `discover' \\ 
\hline
\textbf{(O)ES (SV)}	&	\textit{byti} `be', \textit{priiti} `come, arrive', \textit{chotěti} `want', \textit{rešti} `say', \textit{pomagati} `help', \textit{sěděti} `sit', \textit{iti} `go', \textit{prispěti} `rush, happen', \textit{stojati} ‘stand’, \textit{žiti} `live'	&	\textit{glagolati} `speak, say', \textit{rešti} `say', \textit{viděti} `see', \textit{slyšati} `hear', \textit{chotěti} `want', \textit{byti} `be', \textit{iti} `go', \textit{priiti} `come, arrive', \textit{prositi} `beg, ask', \textit{v\foreignlanguage{russian}{ъ}zęti} `take' \\ 
\hline
\textbf{(O)ES (VS)}	&	\textit{byti} `be', \textit{priiti} `come, arrive', \textit{iti} `go', \textit{sěděti} `sit', \textit{chotěti} `want', \textit{nastati} `begin', \textit{prispěti} `rush, happen', \textit{běžati} `run', \textit{umrěti} `die', \textit{minǫti} `pass'	&	\textit{slyšati} `hear', \textit{viděti} `see', \textit{priiti} `come, arrive', \textit{iti} `go', \textit{uvěděti} `discover', \textit{uslyšati} `hear', \textit{ot\foreignlanguage{russian}{ъ}věštati} `reply', \textit{v\foreignlanguage{russian}{ъ}zęti} `take', \textit{s\foreignlanguage{russian}{ъ}dumati} `think', \textit{priěchati} `come' \\ 
\hline
\textbf{Mar. (SV)}	&	\textit{glagolati} `speak, say', \textit{byti} `be', \textit{zatvoriti} `close', \textit{is\foreignlanguage{russian}{ъ}choditi} `exit, go out', \textit{četvr\foreignlanguage{russian}{ь}tovlast\foreignlanguage{russian}{ь}stvovati} `serve as a tetrarch', \textit{v\foreignlanguage{russian}{ъ}sijati} `begin to shine', \textit{čuditi} `marvel', \textit{plęsati} `dance', \textit{v\foreignlanguage{russian}{ъ}zležati} `lay down', \textit{ugoditi} `please, satisfy'	&	\textit{viděti} `see', \textit{slyšati} `hear', \textit{byti} `be', \textit{priti} `come, arrive', \textit{učiti} `learn, teach', \textit{iměti} `have', \textit{iti} `go', \textit{věděti} `know', \textit{prijęti} `take, receive', \textit{iskušati} `tempt, try' \\ 
\hline
\textbf{Mar. (VS)}	&	\textit{byti} `be', \textit{v\foreignlanguage{russian}{ъ}niti} `go in', \textit{glagolati} `speak, say', \textit{iti} `go', \textit{iziti} `go down', \textit{priti} `come, arrive', \textit{v\foreignlanguage{russian}{ъ}zležati} `lay down', \textit{is\foreignlanguage{russian}{ъ}choditi} `exit, go out', \textit{jasti} `eat', \textit{s\foreignlanguage{russian}{ъ}choditi} `go down'	&	\textit{viděti} `see', \textit{iti} `go', \textit{priti} `come, arrive', \textit{pristǫpiti} `ascend, approach', \textit{slyšati} `hear', \textit{iziti} `go down', \textit{v\foreignlanguage{russian}{ъ}stati} `stand up', \textit{prijęti} `take, receive', \textit{v\foreignlanguage{russian}{ъ}niti} `go in', \textit{v\foreignlanguage{russian}{ъ}z\foreignlanguage{russian}{ь}rěti} `look up at' \\ 
\hline
\textbf{(O)CS + (O)ES + Mar. (left-N)}	&	\textit{byti} `be', \textit{priti} `come, arrive', \textit{glagolati} `speak, say', \textit{iti} `go', \textit{chotěti} `want', \textit{rešti} `say', \textit{sěděti} `sit', \textit{minǫti} `pass', \textit{v\foreignlanguage{russian}{ъ}niti} `go in', \textit{moliti sę} `pray'	&	\textit{viděti} `see', \textit{slyšati} `hear', \textit{byti} `be', \textit{priti} `come, arrive', \textit{iti} `go', \textit{v\foreignlanguage{russian}{ъ}zęti} `take', \textit{prijęti} `take, receive', \textit{v\foreignlanguage{russian}{ъ}stati} `stand up', \textit{s\foreignlanguage{russian}{ъ}tvoriti} `make', \textit{chotěti} `want' \\ 
\hline
\textbf{(O)CS + (O)ES + Mar. (right-N)}	&	\textit{byti} `be', \textit{glagolati} `speak, say', \textit{pomagati} `help', \textit{sěděti} `sit', \textit{cěsar\foreignlanguage{russian}{ь}stvovati} `rule, serve as emperor', \textit{moliti sę} `pray', \textit{chotěti} `want', \textit{stojati} ‘stand’, \textit{s\foreignlanguage{russian}{ъ}pati} `sleep', \textit{bežati} `run'	&	\textit{glagolati} `speak, say', \textit{rešti} `say', \textit{chotěti} `want', \textit{byti} `be', \textit{iměti} `have', \textit{prositi} `beg, ask', \textit{moliti sę} `pray', \textit{tvoriti} `make', \textit{viděti} `see', \textit{slaviti} `praise' \\ 
\hline
\textbf{(O)CS + (O)ES (left-N)}	&	\textit{byti} `be', \textit{priti} `come, arrive', \textit{chotěti} `want', \textit{glagolati} `speak, say', \textit{iti} `go', \textit{rešti} `say', \textit{sěděti} `sit', \textit{minǫti} `pass', \textit{moliti sę} `pray', \textit{v\foreignlanguage{russian}{ъ}niti} `go in'	&	\textit{viděti} `see', \textit{slyšati} `hear', \textit{byti} `be', \textit{priti} `come, arrive', \textit{iti} `go', \textit{v\foreignlanguage{russian}{ъ}zęti} `take', \textit{prijęti} `take, receive', \textit{s\foreignlanguage{russian}{ъ}tvoriti} `make', \textit{chotěti} `want', \textit{v\foreignlanguage{russian}{ъ}stati} `stand up' \\ 
\hline
\textbf{(O)CS + (O)ES (right-N)}	&	\textit{byti} `be', \textit{glagolati} `speak, say', \textit{pomagati} `help', \textit{sěděti} `sit', \textit{cěsar\foreignlanguage{russian}{ь}stvovati} `rule, serve as emperor', \textit{moliti sę} `pray', \textit{bežati} `run', \textit{rešti} `say', \textit{stojati} ‘stand’, \textit{v\foreignlanguage{russian}{ъ}schoditi} `go up'	&	\textit{glagolati} `speak, say', \textit{rešti} `say', \textit{chotěti} `want', \textit{byti} `be', \textit{iměti} `have', \textit{tvoriti} `make', \textit{prositi} `beg, ask', \textit{moliti sę} `pray', \textit{viděti} `see', \textit{slaviti} `praise' \\ 
\hline
\textbf{(O)CS (left)}	&	\textit{byti} `be', \textit{priti} `come, arrive', \textit{glagolati} `speak, say', \textit{rešti} `say', \textit{chotěti} `want', \textit{iti} `go', \textit{doiti} `arrive', \textit{minǫti} `pass', \textit{v\foreignlanguage{russian}{ъ}niti} `go in', \textit{kon\foreignlanguage{russian}{ь}čati} `finish'	&	\textit{byti} `be', \textit{viděti} `see', \textit{slyšati} `hear', \textit{priti} `come, arrive', \textit{iti} `go', \textit{v\foreignlanguage{russian}{ъ}zęti} `take', \textit{prijęti} `take, receive', \textit{s\foreignlanguage{russian}{ъ}tvoriti} `make', \textit{chotěti} `want', \textit{ot\foreignlanguage{russian}{ъ}věštati} `reply' \\ 
\hline
\textbf{(O)CS (right)}	&	\textit{byti} `be', \textit{glagolati} `speak, say', \textit{cěsar\foreignlanguage{russian}{ь}stvovati} `rule, serve as emperor', \textit{podajati} `give', \textit{umrěti} `die', \textit{moliti sę} `pray', \textit{rešti} `say', \textit{zatvoriti} `close', \textit{s\foreignlanguage{russian}{ъ}kon\foreignlanguage{russian}{ь}čavati} `finish', \textit{s\foreignlanguage{russian}{ъ}v\foreignlanguage{russian}{ъ}prašati} `ask'	&	\textit{glagolati} `speak, say', \textit{byti} `be', \textit{chotěti} `want', \textit{iměti} `have', \textit{viděti} `see', \textit{rešti} `say', \textit{tvoriti} `make', \textit{moliti sę} `pray', \textit{prijęti} `take, receive', \textit{slaviti} `praise' \\ 
\hline
\textbf{(O)ES (left)}	&	\textit{byti} `be', \textit{priiti} `come, arrive', \textit{chotěti} `want', \textit{iti} `go', \textit{sěděti} `sit', \textit{nastati} `begin', \textit{prispěti} `rush, happen', \textit{žiti} `live', \textit{stojati} ‘stand’, \textit{minǫti} `pass'	&	\textit{slyšati} `hear', \textit{viděti} `see', \textit{priiti} `come, arrive', \textit{iti} `go', \textit{byti} `be', \textit{v\foreignlanguage{russian}{ъ}zęti} `take', \textit{v\foreignlanguage{russian}{ъ}stati} `stand up', \textit{pojęti} `take', \textit{s\foreignlanguage{russian}{ъ}tvoriti} `make', \textit{chotěti} `want' \\ 
\hline
\textbf{(O)ES (right)}	&	\textit{byti} `be', \textit{pomagati} `help', \textit{sěděti} `sit', \textit{běžati} `run', \textit{v\foreignlanguage{russian}{ъ}schoditi} `go up', \textit{chotěti} `want', \textit{stojati} ‘stand’, \textit{izvoliti} `want, prefer', \textit{moliti sę} `pray', \textit{svitati} `dawn'	&	\textit{glagolati} `speak, say', \textit{rešti} `say', \textit{chotěti} `want', \textit{prositi} `beg, ask', \textit{iměti} `have', \textit{tvoriti} `make', \textit{moliti sę} `pray', \textit{věděti} `know', \textit{velěti} `command', \textit{byti} `be' \\ 
\hline
\textbf{Mar. (left)}	&	\textit{byti} `be', \textit{glagolati} `speak, say', \textit{v\foreignlanguage{russian}{ъ}niti} `go in', \textit{iti} `go', \textit{is\foreignlanguage{russian}{ъ}choditi} `exit, go out', \textit{v\foreignlanguage{russian}{ъ}zležati} `lay down', \textit{jasti} `eat', \textit{iziti} `go down', \textit{priti} `come, arrive', \textit{iměti} `have'	&	\textit{viděti} `see', \textit{priti} `come, arrive', \textit{slyšati} `hear', \textit{iti} `go', \textit{pristǫpiti} `ascend, approach', \textit{iziti} `go down', \textit{prijęti} `take, receive', \textit{v\foreignlanguage{russian}{ъ}stati} `stand up', \textit{v\foreignlanguage{russian}{ъ}z\foreignlanguage{russian}{ь}rěti} `look up at', \textit{byti} `be' \\ 
\hline
\textbf{Mar. (right)}	&	\textit{byti} `be', \textit{pristǫpiti} `ascend, approach', \textit{v\foreignlanguage{russian}{ъ}sijati} `begin to shine', \textit{s\foreignlanguage{russian}{ъ}pati} `sleep', \textit{vlasti} `rule', \textit{v\foreignlanguage{russian}{ъ}niti} `go in', \textit{zatvoriti} `close', \textit{v\foreignlanguage{russian}{ъ}zdati} `give (in homage)', \textit{pospěš\foreignlanguage{russian}{ь}stvovati} `assist, promote', \textit{utvr\foreignlanguage{russian}{ь}ždati} `establish, build'	&	\textit{učiti} `learn, teach', \textit{byti} `be', \textit{iměti} `have', \textit{iskušati} `tempt, try', \textit{viděti} `see', \textit{iskati} `seek', \textit{rešti} `say', \textit{věděti} `know', \textit{moliti sę} `pray', \textit{propovědati} `preach' \\ 
\hline
\caption[Dative absolutes and conjunct participles: ten most-frequent lemmas by subcorpus and position of the participle relative to the matrix clause]{Dative absolutes and conjunct participles: ten most-frequent lemmas by subcorpus and position of the participle relative to the matrix clause. Lemmas for the mixed (O)CS/(O)ES subcorpus are given on the basis of the normalized data. Only lemmas occurring at least twice are included. If fewer than ten lemmas occur at least twice, then only those are included. \textit{Mar.} = Codex Marianus; \textit{(O)ES} = Old East Slavic and Middle Russian; \textit{left} and \textit{right} = pre-matrix and post matrix, respectively; \textit{N} = normalized.}
\label{mostfreqlemmaptcp}
\end{longtable}

The first observation to make is that a number of verbs, such as \textit{byti} and motion verbs like \textit{priti} `come, arrive' are frequent among both conjunct participles and dative absolutes across all configurations. At the same time, however, there are lemmas that are very frequent among the occurrences of one of the two constructions but not present at all, or much less frequent, among the occurrences of the other. Among dative absolutes, we find several verbs denoting activities and states, such as \textit{sěděti} `to be sitting', \textit{s\foreignlanguage{russian}{ъ}pati} `sleep', \textit{žiti} `live', \textit{moliti sę} `pray', \textit{stojati} `be standing', which are most often imperfective and, as is most typical, function as \textsc{frames}, as in (\ref{statedas})-(\ref{activitydas}).

\begin{example}
\gll i se v\foreignlanguage{russian}{ъ} jedin\foreignlanguage{russian}{ъ} dn\foreignlanguage{russian}{ь} \textbf{s\foreignlanguage{russian}{ъ}pęštju} tomu v\foreignlanguage{russian}{ъ} poludnie i se javi sę jemu oc\foreignlanguage{russian}{ь} naš\foreignlanguage{russian}{ь} feodosii glę
{and} {behold} {on} {one.{\sc sg.m.acc}} {day.{\sc sg.m.acc}} {sleep.{\sc ptcp.ipfv.m.dat.sg}} {that.{\sc m.dat.sg}} {at} {midday.{\sc sg.n.acc}} {and} {behold} {appear.{\sc aor.3.sg}} {\sc refl} {\sc 3.sg.m.dat} {father.{\sc m.nom.sg}} {our.{\sc 1.m.nom.sg}} {Feodosij.{\sc m.nom.sg}} {say.{\sc ptcp.ipfv.m.nom.sg}}
\glt `And one day, as he was sleeping at midday, our Father Feodosij appeared to him, saying' (\textit{Life of Feodosij Pečerskij}, Uspenskij Sbornik f. 65b)
\glend
\label{statedas}
\end{example}

\begin{example}
\gll i sice \textbf{molęštim} sę poidoša polovci
and thus pray.\textsc{ptcp.ipfv.dat.pl} \textsc{refl} advance.\textsc{aor.3.pl} Polovtsy.\textsc{nom.pl}
\glt `And as they [the Rusian princes] were praying, the Polovtsy advanced' (\textit{Primary Chronicle}, Codex Laurentianus f. 93d)
\glend
\label{activitydas}
\end{example}

Also among dative absolutes, we find verbs, such as \textit{minǫti} `pass', clearly involved in time expressions, as in (\ref{minotidas}).

\begin{example}
\gll časoma že dvěma \textbf{minǫv\foreignlanguage{russian}{ъ}šema}. raschoždaaše sę {po malu} t\foreignlanguage{russian}{ъ}ma
{hour.{\sc m.dat.du}} {\sc ptc} {two.{\sc m.dat.du}} {pass.{\sc ptcp.pfv.m.dat.du}} {disperse.{\sc impf.3.sg}} {\sc refl} {slowly} {darkness.{\sc f.nom.sg}}
\glt `After two hours had passed, darkness slowly started to disperse' (\textit{Vita of Kodratos}, Codex Suprasliensis f. 56v)
\glend
\label{minotidas}
\end{example}

Very frequent among absolutes are also verbs indicating under which ruler or religious figure a particular event occurs within the narrative. Lemmas such as \textit{cěsar\foreignlanguage{russian}{ь}stvovati} `rule, serve as emperor' and \textit{četvr\foreignlanguage{russian}{ь}tovlast\foreignlanguage{russian}{ь}stvovati} `serve as a tetrarch' lend themselves well as framing devices in a discourse, setting the main event within a timeframe. (\ref{rulingverbdas1}) is one of several such examples.

\begin{example}
\gll v\foreignlanguage{russian}{ъ} lět .dz. ps. vi. \textbf{crstvujuštju}. ol\foreignlanguage{russian}{ь}kse. v\foreignlanguage{russian}{ъ} csrigradě. v\foreignlanguage{russian}{ъ} crstvě isakově. brat svojego. jegože slěpiv\foreignlanguage{russian}{ъ}. a sam\foreignlanguage{russian}{ъ} csrm\foreignlanguage{russian}{ь} sta. a sýna jego oleksu. zatvori v\foreignlanguage{russian}{ъ} stěnach\foreignlanguage{russian}{ъ} vysokych\foreignlanguage{russian}{ъ}. stražeju. jako ne vynidet\foreignlanguage{russian}{ь}. 
{in} {year.{\sc sg.n.acc}} {six thousand} {seven hundred} {twelve} {reign.{\sc ptcp.ipfv.m.dat.sg}} {Oleksa.{\sc m.dat.sg}} {in} {Constantinople.{\sc sg.m.loc}} {in} {reign.{\sc sg.n.loc}} {Isak.{\sc sg.n.loc.strong}} {brother.{\sc sg.m.gen}} {his.{\sc gen.sg}} {who.{\sc sg.m.gen}} {blind.{\sc ptcp.pfv.m.nom.sg}} {and} {self.{\sc m.nom.sg}} {emperor.{\sc m.inst.sg}} {become.{\sc aor.3.sg}} {and} {son.{\sc sg.m.gen}} {his.{\sc gen.sg}} {Oleksa.{\sc sg.m.acc}} {shut.{\sc aor.3.sg}} {in} {wall.{\sc pl.f.loc}} {high.{\sc pl.f.loc}} {guard.{\sc f.inst.sg}} {that} {\sc neg} {go out.{\sc prs.3.sg}}
\glt `In the year 6712 [1204], when Oleksa was reigning in Constantinople: during the reign of his brother Isak, whom Oleksa blinded before becoming emperor himself, Isak shut his son Oleksa in high walls under guard so that he could not get out' (\textit{First Novgorod Chronicle} f. 64r)
\glend
\label{rulingverbdas1}
\end{example}

On the other hand, among conjunct participles we find several telic verbs, such as \textit{prijęti} ‘take, receive’, \textit{v\foreignlanguage{russian}{ъ}zęti} `take', 
\textit{slyšati} `hear' \textit{s\foreignlanguage{russian}{ъ}tvoriti} `make', \textit{viděti} `see', \textit{v\foreignlanguage{russian}{ъ}stati} `stand up'. We can instead more easily envision these as part of a series of foreground clauses following one another, as is often the case with \textsc{independent rhemes}. (\ref{telicxadv1})-(\ref{telicxadv3}) are typical examples of conjunct participles with these lemmas.

\begin{example}
\gll Kagan že \textbf{v\foreignlanguage{russian}{ъ}z\foreignlanguage{russian}{ь}m\foreignlanguage{russian}{ь}} čašu reče
khan \textsc{ptc} take.\textsc{ptcp.pfv.m.nom.sg} cup.\textsc{f.acc.sg} say.\textsc{aor.3.sg}
\glt `The Khan then took the cup and said' (\textit{Vita Constantini} 9)
\glend
\label{telicxadv1}
\end{example}

\begin{example}
\gll plamen\foreignlanguage{russian}{ь} velik\foreignlanguage{russian}{ъ} zělo ot v\foreignlanguage{russian}{ь}r\foreignlanguage{russian}{ь}cha crk\foreignlanguage{russian}{ъ}v\foreignlanguage{russian}{ь}naago \textbf{iš\foreignlanguage{russian}{ь}d\foreignlanguage{russian}{ъ}} i aky komara \textbf{s\foreignlanguage{russian}{ъ}tvoriv\foreignlanguage{russian}{ъ}} sę prěide na drugyi ch\foreignlanguage{russian}{ъ}l\foreignlanguage{russian}{ъ}m\foreignlanguage{russian}{ъ}
{flame.{\sc m.nom.sg}} {big.{\sc m.nom.sg.strong}} {very} {from} {top.{\sc sg.m.gen}} {church's.{\sc sg.m.gen}} {go out.{\sc ptcp.pfv.m.nom.sg}} {and} {like} {vault.{\sc f.nom.sg}} {make.{\sc ptcp.pfv.m.nom.sg}} {\sc refl} {reach.{\sc aor.3.sg}} {on} {other.{\sc sg.m.acc}} {hill.{\sc sg.m.acc}}
\glt `A very tall flame rose from the dome of the church, bent like a vault and reached another hill' (\textit{Life of Feodosij Pečerskij}, Uspenskij Sbornik f. 56a)
\glend
\label{telicxadv2}
\end{example}

\begin{example}
\gll on\foreignlanguage{russian}{ъ} že \textbf{priim\foreignlanguage{russian}{ъ}} jego oglasi jego slovesy dchov\foreignlanguage{russian}{ъ}nyimi
{that.{\sc m.nom.sg}} {\sc ptc} {receive.{\sc ptcp.pfv.m.nom.sg}} {\sc 3.sg.m.gen} {catechize.{\sc aor.3.sg}} {\sc 3.sg.m.gen} {word.{\sc pl.n.inst}} {spiritual.{\sc pl.n.inst}}
\glt `But he received him and instructed him in the holy scriptures' (\textit{Vita of St. Aninas the Wonderworker}, Codex Suprasliensis f. 140r)
\glend
\label{telicxadv3}
\end{example}

Mental-state verbs such as \textit{chotěti} ‘want’ and \textit{věděti} ‘know’ generally involve a different reading than the typical temporal ones seen in the OCS dataset and are often likely to receive a causal interpretation, as in the conjunct participles in (\ref{causalvedeti})-(\ref{causalxoteti}).

\begin{example}
\gll d\foreignlanguage{russian}{ь}javol\foreignlanguage{russian}{ъ} radvaše sę semu. ne \textbf{vědyi} jako bliz\foreignlanguage{russian}{ь} pogibel\foreignlanguage{russian}{ь} chotęše byti emu
{devil.{\sc nom}} {rejoice.{\sc impf.3.sg}} {\sc refl} {this.{\sc dat.sg}} {\sc neg} {know.{\sc ptcp.ipfv.m.nom.sg}} {that} {near} {death.{\sc nom}} {want.{\sc impf.3.sg}} {be.{\sc inf}} {\sc 3.sg.m.dat}
\glt `The devil rejoiced at that, not knowing [i.e. because he did not know] that his death was near' (\textit{Primary Chronicle}, Codex Laurentianus f. 26v)
\glend
\label{causalvedeti}
\end{example}

\begin{example}
\gll on že ne \textbf{chotę} krovi proliti. ne bi sę s nimi
{that.{\sc m.nom.sg}} {\sc ptc} {\sc neg} {want.{\sc ptcp.ipfv.m.nom.sg}} {blood.{\sc gen.sg}} {shed.{\sc inf}} {\sc neg} {fight.{\sc aor.3.sg}} {\sc refl} {with} {\sc 3.pl.m.inst}
\glt `Not wanting to [i.e. since he did not want to] shed blood, he did not fight with him'. (\textit{Suzdal Chronicle}, Codex Laurentianus f. 102r) %273808 102r
\glend
\label{causalxoteti}
\end{example}

This, however, does not necessarily entail overall different functions than the one observed so far: as already mentioned, unless augmented with overt connectives (e.g. ‘because’, ‘when’, ‘while’), the semantics of participial adjuncts is underspecified, which is why a semantic framework integrating discourse relations was needed in the first place to fully account for dative absolutes and conjunct participles. \textit{Chotěti} ‘want’ is found among the most frequent lemmas of both dative absolutes and conjunct participles. Among the latter it is most often used imperfectively and can generally be interpreted as an \textsc{elaboration}, as in (\ref{xotetixadv1}) and (\ref{xotetixadv2}).

\begin{example}
\gll michal\foreignlanguage{russian}{ь} že viděv\foreignlanguage{russian}{ъ} to sskoči s konę \textbf{chotę} pomoči volodimeru
{Michal.{\sc m.nom.sg}} {\sc ptc} {see.{\sc ptcp.pfv.m.nom.sg}} {that.{\sc sg.n.acc}} {jump off.{\sc aor.3.sg}} {from} {horse.{\sc sg.m.gen}} {want.{\sc ptcp.ipfv.m.nom.sg}} {help.{\sc inf}} {Volodimer.{\sc m.dat.sg}}
\glt ‘Michal, when he saw that, jumped off the horse wanting to help Volodimer' (\textit{Suzdal Chronicle}, Codex Laurentianus f. 106r)
\glend
\label{xotetixadv1}
\end{example}

\begin{example}
\gll i v monastyri poidoša ne \textbf{chotjašte} v polon poiti ot svoego grada vo inye grady
{and} {in} {monastery.{\sc pl.f.acc}} {go.{\sc aor.3.pl}} {\sc neg} {want.{\sc ptcp.ipfv.f.nom.pl}} {in} {captivity.{\sc sg.m.acc}} {go.{\sc inf}} {from} {their.{\sc gen.sg}} {town.{\sc sg.m.gen}} {in} {other.{\sc pl.m.acc}} {town.{\sc pl.m.acc}}
\glt `And they went in monasteries, not wishing to go into captivity from their own town to other towns' (\textit{The Taking of Pskov}) %245369
\glend
\label{xotetixadv2}
\end{example}

In dative absolutes, \textit{chotěti} is not always used in its lexical meaning of ‘wanting to’, but rather as a modal auxiliary translatable as ‘being about to’, as in (\ref{xotetidas1})-(\ref{xotetidas2}), where it can be interpreted as a generic \textsc{frame} (‘when \textit{x} was about to do \textit{y}, \textit{z} happened’):

\begin{example}
\gll na sud že jemu \textbf{choteštu} iti. plaka se mati otročete sego. glagoljušti
to judgement.\textsc{acc} \textsc{ptc} \textsc{3.sg.dat} want.\textsc{ptcp.ipfv.m.dat.sg} go.\textsc{inf} cry.\textsc{aor.3.sg} \textsc{refl} mother.\textsc{nom} child.\textsc{gen} this.\textsc{m.gen.sg} say.\textsc{ptcp.ipfv.f.nom.sg}
\glt ‘But when he was ready to go for the Judgement, the mother of the child cried, saying' (\textit{Vita Constantini} 2)
\glend
\label{xotetidas1}
\end{example}

\begin{example}
\gll i \textbf{chotěv\foreignlanguage{russian}{ъ}šu} jemu po dvoju d\foreignlanguage{russian}{ь}niju ot\foreignlanguage{russian}{ъ}plouti javi sę jemu v\foreignlanguage{russian}{ь} s\foreignlanguage{russian}{ь}ně aggel\foreignlanguage{russian}{ъ} gospod\foreignlanguage{russian}{ь}n\foreignlanguage{russian}{ь} glagoľę
and want.\textsc{ptcp.pfv.m.dat.sg} \textsc{3.sg.dat} on second.\textsc{loc} day.\textsc{loc} sail.away.\textsc{inf} appear.\textsc{aor.3.sg} \textsc{refl} \textsc{3.sg.dat} in dream.\textsc{loc} angel.\textsc{nom} lord.\textsc{adj.nom.sg} say.\textsc{ptcp.ipfv.m.nom.sg}
\glt ‘And as he was going to sail away on the second day, an angel of the Lord appeared to him in dream saying’ (\textit{Vita of John the Hesychast}, Codex Suprasliensis f. 13v)
\glend
\label{xotetidas2}
\end{example}

It is interesting to notice that \textit{věděti} ‘know’ (all imperfective) also occurs as the participle lemma among dative absolutes but does not make it to the 10 most frequent lemmas, seemingly because they all occur in a specific literary genre and language variety (the OES chronicles\footnote{Of the 12 total occurrences, 5 are found in the \textit{Suzdal Chronicle} (Codex Laurentianus), 2 in the \textit{Primary Chronicle} (Codex Laurentianus); outside the chronicles, 4 are in the \textit{Life of Feodosij Pečerskij} (Uspenskij Sbornik) and only 1 in the Codex Suprasliensis.}). If we look at these occurrences more closely, we see that their subjects are often names of kings (i.e. \textit{jaroslav\foreignlanguage{russian}{ъ}} ‘Jaroslav’, \textit{m\foreignlanguage{russian}{ь}stislav\foreignlanguage{russian}{ъ}} ‘Mstislav’, \textit{volodimer\foreignlanguage{russian}{ъ}ko} ‘Volodimirko’, \textit{boris\foreignlanguage{russian}{ъ}} ‘Boris’, \textit{izjaslav\foreignlanguage{russian}{ъ}} ‘Izjaslav’; see examples (\ref{mentalstatedas1})-(\ref{mentalstatedas2})). It is tempting to view this as a tendency for dative absolutes to occur in fixed or semi-fixed verb-subject combinations, especially considering that the common noun \textit{k\foreignlanguage{russian}{ъ}njaz\foreignlanguage{russian}{ъ}} ‘king’, despite being among the most-frequent lemmas in the subcorpus, is never found as the subject of \textit{věděti} ‘know’. 

\begin{example}
\gll jaroslavu že \textbf{ne} \textbf{věduštju} ott\foreignlanguage{russian}{ь}ně smrti. varęzi bęchu mnozi ou jaroslava
Jaroslav.\textsc{dat} \textsc{ptc} \textsc{neg} know.\textsc{ptcp.ipfv.m.dat.sg} father.\textsc{adj.f.dat.sg} death.\textsc{f.dat} Varangian.\textsc{nom.pl} be.\textsc{impf.3.pl} much.\textsc{nom.pl} at Jaroslav.\textsc{gen}
\glt ‘While Jaroslav had not heard of his father’s death yet, many Varangians were under his command’ (\textit{Primary Chronicle}, Codex Laurentianus f. 48a)
\glend
\label{mentalstatedas1}
\end{example}

\begin{example}
\gll volodimerku že togo \textbf{ne} \textbf{věduštju} so andrějem. i stasta u mič\foreignlanguage{russian}{ь}ska
Volodimerko.\textsc{dat} \textsc{ptc} that.\textsc{n.gen} \textsc{neg} know.\textsc{ptcp.ipfv.m.dat.sg} with Andrej.\textsc{ins} and set.out.\textsc{aor.3.du} at Mičesk.\textsc{gen}
\glt ‘Volodimerko and Andrej did not know that and went to Mičesk’ (\textit{Suzdal Chronicle}, Codex Laurentianus f. 110r)
\glend
\label{mentalstatedas2}
\end{example}

% It is possible that the lower lexical variation among sentence-initial SV dative absolutes may be due to such common, more `localized' combinations of subject-lemma in that configuration, which is difficult, however, to test quantitatively.\\
The rest of the occurrences of \textit{věděti} ‘know’ among dative absolutes appear instead in clearly fixed expressions meaning ‘without anyone knowing’, ‘not known to anyone’, all in the OES subcorpus (as in (\ref{vedetifixed1})) except one example in \textit{Suprasliensis} (\ref{vedetifixed2}): 

\begin{example}
\gll \textbf{ne} \textbf{věduštju} nikomuže pridoch v pečeru.
\textsc{neg} know.\textsc{ptcp.ipfv.m.dat.sg} nobody.\textsc{dat} come.\textsc{aor.1.sg} in crypt.\textsc{acc}
\glt ‘Without anyone knowing I went to the crypt’ (\textit{Primary Chronicle}, Codex Laurentianus f. 70b)
\glend
\label{vedetifixed1}
\end{example}

\begin{example}
\gll i zatvori i v\foreignlanguage{russian}{ъ} chyzině pęti desęt\foreignlanguage{russian}{ъ} ti pęti lět\foreignlanguage{russian}{ъ} sǫšta \textbf{nikomuže} inomu ot\foreignlanguage{russian}{ъ} družiny \textbf{vědǫštu} jako episkup\foreignlanguage{russian}{ъ} jest\foreignlanguage{russian}{ъ}
and lock-up.\textsc{aor.3.sg} \textsc{3.sg.m.acc} in shack.\textsc{loc} five.\textsc{gen} ten.\textsc{gen} \textsc{ptc} five.\textsc{gen} year.\textsc{gen.pl} be.\textsc{ptcp.ipfv.m.gen.sg} nobody.\textsc{dat.sg} other.\textsc{dat} from company.\textsc{gen} know.\textsc{ptcp.ipfv.m.dat.sg} that bishop.\textsc{nom.sg} be.\textsc{prs.3.sg}
\glt ‘And he locked him up in a shack when he was fifty-five, without anyone else from the company knowing that he was a bishop’ (\textit{Vita of John the Hesychast}, Codex Suprasliensis f. 13r) %139641)
\glend
\label{vedetifixed2}
\end{example}

Overall, lexical variation among the constructions supports what we observed through other variables. Dative absolutes have a relatively consistent framing function throughout different configurations, as emerged from the overall smaller lexical variation among participle lemmas compared to conjunct participles, as well as from the lack of significant differences between sentence-initial dative absolutes in the SV and VS configurations. Conjunct participles, on the other hand, are functionally more varied, and the different functions are reflected in the different levels of lexical variation in different configurations, particularly depending on the position of the subject in sentence-initial occurrences, whereby sentence-initial VS conjunct participles are overall less lexically varied than those in the SV configuration, supporting the interpretation of the former as more typical \textsc{frames}.

% \textsc{independent rhemes}, among which we expect overall high lexical variation (since they introduce new information), but also \textsc{frames}, which are instead expected to be more `predictable' and thus display less lexical variation. Similarly, sentence-initial VS conjunct participles are overall less lexically varied than sentence-initial SV conjunct participles, also supporting the interpretation of the former as more typical \textsc{frames}.

\subsubsection{Summary}
The analysis of lexical variation among \textit{participles} showed that conjunct participles are overall more lexically varied than dative absolutes as a whole, except in the VS configuration, where conjunct participles show less lexical variation. When we cross these results together with the properties of subjects, we see that dative absolutes tend to occur in very predictable participle lemma-subject lemma combinations and, not infrequently, in what seems like fixed or semi-fixed expressions. These can occur across different texts and genres as `formulae'\footnote{I am referring to \textit{formulae} informally here since I have not systematically focussed on testing the formulaicity of dative absolutes, although there are strong reasons to think that such a study is warranted.} such as those seen with elaborating dative absolutes (of the type `God <\textsc{help}> <\textsc{person}>', where \textsc{help} is generally one of a limited set of verbs all with a similar semantics to \textit{help} (e.g., again, \textit{s\foreignlanguage{russian}{ъ}bljudati} `guard, watch over', \textit{pomagati} `help', \textit{ukrěpljati} `give strength'), but also variations of the basic phrase \textit{ne věduštju nikomuže} (`without anyone knowing). But they may also occur as semi-fixed combinations contextually only to a specific text or genre, as in `reigning <\textsc{ruler}> in <\textsc{place}>' (e.g. \textit{crstvujuštju ol\foreignlanguage{russian}{ь}kse v\foreignlanguage{russian}{ъ} csrigradě} `when Oleksa was reigning in Constantinople'), which are, for example, found overwhelmingly more frequently in the chronicles.

\section{Conjunct participles and dative absolutes in strategically-annotated texts}\label{sec5}
% [maybe... just maybe... present this as a case study summarizing findings from your JHS article? saying you present it because: 1) it's from cmpletelty automatically annotated treebanks 2) it shows a specific pattern, namely the clause-bridging one, that did not emerge from the codex marianus (probably not a feature there) or from the standard treebanks (because of lack of information structural annotation). or don't talk about clause bridging at all.]
In this final section, I summarize results from a case study which was carried out in \citet{pedrazzinijhs} on strategically annotated treebanks. \pgcitet{pedrazzinijhs}{31-36} looked at the extent to which corpora with considerably fewer and shallower levels of annotation than deeply annotated and standard treebanks can be exploited to investigate a discourse-driven syntactic phenomenon (i.e. the discourse functions of participle clauses), specifically as a means of corroborating results emerging from treebanks with deeper annotation. This section first provides general frequencies on the position and aspect of conjunct participles and dative absolutes and lexical variation among participles (Section \ref{varsstratsec}) as they emerged from the small corpus of strategically annotated treebanks as presented in Section \ref{identifconstr} of the Introduction. However, the main part of the case study (Section \ref{secbdinski}) presents a `distant-to-close-reading' analysis of a specific text, in the sense that it shows the extent to which closer-reading of automatically extracted occurrences of the constructions may complement potential patterns found from automatic methods. The text selected by \citet{pedrazzinijhs} for this case study is the Slavonic \textit{Story of Abraham of Qidun and his niece Mary} (AQM), contained in the 14th-century Middle Bulgarian Bdinski Sbornik (\citealt{duj1972a}; \citealt{scharpe1973a}). The reason for this choice is mainly incidental to the content of TOROT at the time of this study, where the vast majority of non-liturgical texts are of East Slavic provenance. AQM, despite being a translation from Greek, is instead of South Slavic (Middle Bulgarian) provenance and narrates the life of Abraham Qidun, a 4\textsuperscript{th}-century Syrian hermit, and his niece Mary, in a rather different style and with a different vocabulary range than what we are used to from hagiographies included in previous datasets.\footnote{For a detailed discussion of the text's relation to its Greek sources, including translation techniques in this and related hagiographies, see \citet{stern2013, stern2015, stern2016, stern2018}. I am grateful to Dieter Stern for pointing me to the relevant literature on the topic and for kindly sending me copies of the cited material.}\\
\subsection{Position, aspect and lexical variation of participles}\label{varsstratsec}
Tables \ref{abs-aspect-position-strat} and \ref{xadv-aspect-position-strat} show the distribution of aspect among dative absolutes and conjunct participles preceding and following the matrix clause.

\begin{table}[!h]
\centering
\begin{tabular}{ccc}
\hline
& \textbf{imperfectives} & \textbf{perfectives} \\
\hline
\textbf{pre-matrix} & 47.3\% (124) & 52.7\% (138)\\
\textbf{post-matrix}  & 59.9\% (94) & 40.1\% (63)\\
\textbf{tot.} & 52\% (218) & 48\% (201) \\
\hline
\end{tabular}
\caption[Dative absolutes in strategically annotated treebanks: aspect distribution by sentence position]{Dative absolutes in strategically annotated treebanks: aspect distribution by sentence position (row percentage)}
\label{abs-aspect-position-strat}
\end{table}

\begin{table}[!h]
\centering
\begin{tabular}{ccc}
\hline
& \textbf{imperfectives} & \textbf{perfectives} \\
\hline
\textbf{pre-matrix} & 27.7\% (319) & 72.3\% (833)\\
\textbf{post-matrix}  & 64.5\% (453) & 35.5\% (249)\\
\textbf{tot.} & 41.6\% (772) & 58.4\% (1082) \\
\hline
\end{tabular}
\caption[Conjunct participles in strategically annotated treebanks]{Conjunct participles in strategically annotated treebanks: aspect distribution by sentence position (row percentage)}
\label{xadv-aspect-position-strat}
\end{table}

It is immediately clear that the distribution of aspect among conjunct participles follows the same pattern as the one observed in all the other datasets. Namely, the majority of pre-matrix conjunct participles are perfective, while the majority of post-matrix ones are imperfective. This suggests that, overall, it is very likely that their typical functions as \textsc{independent rhemes} and \textsc{elaborations}, respectively, are also reflected in this fully automatically annotated dataset. The association between aspect and position is statistically significant, with a moderate to strong effect size ($\phi=0.36$, Cramér's $V$): $\chi^{2}(1) = 242.08$, $p<0.01$, with the odds of perfectives in pre-matrix position being 4.75 times higher than in post-matrix position. It is, in fact, not hard to find examples with the expected functions only by looking for different combinations of aspect and position in the sentence. (\ref{stratindrheme1})-(\ref{stratindrheme2}) are examples of pre-matrix, perfective conjunct participles, clearly functioning as \textsc{independent rhemes}. (\ref{stratframe}) is instead more likely an example of \textsc{frame} conjunct participles. The framing function of the latter can also be inferred from the fact that we have two coordinated conjunct participles, suggesting fronting and, therefore, a higher adjunction site than \textsc{independent rhemes}. (\ref{stratelab})-(\ref{stratelab2}) are post-matrix, imperfective conjunct participles, clearly functioning as \textsc{elaborations}.

\begin{example}
\gll abie že \textbf{v\foreignlanguage{russian}{ъ}stav\foreignlanguage{russian}{ь}} \textbf{iz\foreignlanguage{russian}{ъ}m\foreignlanguage{russian}{ь}} nož\foreignlanguage{russian}{ь} svoi i s\foreignlanguage{russian}{ъ}tvori ot drěva krst\foreignlanguage{russian}{ь}
immediately \textsc{ptc} {stand up}.\textsc{ptcp.pfv.m.nom.sg} take.\textsc{ptcp.pfv.m.nom.sg} knife and make\textsc{.aor.3.sg} from tree.\textsc{gen.sg} cross.\textsc{acc.sg}
\glt `At once he stood up, took his knife and made a cross from the tree' (Euthymius, \textit{Praise for St. Constantine and Helena}, Sbornici of Vladislav Grammarian f. 428v)
\glend
\label{stratindrheme1}
\end{example}

\begin{example}
\gll i on\foreignlanguage{russian}{ъ} bo \textbf{ispad\foreignlanguage{russian}{ъ}} ot sědania kon\foreignlanguage{russian}{ь}naago padaet\foreignlanguage{russian}{ъ} na zemę nizvr\foreignlanguage{russian}{ъ}žen\foreignlanguage{russian}{ъ} byv\foreignlanguage{russian}{ъ}
and that\textsc{.nom.sg} because fall.\textsc{ptcp.pfv.m.nom.sg} from seat.\textsc{gen.sg} horse's.\textsc{gen.sg} fall.\textsc{prs.3.sg} on ground.\textsc{acc} overthrow.\textsc{ptcp.pass.pfv.m.nom.sg} be\textsc{.ptcp.pfv.m.nom.sg}
\glt `Because he will fall from the saddle and land on the ground, overthrown' (\textit{Manasses Chronicles} p. 5447)
\glend
\label{stratindrheme2}
\end{example}
% и он\foreignlanguage{russian}{ъ} бо испад\foreignlanguage{russian}{ъ} от сѣданиа кон\foreignlanguage{russian}{ь}нааго падает\foreignlanguage{russian}{ъ} на земѧ низвр\foreignlanguage{russian}{ъ}жен\foreignlanguage{russian}{ъ} бꙑв\foreignlanguage{russian}{ъ}

\begin{example}
\gll {\normalfont [}Sia že v\foreignlanguage{russian}{ъ}si \textbf{slyšavše} i \textbf{viděvše}{\normalfont ]} množaišǫ věrǫ i usr\foreignlanguage{russian}{ъ}die k\foreignlanguage{russian}{ъ} stmu pokazovaachǫ 
this.\textsc{n.acc.pl} \textsc{ptc} this.\textsc{n.acc.pl} hear.\textsc{ptcp.pfv.m.nom.sg} and see.\textsc{ptcp.pfv.m.nom.sg} more.\textsc{f.acc} faith.\textsc{f.acc} and zeal.\textsc{n.acc} to saint.\textsc{m.dat.sg} show.\textsc{impf.3.pl}
\glt `After hearing and seeing all these things, they started showing greater faith and zeal towards the Saint' \textit{Life of Ivan of Rila}, Zografski Sbornik f. 102r)
\glend
\label{stratframe}
\end{example}
% Сїа же в\foreignlanguage{russian}{ъ}си слꙑшавше и видѣвше , множаишѫ вѣрѫ и ꙋср\foreignlanguage{russian}{ъ}дїе к\foreignlanguage{russian}{ъ} ст҃мꙋ покаꙁоваахѫ 

\begin{example}
\gll stašę ereie \textbf{čaušte} cělovanija ego
{stand up}.\textsc{aor.3.pl} priest.\textsc{nom.pl} wait.\textsc{ptcp.ipfv.nom.pl} kiss\textsc{.acc.pl} \textsc{3.sg.gen}
\glt `The priests stood up, waiting for his kiss' (Jagičev Zlatoust, f. 29b)
\glend
\label{stratelab}
\end{example}
% пріде в\foreignlanguage{russian}{ь} ризу без\foreignlanguage{russian}{ь}истлѣния нꙑ облачѧ Jagičev Zlatoust, 44b

\begin{example}
\gll i sl\foreignlanguage{russian}{ъ}zami sebe obliav\foreignlanguage{russian}{ъ} padaet na lici \textbf{molę} sę i \textbf{glagolę}
and tear.\textsc{inst.pl} \textsc{refl} dress.\textsc{ptcp.pfv.m.nom.sg} fall.\textsc{prs.3.sg} on face.\textsc{loc} pray.\textsc{ptcp.ipfv.m.nom.sg} \textsc{refl} and say.\textsc{ptcp.ipfv.m.nom.sg}
\glt `And, in tears, he puts on clothes and falls on his face, praying and saying (\textit{Life of Ivan of Rila}, Zografski Sbornik f. 96v)
\glend
\label{stratelab2}
\end{example}
% и сл\foreignlanguage{russian}{ъ}зами себе ѻблїав\foreignlanguage{russian}{ъ} падает на лици молѧ сѧ и гл҃ѧ Zografski Sbornik (96v)

% абие же в\foreignlanguage{russian}{ъ}став\foreignlanguage{russian}{ь} из\foreignlanguage{russian}{ъ}м\foreignlanguage{russian}{ь} нож\foreignlanguage{russian}{ь} свои и с\foreignlanguage{russian}{ъ}твори от дрѣва крст\foreignlanguage{russian}{ь} и на вꙑсоцѣ в\foreignlanguage{russian}{ъ}злож\foreignlanguage{russian}{ь} дрѣвѣ прдѣ воинством носити повелѣ 
Also among dative absolutes, there is a statistically significant correlation between aspect and position, although with a very low effect size ($\phi=0.12$, Cramér's $V$): $\chi^{2}(1) = 5.69$, $p=0.02$, with the odds of perfectives in pre-matrix position being only 1.66 times higher than in post-matrix position. We can see that, as in standard treebanks, the difference in proportion between the two aspects (47.3\% imperfective versus 52.7\% perfectives) is nearly equal. A further test indicates that the difference is not statistically significant among pre-matrix absolutes,\footnote{Binomial test, $p=0.21$.} whereas it is so among post-matrix ones.\footnote{Binomial test, $p<0.01$.} This result suggests that the observations made previously regarding the differences between pre- and post-matrix absolutes are likely supported by this dataset, namely, post-matrix absolutes may also work as \textsc{elaborations} (temporally identical to their matrix), or as clause-final restrictive adjuncts (temporally closer to \textsc{frames}). \\
\indent As Table \ref{lexvarptcp-strat} shows, the difference in lexical variation among the constructions is overall very similar to the one observed in the previous dataset.\footnote{Note that among post-matrix conjunct participles \textit{glagolati} is once again used imperfectively in its quotative particle-like function, so it has been removed from the frequencies in the table.} If we consider the total occurrences, lexical variation is smaller among dative absolutes than conjunct participles. For both constructions, lexical variation is greater among post-matrix than pre-matrix occurrences, but more evidently so for dative absolutes. The difference in variation is evident among pre-matrix occurrences, with dative absolutes showing a much smaller variation than conjunct participles. Variation among post-matrix occurrences is quite similar between the two constructions, although slightly lower for conjunct participles. 

\begin{table}[!h]
\centering
\begin{tabular}{|c|c|c|c|c|}
\hline
\multirow{2}*{\textbf{Subsample}} & \multicolumn{2}{|c|}{\textbf{Absolutes}} & \multicolumn{2}{|c|}{\textbf{Conjuncts}}\\
\cline{2-5}
 & \textbf{10MFL} & \textbf{MATTR} & \textbf{10MFL}   & \textbf{MATTR}\\
\hline
\textbf{total}	&	33.42\%	& 0.73	& 26.18\%	& 0.79 \\ 
\textbf{left}	&	39.84\%	& 0.67	& 27.55\%	& 0.78 \\
\textbf{right}	&	28.77\%	& 0.82	& 28.83\%	& 0.80\\
\hline
\end{tabular}
\caption[Lexical variation among participles by subcorpus in strategically annotated treebanks]{Lexical variation among participles by subcorpus in strategically annotated treebanks. \textit{10MFL} = ten most frequent lemmas; \textit{MATTR} = moving-average type-token ratio; \textit{left} and \textit{right} = pre-matrix and post matrix, respectively. Note that the closer the MATTR is to 1 (and the lower the percentage for the 10MFL), the greater the lexical variation.}
\label{lexvarptcp-strat}
\end{table}

Table \ref{mostfreq_strat} shows the ten most frequent lemmas among conjunct participles and dative absolutes, both overall and by position relative to the matrix clause.

\begin{table}[!h]
\centering
\begin{tabular}{|p{3cm}|p{4cm}|p{4cm}|}
\hline
\textbf{Subcorpus} & \textbf{Absolutes} & \textbf{Conjuncts}\\
\hline
\textbf{total}	& \textit{byti} `be', \textit{chotěti} `want', \textit{rešti} `say', \textit{prěiti} `pass away', \textit{žiti} `live', \textit{dr\foreignlanguage{russian}{ь}žati} `hold', \textit{priiti} `come, arrive', \textit{ostaviti} `leave', \textit{iměti} `have', \textit{ljubiti} `love' & \textit{byti} `be', \textit{viděti} `see', \textit{chotěti} `want', \textit{slyšati} `hear', \textit{rešti} `say', \textit{v\foreignlanguage{russian}{ъ}zęti} `take, seize', \textit{uvěděti} `know', \textit{iti} `go', \textit{iměti} `have', \textit{ostaviti} `leave' \\
\hline
\textbf{left}	&  \textit{byti} `be', \textit{chotěti} `want', \textit{prěiti} `pass away', \textit{rešti} `say', \textit{ostaviti} `leave', \textit{dr\foreignlanguage{russian}{ь}žati} `hold', \textit{s\foreignlanguage{russian}{ъ}kon\foreignlanguage{russian}{ь}čati} `end, finish', \textit{v\foreignlanguage{russian}{ъ}sijati} `become light', \textit{žiti} `live', \textit{iti} `go'  & \textit{byti} `be', \textit{viděti} `see', \textit{slyšati} `hear', \textit{uvěděti} `know', \textit{iti} `go', \textit{priiti} `come, arrive', \textit{v\foreignlanguage{russian}{ъ}stati} `stand up', \textit{rešti} `say', \textit{v\foreignlanguage{russian}{ъ}zęti} `take, seize', \textit{chotěti} `want'  \\
\hline
\textbf{right}	& \textit{byti} `be', \textit{žiti} `live', \textit{ljubiti} `love', \textit{chotěti} `want', \textit{s\foreignlanguage{russian}{ъ}b\foreignlanguage{russian}{ь}rati} `gather', \textit{priiti} `come, arrive', \textit{iměti} `have', \textit{rešti} `say', \textit{věděti} `know', \textit{obladati} `rule, reign'  & \textit{byti} `be', \textit{chotěti} `want', \textit{viděti} `see', \textit{rešti} `say', \textit{tvoriti} `make', \textit{iměti} `have', \textit{v\foreignlanguage{russian}{ъ}zęti} `take, seize', \textit{nositi} `carry, take', \textit{m\foreignlanguage{russian}{ь}něti} `think', \textit{ostaviti} `leave'\\
\hline
\end{tabular}
\caption{Dative absolutes and conjunct participles in strategically annotated treebanks: ten most-frequent lemmas by position of participle relative to the matrix clause}
\label{mostfreq_strat}
\end{table}

Besides the high frequency of \textit{byti} `be' and certain verbs of motion such as \textit{priti} `come, arrive' among both constructions, there are lemmas occurring only among the most frequent of one of the other construction, most evidently state and activity verbs such as \textit{d\foreignlanguage{russian}{ь}ržati} `hold', \textit{žiti} `live', and \textit{obladati} `rule, reign' among dative absolutes and telic verbs such as \textit{v\foreignlanguage{russian}{ъ}zęti} `take, seize', \textit{viděti} `see', and \textit{slyšati} `hear' among conjunct participles. We can also find verbs among dative absolutes which are likely to occur in time expressions, such as \textit{v\foreignlanguage{russian}{ъ}sijati} `become light', to be most likely found in framing expressions such as `when it dawned, when day came' or similar, as in (\ref{framedastimeofday}).

\begin{example}
\gll Světu že \textbf{v\foreignlanguage{russian}{ъ}siavšu} i pov\foreignlanguage{russian}{ъ}sǫdu \textbf{proliavšu} sę, i světlostiǫ ispl\foreignlanguage{russian}{ъ}nenomu i světozarnu d\foreignlanguage{russian}{ь}nju \textbf{byvšu}, javišǫ sę prěžde nevidimaa, i mračnaa t\foreignlanguage{russian}{ъ}ma prognana b\foreignlanguage{russian}{ъ}st\foreignlanguage{russian}{ъ} sianiem\foreignlanguage{russian}{ъ} světlyich\foreignlanguage{russian}{ъ} luč\foreignlanguage{russian}{ь}
light.\textsc{dat} \textsc{ptc} dawn.\textsc{ptcp.pfv.m.dat.sg} and everywhere spread.\textsc{ptcp.pfv.m.dat.sg} \textsc{refl} and light.\textit{inst.sg} fill.\textsc{ptcp.pass.pfv.m.dat.sg} and bright.\textit{dat.sg} day.\textit{dat.sg} be.\textsc{ptcp.pfv.m.dat.sg} appear.\textsc{aor.3.pl} \textsc{refl} before invisible.\textsc{n.nom.pl} and dark.\textsc{f.nom.sg} darkness.\textsc{f.nom.sg} disperse.\textsc{ptcp.pass.pfv.f.nom.sg} bright.\textsc{gen.pl} ray.\textsc{gen.pl}
\glt `And when the light dawned and spread everywhere, and the day became bright and full of light, all the invisible things appeared and the dark gloom was dispersed by the shining of the bright rays' (\textit{Manasses Chronicles} p. 27)
\glend
\label{framedastimeofday}
\end{example}
% Свѣтоу же в\foreignlanguage{russian}{ъ}сїавшоу и пов\foreignlanguage{russian}{ъ}сѫдоу пролїавшоу сѧ, и свѣтлостїѫ испл\foreignlanguage{russian}{ъ}неномꙋ и свѣтоꙁарноу(bright д(\foreignlanguage{russian}{ь})ню бывшоу, ꙗвишѫ сѧ прѣжде невидимаа, и мранаа т\foreignlanguage{russian}{ъ}ма прогнана б\foreignlanguage{russian}{ъ}ст\foreignlanguage{russian}{ъ} сїанїем\foreignlanguage{russian}{ъ} свѣтлыих\foreignlanguage{russian}{ъ} лоу\foreignlanguage{russian}{ь}.

Overall, aspect, position in the sentence and lexical variation all point to the properties of absolutes and conjunct participles that we observed in deeply annotated and standard treebanks. The results of the closer-reading analysis in the next section, which looks at the \textit{Story of Abraham of Qidun and his niece Mary} from the Bdinski Sbornik, while agreeing with these observations, will also highlight a more specialized potential usage of absolute constructions as clause-bridging (and topic-shifting) devices.

\subsection{Zooming in: dative absolutes and conjunct participles in a Middle Bulgarian text}\label{secbdinski}
\indent Potential conjunct participles and dative absolutes were first extracted using the same criteria used for extracting the constructions from the rest of the strategically annotated treebank: dative absolutes by looking for participles in the dative with an \textsc{advcl} dependency relation, conjunct participles by looking for participles in the nominative also with an \textsc{advcl} dependency relation. \\
\indent Before post-correction, 15 potential dative absolutes and 22 conjunct participles were identified on the basis of automatic annotation alone. All 22 potential conjunct participles were confirmed as actual conjunct participles. Of the potential dative absolutes, two turned out to be part of the same passive absolute construction (\textit{prizvaně byvši} ‘having been called’), where one of the dative participles (\textit{byvši}) is the copula (i.e. two of the potential absolutes were actually one). One occurrence was found to be a finite verb (\textit{nareku} ‘I call, will call’), which was tagged as a dative participle, possibly because of the stem \textit{narek}- (morphologically potentially suggesting a past active participle) and the -\textit{u} ending (suggesting a masculine or neuter dative). Finally, two of the potential occurrences were revealed to be regular dative participles agreeing with a preceding dative noun. All remaining potential absolutes (11 occurrences) were confirmed as such.
% Using the results from the dependency parser meant only retrieving occurrences for which a clear matrix clause had been identified. We can assume that the same is true for conjunct participles, namely that a few occurrences were missed because of sentence missegmentation. However, the relatively small number of occurrences allows us to closely analyze each occurrence and the contexts in which they appear on a case-by-case basis. 
\\ \indent First, predictions were made based on the patterns observed in deeply annotated and standard treebanks. The majority of dative absolutes were expected to be found to the left of the matrix, while conjunct participles were expected to be found in both pre- and post-matrix position with some frequency, with pre-matrix occurrences more likely to be perfective and post-matrix occurrences more likely to be imperfective. This is indeed what we find, as Tables \ref{abs-aspect-position-abrah} and \ref{xadv-aspect-position-abrah} show.

\begin{table}[!h]
\centering
\begin{tabular}{ccc}
\hline
& \textbf{imperfectives} & \textbf{perfectives} \\
\hline
\textbf{pre-matrix} & 4 & 7\\
\textbf{post-matrix}  & 0 & 0 \\
\hline
\end{tabular}
\caption{Dative absolutes in AQM: aspect distribution by sentence position}
\label{abs-aspect-position-abrah}
\end{table}

\begin{table}[!h]
\centering
\begin{tabular}{ccc}
\hline
& \textbf{imperfectives} & \textbf{perfectives} \\
\hline
\textbf{pre-matrix} & 0 & 19 \\
\textbf{post-matrix} & 3 & 0\\
\hline
\end{tabular}
\caption{Conjunct participles in AQM: aspect distribution by sentence position}
\label{xadv-aspect-position-abrah}
\end{table}

\textit{All} dative absolutes occur in pre-matrix position, both imperfectively and perfectively. Conjunct participles occur overwhelmingly more in pre-matrix position, with only three occurrences following the matrix. All pre-matrix occurrences are perfective, whereas the only three post-matrix conjunct participles are imperfective. This is in line with the typical functions of pre-matrix conjunct participles as \textsc{independent rhemes} and of post-matrix ones as \textsc{elaborations}.\\
\indent 

Tables \ref{abslemmasptcp-abr} and \ref{xadvlemmasptcp-abr} show all the subject lemmas and their frequencies for dative absolutes and conjunct participles respectively.

\begin{table}[!h]
\centering
\begin{tabular}{ll}
\hline
\textbf{subject} & \textit{\textbf{n}} \\
\hline
\textit{*i} (3\textsuperscript{rd} pers. pron.) & 6\\
\textit{s\foreignlanguage{russian}{ь}} `this' & 1\\
\textit{ot\foreignlanguage{russian}{ь}c\foreignlanguage{russian}{ь}} ‘father’  & 1\\
\textit{lěto} `year' & 1\\
\textit{utro} `morning' & 1\\
\textit{d\foreignlanguage{russian}{ь}n\foreignlanguage{russian}{ь}} ‘day’  & 1\\
\hline
\end{tabular}
\caption{Dative absolutes in AQM: subject lemmas}
\label{abslemmasptcp-abr}
\end{table}

\begin{table}[!h]
\centering
\begin{tabular}{ll}
\hline
\textbf{subject} & \textit{\textbf{n}} \\
\hline
\textit{on\foreignlanguage{russian}{ъ}} `that one' & 3\\
\textit{sam\foreignlanguage{russian}{ъ}} `self' & 2\\
\textit{znaem\foreignlanguage{russian}{ъ}} ‘acquaintance’  & 1\\
\textit{ogn\foreignlanguage{russian}{ь}} `fire' & 1\\
\textit{poslan\foreignlanguage{russian}{ъ}} `(the) sent (one)' & 1\\
null & 14\\
\hline
\end{tabular}
\caption{Conjunct participles in AQM: subject lemmas}
\label{xadvlemmasptcp-abr}
\end{table}

Very similarly to dative absolutes in both deeply annotated and standard treebanks, the majority of the subjects of dative absolutes are third-person personal pronouns (6 occurrences), as in (\ref{ex32}).

\begin{example}
\gll potom\foreignlanguage{russian}{ь} \textbf{sědeštu} jemu na odrě, reč k\foreignlanguage{russian}{ь} njemu
after sit.\textsc{ptcp.ipfv.m.dat.sg} \textsc{3.sg.dat} on bed.\textsc{loc} say.\textsc{aor.3.sg} to \textsc{3.sg.dat}
\glt ‘After that, as he was was sitting on the bed, she said to him (\textit{AQM}, Bdinski Sbornik f. 10v)
\glend
\label{ex32}
\end{example}

Time-denoting nouns, namely \textit{d\foreignlanguage{russian}{ь}n\foreignlanguage{russian}{ь}} ‘day’ and \textit{lěto} ‘year’ (one occurrence each), also occur as subjects in dative absolutes and appear in constructions very similar to the widespread generic temporal \textsc{frames} found in the Gospels. A particularly clear echo of the latter can be detected in (\ref{utrobyv}), where \textit{utro} ‘morning’ would be analyzed in TOROT as a predicative complement, consistently with similar examples headed by \textit{byti} ‘be’ (e.g. \textit{pozdě byv\foreignlanguage{russian}{ъ}šu} ‘once it got late’, Matthew 14:23). 

\begin{example}
\gll \textbf{outru} že \textbf{byvšu} reč k\foreignlanguage{russian}{ь} nei 
morning.\textsc{dat} \textsc{ptc} be.\textsc{ptcp.pfv.n.dat.sg} say.\textsc{aor.3.sg} to s\textsc{3.sg.dat}
\glt ‘When morning came, he said to her' (\textit{AQM}, Bdinski Sbornik f. 13r)
\glend
\label{utrobyv}
\end{example}

All 10 overt subjects of the dative absolutes follow the participle, which, as we already commented on, can be associated with their framing function, whereby older referents get reinstated in a new discourse segment.\\
\indent Most conjunct participles have a null subject. The demonstrative \textit{on\foreignlanguage{russian}{ъ}} `that one' is the most frequent among overt subjects, which was also the case in the previous dataset. Of the 8 overt subjects, three precede the participle and five follow it. VS conjunct participles are once again clearly more ambiguous between a \textsc{frame} and an \textsc{independent rheme} interpretation. Compare (\ref{xadvframeambiguous}) and (\ref{xadvindrhemeambig}).

\begin{example}
\gll \textbf{slyšavši} že ona ěko kamen\foreignlanguage{russian}{ь} bez\foreignlanguage{russian}{ь}dšen\foreignlanguage{russian}{ь} prěbys v\foreignlanguage{russian}{ь} roukou ego
hear.\textsc{ptcp.pfv.f.nom.sg} \textsc{ptc} that.\textsc{f.nom} as stone.\textsc{nom.sg} without.soul.\textsc{nom.sg} remain.\textsc{aor.3.sg} in arm.\textsc{loc.du} his
\glt ‘After listening, she remained like a soulless stone in his arms' (\textit{AQM}, Bdinski Sbornik f. 12r)
\glend
\label{xadvframeambiguous}
\end{example}

\begin{example}
\gll ona že \textbf{v\foreignlanguage{russian}{ь}stavši} i ob\foreignlanguage{russian}{ь}et\foreignlanguage{russian}{ь} i i načet\foreignlanguage{russian}{ь} lobizati po vii ego
she \textsc{ptc} stand-up.\textsc{ptcp.pfv.f.nom.sg} and embrace.\textsc{aor.3.sg} \textsc{3.sg.m.acc} and start.\textsc{aor.3.sg} kiss.\textsc{inf} on neck.\textsc{loc} \textsc{3.sg.m.gen}
\glt ‘She stood up, and embraced him, and started kissing his neck’ (\textit{AQM}, Bdinski Sbornik f. 8r)
\glend
\label{xadvindrhemeambig}
\end{example}

While (\ref{xadvindrhemeambig}) is unmistakably an \textsc{independent rheme}, a \textsc{frame} reading for (\ref{xadvframeambiguous}) seems more likely. In the latter, one may also notice a causal relation between the participle clause and its matrix, which, as we have seen in the previous section, is more likely to emerge from \textsc{frames} and generally becomes more evident when \textit{framing expressions} follow the matrix.\\
\indent Most null-subject pre-matrix conjunct participles, as in (\ref{bdinkskiindrheme2}), more clearly receive an \textsc{independent rheme} reading.

\begin{example}
\gll \textbf{v\foreignlanguage{russian}{ь}stavši} idi s\foreignlanguage{russian}{ь} nami
{stand up}.\textit{ptcp.pfv.f.nom.sg} go.\textit{imp.2.sg} with us.\textit{inst.pl}
\glt `Stand up and come with us' (\textit{AQM}, Bdinski Sbornik f. 12v)
\glend
\label{bdinkskiindrheme2}
\end{example}

In some instances, a \textsc{frame} reading is instead triggered by the fact that the participle in question is not interpreted as modally dependent on the matrix verb (cf. Section \ref{frameworks} of the Introduction). In (\ref{bdinkskiindrheme1}), for example, \textit{iz\foreignlanguage{russian}{ь}šed\foreignlanguage{russian}{ь}} is not part of the indirect question (signalled by \textit{li}) nor of the negation (i.e. the question is arguably whether the fire will \textit{burn}, not whether it will \textit{come out}).

\begin{example}
\gll ašte že i dr\foreignlanguage{russian}{ь}znu ogn\foreignlanguage{russian}{ь} iz\foreignlanguage{russian}{ь}šed\foreignlanguage{russian}{ь} iz\foreignlanguage{russian}{ь} dverc\foreignlanguage{russian}{ь} těch ne s\foreignlanguage{russian}{ь}žežet li mene
{if} \textsc{ptc} even dare.\textsc{prs.1.sg} fire.\textsc{m.nom.sg} {come out}.\textsc{ptcp.pfv.m.nom.sg} from door.\textit{gen.pl} that.\textsc{gen.pl} \textsc{neg} burn \textsc{q} \textsc{1.sg.gen}
\glt `Even if I dared, won't the fire burn me when it comes out of those doors?' (\textit{AQM}, Bdinski Sbornik f. 4v)
\glend
\label{bdinkskiindrheme1}
\end{example}

Some examples, such as (\ref{bdinskiambig}), are, as is common with sentence-initial conjunct participles, ambiguous between a \textsc{frame} and an \textsc{independent rheme} reading.

\begin{example}
\gll i \textbf{rastr\foreignlanguage{russian}{ь}zavši} ryzi svoe biěše se po licju i chotěše se sama udaviti ot pečaly
and tear.\textsc{part.prf.m.nom.sg} robe.\textsc{acc.pl} {her} beat.\textsc{impf.3.sg} \textsc{refl} on face.\textsc{dat} and want.\textsc{impf.3.sg} \textsc{refl} self suffocate.\textsc{inf} from sorrow.\textsc{gen.sg}
\glt ‘And after tearing her own robes, she started striking her own face and wanted to suffocate herself from sorrow’ (\textit{AQM}, Bdinski Sbornik f. 3v)
\glend
\label{bdinskiambig}
\end{example}

All three post-matrix conjunct participles, shown in (\ref{bdinkskielab1})-(\ref{bdinkskielab2}), are instead very clearly \textsc{elaborations}.

\begin{example}
\gll stoěchu \textbf{plačjušte} se i \textbf{divešte} se
stand.\textsc{impf.3.pl} cry.\textsc{ptcp.ipfv.m.nom.pl} \textsc{refl} and marvel.\textsc{ptcp.ipfv.m.nom.pl} \textsc{refl}
\glt ‘They stood crying and marvelling’ (\textit{AQM}, Bdinski Sbornik f. 16r)
\glend
\label{bdinkskielab1}
\end{example}

\begin{example}
\gll v\foreignlanguage{russian}{ь}z\foreignlanguage{russian}{ь}mše že s\foreignlanguage{russian}{ь} mnoju srebr\foreignlanguage{russian}{ь}nikyi se nně podvizaite se dn\foreignlanguage{russian}{ь} i nošt\foreignlanguage{russian}{ь} kouple \textbf{tvorešte}
take.\textsc{ptcp.pfv.m.nom.sg} but with me silver.\textsc{nom.pl} \textsc{refl} now strive.\textsc{imp.2.pl} \textsc{refl} day.\textsc{acc.sg} and night.\textsc{acc.sg} both create.\textsc{ptcp.ipfv.m.nom.sg}
\glt `When you have earned money with me, make an effort day and night, trading' (\textit{AQM}, Bdinski Sbornik f. 16v)
\glend
\label{bdinkskielab2}
\end{example}

Even though the occurrences are too few to be able to make generalizations regarding lexical variation among participles, we can comment on the types of events they seem to describe. Tables \ref{bdinskixadvlemmas} and \ref{bdinskiabslemmas} show the frequency of lemmas among conjunct participles and dative absolutes, respectively.

\begin{table}[!h]
\centering
\begin{tabular}{ll}
\hline
\textbf{participle} & \textbf{\textit{n}}\\
\hline
\textit{v\foreignlanguage{russian}{ъ}stati} `stand up' & 3\\
\textit{ot\foreignlanguage{russian}{ъ}věštati} `answer' & 2\\
\textit{v\foreignlanguage{russian}{ъ}zęti} `take, seize' & 1\\
\textit{v\foreignlanguage{russian}{ъ}zlězti} `climb, crawl' & 1\\
\textit{diviti sę} `marvel' & 1\\
\textit{plakati} `cry' & 1\\
\textit{v\foreignlanguage{russian}{ъ}saditi} `sit (t.)' & 1\\
\textit{prijęti} `receive' & 1\\
\textit{slyšati} `hear' & 1\\
\textit{v\foreignlanguage{russian}{ъ}splakati} `burst into tears' & 1\\
\textit{poznati} `recognize' & 1\\
\textit{izęti} `remove' & 1\\
\textit{rastr\foreignlanguage{russian}{ь}zati} `tear apart' & 1\\
\textit{v\foreignlanguage{russian}{ъ}sklasti} `place (on)' & 1\\
\textit{iti} `go' & 1\\
\textit{oumoliti sę} `pray' & 1\\
\textit{pomoliti sę} `pray' & 1\\
\textit{iziti} `go out' & 1\\
\textit{tvoriti} `make' & 1\\
\hline
\end{tabular}
\caption{Conjunct participles in AQM: frequency of participle lemmas}
\label{bdinskixadvlemmas}
\end{table}

\begin{table}[!h]
\centering
\begin{tabular}{ll}
\hline
\textbf{participle} & \textbf{\textit{n}}\\
\hline
\textit{sěděti} `sit' & 2 \\
\textit{byti} `be' & 1 \\
\textit{pěti} `sing' & 1    \\
\textit{lobyzati} `kiss' & 1 \\
\textit{umrěti} `die' & 1 \\
\textit{priz\foreignlanguage{russian}{ъ}vati} `summon'  & 1 \\
\textit{v\foreignlanguage{russian}{ъ}niti} ‘enter’ & 1 \\
\textit{prilučiti} (\textit{sę}) `happen' & 1 \\
\textit{kon\foreignlanguage{russian}{ь}čati} `finish' & 1 \\
\textit{pomesti} `change, replace' & 1 \\
\hline
\end{tabular}
\caption{Dative absolutes in AQM: frequency of participle lemmas}
\label{bdinskiabslemmas}
\end{table}

As in previous datasets, \textit{v\foreignlanguage{russian}{ъ}stati} `stand up' and \textit{ot\foreignlanguage{russian}{ъ}věštati} `answer' occur more than once, in sentences such as (\ref{stand}) and (\ref{answer}), respectively.

\begin{example}
\gll i \textbf{v\foreignlanguage{russian}{ь}stavšaa} idosta abie
and stand.up.\textsc{ptcp.pfv.m.nom.du} go.\textsc{aor.3.du} immediately
\glt `And they stood up and left without further ado' (\textit{AQM}, Bdinski Sbornik f. 13r)
\glend
\label{stand}
\end{example}

\begin{example}
\gll ona že \textbf{otvěštavši} reč k\foreignlanguage{russian}{ь} nemu
that.\textsc{nom.f.sg} \textsc{ptc} answer.\textsc{ptcp.pfv.f.nom.sg} say.\textsc{aor.3.sg} to \textsc{3.sg.m.gen}
\glt `She then replied to him saying' (\textit{AQM}, Bdinski Sbornik f. 13r)
\glend
\label{answer}
\end{example}

As in the previous datasets, telic verbs, such as \textit{v\foreignlanguage{russian}{ъ}zęti} `take, seize', \textit{prijęti} `receive', \textit{slyšati} `hear', \textit{izęti} `remove', are particularly common among conjunct participles. As we observed, these can be easily envisioned as part of a series of events following one another, as in (\ref{telicabrah}).

\begin{example}
\gll \textbf{iz\foreignlanguage{russian}{ь}m} že cětu sam\foreignlanguage{russian}{ь} dast\foreignlanguage{russian}{ь} gostinnyku
{take out}.\textsc{ptcp.pfv.m.nom.sg} \textsc{ptc} daric.\textsc{acc} self.\textsc{m.nom.sg} give.\textsc{aor.3.sg} innkeeper.\textsc{dat.sg}
\glt `The blessed man took out a daric and handed it to the innkeeper' (\textit{AQM}, Bdinski Sbornik f. 9r)
\glend
\label{telicabrah}
\end{example}

Verbs of motion, constituting the majority in the previous two case studies, are also represented in the new datasets for both constructions, as in the conjunct participle in (\ref{moveabrahxadv}) and the dative absolute in (\ref{moveabrahabs}).

\begin{example}
\gll i \textbf{v\foreignlanguage{russian}{ь}zlěz\foreignlanguage{russian}{ь}} sěde na odrě
and climb.up.\textsc{ptcp.pfv.m.nom.sg} sit.\textsc{aor.3.sg} on bed.\textsc{loc}
\glt `So he climbed up and sat on the bed' (\textit{AQM}, Bdinski Sbornik f. 10r)
\glend
\label{moveabrahxadv}
\end{example}

\begin{example}
\gll i \textbf{v\foreignlanguage{russian}{ь}šed\foreignlanguage{russian}{ь}šima} že ima. vidě odr\foreignlanguage{russian}{ь} vysok\foreignlanguage{russian}{ь} nastlan\foreignlanguage{russian}{ь}
and enter.\textsc{ptcp.pfv.dat.du} \textsc{ptc} \textsc{3.du.m.dat} see.\textsc{aor.3.sg} bed.\textsc{acc} large.\textsc{acc} spread.\textsc{ptcp.pfv.pass.acc.sg}
\glt ‘As they entered, he saw a large bed made up’ (\textit{AQM}, Bdinski Sbornik f. 10r)
\glend
\label{moveabrahabs}
\end{example}

If taken out of context, some of the lemmas among dative absolutes hardly seem `predictable predications' from the (Old) Church Slavonic standpoint, based on the pool of lemmas observed in the previous dataset. This is the case with \textit{lobyzajušti} `kissing' in (\ref{ex34}).

\begin{example}
\gll ona že v\foreignlanguage{russian}{ь}stavši i ob\foreignlanguage{russian}{ь}jet\foreignlanguage{russian}{ь} i i načet\foreignlanguage{russian}{ь} lobizati po vii jego. \textbf{lobyzajušti} že jee. v\foreignlanguage{russian}{ь}z\foreignlanguage{russian}{ь}vonja ubo vonjeju čr\foreignlanguage{russian}{ь}noriz\foreignlanguage{russian}{ь}č\foreignlanguage{russian}{ь}skoju tělo jego
she \textsc{ptc} stand-up.\textsc{ptcp.pfv.f.nom.sg} and embrace.\textsc{aor.3.sg} \textsc{3.sg.m.acc} and start.\textsc{aor.3.sg} kiss.\textsc{inf} on neck.\textsc{loc} \textsc{3.sg.m.gen} kiss.\textsc{ptcp.ipfv.f.dat.sg} \textsc{ptc} \textsc{3.sg.f.gen} exhale.\textsc{aor.3.sg} then smell.\textsc{ins} monk.\textsc{adj.ins} body.\textsc{nom} \textsc{3.sg.m.gen}
\glt ‘She stood up, and embraced him, and started kissing his neck. As she was kissing him, the smell of monasticism issued from his body’ (\textit{AQM}, Bdinski Sbornik f. 8r)
\glend
\label{ex34}
\end{example}

By close-reading the surrounding discourse, however, it becomes clear that even uncommon lemmas do not present new information, but refer back to events that had already been introduced in the previous discourse. The dative absolute's lemma \textit{lobyzati} `kiss' in (\ref{ex34}), for example, had already occurred as part of the main predication in the previous sentence (\textit{načet\foreignlanguage{russian}{ь} lobizati} `(he) started kissing'), and gets reinstated by the dative absolute not as a main event but as a \textsc{frame} for a following new event.
\\ \indent It is tempting to see this as a general feature of the phenomenon, whereby dative absolutes may often be used as clause-bridging devices. After closely checking the discourse surrounding each dative absolutes, for most of the occurrences, either one of two patterns emerged. In some of the occurrences, the dative absolute's lemma appeared as the main verb in one of the previous two sentences, as in (\ref{ex34}) above and (\ref{ex35}). In other occurrences, the subject of the dative absolute appeared in the previous sentence as a focal\footnote{The information-structural notion of focus is intended here, along with \pgcitet{halliday_1967}{204} as `a message block which he [the speaker] wishes to be interpreted as informative. What is focal is "new" information; not in the sense that it cannot have been previously mentioned, although it is often the case that it has not been, but in the sense that the speaker presents it as not being recoverable from the preceding discourse'. In this sense, it overlaps with the notion of `rheme' in other terminologies.} element or as one of the event participants in a foregrounded eventuality (\ref{ex36})-(\ref{ex37}).

\begin{example}
\gll prizovi mi ju da se poveselju dn\foreignlanguage{russian}{ь}s s njeju. zělo bo jako slyšal\foreignlanguage{russian}{ь} jesm o njei godě mi jest\foreignlanguage{russian}{ь}. \textbf{prizvaně} že \textbf{byvši} ei. pride k njemu
summon.\textsc{imp} \textsc{1.sg.dat} \textsc{3.sg.f.acc} so-that \textsc{refl} enjoy.\textsc{prs.1.sg} today with \textsc{3.sg.f.inst} much for that hear.\textsc{ptcp.result.m.sg} be.\textsc{prs.1.sg} about \textsc{3.sg.f.loc} pleasant.\textsc{adv} \textsc{1.sg.dat} be.\textsc{prs.3.sg} summon.\textsc{ptcp.pfv.pass.f.dat.sg} \textsc{ptc} be.\textsc{ptcp.pfv.f.dat.sg} s\textsc{3.sg.dat} come.\textsc{aor.3.sg} to \textsc{3.sg.dat}
\glt ‘"Summon her, so that I can enjoy myself with her today. From what I've heard of her, I am much attracted by her”. After being summoned, she came to him.’\hfill(\textit{AQM}, Bdinski Sbornik f. 7v)
\glend
\label{ex35}
\end{example}

\begin{example}
\gll k lět, čr\foreignlanguage{russian}{ь}noriz\foreignlanguage{russian}{ь}č\foreignlanguage{russian}{ь}stvova s\foreignlanguage{russian}{ь} nim\foreignlanguage{russian}{ь} jako agnica čstaa, i golubyca neskvr\foreignlanguage{russian}{ь}nnaa. i \textbf{končavšu} se dvadeset\foreignlanguage{russian}{ь}nu lětu vrěmeni. i neistov\foreignlanguage{russian}{ь}stvo s\foreignlanguage{russian}{ь}tvori dijavol\foreignlanguage{russian}{ь} na nju i sět\foreignlanguage{russian}{ь} poleče da ulovit\foreignlanguage{russian}{ь} ju.
twenty year.\textsc{gen.pl} live-monastically.\textsc{aor.3.sg} with \textsc{3.m.inst.sg} like lamb.\textsc{nom} pure.\textsc{nom} and dove.\textsc{nom} innocent.\textsc{nom} and end.\textsc{ptcp.pfv.n.dat.sg} \textsc{refl} twentieth.\textsc{n.dat} year.\textsc{dat} time.\textsc{gen} and fury.\textsc{acc} make.\textsc{aor.3.sg} devil.\textsc{nom} on \textsc{3.sg.f.acc} and snare.\textsc{acc} set-up.\textsc{aor.3.sg} to catch.\textsc{prs.3.sg} \textsc{3.sg.f.acc}
\glt ‘For twenty years she lived monastically after him, like a chaste lamb, and an innocent dove. When twenty years had elapsed, the Devil turned his fury on her and set up nets to ensnare her’ (\textit{AQM}, Bdinski Sbornik 2v)
\glend
\label{ex36}
\end{example}

\begin{example}
\gll iměše blaženy s\foreignlanguage{russian}{ь} brata po pl\foreignlanguage{russian}{ь}ti. imušta d\foreignlanguage{russian}{ь}štere edinočedju. \textbf{umeršu} že ocju jee. prěbys že junotka sir
have.\textsc{impf.3.sg} blessed.\textsc{nom} this.\textsc{nom} brother.\textsc{acc} by flesh.\textsc{dat} have.\textsc{ptcp.ipfv.m.acc.sg} daughter.\textsc{acc} only.\textsc{acc} die.\textsc{ptcp.pfv.m.dat.sg} \textsc{ptc} father.\textsc{dat} \textsc{3.sg.f.gen} remain.\textsc{aor.3.sg} \textsc{ptc} girl.\textsc{nom} orphan.\textsc{nom}
\glt ‘The blessed one [Abraham] had a brother who had an only daughter [Mary]. When the father [of Mary] died, the girl remained orphan.’ (\textit{AQM}, Bdinski Sbornik f. 1v)
\glend
\label{ex37}
\end{example}

In (\ref{ex35}), the dative absolute \textit{prizvane že byvši ei} ‘after she was called’ picks up the lemma of the imperative \textit{prizovi ju} ‘call her’. In (\ref{ex36}), \textit{20 let} ‘for 20 years’ is also focussed (implying `\textit{for as long as} 20 years'), whereas it then reappears as the subject of the dative absolute, where it is topical (\textit{končavšu se dvadeset\foreignlanguage{russian}{ь}nu letu vremeni} ‘after 20 years had passed’). In (\ref{ex37}), \textit{ocju} ‘father’, the dative absolute's subject, and \textit{brata} ‘brother’, the object of the previous sentence, are actually the same referent, but the different lemma is due to a topic-shift, from the referent’s brother (Abraham) to the referent’s daughter (Mary).\\
\indent It is clear that, in this dataset, dative absolutes are quite systematically used as clause-bridging devices, whereby previously foregrounded eventualities or event participants are `repurposed' as part of the background for a new portion of foregrounded discourse to come. As we will discuss in more detail in the second part of the thesis (see Section \ref{sec:bridging} of Chapter 6), the clause-bridging function of absolutes observed in this dataset is consistent with the very clear typical framing function of the construction and find clear parallels in the world's languages (particularly switch-reference systems; cf. Section \ref{sec:bridging} of Chapter 6).

\section{Summary}
This first part of this chapter analysed the properties of dative absolutes and conjunct participles as they emerge from standard treebanks, a much larger (in terms of number of tokens) and more diverse (in terms of genre and historical periods) dataset than the Codex Marianus. As in Chapter 1, the analysis focussed on the extent to which the functions of participle constructions can be derived indirectly from (a combination of) variables such as position in the sentence, aspect distribution, properties of their subjects, and lexical variation among the participles. \\
\indent As in the Codex Marianus, conjunct participles showed a strong association between position relative to their matrix clause and their aspect. This supports the observation, already made in Chapter 1, that the most typical functions of pre-matrix conjunct participles, which are much more likely to be perfective, is as \textsc{independent rhemes}, whereas post-matrix conjunct participles, which are much more likely to be imperfective, are much more typically \textsc{elaborations}. \\
\indent Among dative absolutes, post-matrix occurrences, while still less frequent (13.9\%), are not as uncommon as the first dataset suggested. The majority of post-matrix absolutes are very similar to \textsc{frames} from the temporal semantic perspective, and several of these may be considered, following \citet{fabricius-hansen2012b}, as clause-final restrictive adjuncts, which are very similar to \textsc{frames} in that they restrict the domain of the matrix clause by introducing a \textit{state} within which the main eventuality holds. Like clause-final restrictive adjuncts, several post-matrix absolutes may receive a (strongly or mildly) contrastive reading. However, based on their temporal semantic contribution, we also found evidence that a smaller number of post-matrix occurrences can be interpreted as \textsc{elaborations}, whose runtime is identical to the one of the matrix eventuality (unlike \textsc{frames}, which have their own runtime and may include the matrix runtime).\\
\indent We then analysed the properties of subjects, focussing on sentence-initial participle constructions. The subjects of sentence-initial conjunct participles showed similar properties to those observed in Chapter 1. Anaphoric parts of speech, namely demonstratives and personal pronouns, are much less frequent in the VS configuration, which can be explained by the specific function of VS conjunct participles of reintroducing older or inactive referents in the discourse. Among sentence-initial dative absolutes, demonstratives, as with conjunct participles, are much more common in the SV configuration, whereas personal pronouns are much more frequent so in the VS one. The analysis of lexical variation both among subjects and among the participles themselves supports the intuition that dative absolutes as a whole, but particularly the VS configuration, are more likely to attract highly discourse-prominent referents as their subjects and to be found in fixed expressions (e.g. \textit{ne věduštju nikomuže} `without anyone knowing' and variations thereof) or predictable subject-verb lemma combinations depending on text or genre (e.g. `reigning <\textsc{ruler}> in <\textsc{place}>', as in \textit{crstvujuštju ol\foreignlanguage{russian}{ь}kse v\foreignlanguage{russian}{ъ} csrigradě} `when Oleksa was reigning in Constantinople'). The new data generally suggests, besides the pool of typically framing verbs common to most texts and historical stages, different genres and language varieties might involve different high-frequency verbs and subjects. The phenomena thus seem to lend itself well to genre-based distributional analyses.\footnote{See \citet{kure2006a} for a study on dative absolutes using a genre-based approach, although not from a quantitative perspective.} \\
\indent The patterns observed in strategically annotated treebanks are consistent with what we observed in the other two datasets. The closer-reading analysis of a Middle Bulgarian Slavonic text, the \textit{Story of Abraham and Mary His Niece} from the Bdinski Sbornik, besides confirming the general patterns found previously, showed that absolutes in this text systematically work as a clause-bridging (or topic-shifting) device. This either emerged from the predicates themselves, whereby a previous foreground clause is repurposed as a background clause using the dative absolutes, or by only repeating event participants that may have been focussed (or constituents of a foreground clause) in the previous discourse and are shifted to topics as subjects of a dative absolute.

\chapter{Finite competitors: \textit{jegda}-clauses}
\section{\textit{Jegda}-clauses as finite competitors}\label{introduction}
Distributional overlap between participle constructions and \textit{jegda}-clauses is evident from parallel examples such as (\ref{parall1})-(\ref{parall2}) and (\ref{parall3})-(\ref{parall4}).

\begin{example}
\begin{itemize}
\item[a.]
\gll i \textbf{egda} \textbf{pride} v\foreignlanguage{russian}{ъ} crkv\foreignlanguage{russian}{ъ}. pristǫpišę k\foreignlanguage{russian}{ъ} nemu učęštju. archierei i star\foreignlanguage{russian}{ь}ci ljud\foreignlanguage{russian}{ь}stii
{and} {when} {come.{\sc aor.3.sg}} {in} {temple.{\sc acc.f.sg}} {move forward.{\sc aor.3.pl}} {to} {\sc 3.sg.m.dat} {teach.{\sc ptcp.ipfv.m.sg.dat}} {chief priest.{\sc pl.m.nom}} {and} {elder.{\sc pl.m.nom}} {people.{\sc pl.m.nom}}
\glt
\glend
\item[b.] 
\gll {Kai} {\textbf{elthontos}} {autou} {eis} {to} {hieron} {prosēlthon} {autōi} {didaskonti} {hoi} {arkhiereis} {kai} {hoi} {presbuteroi} {tou} {laou}
{and} {come.{\sc ptcp.pfv.m.gen.sg}} {\sc 3.sg.m.gen} {in} {the.{\sc n.acc.sg}} {temple.{\sc n.acc.sg}} {move forward.{\sc aor.3.pl}} {\sc 3.sg.m.dat} {teach.{\sc ptcp.ipfv.m.dat.sg}} {the.{\sc pl.m.nom}} {chief priest.{\sc m.nom.pl}} {and} {the.{\sc m.nom.pl}} {elder.{\sc m.nom.pl}} {the.{\sc m.gen.sg}} {people.{\sc m.gen.sg}} 
\glt ‘And when he entered the temple, the chief priests and the elders of the people came to him as he was teaching’ (Matthew 21:23) % 47807(GNT MATT 21.23) 
\glend
\label{parall1}
\end{itemize}
\end{example}

\begin{example}
\begin{itemize}
\item[a.]
\gll \textbf{prišed\foreignlanguage{russian}{ъ}šou} že emu v\foreignlanguage{russian}{ъ} dom\foreignlanguage{russian}{ъ}. pristǫ[p]ste k\foreignlanguage{russian}{ъ} nemu slěp\foreignlanguage{russian}{ъ}ca.
{come.{\sc ptcp.pfv.m.sg.dat}} {\sc ptc} {\sc 3.sg.m.dat} {in} {house.{\sc sg.m.acc}} {move forward.{\sc aor.3.du}} {to} {\sc 3.sg.m.dat} {blind man.{\sc du.m.nom}}
\glt
\glend
\item[b.] 
\gll {\textbf{elthonti}} {de} {eis} {tēn} {oikian} {prosēlthon} {autōi} {hoi} {tuphloi}
{come.{\sc ptcp.pfv.m.dat.sg}} {\sc ptc} {in} {the.{\sc f.acc.sg}} {house.{\sc f.acc.sg}} {move forward.{\sc aor.3.pl}} {him.{\sc 3.sg.m.dat}} {the.{\sc m.nom.pl}} {blind man.{\sc m.nom.pl}}
\glt ‘When he entered the house, the blind men came to him’ (Matthew 9:28) % 47807
\glend
\label{parall2}
\end{itemize}
\end{example}

\begin{example}
\begin{itemize}
\item[a.]
\gll \textbf{egda} \textbf{molite} \textbf{sę} glte Ot\foreignlanguage{russian}{ъ}če naš\foreignlanguage{russian}{ъ} iže esi na nbsch
{when} {pray.{\sc prs.2.pl}} {\sc refl} {say.{\sc imp.2.pl}} {Father.{\sc voc}} {\sc 1.pl.m.voc} {that.{\sc m.nom.sg}} {be.{\sc prs.2.sg}} {in} {heaven.{\sc loc.pl}}
\glt
\glend
\item[b.] 
\gll {\textbf{hotan}} {\textbf{proseukhēsthe}} {legete} {pater} {hagiasthētō} {to} {onoma} {sou}
{when} {pray.{\sc sbjv.prs.2.pl}} {say.{\sc imp.prs.2.pl}} {Father.{\sc voc}} {dedicate.{\sc imp.aor.pas.3.sg}} {the.{\sc n.nom.sg}} {name.{\sc n.nom.sg}} {\sc 2.sg.gen}
\glt `When you pray, say: ``Our Father in heaven, Hallowed be Your name. Your kingdom come"' (Luke 11:2) %  
\glend
\label{parall3}
\end{itemize}
\end{example}

\begin{example}
\begin{itemize}
\item[a.]
\gll \textbf{Molęšte} že \textbf{sę} ne licho glte ěkože i języč\foreignlanguage{russian}{ъ}nici
{pray.{\sc ptcp.ipfv.nom.pl}} {\sc ptc} {\sc refl} {\sc neg} {evil.{\sc n.acc.sg}} {say.{\sc imp.2.pl}} {like} {also} {heathen.{\sc m.nom.pl}}
\glt
\glend
\item[b.] 
\gll {\textbf{Proseukhomenoi}} {de} {mē} {battalogēsēte} {hōsper} {hoi} {ethnikoi}
{pray.{\sc ptcp.ipfv.nom.pl}} {\sc ptc} {\sc neg} {speak idly.{\sc sbjv.aor.2.pl}} {like} {the.{\sc m.nom.pl}} {heathen.{\sc m.nom.pl}}
\glt `And when you pray, do not use vain repetitions as the heathen do' (Matthew 6:7) % 38417 
\glend
\label{parall4}
\end{itemize}
\end{example}

(\ref{parall1}) and (\ref{parall2}) are an example of clear functional overlap between \textit{jegda}-clauses and dative absolutes. They involve the same lemma and verbal aspect, and in both examples the subject of the adverbial is not co-referential with that of the matrix clause. Likewise, (\ref{parall3}) and (\ref{parall4}), involving a \textit{jegda}-clause and a conjunct participle, respectively, are structurally very similar, with a framing adverbial (`when praying') followed by a command.\\
\indent Such parallel examples can also be easily found outside of the Gospels. See, for instance, (\ref{parallshallow1})-(\ref{parallshallow2}) and (\ref{parallshallow3})-(\ref{parallshallow4}), showing the use of \textit{jegda}-clauses and conjunct participles, and of \textit{jegda}-clauses and dative absolutes, respectively, in very similar contexts. 

\begin{example}
    \gll i \textbf{egda} \textbf{vidit\foreignlanguage{russian}{ь}} mest\foreignlanguage{russian}{ь} rucě svoi umyet\foreignlanguage{russian}{ь} v krovi grěšnika
    {and} {when} {see.{\sc prs.3.sg}} {vengeance.{\sc f.acc.sg}} {hand.{\sc f.du.acc}} {his.{\sc 3.du.f.acc}} {wash.{\sc prs.3.sg}} {in} {blood.{\sc loc}} {sinner.{\sc m.gen.sg}}
    \glt `And when he sees vengeance, he will wash his hands in the blood of the sinner' (\textit{Primary Chronicle}, Codex Laurentianus f. 78d)
    \glend
    \label{parallshallow1}
\end{example}

\begin{example}
    \gll on že \textbf{vidę} się i stuživ si ostavlęet pustynju kupno ž i brata svoego
    {that.{\sc m.sg.nom}} {\sc ptc} {see.{\sc ptcp.ipfv.m.sg.nom}} {\sc refl} {and} {grieve.{\sc ptcp.pfv.m.sg.nom}} {this.{\sc 3.sg.m.dat}} {leave.{\sc 3.sg.prs.act}} {hermitage.{\sc f.sg.acc}} {together} {with} {also} {brother.{\sc m.gen.sg}} {his.{\sc 3.sg.m.gen}}
    \glt `When he sees these things and starts grieving, he will leave the hermitage, as well as his own brother' (\textit{Life of Sergij of Radonezh}, f. 61r)
    \glend
    \label{parallshallow2}
\end{example}

\begin{example}
    \gll \textbf{egda} \textbf{ispolni} \textbf{sę} m dnii povelě im\foreignlanguage{russian}{ъ} iti v goru elevon\foreignlanguage{russian}{ь}skuju
    {when} {finish.{\sc aor.3.sg}} {\sc refl} {forty} {day.{\sc m.gen.pl}} {command.{\sc aor.3.sg}} {\sc 3.pl.m.dat} {go.{\sc inf}} {in} {mount.{\sc sg.f.acc}} {olive's.{\sc sg.f.acc}}
    \glt `When forty days had passed, he commanded them to go to the Mount of Olives' (\textit{Primary Chronicle}, Codex Laurentianus f. 35v)
    \glend
    \label{parallshallow3}
\end{example}

\begin{example}
    \gll i \textbf{koncęjuštju} \textbf{sę} lět tomu. vygnaša žiroslava is posadnic\foreignlanguage{russian}{ь}stva. i daša zavidu nerevinicju
    {and} {finish.{\sc ptcp.ipfv.n.dat.sg}} {\sc refl} {year.{\sc sg.n.dat}} {that.{\sc sg.n.dat}} {drive out.{\sc aor.3.pl}} {Žiroslav.{\sc sg.m.gen}} {from} {posadnichestvo.{\sc sg.n.gen}} {and} {give.{\sc aor.3.pl}} {Zavid.{\sc sg.m.dat}} {Nerevinič.{\sc sg.m.dat}}
    \glt `And when that year was ending, they drove Žiroslav out of the \textit{posadnichestvo} and gave it to Zavid Nerevinič' (\textit{Novgorod First Chronicle} f. 40r)
    \glend
    \label{parallshallow4}
\end{example}

The competition between \textit{jegda}-clauses and participle constructions was also clear from some of the mismatches between Greek and Slavic seen in Section \ref{greekocscompa} of Chapter 1. In (\ref{absoluteegda}) and (\ref{AcIegda}), for instance, Greek uses a genitive absolute and a nominalized accusative with infinitive construction, respectively, whereas in both cases Slavic has a \textit{jegda}-clause.

\begin{example}
\begin{itemize}
\item[a.]
\gll da ne \textbf{egda} \textbf{položit\foreignlanguage{russian}{ъ}} osnovaniě. i ne možet\foreignlanguage{russian}{ъ} s\foreignlanguage{russian}{ъ}vr\foreignlanguage{russian}{ъ}šiti. v\foreignlanguage{russian}{ь}si vidęštei nač\foreignlanguage{russian}{ъ}nǫt\foreignlanguage{russian}{ъ} rǫgati sę emu ěko
{so that} \textsc{neg} when lay.\textsc{prs.3.sg} foundation.\textsc{acc} and \textsc{neg} {be able}.\textsc{prs.3.sg} finish.\textsc{inf} all.\textsc{nom.pl} see.\textsc{ptcp.ipfv.nom.pl} begin.\textsc{prs.3.pl} mock.\textsc{inf} \textsc{refl} \textsc{3.sg.dat} that
\glt
\glend
\item[b.]
\gll hina mēpote \textbf{thentos} autou themelion kai mē ischyontos ektelesai pantes hoi heōrountes arxōntai autōi empaizein legontes hoti
thus not-ever lay.\textsc{ptcp.pfv.m.gen.sg} he.\textsc{gen.sg} foundation.\textsc{acc} and \textsc{neg} be-able.\textsc{ptcp.ipfv.m.gen.sg} finish.\textsc{aor.inf} all.\textsc{nom.pl} the see.\textsc{ptcp.ipfv.m.nom.pl} begin.\textsc{sbjv.aor.mid} he.\textsc{dat} mock.\textsc{inf.prs} say.\textsc{ptcp.ipfv.nom.pl} that
\glt ‘lest, after he has laid the foundation, and is not able to finish, all who see it begin to mock him’ (Luke 14:29-30)
\glend
\label{absoluteegda}
\end{itemize}
\end{example}

\begin{example}
\begin{itemize}
\item[a.]
\gll byst\foreignlanguage{russian}{ъ} že \textbf{egda} \textbf{približi} \textbf{sę} is\foreignlanguage{russian}{ъ} v\foreignlanguage{russian}{ъ} erichǫ. slěpec\foreignlanguage{russian}{ъ} eter\foreignlanguage{russian}{ъ} sěděaše pri pǫti prosę
happen.\textsc{aor.3.sg} \textsc{ptc} when approach.\textsc{aor.3.sg} \textsc{refl} Jesus.\textsc{nom} in Jericho.\textsc{acc} blind-man.\textsc{nom} certain.\textsc{nom} sit.\textsc{impf.3.sg} beside way.\textsc{loc} beg.\textsc{ptcp.ipfv.nom.sg}
\glt
\glend
\item[b.]
\gll egeneto de \textbf{en} \textbf{tōi} \textbf{engizein} auton eis Hiereichō typhlos tis ekathēto para tēn hodon epaitōn.
happen.\textsc{aor.3.sg} \textsc{ptc} in the.\textsc{n.dat} approach.\textsc{prs.inf} he.\textsc{acc} to Jericho.\textsc{acc} blind-man.\textsc{nom} certain.\textsc{nom} sit.\textsc{impf.3.sg} beside the road.\textsc{acc} beg.\textsc{ptcp.ipfv.m.nom.sg}
\glt ‘It came to pass that, as Jesus approached Jericho, a blind man was sitting by the road begging’ (Luke 18:35)
\glend
\label{AcIegda}
\end{itemize}
\end{example}

In Chapter 1, we observed that, even though a large number of dative absolutes correspond to genitive absolutes, there is one coherent group of mismatches where Greek has a nominalized accusative with infinitive and Old Church Slavonic has a dative absolutes. We also saw that \textit{byst\foreignlanguage{russian}{ъ}}-/\textit{egeneto}-clauses are somewhat formulaic, in that they generally involve slots filled by constructions with predictable discourse functions---with the slot between \textit{byst\foreignlanguage{russian}{ъ}} and the main verb occupied by a framing, forward-looking construction, in many cases corresponding to an absolute construction. In (\ref{AcIegda}) the \textit{jegda}-clause fills that same slot and, in fact, if we compare the Greek constructions corresponding to \textit{jegda}-clauses in Slavic (Table \ref{egdamismatches_1}), we can see that accusative with infinitives, just like with dative absolutes, are translated relatively often with a \textit{jegda}-clause.

\begin{table}[!h]
\centering
\begin{tabular}{ll}
\hline
\textbf{construction} & \textit{\textbf{n}}\\
\hline
\textbf{finite subordinate}         & 124   \\
\textbf{accusative with infinitive}                        & 24  \\
\textbf{genitive absolute}          & 2 \\
\textbf{backward-controlled participle} & 1   \\
\textbf{conjunct participle}                       & 1   \\
\hline
\end{tabular}
\caption{Ancient Greek constructions corresponding to a \textit{jegda}-clause in Old Church Slavonic}
\label{egdamismatches_1}
\end{table}

With the goal of clarifying what drives the choice between a \textit{jegda}-clause and participle constructions, this chapter and the next look into the issue of competition from the perspectives of temporal relations and discourse interpretation. Our approach to the analysis of \textit{jegda}-clauses will necessarily differ in some important respects from the one taken so far to the analysis of participle clauses. \\
\indent First, because of their finiteness, the addition of the tense variable to the aspectual one introduces some complexity at the level of temporal reference that did not concern the temporal interpretation of participle clauses. As discussed in Section \ref{whatswithwhen} of the Introduction, there are several distinctions at the broad level of event structure (e.g. event/states, durativity/punctuality, telicity/atelicity) that interact with tense-aspect in the interpretation of the temporal relations between finite temporal adverbials and their matrix clause as much as between sequences of independent clauses. As noted in the Introduction, this type of complexity, specific to finite clauses, is reflected, for instance, in the temporal symmetry (or absence thereof) that has long been observed between \textit{when}-clauses and their matrix clauses when either of the two is imperfective (\citealt{kamprohrer,partee84,hinrichs86,sandstrom,bonomi97,s2012a}), as in (\ref{russiansymmetry}).

\begin{example}
    \begin{itemize}
        \item[a.] Irina and I were preparing the documents when Boris called.
        \item[b.] When Irina and I were preparing the documents Boris called.
    \end{itemize}
    \label{russiansymmetry}
\end{example}

The temporal relation between the imperfective clause (\textit{were preparing}) and the perfective clause (\textit{called}) remains the same if the roles of the two clauses are swapped: in this case, the eventuality described by the perfective clause is included in the eventuality described by the imperfective clauses in both (\ref{russiansymmetry}a) and (\ref{russiansymmetry}b). What may have changed is their information structure and/or the discourse relations involved. We can assume a similar situation to hold in Early Slavic. In (\ref{egdasymmetry}a), for example, the \textit{jegda}-clause is imperfective, whereas its matrix clause is perfective, and their temporal relation is one of inclusion of the bounded matrix in the unbounded \textit{jegda}-clause. This same temporal relation would be preserved, without having to change the tense-aspect of either of the predicates, if the roles were swapped, which would give (\ref{egdasymmetry}b).\footnote{The asterisk here does not indicate ungrammaticality, but simply that it is not an attested example.}

\begin{example}
\begin{itemize}
\item[a.]
\gll [i \textbf{egda} \textbf{sěaše}] [ovo pade pri pǫti]
and when sow.\textsc{impf.3.sg} this.\textsc{nom.sg} fall\textsc{.aor.3.sg} along path.\textsc{loc}
\glt ‘And when he was sowing some [seeds] fell along the path’ (Luke 8:5)
\glend
\item[*b.]
\gll [i sěaše] [\textbf{egda} \textbf{ovo} \textbf{pade} \textbf{pri} \textbf{pǫti}]
and sow.\textsc{impf.3.sg} when this\textsc{.nom.sg} fall.\textsc{aor.3.sg} along path.\textsc{loc}
\glt ‘And he was sowing when some [seeds] fell along the path’
\glend
\label{egdasymmetry}
\end{itemize}
\end{example}

While (\ref{egdasymmetry}b) is not, of course, a real example, similar readings are easily found, as in (\ref{egdasymmetryreal}).

\begin{example}
    \gll Toma že edin\foreignlanguage{russian}{ъ} ot\foreignlanguage{russian}{ъ} {oboju na desęte} naricaemy bliznec\foreignlanguage{russian}{ъ}. ne bě tu s\foreignlanguage{russian}{ъ} nimi \textbf{egda} \textbf{pride} \textbf{is}
    {Thomas.{\sc sg.m.nom}} {\sc ptc} {one.{\sc sg.m.nom}} {from} {twelve} {called} {Twin.{\sc sg.m.nom}} {\sc neg} {be.{\sc impf.3.sg}} {here} {with} {\sc 3.pl.m.inst} {when} {come.{\sc aor.3.sg}} {Jesus.{\sc nom}}
    \glt `Now Thomas, one of the twelve, called the Twin, was not with them when Jesus came' (John 20:24)
    \glend
    \label{egdasymmetryreal}
\end{example}

Now, the same observations do not seem to apply to non-finite adjuncts. (\ref{dasymmetry}a), for instance, is an almost identical example to (\ref{egdasymmetry}) above, except that a dative absolute is used instead of a \textit{jegda}-clause. The dative absolute is imperfective, whereas the matrix clause is perfective, and the temporal relation between the two, once again, is one of inclusion of the perfective event in the imperfective event. Albeit a very dubious procedure in itself when it comes to nonfinite adjuncts, to test the possibility of temporal symmetry between a participle construction and its matrix clause, in order to swap their role, we would have to add tense distinctions to the first clause and, conversely, remove them from the second clause. Potentially, if we made the first clause a past-tense one, we would obtain an imperfect, as in the \textit{jegda} example in (\ref{egdasymmetry}a), whereas the second clause would become a perfective dative absolute, presumably resulting in (\ref{dasymmetry}b).

\begin{example}
\begin{itemize}
\item[a.]
\gll [i \textbf{sějǫštoumou}] [ova oubo padǫ pri pǫti]
and sow.\textsc{ptcp.ipfv.m.dat.sg} this.\textsc{nom.pl} fall.\textsc{aor.3.pl} along path
\glt ‘And when he was sowing some [seeds] fell along the path’ (Matthew 13:4)
\glend
\item[*b.]
\gll [i sěaše] [ověm\foreignlanguage{russian}{ъ} oubo \textbf{pad\foreignlanguage{russian}{ъ}šim\foreignlanguage{russian}{ъ}} pri pǫti]
and sow.\textsc{impf.3.sg} this.\textsc{dat.pl} fall.\textsc{ptcp.ipfv.m.dat.pl} along path
\glt ?‘And he was sowing, as some [seeds] had fallen along the path’
\glend
\label{dasymmetry}
\end{itemize}
\end{example}

If such configuration was at all possible, then the event described by the dative absolute in (\ref{dasymmetry}b) would most likely receive a pluperfect interpretation, whereby it is interpreted as having occurred sometime before the imperfect clause, not one of inclusion of the event described by the absolute construction in the matrix event. We find very few of such occurrences (i.e. an imperfect matrix followed by a perfective dative absolute, but (\ref{dasymmetryreal}) is one such example, where the absolute is indeed interpreted as a pluperfect.

\begin{example}
    \gll ězdęchu po onoi storoně dněpra ljudi emljušte a drugyja sěkušte ně \textbf{outęgšim\foreignlanguage{russian}{ъ}} perevesti sę im\foreignlanguage{russian}{ъ}
    {ride.{\sc impf.3.pl}} {along} {one.{\sc sg.f.dat}} {side.{\sc sg.f.dat}} {Dnepr.{\sc sg.m.gen}} {people.{\sc pl.m.acc}} {capture.{\sc ptcp.ipfv.nom.pl}} {and} {other.{\sc pl.m.acc}} {kill.{\sc ptcp.ipfv.nom.pl}} {\sc neg} {manage.{\sc ptcp.pfv.dat.pl}} {cross over.{\sc inf}} {\sc refl} {\sc 3.pl.m.dat}
    \glt `They were riding along the other side of the river Dnepr, capturing some people and killing others, as they had not been able to cross over' (\textit{Kiev Chronicle}, year 6643, Codex Hypatianus)
    \glend
    \label{dasymmetryreal}
\end{example}

Second, \textit{when}-clauses are \textit{overtly} \textit{subordinated} temporal adverbials, even though the temporal relation conveyed by the subordinator \textit{when} is semantically quite underspecified compared to more explicit temporal connectives, such as \textit{after} or \textit{before}. The temporal subordinator \textit{when} has, in fact, long been identified as a clue to infer rhetorical relations closely connected to frame adverbials: most notably, SDRT considers it an `easily recognized trigger' of \textsc{Background} structures (\pgcitealt{asher2007a}{13}), which immediately places \textit{jegda}-clauses in competition specifically to participle \textsc{frames}, rather than participle constructions in any configurations. This narrows down our investigation from the very start and allows us to focus particularly on differences and similarities between \textit{jegda}-clauses and participle constructions functioning as \textsc{frames}. \\
\indent If the main point of reference for the study of participle clauses was \posscitet{baryhaug2011} framework and \posscitet{haug2012a} corpus evidence, the analysis of \textit{jegda}-clauses draws from the plethora of studies on the topic of temporal relations between temporal adverbials, including \textit{when}-clauses, and their matrix clause, particularly \citet{hinrichs86}, \citet{sandstrom}, and \citet{declerck}. It also engages with previous work specifically on the competition between finite and non-finite adjunct clauses, particularly \citet{behrens2012a}, and attempts to reconcile \textit{jegda}-clauses and participle constructions in Early Slavic within the same discourse representation analysis drawing from SDRT accounts of the rhetorical relations triggered by various temporal connectives. \\
\indent The analysis is split across this chapter and the next. This chapter provides a descriptive and quantitative analysis of \textit{jegda}-clauses as they appear in the Early Slavic corpus with a focus on variables that may capture differences in their properties compared to dative absolutes and conjunct participles so as to motivate their distribution. The next chapter analyses the range of temporal readings triggered by \textit{jegda}-clauses based on different attested combinations of tense-aspect between the \textit{jegda}-clause and its matrix clause and provides a discourse-representation-based analysis of \textit{jegda}- and participle clauses.\\
\indent Similarly to the previous two chapters, the corpus analysis in this chapter looks at position of \textit{jegda}-clauses in the sentence and the distribution of tense-aspect in different configurations (Section \ref{egda-order}); the properties of subjects, including their information-structural properties (e.g. anaphoric distance to their antecedent) where possible (i.e. in deeply annotated treebanks) (Section \ref{egda-subjects}); the level of lexical variation of verbs in \textit{jegda}-clauses compared to participle constructions in different configurations; (Section \ref{egda-var}). Finally, it also touches on the average clause length (in number of words) of \textit{jegda}-clauses as opposed to participle clauses, as this has been brought forward in previous literature, as we will see, as a potential functionally motivated difference between finite and nonfinite competitors (Section \ref{lengthsec}).
% \indent Second, it analyses the range of temporal readings triggered by \textit{jegda}-clauses based on different attested combinations of tenses-aspects between the \textit{jegda}-clause and its matrix clause. That section (\ref{egda-tenseaspect}) draws from relevant literature on the topic of temporal relations between temporal adverbials, including \textit{when}-clauses, and their matrix clause, comparing the claims made there with what we are able to observe in the Early Slavic data. \\
% \indent Third, it provides a discourse-representation-based analysis of \textit{jegda}- and participle clauses, this time drawing mainly from SDRT accounts of rhetorical relations triggered by different temporal connectives (Section \ref{egdarethorical}).

\section{\textit{Jegda}-clauses in the corpus}
\subsection{Position in the sentence and tense-aspect distribution}\label{egda-order}
Tables \ref{egdaposition} and \ref{egdapositionshallow} show the position of \textit{jegda}-clauses relative to the matrix clause, in the Codex Marianus and in standard treebanks, respectively. In both subcorpora, \textit{jegda}-clauses to the left of the matrix clause are significantly more frequent than those to the right ($p<0.01$, binomial test, one-tailed).\footnote{The \textit{jegda}-matrix configuration in all the following frequencies also includes 17 occurrences of \textit{byst\foreignlanguage{russian}{ъ}}-clauses where, syntactically, the \textit{jegda}-clause follows the main verb (\textit{byst\foreignlanguage{russian}{ъ}}). For the same reasons pointed out in Chapter 1, these have been included among \textit{jegda}-clauses to the left of the matrix.}

\begin{table}[!h]
\centering
\begin{tabular}{ccc}
\hline
\textbf{\textit{jegda}-matrix} & \textbf{matrix-\textit{jegda}}\\
\hline
84.9\% (129)           & 15.1\% (23) \\
\hline
\end{tabular}
\caption{Relative order of \textit{jegda}-clause and matrix clause in the Codex Marianus}
\label{egdaposition}
\end{table}

\begin{table}[!h]
\centering
\begin{tabular}{ccc}
\hline
\textbf{\textit{jegda}-matrix} & \textbf{matrix-\textit{jegda}} & \textit{NA}\\
\hline
62.7\% (158)           & 29\% (73)    & 8.3\% (21)   \\
\hline
\end{tabular}
\caption{Relative order of \textit{jegda}-clause and matrix clause in standard treebanks}
\label{egdapositionshallow}
\end{table}

Some occurrences do not have an identifiable matrix clause (\textit{NA} in Table \ref{egdapositionshallow}) and, upon closer inspection, these appear largely in contexts such as (\ref{noheadegda}), namely as part of preambles preceding the narrative proper. In such contexts, \textit{jegda}-clauses are used to either specify the time and location in which the narrative takes place or to summarize it. 

\begin{example}
    \gll \textbf{EGDA} \textbf{POS\foreignlanguage{russian}{ъ}LA} SAOUL\foreignlanguage{russian}{ъ} I S\foreignlanguage{russian}{ъ}CHRANI DOM\foreignlanguage{russian}{ъ} EGO OUBITI I
    {when} {send.{\sc aor.3.sg}} {Saul.{\sc sg.m.nom}} {and} {watch.{\sc aor.3.sg}} {house.{\sc sg.m.acc}} {\sc 3.sg.m.gen} {kill.{\sc inf}} {3.sg.m.acc}
    \glt `When Saul had sent men to watch David’s house in order to kill him' (\textit{Psalterim Sinaiticum}, Psalm 58)
    \glend
    \label{noheadegda}
\end{example}

As Tables \ref{egdatense} and \ref{egdatenseshallow} show, present-tense \textit{jegda}-clauses are overall the most frequent across the corpus, with aorist \textit{jegda}-clauses not far behind. Imperfect and future \textit{jegda}-clauses (the latter all forms of \textit{byti}, i.e. \textit{bǫd}-) are also found with some frequency.

\begin{table}[!h]
\centering
\begin{tabular}{ccccc}
\hline
& \textbf{present} & \textbf{aorist} & \textbf{imperfect} & \textbf{future}\\ \hline
\textbf{\textit{jegda}-matrix} & 40.3\% (52) & 39.5\% (51) & 11.6\% (15) & 8.5\% (11) \\
\textbf{matrix-\textit{jegda}} & 56.5\% (13) & 34.8\%(8) & 8.7\% (2) & 0\% (0)  \\
\textbf{Tot.} & 42.8\% (65) & 38.8\% (59) & 11.2\% (17) & 7.2\% (11)  \\
\hline
\end{tabular}
\caption[\textit{Jegda}-clauses in the Codex Marianus: tense distribution by position relative to the matrix clause]{\textit{Jegda}-clauses in the Codex Marianus: tense distribution by position relative to the matrix clause (row percentage)}
\label{egdatense}
\end{table}

\begin{table}[!h]
\centering
\begin{tabular}{ccccc}
\hline
& \textbf{present} & \textbf{aorist} & \textbf{imperfect} & \textbf{future}\\ \hline
\textbf{\textit{jegda}-matrix} & 39.9\% (63) & 36.7\% (58) & 20.9\% (33) & 2.5\% (4) \\
\textbf{matrix-\textit{jegda}} & 54.8\% (40) & 27.4\%(20) & 16.4\% (12) & 1.4\% (1)  \\
\textit{NA} & 4.8\% (1) & 57.1\%(12) & 38.1\% (8) & 0\% (0)  \\
\textbf{Tot.} & 41.3\% (104) & 35.7\% (90) & 21\% (53) & 2\% (5)  \\
\hline
\end{tabular}
\caption[\textit{Jegda}-clauses in standard treebanks: tense distribution by position relative to the matrix clause]{\textit{Jegda}-clauses in standard treebanks: tense distribution by position relative to the matrix clause (row percentage)}
\label{egdatenseshallow}
\end{table}

It is hard to make a direct comparison with the distribution of aspects among participle clauses. On the one hand, we can confidently establish the aspect of past-tense forms on the basis of inflectional morphology alone, with the aorist indicating perfective aspect and the imperfect indicating imperfective aspect. Establishing the aspect of nonpast forms is a more complex matter, since the modern Slavic derivational or `lexical' aspect (i.e. aspectual-pair) system was still developing in Old Church Slavonic (cf. \citealt{eckhoff2014a}). Table \ref{egdaasp} reports the frequency of perfectives and imperfectives among \textit{jegda}-clauses in the Codex Marianus after establishing the aspect of present-tense forms manually on a case-by-case basis. As a general rule, if a present-tense form allows a future interpretation (i.e. it is seen as a completed whole), then it was considered perfective (which is \posscitet{dostal} main diagnostic for present forms); otherwise, it was considered imperfective. All aorist forms were instead as perfective, all imperfect forms as imperfective, and all future forms of \textit{byti} (\textit{bǫd}-) as imperfective (cf. \pgcitealt{kamphuis2020a}{36}).\footnote{The dataset with the added `aspect' variable can be found in the project repository (\url{https://doi.org/10.6084/m9.figshare.24166254}).}

\begin{table}[!h]
\centering
\begin{tabular}{ccc}
\hline
& \textbf{imperfectives} & \textbf{perfectives} \\ \hline
\textbf{\textit{jegda}-matrix} & 48.5\% (63) & 51.5\% (67) \\
\textbf{matrix-\textit{jegda}} & 18.2\% (4) & 81.8\% (18)  \\
\textbf{Tot.} & 44.1\% (67) & 55.9\% (85)  \\
\hline
\end{tabular}
\caption[\textit{Jegda}-clauses in the Codex Marianus: aspect distribution by position relative to the matrix clause]{\textit{Jegda}-clauses in the Codex Marianus: aspect distribution by position relative to the matrix clause (row percentage)}
\label{egdaasp}
\end{table}

Based on the manually-tagged aspect, perfectives seem slightly more frequent than imperfectives in both pre-matrix and post-matrix position. A binomial test suggests, however, that the difference in proportion is not statistically significant overall ($p=0.08$, one-tailed), nor in pre-matrix position ($p=0.39$, one-tailed). In the post-matrix position, we get a significant value ($p<0.01$, one-tailed), although we should mind that occurrences are relatively few in that position.

\subsection{Properties of subjects}\label{egda-subjects}
% [all else being equal (pre-matrix, switch reference), egda- and abs are different in the fact that abs have longer anaphoric chains (i.e. used with more salient subjects) and their subject is much more often mentioned in the immediately preceding sentence.]
As shown in Table \ref{egdasubjreal}, in the Codex Marianus \textit{jegda}-clauses with an overt subject are slightly more frequent than those with a null subject, among both pre- and post-matrix occurrences. However, the difference in proportion is not clear-cut and is not, in fact, statistically significant ($p=0.08$, two-tailed binomial test). Outside of the Codex Marianus, as shown in Table \ref{egdasubjrealshallow}, \textit{jegda}-clauses with a null subject are instead slightly more frequent, but once again, the difference between overt- and null-subject \textit{jegda}-clauses is not statistically significant ($p=0.11$, two-tailed binomial test).

\begin{table}[!h]
\centering
\begin{tabular}{lll}
\hline
& \textbf{Overt subject} & \textbf{Null subject} \\
\hline
\textbf{\textit{jegda}-matrix} & 54.3\% (70)  & 45.7\% (59) \\
\textbf{matrix-\textit{jegda}} & 73.9\% (17) & 26.1\% (6)  \\
\textbf{tot.} & 57.2\% (87) & 42.8\% (65)   \\    
\hline
\end{tabular}
\caption[Overt and null subjects in \textit{jegda}-clauses in the Codex Marianus]{Overt and null subjects in \textit{jegda}-clauses in the Codex Marianus (row percentage)}
\label{egdasubjreal}
\end{table}

\begin{table}[!h]
\centering
\begin{tabular}{lll}
\hline
&   \textbf{overt subject} & \textbf{null subject}\\
\hline
\textbf{\textit{jegda}-matrix} & 44.3\% (70)  & 55.7\% (88)\\
\textbf{matrix-\textit{jegda}} & 41.1\% (30) & 58.9\% (43)\\
\textit{NA} & 61.9\% (13) & 38.1\% (8)\\    
\textbf{tot.} & 44.8\% (113) & 55.2\% (139)\\   
\hline
\end{tabular}
\caption[Overt and null subjects in \textit{jegda}-clauses in standard treebanks]{Overt and null subjects in \textit{jegda}-clauses in standard treebanks (row percentage)}
\label{egdasubjrealshallow}
\end{table}

Thanks to the anaphoric-link annotation on the Greek New Testament in the PROIEL treebanks, from nominal referents to their antecedents (including null arguments), we can infer whether the subject of a \textit{jegda}-clause in the Codex Marianus has the same referent as the subject of its matrix clause. Table \ref{egdasubjcoref} shows the number of coreferential versus non-coreferential \textit{jegda}-clauses by position relative to the matrix.

\begin{table}[!h]
\centering
\begin{tabular}{lll}
\hline
& \textbf{coreferential} & \textbf{non-coreferential} \\
\hline
\textbf{\textit{jegda}-matrix} & 37.2\% (48)  & 62.8\% (81) \\
\textbf{matrix-\textit{jegda}} & 30.4\% (7) & 69.6\% (16)  \\
\textbf{Tot.} & 36.8\% (56) & 63.2\% (96)   \\    
\hline
\end{tabular}
\caption{Subject coreference and non-coreference between \textit{jegda}-clauses and their matrix clause in the Codex Marianus}
\label{egdasubjcoref}
\end{table}

\begin{figure}[!h]
\begin{subfigure}{0.5\textwidth}
\includegraphics[width=0.9\linewidth, height=6cm]{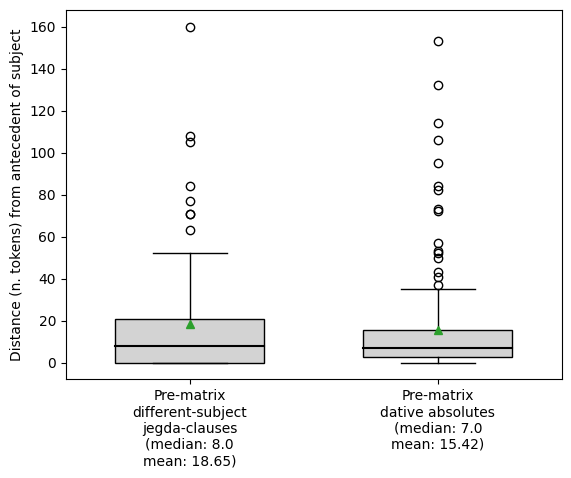} 
\caption[]{}
\label{distance_jegdadas}
\end{subfigure}
\begin{subfigure}{0.5\textwidth}
\includegraphics[width=0.9\linewidth, height=6cm]{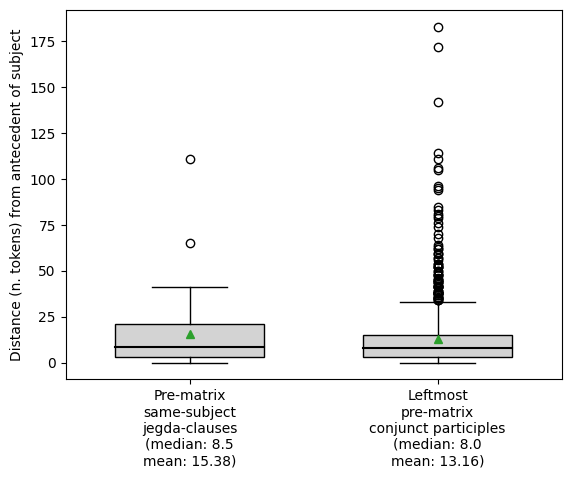}
\caption[]{}
\label{distance_jegdaxadv}
\end{subfigure}
\caption[Average anaphoric distance between the subject and immediate antecedent in pre-matrix non-coreferential/coreferential \textit{jegda}-clauses and pre-matrix dative absolutes/conjunct participles]{(a) Average anaphoric distance between the subject and immediate antecedent in pre-matrix non-coreferential \textit{jegda}-clauses and pre-matrix dative absolutes. Test results: Welch’s $t$-test: 0.82, $p$-value $=0.41$; one-tailed Mann–Whitney $U$-test: 6006.5, $p$-value $=0.78$. (b) Average anaphoric distance between the subject and immediate antecedent in pre-matrix coreferential \textit{jegda}-clauses and (the leftmost) pre-matrix conjunct participles. Test results: Welch’s $t$-test: 0.74, $p$-value $=0.45$; one-tailed Mann–Whitney $U$-test: 21547.5, $p$-value $=0.40$.}
\end{figure}

\textit{Jegda}-clauses with a different subject from the one of the matrix are significantly more frequent than co-referential \textit{jegda}-clauses ($p<0.01$, two-tailed binomial test). 
% Since the vast majority of dative absolutes in the Codex Marianus also have a different subject from that of the matrix, it is tempting to expect \textit{jegda}-clauses to be competing more clearly with dative absolutes from the discourse perspective. This is not at all unlikely from the discourse perspective. We have seen that dative absolutes in the Codex Marianus function by and large as \textsc{frames}. As we have discussed in Section \ref{} of the Introduction, \posscitet{baryhaug2011} \textsc{frames}, albeit specifically referring to semantic properties and discourse functions modeled on participle clauses, are very similar to \textit{frame setters} and are closely related to frame adverbials in formal frameworks of discourse representations, such as SDRT. \\
Intuitively, it is non-coreferential \textit{jegda}-clauses that should show the clearest functional overlap with dative absolutes, since, in the Codex Marianus, the vast majority of absolutes have a different subject from that of the matrix clause. Conversely, coreferential \textit{jegda}-clauses should be especially useful for understanding their competition with conjunct participles, since the latter predominantly have the same subject as the matrix one.\\
\indent As we did in Chapter 1 for dative absolutes and conjunct participles, we can look at the average distance of the immediate antecedent of the subject of \textit{jegda}-clauses and compare it to the results obtained on the subjects of dative absolutes and conjunct participles. Significant differences between average distances may point to functional differences between the constructions. In Section \ref{subjpositionsecdeep} of Chapter 1, for example, we argued that the greater average distance of the antecedent of subjects in VS conjunct participles compared to SV conjunct participles may be indicative of the framing function of the former (whereby older referents are reinstated as part of a \textsc{frame}). In Chapter 1, we only compared sentence-initial dative absolutes and conjunct participles, since that is where the clearest functional overlap occurs between the two and because that is the position in which conjunct participles may head an overt subject. This time, however, even though over 70\% of pre-matrix \textit{jegda}-clauses (93) \textit{are} sentence-initial, if we were to further subset these into two subgroups of \textit{jegda}-clauses (coreferential and non-coreferential, to compare them with conjunct participles and dative absolutes, respectively), the number of sentence-initial occurrences in each subset would be too few to allow for a comparison with sentence-initial participle constructions. We shall therefore consider \textit{pre-matrix}, rather than strictly sentence-initial, \textit{jegda}-clauses, dative absolutes, and conjunct participles. From conjunct participles, only the \textit{leftmost} occurrence is considered (which, most of the time, but not always, coincides with the sentence-initial position), since, as we have seen, that is where they may syntactically head a subject. Figure \ref{distance_jegdadas} compares pre-matrix dative absolutes and pre-matrix \textit{jegda}-clauses whose subject is \textit{not} coreferential with the one of the matrix. Figure \ref{distance_jegdaxadv} compares (the leftmost) pre-matrix conjunct participles and pre-matrix \textit{jegda}-clauses whose subject \textit{is} coreferential with the one of the matrix. 
% As Figure \ref{distanceallcomparison} shows, there is no significant difference between the average distance of the subject's antecedent in the three constructions in sentence-initial position.

% \begin{figure}[!h]
% \centering
% \includegraphics[width=0.8\textwidth]{distance_sentinitall.png}
% \caption{\label{distance_sentinitall}Average anaphoric distance between subject and immediate antecedent in sentence-initial \textit{jegda}-clauses, conjunct participles and dative absolutes. Test results: 1. \textit{jegda}-clauses versus dative absolutes: Welch’s $t$-test: -0.71, $p$-value $=0.47$; one-tailed Mann–Whitney $U$-test: 2366.5, $p$-value $=0.22$; 2. \textit{jegda}-clauses versus conjunct participles: Welch’s $t$-test: 0.42, $p$-value $=0.67$; one-tailed Mann–Whitney $U$-test: 9768.5, $p$-value $=0.89$.}
% \end{figure}
% 

As the test results show, no significant differences emerge from such comparison, namely, the average distance of the antecedent of the subjects of \textit{jegda}-clauses in pre-matrix position is similar to that of the subjects of pre-matrix dative absolutes and conjunct participles regardless of subject coreference. Note that, even when we compare the three constructions in sentence-initial position (regardless of subject co-reference), we still do not observe significant differences.

\begin{figure}[!h]
\centering
\includegraphics[width=0.8\textwidth]{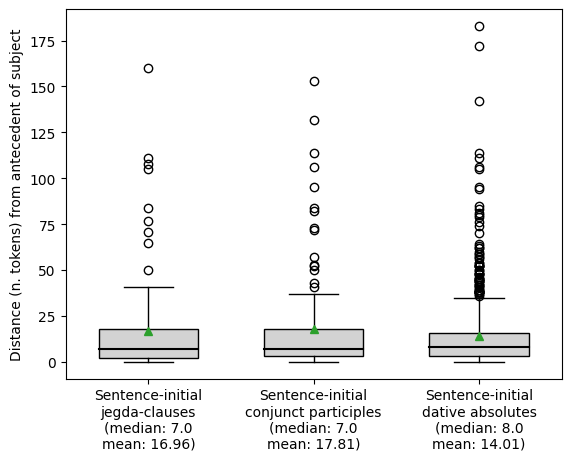}
\caption[Average anaphoric distance between subject and immediate antecedent in sentence-initial \textit{jegda}-clauses, conjunct participles and dative absolutes (all subjects)]{\label{distance_sentinitall}Average anaphoric distance between subject and immediate antecedent in sentence-initial \textit{jegda}-clauses, conjunct participles and dative absolutes (all subjects). Test results: 1. \textit{jegda}-clauses versus dative absolutes: Welch’s $t$-test: -0.71, $p$-value $=0.47$; one-tailed Mann–Whitney $U$-test: 2366.5, $p$-value $=0.22$; 2. \textit{jegda}-clauses versus conjunct participles: Welch’s $t$-test: 0.42, $p$-value $=0.67$; one-tailed Mann–Whitney $U$-test: 9768.5, $p$-value $=0.89$.}
\end{figure}

Some potentially interesting observations can instead be made on the pick-up \textit{rates} of the subject referents in the preceding discourse. In Section \ref{subjpositionsecdeep} of Chapter 1, we compared the average number of mentions (`relative saliency') of the subject referents of sentence-initial conjunct participles and dative absolutes in the preceding 1, 5, 30, and 60 sentences, to gauge whether there are differences in how many times the subject referents of participle constructions in sentence-initial position are mentioned in the near or distant preceding discourse. We observed that the subjects of sentence-initial participle constructions show overall very similar relative saliency. Similarly, Table \ref{relsaliencymeansmedstd_jegdaabs} compares the average relative saliency of subject referents in pre-matrix dative absolutes and pre-matrix non-coreferential \textit{jegda}-clauses. Table \ref{relsaliencymeansmedstd_jegdaxadv} compares the average relative saliency of (the leftmost) pre-matrix conjunct participles and pre-matrix \textit{jegda}-clauses sharing the subject with the matrix clause. For each window of preceding sentences (\textit{w}), the mean ($\bar{x}$), median ($med$) and standard deviation ($\sigma$) are provided, together with the results of the statistical tests (Welch’s $t$-test and one-tailed Mann–Whitney $U$-test) comparing the constructions at the relevant window.
% It seems reasonable, then, to consider null-subject \textit{jegda}-clauses to be affecting the results of the comparison between all subjects. Since the vast majority of the subjects of dative absolutes are overt, the former may then reflect a difference between overt and null subjects, whereby if the subject is 

% \caption{Comparison between the referent pick-up rates of overt subjects in dative absolutes and non-coreferential \textit{jegda}-clauses, in pre-matrix position, based on 1, 5, 30, and 60 preceding sentences.  The green triangle represents the mean. Test results: (a) Welch’s $t$-test: 3.02, $p<0.01$; one-tailed Mann–Whitney $U$-test: 2584, $p=0.03$. (b) Welch’s $t$-test: 1.36, $p=0.08$; one-tailed Mann–Whitney $U$-test: 2371.5, $p=0.16$. (c) Welch’s $t$-test: 0.80, $p=0.21$; one-tailed Mann–Whitney $U$-test: 2314, $p=0.22$. (d) Welch’s $t$-test: 0.27, $p=0.39$; one-tailed Mann–Whitney $U$-test: 2315, $p=0.22$.}

\begin{table}[!h]
\centering
\begin{tabular}{c|c|c|c|c|c|c|c|c|c|c|c|}
\cline{2-12}
 & \multirow{2}*{\textbf{\textit{w}}} & \multicolumn{3}{c|}{\textbf{non-coref. \textit{jegda}}} & \multicolumn{3}{c|}{\textbf{absolutes}} & \multicolumn{4}{c|}{\textbf{tests}} \\
\cline{3-12}
  & & $\bar{x}$ & $med$ & $\sigma$ & $\bar{x}$  & $med$ & $\sigma$ & $t$ & $p$ & $U$ & $p$ \\
\hline
\multicolumn{1}{|l|}{\multirow{4}{*}{\rotatebox[origin=c]{90}{\textbf{\textit{all subj.}}}}} & \textbf{1} & 1.08 &  1.0 &  0.98 & 1.39 &  1.0 &  1.09 & -2.21 &  \textbf{0.03} &  4912.5 &  \textbf{0.03}\\
\multicolumn{1}{|l|}{} & \textbf{5} & 2.52 &  1.0 &  2.96 & 3.24 &  3.0 &  2.99 & -1.75 &  0.08 &  4983.5 &  \textit{0.05}\\
\multicolumn{1}{|l|}{} & \textbf{30} &  8.6 &  1.0 &  11.8 & 9.27 &  5.0 &  10.38 & -0.42 &  0.67 &  5284.5 &  0.2\\
\multicolumn{1}{|l|}{}  & \textbf{60} & 12.08 &  1.0 &  18.53 & 13.36 &  5.0 &  17.79 & -0.51 &  0.61 &  5175.5 &  0.15\\
\hline
\hline
\multicolumn{1}{|l|}{\multirow{4}{*}{\rotatebox[origin=c]{90}{\textbf{\textit{overt}}}}} & \textbf{1} & 0.9 &  1.0 &  0.88 & 1.42 &  1.0 &  1.09 & -3.32 &  \textbf{<$\alpha$} &  2501.5 &  \textbf{<$\alpha$} \\
\multicolumn{1}{|l|}{} & \textbf{5} & 2.35 &  1.0 &  3.14 & 3.34 &  3.0 &  2.99 & -1.9 &  \textit{0.06} &  2612.0 &  \textbf{0.01}\\
\multicolumn{1}{|l|}{} & \textbf{30} & 6.56 &  1.0 &  10.48 & 9.56 &  6.0 &  10.44 & -1.72 &  0.09 &  2571.5 &  \textbf{0.01}\\
\multicolumn{1}{|l|}{}  & \textbf{60} & 7.85 &  1.0 &  13.13 & 13.81 &  6.0 &  17.94 & -2.46 &  \textbf{0.02} &  2490.5 &  \textbf{0.01}\\
\hline
\end{tabular}
\caption[Pick-up rates of subject referents in the previous discourse: pre-matrix non-coreferential \textit{jegda}-clauses and pre-matrix dative absolutes, by subject realization and window of preceding sentences]{Pick-up rates of subject referents in the previous discourse: pre-matrix non-coreferential \textit{jegda}-clauses and pre-matrix dative absolutes, by subject realization and window of preceding sentences. \textit{w} = window of preceding sentences; \textit{non-coref.} = non-coreferential; \textit{all subj}. =  null and overt subjects; \textit{overt} = overt subjects only; $\bar{x}$ = mean; $med$ = median; $\sigma$ = standard deviation; $t$ = Welch's $t$-test statistics; $U$ = two-tailed Mann-Whitney $U$-test; $p$ = $p$-value relative to the statistics to the left; <$\alpha$ = $<0.01$, i.e., highly significant; bold = significant; italics = marginally significant. The $t$-test statistics reflect the order of the column, so if it is negative, \textit{jegda}-clauses have a lower average value than absolutes; if positive, they have a higher average value.}
\label{relsaliencymeansmedstd_jegdaabs}
\end{table}

\begin{table}[!h]
\centering
\begin{tabular}{c|c|c|c|c|c|c|c|c|c|c|c|}
\cline{2-12}
 & \multirow{2}*{\textbf{\textit{w}}} & \multicolumn{3}{c|}{\textbf{coref. \textit{jegda}}} & \multicolumn{3}{c|}{\textbf{conjuncts}} & \multicolumn{4}{c|}{\textbf{tests}}\\
\cline{3-12}
  & & $\bar{x}$ & $med$ & $\sigma$ & $\bar{x}$  & $med$ & $\sigma$  & $t$ & $p$ & $U$ & $p$  \\
\hline
\multicolumn{1}{|l|}{\multirow{4}{*}{\rotatebox[origin=c]{90}{\textbf{\textit{all subj.}}}}} & \textbf{1} & 1.77 &  1.0 &  1.61 & 1.51 &  1.0 &  1.21 & 1.09 &  0.28 &  21730.5 &  0.53\\
\multicolumn{1}{|l|}{} & \textbf{5} & 4.17 &  3.5 &  3.39 & 3.5 &  3.0 &  3.08 & 1.33 &  0.19 &  23042.5 &  0.18\\
\multicolumn{1}{|l|}{} & \textbf{30} & 13.67 &  7.0 &  15.78 & 10.77 &  6.0 &  11.54  & 1.25 &  0.22 &  22451.0 &  0.31\\
\multicolumn{1}{|l|}{}  & \textbf{60} & 20.08 &  7.0 &  26.87 & 15.81 &  6.0 &  19.97 & 1.08 &  0.28 &  22256.5 &  0.37\\
\hline
\hline
\multicolumn{1}{|l|}{\multirow{4}{*}{\rotatebox[origin=c]{90}{\textbf{\textit{overt}}}}} & \textbf{1} & 1.05 &  1.0 &  0.65 & 1.7 &  1.0 &  1.14 & -2.1 &  \textbf{0.04} &  4737.0 &  0.39\\
\multicolumn{1}{|l|}{} & \textbf{5} & 3.0 &  2.5 &  2.65 & 3.73 &  3.0 &  2.99 & -0.56 &  0.58 &  5171.0 &  0.86\\
\multicolumn{1}{|l|}{} & \textbf{30} & 10.95 &  4.0 &  13.99 & 10.62 &  6.0 &  11.01 & 0.02 &  0.98 &  5204.0 &  0.9\\
\multicolumn{1}{|l|}{}  & \textbf{60} & 15.09 &  4.0 &  21.11 & 15.6 &  6.0 &  19.55 & -0.19 &  0.85 &  5198.5 &  0.89\\
\hline
\end{tabular}
\caption[Pick-up rates of subject referents in the previous discourse: pre-matrix coreferential \textit{jegda}-clauses and (leftmost) pre-matrix conjunct participles, by subject realization and window of preceding sentences]{Pick-up rates of subject referents in the previous discourse: pre-matrix coreferential \textit{jegda}-clauses and (leftmost) pre-matrix conjunct participles, by subject realization and window of preceding sentences; \textit{non-coref.} = non-coreferential; \textit{all subj}. =  null and overt subjects; \textit{overt} = overt subjects only; $\bar{x}$ = mean; $med$ = median; $\sigma$ = standard deviation; $t$ = Welch's $t$-test statistics; $U$ = two-tailed Mann-Whitney $U$-test; $p$ = $p$-value relative to the statistics to the left; <$\alpha$ = $<0.01$, i.e., highly significant; bold = significant; italics = marginally significant. The $t$-test statistics reflect the order of the column, so if it is negative, \textit{jegda}-clauses have a lower average value than conjunct participles; if positive, they have a higher average value.}
\label{relsaliencymeansmedstd_jegdaxadv}
\end{table}

As the test results indicate, there are no significant differences between the pick-up rates of the subject referents of pre-matrix coreferential \textit{jegda}-clauses and (the leftmost) pre-matrix conjunct participles, at any window of preceding sentences. Instead, when we compare pre-matrix non-coreferential \textit{jegda}-clauses and pre-matrix dative absolutes, we observe that the referents of the subjects of dative absolutes are overall significantly more activated in the previous discourse than the referents of the subjects of \textit{jegda}-clauses, with some differences depending on whether we consider all subjects (null \textit{and} overt) or overt subjects only. When we consider all subjects, we observe a significant difference in the number of mentions only if the immediately preceding sentence is considered and a marginally significant difference if the 5 preceding sentences are considered. Namely, when we consider the nearby preceding discourse, the subjects of pre-matrix dative absolutes are referentially more activated than the subjects of pre-matrix \textit{jegda}-clauses, regardless of realization (i.e. this is true of both null- and overt-subject absolutes and \textit{jegda}-clauses). When longer stretches of discourse are considered, there are instead no significant differences in their level of referential activation. When we only consider overt subjects, then the difference in relative saliency between the subjects of the two constructions is highly significant in both the preceding nearby discourse (1, 5 preceding sentences) and the preceding more distant discourse (30, 60 preceding sentences), with the subject referents of dative absolutes showing significantly more activation. \\
\indent Recall that \textit{jegda}-clauses equally have null and overt subjects, whereas dative absolutes in the Gospels virtually always have overt subjects. Null subjects are generally much more likely to signal topic continuation from the preceding discourse and are, therefore, generally the most likely candidate for topics in the sentence in which they appear (cf. \citealt{frascarelli2007, rizzi2018, didomenull2020}). If we cross the results on relative saliency with those on the average distance of subject antecedents, it is tempting to think that, when choosing a framing construction involving a change of subject from the \textsc{frame} itself to the matrix clause (as is most often the case with dative absolutes in the Codex Marianus and with non-coreferential \textit{jegda}-clauses), then, \textit{with very salient subjects} \textit{jegda}-clauses may be preferred if there is a topic continuation from the previous discourse, dative absolutes if there is a topic shift. This is potentially suggested by the equally high relative saliencies obtained from larger sentence windows and both null and overt subjects. However, \textit{jegda}-clauses may be preferred as topic-shifter when the \textsc{frame} involves \textit{less salient} subjects. This may instead be suggested by the fact that overt-subject non-coreferential \textit{jegda}-clauses are significantly less activated (at any window of preceding sentences) than overt-subject dative absolutes. \\
\indent The PROIEL treebank does not include annotation of complex information-structural categories such as topic or focus, since these are harder to assign with confidence. It does, however, include several other levels of annotation that can be used to collectively estimate the \textit{topicworthiness} of nominal referents relative to other nominals in the surrounding discourse. Topicworthiness is here meant as `an inherent property of an NP, such that it is more likely to serve as the topic (although it need not do so), and is closely related to such concepts as the hierarchies of grammatical person, animacy, definiteness and salience' (\pgcitealt{comrietriggerhappy}{329}). We can use a well-tested algorithm\footnote{The algorithm was written by Dag Haug and Hanne Eckhoff and is described in great detail in \citet{eckhoff2018b}.} to assign a \textit{topic score} measuring the relative topicworthiness of nominal referents based on properties known to correlate with aboutness topics cross-linguistically (\pgcitealt{eckhoff2018b}{34}). Summarizing from \citet{eckhoff2018b}, these properties and the respective scores for their values are:

\begin{itemize}
    \item givenness status, namely \textsc{old} (+15), \textsc{new} (+0), and different types of referent accessibility: \textsc{acc-inf} (accessible by inference, +10), \textsc{acc-gen} (accessible from world knowledge, +5), and \textsc{acc-sit} (accessible by deixis, +13);
    \item \textit{relative saliency}, measured as the number of mentions in the 30 preceding sentences (+10 for the candidate in the sentence with the highest relative saliency);
    \item word order (+15 to the linearly first topic candidate in the sentence, since topics tend to be fronted);
    \item realization: +30 for null subjects (except for those with a \textsc{non-spec}, +5 for personal pronouns, +5 for human proper nouns, \textsc{kind}, or \textsc{new} givenness status), since pro-drops, personal pronouns, and (human) proper nouns are more likely to be topics;
    \item syntactic relation: +10 for \textsc{sub} (since subjects are more likely to be topics), +5 for \textsc{obj}, +2 for \textsc{obl}, +1 for \textsc{comp} and \textsc{adv} clauses.
    \item animacy hierarchy: based on the animacy of referents established at lemma level (see \citealt{haug-etal-2009-computational}), +10 \textsc{human} (since nominals with human referents are more likely to be topics), +5 \textsc{org} (human, collective), +3 \textsc{animal}, +3 \textsc{concrete}, +0 \textsc{time}, \textsc{place}, \textsc{nonconc}, \textsc{veh}.
    \item the properties of the immediate antecedent and of the intervening competing candidates (+2 for the candidate whose immediate antecedent outranks intervening candidates on syntactic relation, and animacy and givenness hierarchies).
\end{itemize}

\begin{figure}[!h]
\begin{subfigure}{0.5\textwidth}
\includegraphics[width=0.9\linewidth, height=6cm]{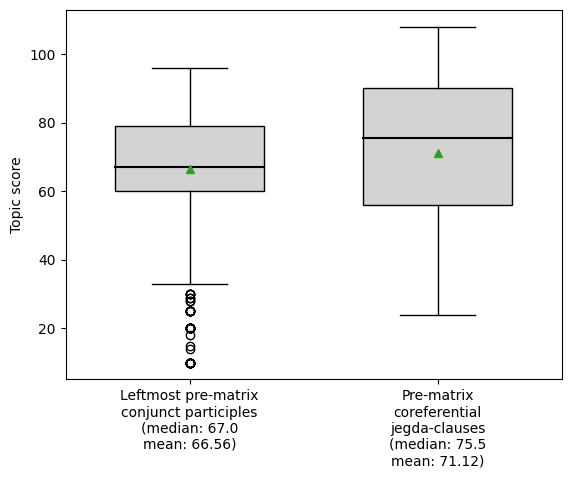} 
\caption[]{}
\label{topic_jegda_xadvs_all}
\end{subfigure}
\begin{subfigure}{0.5\textwidth}
\includegraphics[width=0.9\linewidth, height=6cm]{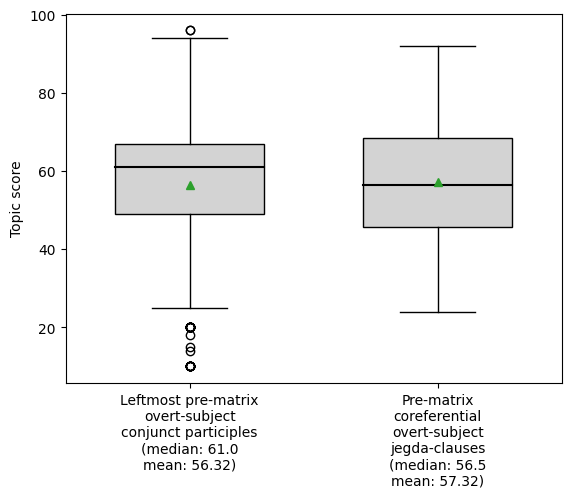}
\caption[]{}
\label{topic_jegda_xadvs_overt}
\end{subfigure}
\caption[Average topic score of subjects in (the leftmost) pre-matrix conjunct participles and pre-matrix coreferential \textit{jegda}-clauses, by subject realization]{Average topic score of subjects in (the leftmost) pre-matrix conjunct participles and pre-matrix coreferential \textit{jegda}-clauses, by subject realization. \textbf{(a.)} Null and overt subjects:  Welch’s $t$-test: 0.90, $p=0.37$; one-tailed Mann–Whitney $U$-test: 75994, $p=0.10$. \textbf{(b.)} Overt subjects only: Welch’s $t$-test: -0.10, $p=0.91$; one-tailed Mann–Whitney $U$-test: 25643, $p=0.42$.}
\end{figure}

\indent The subjects of coreferential pre-matrix \textit{jegda}-clauses and those of pre-matrix conjunct participles have overall very similar average topic scores, whether we consider all subjects (Figure \ref{topic_jegda_xadvs_all}) or overt subject only (Figure \ref{topic_jegda_xadvs_overt}).

\begin{figure}[!h]
\begin{subfigure}{0.5\textwidth}
\includegraphics[width=0.9\linewidth, height=6cm]{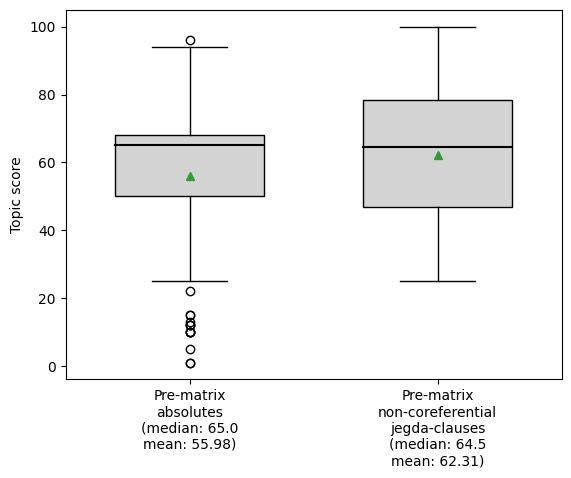} 
\caption[]{}
\label{topic_jegda_das_all}
\end{subfigure}
\begin{subfigure}{0.5\textwidth}
\includegraphics[width=0.9\linewidth, height=6cm]{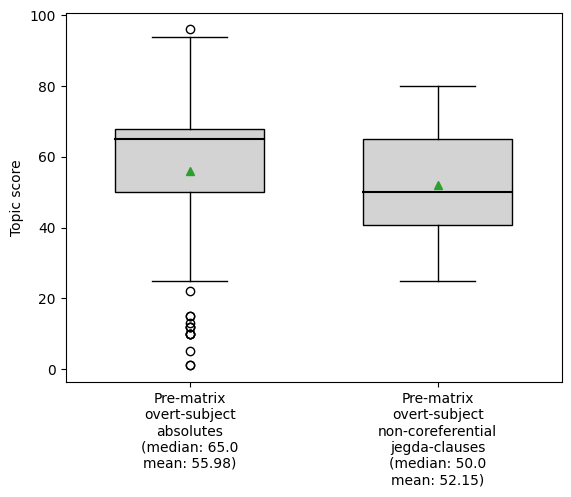}
\caption[]{}
\label{topic_jegda_das_overt}
\end{subfigure}
\caption[Average topic score of subjects in pre-matrix dative absolutes and pre-matrix non-coreferential \textit{jegda}-clauses, by subject realization]{Average topic score of subjects in pre-matrix dative absolutes and pre-matrix non-coreferential \textit{jegda}-clauses, by subject realization. \textbf{(a.)} Null and overt subjects: Welch’s $t$-test: -2.33, $p=0.01$; one-tailed Mann–Whitney $U$-test: 4563, $p=0.06$. \textbf{(b.)} Overt subjects only: Welch’s $t$-test: 1.41, $p=0.08$; one-tailed Mann–Whitney $U$-test: 3874, $p<0.01$.}
\end{figure}

The subjects of non-coreferential pre-matrix \textit{jegda}-clauses and pre-matrix dative absolutes show instead some differences depending on subject realization. Figures \ref{topic_jegda_das_all} and \ref{topic_jegda_das_overt} compare the average topic scores of subjects in non-coreferential pre-matrix \textit{jegda}-clauses and pre-matrix dative absolutes. Figure \ref{topic_jegda_das_all} includes the scores for both overt and null subjects, whereas Figure \ref{topic_jegda_das_overt} includes those for overt subjects only.

\indent On average, when considering both null and overt subjects, the subjects of \textit{jegda}-clauses have a higher topic score than the subjects of dative absolutes. According to the Mann–Whitney $U$-test, the difference is not significant or only marginally significant, with $p=0.06$, whereas according to Welch’s $t$-test the difference is highly significant at $p=0.01$. This is as expected, since, as we have seen, \textit{jegda}-clauses equally have null and overt subjects, whereas dative absolutes most often have an overt subject. Null subjects are much more likely candidates as aboutness topics in the sentence in which they appear and we generally mainly expect relatively topicworthy nominal referents to be able to be prodropped in the first place. \\
\indent Instead, when we only consider overt subjects, the subjects of dative absolutes have an overall greater topic score than the subjects of \textit{jegda}-clauses. In this case, according to Welch's $t$-test the difference is not significant at $p=0.08$, whereas according to the Mann–Whitney $U$-test, the difference is highly significant at $p<0.01$. These figures seem to support what we observed about the relative saliency of the subject referents in the two constructions. Namely, given a change in subject from a framing construction to the matrix clause (i.e. the need to use a non-coreferential adjunct), \textit{jegda}-clauses may preferred when there is a topic continuation from the previous discourse, which may license subject prodropping, or when there is a topic shift from the previous discourse but involving less salient or less topicworthy subjects. Dative absolutes may instead overall be preferred when there is a topic shift from the previous discourse, especially if more salient or more topicworthy subjects are involved. \\

\subsection{Aktionsart and lexical variation}\label{egda-var}
As seen in Chapter 1, a large portion of the Greek New Testament in PROIEL includes annotation on the Aktionsart of verbs, including the categories \textsc{activity}, \textsc{semelfactive}, \textsc{state}, and \textsc{telic}. In Section \ref{deep-variation} of Chapter 1, we saw that telic verbs are the most common among both conjunct participles and dative absolutes, although more prominently so among the former, whereas activity verbs are markedly more common among dative absolutes. Activity verbs are intuitively easily found as part of framing constructions (e.g. \textit{while they were singing}, \textit{when you pray}), whereas telic verbs can be expected to be particularly frequent in series of events with narrative progression (as is the case with \textsc{independent rhemes}).\\
\indent As Table \ref{aktionsartegda} shows, telic verbs are also the most frequent among \textit{jegda}-clauses. Stative and activity verbs are relatively more frequent among \textit{jegda}-clauses than conjunct participles (where they represented 14.82\% and 3.58\% of the verbs, respectively). This makes the Aktionsarten of verbs among \textit{jegda}-clauses overall more similar to dative absolutes, even though activity verbs remain markedly more frequent among the latter (20.47\% of all dative absolutes). The overall similarity between absolutes and \textit{jegda}-clauses in this respect is not surprising, since both activity and stative verbs can be expected to occur in framing constructions relatively often.

\begin{table}[!h]
\centering
\begin{tabular}{cc|c|c|c|c|c|}
\cline{3-7}
 & & \multicolumn{4}{c|}{\textbf{matrix}} & \multirow{2}*{\textbf{\textit{tot.}}}\\
\cline{3-6}
  & & \textsc{act} & \textsc{sem}   & \textsc{st}  & \textsc{tel} & \\
\hline
\multicolumn{1}{|l|}{\multirow{4}{*}{\rotatebox[origin=c]{90}{\textbf{\textit{jegda}}}}} & \textsc{act} & 1.38 (2)  & 0 & 2.1 (3) & 6.9\% (10) & 10.34\% (15) \\
\multicolumn{1}{|l|}{} & \textsc{sem} & 0 & 0 & 0 & 0 & 0\\
\multicolumn{1}{|l|}{} & \textsc{st} & 2.1 (3) & 0 & 8.27\% (12) & 9\% (13)& 19.31\% (28) \\
\multicolumn{1}{|l|}{}  & \textsc{tel} & 3.4\% (5) & 0 & 11.72\% (17) & 55.17\% (80) & 70.34\% (102)\\
\hline
\end{tabular}
\caption[Aktionsarten of \textit{jegda}-clauses and their matrix clause in the Codex Marianus]{Aktionsarten of \textit{jegda}-clauses and their matrix clause in the Codex Marianus. \textsc{act} = activity, \textsc{sem} = semelfactive, \textsc{st} = state, \textsc{tel} = telic.}
\label{aktionsartegda}
\end{table}

\begin{figure}[!h]
\begin{subfigure}{0.5\textwidth}
\includegraphics[width=0.9\linewidth, height=6cm]{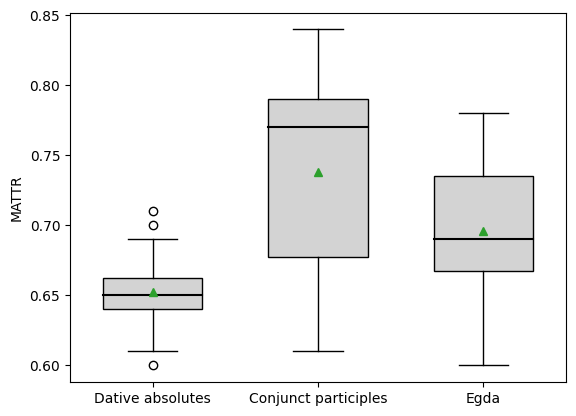} 
\caption[]{}
\label{egdamattrsverb}
\end{subfigure}
\begin{subfigure}{0.5\textwidth}
\includegraphics[width=0.9\linewidth, height=6cm]{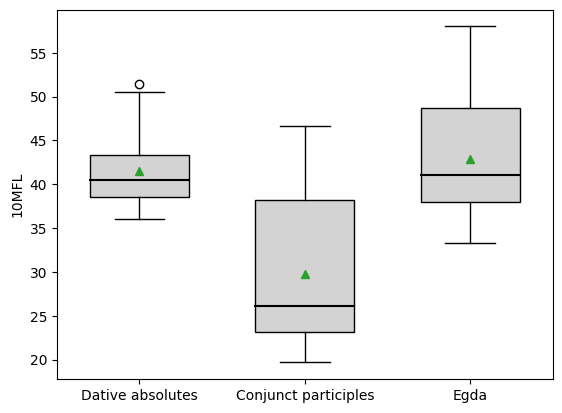}
\caption[]{}
\label{egda10mflverb}
\end{subfigure}
\caption[Lexical variation among verbs in \textit{jegda}-clauses, conjunct participles and dative absolutes]{Lexical variation among verbs in \textit{jegda}-clauses, conjunct participles and dative absolutes. \textbf{(a.)} average moving-average type-token ratio (MATTR). Test results: 1. \textit{jegda}-clauses versus dative absolutes: Welch’s $t$-test: -3.16, $p<0.01$; one-tailed Mann–Whitney $U$-test: 93, $p<0.01$; 2. \textit{jegda}-clauses versus conjunct participles: Welch’s $t$-test: 2.1, $p=0.04$; one-tailed Mann–Whitney $U$-test: 283, $p=0.02$. \textbf{(b.)}: average percentage of subjects belonging to the ten most frequent lemmas (10MFL). Test results: 1. \textit{jegda}-clauses versus dative absolutes: Welch’s $t$-test: -0.8, $p=0.4$; one-tailed Mann–Whitney $U$-test: 186, $p=0.7$; 2. \textit{jegda}-clauses versus conjunct participles: Welch’s $t$-test: -5.1, $p<0.01$; one-tailed Mann–Whitney $U$-test: 72, $p<0.01$.}
\label{egdamattrs10mflverb}
\end{figure}

The analysis of the lexical variation among participles in Section \ref{lexicalvarptcpshallowsec} of Chapter 2 also provided additional evidence regarding the different functions of conjunct participles and dative absolutes. As in Chapter 2, Table \ref{lexvar} shows the percentage of verbs belonging to the ten most frequent lemmas (10MFL), as well as the MATTR with a moving window of 40 lemmas, for each construction in different subcorpora and configurations. The values for absolutes and conjunct participles are repeated from Chapter 2 for ease of comparison.  Figures \ref{egdamattrsverb} and \ref{egda10mflverb} compare the average 10MFL and MATTR from Table \ref{lexvar} for each construction.

\begin{sidewaystable}
\centering
\begin{tabular}{|c|c|c|c|c|c|c|}
\hline
\multirow{2}*{\textbf{Subsample}} & \multicolumn{2}{|c|}{\textbf{Absolutes}} & \multicolumn{2}{|c|}{\textbf{Conjuncts}} & \multicolumn{2}{|c|}{\textbf{Jegda}}\\
\cline{2-7}
 & \textbf{10MFL} & \textbf{MATTR} & \textbf{10MFL}   & \textbf{MATTR} & \textbf{10MFL}   & \textbf{MATTR}\\
\hline
\textbf{(O)CS + (O)ES + Mar. (non-N)} & 36.15\% & 0.65 & 20.51\% & 0.78 & 35.15\% & 0.69\\ 
\textbf{(O)CS + (O)ES + Mar. (N)} & 38.53\% & 0.65 & 24.29\% & 0.78 & 38.12\% & 0.69\\ 
\textbf{(O)CS + (O)ES (non-N)} & 36.05\% & 0.65 & 19.67\% & 0.79 & 33.33\% & 0.73\\ 
\textbf{(O)CS + (O)ES (N) }& 38.46\% & 0.65 & 23.20\% & 0.79 & 35.71\% & 0.72\\ 
\textbf{(O)CS + Mar.} & 42.74\% & 0.64 & 25.12\% & 0.77 & 41.38\% & 0.65\\ 
\textbf{(O)CS} & 43.31\% & 0.64 & 23.32\% & 0.80 & 40.72\% & 0.67\\ 
\textbf{(O)ES} & 37.93\% & 0.66 & 23.49\% & 0.78 & 37.65\% & 0.78\\ 
\textbf{Mar.} & 50.54\% & 0.61 & 36.79\% & 0.69 & 50\% & 0.60\\ 
\textbf{(O)CS + (O)ES + Mar. (left, N)} & 39.83\% & 0.65 & 28.72\% & 0.74 & 40.62\% & 0.68\\ 
\textbf{(O)CS + (O)ES + Mar. (right, N)} & 40.71\% & 0.69 & 22.90\% & 0.80 & 48.42\% & 0.69\\ 
\textbf{(O)CS + (O)ES (left, N)} & 39.36\% & 0.66 & 27.22\% & 0.76 & 36.08\% & 0.72\\ 
\textbf{(O)CS + (O)ES (right, N)} & 41.31\% & 0.70 & 22.85\% & 0.80 & 49.32\% & 0.68\\ 
\textbf{(O)CS (left)} & 44.22\% & 0.64 & 27.61\% & 0.77 & 43.96\% & 0.68\\ 
\textbf{(O)CS (right)} & 45.78\% & 0.71 & 19.98\% & 0.84 & 50.85\% & 0.66\\ 
\textbf{(O)ES (left) }& 39.47\% & 0.67 & 27.20\% & 0.75 & 40.30\% & 0.77\\ 
\textbf{(O)ES (right)} & 46.92\% & 0.67 & 25.17\% & 0.77 & \textit{NA} & \textit{NA}\\ 
\textbf{Mar. (left)} & 51.46\% & 0.60 & 44.12\% & 0.63 & 53.08\% & 0.60\\ 
\textbf{Mar. (right) }& \textit{NA} & \textit{NA} & 30.11\% & 0.78 & \textit{NA} & \textit{NA} \\ 
\hline
\end{tabular}
\caption[Lexical variation among the verbs in \textit{jegda}-clauses and participle constructions, by subcorpus and configuration]{Lexical variation among the verbs in \textit{jegda}-clauses and participle constructions, by subcorpus and configuration. \textit{10MFL} = ten most frequent lemmas; \textit{MATTR} = moving-average type-token ratio; \textit{Mar} = Codex Marianus; (O)ES = Old East Slavic and Middle Russian; \textit{left} and \textit{right} = pre-matrix and post matrix, respectively; \textit{N} = normalized. Note that the closer the MATTR is to 1 (and the lower the percentage for the 10MFL), the greater the lexical variation.}
\label{lexvar}
\end{sidewaystable}

When adding \textit{jegda}-clauses to the picture, it appears that \textit{jegda}-clauses, overall, occupy an intermediate position between conjunct participles and dative absolutes. Namely, lexical variation among \textit{jegda}-clauses is generally greater than or equal to dative absolutes but smaller than variation among conjunct participles. As the statistics in Figure \ref{egdamattrs10mflverb} indicate, we get a significant difference between conjunct participles and both \textit{jegda}-clauses and dative absolutes, according to both the MATTR and the 10MFL scores. This is as expected, since the most typical functions of conjunct participles are as \textsc{independent rhemes} and as \textsc{elaborations}, both of which normally encode new information, as opposed to absolutes and \textit{jegda}-clauses, which are more typically framing constructions and as such refer to previously mentioned or inferrable events, resulting in overall more `predictable' predications.\\
\indent Variation among dative absolutes is only significantly smaller than \textit{jegda}-clauses when considering the MATTR but not the 10MFL value. Judging from the types of variation the two metrics tend to capture, this suggests that, similarly to absolutes, many \textit{jegda}-clauses are accounted for by the 10 most frequent lemmas, that is, just like absolutes, \textit{jegda}-clauses are used very often with the same few verbs. However, outside of those 10 lemmas, \textit{jegda}-clauses show significantly more variation than absolutes, as the difference in MATTR indicates. This suggests that \textit{jegda}-clauses, while being overall functionally very similar to dative absolutes, may also function as less `predictable' frames than absolutes, and also points to the nearly-formulaic character of absolutes, which often occur in what appears to be fixed or semi-fixed verb-subject combinations.

\subsection{Clause length}\label{lengthsec}
\pgcitet{behrens2012a}{182} suggest that finite adverbial clauses, having explicit tense and a subordinating conjunction, tend to make a ‘less context-dependent, more precise contribution’ to the interpretation of the sentence in which they occur than their non-finite competitors, ‘highlighting one single aspect of the relation between the events or propositions involved, or narrowing down the set of possible interpretations’ (cf. also \citealt{s2011a}). According to \citeauthor{behrens2012a}, non-finite clauses may therefore be chosen merely because of their semantic underspecifity and economy from the production perspective, since they typically distinguish fewer morphosyntactic categories than finite competitors. In the case of Old Church Slavonic and Greek, participles make aspect, but not tense, distinctions. They are, however, inflected for gender and case, unlike finite verb forms. \citet{behrens2012a} also claim that non-finite adjuncts are less explicit in the semantic dimensions relevant to finite adverbial clauses because the former are shorter than the latter, since they lack a subjunction and `have no explicit subject'. This is obviously referring specifically to \textit{open} adjuncts (in our case, conjunct participles, but not absolute constructions), which share their subject with the matrix clause, but they also claim that the observations they make on the basis of open adjuncts are overall valid for \textit{closed} adjuncts, too (i.e. absolutes) (\pgcitealt{behrens2012a}{197--198}). We can test the claims on the length of participle adjuncts and \textit{jegda}-clauses by leveraging dependency annotation to count the number of tokens occurring inside participle clauses and \textit{jegda}-clauses. \pgcitet{haug2012a}{314} already compared the length of Ancient Greek participle clauses in different configurations (e.g. SV, VS, pre-matrix, post-matrix) to test the intuition that sentence-initial participles, particularly VS ones, should be shorter than others since they are more likely to function as \textsc{frames}, which introduce known or presupposed events used to anchor other events and are therefore expected to be more `predictable'. His results confirmed that this was, in fact, the case.\\
\indent We can reproduce the same figures as those in \citet{haug2012a} on Early Slavic, while also adding \textit{jegda}-clauses, to test whether \posscitet{behrens2012a} claim on the length of non-finite adjuncts and finite adverbials applies in our case. Mean, median, and standard deviation of the length of clauses in different configurations are provided in Table \ref{clauselenghtstab}.

\begin{table}[!h]
\centering
\begin{tabular}{cccc}
\hline
 & \textbf{mean} & \textbf{median} & \textbf{standard deviation}\\
\hline
\textbf{sub\textunderscore ptcp} & 3.65 & 3.0 & 3.17\\ 
\textbf{ptcp\textunderscore sub} & 3.39 & 3.0 & 2.66\\ 
\textbf{sub\textunderscore da} & 4.62 & 4.0 & 3.24\\ 
\textbf{da\textunderscore sub }& 4.67 & 4.0 & 3.31\\ 
\textbf{sub\textunderscore xadv} & 4.13 & 3.0 & 3.93\\ 
\textbf{xadv\textunderscore sub} & 3.17 & 2.5 & 2.53\\ 
\textbf{da\textunderscore left} & 4.2 & 4.0 & 2.53\\ 
\textbf{da\textunderscore right} & 4.87 & 4.0 & 4.14\\ 
\textbf{xadv\textunderscore left} & 3.23 & 3.0 & 2.5\\ 
\textbf{xadv\textunderscore right} & 4.63 & 3.0 & 4.4\\ 
\textbf{jegda\textunderscore left} & 5.22 & 4.0 & 3.0\\ 
\textbf{jegda\textunderscore right} & 5.6 & 5.0 & 2.94\\ 
\hline
\end{tabular}
\caption[Mean, median and standard deviation of the length (in number of tokens) of participle constructions and \textit{jegda}-clauses in different configurations]{Mean, median and standard deviation of the length (in number of tokens) of participle constructions and \textit{jegda}-clauses in different configurations. \textit{sub\textunderscore ptcp} = sentence-initial SV participle constructions (absolutes and conjunct participles), \textit{ptcp\textunderscore sub} = sentence-initial VS participle constructions, \textit{sub\textunderscore da} = sentence-initial SV absolutes, \textit{da\textunderscore sub} = sentence-initial VS absolutes, \textit{sub\textunderscore xadv} = sentence-initial SV conjunct participles, \textit{xadv\textunderscore sub} = sentence-initial VS conjunct participles, \textit{da\textunderscore left} = pre-matrix absolutes, \textit{da\textunderscore right} = post-matrix absolutes, \textit{xadv\textunderscore left} = pre-matrix conjunct participles, \textit{xadv\textunderscore right} = post-matrix conjunct participles, \textit{jegda\textunderscore left} = pre-matrix \textit{jegda}-clauses, \textit{jegda\textunderscore right} = post-matrix \textit{jegda}-clauses.}
\label{clauselenghtstab}
\end{table}

The clear-cut differences between participle configurations observed by \citet{haug2012a} on Ancient Greek are also found in the Early Slavic data to a large extent. \pgcitet{haug2012a}{315} reports that sentence-initial VS participle constructions (`ptcp\textunderscore sub' in the table) and post-matrix participles as a whole (`ptcp\textunderscore right') stand out as significantly shorter and longer than the average, respectively, which fits the observation that \textsc{frames} are presupposed or known eventualities, hence presumably `more predictable' and thus shorter on average, whereas \textsc{elaborations} convey new information and are thus generally expected to involve longer clauses. As Table \ref{clauselenghtstab} shows, both absolutes and conjunct participles are shorter, on average, in pre-matrix than in post-matrix position.\footnote{Pre-matrix absolutes versus post-matrix absolutes: Welch’s $t$-test: -2.34, \textit{\textbf{p=}}\textbf{0.01}; one-tailed Mann–Whitney $U$-test: 127016, $p=0.10$. Pre-matrix conjunct participles versus post-matrix conjunct participles: Welch’s $t$-test: -11.64, \textbf{\textit{p}<0.01}; one-tailed Mann–Whitney $U$-test: 10382926.5, \textbf{\textit{p}<0.01}.} Sentence-initial SV conjunct participles are also significantly longer than VS conjunct participles.\footnote{Welch’s $t$-test: 9.71, \textbf{\textit{p}<0.01}; one-tailed Mann–Whitney $U$-test: 10449661, \textbf{\textit{p}<0.01}.} Among sentence-initial dative absolutes, the difference in length between those in the SV and VS configuration is instead not statistically significant.\footnote{Welch’s $t$-test: -0.33, \textit{p}=0.73; one-tailed Mann–Whitney $U$-test: 318913.5, \textit{p}=0.65.} The average length of \textit{jegda}-clauses, in both pre- and post-matrix position, although particularly in the post-matrix one (`egda\textunderscore right' in the table), is greater than participle constructions in any configurations. This suggests that, although the lack of an explicit subject or a subjunction does not exclude relatively long clauses, it does affect the average length. The length of absolute constructions, in fact, also stands out, overall, compared to that of conjunct participles. Once again, the presence of an overt, internal subject in absolute constructions in the vast majority of cases is likely to have an impact on the average length. 

\subsection{Summary}
The analysis of the corpus data in this section has pointed to some potential functional differences between \textit{jegda}-clauses and participle constructions. \\
\indent \textit{Jegda}-clauses appear to be distributionally more similar to dative absolutes, which reflects the intuition that their typical function is as framing, backgrounding devices, which is also supported by the fact that, similarly to absolutes, \textit{jegda}-clauses are significantly more frequent in pre-matrix than post-matrix position. This is to be expected from frames since they set the scene for the discourse to come and are thus more likely to occur sentence-initially.\\
\indent We saw that establishing the aspect of \textit{jegda}-clauses is not as straightforward as with participle constructions due to frequent present-tense forms, which cannot always be confidently disambiguated as perfective or imperfective on the basis of morphology alone since the modern Slavic derivational-aspect system had not fully developed in Old Church Slavonic. Although this makes a direct comparison of aspect distribution among participles and \textit{jegda}-clauses using quantitative methods quite challenging, manual annotation of a subset of our dataset with information on aspect showed no significant difference in the number of perfectives and imperfectives among \textit{jegda}-clauses.\\
\indent The analysis of some of the properties of subjects among \textit{jegda}-clauses indicated that there is overall more evident competition with absolutes than with conjunct participles. In the Gospels, \textit{jegda}-clauses with a different subject than their matrix were found to be significantly more frequent than those with co-referential subjects. However, null subjects are not infrequent even among \textit{jegda}-clauses with a different subject from their matrix, suggesting that \textit{jegda}-clauses may be preferred when there is a topic continuation from the immediately preceding discourse. This also seems to be confirmed, quite unsurprisingly, when we compare the average topic scores of subject referents in non-coreferential \textit{jegda}-clauses and dative absolutes, the former being significantly higher than the latter. When we only consider overt subjects, however, it becomes clear that this result is chiefly obtained \textit{because} of null-subject \textit{jegda}-clauses. With overt subjects, which we can take to indicate potential topic-shifting (though, of course, not necessarily), their difference in topic score is not significant, and the subjects of dative absolutes are found to have referents that are significantly more activated in the previous sentence than \textit{jegda}-clauses. This may point to the frequent topic-shifting function of dative absolutes observed in Chapter 2, whereby a highly salient referent in the immediately preceding discourse, possibly part of a rhemic constituent, is repurposed as the topic of a new portion of discourse. \\
\indent Finally, the analysis of lexical variation among the verbs indicates that \textit{jegda}-clauses may be somewhat less predictable frames than absolute constructions, as suggested by the significantly greater MATTR among the former than the latter. However, when using the ten most frequent lemmas as the criterion, we observe no significant difference in lexical variation among the two, regardless of subcorpus or configuration, which suggests that, similarly to absolutes, \textit{jegda}-clauses involve few lemmas occurring very often, but unlike the latter, they also involve several lemmas occurring very infrequently. Compared to conjunct participles, however, \textit{jegda}-clauses show significantly smaller lexical variation among verbs on both fronts, as was also expected given the more typical \textsc{independent rheme} or \textsc{elaboration} function of conjunct participles, which are thus more likely to introduce new, less predictable information.

\chapter{Temporal relations and discourse representation}

The previous chapter provided a descriptive and quantitative analysis of \textit{jegda}-clauses as they appear in the Early Slavic corpus and focussed on variables that could capture differences in their properties compared to dative absolutes and conjunct participles, as they had emerged from the analyses in Chapters 1 and 2. The analysis of their distribution in the sentence and of the properties of their subjects suggested that \textit{jegda}-clauses are, overall, functionally more similar to dative absolutes, with significant differences in the information status of their typical subjects. In particular, the results on subject realization and the analysis of the relative saliency and topicworthiness of their subject referents suggested that \textit{jegda}-clauses may be preferred over dative absolutes when there is a topic continuation from the immediately preceding discourse or when there is a topic shift from the previous discourse but involving less salient or less topicworthy subjects. The analysis of the lexical variation among verbs indicated that \textit{jegda}-clauses may also occur as much less `predictable' \textsc{frames} than dative absolutes, which instead seem to be formed from a much more limited pool of lemmas and not infrequently appear in fixed or formulaic verb-subject combinations.\\
\indent This chapter discusses the temporal interpretation of \textit{jegda}-clauses based on the different attested combinations of tense-aspect between the \textit{jegda}-clause and its matrix clause (Section \ref{egda-tenseaspect}). It also proposes a possible formalization of the rhetorical relations typically introduced by \textit{jegda}-clauses and participle constructions within an SDRT framework (Section \ref{egdarethorical}). 

\section{\textit{Jegda}-clauses and temporal relations interpretation}\label{egda-tenseaspect}
As shown in Table \ref{egdamatrtense}, most aorist \textit{jegda}-clauses have an aorist matrix clause and most present-tense \textit{jegda}-clauses have a present-tense matrix clause. Imperfect \textit{jegda}-clauses can occur with both an aorist matrix or an imperfect matrix clause. This is unlike Ancient Greek, where an imperfect \textit{hote}/\textit{hotan}-clause with an aorist matrix is instead exceedingly rare.\footnote{As observed in the Introduction, only 1 occurrence of pre-matrix \textit{hóte}-clause with an imperfect verb followed by an aorist main verb was found (in Herodotus), and none of \textit{hótan}.}

\begin{table}[!h]
\centering
\begin{tabular}{c|cccccc}
\hline
\backslashbox{egda}{matrix} & \textbf{aorist} & \textbf{future} & \textbf{imperfect} & \textbf{present} & NA\\
\hline
\textbf{aorist} &       26.5\% (102)  & -            & 7\% (29)    & 1.3\% (6)    & 2.4\% (12) \\
\textbf{future} &       -           & 0.7\% (3)    & -              & 3.9\% (13)   & - \\
\textbf{imperfect} &    5.4\% (22) & -            & 8.5\% (37)    & 0.6\% (3)    & 1.6\% (8) \\
\textbf{present} &      1.6\% (8)  & 2.4\% (9)    & -              & 37.5\% (150)  & 0.5\% (2) \\
\hline
\end{tabular}
\caption{Tense of \textit{jegda}-clauses and their matrix clause in the Codex Marianus and standard treebanks (combined)}
\label{egdamatrtense}
\end{table}

In the past tense, an aorist \textit{jegda}-clause followed by an aorist main clause normally produces narrative progression, whereby the matrix event is interpreted as occurring after the \textit{jegda}-event. (\ref{aoraor1}) and (\ref{aoraor2}) are two such examples.

\begin{example}
    \gll i \textbf{egda} \textbf{porǫgašę} sę emu. s\foreignlanguage{russian}{ъ}vlěšę s\foreignlanguage{russian}{ъ} nego chlamidǫ i oblěšę i v\foreignlanguage{russian}{ъ} rizy svoję.
    {and} {when} {mock.{\sc aor.3.sg}} {\sc refl} {\sc 3.sg.m.dat} {take off.{\sc aor.3.pl}} {from} {\sc 3.sg.m.gen} {robe.{\sc sg.f.acc}} {and} {dress.{\sc aor.3.pl}} {\sc 3.sg.m.acc} {in} {clothes.{\sc pl.f.acc}} {own.{\sc 3.pl.acc}}
    \glt `After they had mocked him, they took off the robe and put his own clothes on him' (Matthew 27:31)
    \glend
    \label{aoraor1}
\end{example}

\begin{example}
    \gll \textbf{Egda} že \textbf{slyša} pilat\foreignlanguage{russian}{ъ} se slovo. pače ouboě sę
    {when} {\sc ptc} {hear.{\sc aor.3.sg}} {Pilate.{\sc nom}} {this.{\sc sg.n.acc}} {word.{\sc sg.n.acc}} {more.{\sc comp}} {be frightened.{\sc aor.3.sg}} {\sc refl}
    \glt `When Pilate heard this, he was even more afraid' (John 19:8–9) %23242 Greek
    \glend
    \label{aoraor2}
\end{example}

According to \citet{partee84}, a pre-posed eventive \textit{when}-clause introduces a new temporal referent locating the main eventuality \textit{just} \textit{after} the event described by the \textit{when}-clause. This is the temporal effect we see in (\ref{aoraor1}) and (\ref{aoraor2}), and it is similar to that created by perfective \textsc{frames} and \textsc{independent rhemes}. \pgcitet{hinrichs86}{75--76} instead claimed that two eventive clauses (accomplishments and achievements, that is, \textit{telic} expressions) linked by \textit{when} in English can have any temporal order with respect to each other and brings (\ref{hinrichs}a)-(\ref{hinrichs}c) as evidence of this. 

\begin{figure}[!h]
    \centering
\includegraphics[scale=0.5]{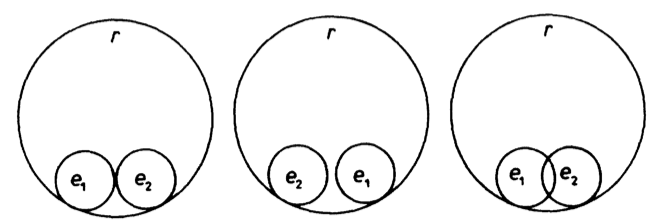}
    \caption{Possible orders of a non-durative \textit{when}-clause and its non-durative matrix clause according to \pgcitet{hinrichs86}{78}}
    \label{hinrichstelics}
\end{figure}

\begin{example}
    \begin{itemize}
        \item[a.] John broke his arm when he wrecked the Pinto.
        \item[b.] When the Smiths moved in, they threw a party.
        \item[c.] When the Smiths threw a party, they invited all their old friends.
    \end{itemize}
    \label{hinrichs}
\end{example}
 
where (\ref{hinrichs}a), according to Hinrichs, conveys simultaneity, whereas in (\ref{hinrichs}b) and (\ref{hinrichs}c) the event described by the \textit{when}-clause precedes and follows the one described by the main clause, respectively. The argument is that when both the \textit{when}-clause and the main clause are non-durative, then they are \textit{both} contained within the reference time, or \textit{frame} (\pgcitealt{hinrichs86}{78}), introduced by \textit{when}. Inside this reference time, the two events can have any order, as represented in Figure \ref{hinrichstelics} from \pgcitet{hinrichs86}{75}.

There are some clear shortcomings in Hinrichs' account---apart from relying only on native speakers' intuitions (cf. \pgcitealt{hinrichs86}{74}). First, as also noted by \citet{sandstrom}, he seemingly considers the \textit{when}-clause and the main clause equal, rather than the first being subordinated to the second, as is clear from Hinrichs disregarding the order of the clauses in (\ref{hinrichs}a), which he compares to (\ref{hinrichs}b)-(\ref{hinrichs}c), even if the linear order has obvious importance in the organization of discourse, which, in turn, has clear effects on the range of possible readings. Another weakness of \possciteauthor{hinrichs86} account is that, despite working within a coarse stative versus eventive binary ontology as in DRT, it seems to make an equivalence between lack of progressive form on telic predicates in English and eventivity. Arguably, the \textit{when}-clause in (\ref{hinrichs}a), while being a telic, bounded expression, is pragmatically not the same as the one in (\ref{hinrichs}b), for example, which is also telic and bounded. Even DRT and SDRT, which strictly work within that same coarse eventive-stative binary, recognize that while in languages such as French and English the distinction is \textit{generally} marked by tense in narrative texts, it can result from a combination of several factors, including, crucially, the larger discourse context, so that finer-grained distinctions are often needed to distinguish the two (cf. \citealt{kamp1993a,swart95,swart98,caudal05}). A more satisfactory account, in some respects, of the possible temporal interpretation of \textit{when}-clauses is the one by \citet{declerck}, who gives default, `unmarked' interpretations based on the combination of durativity, telicity, and boundedness, but always leaves room for pragmatic incompatibility, where \textit{generally} another default interpretation is given. A recurring `exception' to the default interpretation of different combinations of \textit{when}-clause and main clause is what \citeauthor{declerck} calls `occasion'-referring \textit{when}-clauses, which he describes as those not introducing a time referent, but defining an occasion in which the main eventuality actualizes, as in examples (\ref{declerckoccasion1})-(\ref{declerckoccasion3}), from \pgcitet{declerck}{218-226}.

\begin{example}
    When John built his sailboat, he bought some timber from me
    \label{declerckoccasion1}
\end{example}

\begin{example}
    The chairman resigned when the secretary wrote a libellous article about him in the local newspaper
    \label{declerckoccasion2}
\end{example}

\begin{example}
    When he found the lost ring, he also found some secret letters
    \label{declerckoccasion3}
\end{example}

In such cases, according to \citeauthor{declerck}, the order in which the two eventualities occur can remain unspecified. In a way, this is very similar to what is claimed by \citet{hinrichs86}, except that \textit{inclusion} of one unbounded eventuality into another (the \textit{when}-situation into the main situation or vice versa) does not appear to be an option in his account, as Figure \ref{hinrichstelics} indicates. I argue that what \citeauthor{declerck} refers to as \textit{occasions} can often be loosely represented as durative, but bounded, even when the \textit{when}-clause in isolation is punctual, in which case a non-durative (punctual) matrix can be included in the \textit{when}-situation. This is arguably the case in (\ref{declerckoccasion3}), as well as (\ref{hinrichs}a), as already discussed, but also (\ref{hinrichs}c). Example (\ref{hinrichs}c) is also discussed by \citet{sandstrom}, among others, to show how temporal inclusion of an eventive main clause into an eventive \textit{when}-clause is not only possible, but it involves a part-of relation, whereby the main clause event is a subevent, or a constitutive part (\citealt{moens}), of the \textit{when}-clause event (but not vice-versa). Note that the part-of relation between an eventive when clause and an eventive main clause is meant by \citet{sandstrom} in terms of Boolean inclusion $\subseteq$, whereby identity relations are also considered as part-of relations while keeping the order of \textit{when}-clause and main clause relevant. For example, \ref{subevent}a), from \pgcitet{sandstrom}{197}, is obviously not the same as (\ref{subevent}b).

\begin{example}
\begin{itemize}
    \item[a.] When Sue killed her husband, she put poison in his whisky.
    \item[b.] When Sue put poison in her husband's whiskey, she killed him.\footnote{\citet{sandstrom} gives the example \textit{When Sue put poison in her husband's whisky, he died}, but (\ref{subevent}b) is properly parallel to (\ref{subevent}a) and gives a clearer impression of the unidirectionality of the inclusion relation.}
\end{itemize}
    \label{subevent}
\end{example}

In (\ref{subevent}a), or [when $e_1$, $e_2$], the subevent relation $e_1$$\subseteq$$e_2$ is in fact identity, whereby the poisoning constitutes the killing. Note, however, that the \textit{when}-clause and the main clause are not equal, so it \textit{does} matter what goes into what. If we switch the order in which the eventualities are presented, we do not get a $\subseteq$ relation, but a consequentiality one, temporally $e_1$$\prec$$e_2$.\\
\indent The subevent relation between an eventive \textit{jegda}-clause and an eventive main clause can be exemplified by (\ref{egdaoccasion}).

\begin{example}
\begin{itemize}
    \item[a.]
    \gll Byst\foreignlanguage{russian}{ъ} že \textbf{egda} \textbf{kr\foreignlanguage{russian}{ъ}stišę} \textbf{sę} v\foreignlanguage{russian}{ь}si ljud\foreignlanguage{russian}{ь}e. isou kr\foreignlanguage{russian}{ь}št\foreignlanguage{russian}{ь}šju sę i molęštju sę. otvr\foreignlanguage{russian}{ъ}ze sę nebo
    {happen.{\sc aor.3.sg}} {\sc ptc} {when} {baptize.{\sc aor.3.pl}} {\sc refl} {all.{\sc pl.m.nom}} {people.{\sc pl.m.nom}} {Jesus.{\sc dat}} {baptize.{\sc ptcp.pfv.m.sg.dat}} {\sc refl} {and} {pray.{\sc ptcp.ipfv.m.sg.dat}} {\sc refl} {open.{\sc aor.3.sg}} {\sc refl} {heaven.{\sc sg.n.nom}}
    \glt
    \glend
    \item[b.]
    \gll {Egeneto} {de} {\textbf{en}} {\textbf{tōi}} {\textbf{baptisthēnai}} {hapanta} {ton} {laon} {kai} {Iēsou} {baptisthentos} {kai} {proseukhomenou} {aneōikhthēnai} {ton} {ouranon}
    {happen.{\sc aor.3.sg}} {\sc ptc} {in} {the.{\sc dat.sg}} {wash ceremonially.{\sc inf.aor.pas}} {all.{\sc sg.m.acc}} {the.{\sc sg.m.acc}} {people.{\sc sg.m.acc}} {even} {Jesus.{\sc gen}} {wash ceremonially.{\sc ptcp.pas.pfv.m.gen.sg}} {and} {pray.{\sc ptcp.ipfv.m.gen.sg}} {open.{\sc inf.pas.aor}} {the.{\sc sg.m.acc}} {heaven.{\sc sg.m.acc}}
    \glt `It came to pass that, when all the people were baptized, as Jesus was baptized and as he was praying, the heaven was opened' (Luke 3:21)
    \glend
    \label{egdaoccasion}
    \end{itemize}
\end{example}

Incidentally, we can also observe that in (\ref{egdaoccasion}b) Greek does not have a finite \textit{when}-clause equivalent (e.g. \textit{hōte} or \textit{hōtan}), but a nominalized accusative with infinitive, as is often the case in \textit{egeneto}-constructions, which suggests that this usage of \textit{jegda}, in common with English \textit{when}, is found in Old Church Slavonic regardless of Greek.\\
\indent When an aorist \textit{jegda}-clause is followed by an imperfect matrix verb, the main eventuality may receive an inchoative reading (\ref{aorimp1}) or a reading where the event described by the \textit{jegda}-clause could be interpreted as temporally included in the main eventuality (\ref{aorimp2}).

\begin{example}
\begin{itemize}
    \item[a.]
    \gll i \textbf{egda} \textbf{v\foreignlanguage{russian}{ь}nide} v\foreignlanguage{russian}{ъ} dom\foreignlanguage{russian}{ъ} ot\foreignlanguage{russian}{ъ} naroda. v\foreignlanguage{russian}{ъ}prašaachǫ i učenici ego o prit\foreignlanguage{russian}{ъ}či. 
    {and} {when} {enter.{\sc aor.3.sg}} {in} {house.{\sc sg.m.acc}} {from} {crowd.{\sc sg.m.gen}} {ask.{\sc impf.3.pl}} {\sc 3.sg.m.acc} {disciple.{\sc pl.m.nom}} {\sc 3.sg.m.gen} {about} {parable.{\sc sg.f.acc}}
    \glt
    \glend
    \item[b.]
    \gll {Kai} {\textbf{hote}} {\textbf{eisēlthen}} {eis} {ton} {oikon} {apo} {tou} {okhlou} {epērōtōn} {auton} {hoi} {mathētai} {autou} {tēn} {parabolēn}
    {and} {when} {enter.{\sc aor.3.sg}} {in} {the.{\sc sg.m.acc}} {house.{\sc sg.m.acc}} {from} {the.{\sc sg.m.gen}} {crowd.{\sc sg.m.gen}} {ask.{\sc impf.3.pl}} {\sc 3.sg.m.acc} {the.{\sc pl.m.nom}} {disciple.{\sc pl.m.nom}} {\sc 3.sg.m.gen} {the.{\sc sg.f.acc}} {parable.{\sc sg.f.acc}}
    \glt 'When he entered a house away from the crowd, his disciples started asking him about the parable.' (Mark 7:17)%36715
    \glend
    \label{aorimp1}
\end{itemize}
\end{example}

\begin{example}
\begin{itemize}
    \item[a.]
    \gll Byst\foreignlanguage{russian}{ъ} že \textbf{egda} \textbf{približi} \textbf{sę} is\foreignlanguage{russian}{ъ} v\foreignlanguage{russian}{ъ} erichǫ. slěpec\foreignlanguage{russian}{ъ} eter\foreignlanguage{russian}{ъ} sěděaše pri pǫti prosę. 
    {happen.{\sc aor.3.sg}} {\sc ptc} {when} {approach.{\sc aor.3.sg}} {\sc refl} {Jesus.{\sc nom}} {in} {Jericho.{\sc acc}} {blind man.{\sc sg.m.nom}} {certain.{\sc sg.m.nom}} {sit.{\sc impf.3.sg}} {by} {road.{\sc sg.m.loc}} {beg.{\sc ptcp.ipfv.m.nom.sg}}
    \glt
    \glend
    \item[b.]
    \gll {Egeneto} {de} {en} {tōi} {eggizein} {auton} {eis} {Hiereikhō} {tuphlos} {tis} {ekathēto} {para} {tēn} {hodon} {epaitōn}
    {happen.{\sc aor.3.sg}} {\sc ptc} {in} {the.{\sc sg.m.dat}} {approach.{\sc inf.prs}} {\sc 3.sg.m.acc} {in} {Jericho} {blind man.{\sc sg.m.nom}} {certain.{\sc sg.m.nom}} {sit.{\sc impf.3.sg}} {by} {the.{\sc sg.f.acc}} {road.{\sc sg.f.acc}} {beg .{\sc ptcp.ipfv.m.nom.sg}}
    \glt `It came to pass that, when he approached Jericho, a certain blind man was sitting by the road begging.' (Luke 18:35)%41127
    \glend
    \label{aorimp2}
\end{itemize}
\end{example}

If we followed \posscitet{hinrichs86} theory, the sitting eventuality should include the approaching event in (\ref{aorimp2}). According to \citeauthor{hinrichs86}, the temporal relation between an eventive \textit{when}-clause and a stative main clause is the same as the one between a stative \textit{when}-clause and an eventive main clause, namely one of inclusion of the eventive situation in the stative one. If, instead, we followed \posscitet{partee84} account, the sitting eventuality should occur \textit{just after} the approaching event. It seems, however, that the temporal extension of the main clause in (\ref{aorimp2}) is irrelevant in this context, which merely appears to report that a \textit{sitting} state holds within the time referent introduced by the \textit{jegda}-clause, which introduces a bounded referent. This is also in line with \pgcitet{sandstrom}{192--195}, who argues that in such cases it is clear how a \textit{when}-clause `provides a temporal referent, which the head clause says something about' and that like a `temporal adverbial in general [it] serves to narrow down the section of the timeline in which the temporal location of the event is to be placed'--which is how framing adverbials and time locating adverbs are also generally described. As framing adverbials, sentence-initial \textit{when}-clauses provide the common temporal frame for both the \textit{when}-situation and the main clause situation, restricting the interpretation of the main eventuality to the specific domain introduced by the \textit{when}-clause itself (cf. \citealt{charolles05}). As observed by \citet{declerck} on English \textit{when}-clauses, when a durative-unbounded main clause follows a punctual \textit{when}-clause, the default is to represent the main clause, as it were, as a snapshot of a point in time---the one provided by the \textit{when}-clause and corresponding to the common temporal frame. This is what we get in (\ref{aorimp2}). If the default reading is pragmatically inaccessible, then we get what Declerck calls `sloppy W-simultaneity', namely the succession of one eventuality immediately after the other. With a durative-unbounded main clause, the effect of sloppy simultaneity is that the main eventuality is interpreted as either left- or right-bounded, namely as inchoative or terminative. The coincidence between the punctual framing adverbial and the durative-unbounded main clause is, in this case, with the starting or ending point of the main clause. \citet{sandstrom} also gives the inchoative reading as a possible interpretation of the temporal relation between an eventive \textit{when}-clause and a stative \textit{main}-clause, in which case, however, she also posits a consequentiality relation, whereby the stative main clause describes \textit{reactions} to the \textit{when}-clause event, which introduces a `change as a result of which the state predicate holds' (\pgcitealt{sandstrom}{194-195}). The consequentiality relation is certainly possible in (\ref{aorimp1}), where the disciples start asking Jesus questions as a result of now being in the house, where he was not before the \textit{jegda}-event.\\
\indent The inchoative effect is also found with a perfective participle followed by an imperfect matrix, as shown by the dative absolute in (\ref{absinchoa}) and the conjunct participle in (\ref{xadvinchoa}).

\begin{example}
\begin{itemize}
    \item[a.]
    \gll i \textbf{v\foreignlanguage{russian}{ъ}šed\foreignlanguage{russian}{ъ}šju} emu v\foreignlanguage{russian}{ъ} dom\foreignlanguage{russian}{ъ}. učenici ego v\foreignlanguage{russian}{ъ}prašachǫ i edinogo. ěko 
    {and} {enter.{\sc ptcp.pfv.m.dat.sg}} {\sc 3.sg.m.dat} {in} {house.{\sc sg.m.acc}} {disciple.{\sc pl.m.nom}} {\sc 3.sg.m.gen} {ask.{\sc impf.3.pl}} {\sc 3.sg.m.acc} {alone.{\sc sg.m.gen}} {that}
    \glt
    \glend
    \item[b.]
    \gll {kai} {\textbf{eiselthontos}} {autou} {eis} {oikon} {hoi} {mathētai} {autou} {kat'} {idian} {epērōtōn} {auton}
    {and} {enter.{\sc ptcp.pfv.m.gen.sg}} {\sc 3.sg.m.gen} {in} {house.{\sc sg.m.acc}} {the.{\sc pl.m.nom}} {disciple.{\sc pl.m.nom}} {\sc 3.sg.m.gen} {in} {private.{\sc sg.f.acc}} {ask.{\sc impf.3.pl}} {\sc 3.sg.m.acc}
    \glt `And when He had come into the house, His disciples started to ask Him privately' (Mark 9:28) %7012
    \glend
    \label{absinchoa}
\end{itemize}
\end{example}

\begin{example}
\begin{itemize}
    \item[a.]
    \gll \textbf{Prišed\foreignlanguage{russian}{ъ}} že is v\foreignlanguage{russian}{ъ} stranǫ kesariję filipovy. v\foreignlanguage{russian}{ъ}prašaše učeniky svoję glę
    {arrive.{\sc ptcp.pfv.m.nom.sg}} {\sc ptc} {Jesus.{\sc nom}} {in} {region.{\sc sg.f.acc}} {Caesarea.{\sc sg.f.gen}} {Philippi.{\sc sg.f.gen}} {ask.{\sc impf.3.sg}} {disciple.{\sc pl.m.acc}} {own.{\sc 3.pl.acc}} {say.{\sc ptcp.ipfv.m.nom.sg}}
    \glt
    \glend
    \item[b.]
    \gll {\textbf{Elthōn}} {de} {ho} {Iēsous} {eis} {ta} {merē} {Kaisareias} {tēs} {Filippou} {ērōta} {tous} {mathētas} {autou} {legōn}
    {come.{\sc ptcp.pfv.m.nom.sg}} {\sc ptc} {the.{\sc nom}} {Jesus.{\sc nom}} {in} {the.{\sc pl.n.acc}} {region.{\sc pl.n.acc}} {Caesarea.{\sc sg.f.gen}} {the.{\sc sg.f.gen}} {Philippi.{\sc sg.m.gen}} {ask.{\sc impf.3.sg}} {the.{\sc pl.m.acc}} {disciple.{\sc pl.m.acc}} {\sc 3.sg.m.gen} {say.{\sc ptcp.ipfv.m.nom.sg}}
    \glt `When Jesus arrived to the region of Caesarea Philippi, he began asking his disciples' (Matthew 16:13) %39004
    \glend
    \label{xadvinchoa}
\end{itemize}
\end{example}

Proper temporal inclusion or coincidence of a \textsc{frame} or an \textsc{independent rheme} perfective participle with one point in an imperfect matrix clause, as with the \textit{jegda}-clause in (\ref{aorimp2}), is instead an unlikely reading. Perfective \textsc{independent rhemes} and \textsc{frames} always induce narrative progression (\citealt{baryhaug2011}), so that while a reading where the eventuality described by an imperfective finite verb starts immediately after the perfective participle event (inchoative reading) is possible, the inclusion of the latter in the former should not be. \textsc{Elaborations}, which instead never induce narrative progression, will always temporally overlap with the matrix even in the uncommon scenario in which they are perfective and precede an imperfective matrix, as in (\ref{elabperf}), where an \textsc{elaboration} interpretation seems to be preferred considering the context.

\begin{example}
\begin{itemize}
    \item[a.]
    \gll \textbf{ouboěv\foreignlanguage{russian}{ъ}še} že \textbf{sę} i \textbf{pristraš\foreignlanguage{russian}{ъ}ni} \textbf{byv\foreignlanguage{russian}{ъ}še}. m\foreignlanguage{russian}{ь}něachǫ dch\foreignlanguage{russian}{ъ} vidęšte
     {be scared.{\sc ptpc.pfv.nom.pl}} {\sc ptc} {\sc refl} {and} {frightened.{\sc pl.m.nom}} {be.{\sc ptcp.pfv.m.nom.pl}} {think.{\sc impf.3.pl}} {ghost.{\sc sg.m.acc}} {see.{\sc ptcp.ipfv.nom.pl}}
    \glt
    \glend
    \item[b.]
    \gll {\textbf{ptoēthentes}} {de} {kai} {\textbf{emphoboi}} {\textbf{genomenoi}} {edokoun} {pneuma} {theōrein}
    {frighten.{\sc pl.pfv.pst.ptcp.pas.m.nom}} {\sc ptc} {and} {terrified.{\sc pl.m.nom}} {become.{\sc ptcp.pfv.nom.pl}} {think.{\sc impf.3.pl}} {ghost.{\sc sg.n.acc}} {see.{\sc inf.prs}}
    \glt `Startled and frightened, they thought they were seeing a ghost' (Luke 24:37)
    \glend
    \label{elabperf}
\end{itemize}
\end{example}

With an imperfect \textit{jegda}-clause, an aorist matrix is normally interpreted as temporally included in the situation described by the \textit{jegda}-clause, as in (\ref{impaor1})-(\ref{impaor3}), similarly to what typically happens with imperfective dative absolutes. 

\begin{example}
    \gll \textbf{jegda} že i \textbf{nesęchut\foreignlanguage{russian}{ь}} k grobu div\foreignlanguage{russian}{ь}no znamen\foreignlanguage{russian}{ь}e bys na nebesi i strašno
    {when} {\sc ptc} {\sc 3.m.acc.sg} {carry.{\sc impf.3.pl}} {to} {tomb.{\sc sg.m.dat}} {divine.{\sc n.nom.sg}} {sign.{\sc n.nom.sg}} {be.{\sc aor.3.sg}} {in} {sky.{\sc n.loc.sg}} {and} {terrible.{\sc n.nom.sg}}
    \glt `When they were carrying him to the tomb, a divine and terrible sign appeared in the sky' (\textit{Suzdal Chronicle}, Codex Laurentianus f. 103r)
    \glend
    \label{impaor1}
\end{example}

\begin{example}
    \gll i \textbf{jegda} \textbf{chotjachu} stran\foreignlanguage{russian}{ь}nii ot\foreignlanguage{russian}{ъ}iti v\foreignlanguage{russian}{ъ}zvěstiša unoši svoi otchod\foreignlanguage{russian}{ъ}
    {and} {when} {want.{\sc impf.3.pl}} {wanderer.{\sc m.nom.pl}} {leave.{\sc inf}} {inform.{\sc aor.3.pl}} {young man.{\sc m.dat.sg}} {their.{\sc m.acc.sg}} {departure.{\sc m.acc.sg}}
    \glt `When the wanderers were about to leave, they informed the young man of their departure` (\textit{Life of Feodosij Pečerskij}, Uspenskij Sbornik f. 28v) %277921
    \glend
    \label{impaor2}
\end{example}

\begin{example}
\begin{itemize}
    \item[a.]
    \gll i \textbf{egda} \textbf{sěaše} ovo pade pri pǫti i pop\foreignlanguage{russian}{ъ}rano byst\foreignlanguage{russian}{ъ}. i pticę nebskyję pozobašę e
    {and} {when} {sow.{\sc impf.3.sg}} {some.{\sc sg.n.nom}} {fall.{\sc aor.3.sg}} {along} {way.{\sc m.loc.sg}} {and} {trample.{\sc ptcp.pas.pfv.n.nom.sg}} {be.{\sc aor.3.sg}} {and} {bird.{\sc f.nom.pl}} {sky.{\sc f.nom.pl}} {devour.{\sc aor.3.pl}} {\sc 3.sg.n.acc}
    \glt
    \glend
    \item[b.]
    \gll {kai} {\textbf{en}} {\textbf{tōi}} {\textbf{speirein}} {auton} {ho} {men} {epesen} {para} {tēn} {hodon} {kai} {katepatēthē} {kai} {ta} {peteina} {tou} {ouranou} {katephagen} {auto}
    {and} {in} {the.{\sc n.dat.sg}} {sow seed.{\sc inf.prs}} {\sc 3.sg.m.acc} {\sc 3.sg.n.nom} {\sc ptc} {fall.{\sc aor.3.sg}} {along} {the.{\sc f.acc.sg}} {path.{\sc f.acc.sg}} {and} {trample.{\sc aor.pas.3.sg}} {and} {the.{\sc n.nom.pl}} {bird.{\sc n.nom.pl}} {the.{\sc m.gen.sg}} {air.{\sc m.gen.sg}} {devour.{\sc aor.3.sg}} {\sc 3.sg.n.acc}
    \glt 'And as he sowed, some fell by the wayside; and it was trampled down, and the birds of the air devoured it.' (Luke 8:5) %40385
    \glend
    \label{impaor3}
\end{itemize}
\end{example}

These three examples are to be analysed as a stative time-locating adverbial followed by an eventive main clause. Following \posscitet{hinrichs86} treatment, the reference time introduced by the \textit{jegda}-clause is a subinterval of the state it describes.\\
\indent When both the \textit{jegda}-clause and the matrix clause are in the imperfect, the two situations are generally interpreted as either temporally equivalent, as in (\ref{impimp1}), or as habitual, repeated occurrences, as in (\ref{impimp2}).

\begin{example}
\begin{itemize}
    \item[a.]
    \gll \textbf{egda} \textbf{iděaše} narodi ougnětaachǫ i
    {when} {go.{\sc impf.3.sg}} {crowd.{\sc m.nom.pl}} {throng.{\sc impf.3.pl}} {\sc 3.sg.m.acc}
    \glt
    \glend
    \item[b.]
    \gll {\textbf{en}} {de} {\textbf{tōi}} {\textbf{hupagein}} {auton} {hoi} {okhloi} {sunepnigon} {auton}
    {in} {\sc ptc} {the.{\sc m.dat.sg}} {go away.{\sc inf.prs}} {\sc 3.sg.m.acc} {the.{\sc m.nom.pl}} {crowd.{\sc m.nom.pl}} {choke.{\sc impf.3.pl}} {\sc 3.sg.m.acc}
    \glt `As he [Jesus] went, the crowds thronged him' (Luke 8:42) %51377
    \glend
    \label{impimp1}
\end{itemize}
\end{example}

\begin{example}
\begin{itemize}
    \item[a.]
    \gll i \textbf{egda} \textbf{viděachǫ} i dsi nečistii pripadaachǫ k\foreignlanguage{russian}{ъ} nemu. i v\foreignlanguage{russian}{ъ}piěchǫ glšte ěko
    {and} {when} {see.{\sc impf.3.pl}} {\sc 3.sg.m.acc} {spirit.{\sc m.nom.pl}} {unclean.{\sc m.nom.pl}} {fall down.{\sc impf.3.pl}} {to} {\sc 3.sg.m.dat} {and} {cry out.{\sc impf.3.pl}} {say.{\sc ptcp.ipfv.nom.pl}} {that}
    \glt
    \glend
    \item[b.]
    \gll {kai} {ta} {pneumata} {ta} {akatharta} {\textbf{hotan}} {auton} {\textbf{etheōroun}} {prosepipton} {autōi} {kai} {ekrazon} {legontes} {hoti}
    {and} {the.{\sc n.nom.pl}} {sprit.{\sc n.nom.pl}} {the.{\sc n.nom.pl}} {unclean.{\sc n.nom.pl}} {when} {\sc 3.sg.m.acc} {see.{\sc impf.3.pl}} {fall down before.{\sc impf.3.pl}} {\sc 3.sg.m.dat} {and} {cry out.{\sc impf.3.pl}} {say.{\sc ptcp.ipfv.nom.pl}} {that}   
    \glt `Whenever they saw him, the unclean spirits fell down before Him and cried out, saying' (Mark 3:11) %36479
    \glend
    \label{impimp2}
\end{itemize}
\end{example}

The range of temporal relations of nonpast-tense \textit{jegda}-clauses to their matrix clause (always nonpast-tense as well) is similar to the one we observe with past-tense clauses. Once again, what determines the interpretation can be boiled down to whether the \textit{jegda}-clause and the matrix clause present an eventuality as durative or non-durative, and as bounded or unbounded. As already discussed, establishing the aspect of nonpast forms is not always straightforward. It is often possible for perfective nonpast indicatives in Old Church Slavonic to receive a future interpretation (cf. \citealt{dostal}; \citealt{eckhoff2014a}; \citealt{eckhoff2015b}; \citealt{kamphuis2020a}), but that is not always clearly the case (\pgcitealt{eckhoff2014a}{253}). In (\ref{prespres2}), for example, the verb \textit{s\foreignlanguage{russian}{ъ}zrěti} `to ripen', arguably perfective, does not receive a future interpretation. 

\begin{example}
\begin{itemize}
    \item[a.]
    \gll \textbf{egda} že \textbf{soz\foreignlanguage{russian}{ъ}rěat\foreignlanguage{russian}{ъ}} plod\foreignlanguage{russian}{ъ}. abie pos\foreignlanguage{russian}{ъ}let\foreignlanguage{russian}{ъ} sr\foreignlanguage{russian}{ъ}p\foreignlanguage{russian}{ъ}. ěko nastoit\foreignlanguage{russian}{ъ} žętva
    {when} {\sc ptc} {ripen.{\sc prs.3.sg}} {grain.{\sc m.nom.sg}} {immediately} {send.{\sc prs.3.sg}} {sickle.{\sc m.acc.sg}} {because} {be present.{\sc prs.3.sg}} {harvest.{\sc f.nom.sg}}
    \glt
    \glend
    \item[b.]
    \gll {\textbf{hotan}} {de} {\textbf{paradoi}} {ho} {karpos} {euthus} {apostellei} {to} {drepanon} {hoti} {parestēken} {ho} {therismos}
    {when} {\sc ptc} {hand over.{\sc sbjv.pfv.3.sg}} {the.{\sc m.nom.sg}} {grain.{\sc sg.m.nom}} {immediately} {send away.{\sc prs.3.sg}} {the.{\sc n.acc.sg}} {sickle.{\sc n.acc.sg}} {that} {be present.{\sc prf.3.sg}} {the.{\sc m.nom.sg}} {harvest.{\sc m.nom.sg}} 
    \glt `But when the grain ripens, immediately he puts in the sickle, because the harvest has come' (Mark 4:29) %36544
    \glend
    \label{prespres2}
\end{itemize}
\end{example}

It does, however, induce narrative progression, similarly to what occurs between an eventive \textit{jegda}-clause in the aorist followed by an aorist main clause. The \textit{jegda}-clause in this example is habitual-repetitive, conveying a similar meaning to English \textit{whenever}. As \pgcitet{declerck}{232} argues, the basic temporal relation between the situation described by a \textit{whenever}-clause and a main clause is interpreted in the same way as the non-habitual-repetitive counterpart.\\
\indent Similarly to the temporal relation between an aorist \textit{jegda}-clause and an imperfect matrix, a stative matrix clause following a perfective nonpast \textit{jegda}-clause may receive an inchoative reading, as (\ref{prespres4}) exemplifies.

\begin{example}
\begin{itemize}
    \item[a.]
    \gll a ch\foreignlanguage{russian}{ъ} \textbf{egda} \textbf{pridet\foreignlanguage{russian}{ъ}} niktože ne věst\foreignlanguage{russian}{ъ} {ot\foreignlanguage{russian}{ъ} kǫdou} bǫdet\foreignlanguage{russian}{ъ}.
    {but} {Christ.{\sc nom}} {when} {come.{\sc prs.3.sg}} {nobody.{\sc nom}} {\sc neg} {know.{\sc prs.3.sg}} {whence} {be.{\sc fut.3.sg}}
    \glt
    \glend
    \item[b.]
    \gll {ho} {de} {Xristos} {\textbf{hotan}} {\textbf{erkhētai}} {oudeis} {ginōskei} {pothen} {estin}
    {the.{\sc m.nom.sg}} {\sc ptc} {Christ.{\sc m.nom.sg}} {when} {come.{\sc sbjv.prs.3.sg}} {nobody.{\sc nom}} {know.{\sc prs.3.sg}} {whence} {be.{\sc prs.3.sg}}
    \glt `When Christ comes, no one will know where he is from' (John 7:27) %42068
    \glend
    \label{prespres4}
\end{itemize}
\end{example}

The state of (not) \textit{knowing} starts \textit{just after} the \textit{when}-clause event. In other words, the temporal referent for the main clause, introduced by the \textit{jegda}-clause, is the (left-bounded) interval corresponding to the consequent state of Christ \textit{coming}.\\
\indent Finally, the temporal relation between a durative nonpast \textit{jegda}-clause followed by a stative main clause is one of overlap. In (\ref{prespres1}), the state described by the main verb \textit{sǫt\foreignlanguage{russian}{ъ}} holds at least as long as the \textit{guarding} eventuality holds.

\begin{example}
\begin{itemize}
    \item[a.]
    \gll \textbf{egda} krěp\foreignlanguage{russian}{ъ}ky ouorǫž\foreignlanguage{russian}{ь} sę \textbf{chranit\foreignlanguage{russian}{ъ}} svoi dvor\foreignlanguage{russian}{ъ}. v\foreignlanguage{russian}{ь} mirě sǫt\foreignlanguage{russian}{ъ} iměniě ego
    {when} {strong man.{\sc m.nom.sg}} {arm.{\sc ptcp.pfv.m.sg.nom}} {\sc refl} {guard.{\sc prs.3.sg}} {his.{\sc m.acc.sg}} {house.{\sc m.acc.sg}} {in} {peace.{\sc m.loc.sg}} {be.{\sc prs.3.pl}} {possession.{\sc n.nom.pl}} {\sc 3.sg.m.gen}
    \glt
    \glend
    \item[b.]
    \gll {\textbf{hótan}} {ho} {iskhuros} {kathōplismenos} {\textbf{phulassēi}} {tēn} {heautou} {aulēn} {en} {eirēnēi} {estin} {ta} {huparkhonta} {autou}
    {when} {the.{\sc m.nom.sg}} {strong man.{\sc m.nom.sg}} {arm.{\sc ptcp.pas.prf.m.nom.sg}} {guard.{\sc sbjv.prs.3.sg.act}} {the.{\sc f.acc.sg}} {\sc 3.sg.m.gen} {house.{\sc f.acc.sg}} {in} {peace.{\sc f.dat.sg}} {be.{\sc prs.3.sg}} {the.{\sc n.nom.pl}} {possession.{\sc n.nom.pl}} {\sc 3.sg.m.gen}
    \glt `When a strong man, fully armed, guards his own house, his possessions are safe' (Luke 11:21) %40675
    \glend
    \label{prespres1}
\end{itemize}
\end{example}

% DA durative + durative past
% ѡвѣм же б\foreignlanguage{russian}{ь}ющим с града и стрѣлѧющим межи собою идѧху стрѣлы акы дожд\foreignlanguage{russian}{ь}
% when the besieged fought from the city walls, and there was a shootout between them, the arrows flew like rain

% \begin{example}
%     \gll Egda bo iz mr\foreignlanguage{russian}{ъ}tvych\foreignlanguage{russian}{ъ} v\foreignlanguage{russian}{ъ}skr\foreignlanguage{russian}{ъ}snǫt\foreignlanguage{russian}{ъ}. ni ženęt\foreignlanguage{russian}{ъ} sę ni posagajǫt\foreignlanguage{russian}{ъ}. n\foreignlanguage{russian}{ъ} sǫt\foreignlanguage{russian}{ъ} ěko anǵi na nebesech\foreignlanguage{russian}{ъ}
%     {NOGLOSS} {NOGLOSS} {NOGLOSS} {NOGLOSS.{\sc pl.m.gen}} {NOGLOSS.{\sc 3.pl.prs.act}} {NOGLOSS} {NOGLOSS.{\sc 3.pl.prs.act}} {NOGLOSS.{\sc 3.sg.m.acc}} {NOGLOSS} {NOGLOSS.{\sc 3.pl.prs.act}} {NOGLOSS} {NOGLOSS.{\sc 3.pl.prs.act}} {as} {NOGLOSS.{\sc pl.m.nom}} {NOGLOSS} {NOGLOSS.{\sc pl.n.loc}}
%     \glt `For when the dead rise, they will neither marry nor be given in marriage; they will be like the angels in heaven' (Mark 12.25) %37070
%     \glend
%     \label{prespres3}
% \end{example}
%
% [difference 1 between participles and egda clauses: perfectives followed by imperfectives]
%

\subsection{Summary}
This section presented an overview of the main temporal relations holding between \textit{jegda}-clauses and their matrix clause, on the basis of the different attested combinations of tense-aspect in each of them. We have observed that the range of readings inferred for English \textit{when}-clauses, as gathered from some of the major studies on the subject (e.g. \citealt{partee84,hinrichs86,sandstrom,declerck}), can generally be observed to hold in Early Slavic \textit{jegda}-clauses as well. In particular, we have observed that accounts of temporal relations taking into account not only boundedness \textit{and} durativity of eventualities, but the larger discourse context and pragmatic incompatibility (e.g. \citealt{moens,sandstrom,declerck}), seem more suitable to explain a wider range of readings beyond the default interpretations based on tense-aspect only (e.g. \citealt{partee84, hinrichs86}). This is the case, for instance, of what \citet{declerck} calls \textit{occasion-referring} \textit{when}-clauses and, similarly, what \citet{sandstrom} refers to as \textit{subevent} relation (the `constitutive part' relation of \citealt{moens}). This type of temporal relation can be observed in Early Slavic between an aorist \textit{jegda}-clause followed by an \textit{aorist} matrix clause, which can \textit{also} (perhaps typically, but certainly not exclusively) invite a reading whereby the matrix event immediately follows the \textit{jegda}-clause event (i.e. \posscitet{partee84} and \posscitet{dowty86} `time just after').\\
\indent We have also seen how imperfect \textit{jegda}-clauses \textit{do} occur relatively often with an aorist matrix clause, unlike Greek imperfect \textit{hóte}-/\textit{hótan}-clauses, for which an aorist matrix is exceedingly uncommon. Such configuration in Early Slavic normally involves a reading whereby the matrix event is included in the \textit{jegda}-clause (stative) eventuality, similarly to the reading obtained from an imperfective participle followed by an aorist matrix clause. 

\section{Rhetorical relations and the discourse structure of \textit{jegda}- and participle clauses}\label{egdarethorical}
\subsection{Competing motivations, competing relations}
\posscitet{behrens2012a} analysis of competing structure, while mentioning finite subordinate clauses as one of the possible competitors to non-finite adjuncts, focuses on the semantic-pragmatic parameters affecting the choice between non-finite adjunction and syntactic coordination.\footnote{The authors, unfortunately, largely dismiss the competition between finite subordinates and non-finite adjuncts to focus on the competition of the latter with coordinated and juxtaposed structure, since, according to them, finite subordinate `differ from non-finite adjuncts mainly by expressing a more precise semantic relation to the host' and are therefore `less interesting competitors than are the non-subordinate alternatives, namely coordination and juxtaposition’ (\pgcitealt{behrens2012a}{201}).} Their main observation is that fronted adjuncts typically trigger an Accompanying Circumstance\footnote{`Accompanying circumstance (AC), as a clause-linking relation, is informally defined as relating two events or states to each other, which, "from the point of view of the speaker/writer, form a unit mentally" (\pgcitealt{kortmann1991a}{122})' (\citealt{accompcircum}). Note that, as the authors also point out, \textit{Accompanying circumstance} does not correspond to any SDRT relation in particular, while being compatible with several of them (e.g. Background, Elaboration, Result).} or SDRT's \textit{Background} relation to the matrix clause, often with a (weakly) causal meaning (\pgcitealt{behrens2012a}{207}) and that their prototypical uses are generally incompatible with SDRT coordinating discourse relations such as \textit{Narration}, which imply temporal succession of events, whereas syntactic coordination is instead taken to signal discourse coordination.\footnote{From an SDRT standpoint, the equivalence between syntactic coordination and discourse coordination is not as straightforward: see for example \pgcitet{ashervieusubcoorddisc}{598-599} on this issue.} They also make a difference in `discourse potential' (i.e. presumably the set of compatible discourse relations) between VP coordination, where the conjuncts share the subject, and clause coordination, where they do not, as in (\ref{vpclausalcoord}a), from \pgcitet{blakemorecarston05}{570} and (\ref{vpclausalcoord}b), from \pgcitet{levinson2000}{121}, respectively. They suggest that VP coordination, given the closer tie between the first and the second conjunct, is more likely to establish a causal background for the next event in the sequence, encoded by the second conjunct, which makes VP coordination closer to adjunction than clausal coordination. 

\begin{example}
\begin{itemize}
    \item[a.] She jumped on the horse \textit{and} rode into the sunset.
    \item[b.] John turned the key \textit{and} the engine started.
\end{itemize}
\label{vpclausalcoord}
\end{example}

They also add that some languages may have VP coordination as a felicitous alternative to clause-final non-finite adjuncts that describe an unbounded state or activity and that introduce an \textit{Accompanying Circumstance} to the matrix, in which case they `contribute to the main storyline on a par with their host' (\pgcitealt{behrens2012a}{216}). However, we know that \textsc{independent rhemes} are discourse-structurally very similar to independent clauses and, in many cases, a \textit{Background} or \textit{Accompanying Circumstance} reading is unambiguously not available, as in (\ref{unambindrheme}) (repeated from Chapter 1) and (\ref{unambindrheme2}).

\begin{example}
\begin{itemize}
\item[a.]
\gll i abie \textbf{tek\foreignlanguage{russian}{ъ}} edin\foreignlanguage{russian}{ъ} ot\foreignlanguage{russian}{ъ} nich\foreignlanguage{russian}{ъ}. i \textbf{priem\foreignlanguage{russian}{ъ}} gǫbǫ. \textbf{ispl\foreignlanguage{russian}{ь}n\foreignlanguage{russian}{ь}} oc\foreignlanguage{russian}{ь}ta. i \textbf{v\foreignlanguage{russian}{ь}znez\foreignlanguage{russian}{ъ}} na tr\foreignlanguage{russian}{ь}st\foreignlanguage{russian}{ь}. napaěše i
and immediately run.\textsc{ptcp.pfv.m.nom.sg} one.\textsc{m.nom.sg} from he.\textsc{gen.pl} and take.\textsc{ptcp.pfv.m.nom.sg} sponge.\textsc{acc.sg} fill.\textsc{ptcp.pfv.m.nom.sg} vinegar.\textsc{gen.sg} and put.\textsc{ptcp.pfv.m.nom.sg} on reed.\textsc{acc.sg} give.to.drink.\textsc{impf.3.sg} he.\textsc{acc.sg}
\glt
\glend
\item[b.]
\gll kai eutheōs \textbf{dramōn} heis ex autōn kai \textbf{labōn} spongon \textbf{plēsas} te oxous kai \textbf{peritheis} kalamō epotizen auton
and immediately run.\textsc{ptcp.pfv.m.nom.sg} one.\textsc{nom.sg} from he.\textsc{gen.pl} and take.\textsc{ptcp.pfv.m.nom.sg} sponge.\textsc{acc.sg} fill.\textsc{ptcp.pfv.m.nom.sg} with vinegar.\textsc{gen.sg} and put.\textsc{ptcp.pfv.m.nom.sg} reed.\textsc{dat.sg} give.to.drink.\textsc{impf.3.sg} he.\textsc{acc.sg}
\glt ‘Immediately one of them ran and took a sponge, filled it with sour wine and put it on a reed, and offered it to him to drink’ (Matthew 27:48)
\glend
\label{unambindrheme}
\end{itemize}
\end{example}

\begin{example}
    \gll i viděv\foreignlanguage{russian}{ъ} ja bž\foreignlanguage{russian}{ь}stv\foreignlanguage{russian}{ь}nyi unoša i rad\foreignlanguage{russian}{ъ} byv\foreignlanguage{russian}{ъ} \textbf{tek\foreignlanguage{russian}{ъ}} pokloni sja im\foreignlanguage{russian}{ъ}
    {and} {see.{\sc ptcp.pfv.m.sg.nom}} {\sc 3.pl.m.acc} {divine.{\sc m.nom.sg}} {joung man.{\sc m.nom.sg}} {and} {happy.{\sc m.nom.sg}} {be.{\sc ptcp.pfv.m.sg.nom}} {run.{\sc ptcp.pfv.m.sg.nom}} {bow.{\sc aor.3.sg}} {\sc refl} {\sc 3.pl.m.dat}
    \glt `And when he saw them, the divine young man rejoiced, ran up to them and bowed' (\textit{Life of Feodosij Pečerskij}, Uspenskij Sbornik f. 28v)
    \glend
    \label{unambindrheme2}
\end{example}

Indeed, the sentence in (\ref{vpclausalcoord}a) intuitively corresponds to the typical contexts in which \textsc{independent rhemes} occur, whereas (\ref{vpclausalcoord}b), involving subject switch-reference, may be realistically expected as a sequence of absolute construction, or \textit{jegda}-clause, and main clause, even if this may trigger somewhat different discourse interpretations than a sequence of main clauses. Compare (\ref{vpclausalcoord}) with (\ref{whenclausalcoord}).

\begin{example}
\begin{itemize}
    \item[a.] \textit{Jumping on the horse}, she rode into the sunset.
    \item[b.] \textit{When John turned the key}, the engine started.
\end{itemize}
    \label{whenclausalcoord}
\end{example}

Even from the English non-finite adjunct in (\ref{whenclausalcoord}a), it is clear that the relation between the adjunct and the main clause is not \textit{Background}, since it involves two events, whereas to license \textit{Background}, in SDRT terms, one of the two eventualities should be a state (cf. \citealt{asherlasca2003,asher2007a}): \textit{Event}($e_{\alpha}$) $\land$ \textit{State}($e_{\beta}$) $\land$ ?(\textit{$\alpha$,$\beta$}) $>$ Background($\alpha$,$\beta$), as well as \textit{State}($e_{\beta}$) $\land$ \textit{Event}($e_{\alpha}$) $\land$ ?(\textit{$\alpha$,$\beta$}) $>$ Background($\alpha$,$\beta$).\footnote{The question mark in the SDRT notation simply indicates that two discourse constituent are linked by a semantically underspecified relation.} But it is not \textit{Accompanying Circumstance} either, since the two eventualities are understood as occurring in sequence, with the \textit{jumping} event almost enabling the \textit{riding} one, which goes against the co-temporality requirement of Accompanying Circumstances (\citealt{accompcircum}). While both (\ref{vpclausalcoord}b) and (\ref{whenclausalcoord}b) invite a causal/enablement reading, in (\ref{whenclausalcoord}b) the consequentiality relation seems stronger. A possible explanation for this is that \textit{when}, as highlighted by \citet{behrens2012a} among the potential differences between coordination and adjunction/subordination, can have a connective function by referring back to the preceding discourse and link a known or inferrable eventuality to the matrix clause, whereas, according to them, a first conjunct in a coordinated construction cannot because `it has to introduce a new event referent' (\pgcitealt{behrens2012a}{221}). Moreover, besides their `connective function', which is in common with non-finite adjuncts, \pgcitet{behrens2012a}{219} also argue that finite adverbial clauses `usually highlight one particular (temporal, causal, concessive) aspect of the full co-eventuality relation at the cost of others, thus narrowing down the interpretation as compared to a non-finite adjunct'. In (\ref{whenclausalcoord}b), the intuition is that the use of a \textit{when}-clause would be maximally coherent if, for instance, the engine started \textit{only} when John turned the key (e.g. after trying other ways) or when John turned the key \textit{this time} (e.g. as opposed to other times), perhaps after fixing something meanwhile, and so on. \posscitet{behrens2012a} observations thus seem useful for explaining some of the differences between coordinated independent clauses and a non-finite adjunct followed by the main clause when the adjunct functions as a \textsc{frame}, as well as between participle adjuncts and a \textit{jegda}-clause. However, they are still not helpful for accounting for the choice between coordinated independent clauses and an \textsc{independent rheme} participle followed by the matrix. In particular, unlike what \citet{behrens2012a} observe regarding non-finite adjuncts, \textsc{independent rhemes} introduce new events in the discourse, thus inducing narrative progression (cf. \pgcitealt{baryhaug2011}{13-16}), and they are not only \textit{compatible} with \textit{Narration}, but they \textit{typically} license it. In fact, as argued by \citet{baryhaug2011}, they \textit{always} induce narrative progression, which is instead only the default between two eventive main clauses and is among the rationales for referring to Ancient Greek participles as \textit{grammaticalized} versions of SDRT's rhetorical relations (\pgcitealt{baryhaug2011}{5}).

\subsection{INDEPENDENT RHEMES as triggers for Strong Narration}
As argued by \pgcitet{asherlasca2003}{20}, the coherence of a discourse is scalar, in the sense that its quality can vary from minimally to maximally coherent (or `relevant', broadly in the now classical sense of \citealt{sperberwilson86}) depending on several factors, including how many rhetorical relation hold between two discourse segments (the more relations, the more coherent the interpretation), the number of anaphoric expressions with a resolved antecedent (the more expressions, the greater the quality of the coherence), and the specific rhetorical relation(s) involved in the interpretation. Similarly, \citet{altshulertrusswell22} argue that the acceptability of a discourse with a given assigned structure is a gradient matter, which can be boiled down to two major components, namely the `gradient quality' of some discourse relations, such as \textit{Parallel} or \textit{Contrast} (cf. \pgcitealt{altshulertrusswell22}{245-248}), and the `more slippery notion' of how strongly an inferred discourse structure is supported by world knowledge (ibi: 215-216). Already \citet{asherlasca2003} had argued that some rhetorical relations are inherently scalar. The `quality' of a \textit{Narration} relation, for instance, depends on the specificity of the common topic that summarizes the content of the `story' within which the units linked by \textit{Narration} occur.\\
\indent \citet{brasetal2001}, for instance, analysed the different ways certain temporal connectives in French, such as \textit{puis} `then' and \textit{un peu plus tard} `a bit later', affect the temporal structure of a discourse and license somewhat different versions of the same rhetorical relation \textit{Narration}. 
\textit{Puis} triggers \textit{Strong Narration}, which has a stronger requirement on the presence of a common topic summarizing the eventuality linked by the relation and signals that no relevant intermediate event $e_{\gamma}$ can occur between the two discourse units $e_{\alpha}$ and $e_{\beta}$ connected by \textit{Narration} without the post-state of $e_{\alpha}$ being incompatible with the pre-state of $e_{\beta}$. \textit{Un peu plus tard} instead licenses \textit{Weak} \textit{Narration}, which merely signals temporal succession and can link two eventualities with a much weaker common topic. The temporal sequentiality (or \textit{abutment}, \pgcitealt{kamp1993a}{573}) entailed by \textit{Strong Narration} is therefore not present with the \textit{Weak Narration} relation.\footnote{See also \posscitet{kehler2002} relation \textit{Occasion}, which is largely equivalent to \posscitet{brasetal2001} \textit{Strong Narration}, and its discussion in \citet{altshulertrusswell22}.} Moreover, there are different non-temporal effects typically allowed by \textit{Strong} and \textit{Weak Narration}. Both signal temporal succession of two events, but the former does not seem to allow \textit{Result} to additionally hold between the two, while the latter does; conversely, the former allows some form of \textit{Enablement} relation, whereas the latter does not.\footnote{Note that \textit{Enablement}, as \citet{brasetal2001} already pointed out, has not been formalized within SDRT, but it is informally borrowed from \citet{sandstrom}, who posited it as a non-temporal effect holding between a \textit{when}-clause and its matrix in several contexts.}\\
\indent Parallels can be made between the discourse properties of \textit{puis} observed by \citet{brasetal2001} and the temporal and non-temporal semantic effects of \textsc{independent rhemes}. First, the typical usage of \textsc{independent rhemes} is in contexts where one event immediately follows another, often with the non-temporal effect that the first loosely enables the second. This can occur in relatively long chains of conjunct participles leading up to a finite main verb, inviting a reading in which the events follow one another without interruptions. Second, sequences of conjunct participles are necessarily interpreted as part of the same storyline as the event described by the matrix clauses which they lead up to. \textsc{Independent rhemes} may thus be considered as explicitly signalling that the relationship between the events described is one of \textit{Strong Narration}, namely that the matrix clause event and any event encoded by an \textsc{independent rheme} syntactically dependent on it are to be interpreted strictly as `telling the “same story” (topic)' (\pgcitealt{brasetal2001}{8}) and as temporally \textit{abutted}. The `same story' requirement (i.e. the presence of a clearly identifiable FT), in Early Slavic (as in Greek) is often fulfilled by an explicit sentence-initial \textsc{frame}---i.e. another conjunct participle, dative absolute, or finite temporal subordinate, if not several of these together. (\ref{frameplusind1})-(\ref{frameplusind3}) are examples of this discourse structure.

\begin{example}
    \gll i prozvuteru tomu \textbf{suštju}$_{[\textsc{frame}]}$. i č\foreignlanguage{russian}{ь}rnoriz\foreignlanguage{russian}{ь}cju iskus\foreignlanguage{russian}{ь}nu. iže i \textbf{poim\foreignlanguage{russian}{ъ}}$_{[\textsc{indrheme}]}$ blaženago ḟeodosia. i po obyčaju svętyich\foreignlanguage{russian}{ъ} oc\foreignlanguage{russian}{ь} \textbf{ostrigy}$_{[\textsc{indrheme}]}$ i obleče i v\foreignlanguage{russian}{ъ} m\foreignlanguage{russian}{ь}niš\foreignlanguage{russian}{ь}skuju odežju
    {and} {priest.{\sc sg.m.dat}} {that.{\sc sg.m.dat}} {be.{\sc ptcp.ipfv.m.sg.dat}} {and} {monk.{\sc sg.m.dat}} {skillful.{\sc sg.m.dat}} {which.{\sc sg.m.nom}} {and} {take.{\sc ptcp.pfv.m.sg.nom}} {blessed.{\sc sg.m.gen}} {Feodosij.{\sc sg.m.gen}} {and} {according to} {custom.{\sc sg.m.dat}} {holy.{\sc pl.m.gen}} {father.{\sc pl.m.gen}} {tonsure.{\sc ptcp.pfv.m.sg.nom}} {\sc 3.sg.m.acc} {clothe.{\sc 3.sg.aor.act}} {\sc 3.sg.m.acc} {in} {monastic.{\sc sg.f.acc}} {clothes.{\sc sg.f.acc}}
    \glt `As he was a priest himself and a skillful monk, he tonsured the Blessed Theodosius according to the custom of the Holy Fathers and put him in monastic clothes' (\textit{Life of Feodosij Pečerskij}, Uspenskij Sbornik f. 31v) %278054
    \glend
    \label{frameplusind1}
\end{example}

\begin{example}
    \gll andrěi že \textbf{poklonšju}$_{[\textsc{frame}]}$ \textbf{sę} ocju i \textbf{šed\foreignlanguage{russian}{ъ}}$_{[\textsc{indrheme}]}$ sěde v peresopnici
    {Andrej.{\sc nom}} {\sc ptc} {bow.{\sc ptcp.pfv.m.sg.dat}} {\sc refl} {father.{\sc m.dat.sg}} {and} {go.{\sc ptcp.pfv.m.sg.nom}} {sit.{\sc aor.3.sg}} {in} {Peresopnitsa.{\sc loc}}
    \glt `Andrej, after bowing to his father, went and sat in Peresopnitsa' (\textit{Suzdal Chronicle}, Codex Laurentianus f. 109v) %274361
    \glend
    \label{frameplusind2}
\end{example}

\begin{example}
    \gll i žena edina \textbf{sǫšti}$_{[\textsc{frame}]}$ v\foreignlanguage{russian}{ъ} točenii kr\foreignlanguage{russian}{ъ}ve lět\foreignlanguage{russian}{ъ} {d\foreignlanguage{russian}{ь}vě na desęte}. i mnogo \textbf{postradav\foreignlanguage{russian}{ъ}ši}$_{[\textsc{frame}]}$ ot\foreignlanguage{russian}{ъ} m\foreignlanguage{russian}{ъ}nog\foreignlanguage{russian}{ъ} balii. i \textbf{iždiv\foreignlanguage{russian}{ъ}ši}$_{[\textsc{frame}]}$ v\foreignlanguage{russian}{ь}se svoe. i ni edinoję pol\foreignlanguage{russian}{ь}dzę \textbf{obrět\foreignlanguage{russian}{ъ}ši}$_{[\textsc{frame}]}$. n\foreignlanguage{russian}{ъ} pače v\foreignlanguage{russian}{ъ} gore \textbf{priš\foreignlanguage{russian}{ь}d\foreignlanguage{russian}{ъ}ši}$_{[\textsc{frame}]}$. \textbf{slyšav\foreignlanguage{russian}{ъ}ši}$_{[\textsc{frame}]}$ o isě. \textbf{prišed\foreignlanguage{russian}{ъ}ši}$_{[\textsc{indrheme}]}$ v\foreignlanguage{russian}{ъ} narodě. s\foreignlanguage{russian}{ъ} zadi prikosno sę rizě ego.
    {and} {woman.{\sc sg.nom}} {some.{\sc sg.nom}} {be.{\sc ptcp.ipfv.f.sg.nom}} {in} {flow.{\sc loc}} {blood.{\sc gen}} {year.{\sc gen.pl}} {twelve} {and} {much} {suffer.{\sc ptcp.pfv.f.sg.nom}} {from} {much.{\sc gen.pl}} {doctor.{\sc gen.pl}} {and} {spend.{\sc ptcp.pfv.f.sg.nom}} {all.{\sc n.acc.sg}} {her.{\sc n.acc.sg}} {and} {\sc neg} {one.{\sc f.gen.sg}} {benefit.{\sc f.gen.sg}} {receive.{\sc ptcp.pfv.f.sg.nom}} {but} {even} {in} {worse.{\sc n.acc.sg}} {come.{\sc ptcp.pfv.f.sg.nom}} {listen.{\sc ptcp.pfv.f.sg.nom}} {about} {Jesus.{\sc loc}} {come.{\sc ptcp.pfv.f.sg.nom}} {in} {crowd.{\sc m.loc.sg}} {from} {behind} {touch.{\sc aor.3.sg}} {\sc refl} {garment.{\sc f.loc.sg}} {\sc 3.sg.m.gen}
    \glt `Now a certain woman had a flow of blood for twelve years, and had suffered many things from many physicians. She had spent all that she had and was no better, but rather grew worse. When she heard about Jesus, she came behind Him in the crowd and touched His garment' (Mark 5:25-27)
    \glend
    \label{frameplusind3}
\end{example}

In (\ref{frameplusind1}), the independent rhemes \textit{poim\foreignlanguage{russian}{ъ}} `took' and \textit{ostrigy} `tonsured' are preceded by the dative absolute \textit{suštju}, functioning as a \textsc{frame}. In (\ref{frameplusind2}), the \textsc{independent rheme} \textit{šed\foreignlanguage{russian}{ъ}} `went' is also preceded by a dative absolute \textit{polkonšju} `after bowing', which is in fact co-referential with the \textsc{independent rheme} (and the matrix verb), clearly signalling the different discourse status of the first participle compared to the following one. In (\ref{frameplusind3}), we see a much more complex \textsc{frame}, made of several conjunct participles, leading up to what seems to be the only conjunct participle in the sentence potentially functioning as an \textsc{independent rheme}, namely \textit{prišed\foreignlanguage{russian}{ъ}ši} `came'.

\subsection{Formalizing FRAMES as Background-triggers}
\begin{figure}[!h]
    \centering
    \includegraphics[scale=0.5]{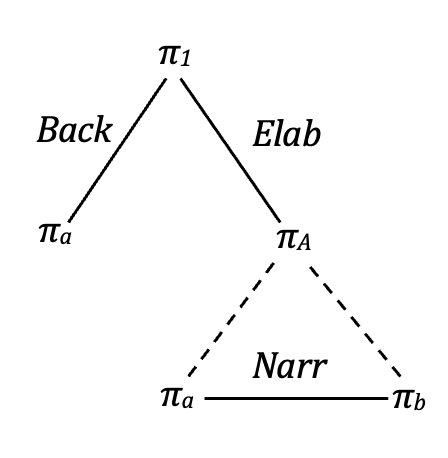}
    \caption{Basic discourse structure of a discourse unit involving a Background$_{forward}$ relation in SDRT according to \pgcitet{asher2007a}{14}}
    \label{fbp}
\end{figure}

All the examples so far involved participles clauses in pre-matrix position, which, as we have repeatedly seen, is where participles are typically either \textsc{frames} or \textsc{independent rhemes}. In SDRT terms, we have seen how, most typically, \textsc{independent rhemes} are likely to trigger a \textit{Narration} relation. When preceded by framing constructions, the latter are most likely to license \textit{Background}. The formalization in \citet{asher2007a} of the SDRT relation \textit{Background} in its \textit{Back\-ground}$_{Forward}$ version, namely when the backgrounded eventuality occurs linearly before the foregrounded eventuality, in fact, seems to capture the \textit{typical} discourse relations between pre-matrix participle clauses and their matrix particularly well. The schematized structure in Figure (\ref{fbp}), from \pgcitet{asher2007a}{14}, could in principle be applied to all the examples seen so far.

As Figure \ref{fbp} shows, \textit{Background}$_{Forward}$ does not directly attach to the foregrounded constituent, but to a constructed framing topic (FT), $\pi_{1}$ in Figure \ref{fbp}, to which the foreground is then attached via \textit{Elaboration}. The foregrounded constituent can then be complex and contain, for example, several discourse units attached to each other via \textit{Narration}.\\
\indent The FT is constructed by `percolating' up a discourse referent from the backgrounded constituent ($\pi_{a}$) to make it available for the foreground ($\pi_{A}$). Different factors are considered to affect the potential of a referent to percolate up to the topic, including information structure, the position of the referent in the givenness scale, the syntactic and thematic role of a referent, and, more generally, the salience of the referents (\pgcitealt{asher2007a}{17}). What is relevant to our discussion here is that the FT required by the \textsc{independent rhemes} attached to the matrix by (\textit{Strong}) \textit{Narration} is the same as the one which is constructed when a background constituent is available in the discourse structure. When a \textsc{frame} is available, then the framing participles (or \textit{jegda}-clause) constitute the backgrounded constituent, whose referents percolate up to construct an FT within which the foregrounded constituent(s) (i.e. typically \textsc{independent rhemes}, if present, plus the main verb) are located. Basic structures like the one in Figure \ref{fbp} reflect the very common case in which one or several participles can be stacked up before the matrix verb and that only the first is, potentially, a \textsc{frame}, as \citet{baryhaug2011} had already observed. This also fits the observation that very often the first clauses in such series are dative absolutes or \textit{jegda}-clauses, followed by one or more conjunct participles. Note that the backgrounded constituent can certainly be a complex one, as \pgcitet{declerck}{114, 130-131, i.a.} observed about multiple temporal adverbials, which are normally understood as including one another and all eventually contribute to the specification of a single time. See, for example, (\ref{complexback1})-(\ref{complexback3}).

\begin{example}
\gll {\normalfont \{}i \textbf{egda} \textbf{byst\foreignlanguage{russian}{ъ}}$_{[\textsc{frame}]}$ {d\foreignlanguage{russian}{ъ}voju na desęte} lětou \textbf{v\foreignlanguage{russian}{ъ}šed\foreignlanguage{russian}{ъ}šem\foreignlanguage{russian}{ъ}}$_{[\textsc{frame}]}$ im\foreignlanguage{russian}{ъ} v\foreignlanguage{russian}{ъ} im\foreignlanguage{russian}{ъ} po obyčajǫ prazd\foreignlanguage{russian}{ь}nika i \textbf{kon\foreignlanguage{russian}{ь}čav\foreignlanguage{russian}{ъ}šem\foreignlanguage{russian}{ъ}}$_{[\textsc{frame}]}$ d\foreignlanguage{russian}{ъ}ni \textbf{v\foreignlanguage{russian}{ъ}zvraštajǫštem\foreignlanguage{russian}{ъ}}$_{[\textsc{frame}]}$ \textbf{sę} im\foreignlanguage{russian}{ъ}{\normalfont \}} osta otrok\foreignlanguage{russian}{ъ} is v\foreignlanguage{russian}{ъ} imě i ne čju iosif\foreignlanguage{russian}{ъ} i mati ego
{and} {when} {be.{\sc aor.3.sg}} {twelve} {year.{\sc gen}} {go in.{\sc ptcp.pfv.dat.pl}} {\sc 3.pl.m.dat} {in} {Jerusalem.{\sc acc}} {according to} {custom.{\sc dat.sg}} {festival.{\sc gen.sg}} {and} {finish.{\sc ptcp.pfv.m.dat.sg}} {day.{\sc m.acc.pl}} {return.{\sc ptcp.ipfv.dat.pl}} {\sc refl} {\sc 3.pl.m.dat} {leave.{\sc aor.3.sg}} {boy.{\sc m.nom.sg}} {Jesus.{\sc nom}} {in} {Jerusalem.{\sc loc}} {and} {\sc neg} {realize.{\sc aor.3.sg}} {Joseph.{\sc nom}} {and} {mother.{\sc nom.sg}} {\sc 3.sg.m.gen}
\glt `When he was twelve years old, they went up to the festival according to the custom and, after the festival was over, while they were returning home, the boy Jesus stayed behind in Jerusalem and Mary and Joseph did not realize' (Luke 2:42–43) %40046
\glend
\label{complexback1}
\end{example}

\begin{example}
\gll {\normalfont \{}i \textbf{egda} \textbf{vorotiša}$_{[\textsc{frame}]}$ sę opęt\foreignlanguage{russian}{ь} polovci knęz\foreignlanguage{russian}{ь} že jaropolk\foreignlanguage{russian}{ъ} \textbf{ukrěpiv\foreignlanguage{russian}{ъ}}$_{[\textsc{frame}]}$ sę b\foreignlanguage{russian}{ь}eju pomošt\foreignlanguage{russian}{ь}ju ne \textbf{žda}$_{[\textsc{frame}]}$ inoe pomošti ni brata ni drugago{\normalfont \}} tokmo s perejaslavci svoimi ispostiže ja u polk\foreignlanguage{russian}{ъ}stěnę
{and} {when} {return.{\sc aor.3.pl}} {\sc refl} {again} {Polovtsians.{\sc nom.pl}} {king.{\sc nom.sg}} {\sc ptc} {Yaropolk.{\sc nom}} {strengthen.{\sc ptcp.pfv.m.sg.nom}} {\sc refl} {God's.{\sc f.inst.sg}} {help.{\sc f.inst.sg}} {\sc neg} {expect.{\sc ptcp.ipfv.m.sg.nom}} {other.{\sc f.gen.sg}} {help.{\sc f.gen.sg}} {neither} {brother.{\sc m.gen.sg}} {nor} {other.{\sc m.gen.sg}} {only} {with} {Pereyaslavs.{\sc inst}} {his.{\sc inst.pl}} {overtake.{\sc aor.3.sg}} {\sc 3.pl.m.acc} {at} {Polkosten.{\sc gen}}
\glt `And when the Polovtsians returned again, Prince Yaropolk strengthened himself with God's help and, expecting no other help, neither from a brother nor from others, only with his Pereyaslavs he overtook them at Polkosten' (\textit{Suzdal Chronicle}, Codex Laurentianus f. 98r) %40046
\glend
\label{complexback2}
\end{example}

\begin{example}
\gll {\normalfont \{}\textbf{egda} že \textbf{priide}$_{[\textsc{frame}]}$ nošt\foreignlanguage{russian}{ь} drugaja po pravilě \textbf{vozlegšu}$_{[\textsc{frame}]}$ mi i ne \textbf{spja}$_{[\textsc{frame}]}${\normalfont \}} molitvy govorju
{when} {\sc ptc} {come.{\sc aor.3.sg}} {night.{\sc f.nom.sg}} {other.{\sc f.nom.sg}} {after} {Prayer Rules.{\sc loc}} {lay down.{\sc ptcp.pfv.m.sg.dat}} {\sc 1.sg.m.dat} {and} {\sc neg} {sleep.{\sc ptcp.ipfv.m.sg.nom}} {prayer.{\sc f.acc.pl}} {say.{\sc prs.1.sg}}
\glt `When another night came, as I lay down after my Prayer Rules and I could not sleep, I pray' (\textit{The Life of Avvakum}, Pustozerskij Sbornik f. 99r)
\glend
\label{complexback3}
\end{example}

All these examples have a complex frame, made of multiple adverbial clauses incrementally building a common FT for the foregrounded eventuality. If we break down these examples and construct a discourse structure (SDRS) within an SDRT framework, we can see how, in fact, the construction of a common FT for the main eventuality occurs with every incoming discourse unit. Each discourse unit adding to the top-level FT (i.e. the one serving as FT for the main eventuality) can, in turn, introduce different discourse relations to the other units that contribute to the main FT. Let us look more closely at one of the examples, (\ref{complexback1}), to illustrate this. Let (a)-(f) be more literal renditions of each discourse unit in (\ref{complexback1}):

\begin{itemize}
    \item[($a$)] \textit{i egda byst\foreignlanguage{russian}{ъ} d\foreignlanguage{russian}{ъ}voju na desęte lětou} [And when he was twelve]
    \item[($b$)] \textit{v\foreignlanguage{russian}{ъ}šed\foreignlanguage{russian}{ъ}šem\foreignlanguage{russian}{ъ} im\foreignlanguage{russian}{ъ} v\foreignlanguage{russian}{ъ} im\foreignlanguage{russian}{ъ} po obyčajǫ prazd\foreignlanguage{russian}{ь}nika} [they having gone up to Jerusalem according to the custom of the feast]
    \item[($c$)] \textit{i kon\foreignlanguage{russian}{ь}čav\foreignlanguage{russian}{ъ}šem\foreignlanguage{russian}{ъ} d\foreignlanguage{russian}{ъ}ni} [and having completed the days]
    \item[($d$)] \textit{v\foreignlanguage{russian}{ъ}zvraštajǫštem\foreignlanguage{russian}{ъ} sę im\foreignlanguage{russian}{ъ}} [while they were returning]
    \item[($e$)] \textit{osta otrok\foreignlanguage{russian}{ъ} is v\foreignlanguage{russian}{ъ} imě} [the boy Jesus remained behind in Jerusalem]
    \item[($f$)] \textit{i ne čju iosif\foreignlanguage{russian}{ъ} i mati ego} [and Mary and Joseph did not realize]
\end{itemize}

\begin{figure}[!h]
    \centering
    \includegraphics[scale=0.5]{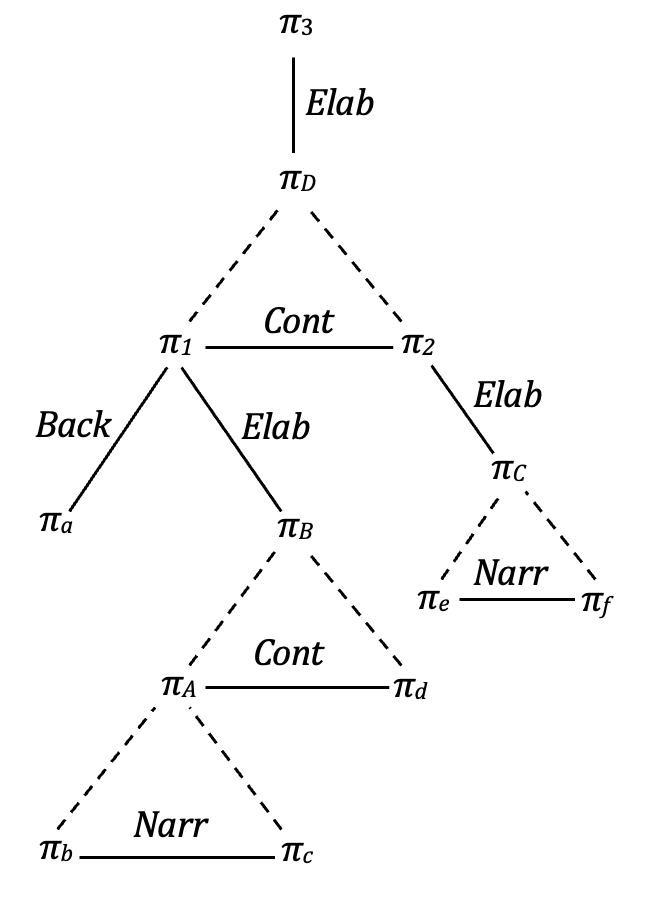}
    \caption[Discourse representation structure for a complex discourse unit in Old Church Slavonic involving several \textsc{frames} contributing to a common Framing Topic for a foreground constituent]{Discourse representation structure for (\ref{complexback1}), involving a complex discourse unit consisting of several \textsc{frames} contributing to a common Framing Topic for a foreground constituent}
    \label{drscomplexback}
\end{figure}

All discourse units from (a) to (d) eventually contribute to the FT, which the foregrounded eventuality, also a complex one and including both (e) and (f), elaborate on. A possible SDRS for (\ref{complexback1}) would look like (\ref{drscomplexback}). As is conventional in SDRT, $\pi$ is any discourse unit, elementary (simple) or complex. Subscript lowercase refers to elementary discourse unit (EDU), namely each of the units listed as (a) to (f). Subscript numbers refer to FTs, whereas uppercase refer to complex discourse units, namely discourse-coordinated units (both EDUs and FTs) dominated by the same node. \\
\indent First, $\pi_{a}$ introduces an FT, which the complex unit $\pi_{B}$, including all EDUs from (b) to (d), elaborate on. The EDUs inside $\pi_{B}$ are not all on the same level: $\pi_{b}$ and $\pi_{c}$ are linked via \textit{Narration}, whereas $\pi_{d}$ is linked to both $\pi_{b}$ and $\pi_{c}$ (i.e. $\pi_{A}$) via \textit{Continuation}, which signals that the second constituent ($\pi_{d}$) elaborates on the same topic as the first. This allows $\pi_{d}$ to be stative, and $\pi_{b}$ and $\pi_{c}$ to be eventive. Now, not only ($\pi_{1}$) but also referents from ($\pi_{B}$) can percolate up and become available for constructing an FT for the whole discourse ($\pi_{3}$). The FT ($\pi_{2}$) for the foregrounded eventuality, drawing discourse material from ($\pi_{3}$), is linked to ($\pi_{1}$) via \textit{Continuation}, since ($\pi_{2}$) elaborates on the same topic as ($\pi_{1}$). 

\begin{figure}[!h]
    \centering
    \includegraphics[scale=0.5]{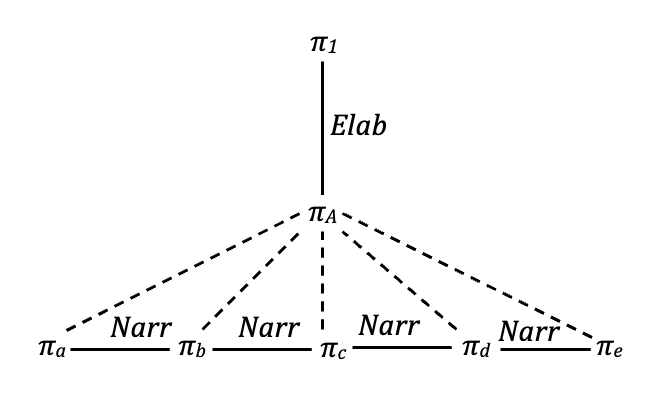}
    \caption[Discourse representation structure for a complex discourse unit involving an implicit Framing Topic]{Discourse representation structure for (\ref{unambindrheme}), involving an implicit Framing Topic}
    \label{drscomplexback2}
\end{figure}

We have thus established that complex frames all contribute to the construction of a discourse topic to which foregrounded narrative sequences may then be attached. The structure (\ref{drscomplexback}) is, of course, one specific scenario among the ways in which a framing topic can be constructed and in which a foreground can attach to a background. In particular, ($\pi_{C}$) can, in principle, be made of a chain of conjunct participles functioning as \textsc{independent rhemes}. In other cases, the construction of the framing topic may not be explicit, i.e. it might not be achieved with framing constructions, but merely by reference to the discourse at large---in fact, the accommodation of presupposition is one of the main functions of the SDRT relation Background (specifically in its \textit{Background$_{forward}$} `flavour') (\citealt{asherlasca1998,asher2007a}). A simpler discourse structure showing both the presence of \textsc{independent rhemes} in the foreground part of the structure in (\ref{fbp}) \textit{and} implicit background is the one for (\ref{unambindrheme}) above, whose structure might look as shown in Figure (\ref{drscomplexback2}). In Figure \ref{drscomplexback2} each discourse segment ($\pi_{a}$) to ($\pi_{d}$) corresponds to each conjunct participle in (\ref{unambindrheme}), while ($\pi_{e}$) corresponds to the main verb. The ($\pi_{1}$) constituent is structurally the same as ($\pi_{2}$) in Figure \ref{drscomplexback}, but instead of attaching to some discourse constituent ($\pi_{a}$) in the sentence, it will attach to the preceding discourse (via \textit{Background}) by solving the presupposition $\delta(\pi,\pi_{1})$. 

\section{Summary}
This chapter provided a tentative account of the typical functions of \textit{jegda}-clauses and participle constructions by drawing from frameworks for discourse representation (SDRT), particularly from accounts of temporal adverbials and time-locating adverbs.\\
\indent We have seen that previous work (such as \citealt{behrens2012a}) on the competition between finite clauses and non-finite adjuncts (loosely) based on discourse representation theories started from the assumption that the main difference between finite subordinates and non-finite adjuncts is that the former expresses a more precise semantic relation to the matrix. This observation potentially holds when comparing clearly framing \textit{perfective} participle constructions with \textit{perfective} \textit{jegda}-clauses. In that case, it is possible that greater semantic `preciseness’ brought up by \citet{behrens2012a} can be seen in the fact that framing perfective participles always cause narrative progression, which often has the effect that the matrix eventuality is located \textit{at some point soon} after the event referred to by the participle (within the consequent state). With perfective \textit{jegda}-clauses, this is not always necessarily the case, and we can, in fact, get an interpretation whereby the perfective \textit{jegda}-event includes a perfective matrix event (as in (\ref{egdaoccasion}) above, which \citet{declerck} would define as `occasion'-referring \textit{when}-clauses), that is, it is a more precise contribution only insofar as the \textit{jegda}-clause introduces a bounded temporal referent within which the matrix event is set. On the other hand, perfective framing participles seem to always introduce a consequent state and set the matrix event within it. \\
\indent We have then observed that, unlike claimed by \citet{behrens2012a}, not only do \textsc{independent rhemes} seem to typically trigger the SDRT relation \textit{Narration}, but they may specifically be introducing a subtype of it, which \citet{brasetal2001} call \textit{Strong Narration}. Strong Narration involves the pragmatic inference that the temporal relation between the event expressed by an \textsc{independent rheme} and the following event in line (another \textsc{independent rheme} or its matrix) is one of \textit{abutment}, namely of strict sequentiality with a stronger discourse topic requirement, which fits the idea that participle clauses as a whole are particularly well-suited as devices for narrative cohesion.\\
\indent Finally, we formalized a typical Early Slavic example containing multiple framing clauses on the basis of the SDRT accounts of the rhetorical relation \textit{Background}. The formalization suggested that the discourse properties of this relation and related observations on \textit{Narration} are consistent with the observations made on both framing constructions (\textit{jegda}-clauses and participles) and independent rhemes, particularly the discourse topic requirement and its attachment to the subsequent discourse via \textit{Narration} and to the previous discourse via \textit{Background}.

\chapter{The semantic map of WHEN and its typological parallels}

\section{Massively parallel Bible corpora: beyond ancient Indo-European translations}\label{intro}
The previous chapters have shown that the discourse functions of Early Slavic predicative participle can partly be inferred compositionally from a combination of syntactic configuration, tense-aspect of both matrix and participle clause, and lexical fillers.\footnote{This chapter is the result of a collaboration with Dag Haug (\citealt{haugpedrazzini}).} Following \posscitet{baryhaug2011} treatment of Ancient Greek predicative participles, Early Slavic participle clauses were treated as grammaticalized discourse relations, with a main tripartite distinction between \textsc{frames}, \textsc{independent rhemes}, and \textsc{elaborations}. We have seen that conjunct participles and absolute constructions have overall quite different distributions and discourse functions in the Early Slavic corpus, despite their partial competition as \textsc{frames} at the beginning of a discourse segment. We have also seen that finite subordinates introduced by \textit{jegda} `when' compete with both constructions as frame setters, but overall appear to be distributionally more similar to absolute clauses.\\
\indent To allow us to make generalizations about core semantic-pragmatic properties of different competing temporal constructions, this chapter takes a temporary step back from Early Slavic and Greek with the goal of investigating whether the division of labour between different participle clauses and finite temporal subordinates in Early Slavic and Greek corresponds to well-attested cross-linguistic patterns. English \textit{when} is used as the new point of departure, for two main reasons. First, the scope of \textit{when}-clauses spans both that of \textit{jegda}-clauses and that of conjunct participles and absolute constructions, as can be evinced by the counterparts to each of these constructions in different English versions of the Bible, that is, all three OCS constructions can correspond to a \textit{when}-clause in English. In fact, any of the three Early Slavic competitors would be considered under the generic label of \textsc{when}-clauses in some typological frameworks to temporal subordination (cf. \citealt{cristofarowals}).\footnote{Small caps \textsc{when} will be used throughout the chapter to refer to the semantic concept, and italicized \textit{when} for the English lexical item.} Second, their semantic properties have been more widely studied and are relatively well understood, despite their complexity, as we have seen in Chapter 3. \\
\indent To explore the expression of \textsc{when} cross-linguistically and see how the ground covered by English \textit{when} is expressed across languages, we use \citeauthor{mayer-cysouw}'s (\citeyear{mayer-cysouw}) massively parallel corpus, which contains the New Testament in more than 1400 languages. The data is described in detail in Section \ref{sec:data}.\\
\indent To explore the data, probabilistic semantic maps are employed, as they are now a well-established tool in language typology for capturing universal correspondences between classes of forms and ranges of highly similar situational meanings across `massively cross-linguistic' datasets (\citealt{walchli-cysouw2012}). Probabilistic semantic maps can deal with very large datasets containing great degrees of variation within and across languages (\citealt{croft-poole-2008}), and unlike traditional implicational semantic maps (\citealt{haspelmath2003}) they do not rely on a limited set of posited abstract functions and translational equivalents. The premises and central aim of this thesis---to investigate the competing motivations governing the choice of similar temporal constructions---strikes a clear chord with the theoretical bases of probabilistic semantic maps, namely:
\begin{itemize}
    \item \textit{similarity semantics}, whereby similarity is considered to be a more basic notion than identity (see \citealt{walchli-cysouw2012});
    \item \textit{exemplar semantics}  (\citealt{croft-exemplarsem}), whereby the abstract functional domains assumed by traditional semantic-map approaches (\citealt{haspelmath2003}) are in fact clusters of `contextually embedded' situations.
\end{itemize}
\pgcitet{walchli-cysouw2012}{677} specifically argue for probabilistic semantic maps as a good fit for analyses in which usage-related factors, including discourse structure, are particularly relevant, which is a welcome observation in this study, since \textit{jegda}- and participle clauses play an important role in discourse-structuring, as seen in the previous chapters. The methods used to generate and analyse the semantic maps for the analysis are described in detail in Section~\ref{sec:methods}.

\section{Data}\label{sec:data}
\citet{mayer-cysouw} massively-parallel Bible corpus\footnote{I am grateful to Michael Cysouw for sharing the repository with the full parallel corpus with me and Dag Haug.} comprises translations representing 1465 ISO 639-3 language codes.\footnote{As of January 2023.} As noted in \citet{cysouwgood}, an ISO 639-3 code should be understood as referring to a \textsc{languoid}, a generalization of the term \textit{language} referring to the grouping of varieties as represented in specific resources (\textsc{doculects}) without the common constraints associated with the definition of language, dialect or family. This is crucial for avoiding creating the misconception that the `languages’ represented in the dataset are defined as such by virtue of their sociolinguistic status. Rather, each of them can be considered as sets of \textsc{doculects} at some level of hierarchical grouping. For practical purposes, the varieties represented by each Bible translation in the parallel corpus will be referred to as `languages’, with the caveat in mind that not all the varieties with an ISO 639-3 code will equally correspond to what is generally considered a ‘language’.\footnote{The Glottolog database (\url{https://glottolog.org}; \citealt{nordhoffhammarstrom}, \citealt{glottolog2021}), for example, which adopts a \textsc{doculect}-based approach while also grouping languoids into successively larger `levels' (such as subdialects, dialects, languages, subfamilies and families) classifies 15 of the languages in our dataset as dialects. Norwegian Bokmål (\textsc{nob}) and Norwegian Nynorsk (\textsc{nno}), for example, are considered ‘dialects’ of Norwegian (\textsc{nor}), even though the latter is in fact defined collectively by the combination of the former two (among other `dialects'). ‘Norwegian’ (\textsc{nor}), then, could therefore be considered as a languoid at a higher hierarchical level than the languoids Norwegian Bokmål and Norwegian Nynorsk.}

\begin{table}[!h]
\centering
\begin{tabular}{rrrrr}
\hline
\textbf{family} & \textbf{bible\textunderscore raw} & \textbf{bible\textunderscore rel} & \multicolumn{1}{l}{\textbf{world\textunderscore raw}} & \multicolumn{1}{l}{\textbf{world\textunderscore rel}} \\
\hline
\textbf{\small Atlantic-Congo} & 249 & 17.2\% & 1380 & 18.1\% \\
\textbf{\small Austronesian} & 246 & 17.0\% & 1289 & 16.9\% \\
\textbf{\small Indo-European} & 110 & 7.6\% & 595 & 7.8\% \\
\textbf{\small Nuclear Trans New Guinea} & 94 & 6.5\% & 313 & 4.1\% \\
\textbf{\small Sino-Tibetan} & 90 & 6.2\% & 441 & 5.8\% \\
\textbf{\small Otomanguean} & 79 & 5.5\% & 180 & 2.4\% \\
\textbf{\small Afro-Asiatic} & 47 & 3.3\% & 371 & 4.9\% \\
\textbf{\small Quechuan} & 27 & 1.9\% & 45 & 0.6\% \\
\textbf{\small Uto-Aztecan} & 26 & 1.8\% & 64 & 0.8\% \\
\textbf{\small Mayan} & 25 & 1.7\% & 35 & 0.5\% \\
\hline
\end{tabular}
\caption[The 10 most frequent language families in our dataset compared to the 10 most frequent families among the world's languages according to the Glottolog classification]{The 10 most frequent language families in our dataset compared to the 10 most frequent families among the world's languages according to the Glottolog classification. \textit{bible\textunderscore } refers to the former, \textit{world\textunderscore } to the latter. \textit{raw} are raw number of languages belonging to the relevant family, \textit{rel} is the relative frequency of these in relation to the total number of languages in the respective dataset (the parallel Bible dataset for \textit{bible\textunderscore }, the whole Glottolog language database for \textit{world\textunderscore }).}\label{tab:10mostfreqparallelbible}

\vspace{1cm}
\centering
\begin{tabular}{rrrrr}
\hline
\textbf{family} & \textbf{world\textunderscore raw} & \textbf{world\textunderscore rel} & \textbf{bible\textunderscore raw} & \textbf{bible\textunderscore rel} \\
\hline
\textbf{\small Atlantic-Congo} & 1380 & 18.1\% & 249 & 17.2\% \\
\textbf{\small Austronesian} & 1289 & 16.9\% & 246 & 17.0\% \\
\textbf{\small Indo-European} & 595 & 7.8\% & 110 & 7.6\% \\
\textbf{\small Sino-Tibetan} & 441 & 5.8\% & 90 & 6.2\% \\
\textbf{\small Afro-Asiatic} & 371 & 4.9\% & 47 & 3.3\% \\
\textbf{\small Nuclear Trans New Guinea} & 313 & 4.1\% & 94 & 6.5\% \\
\textbf{\small Pama-Nyungan} & 268 & 3.5\% & 13 & 0.9\% \\
\textbf{\small Otomanguean} & 180 & 2.4\% & 79 & 5.5\% \\
\textbf{\small Austroasiatic} & 160 & 2.1\% & 10 & 0.7\% \\
\textbf{\small Tai-Kadai} & 91 & 1.2\% & 3 & 0.2\% \\
\hline
\end{tabular}
\caption[The 10 most frequent families among the world's languages according to the Glottolog classification compared to the 10 most frequent language families in our dataset]{The 10 most frequent families among the world's languages according to the Glottolog classification compared to the 10 most frequent language families in our dataset. The same labelling as Table \ref{tab:10mostfreqparallelbible} applies.}\label{tab:10mostfreqglottolog}
\end{table}

Several of the languages in \citeauthor{mayer-cysouw}’s \citeyear{mayer-cysouw} parallel corpus have multiple translations and a few contain only (or predominantly) the Old Testament. To obtain the best textual coverage for the largest number of varieties possible, only languages with a version of the New Testament were considered. For languages with multiple translations, the New Testament version with the widest coverage in terms of verses was selected. If the difference in coverage between versions was less than 2000 verses, the different versions were considered as having the same coverage, in which case the most recent one was chosen.\\
\indent Although \citeauthor{mayer-cysouw}'s corpus already contains versions for some historical languages, for Ancient Greek, Church Slavonic, Latin, Gothic, and Classical Armenian, their versions from the PROIEL Treebank \citep{proiel} were used to facilitate the potential integration of their several layers of linguistic annotations in the semantic maps.\\
\indent The final dataset comprises 1444 languages (around 19\% of the world's languages), representing, following the Glottolog classification, 121 families and 16 language isolates. In comparison, the world's languages are currently classified into 233 families and 167 isolates.\footnote{These numbers do not include some of the `non-genealogical trees' to which some languages are assigned by Glottolog, specifically \textsc{Unclassifiable}, \textsc{Unattested}, and \textsc{Speech Register}. \textsc{Sign languages}, \textsc{Mixed languages}, and \textsc{Pidgins} are instead considered in the numbers and they are therefore counted in the frequencies in Tables \ref{tab:10mostfreqparallelbible}-\ref{tab:10mostfreqglottolog}. So-called \textsc{Bookkeeping languoids} are also excluded from the counts. These exclusions explain why the figures reported here are slightly different from those reported on the Glottolog webpage (\url{https://glottolog.org/glottolog/glottologinformation}).} Tables \ref{tab:10mostfreqparallelbible}-\ref{tab:10mostfreqglottolog} give an overview of the language families most represented in our dataset compared to their frequency in the world's languages according to the Glottolog database. The top three families among the world's languages, the Atlantic-Congo, Austronesian, and Indo-European occupy the same position in our dataset and show very similar relative frequencies to those found in Glottolog. By comparing Table \ref{tab:10mostfreqparallelbible} with Table \ref{tab:10mostfreqglottolog} we can see that the Nuclear Trans New Guinea, Sino-Tibetan, Otomanguean are slightly overrepresented in our dataset compared to the world's languages, while this is certainly the case for the Quechuan, Uto-Aztecan, and Mayan families. On the other hand, the Afro-Asiatic, Pama-Nyungan, Austroasiatic, and Tai-Kadai families are rather heavily underrepresented in our dataset. The families not represented at all in our dataset constitute around 48\% of the world's families and comprise, for the most part, fewer than 10 languages.\footnote{The complete list of the world's language families used to extract the counts reported here, their frequency according to Glottolog and in our dataset can be found in the Appendix.}

\begin{figure}[!h]
\centering
\includegraphics[width=140mm]{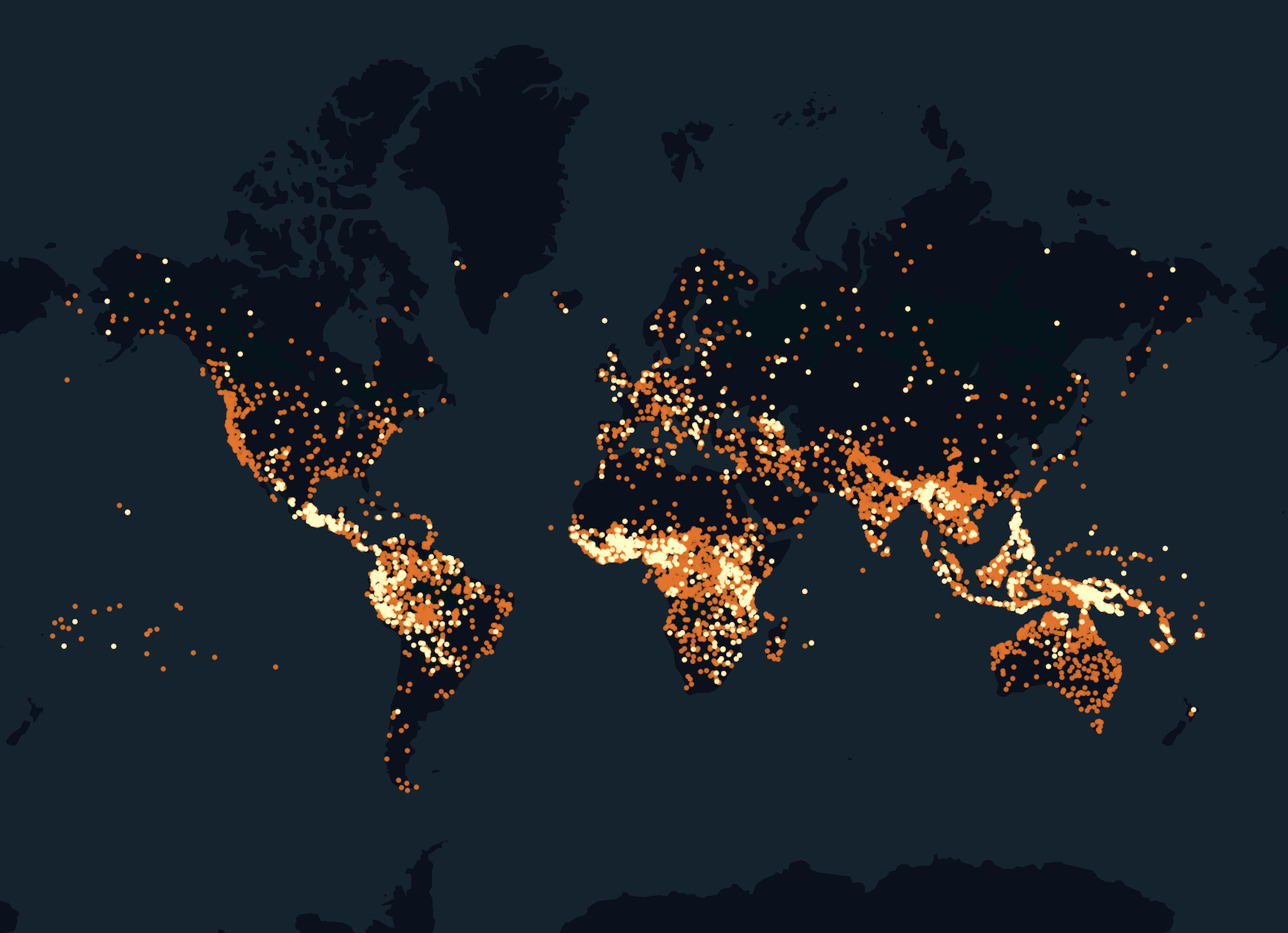}
\caption[Approximate areal distribution of the languages in the massively parallel dataset among the world's languages]{Approximate areal distribution of the languages in our dataset (light yellow) among the world's languages (orange)} \label{arealdistr} 
\end{figure}

In terms of areal distribution, following the Glottolog classification into six main macro-areas (Africa, Australia, Eurasia, North America, South America, Papunesia), as Table \ref{tab:arealfreq} shows languages from Africa and Australia are underrepresented in our dataset, particularly from the former, while languages from the Americas are somewhat overrepresented. Figure \ref{arealdistr} maps the distribution of the languages in our dataset among the world's languages.\footnote{The points in the maps are obviously an approximation of where a particular language is used. The coordinates for the map in Figure \ref{arealdistr} are from Glottolog.}

\begin{table}[]
\centering
\begin{tabular}{rrrrr}
\hline
\textbf{macro-areas} & \textbf{bible\textunderscore raw} & \textbf{bible\textunderscore rel} & \textbf{world\textunderscore raw} & \bf{world\textunderscore rel} \\
\hline
\textbf{Papunesia} & 415 & 28.7\% & 2136 & 28.1\% \\
\textbf{Eurasia} & 336 & 23.3\% & 1743 & 22.9\% \\
\textbf{Africa} & 335 & 23.2\% & 2196 & 28.9\% \\
\textbf{North America} & 181 & 12.5\% & 674 & 8.9\% \\
\textbf{South America} & 157 & 10.9\% & 488 & 6.4\% \\
\textbf{Australia} & 20 & 1.4\% & 371 & 4.9\% \\
\hline
\end{tabular}
\caption{Areal distribution of the languages in our dataset compared to the world's languages, following the classification into macro-areas from Glottolog.}\label{tab:arealfreq}
\end{table}

\section{Methods}\label{sec:methods}
The texts in the target languages were aligned to the English one at the word level, using SyMGIZA++ (\citealt{symgiza}), a modification of the well-known GIZA++ program (\citealt{gizapp}) that allows to train two-directed word alignment models in parallel. The results are one-to-one alignment models, namely one token in the source language corresponding to only one token, or no token at all, in the target language (as opposed to a one-to-many or many-to-one alignment).\\
\indent SyMGIZA++ was chosen over the popular and much faster FastAlign model (\citealt{fastalign}) after applying some heuristics to gauge the quality of their results. Because of the sheer number of languages in the parallel corpus, some bias in the evaluation method will necessarily be introduced, since it requires familiarity with both source and target language. Rather than evaluating the alignment across the board, we checked a randomly selected subsample (10\%) of all the sentences (= 876) containing \textit{when}-clauses in English and calculated the accuracy of the alignment between the token \textit{when} and its respective forms, or lack thereof, in the Norwegian, Russian and Italian versions. 
SyMGIZA++ yielded 96.5\% accuracy on the Norwegian test set, 90\% on the Russian, and 77.9\% on the Italian one, whereas FastAlign only yielded 77.9\%, 75.9\% and 59.3\%, respectively.
Overall, SyMGIZA++ and FastAlign performed similarly at identifying the correct parallel when the target language uses a subordinator, but FastAlign generally aligned \textit{when} to some other token in the absence of a direct parallel (e.g. to a conjunction or an auxiliary verb), whereas SyMGIZA++ more often explicitly indicated the lack of a parallel with a `NULL' alignment, which intuitively means that the target language uses a construction with no subordinator (e.g. an independent clause) or a construction where the subordination is expressed morphologically (e.g. a converb).

Before training the final models with SyMGIZA++, minimal preprocessing (lowercasing and punctuation removal) was applied. The occurrences of English \textit{when} and its parallels in all the target languages were then extracted. Each instance of \textit{when} and its parallels was treated as one usage point for the hypothesized semantic concept \textsc{when}, and Hamming distance was used as a measure of similarity between pairs of contexts, leveraging the number of times a language uses two different linguistic means where English uses one (\textit{when}). For example, based on the six languages shown in Table \ref{fig:datamatrix}, the distance between the two contexts is 3, because they differ in the word choice in Maori (mri), Finnish (fin) and Kazakh (kaz). 

\begin{table}[!h]
\centering
\begin{tabular}{ccccccccc}
\hline
& \textbf{eng} & \textbf{mri} & \textbf{por} & ... & \textbf{fin} & \textbf{kaz} & \textbf{kor} \\ \hline
1 & when & no & quando & ... & kun & \foreignlanguage{russian}{қашан} & \begin{CJK}{UTF8}{mj}때에\end{CJK}\\
2 & when & ka & quando & ... & jolloin & \foreignlanguage{russian}{кейін} & \begin{CJK}{UTF8}{mj}때에\end{CJK}\\
\textit{\textbf{n}} & ... & ... & ... & ... & ... & ... & ... \\
\hline
\end{tabular}
\caption{Matrix of \textit{when} and aligned tokens (sample).}
\label{fig:datamatrix}
\end{table}

After calculating the Hamming distance between all pairs of contexts, we obtain a $n \times n$ Hamming-distance matrix, where both column and rows correspond to the contexts, and each cell indicates the Hamming distance between the respective two contexts.

\begin{figure}[!h]
\centering
\includegraphics[width=150mm]{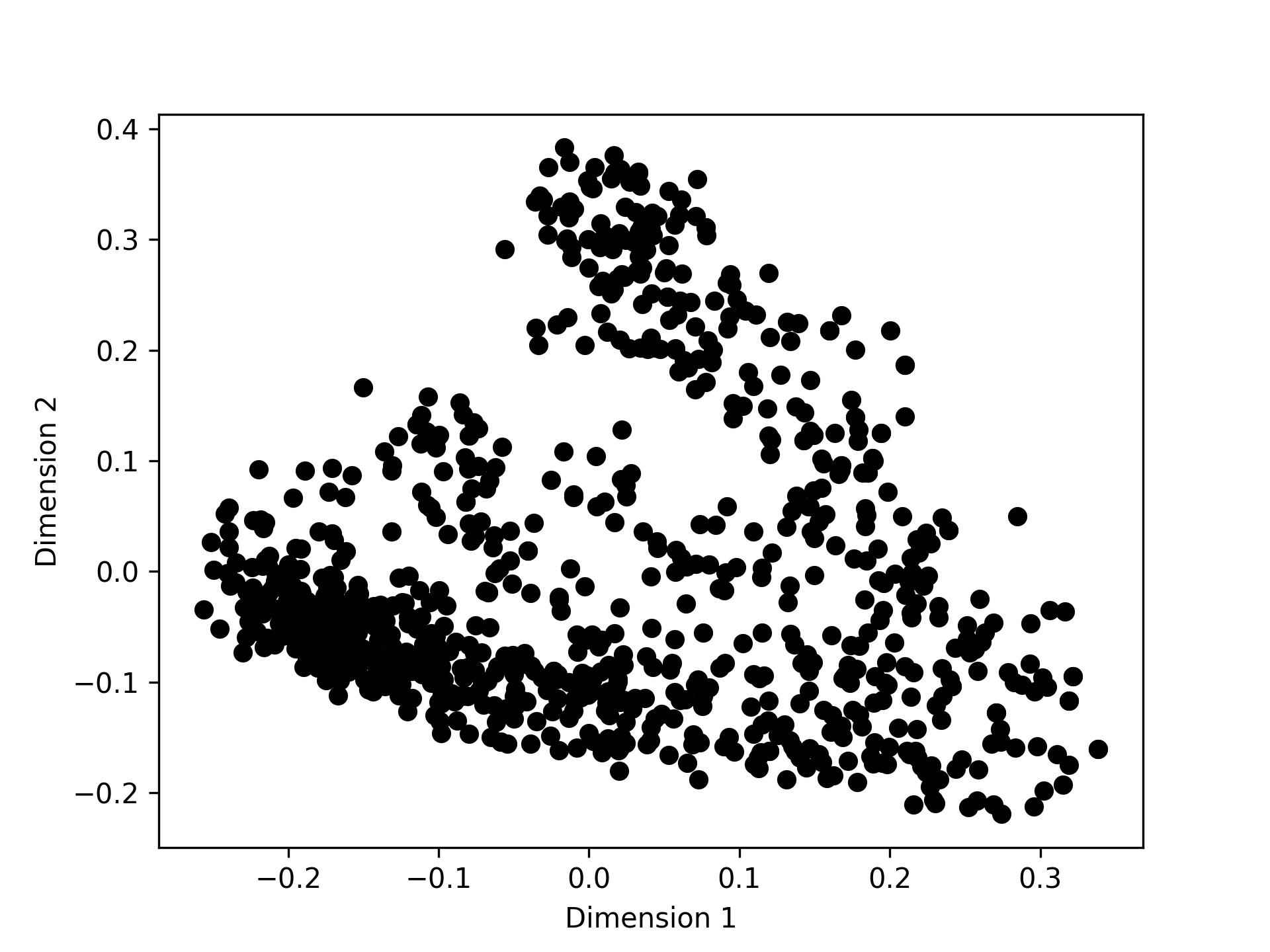}
\caption{\label{plainmds}Basic semantic map of \textsc{when} resulting from plotting the first two dimensions of the MDS matrix}
\end{figure}

Multidimensional scaling (MDS), as implemented by the \texttt{cmdscale} function from the R package \textit{stats} (\citealt{cmdscalecit}), is then applied to the resulting matrix to visualize the distance between each occurrence of \textit{when} two-dimensionally. This output is shown in Figure \ref{plainmds}. Each dot represents a usage point for \textsc{when} (i.e., a Bible verse). If two dots are far apart, they tend to be expressed with different lexical items across the languages in the corpus.

Clusters of semantically similar observations are identified and analysed in two main ways. First, similarly to \citet{hartcysouwhaspl}, starting from the MDS matrix, Kriging is applied as an interpolation method that uses a limited set of sampled data points (each observation in the target languages) to estimate the value of a variable in an unsampled location.

\begin{figure}[!h]
\centering
\includegraphics[width=120mm]{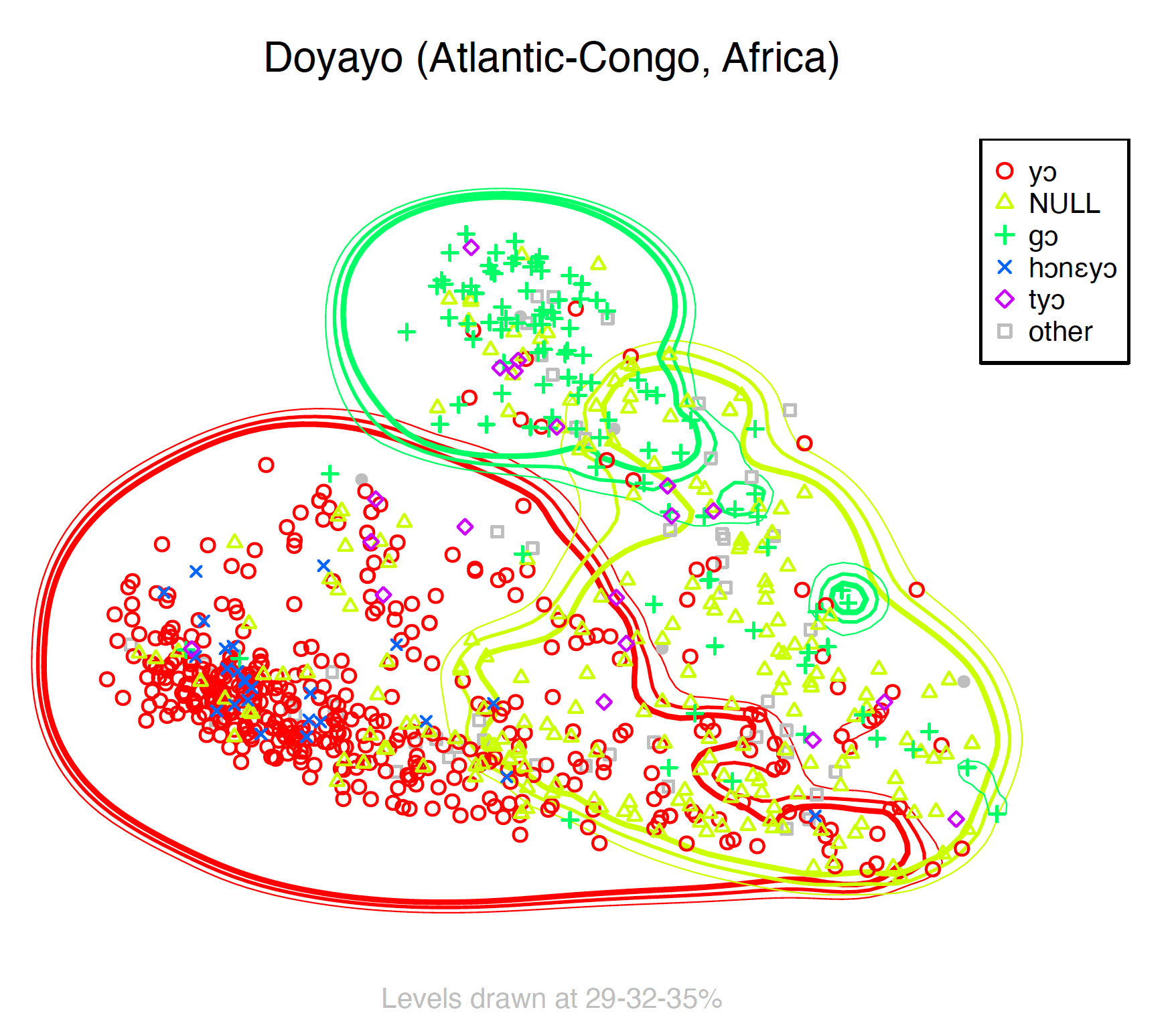}
\caption[Kriging map for Doyayo]{\label{dow}Kriging map for \textit{when}-parallels in Doyayo (Atlantic-Congo, Africa).}
\end{figure}

As an example, Figure \ref{dow} shows the resulting semantic map for Doyayo (Atlantic-Congo, Africa) after applying Kriging to the MDS matrix by using the parallels to English \textit{when} in the language to interpolate the areas shown in green, red, and yellow in the figure. Unlike \citet{hartcysouwhaspl}, we start from one single means (\textit{when}), without pre-emptively assigning a semantic label to the different \textit{when}-situations in English, so that the discernible Kriging-areas in the semantic maps of the target languages must be interpreted on the basis of comparison between similar cross-linguistic patterns. Like \citet{hartcysouwhaspl}, the \verb|Krig| function from the R package \textit{fields} (\citealt{fields}) was used to draw lines at different levels of probability distributions (35\%, 32\% and 29\%). Unlike traditional semantic maps, where boundaries are drawn around all observations of the same type (i.e. the same means in a given language), the lines in the Kriging map in Figure \ref{dow} represent the probability for a particular means to be dominant within their boundaries. This is why, for instance, red points in Figure \ref{dow} can also be found outside the red area identified by Kriging, but it also explains why relatively large areas can overlap, as the points between the red and yellow areas in the figure show. The latter is a welcome feature of this approach, albeit more challenging to account for than non-probabilistic ones, since it reflects the intuition that situations encoded by any two given competing constructions may be envisioned as the union of the sets of all situations which can be encoded by either of those constructions.

\begin{figure}[!h]
\centering
\includegraphics[width=150mm]{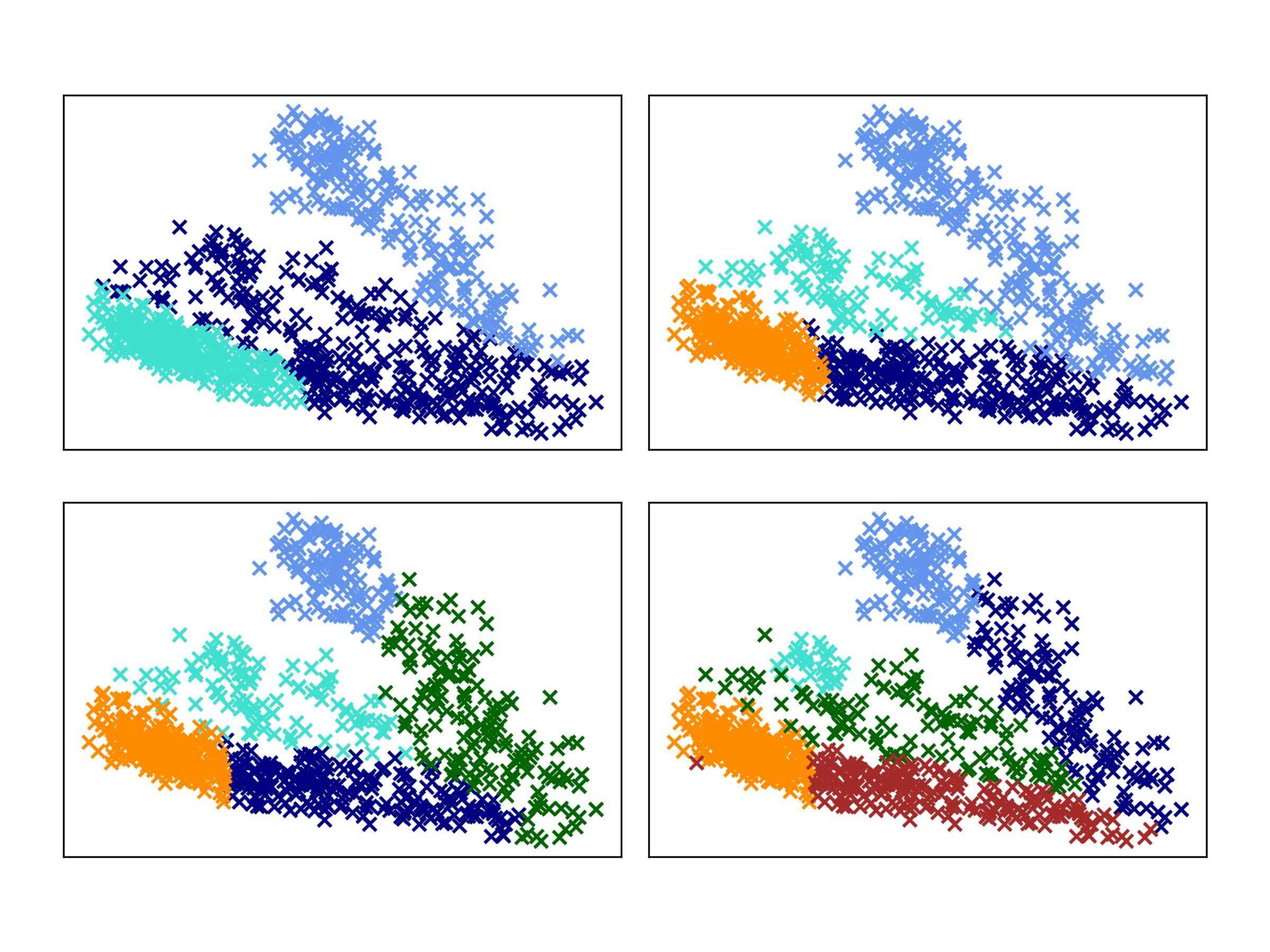}
\caption{\label{gmm}Clusters identified via Gaussian Mixture Modelling and \textit{k}-means initialization, based on the first two dimensions of the MDS matrix: comparison between 3, 4, 5 and 6 clusters.}
\end{figure}

Second, we fit a Gaussian Mixture Model (GMM) to the MDS matrix, to identify clusters which are more likely to correspond to separate universal functions of \textsc{when}, regardless of how much variation a particular language shows within any of the clusters (which could go from no variation across the whole map or across one cluster, to several linguistic means in a single cluster). The result is shown in Figure~\ref{gmm}.

\begin{figure}[!h]
\centering
\begin{tabular}{cc}
  \includegraphics[width=0.5\linewidth]{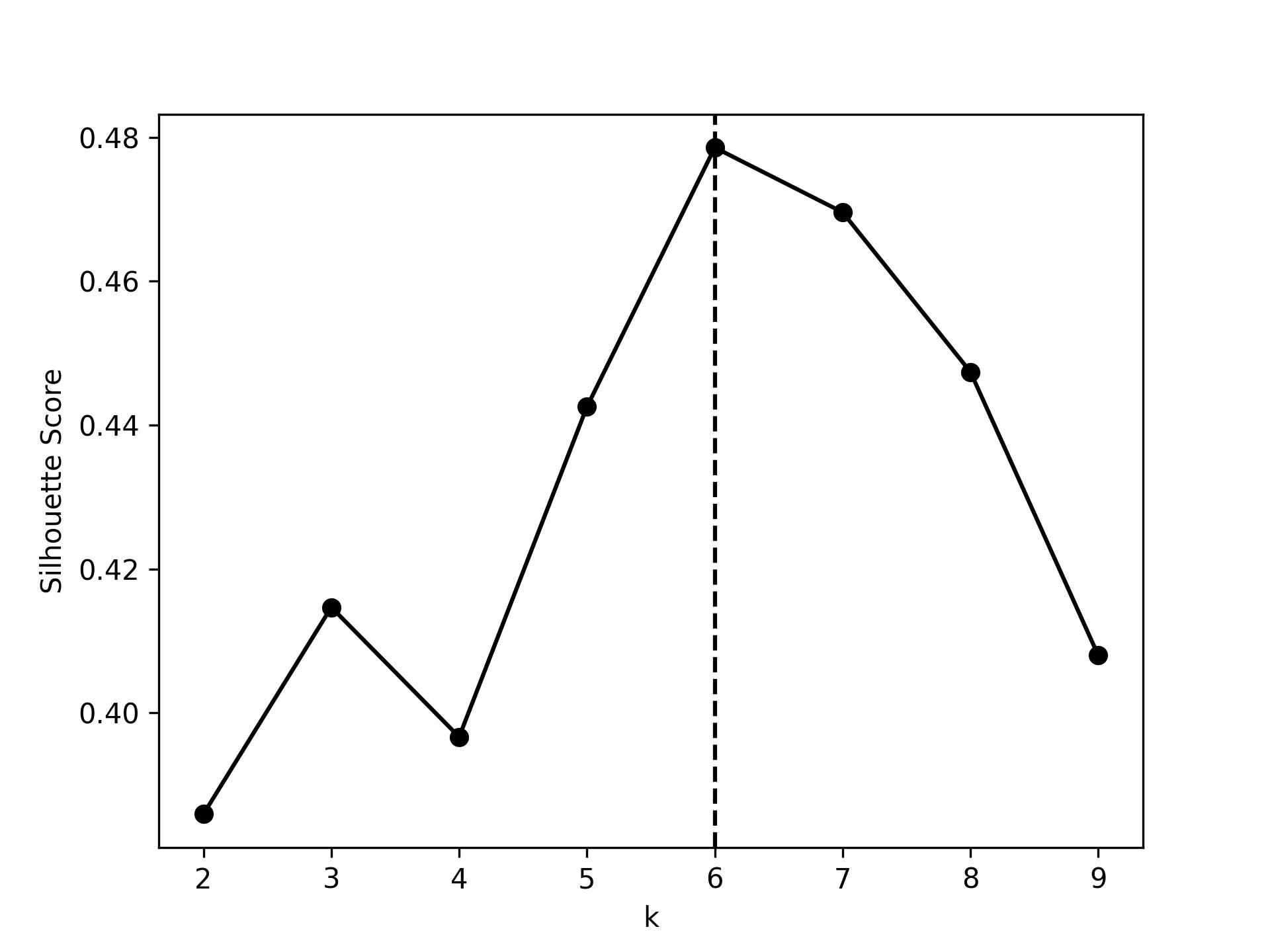} &   \includegraphics[width=0.5\linewidth]{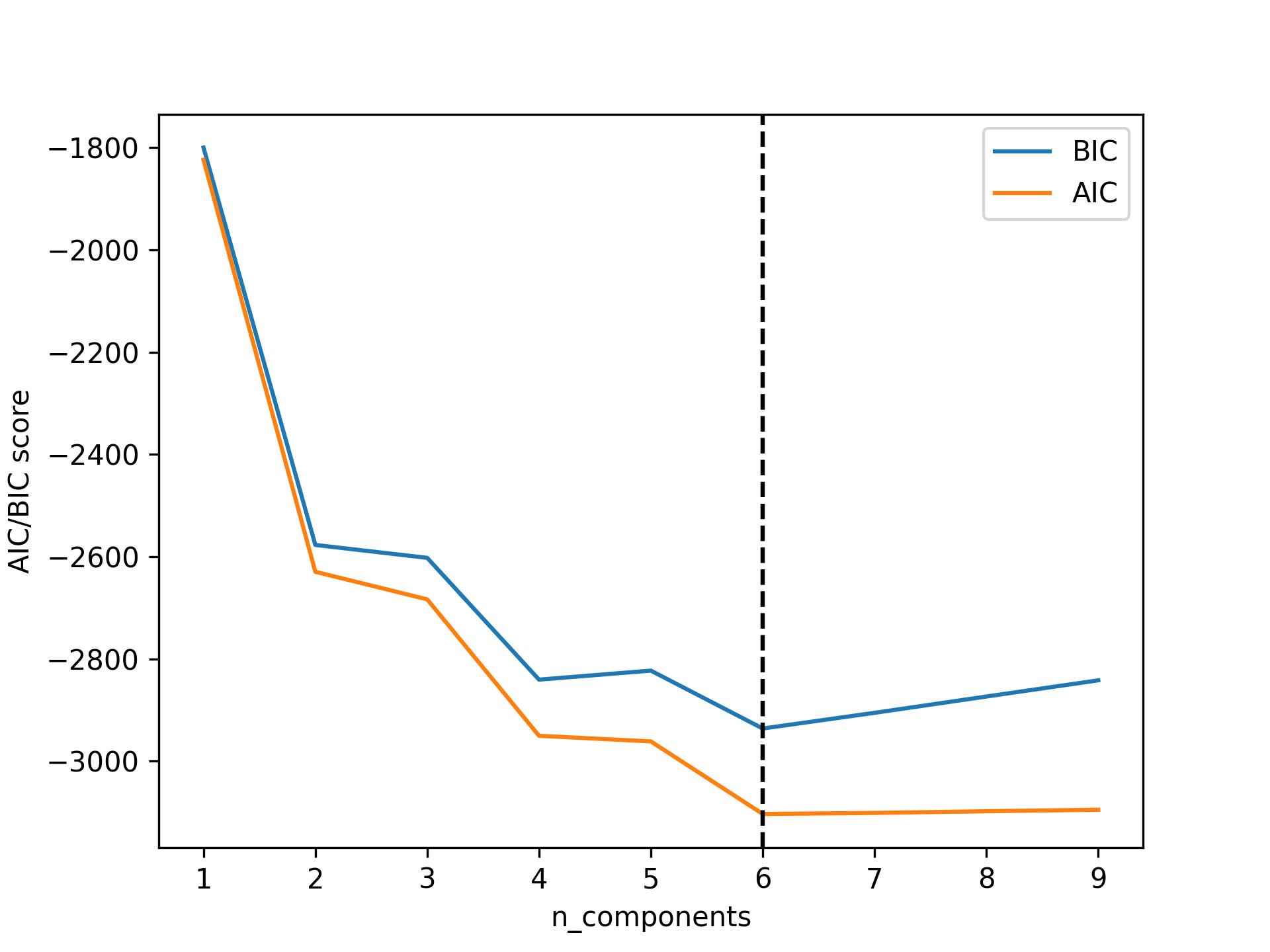}\\
\end{tabular}
\caption[Evaluation of the optimal number of clusters for a GMM model of the two dimensional MDS matrix (Silhouette, AIC and BIC)]{Evaluation of the optimal number of clusters for a Gaussian Mixture Model of the two-dimensional MDS matrix. \textbf{(Left)} Silhouette score: higher value is better. \textbf{(Right)} Akaike information criterion (AIC) and Bayesian information criterion (BIC): lower value is better.}
    \label{aicbicsilhouette}
\end{figure}

GMM assigns data to a given number of clusters based on probability distributions rather than on the distance from a centroid, as in other well-established clustering algorithms (e.g. \textit{k}-means). This means that GMMs also allow for elliptical clusters, unlike \textit{k}-means, which will always result in roughly spherical shapes, since they are based on mere Euclidean distance and thus cannot sensibly approximate the semantic map of competing constructions, which are, by definition, more of a continuum than a set of clearly separate areas.\\
\indent The number of components (clusters) for the GMM models is chosen with the help of different heuristics. In Figure \ref{aicbicsilhouette} (left), we can see that the optimal number (based on the first two dimensions of the MDS matrix) is 6 clusters. Figure \ref{aicbicsilhouette} (right) indicates the best number of clusters based on the Akaike information criterion (AIC) \citep{aic} and the Bayesian information criterion (BIC) \citep{bic}. Both AIC and BIC agree with the Silhouette score that 6 is the number of clusters for the best trade-off between model fit and complexity.

\begin{figure}[!h]
\centering
\includegraphics[width=120mm]{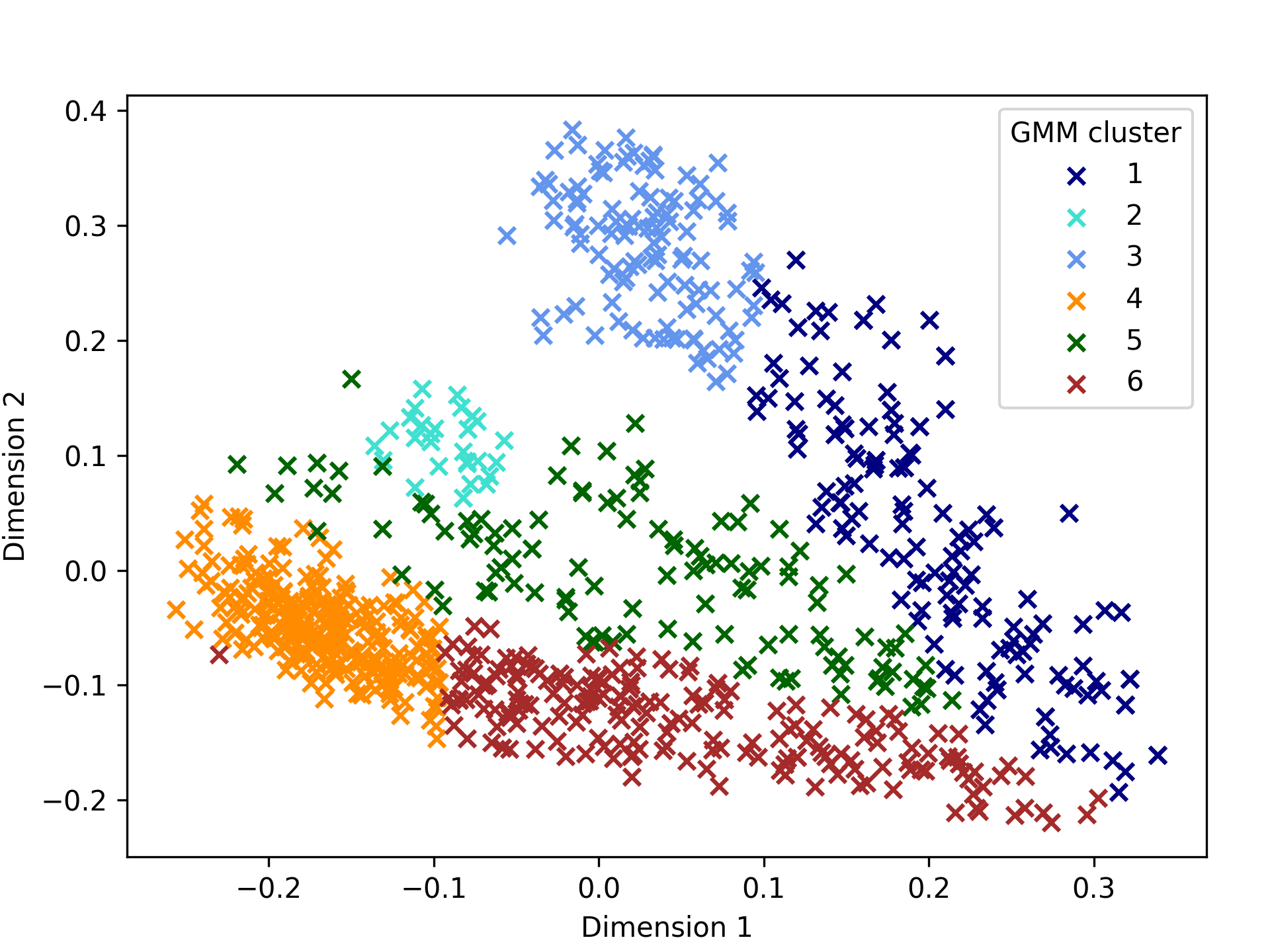}
\caption{\label{gmm6}Gaussian Mixture Model of the two-dimensional MDS matrix (6 clusters)}
\end{figure}

\indent These methods are meant to indicate how many clusters can be considered maximally separate from each other. However, empirically, we know that the temporal constructions under consideration are often competing and that their scopes are not at all clear-cut. While taking this into account, the GMM model consisting of 6 clusters (Figure \ref{gmm6}) was selected, given the agreement between the Silhouette and the BIC/AIC scores.

Each cluster in Figure \ref{gmm6} \emph{potentially} corresponds to a specific functional domain of \textsc{when}.
To test whether this is in fact the case, we check to what extent the languages have lexical items that align well with each GMM cluster.
Concretely, all the attested means used by each language are first extracted for the individual clusters. For each attested means, its occurrences in that cluster are counted as \textit{true positives}, its occurrences outside that cluster as \textit{false positives}, and the occurrences of other means in that cluster as \textit{false negatives}. The \textit{precision} of a means as a rendering of the \textit{when}-clauses corresponding to that cluster, then, is the number of true positives divided by the sum of true positives and false positives; the \textit{recall} is the number of true positives divided by the sum of true positives and false negatives. We can then compute the \textit{F1 score} (harmonic mean of precision and recall) for each means and, for each language, plot the precision and recall of the word with the highest F1 scores. A means with a high F1 score will correspond to a likely lexification of the relevant GMM cluster in its particular language. A frequent high F1 score across several languages may instead indicate a common lexification pattern, as we will see in Section \ref{sec:analysis}.

\indent Finally, leveraging the advantages of both the Kriging and the GMM methods, the Kriging areas that best correspond to each GMM cluster in each language can be identified. As we will see in Section~\ref{sec:analysis}, this allows us to study colexification patterns across languages. The alignment of Kriging areas and GMM clusters runs as follows.

\begin{itemize}
    \item[1.] For each of \textit{n} number clusters, across which cross-linguistic variation in co-lexification is to be investigated, calculate its centroid. This is the sum of the coordinates of the points belonging to each cluster, divided by the number of points in the cluster, namely:
    
    \[(\frac{1}{j}\sum_{i=1}^{j} x_i , \frac{1}{j}\sum_{i=1}^{j} y_i )\]

    where $j$ is the number of points in a GMM cluster, $x$ are the x-coordinates (i.e. dimension 1 of the MDS matrix) and $y$ the y-coordinates (i.e. dimension 2 of the MDS matrix). The centroid of a GMM cluster is preliminarily assumed to be the best representation of that cluster. Note that it is unlikely to correspond to an actual observation in the target languages.\footnote{The procedure can, in principle, have groups of observations drawn from clusters obtained with any method as a starting point. A group can also be made of one individual observation.} 

    \item[2.] For each GMM cluster, extract $k$ actual observations corresponding to the \textit{k}-nearest neighbours of the centroid of that cluster. The value $k$ should be adjusted in a trial-and-error fashion; we set ours to 30, i.e. 30 points are extracted for each cluster. The nearest neighbours were identified using the balltree approach (\citealt{balltree}), a space partitioning system which can be applied to multi-dimensional space for nearest neighbour search.\footnote{We used the implementation of balltree by Scikit-Learn (\citealt{scikit-learn}).} The result of the search is a group of `core' points surrounding the centroid of the GMM cluster. Figure \ref{fig:centroidwithballtree} shows the three groups from our experiment.

    \begin{figure}[!h]
    \centering
    \includegraphics[width=120mm]{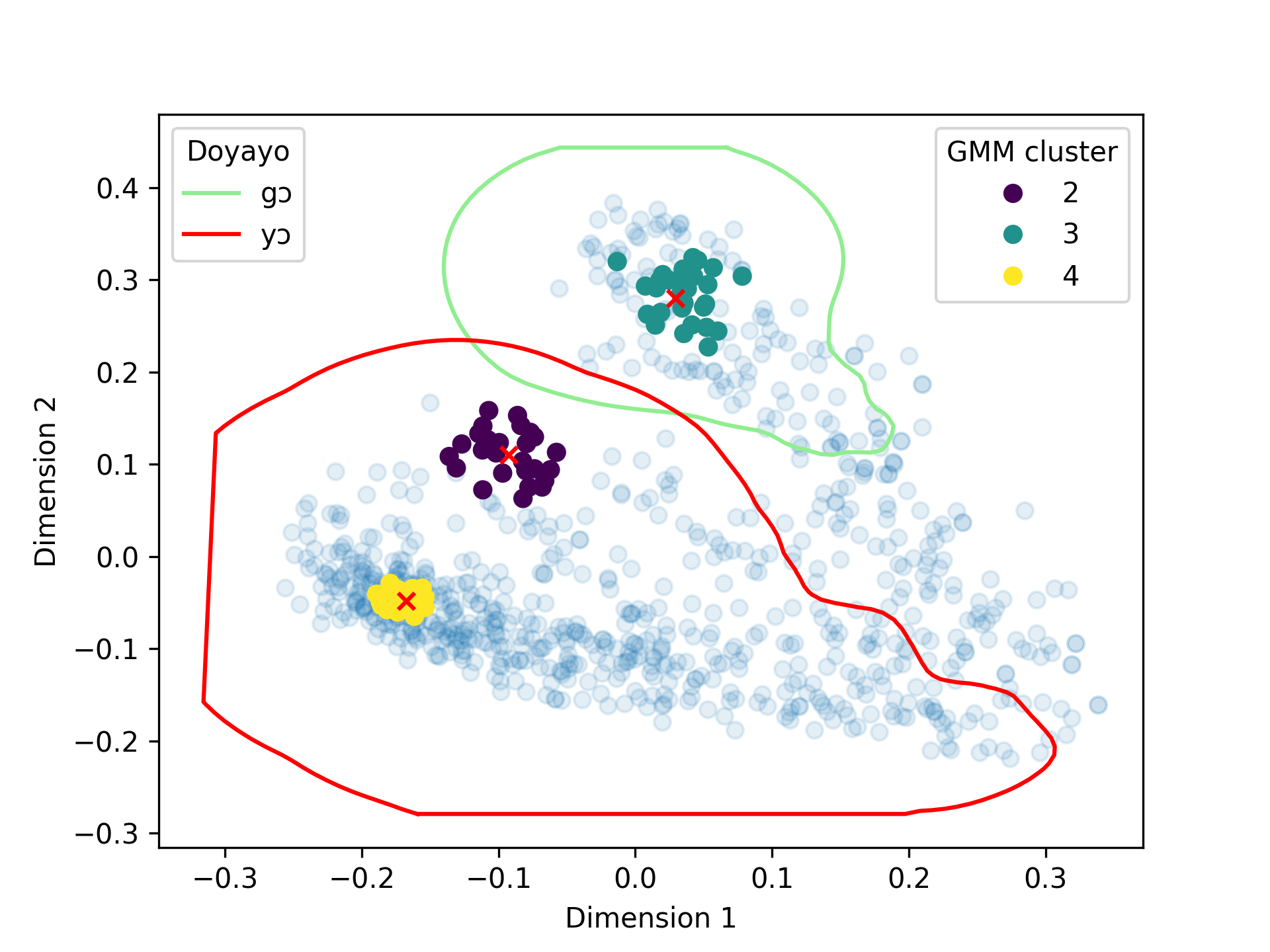}
    \caption[30-nearest-neighbour search using the balltree method, with an example of its application to Doyayo (Atlantic-Congo, Africa)]{\label{fig:centroidwithballtree}Result of the 30-nearest-neighbour search using the balltree method, with an example of its application to Doyayo (Atlantic-Congo, Africa). The red marks are the centroid of the respective GMM clusters (as represented in Figure \ref{gmm6}), while the points in which they are embedded are their 30 nearest neighbours. The contour lines in green and red correspond to the Kriging areas for Doyayo \textit{\textipa{gO}} and \textit{\textipa{yO}} at 29\% probability.}
    \end{figure}

        \item[3.] For each language, check which Kriging area, if any, contains each of the groups in Figure \ref{fig:centroidwithballtree} and, for each language, create a dictionary to take note of the mapping between groups and Kriging areas.\footnote{To obtain this information, we used Kriging areas at 29\% of probability.} For example, the group of points corresponding to GMM cluster 3 in Figure \ref{fig:centroidwithballtree} are contained within the Kriging area for Doyayo \textit{g\textipa{O}}, those corresponding to GMM cluster 2 and 4 are contained within the Kriging area for \textit{y\textipa{O}}. The resulting dictionary for Doyayo, then, is \texttt{\{group-3: g\textipa{O}, group-2: y\textipa{O}, group-4: y\textipa{O}\}}, meaning that all the points of each group are contained within \textit{one} Kriging area only. This is the simpler scenario. 
    
    The more complex scenario is one in which more than one Kriging area includes points from the same group.  
    
    For example, the dictionary for Patep (Austronesian, Papunesia) is \texttt{\{group-3: [obêc, buc], group-2: buc, group-4: NULL\}}, meaning that the Kriging area for \textit{buc} contains points from groups 2 and group 3, but points from group 3 are also found in the Kriging area for \textit{obêc}. In such cases, we apply the following heuristics to infer whether more than one Kriging area should be considered meaningful in that group for the purpose of looking at patterns of coexpression.

    \begin{itemize}
        \item[a.] If one of the two Kriging areas is unique to a given group (e.g. \textit{obêc} in the Patep example), while the other is not (e.g. \textit{buc} in the Patep example), consider the former as meaningful, regardless of how many points from that group it contains. Instead, consider the latter as meaningful only after running a test of proportion with the competing Kriging area. If it contains significantly more points than the competing area, or if the difference in proportion is not statistically significant, then both Kriging areas are kept in the dictionary. To determine this, we use Fisher's exact test with $ \alpha=0.01 $. In the Patep example above, \textit{obêc} is considered a meaningful Kriging area for group 3 because it is only found there. On the other hand, to decide whether to also keep the Kriging area for \textit{buc}, we run a Fisher's test, which indicates that the difference in proportion is not significant (26 out of 30 points are found in the Kriging area for \textit{obêc}, 21 out of 30 in the one for \textit{buc}; $p=0.32$), so both Kriging areas are considered meaningful lexicalization for group 3.
        \item[b.] If neither of any two competing Kriging areas is unique to a particular group, then a Fisher's test is used to establish which one to consider meaningful. For example, the dictionary for Yucatec Maya (Mayan, North America) is \texttt{\{group-3: ken, group-2: [ken, ka], group-4: ka\}}. A Fisher's test indicates that the Kriging area for \textit{ken} contains significantly more points from group 2 than the Kriging area for ka ($p<0.01$), so the dictionary is modified to \texttt{\{group-3: ken, group-2: ken, group-4: ka\}}.
        \item[c.] Give lexical items a greater weight than NULLs. Only consider a NULL Kriging area as meaningful if it is the only one containing a particular group. For example, the dictionary for Manam (Austronesian, Papunesia) is \texttt{\{group-3: [bong, NULL], group-2: bong, group-4: [bong, NULL]\}}, which, for the purpose of looking at lexicalization patterns is then modified to \texttt{\{group-3: bong, group-2: bong, group-4: bong\}}. On the other hand, the dictionary for Hills Karbi (Sino-Tibetan, Eurasia) is \texttt{\{group-3: ahut, group-2: ahut, group-4: NULL\}}, in which case the only Kriging area containing points from group 4 is a NULL area. 
    \end{itemize}
    \item[5.] Assign patterns of lexicalization based on the Kriging areas considered meaningful for each group.
\end{itemize}

As shown in Figure \ref{fig:centroidwithballtree}, for example, the points in group 3 all fall within the Kriging area for Doyayo \textit{\textipa{gO}}, while those in both group 2 and 4 are all contained within the Kriging area for \textit{\textipa{yO}}. On the basis of this, we can assign languages behaving like Doyayo to a pattern in which the top left area (cluster 3) is the domain of one word, whereas the mid and bottom left areas are colexicalized by a different word. This---which we will call `pattern C' in Section \ref{sec:analysis}---is one of five basic patterns which can be observed on the basis of the three groups of core points represented in Figure \ref{fig:centroidwithballtree} (one for each possible logical combinations between the groups). An overview of the patterns will be given in Section \ref{sec:classif}.

\section{Analysis}\label{sec:analysis}
\begin{figure}[!h]
\begin{subfigure}{0.50\textwidth}
\includegraphics[width=0.9\linewidth]{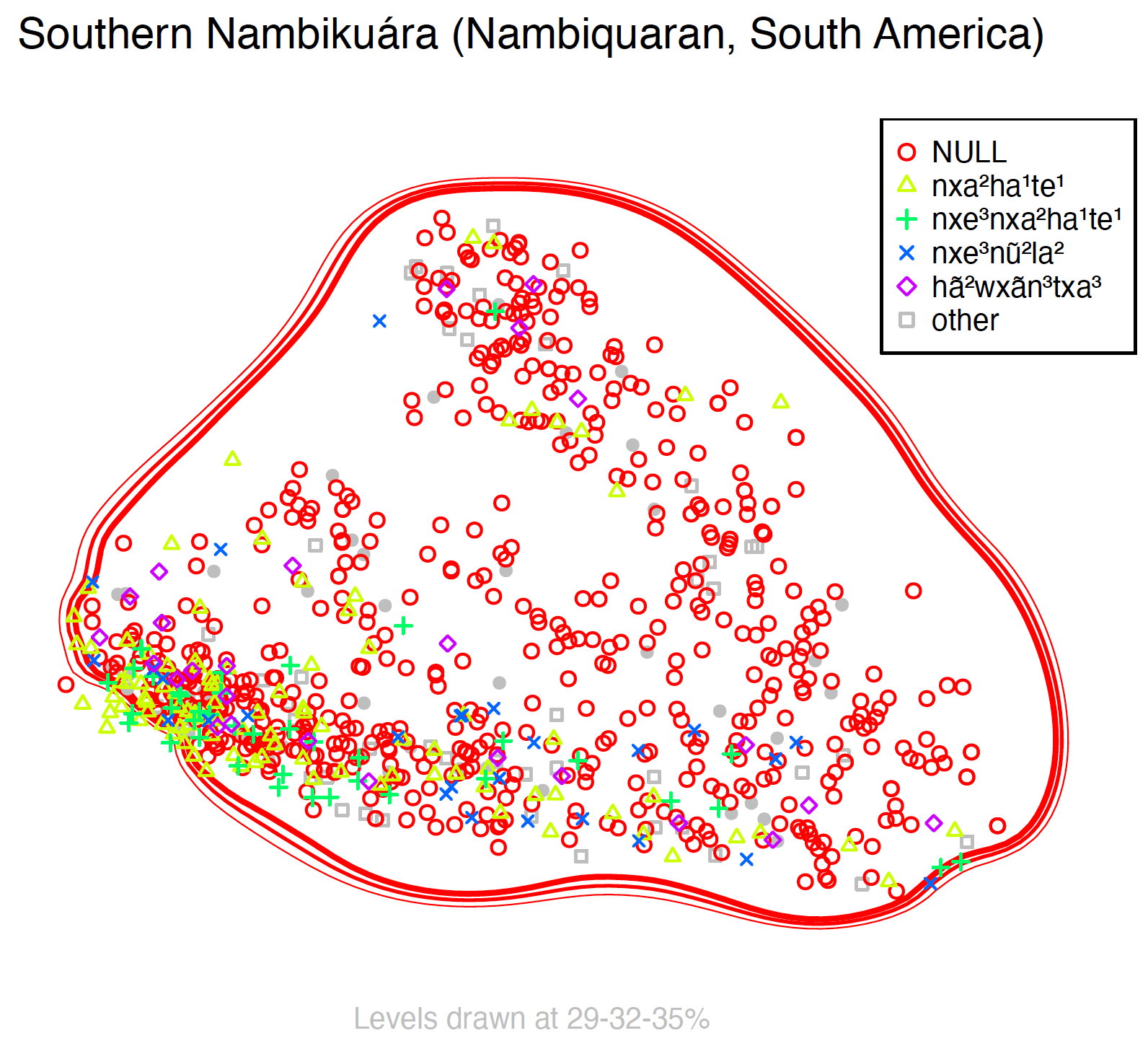} 
\caption[Kriging maps for Southern Nambikuára (Nambiquaran, South America)]{}
\label{nab-when}
\end{subfigure}
\begin{subfigure}{0.50\textwidth}
\includegraphics[width=0.9\linewidth]{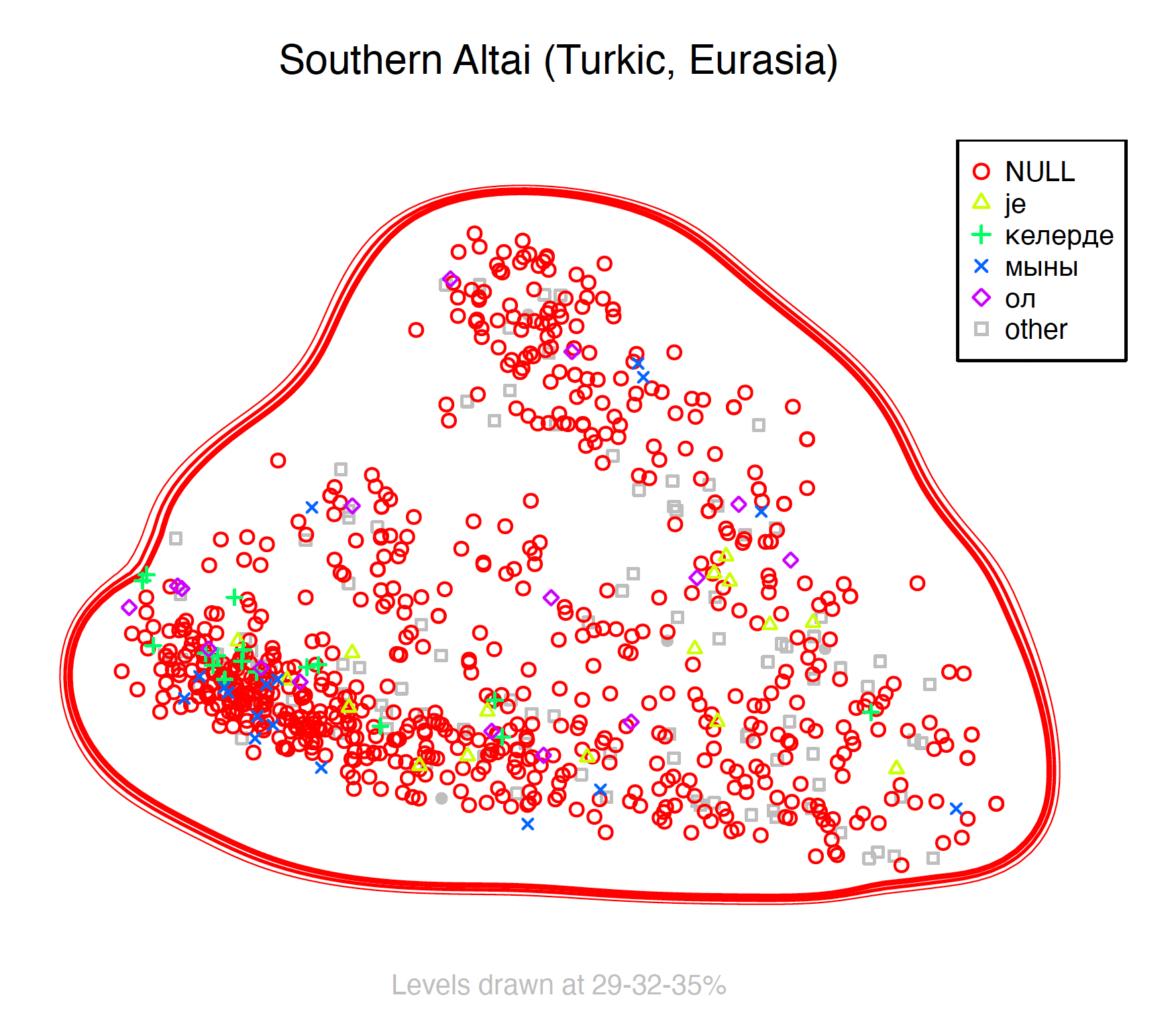}
\caption[Kriging maps for Southern Altai (Turkic, Eurasia)]{}
\label{alt-when}
\end{subfigure}
\caption[]{Kriging maps for Southern Nambikuára (Nambiquaran, South America) and Southern Altai (Turkic, Eurasia)}
\end{figure}

The very first, perhaps obvious, observation to make is that no target language has a one-to-one match with English. All target languages show at least some variation, but a common feature in the maps for all the languages is that if the interpolation method assigned all observations to one common area, then that area is a `null' one, rather than a specific lexical item (i.e. a subordinating conjunction). Figures \ref{nab-when} and \ref{alt-when} are two examples of null-only maps from Southern Nambikuára (Nambiquaran) and Southern Altai (Turkic).

As already mentioned in the Method section, \textit{null} refers to either no parallel in the target language (e.g. the token \textit{when} has no parallel in a typical participle construction), or items that have been reassigned a \textit{null} alignment because their low frequency in the relevant languages likely suggested a misalignment (e.g. in a participle construction with an auxiliary, such as \textit{\textbf{having}} \textit{seen}, the auxiliary may have been considered by the model as a \textit{when}-counterpart in that language). \textit{Null} is also always one of any number of areas detected by Kriging for any language. That is, if Kriging detected two areas, for example, one of these is a \textit{null} area.\\
\indent Throughout the analysis, null areas will not simply be regarded as the \textit{lack} of a construction, but as corresponding to different means of juxtaposing clauses to achieve what the English version we are using achieved with a \textit{when}-clause. This approach is in line with previous typological literature \citep{cristofarowals}, which defines \textsc{when}-clauses in functional terms, classifying as such not only those introduced by specific temporal conjunctions (e.g., English \textit{when X did Y} or \textit{when doing Y}), but also clauses that are simply juxtaposed and whose function must be contextually inferred, as in \ref{canelaex}.  In some other cases, languages may use specific verb forms to mark adverbial subordination, without, however, specifying their semantic relation to the main clause. This is the case of cross-linguistically well-attested converbs and predicative participles \citep{converbs}, as in examples \ref{kumykchain}-\ref{grcex} from our data.\footnote{All glossing my own unless otherwise stated.}

\begin{example}
   \label{canelaex} Canela-Krahô (Macro-Gê; \citealt{cristofarowals})
    \gll \textit{\textbf{pê}}	\textit{\textbf{wa}}	\textit{\textbf{i-pỳm}},	\textit{pê}	\textit{inxê}	\textit{ty}
    \textsc{pst}	\textsc{1}	\textsc{1}-fall	\textsc{pst}	mother	die
    \glt `My mother died when I was born'.
    \glend
\end{example}

\begin{example}
    Kumyk (Turkic)
    \gll Etti èkmekni de, balyk\textsuperscript{$\prime$$\prime$}ny da \textbf{al-yp}, bir \textbf{bašl-ap} š\foreignlanguage{russian}{ъ}k\foreignlanguage{russian}{ъ}r dua ète, song \textbf{syndyr-yp} âk\textsuperscript{$\prime$$\prime$}čylaryna bere, olar da halk\textsuperscript{$\prime$$\prime$}g\textsuperscript{$\prime$$\prime$}a pajlajlar.
    seven bread \textsc{ptc} fish \textsc{ptc} take-\textsc{cvb} first start-\textsc{cvb} thanks prayer do.\textsc{inf} then break-\textsc{cvb} disciples.\textsc{dat} give.\textsc{inf} they \textsc{ptc} people.\textsc{dat} distribute.\textsc{pst}
    \glt `Then he took the seven loaves and the fish, and when he had given thanks, he broke them and gave them to the disciples, and they in turn to the people' (Matthew 15:36)
    \glend
    \label{kumykchain}
\end{example}
% % Етти экмекни де , балык\foreignlanguage{russian}{ъ}ны да алып , бир башлап шюкюр дуа эте , сонг сындырып як\foreignlanguage{russian}{ъ}чыларына бере , олар да халк\foreignlanguage{russian}{ъ}г\foreignlanguage{russian}{ъ}а пайлайлар 

\begin{example}
    \label{ukrex} Ukrainian (Indo-European) 
    \gll \textit{A} \textit{\textbf{pobačivši}}, \textit{rozpovily} \textit{pro} \textit{vse} \textit{te}, \textit{ščo} \textit{pro} \textit{Cju} \textit{Dytynu} \textit{bulo} \textit{jim} \textit{zviščeno}.
    and see.\textsc{pfv}.\textsc{cvb.pst} recount\textsc{.pfv}.\textsc{pst.pl} about all that which about this child was \textsc{3.dat} reported 
    \glt ‘And when they saw it, they made known the saying that had been told them concerning this child’ (Luke 2:17).
    \glend 
\end{example}

\begin{example}
  \label{korex} Korean (Language isolate)
    \gll \textit{geudeureun} \textit{i} \textit{malsseumeul} \textit{\textbf{deut-ko}} \textit{keuge} \textit{nollamyeo} \textit{geubunegeseo} \textit{tteonagassda}
    they this word hear-\textsc{cvb} greatly surprised from-him went-away
    \glt ‘When they heard it, they marveled. And they left him and went away.’ (Matthew 22:22).
    \glend 
\end{example}

\begin{example}
\label{itaex} Italian (Indo-European)
\begin{itemize}
   \item[a.]
   \gll \textit{ma} \textit{\textbf{essendosi}} \textit{i} \textit{discepoli} \textit{\textbf{radunati}} \textit{intorno} \textit{a} \textit{lui} \textit{egli} \textit{si} \textit{alzò} \textit{ed} \textit{entrò} \textit{in} \textit{città} 
   but be\textsc{.ger.refl} the disciples gather\textsc{.pst.ptcp.pl} around to him he refl rise\textsc{.3sg.pst.pfv} in city
   \glt `But when the disciples gathered about him, he rose up and entered the city' (Acts 14:20).
   \glend
   \item[b.] \label{itaexb}
   \gll \textit{e} \textit{il} \textit{centurione} \textit{\textbf{avendo}} \textit{\textbf{sentito}} \textit{parlare} \textit{di} \textit{Gesú} \textit{gli} \textit{mandò} \textit{alcuni} \textit{anziani} \textit{dei} \textit{giudei}
   and the centurion have\textsc{.ger} hear\textsc{.pst.ptcp} speak about Jesus \textsc{3.dat} send\textsc{.3sg.pst.pfv} some elders of-the Jews
   \glt `When the centurion heard about Jesus, he sent to him elders of the Jews, asking him to come and heal his servant' (Luke 7.3).
   \glend
   \label{itaexa}
\end{itemize}
\end{example}

\begin{example}
\label{grcex} Ancient Greek (Indo-European)
\gll {\it \textbf{eparantes}} {\it de} {\it tous} {\it ophthalmous} {\it autōn} {\it oudena} {\it eidon} {\it ei} {\it mē} {\it ton} {\it Iēsoun} {\it monon}
{lift-up.{\sc pl.pfv.pst.ptcp.m.nom}} {\sc ptc} {the} {eyes} {their} {nobody} {see.{\sc 3pl.pfv.pst}} {if} {not} {the} {Jesus} {alone}
\glt `And when they lifted up their eyes, they saw no one but Jesus only.' (Matthew 17.8)
\glend
\end{example}

Examples such as \ref{kumykchain}-\ref{grcex} above are captured as null alignments by the alignment models. This is because the verb form itself, be it a converb, a participle, or a finite (main) verb, is aligned to the verb form in the English \textit{when}-clause. Since we chose a one-to-one alignment between source and target language, the connector \textit{when} has no parallel in these examples and is therefore aligned to a `null' token. Despite some inevitable noise in the alignment, due, for instance, to particularly frequent juxtaposed verb forms that may be interpreted as a recurring parallel to the token \textit{when} itself (e.g. the auxiliary in \textit{having done}), the overall reliability of the alignments is clear from the clear-cut pattern across the world's languages regarding where in the semantic map of \textsc{when} null constructions tend to cluster, as we will see in Section \ref{nullovertsec}. Null alignments thus have significance in themselves--recall, in fact, that the abundance of null alignments was the very reason for choosing the specific model we are using, as opposed to others which, instead, aligned \textit{when} to some other token in the absence of a direct parallel. Since our focus is on the competition between \textit{jegda}-clauses and participle constructions, understanding where null constructions cluster and what kinds of meanings they encode compared to lexified connectors (such as \textit{jegda}) is crucial.\\
\indent In the overview that follows, it is important to keep in mind what the Kriging maps and the GMM clustering are meant to capture, which are, in a way, complementary perspectives on the data. Kriging maps, such as those in Figures \ref{dow} and \ref{alt-when}-\ref{nab-when}, are used to identify the means dedicated to expressing a particular function or meaning of \textsc{when} in a specific language. Conversely, the aim of GMM clustering is to abstract from language-specific observations in order to understand which of those functions or meanings (for which individual languages may have dedicated means) tend to have dedicated means cross-linguistically. \\
\indent In other words, our application of Kriging approximates the search for potential \textit{grammatical morphemes} (\textsc{grams}), while the GMM method is used to look for potential \textsc{gram types}. A \textsc{gram}, according to \citet{bybeedahl}, is a linguistic item---a bound morpheme, a lexical item or a complex construction---with a specific function or meaning. \posscitet{bybeedahl} definition thus encompasses both specific lexical items aligned to English \textit{when} (e.g. a subordinator) \textit{and} null constructions. The concept of \textit{gram} is more abstract than that of \textit{morph} and is in many ways analogous to the concept of \textit{construction} in Construction Grammar. It is language-specific and does not necessarily correspond to a crosslinguistic \textsc{gram type}, which \citet{dahl} defines as `a cluster of language-specific grams whose closeness in meanings and functions is reflected in similar distributions in a parallel corpus'. Together, the gram types make up the semantics atoms in the grammatical space that English \textit{when} covers. \\
\indent The morphosyntactic diversity exemplified by (\ref{kumykchain})-(\ref{grcex}) echoes \citeauthor{dahl}'s (\citeyear{dahl}) remark that grams differ in how \emph{transparent} they are, namely in how constant and isolable their form is, which has bearing on how easily they can be automatically identified via methods such as ours. English \textit{when} is maximally transparent, as it is a single word with a constant form and can therefore be automatically identified with little obstacle. 
The Italian past participle clauses in (\ref{itaex}), on the other hand, are less transparent, as the auxiliary, though marked as a \textsc{gerund} and isolable from the participle form, can be either \textit{essendo} (lit. `being') or \textit{avendo} (lit. `having'), depending on the chosen participle verb. The Kumyk converb markers -\textit{ap}/-\textit{yp} in (\ref{kumykchain}) are much less transparent since they are not easily isolable and are only two of several allomorphs. Cases like the Ancient Greek predicative participle in (\ref{grcex}) are maximally opaque since their form depends on the constituent in the matrix clause with which they agree. Moreover, while participles in their predicative function are similar to converbs, they can often also occur in other, e.g. attributive contexts.

\subsection{Overtly-subordinated versus `null' constructions}\label{nullovertsec}
\begin{figure}[!h]
    \centering   
    \includegraphics[width=120mm]{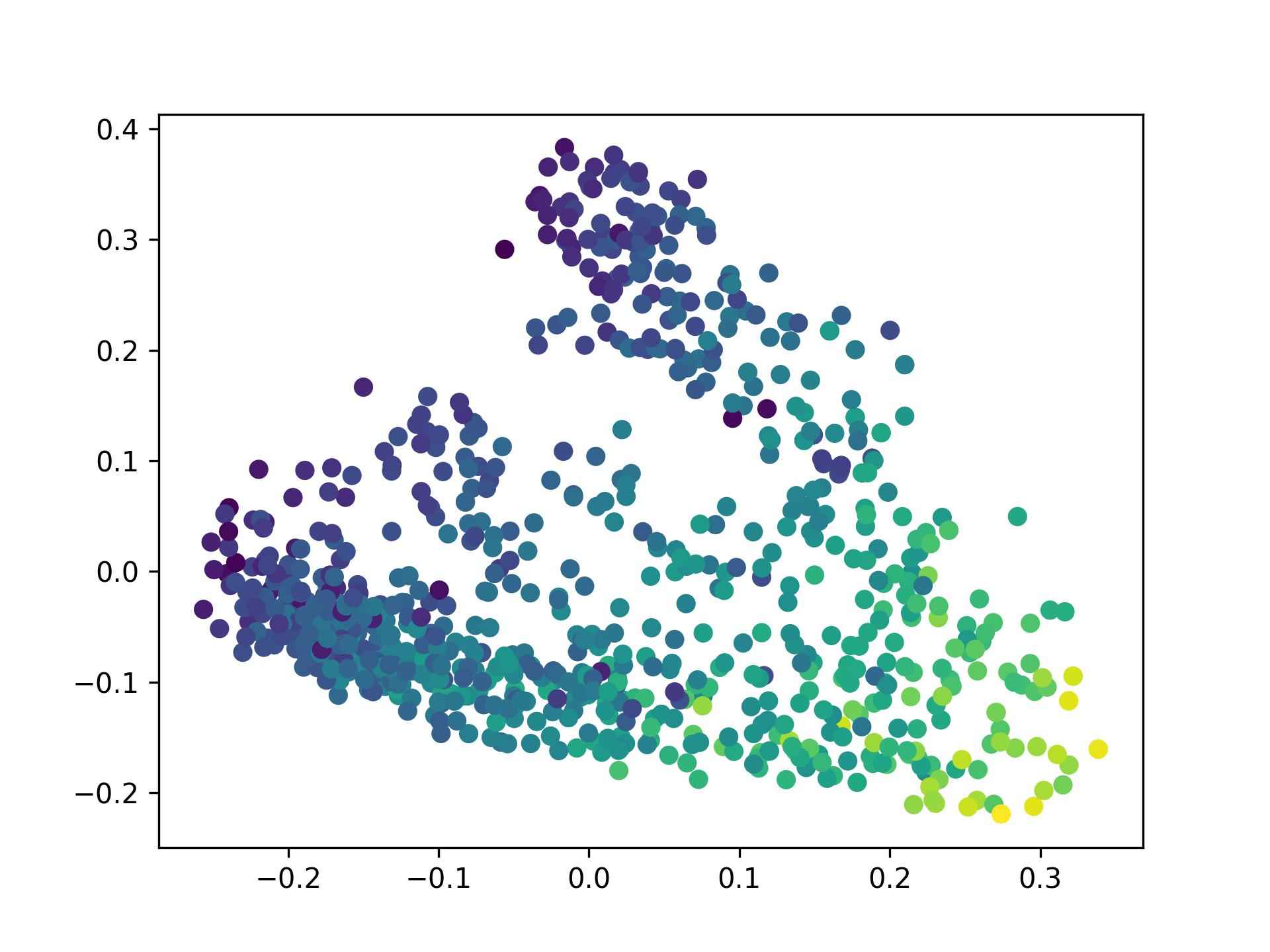}
    \caption[Heatmap of languages expressing each observation in the semantic map with a null construction]{Heatmap showing the concentration of languages expressing a data point with a null construction as opposed to lexified means. The warmer the colour of a point, the more languages use a null construction.}
    \label{fig:nulls}
\end{figure}

As mentioned above, null constructions tend to cluster in a specific area of the semantic maps. We can capture this pattern across languages by counting how many languages in the dataset express a particular observation in the semantic map with a null construction as opposed to some other means. The heatmap in Figure~\ref{fig:nulls} shows the concentration of languages expressing data points in the map with a null construction: the warmer the colour of a point, the more languages use a null construction, and the colder the colour, the more languages use a lexified means.\footnote{An interactive version of the heatmap is available in the Observable notebook at the following address: \url{https://observablehq.com/@npedrazzini/3d-mds-when-map}.}

The closer we get to the lower right corner of the map, the more likely it is that a language will use a null construction. 869 languages, for example, use a null construction for the verse in \ref{869nulls} from the very bottom right corner of the map. 

\begin{example}
    And he took bread, and \textbf{when he had given thanks}, he broke it and gave it to them \hfill (Luke 22:19) \label{869nulls}
\end{example}

In fact, several other English translations also use a null construction for the same verse: (e.g. King James Version, New King James Version and  New International Version: \textit{and he took bread, \textbf{gave thanks} and broke it, and gave it to them, saying}). Note how Old Church Slavonic (\ref{ocsgrcindrheme}a) and Ancient Greek (\ref{ocsgrcindrheme}b) use an \textsc{independent rheme} participle instead.

\begin{example}
\begin{itemize}
    \item[a.]
    \gll i priim\foreignlanguage{russian}{ъ} chlěb\foreignlanguage{russian}{ъ} \textbf{chvalǫ} \textbf{v\foreignlanguage{russian}{ъ}zdav\foreignlanguage{russian}{ъ}} prělomi. i dast\foreignlanguage{russian}{ъ} im\foreignlanguage{russian}{ъ} glę.
    {and} {take.{\sc ptcp.pfv.m.nom.sg}} {bread.{\sc acc}} {blessing.{\sc acc}} {give.{\sc ptcp.pfv.m.nom.sg}} {break.{\sc aor.3.sg}} {and} {give.{\sc aor.3.sg}} {\sc 3.pl.m.dat} {say.{\sc ptcp.ipfv.m.nom.sg}}
    \glt
    \glend
    \item[b.]
    \gll {Kai} {labōn} {arton} {\textbf{eukharistēsas}} {eklasen} {kai} {edōken} {autois} {legōn}
    {and} {take.{\sc ptcp.pfv.m.nom.sg}} {bread.{\sc acc}} {thank.{\sc ptcp.pfv.m.nom.sg}} {break.{\sc aor.3.sg}} {and} {give.{\sc aor.3.sg}} {\sc 3.pl.m.dat} {say.{\sc ptcp.ipfv.m.nom.sg}}
    \glt `Then he took the bread, said the blessing, broke it, and gave it to them, saying' (Luke 22:19) %21628
    \glend
\end{itemize}
\label{ocsgrcindrheme}
\end{example}

At the other end of the spectrum, from the top left of the semantic map, we find examples like \ref{120nulls}, which is expressed with a null construction by only 120 languages.

\begin{example}
\textbf{When all things are subjected to him}, then the Son himself will also be subjected to him who put all things in subjection under him \hfill (1 Corinthians 15:28) \label{120nulls}
\end{example} 

A \textit{hótan}-clause is also found in the Ancient Greek version (the Old Church Slavonic version does not have a translation for Bible verses outside of the Gospels), as (\ref{grcotan}) shows.

\begin{example}
    \gll {\textbf{hotan}} {\textbf{de}} {\textbf{hupotagēi}} {\textbf{autōi}} {\textbf{ta}} {\textbf{panta}} {tote} {kai} {autos} {ho} {huios} {hupotagēsetai} {tōi} {hupotaksanti} {autōi} {ta} {panta} {hina} {ēi} {ho} {theos} {ta} {panta} {en} {pasin}
    {when} {\sc ptc} {subordinate.{\sc sbjv.pas.pfv.3.sg}} {\sc 3.sg.m.dat} {the.{\sc n.nom.pl}} {all things.{\sc n.nom.pl}} {then} {even} {self.{\sc m.nom.sg}} {the.{\sc m.nom.sg}} {son.{\sc m.nom.sg}} {subordinate.{\sc fut.pas.3.sg}} {the.{\sc dat.sg}} {subordinate.{\sc ptcp.pfv.pst.m.dat.sg}} {\sc 3.sg.m.dat} {the.{\sc n.acc.pl}} {all things.{\sc n.acc.pl}} {that} {be.{\sc sbjv.prs.3.sg}} {the.{\sc m.nom.sg}} {god.{\sc nom}} {the.{\sc n.nom.sg}} {all things.{\sc n.nom.pl}} {in} {all.{\sc n.dat.pl}}
    \glt
    \glend
\label{grcotan}
\end{example}

\begin{figure}[!h]
\begin{subfigure}{0.50\textwidth}
\includegraphics[width=0.9\linewidth]{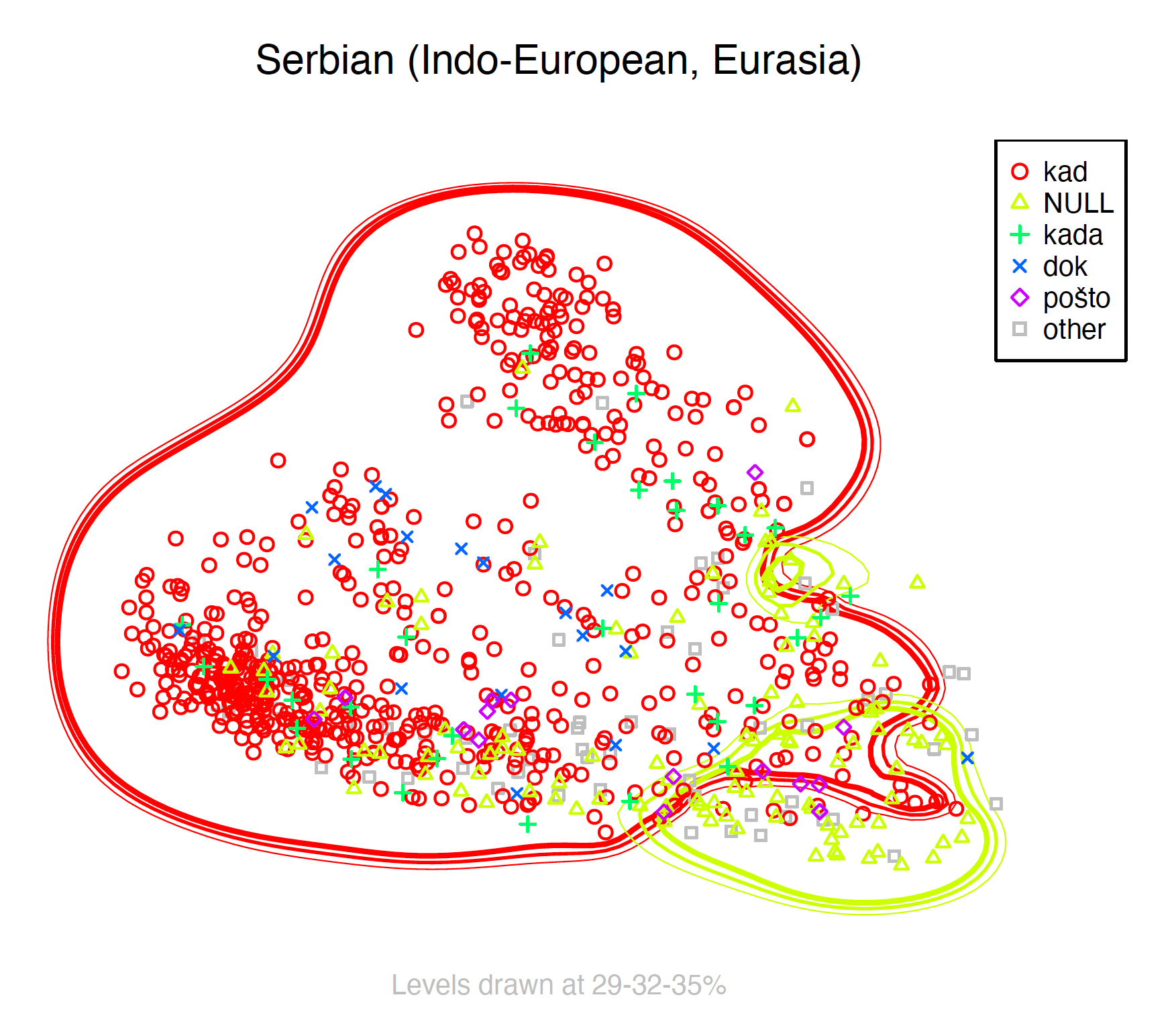} 
\caption[Kriging map for Serbian (Indo-European, Eurasia)]{}
\label{srp-when}
\end{subfigure}
\begin{subfigure}{0.50\textwidth}
\includegraphics[width=0.9\linewidth]{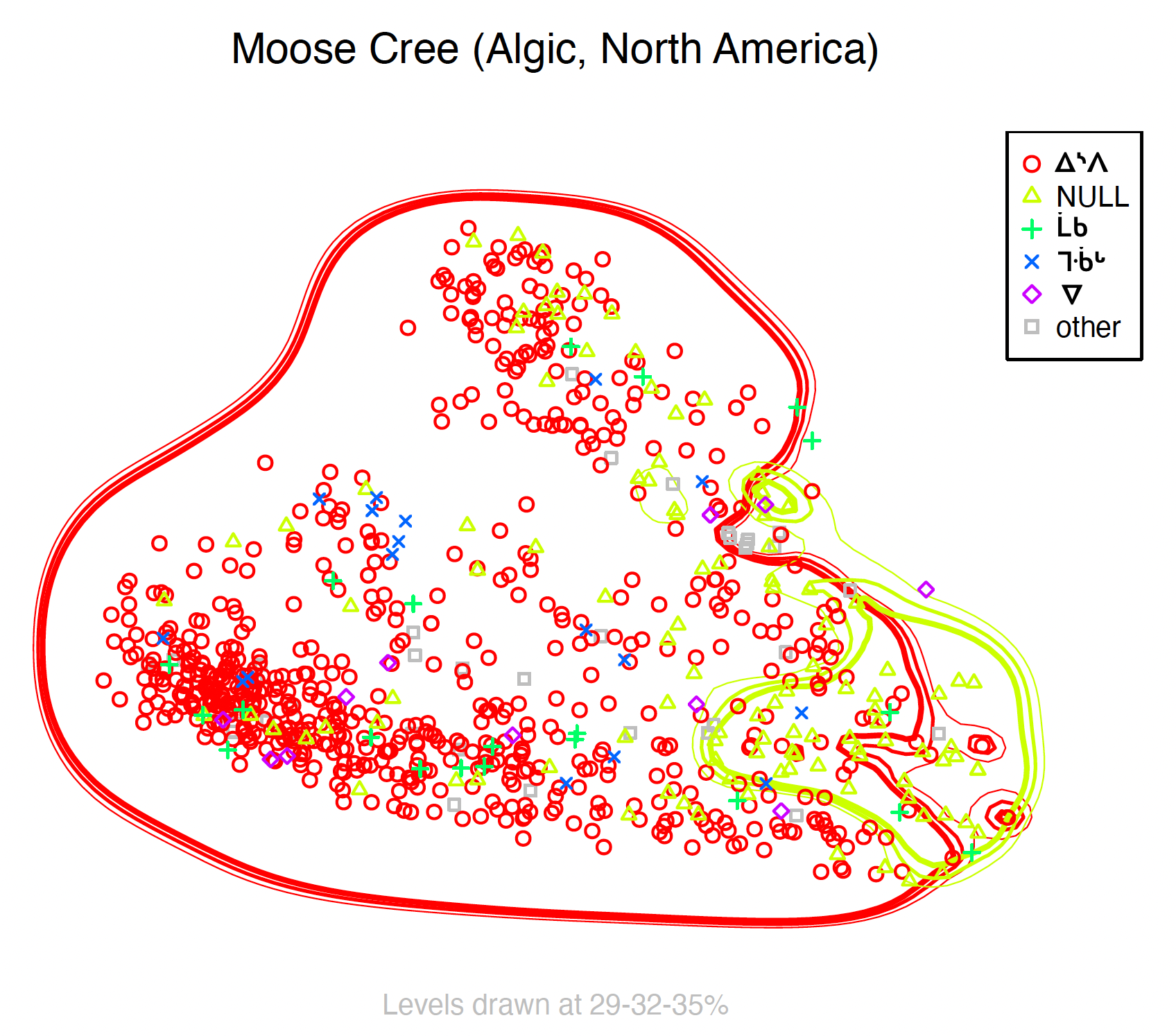}
\caption[Kriging map for Moose Cree (Algic, North America)]{}
\label{crm-when}
\end{subfigure}
\caption[]{Kriging maps for Serbian (Indo-European, Eurasia) and Moose Cree (Algic, North America)}
\end{figure}

It seems relatively common for languages to show an overall split between null- and non-null situations roughly to the right and to the left of the map, respectively, as is the case, for example, in Serbian (Figure \ref{srp-when}) and Moose Cree (Figure \ref{crm-when}), among several others.

\begin{figure}[!h]
\begin{subfigure}{0.50\textwidth}
\includegraphics[width=0.9\linewidth]{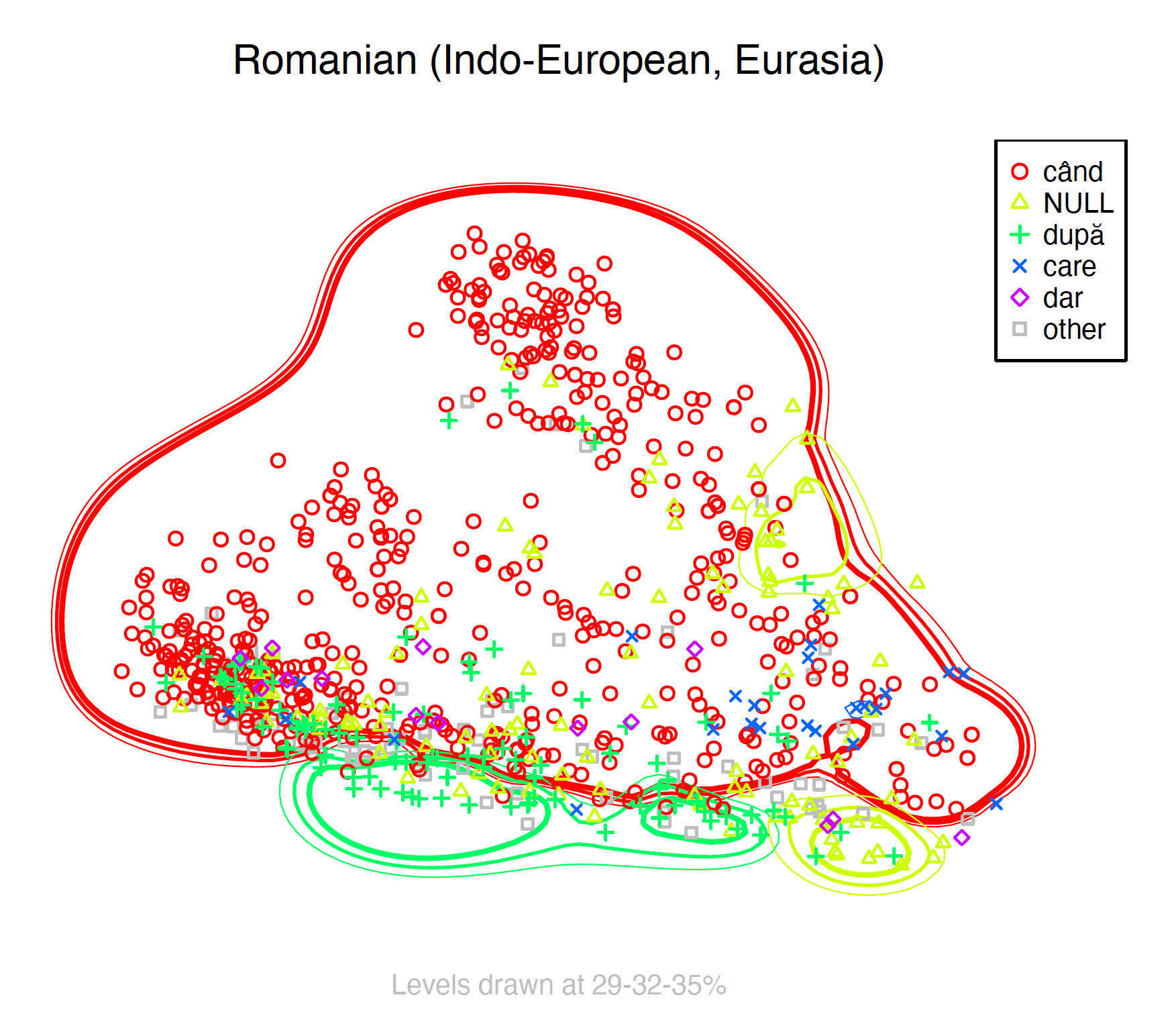} 
\caption[Kriging map for Romanian (Indo-European, Eurasia)]{}
\label{ron-when}
\end{subfigure}
\begin{subfigure}{0.50\textwidth}
\includegraphics[width=0.9\linewidth]{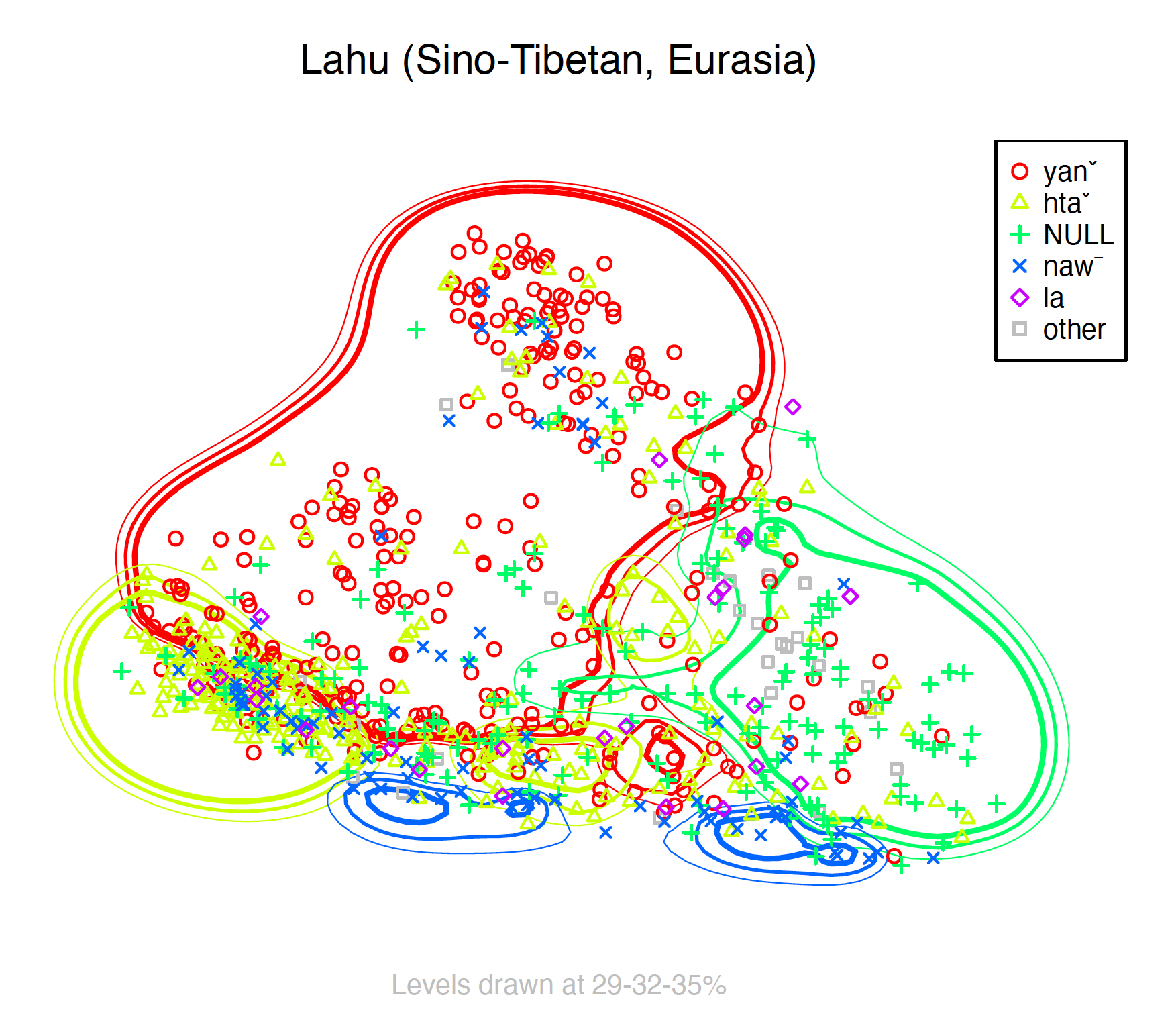}
\caption[Kriging map for Lahu (Sino-Tibetan, Eurasia)]{}
\label{lhu-when}
\end{subfigure}
\caption[]{Kriging maps for Romanian (Indo-European, Eurasia) and Lahu (Sino-Tibetan, Eurasia)}
\end{figure}

\textit{After}-clauses, if present in a language, appear to generally cluster at the bottom of the map, where competition between different linguistic means also seems to be more intense than in other areas. Romance languages are an example of maps with a dedicated \textit{after}-area detected by Kriging, as in Romanian (Figure \ref{ron-when}), but several other languages show a similar pattern, like Lahu (Sino-Tibetan; Figure \ref{lhu-when}).

\begin{figure}[!h]
\centering
\includegraphics[width=0.6\textwidth]{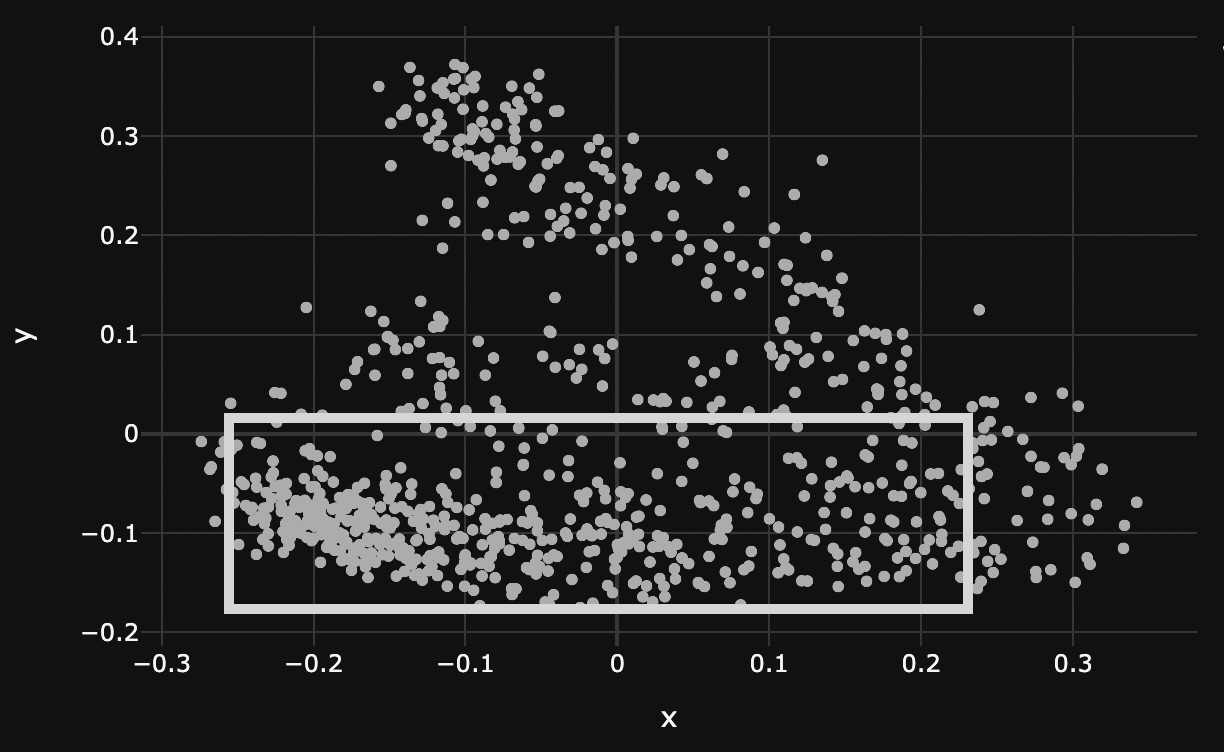}
\caption[High-competition area in the semantic map]{\label{highcomp}Approximate area in the maps where competition appears to be particularly high across all languages}
\end{figure}

In Figures \ref{ron-when} and \ref{ron-when}, respectively, Romanian \textit{după} `after' and Lahu \textit{lhu$\bar{ }$} (part of the phrase \textit{hk'aw naw$\bar{ }$} `after'; \citealt{lahudict}) occupy similar positions in the map, and in both cases there is a clear overlap between different linguistic means on or just around the area of \textit{after}-clauses. It is, in fact, quite clear that the competition is particularly high across all languages roughly in the whole area highlighted in Figure \ref{highcomp}.

\begin{figure}[!h]
\begin{subfigure}{0.50\textwidth}
\includegraphics[width=0.9\linewidth]{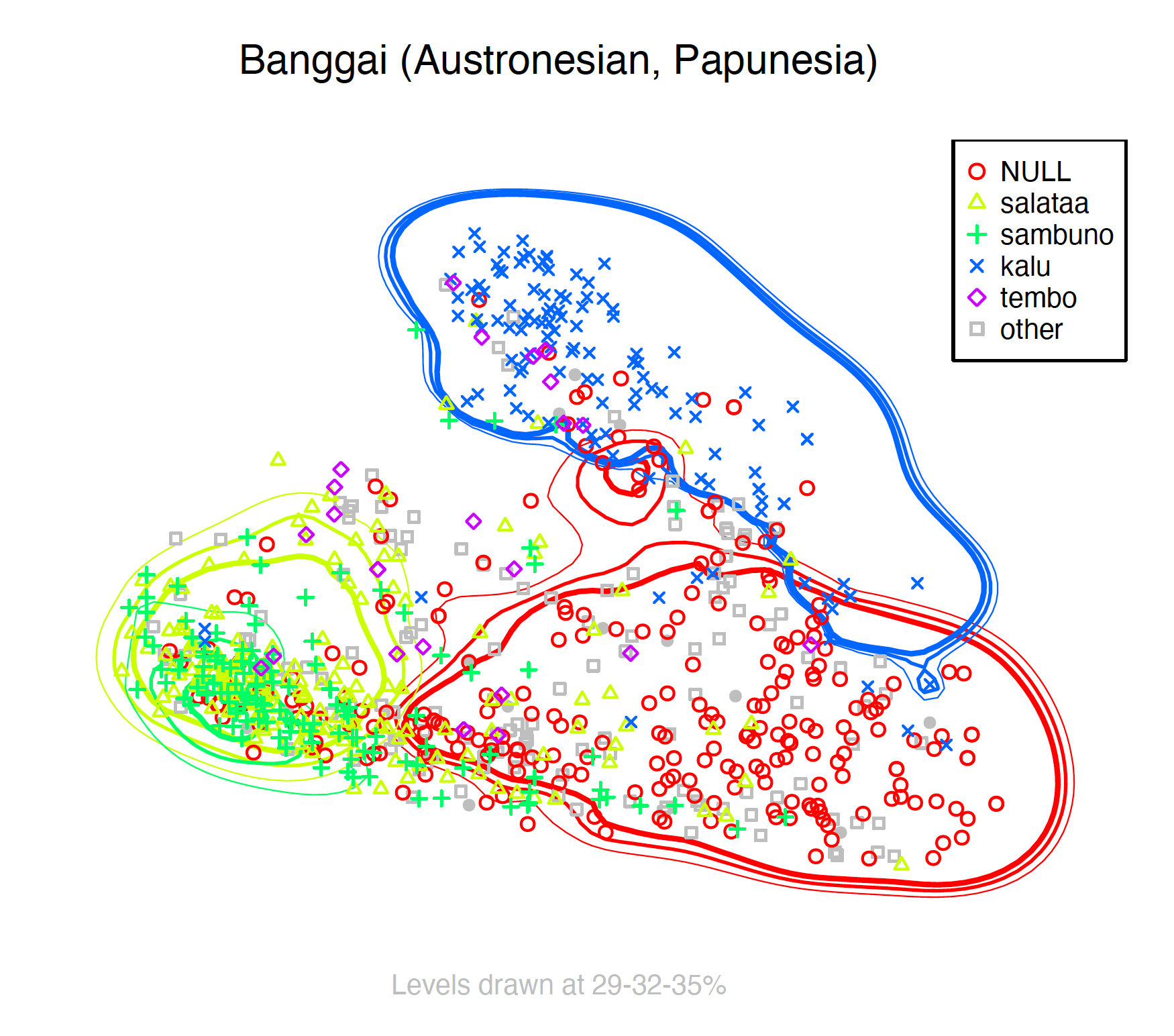} 
\caption[Kriging map for Banggai (Austronesian, Papunesia)]{}
\label{bgz-when}
\end{subfigure}
\begin{subfigure}{0.50\textwidth}
\includegraphics[width=0.9\linewidth]{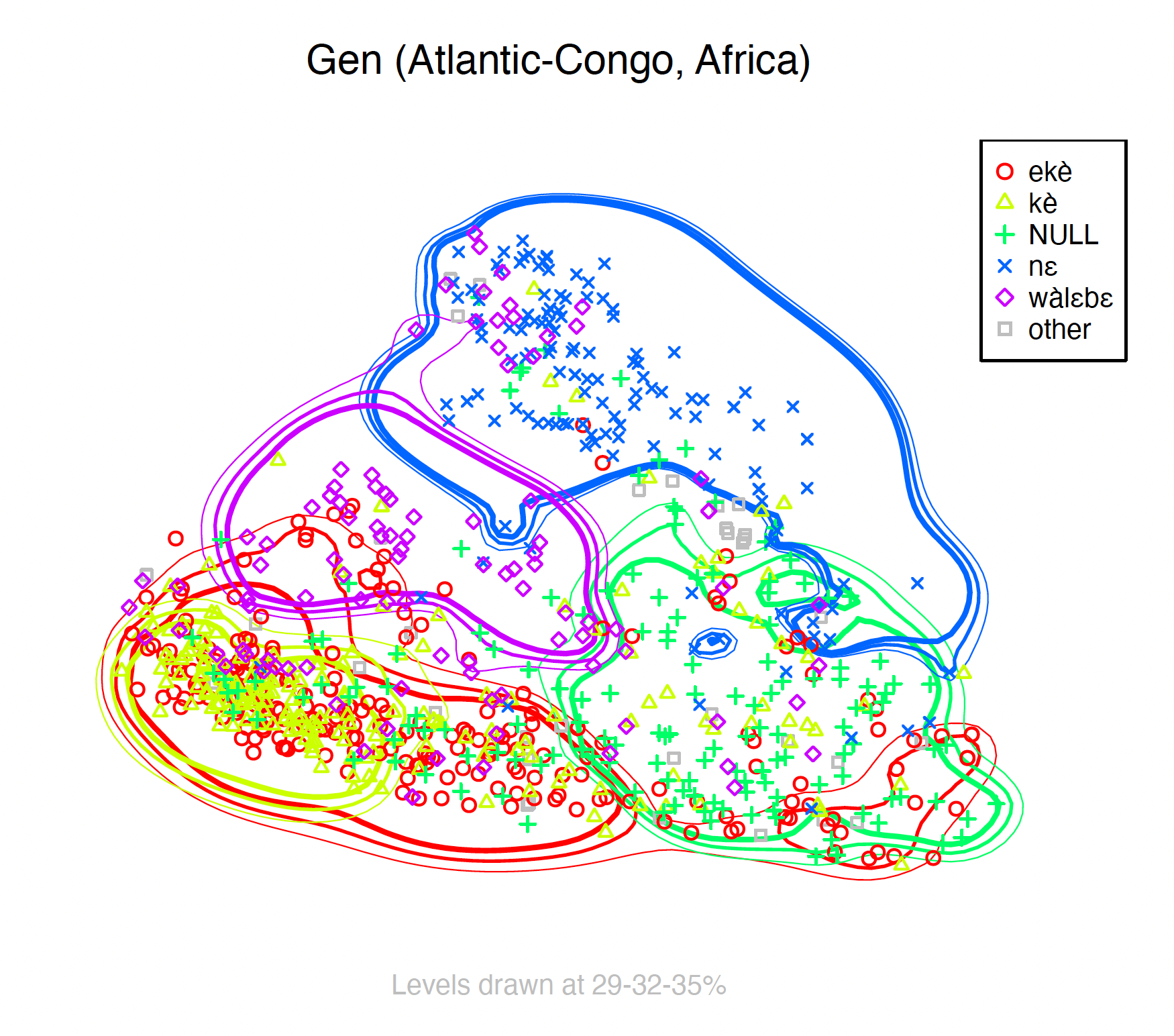}
\caption[Kriging map for Gen (Atlantic-Congo, Africa)]{}
\label{gej-when}
\end{subfigure}
\caption[]{Kriging maps for Banggai (Austronesian, Papunesia) and Gen (Atlantic-Congo, Africa)}
\end{figure}

The maps for several of the target languages even show two distinct overlapping areas (or areas within areas) in the bottom left of the map, as in Banggai (Figure \ref{bgz-when}) and Gen (Figure \ref{gej-when}).

\begin{figure}[!h]
\begin{subfigure}{0.50\textwidth}
\includegraphics[width=0.9\linewidth]{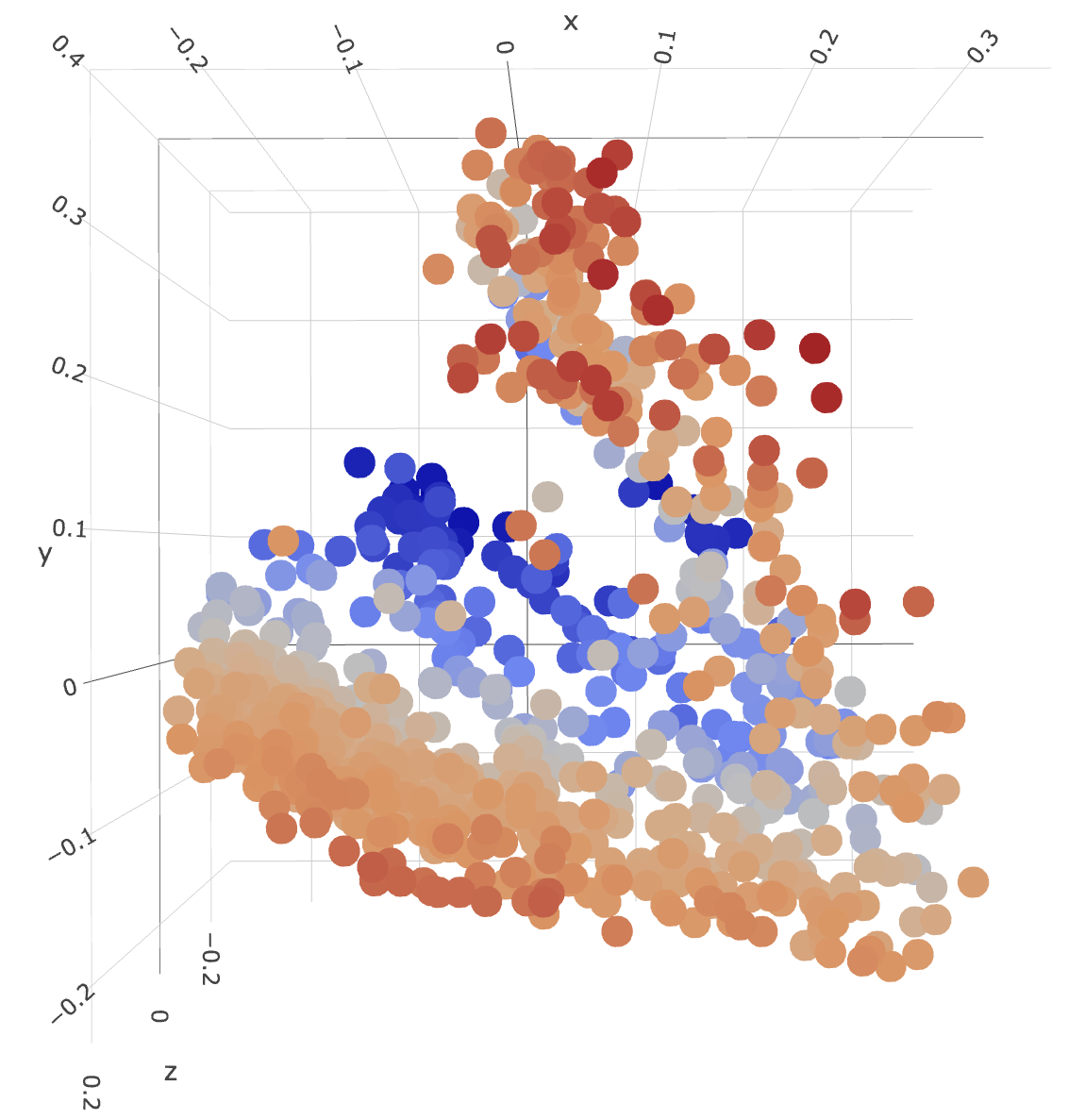} 
\caption[]{}
\label{3d-xy}
\end{subfigure}
\begin{subfigure}{0.50\textwidth}
\includegraphics[width=0.9\linewidth]{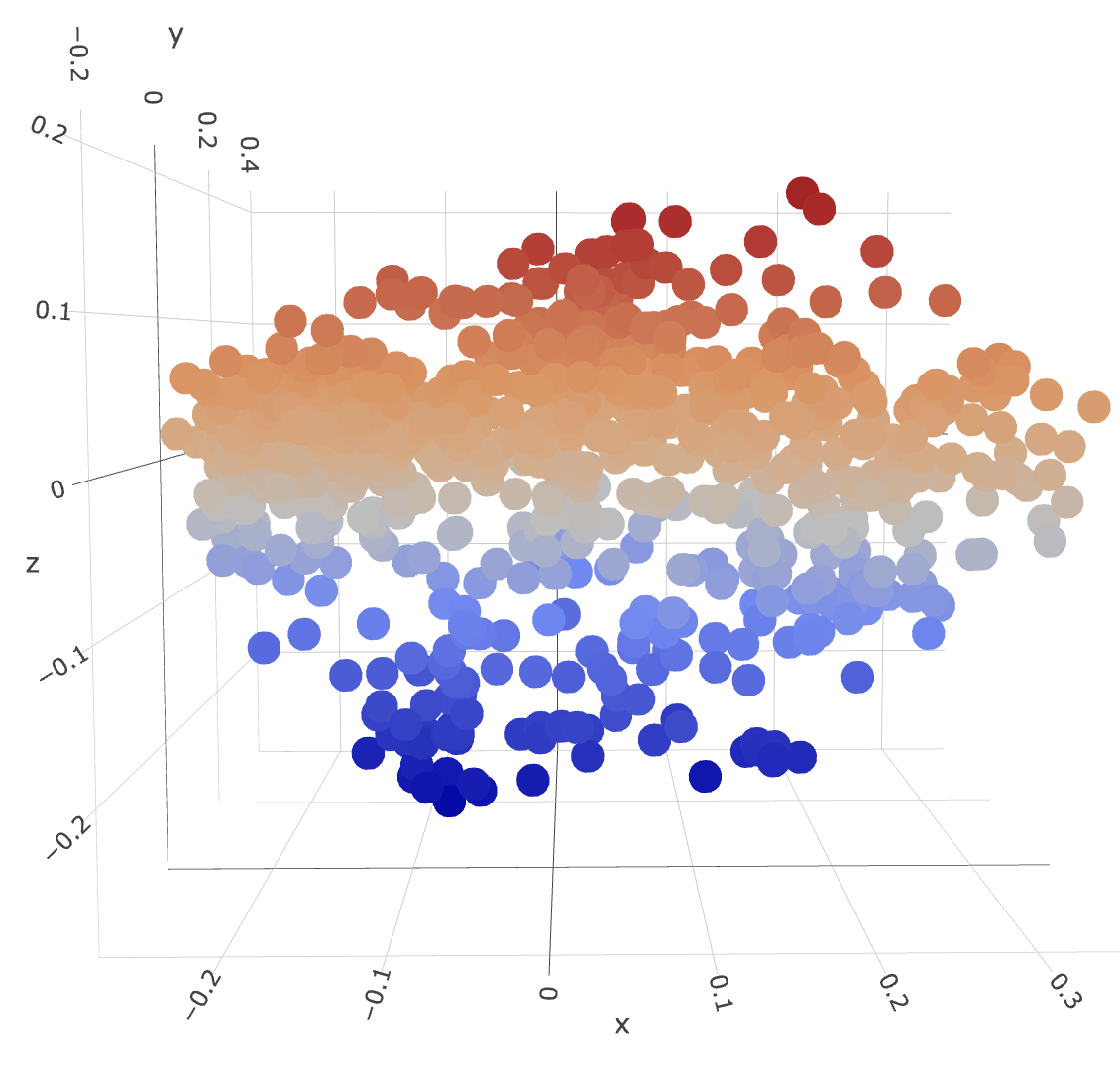} 
\caption[]{}
\label{3d-zx}
\end{subfigure}
\begin{subfigure}{0.50\textwidth}
\includegraphics[width=0.9\linewidth]{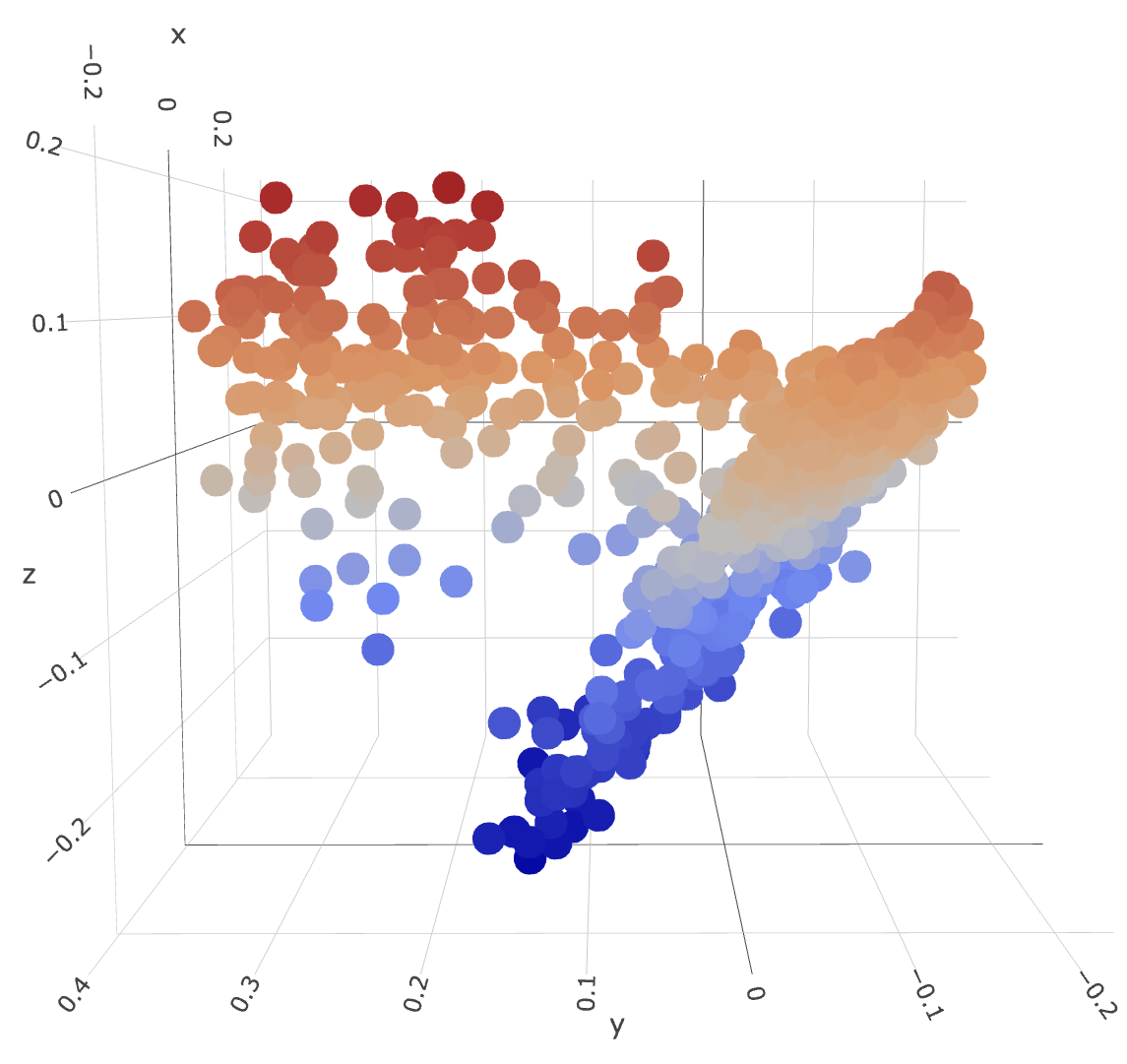} 
\caption[]{}
\label{3d-yz}
\end{subfigure}
\begin{subfigure}{0.50\textwidth}
\includegraphics[width=0.9\linewidth]{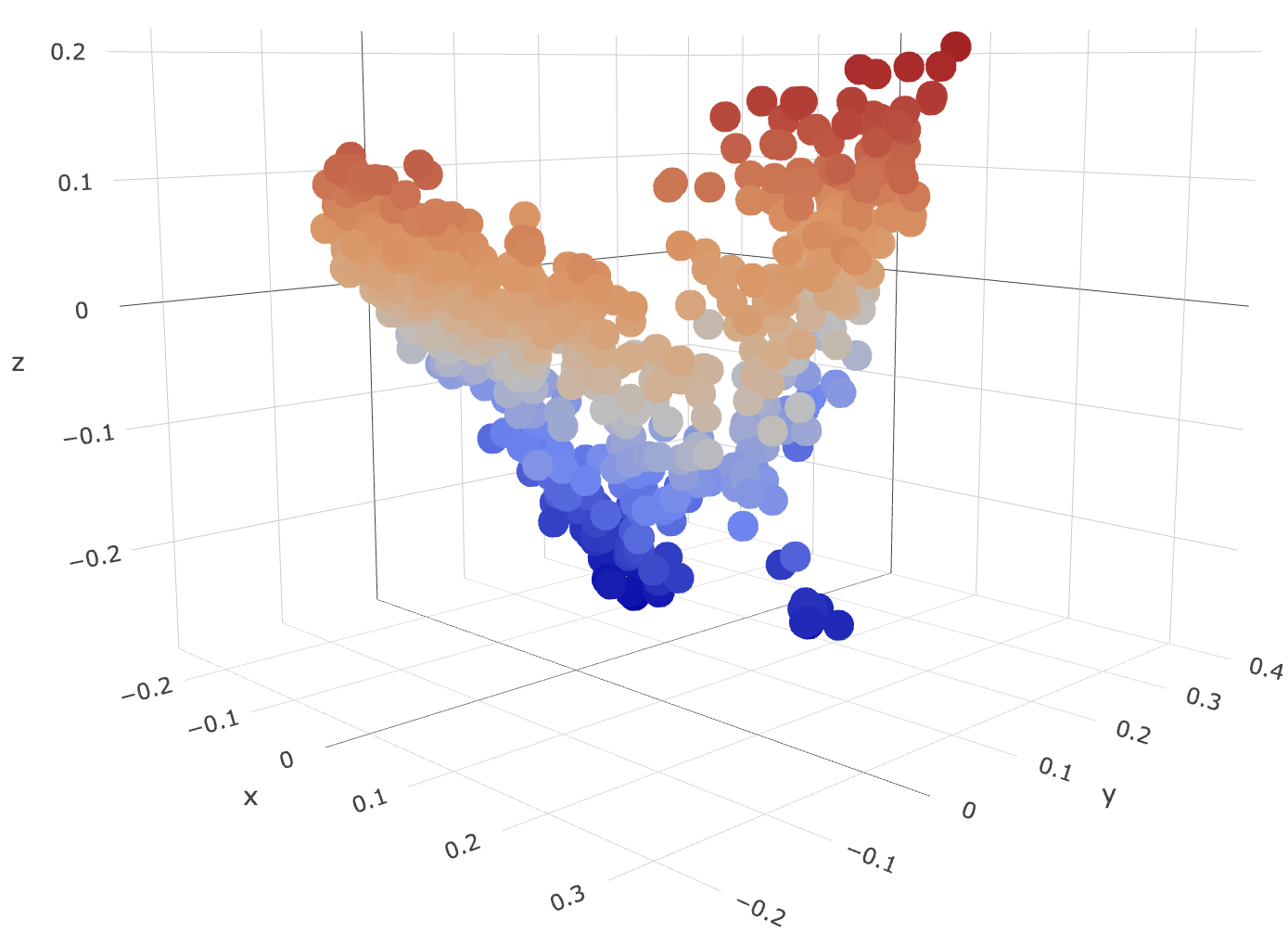} 
\caption[]{}
\label{3d-xyz}
\end{subfigure}
\caption[Three-dimensional unlabelled MDS map from the perspective of different axes]{Three-dimensional unlabelled MDS map from the perspective of different axes}
\label{3dmds}
\end{figure}

In fact, by adding a third dimension to the map, we can see that the observations in that area are distributed across the length of the $z$-axis, so that any two points that are close to each other in a two-dimensional plot may actually be very distant on a third dimension. Figure \ref{3dmds} shows snapshots of the three-dimensional MDS map from different angles, with observations coloured by how distant the observations are from the two-dimensional plane formed by the $x$ and $y$ axes only: the warmer the colour (red), the closer to the plane, the colder the colour (blue), the farther from the plane. (\ref{3d-xy}) correspond to the view with the $x$ and $y$ axes in the foreground and the $z$-axis on the background, namely the same perspective as the one obtained from the two-dimensional map in Figure \ref{plainmds} and elsewhere. (\ref{3d-zx}) shows the map from the perspective of the $x$-$z$ plane, namely the one obtained by rotating (\ref{3d-xy}) 90° clockwise around the $x$-axis (i.e. by rotating (\ref{3d-xy}) upwards/backwards). (\ref{3d-yz}) shows the map from the perspective of the $y$-$z$ plane, namely the one obtained by rotating (\ref{3d-zx}) 90° clockwise around the $z$-axis (i.e. by rotating (\ref{3d-zx}) rightwards). Finally, (\ref{3d-xyz}) shows the three-dimensional map as if facing the observations from the bottom right corner of Figure \ref{3d-zx} or \ref{plainmds}.\footnote{An interactive version of the 3D plot can be found in this Observable notebook: \url{https://observablehq.com/@npedrazzini/3d-mds-when-map}.}

\begin{figure}[!h]
\centering
\includegraphics[width=0.6\textwidth]{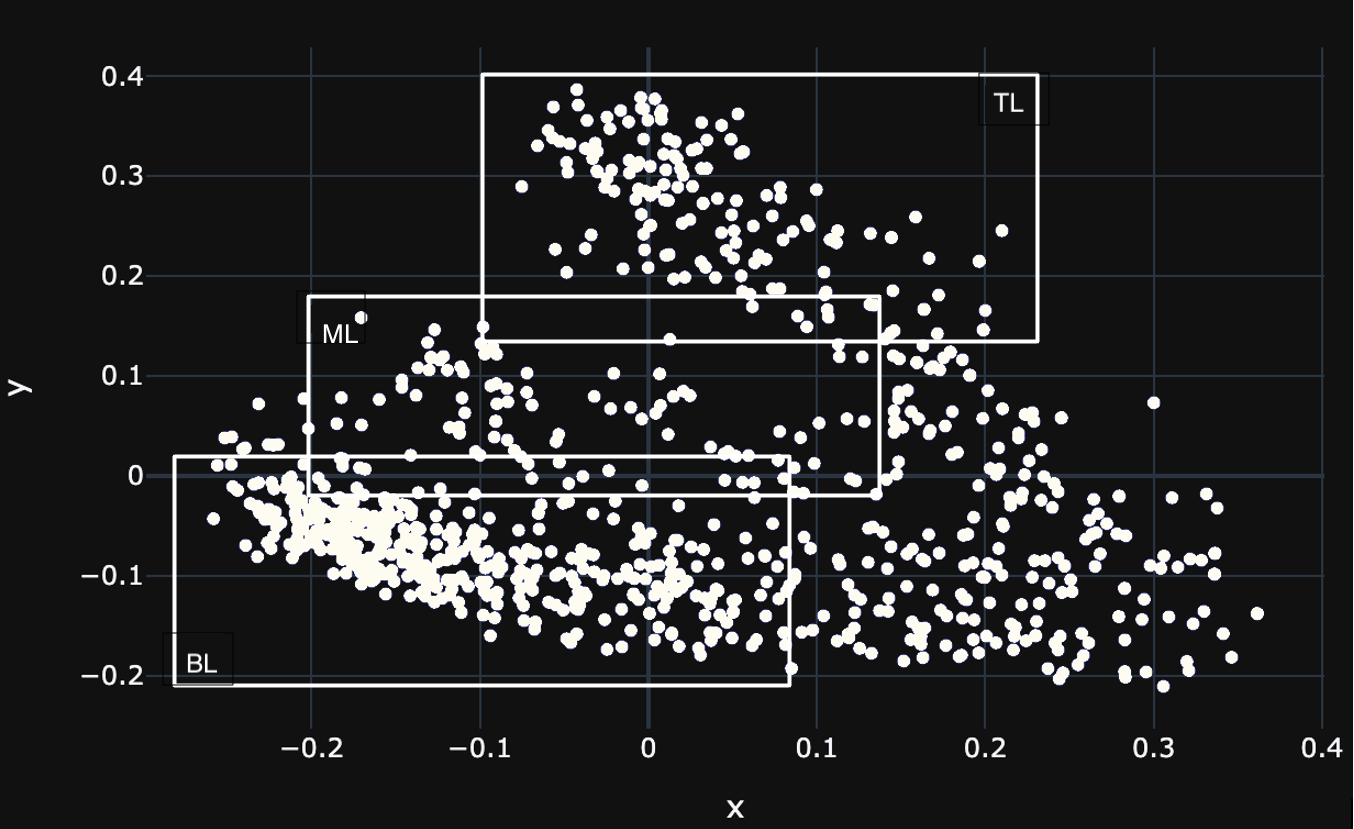}
\caption{\label{threeway}Potential three-way split of non-null situations.}
\end{figure}

Abstracting from the variation observed on the left side of all the maps, we can identify a main three-way split in the possible patterns of lexification (Figure \ref{threeway}). Let us informally refer to the three areas in this split as TL (top-left), ML (mid-left), and BL (bottom-left). 
    
We can speculate that area BL could actually also contain an area BL-2, since that is a configuration found in several languages, as Figures \ref{bgz-when} and \ref{gej-when} showed.

\begin{figure}[!h]
\centering
\includegraphics[width=0.6\textwidth]{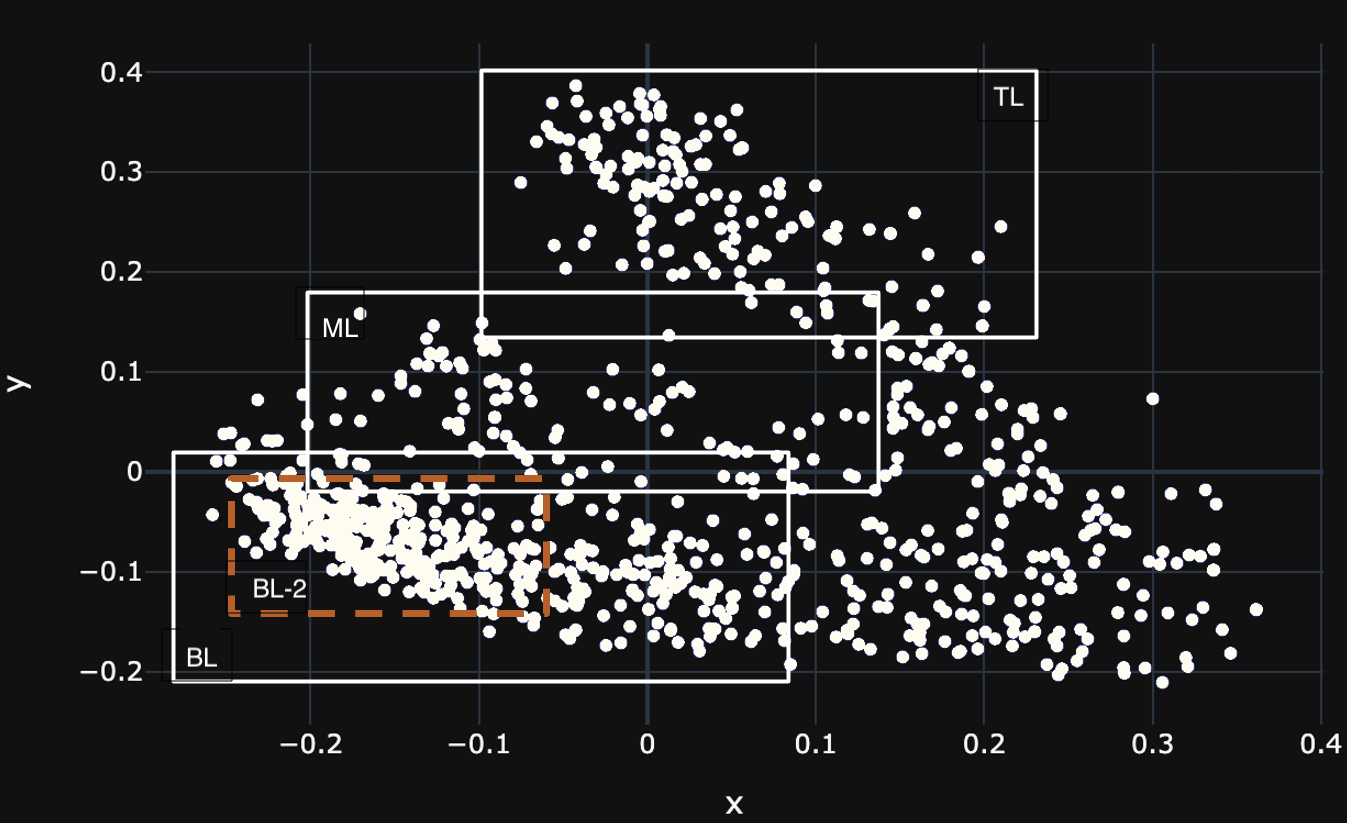}
\caption{\label{threeway2}Potential four-way split of non-null situations.}
\end{figure}

\begin{figure}[!h]
\centering
\includegraphics[width=0.6\textwidth]{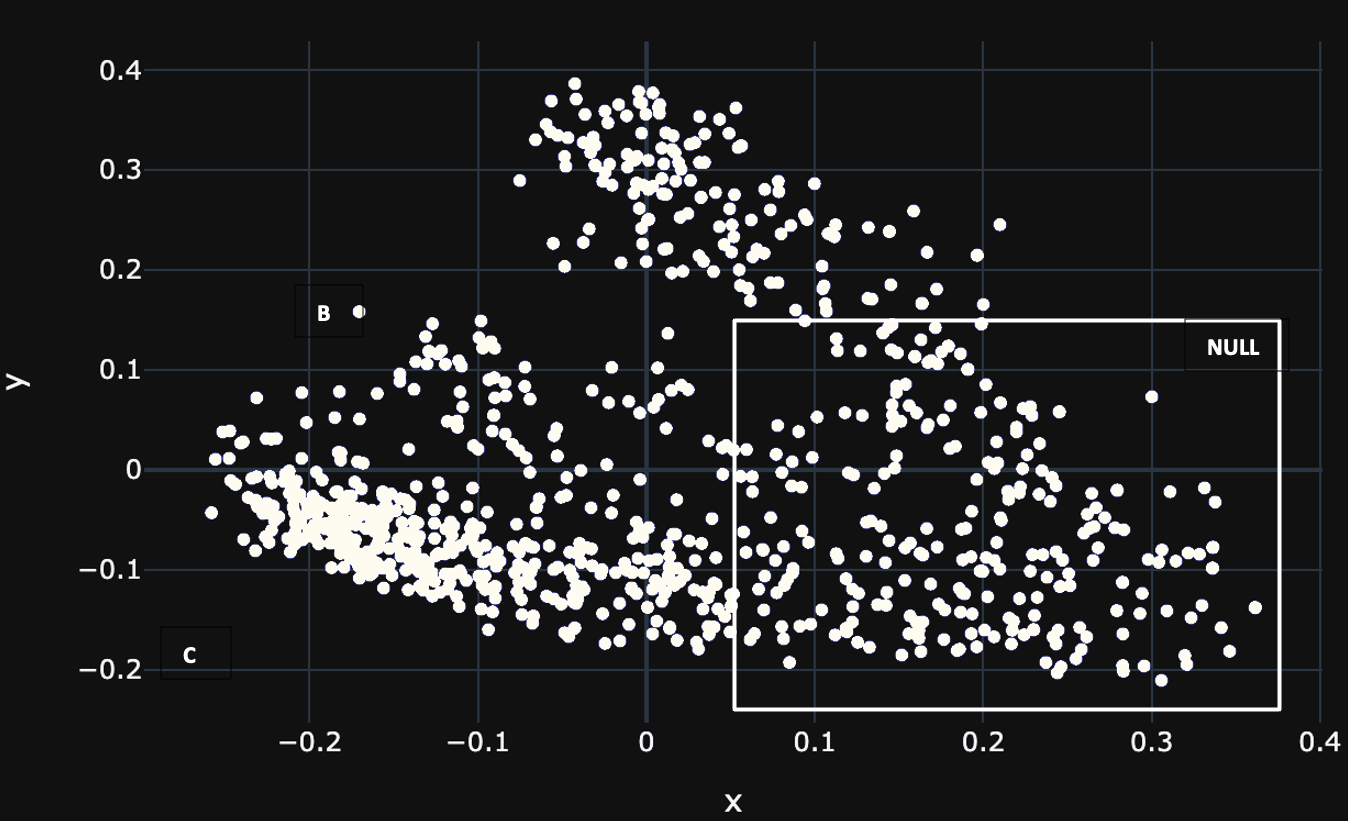}
\caption{\label{NULLS} Approximate null area across all target languages}
\end{figure}

On the basis of a two-dimensional map, it seems safe to use the three-way split shown in Figure \ref{threeway} as a template that is potentially applicable to roughly the left half of the map for most of the languages in the corpus. The cores of the endpoints of these three areas approximately correspond to the GMM cluster centroids identified in Figure~\ref{fig:centroidwithballtree} above. Virtually all languages possessing non-null areas seem to group the three clusters in Figure \ref{threeway} together following (approximately) one of their boundaries, either individually or by merging two or all three of them. The rightmost area (Figure \ref{NULLS}) is instead always predominantly occupied by `null' situations across all languages.

Looking more closely at the attested configurations to the left of the maps, we observe several different patterns of colexification for areas TL, ML, and BL (following the lettering from Figure \ref{threeway}). Using the points within GMM clusters 3, 2 and 5 identified by the balltree approach as the core points for areas TL, ML, and BL, respectively, following the method described in Section \ref{sec:methods}, we can now automatically classify languages based on how their Kriging maps fit those GMM clusters.  

\subsection{Classification}\label{sec:classif}
In the majority of languages (1165 out of 1452), there is one Kriging area that is the best correspondence to each GMM area. This does not mean that the \textit{same} Kriging area cannot span two or all three of the GMM clusters under consideration, but only that the competition around the centroid of each of those clusters is low enough for 1165/1452 languages to have one predominant means. For such languages, then, there are five logically possible colexification patterns. Table~\ref{tbl:colex} shows these with their frequencies. Note that a null Kriging area is considered as a separate means in its own right. The table thus also indicates how many languages (`freq null') have a null Kriging area rather than a lexified one for any of the three clusters (`null in' indicates the cluster which the counts in `freq null' refer to).

\begin{table}[!h]
\centering
\begin{tabular}{|l|l|l|l|l|}
\hline
 & \textbf{pattern} & \textbf{freq} & \textbf{null in} & \textbf{freq null} \\\hline
\textbf{A} & TL = ML = BL  &  636 & all & 250 \\\hline
\textbf{B} & (TL = ML) $\neq$ BL  & 146 & TL,ML & 23\\
& & & BL & 84 \\\hline
\textbf{C} & TL $\neq$ (ML = BL) & 198 & TL & 14 \\
& & & ML,BL & 53 \\\hline
\textbf{D} & TL $\neq$ ML $\neq$ BL & 110 & TL & 6 \\
& & & ML & 21 \\
& & & BL & 29 \\\hline
\textbf{E} & (TL = BL) $\neq$ ML & 75  & ML & 6 \\
& & & TL,BL & 47 \\
\hline
    \end{tabular}
    \caption{Frequency of lexification patterns}
    \label{tbl:colex}
\end{table}

An additional 222 languages have significant competition with at least one more Kriging area within one of the three GMM clusters, but these could be subsumed to one of the five main patterns in Table \ref{tbl:colex}. If, for example, the TL area has two different competing Kriging areas, while the ML and BL areas are together under the scope of yet another Kriging area, then the main pattern could still be considered C (TL $neq$ (ML = BL)), even though the TL area has two, rather than one prominent means. The updated frequencies with the addition of these 222 languages to the respective patterns are shown in Table \ref{tbl:colex2}.\footnote{I have included both classifications for all languages in the Appendix, indicating the main pattern and the subpattern (if any), as well as a breakdown of how each subpattern was assigned.}

\begin{table}[!h]
\centering
\begin{tabular}{|l|l|l|l|l|}
\hline
 & \textbf{pattern} & \textbf{freq} & \textbf{null in} & \textbf{freq null} \\\hline
\textbf{A} & TL = ML = BL  &  639 & all & 250 \\\hline
\textbf{B} & (TL = ML) $\neq$ BL  & 171 & TL,ML & 24\\
& & & BL & 84 \\\hline
\textbf{C} & TL $\neq$ (ML = BL) & 277 & TL & 14 \\
& & & ML,BL & 59 \\\hline
\textbf{D} & TL $\neq$ ML $\neq$ BL & 195 & TL & 9 \\
& & & ML & 26 \\
& & & BL & 40 \\\hline
\textbf{E} & (TL = BL) $\neq$ ML & 105  & ML & 6 \\
& & & TL,BL & 47 \\
\hline
    \end{tabular}
    \caption{Frequency of lexification patterns, including subpatterns within a main pattern}
    \label{tbl:colex2}
\end{table}

\begin{figure}[!h]
\begin{subfigure}{0.50\textwidth}
\includegraphics[width=0.9\linewidth]{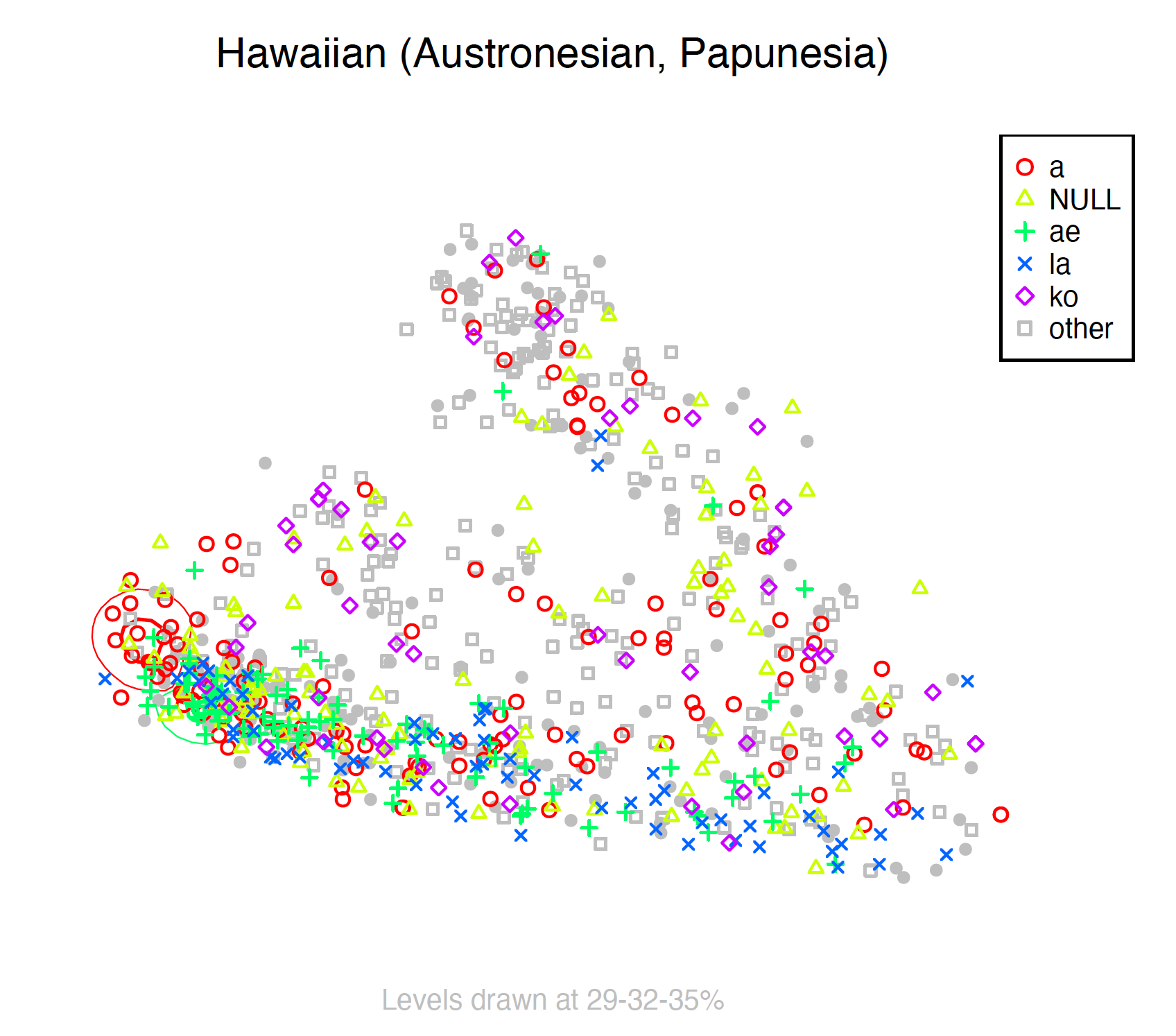} 
\caption[Kriging map for Hawaiian (Austronesian, Papunesia)]{}
\label{hawaiian}
\end{subfigure}
\begin{subfigure}{0.50\textwidth}
\includegraphics[width=0.9\linewidth]{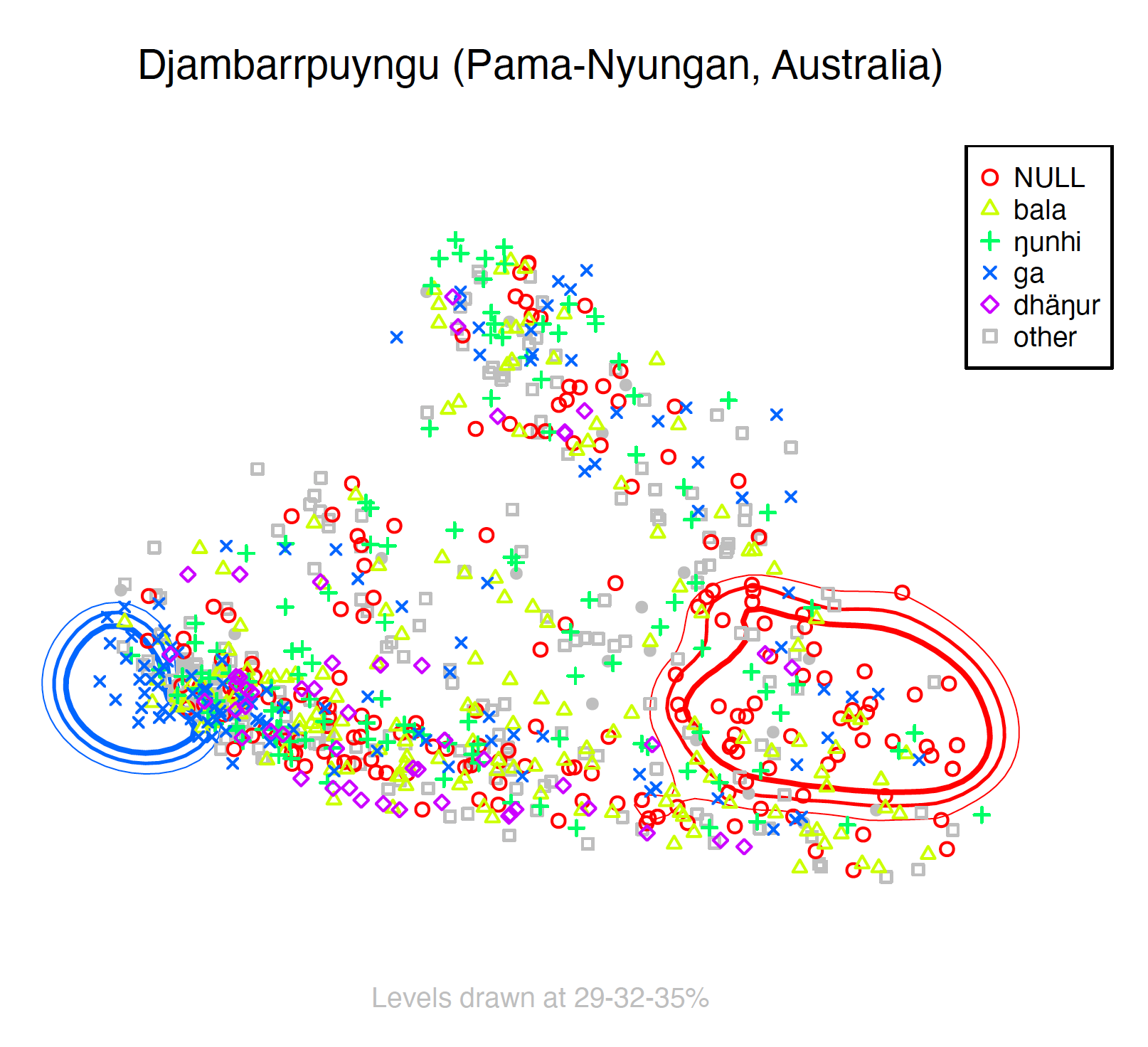}
\caption[Kriging map for Djambarrpuyngu (Pama-Nyungan, Australia)]{}
\label{djambarrpuyngu}
\end{subfigure}
\begin{subfigure}{0.50\textwidth}
\includegraphics[width=0.9\linewidth]{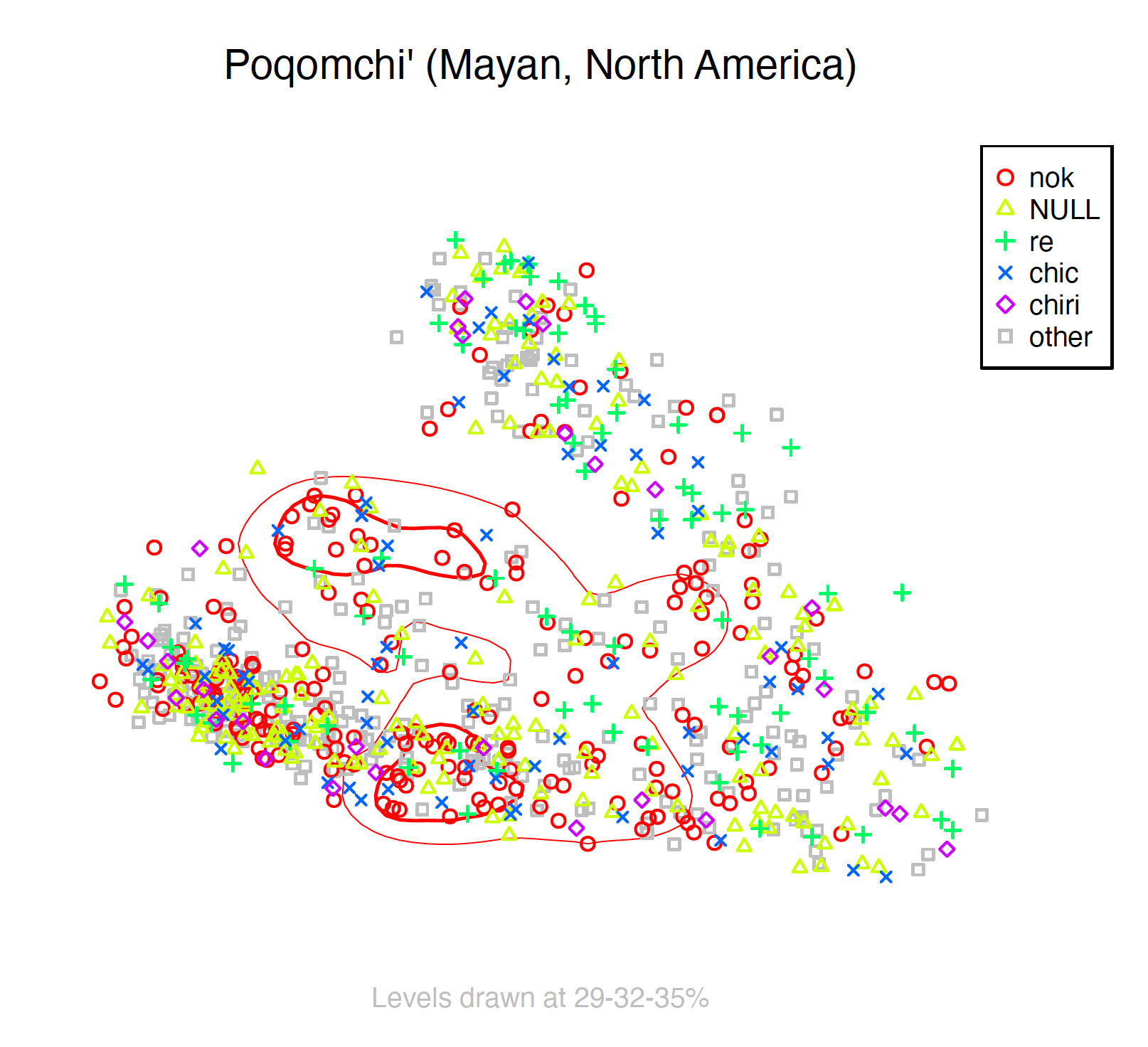}
\caption[Kriging map for Poqomchi' (Mayan, North America)]{}
\label{poqomchi}
\end{subfigure}
\begin{subfigure}{0.50\textwidth}
\includegraphics[width=0.9\linewidth]{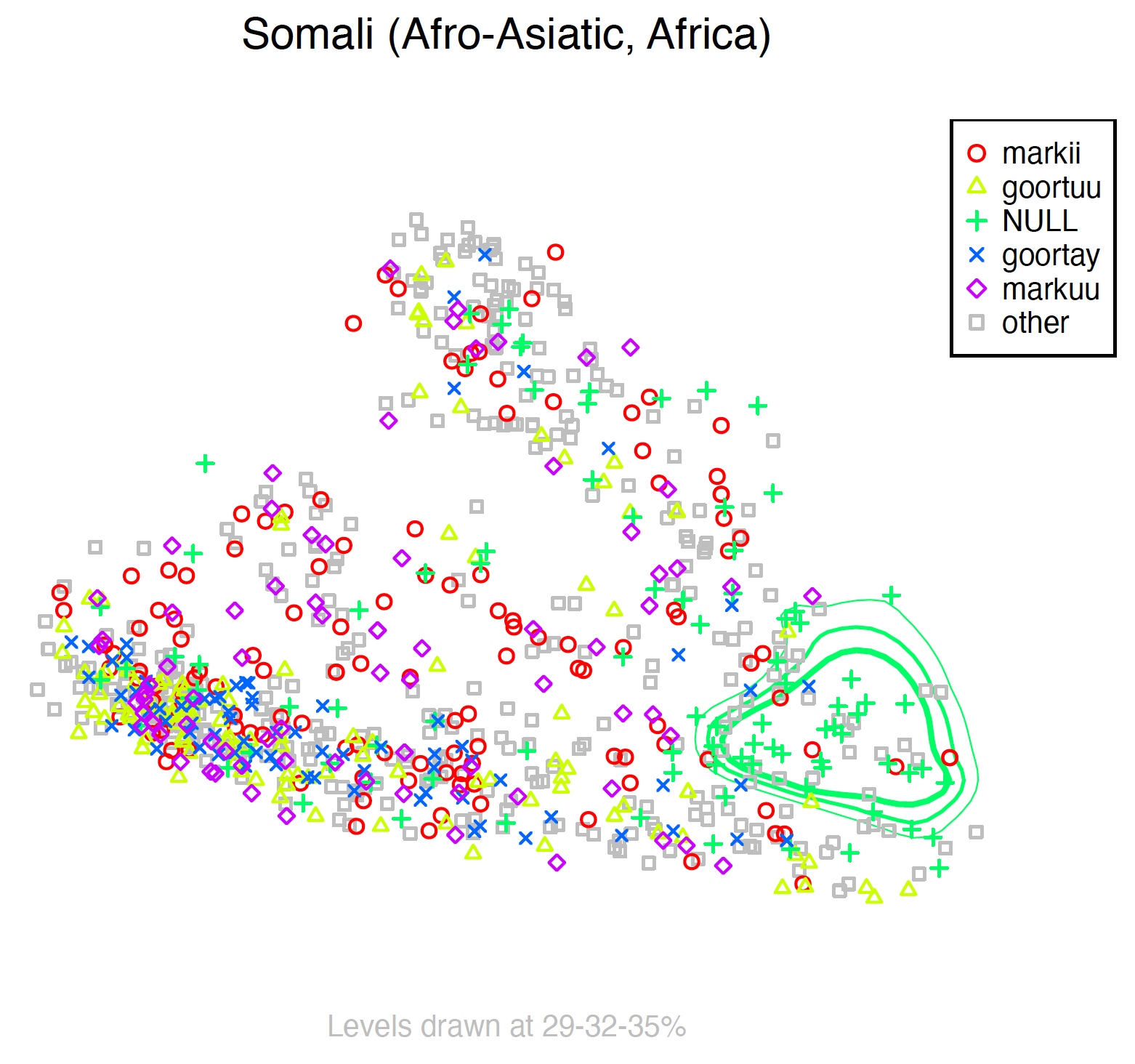}
\caption[Kriging map for Somali (Afro-Asiatic, Africa)]{}
\label{somali}
\end{subfigure}
\caption[]{Kriging maps for Hawaiian (Austronesian, Papunesia), Djambarrpuyngu (Pama-Nyungan, Australia), Poqomchi' (Mayan, North America), Somali (Afro-Asiatic, Africa)}
\end{figure}

40 languages in the dataset have at least one GMM cluster in which there is no Kriging area (i.e. there is not one particular means that is significantly more prominent than others), so we are not able to assign them to any of the five main patterns. This is the case, for example, of Hawaiian (Figure \ref{hawaiian}), Djambarrpuyngu (Figure \ref{djambarrpuyngu}), Poqomchi' (Figure \ref{poqomchi}), and Somali (Figure \ref{somali}). 

\begin{figure}[!h]
\begin{subfigure}{0.50\textwidth}
\includegraphics[width=0.9\linewidth]{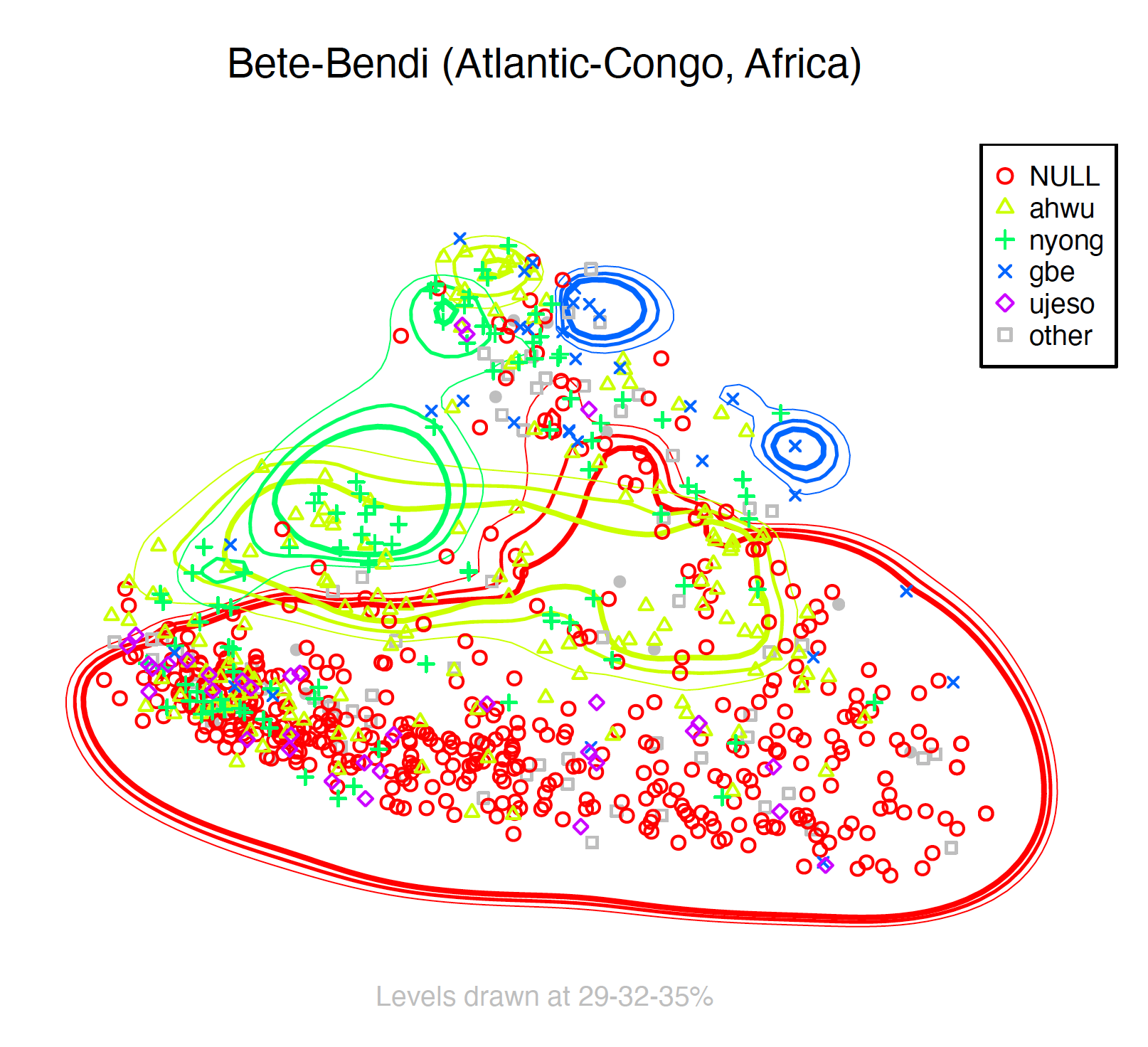} 
\caption[Kriging map for Bete-Bendi (Atlantic-Congo, Africa)]{}
\label{betebendi}
\end{subfigure}
\begin{subfigure}{0.50\textwidth}
\includegraphics[width=0.9\linewidth]{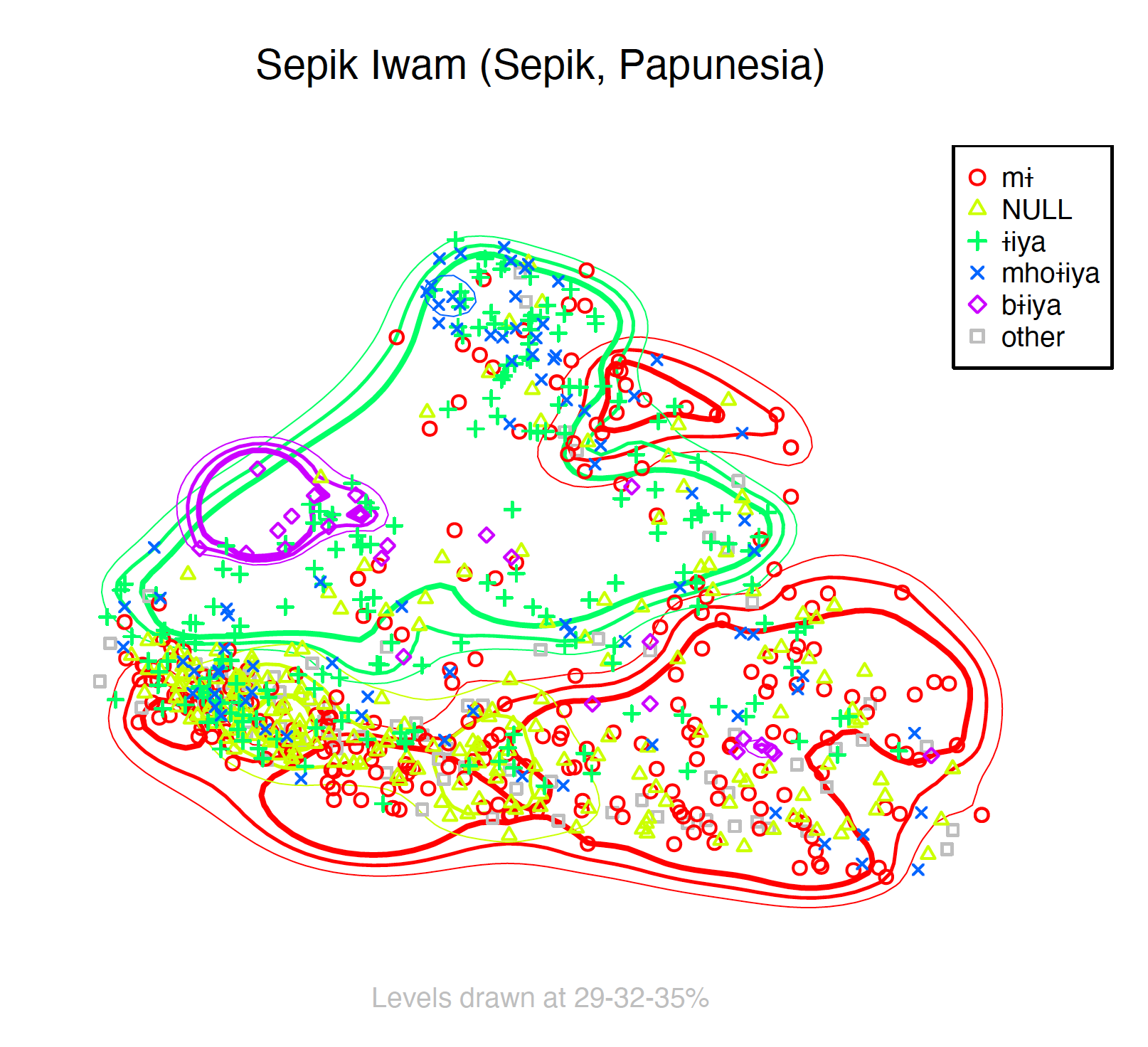}
\caption[Kriging map for Sepik Iwam (Sepik, Papunesia)]{}
\label{sepikiwam}
\end{subfigure}
\caption[]{Kriging maps for Bete-Bendi (Atlantic-Congo, Africa) and Sepik Iwam (Sepik, Papunesia)}
\end{figure}

A small number of languages (17) have at least one main Kriging area per GMM cluster, but their pattern cannot be easily subsumed to any of the 5 main patterns. This is the case, for example, in Bete-Bendi (Figure \ref{betebendi}) and Sepik Iwam (Figure \ref{sepikiwam}).

\begin{figure}[!h]
\begin{subfigure}{0.50\textwidth}
\includegraphics[width=0.9\linewidth]{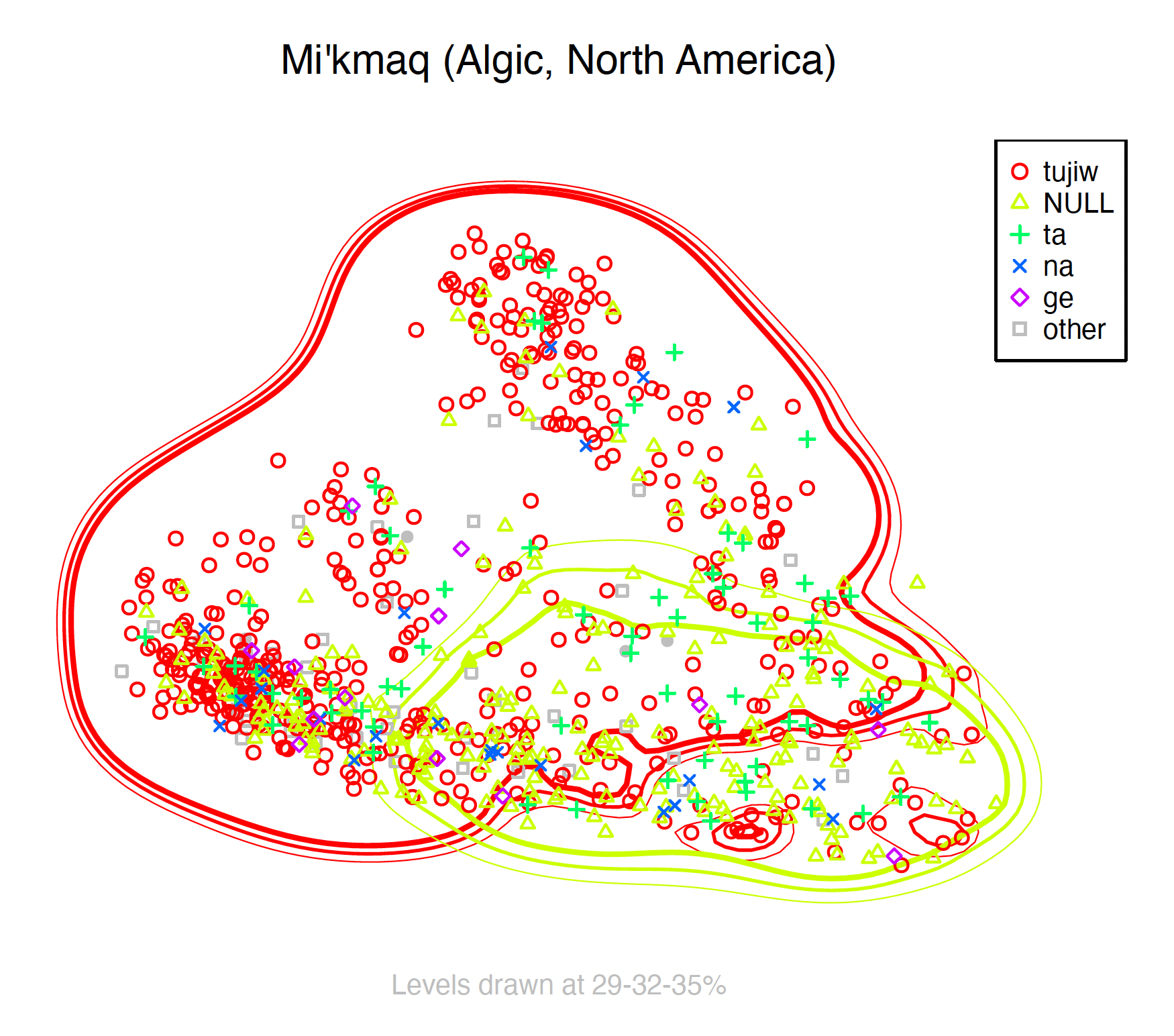} 
\caption[Kriging map for Mi’kmaq (Algic, North America)]{}
\label{mic-when}
\end{subfigure}
\begin{subfigure}{0.50\textwidth}
\includegraphics[width=0.9\linewidth]{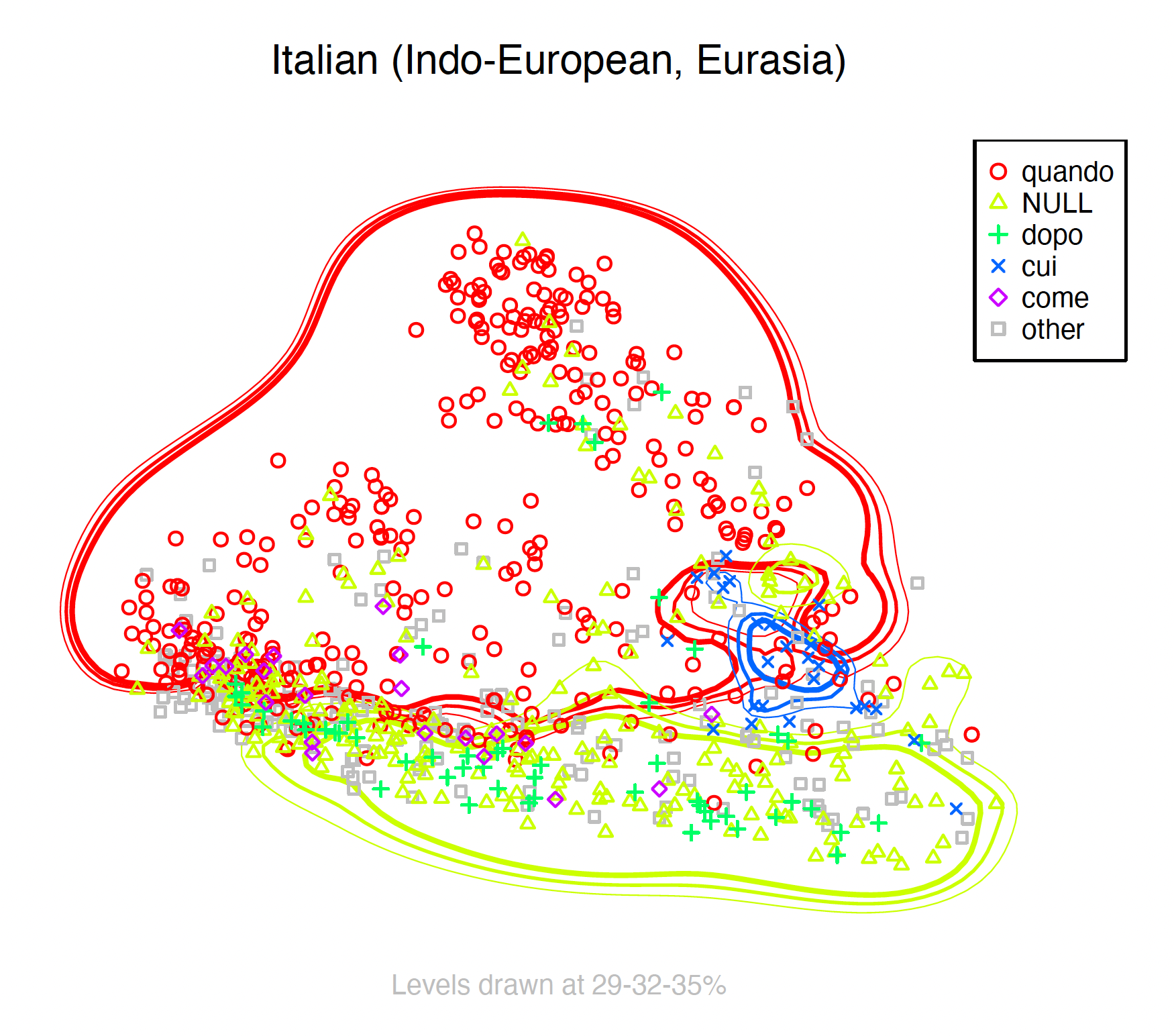}
\caption[Kriging map for Italian (Indo-European, Eurasia)]{}
\label{ita-when}
\end{subfigure}
\caption[]{Kriging maps for Mi’kmaq (Algic, North America) and Italian (Indo-European, Eurasia)}
\end{figure}

Using Table \ref{tbl:colex2} as our baseline, we see that the most common pattern among the world's languages is A, where one main Kriging area encompasses all three GMM clusters, as is the case of Serbian and Moose Cree seen above (Figures \ref{srp-when} and \ref{crm-when}. For some of these languages, such as Mi’kmaq (Figure \ref{mic-when}) and most Romance languages (e.g. Italian; Figure \ref{ita-when}), the competition in the BL area is more evident from the Kriging map despite being assigned to pattern A. Note, for example, how the scope of \textit{quando} `when' in Italian does not cover the whole BL area (and similarly in the other Romance languages). For 250 out of the 639 languages (around 40\%) with pattern A, the Kriging area is a null one. Given the higher likelihood of using null constructions on the right-hand side of the map, it is very likely that, for these languages, this is the case for the whole map. Examples of the latter are the maps of Southern Nambikuára and Southern Altai seen above (Figures \ref{nab-when} and \ref{alt-when}). 

\begin{figure}[!h]
\begin{subfigure}{0.50\textwidth}
\includegraphics[width=0.9\linewidth]{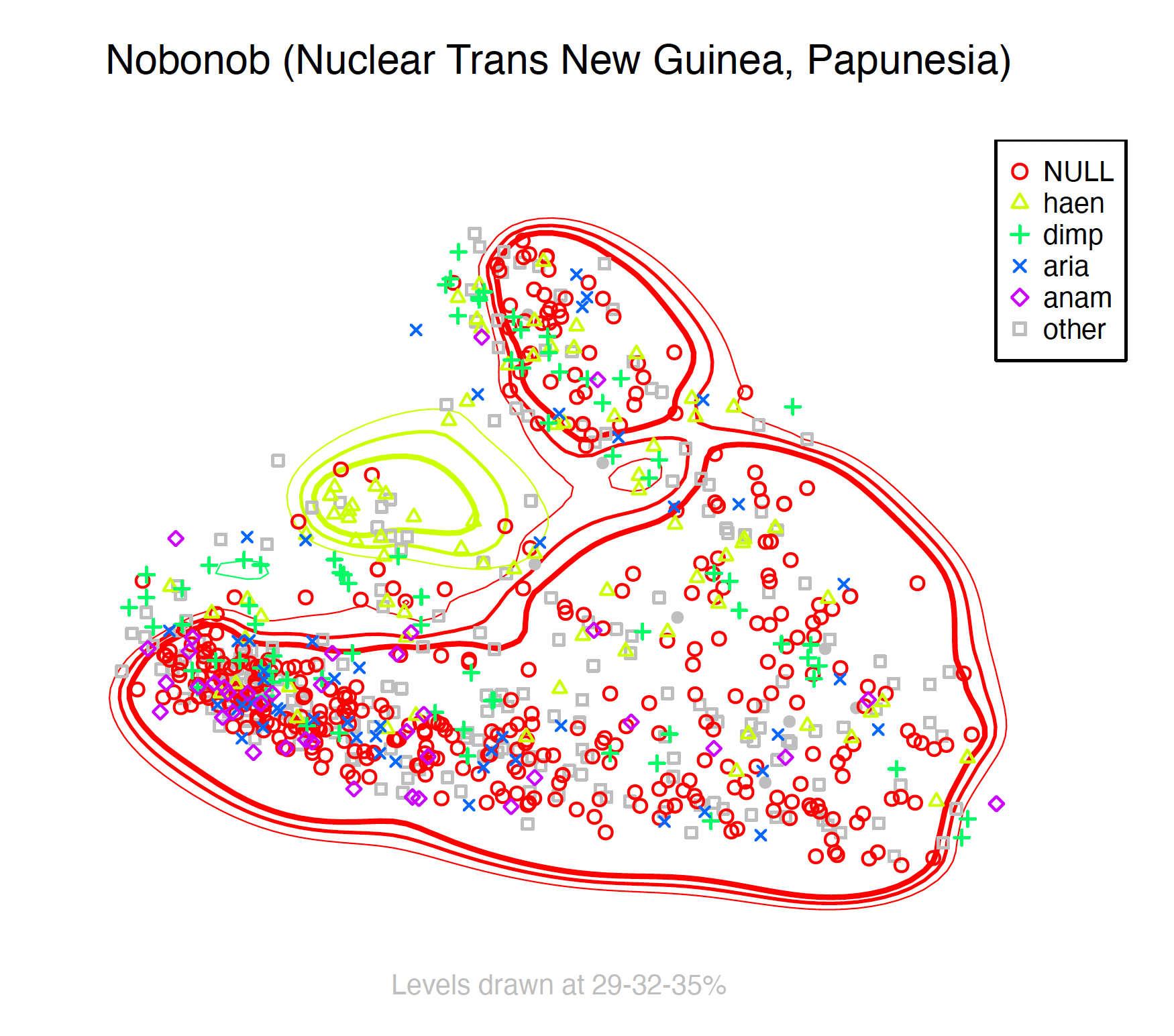} 
\caption[Kriging map for Nobonob (Nuclear Trans New Guinea, Papunesia)]{}
\label{gaw-when}
\end{subfigure}
\begin{subfigure}{0.50\textwidth}
\includegraphics[width=0.9\linewidth]{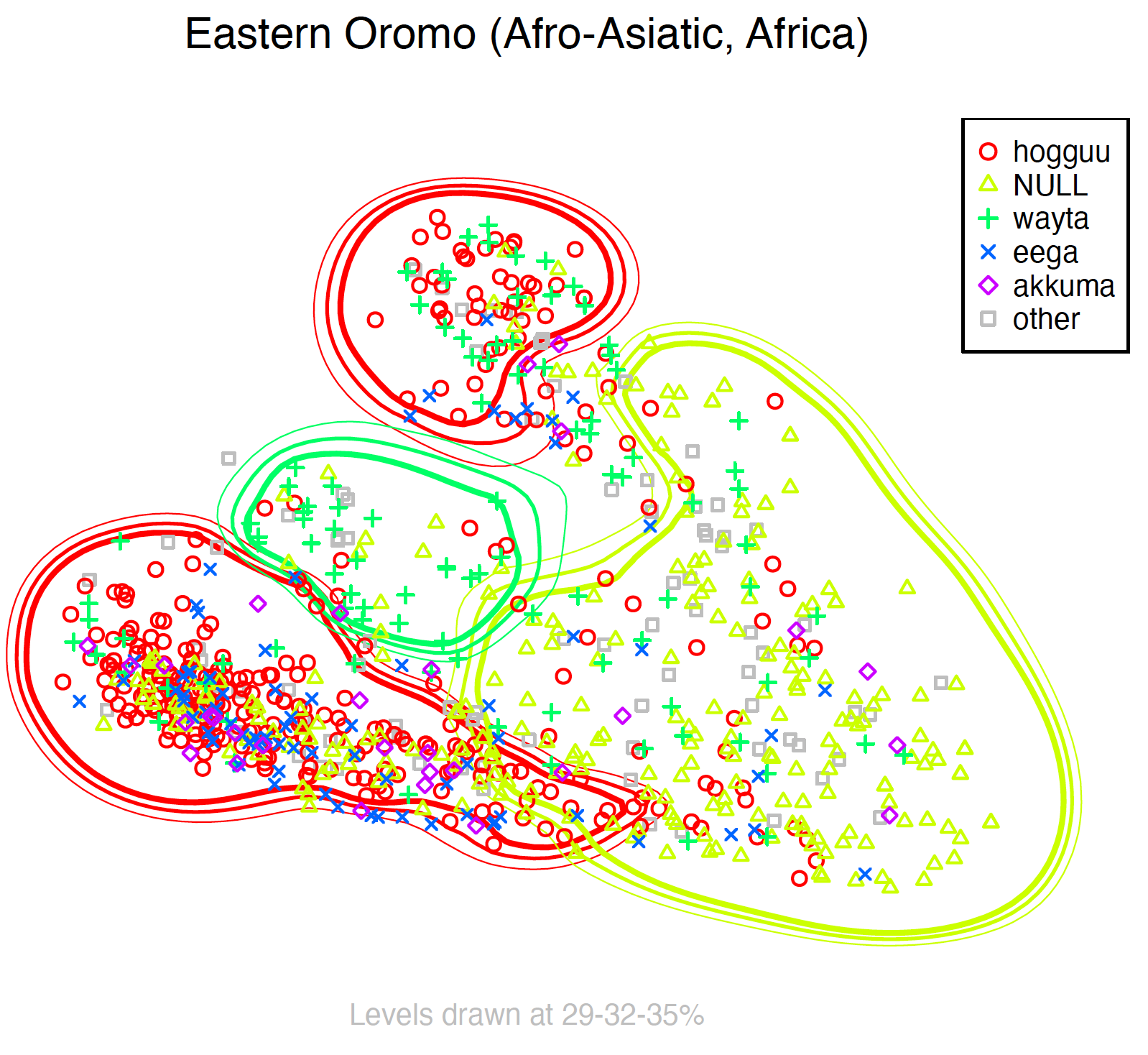}
\caption[Kriging map for Eastern Oromo (Afro-Asiatic, Africa)]{}
\label{hae-when}
\end{subfigure}
\caption[]{Kriging maps for Nobonob (Nuclear Trans New Guinea, Papunesia) and Eastern Oromo (Afro-Asiatic, Africa)}
\end{figure}

The least common pattern is E, consisting of the colexification of the TL and BL areas, with area ML under the scope of a different lexical item. The relative infrequency of this pattern is expected, given that the TL and BL areas are not contiguous in the semantic map. Languages with this pattern are, for example, Nobonob (Figure \ref{gaw-when}), where TL and BL are expressed by null constructions, and Eastern Oromo (Figure \ref{hae-when}), where TL and BL are colexified.

\begin{figure}[!h]
\begin{subfigure}{0.50\textwidth}
\includegraphics[width=0.9\linewidth]{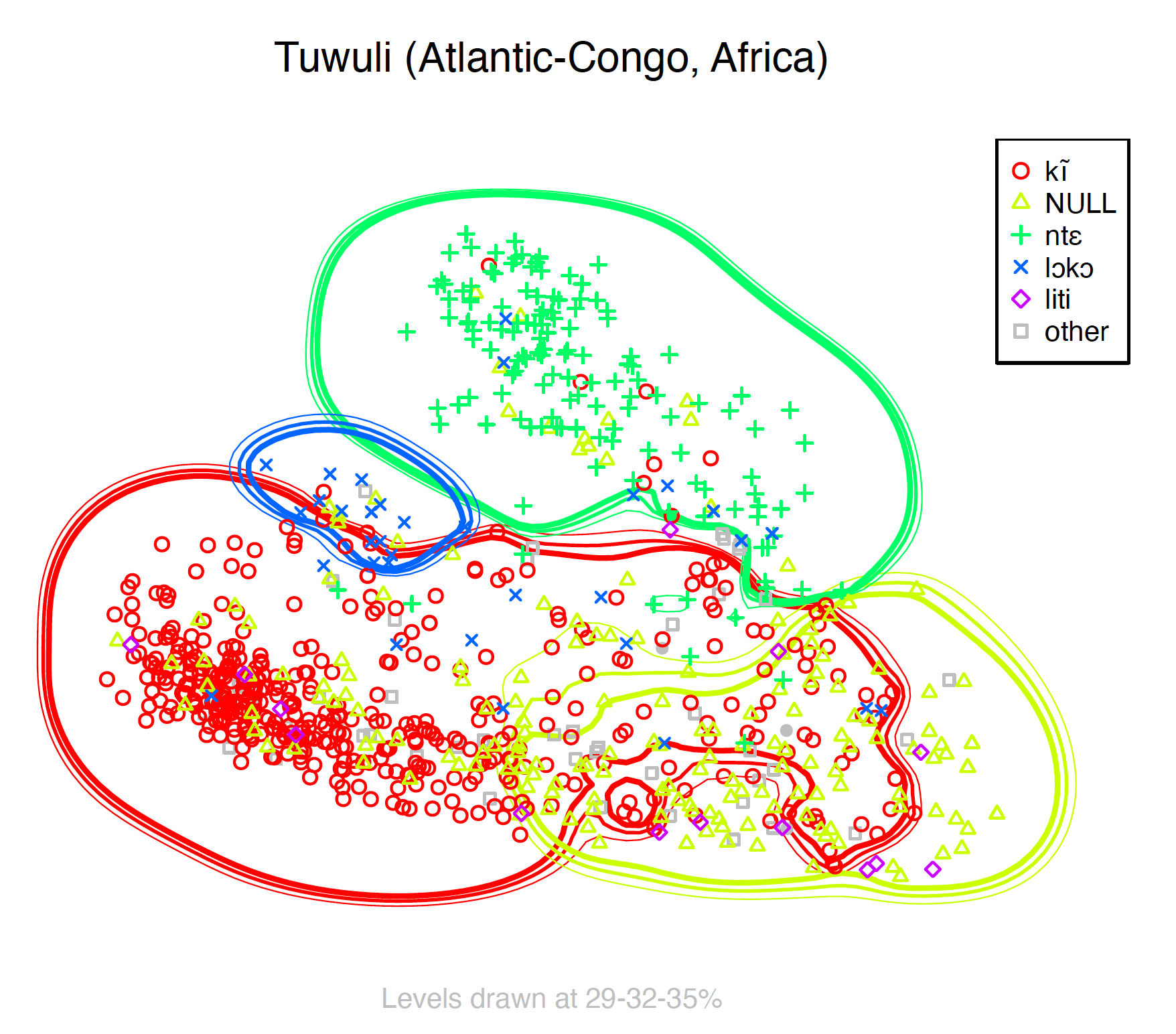} 
\caption[Kriging map for Tuwuli (Atlantic-Congo, Africa)]{}
\label{tuwuli}
\end{subfigure}
\begin{subfigure}{0.50\textwidth}
\includegraphics[width=0.9\linewidth]{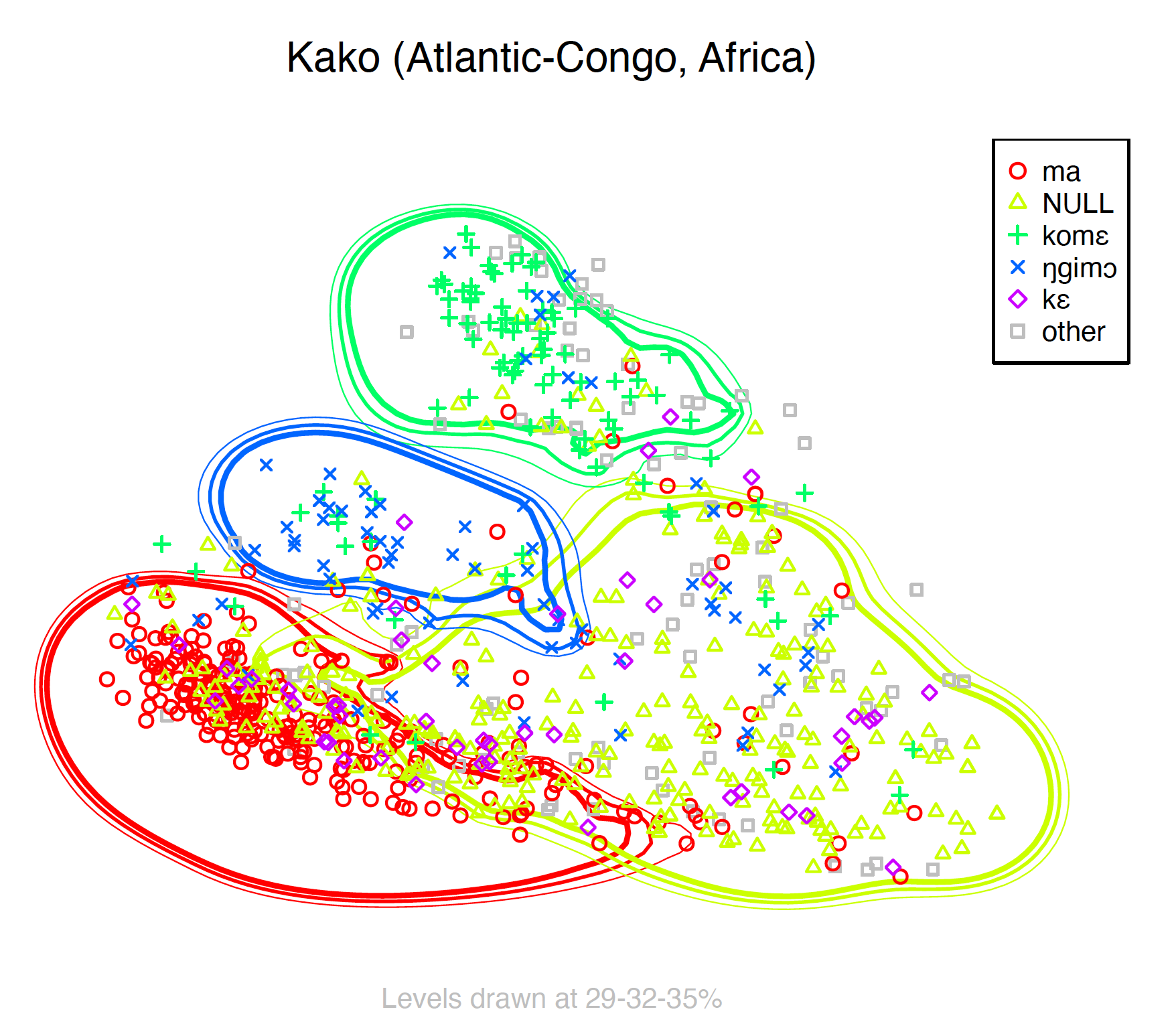}
\caption[Kriging map for Kako (Atlantic-Congo, Africa)]{}
\label{kako}
\end{subfigure}
\caption[]{Kriging maps for Tuwuli (Atlantic-Congo, Africa) and Kako (Atlantic-Congo, Africa)}
\end{figure}

In patterns B, C and D, the TL and BL clusters are under the scope of different Kriging areas, and they differ in whether the ML area is lexified separately or together with either the TL cluster or the BL cluster. \\
\indent Pattern-D languages have a separate Kriging area for the ML cluster, as in Tuwuli (Figure \ref{tuwuli}) and Kako (Figure \ref{kako}). 

\begin{figure}[!h]
\begin{subfigure}{0.50\textwidth}
\includegraphics[width=0.9\linewidth]{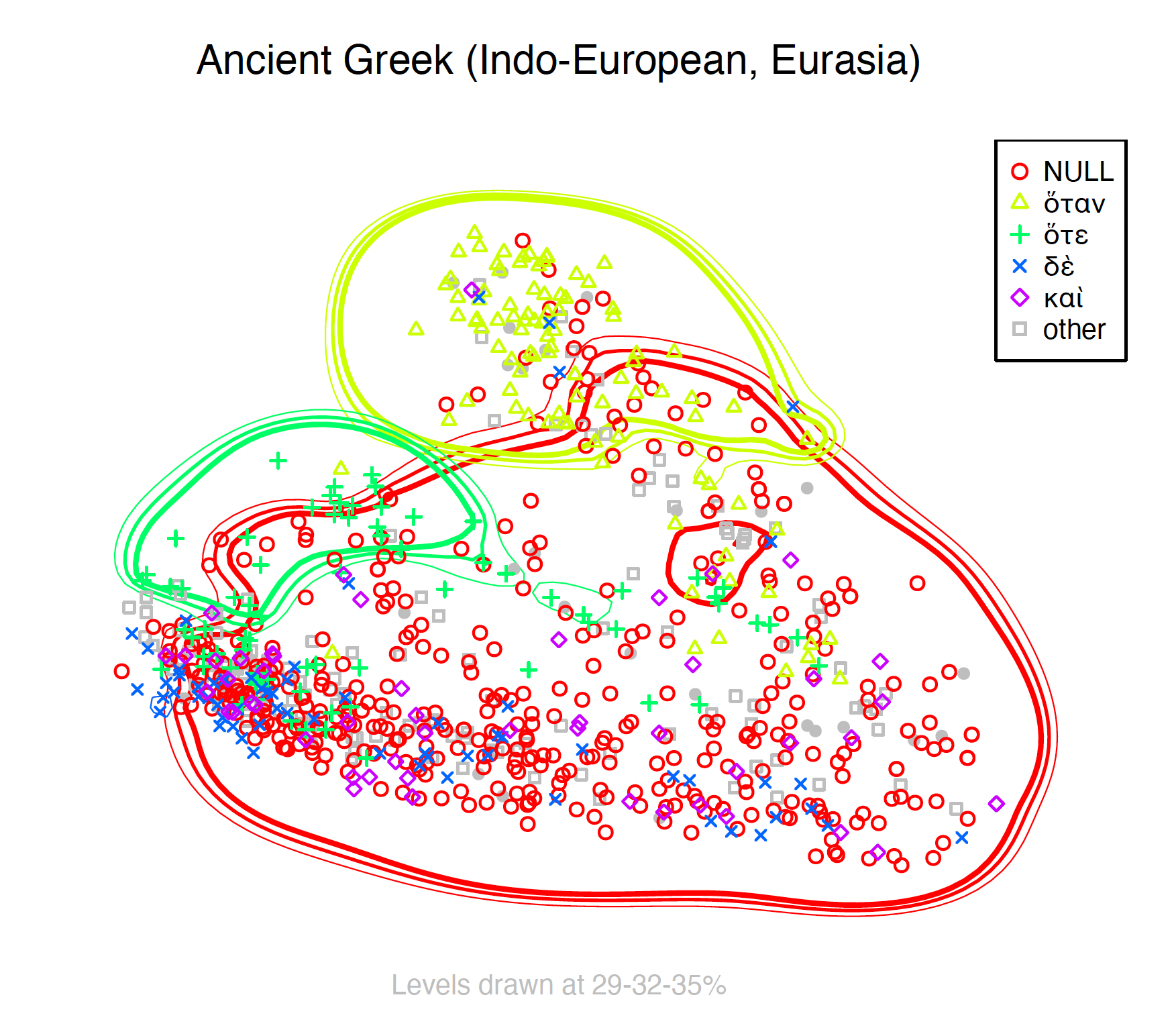} 
\caption[Kriging map for Ancient Greek (Indo-European, Eurasia)]{}
\label{proielgrc0}
\end{subfigure}
\begin{subfigure}{0.50\textwidth}
\includegraphics[width=0.9\linewidth]{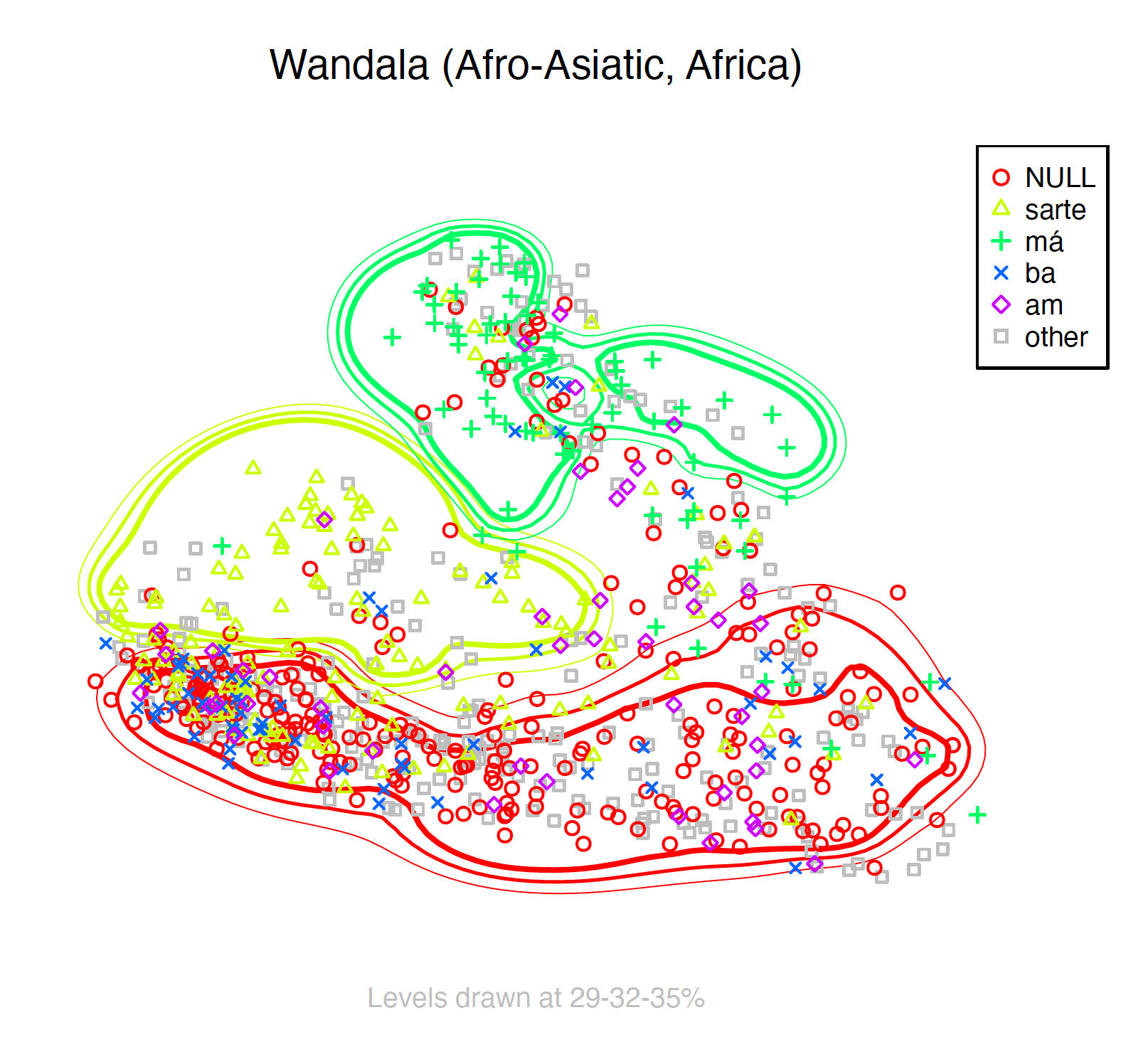}
\caption[Kriging map for Wandala (Afro-Asiatic, Africa)]{}
\label{wandala}
\end{subfigure}
\begin{subfigure}{0.50\textwidth}
\includegraphics[width=0.9\linewidth]{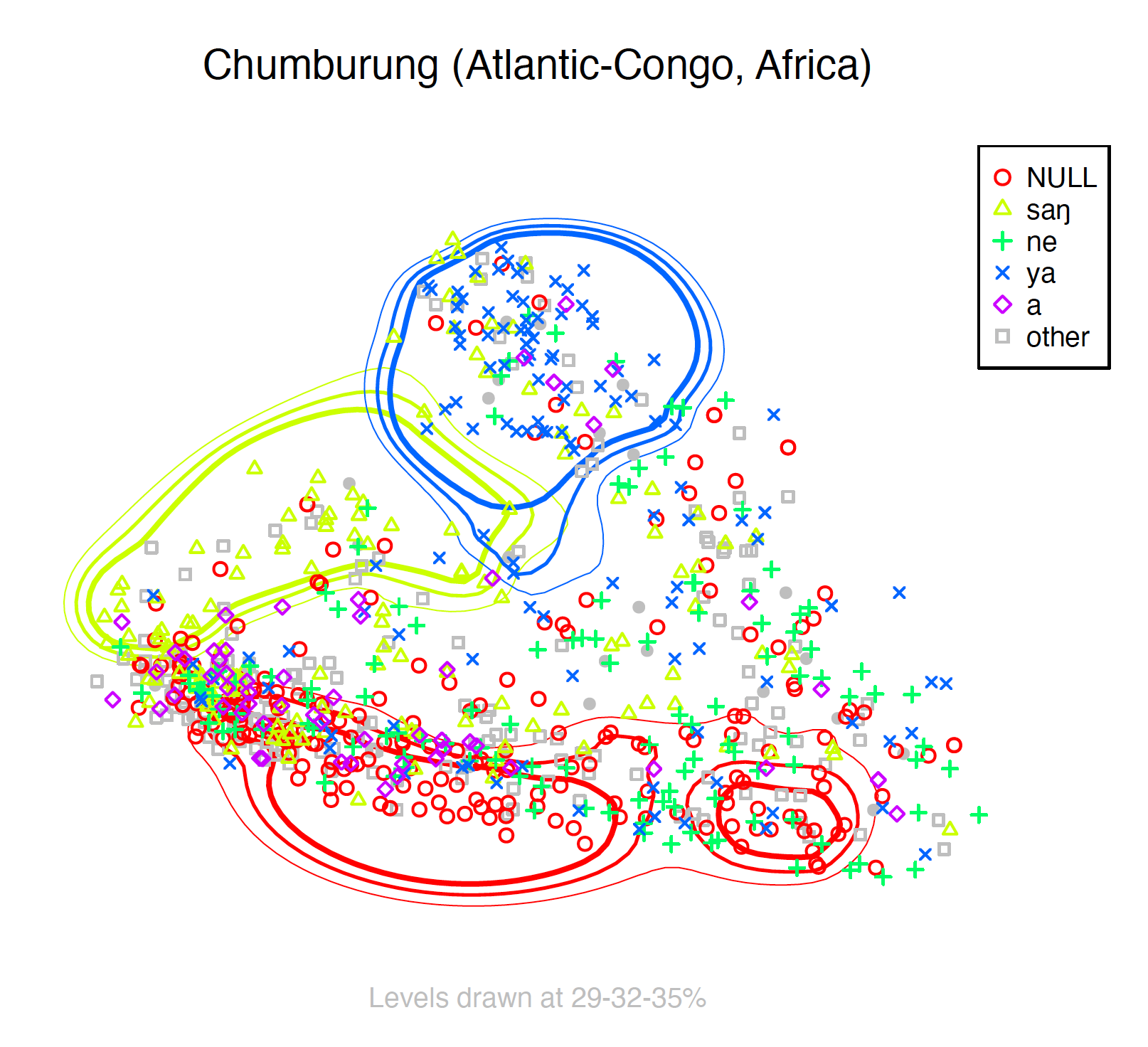}
\caption[Kriging map for Chumburung (Atlantic-Congo, Africa)]{}
\label{chumburung}
\end{subfigure}
\begin{subfigure}{0.50\textwidth}
\includegraphics[width=0.9\linewidth]{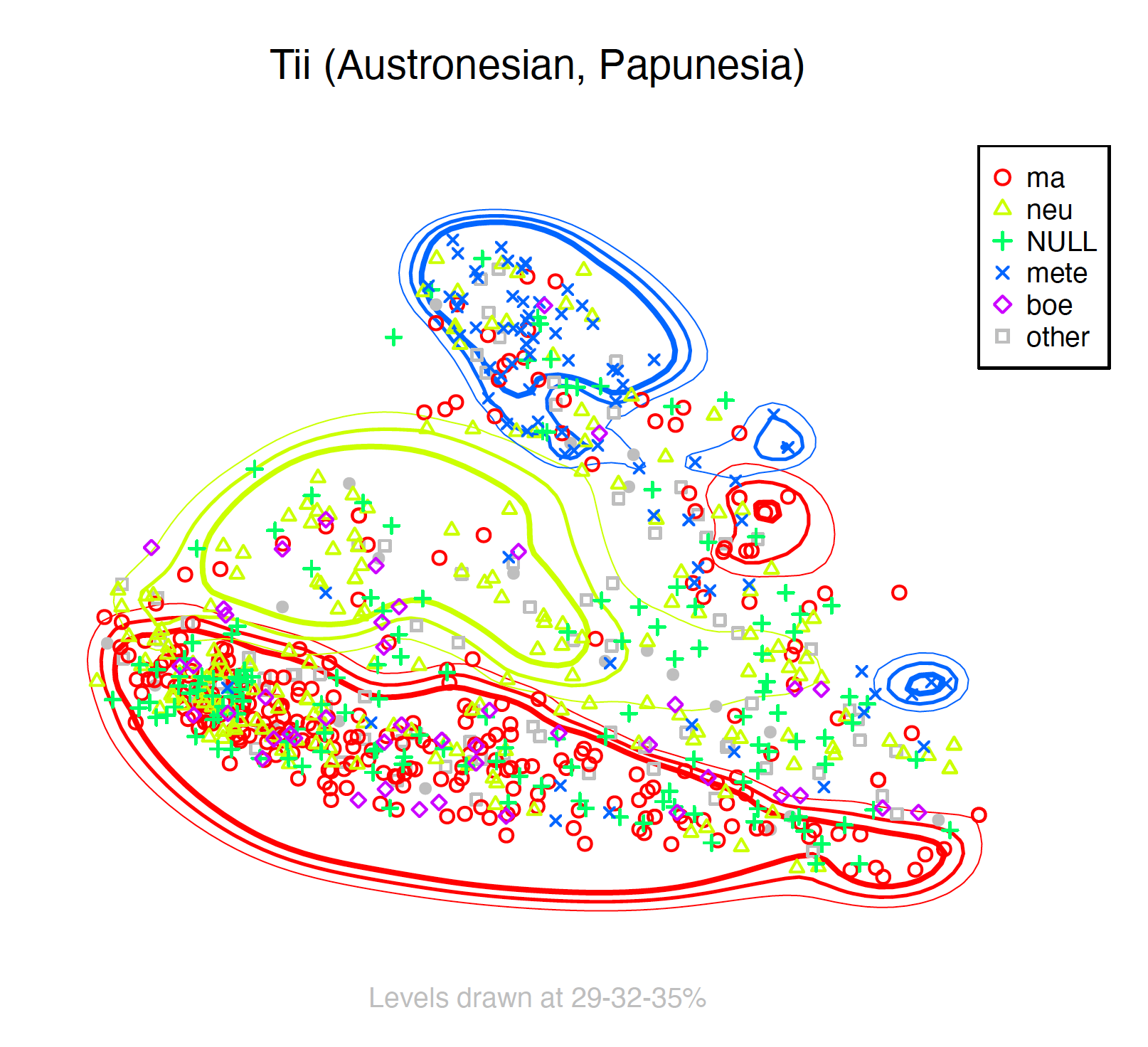}
\caption[Kriging map for Tii (Austronesian, Papunesia)]{}
\label{tii}
\end{subfigure}
\caption[]{Kriging maps for Ancient Greek (Indo-European, Eurasia), Wandala (Afro-Asiatic, Africa), Chumburung (Atlantic-Congo, Africa), Tii (Austronesian, Papunesia)}
\end{figure}

Similarly to what we noticed with A-pattern languages, among pattern D-languages, the scope of the ML Kriging area is quite varied. For several of these, the ML area has scope on part of the BL area as well, where competition among different means is, as already observed, generally more intense cross-linguistically. Ancient Greek (Figure \ref{proielgrc0}) is one such language, as well as, for example, Wandala (Figure \ref{wandala}), Chumburung (Figure \ref{chumburung}), and Tii (Figure \ref{tii}), among several others.

\begin{figure}[!h]
\begin{subfigure}{0.50\textwidth}
\includegraphics[width=0.9\linewidth]{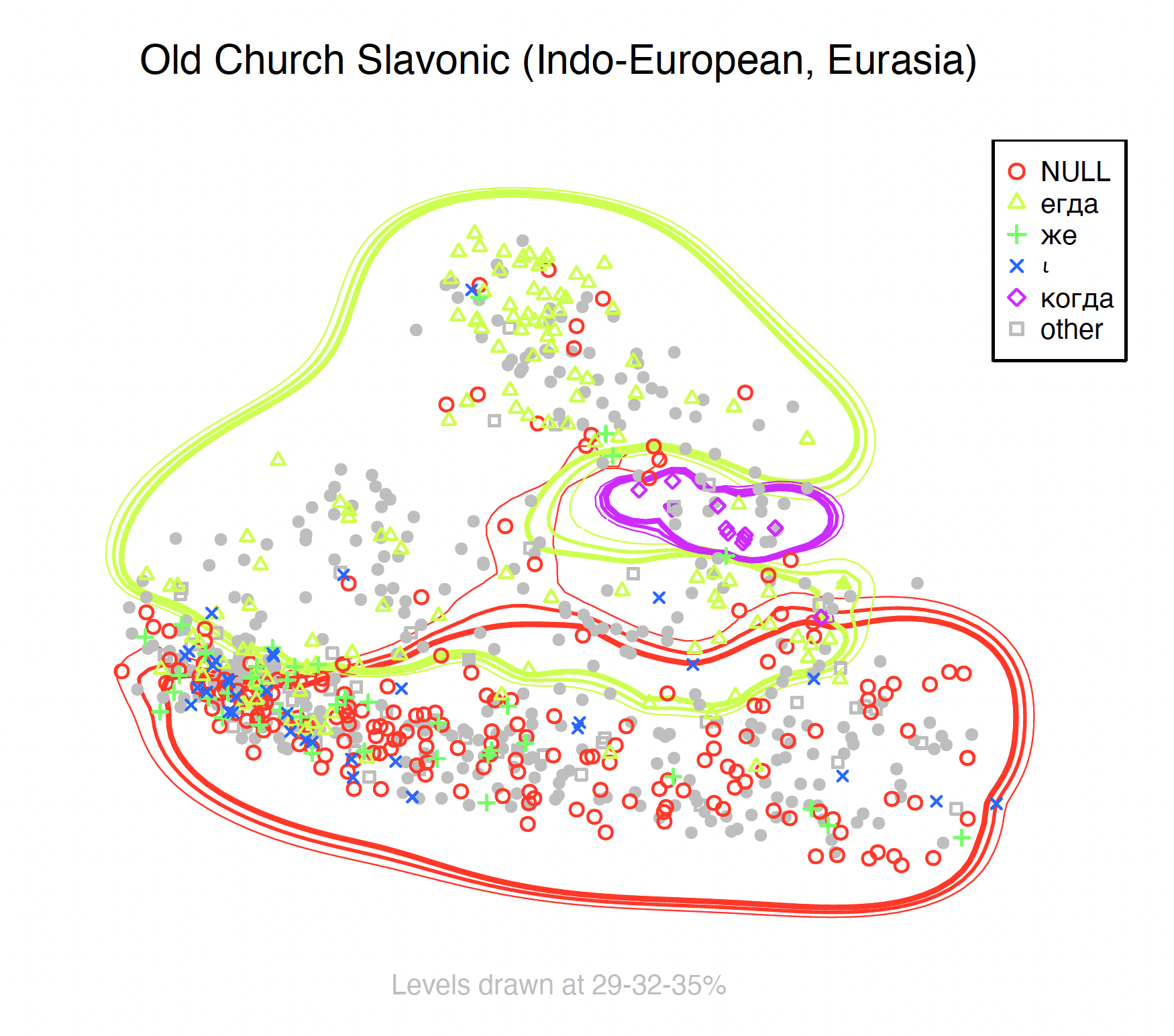} 
\caption[Kriging map for Old Church Slavonic (Indo-European, Eurasia)]{}
\label{proielchu0}
\end{subfigure}
\begin{subfigure}{0.50\textwidth}
\includegraphics[width=0.9\linewidth]{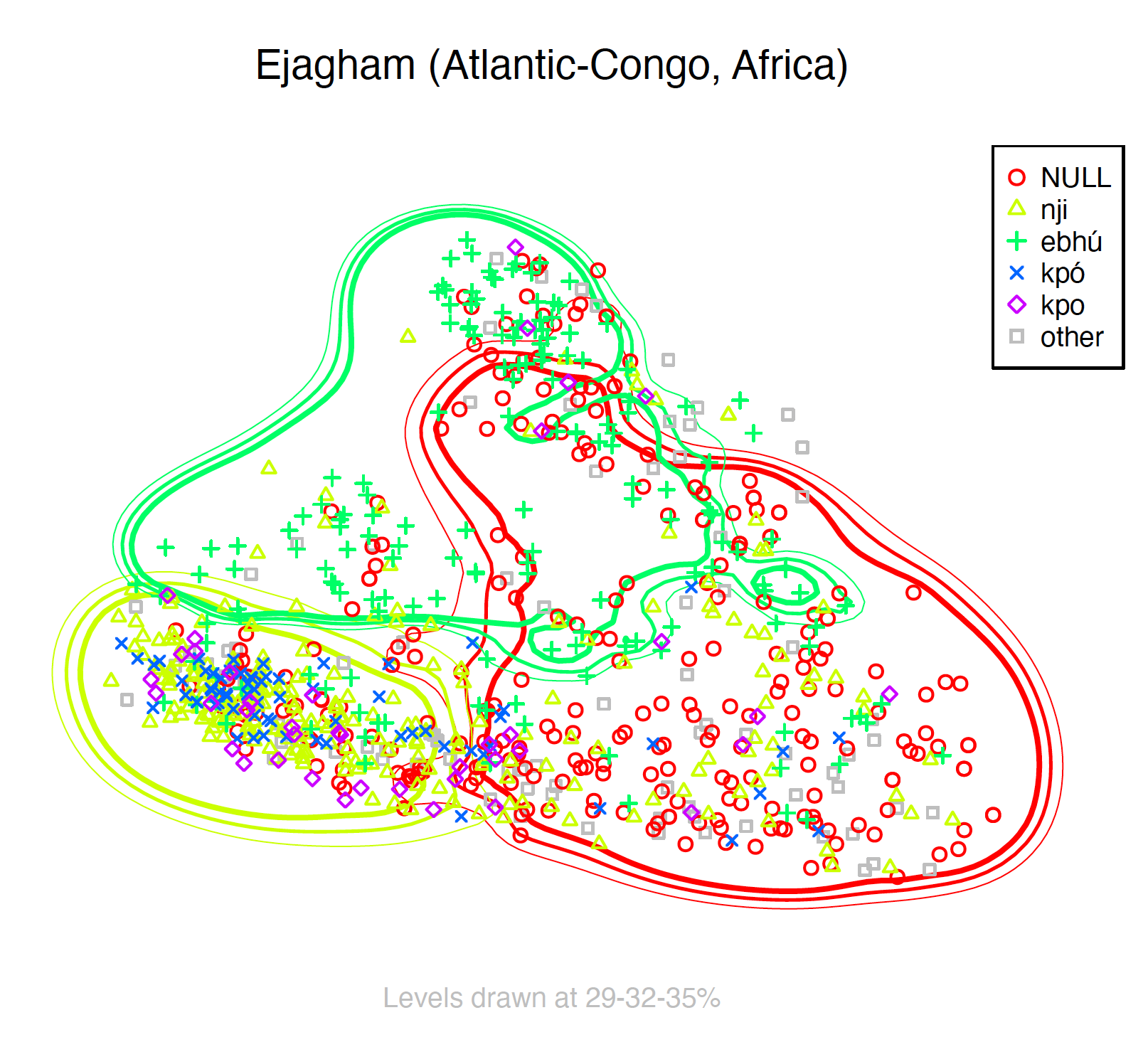}
\caption[Kriging map for Eigham (Atlantic-Congo, Africa)]{}
\label{ejgham}
\end{subfigure}
\begin{subfigure}{0.50\textwidth}
\includegraphics[width=0.9\linewidth]{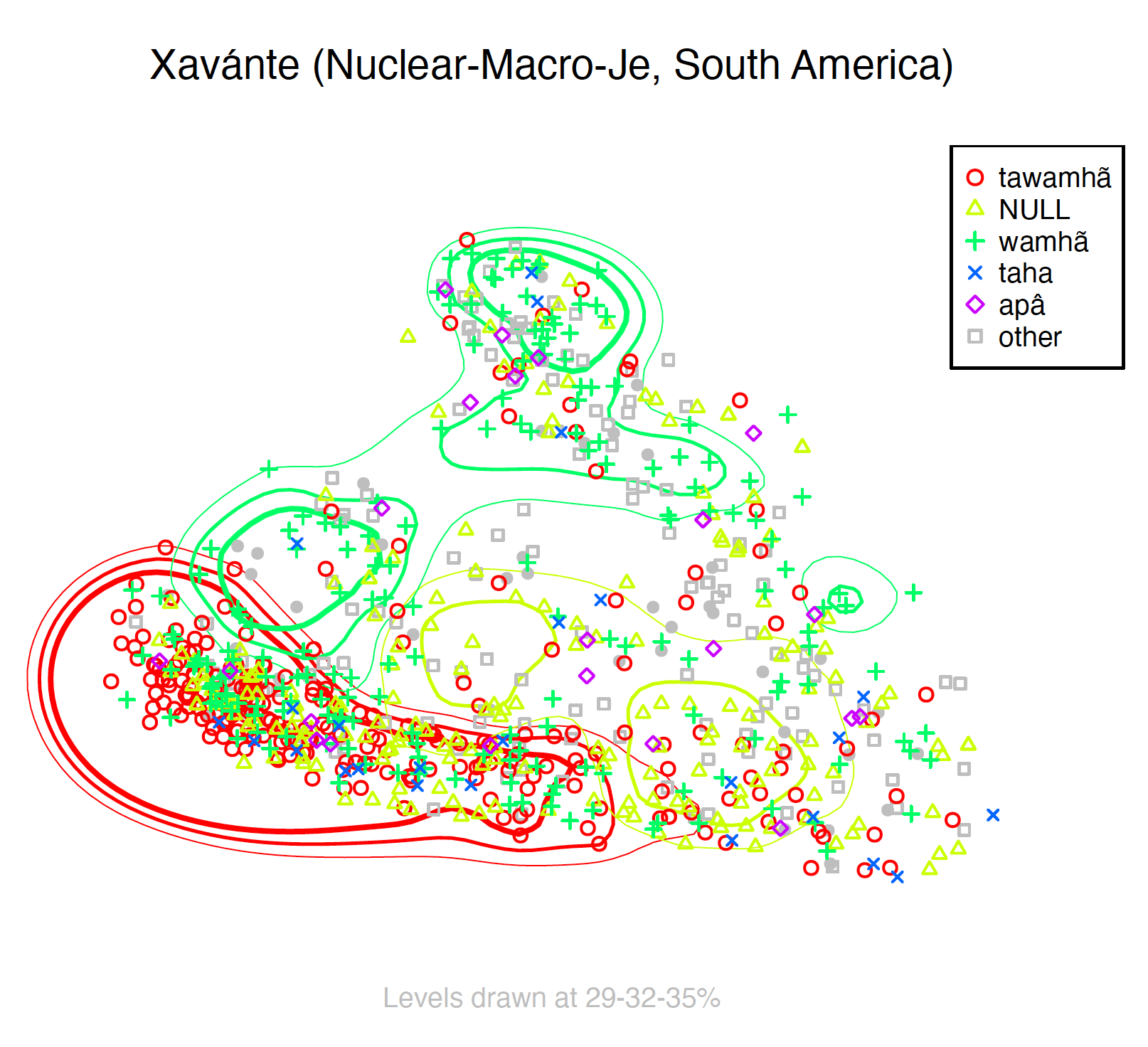}
\caption[Kriging map for Xavánte (Nuclear-Macro-Je, South America)]{}
\label{xavante}
\end{subfigure}
\begin{subfigure}{0.50\textwidth}
\includegraphics[width=0.9\linewidth]{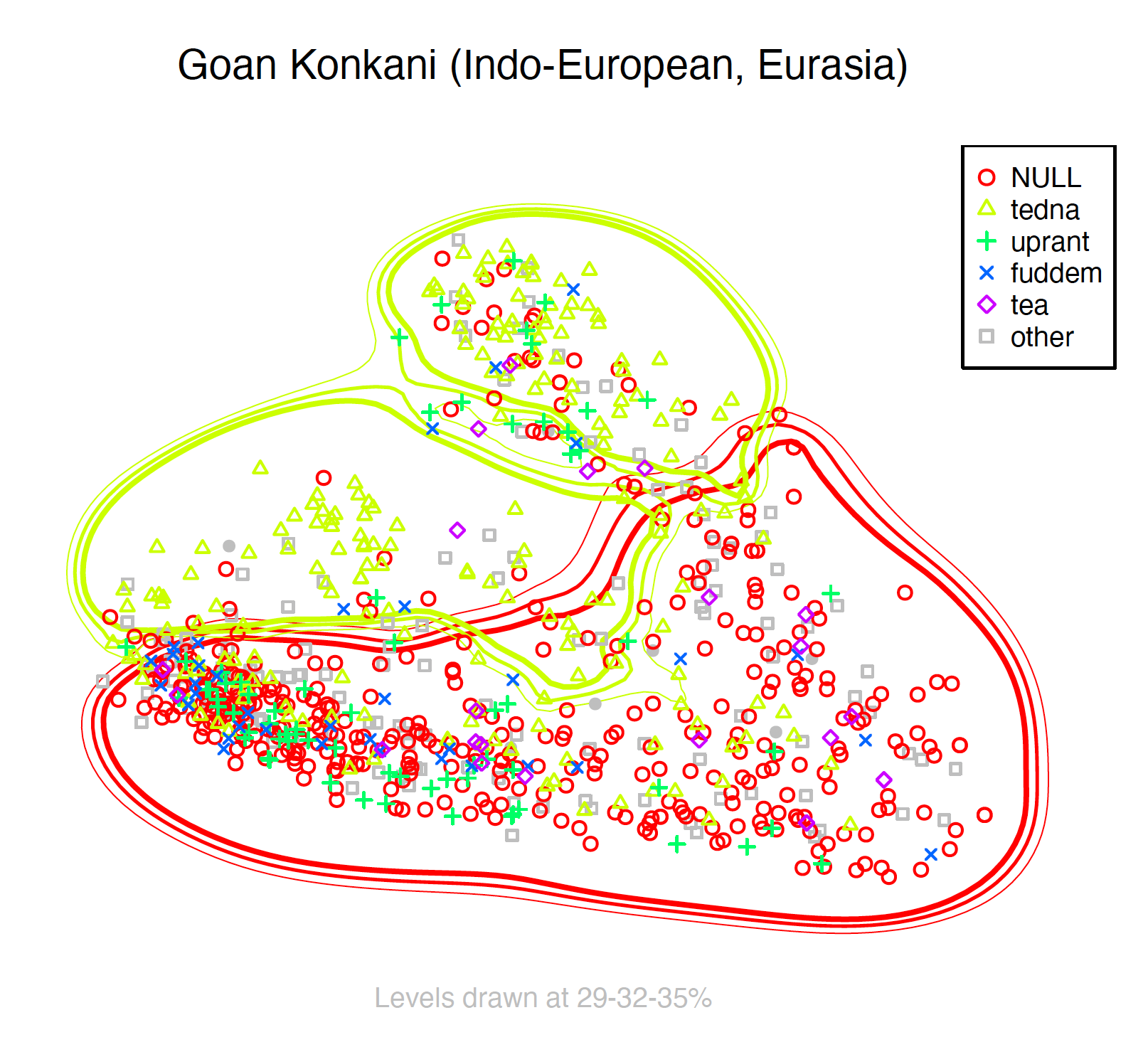}
\caption[Kriging map for Goan Konkani (Indo-European, Eurasia)]{}
\label{goankonkani}
\end{subfigure}
\caption[]{Kriging maps for Old Church Slavonic (Indo-European, Eurasia), Eigham (Atlantic-Congo, Africa), Xavánte (Nuclear-Macro-Je, South America), Goan Konkani (Indo-European, Eurasia)}
\end{figure}

Patterns B and C are, in a way, asymmetrical. That is, they differ in whether the ML area is colexified with the TL or the BL area. In pattern-B languages, the TL area is colexified with the ML area. Old Church Slavonic (Figure \ref{proielchu0}) belongs to this pattern, with the TL and ML areas having \textit{jegda} as the predominant Kriging area, while the BL area is dominated by null constructions.\footnote{At the top-right of the Old Church Slavonic map, we can see a small Kriging area (13 occurrences) for \textit{kogda}, which is `when' as a question word, as in: \textit{\textbf{kogda} že tę viděchom\foreignlanguage{russian}{ъ} bolęšta ili v\foreignlanguage{russian}{ь} tem\foreignlanguage{russian}{ь}nici i pridom\foreignlanguage{russian}{ъ} k\foreignlanguage{russian}{ъ} tebě} `and when did we see you sick or in prison and visit you?' (Matthew 25:39). \textit{When} as a question word is not analyzed separately in this thesis.} Other examples are Eigham (Figure \ref{ejgham}), Xavánte (Figure \ref{xavante}), and Goan Konkani (Figure \ref{goankonkani}).

\begin{figure}[!h]
\begin{subfigure}{0.50\textwidth}
\includegraphics[width=0.9\linewidth]{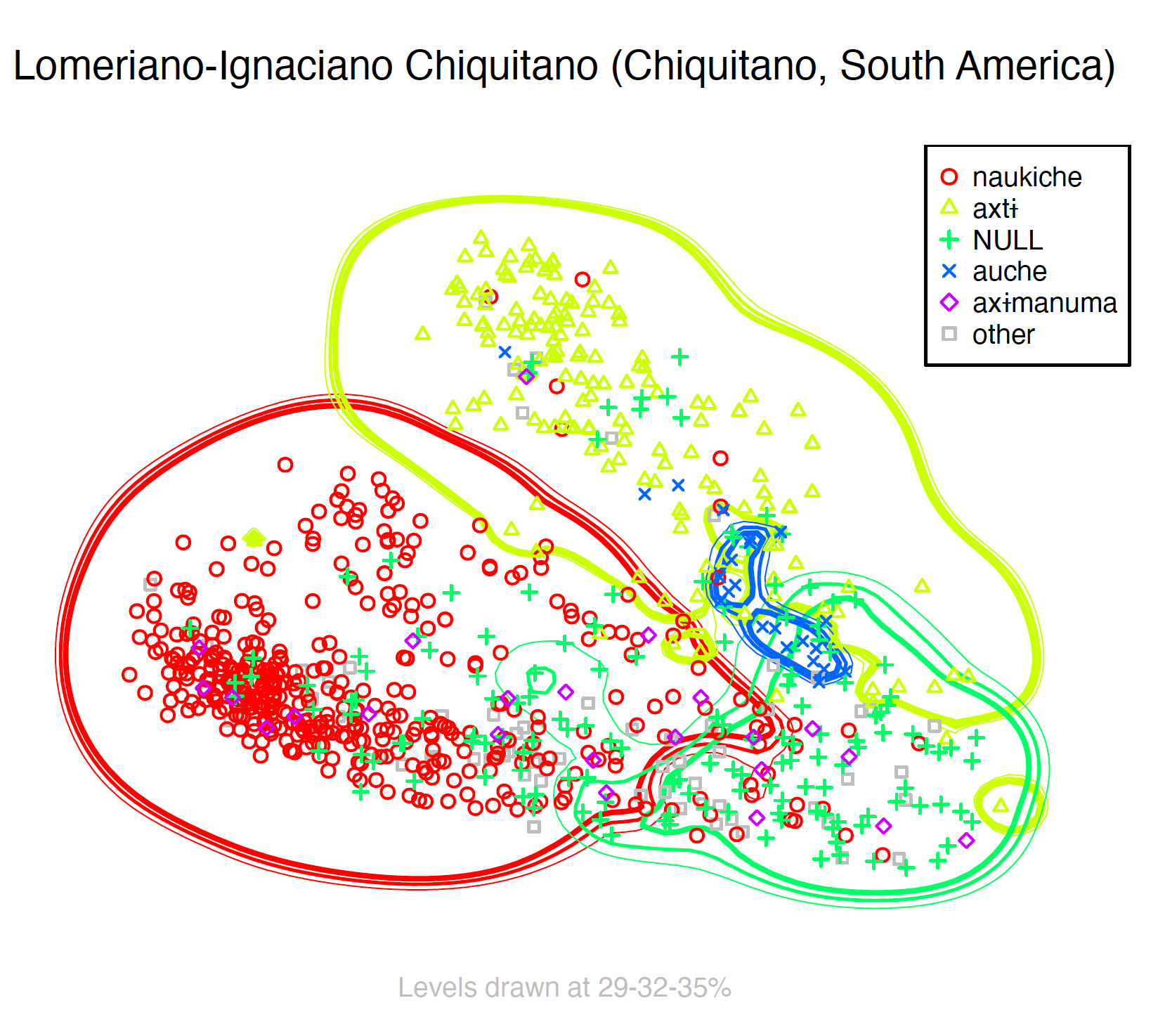} 
\caption[Kriging map for Lomeriano-Ignaciano Chiquitano (Chiquitano, South America)]{}
\label{lomeriano}
\end{subfigure}
\begin{subfigure}{0.50\textwidth}
\includegraphics[width=0.9\linewidth]{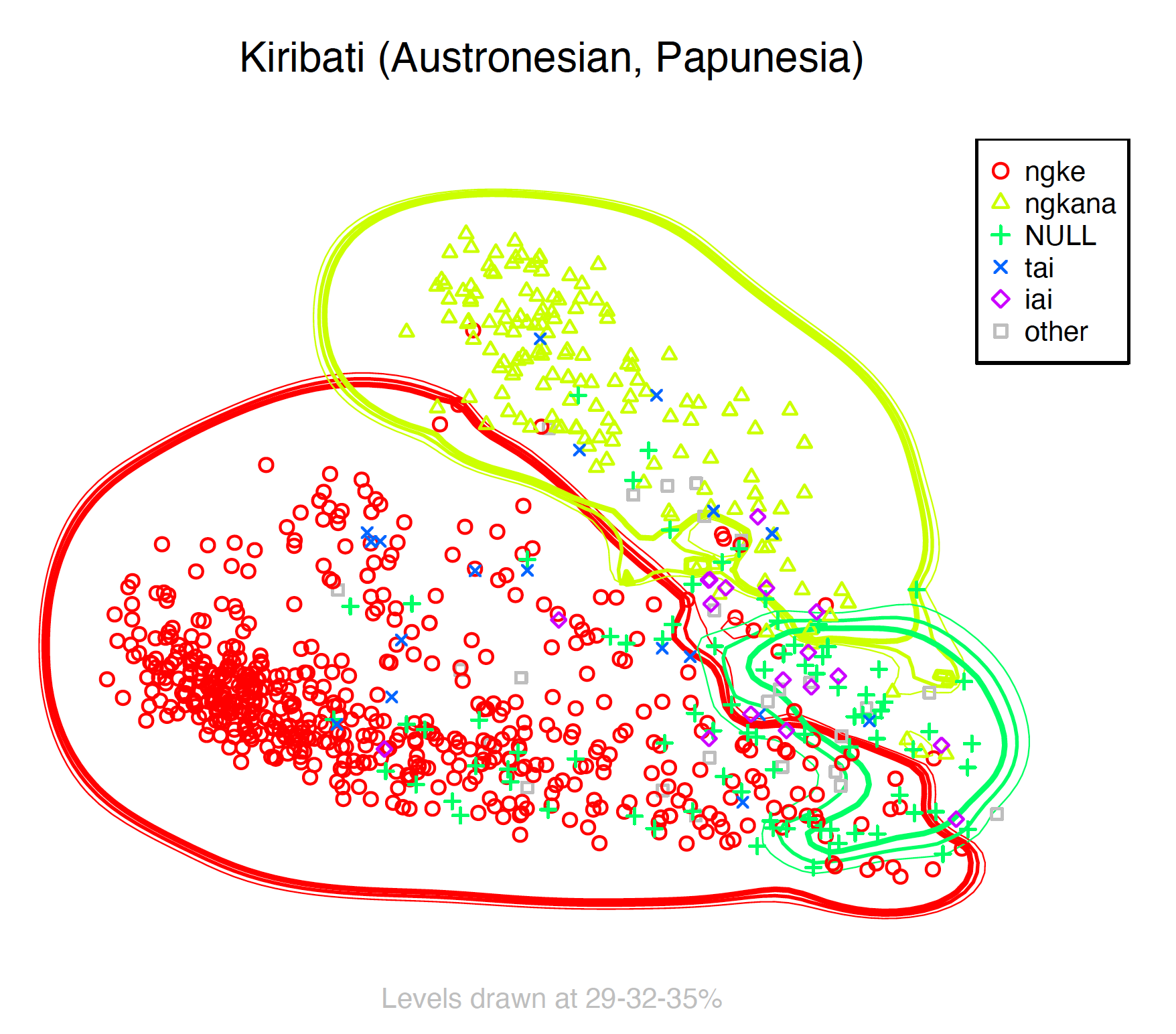}
\caption[Kriging map for Kiribati (Austronesian, Papunesia)]{}
\label{kiribati}
\end{subfigure}
\begin{subfigure}{0.50\textwidth}
\includegraphics[width=0.9\linewidth]{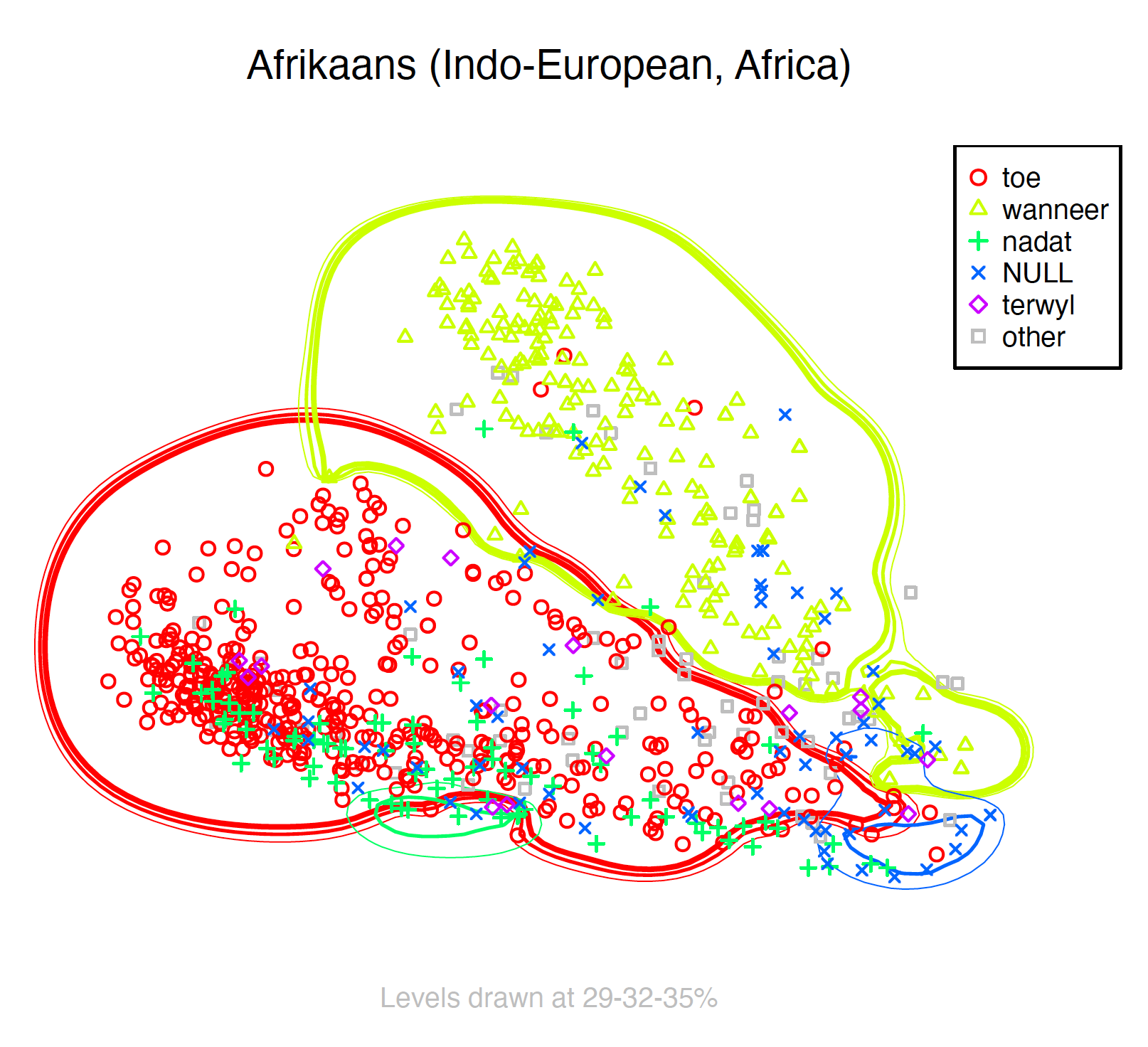}
\caption[Kriging map for Afrikaans (Indo-European, Africa)]{}
\label{afrikaans}
\end{subfigure}
\begin{subfigure}{0.50\textwidth}
\includegraphics[width=0.9\linewidth]{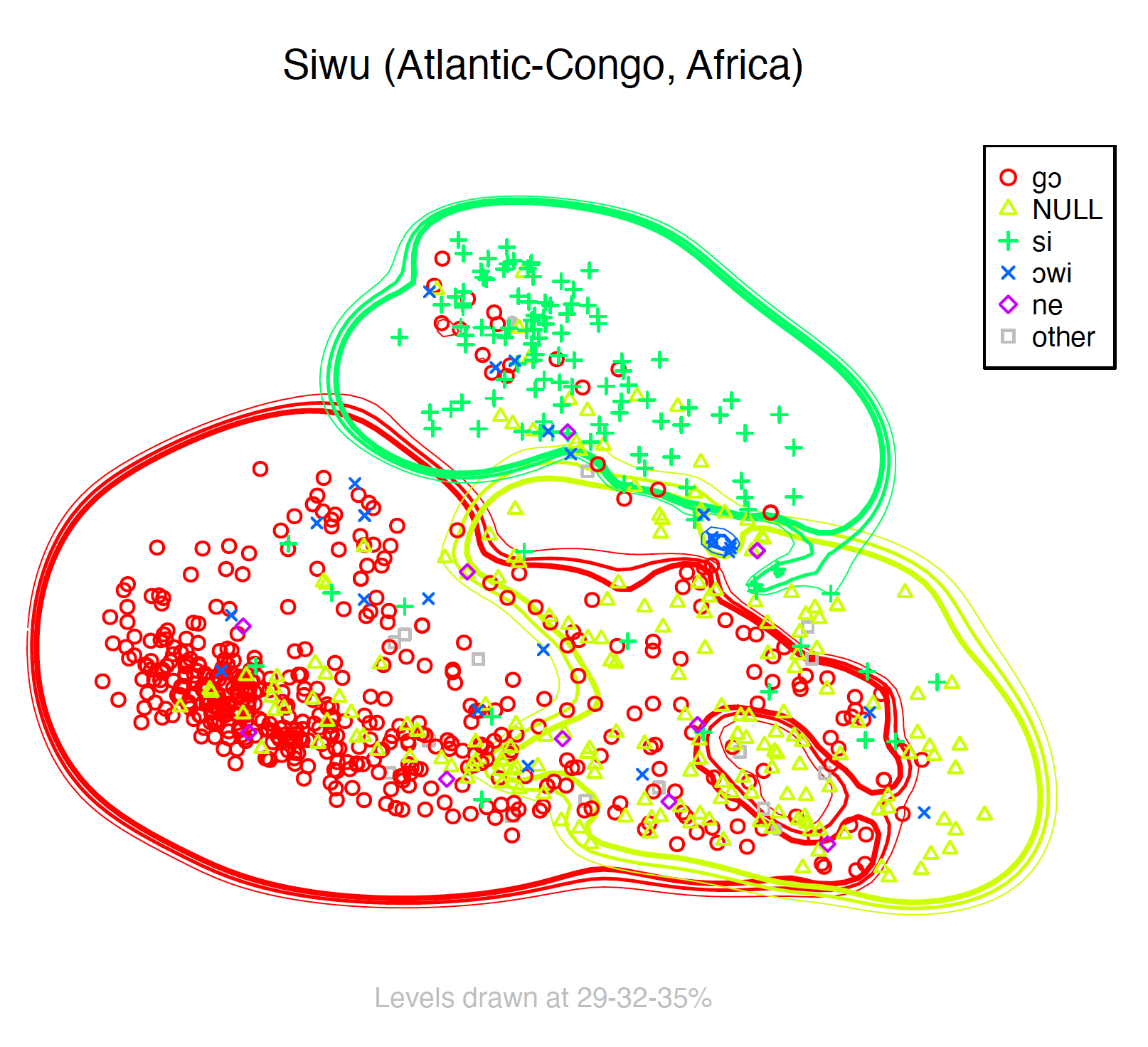}
\caption[Kriging map for Siwu (Atlantic-Congo, Africa)]{}
\label{siwu}
\end{subfigure}
\caption[]{Kriging maps for Lomeriano-Ignaciano Chiquitano (Chiquitano, South America), Kiribati (Austronesian, Papunesia), Afrikaans (Indo-European, Africa), Siwu (Atlantic-Congo, Africa)}
\end{figure}

Patter C is the second most-frequent pattern among the world's languages. Some examples from this very common pattern are Lomeriano-Ignaciano Chiquitano (Figure \ref{lomeriano}), Kiribati (Figure \ref{kiribati}) and Siwu (Figure \ref{siwu}). It also includes most Germanic languages, for example Afrikaans (Figure \ref{afrikaans}). 

Like the Germanic languages in this group, pattern-C languages are likely to have one means for expressing the meaning of \textsc{when} to refer to repeated events in past, present or future, as well as for singular events in the future, and one means for expressing singular events in the past. In other words, they have one means for \textit{universal} \textsc{when} irrespective of tense, which is also used for \textit{existential} \textsc{when} in the future, and one means for \textit{existential} \textsc{when} in the past (cf. \citealt{sabo2011}; \pgcitealt{saboeHansen}{2}). This contrast can be exemplified by the minimal pair in the German example in (\ref{deuexistvsuniv}).

\begin{example}
\begin{itemize}
    \item[a.]
    \gll \textbf{Als} ich ins Bett ging, konnte ich nicht einschlafen
    when.\textsc{exist} I in bed went, could I not sleep
    \glt
    \glend
    \item[b.]
    \gll \textbf{Wenn} ich ins Bett ging, konnte ich nicht einschlafen
    when.\textsc{univ} I in bed went, could I not sleep
    \glt
    \glend
\end{itemize}
\label{deuexistvsuniv}
\end{example}

In the Afrikaans example in Figure \ref{afrikaans}, the TL area is under the scope of \textit{wanneer} `when' (universal and nonpast existential), whereas the ML/BL area is under the scope of \textit{toe} `when' (past existential).\\
\indent It is relatively uncommon for pattern C languages to have a null Kriging area in the TL cluster (59 out of 277 languages), compared to pattern-B languages, where that is the case in about half of them (84 out of 171 languages).\\
\indent Overall, it seems exceedingly rare for languages with only one Kriging area besides nulls to use it exclusively to encode the situations in the BL cluster. On the other hand, it is very common to find languages with a dedicated Kriging area for the TL cluster only. As a hypothetical generalization, if languages make extensive use of null constructions and if they have one subjunction that is equivalent to \textit{when}, then it seems very likely that it will be used most typically for the situations in the TL area without much competition with null constructions. This is what we observe in the Old Church Slavonic and, albeit with two rather than one lexified means, in the Ancient Greek map. The observations in the TL area appear to be more clearly under the domain of the \textit{when}-equivalents available in the language (i.e. \textit{jegda} and \textit{hótan}) than the ML and BL areas are. In Ancient Greek and Old Church Slavonic, the Kriging area for \textit{hóte}/\textit{jegda} and the one for null constructions show a high degree of overlap, suggesting that the competition between finite temporal subordinates and participle constructions, which we can expect to be predominant within their null areas, is particularly clear in between the ML and the BL areas. 
\newpage

\subsection{Overall precision and recall for the GMM clusters}
\begin{figure}[!h]
\centering
\begin{tabular}{cc}
  \includegraphics[width=72mm]{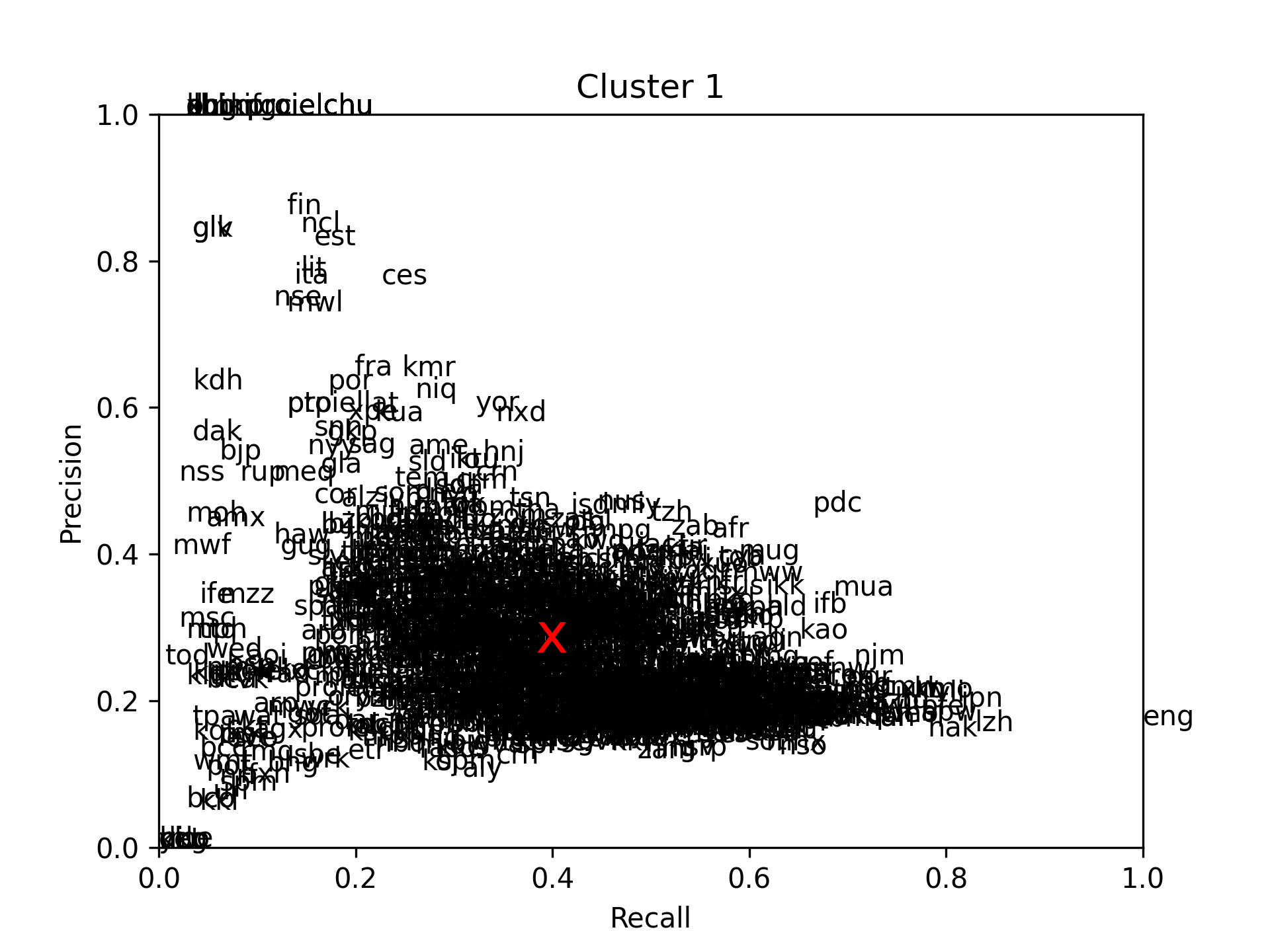} &   \includegraphics[width=72mm]{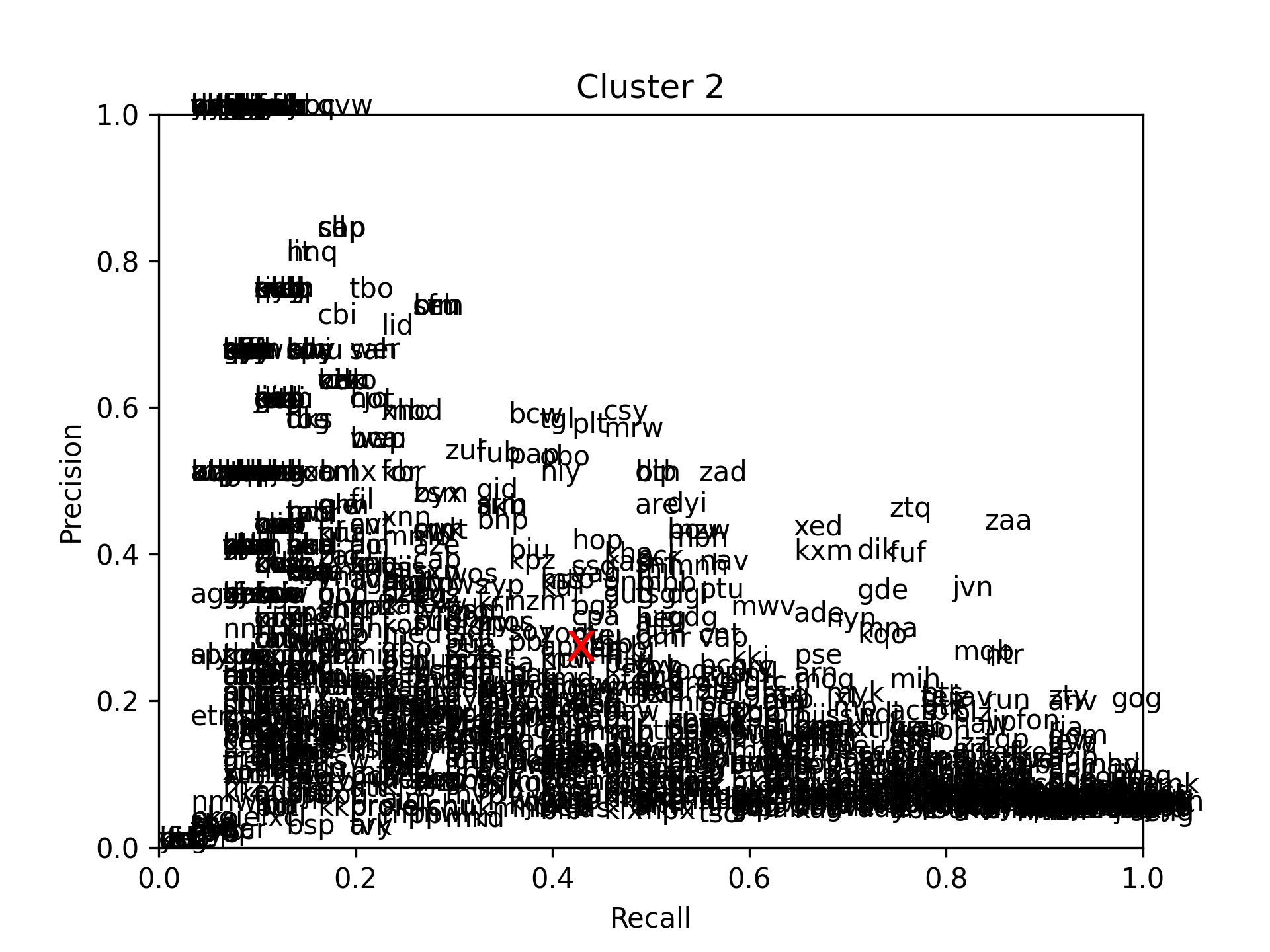}\\
 \includegraphics[width=72mm]{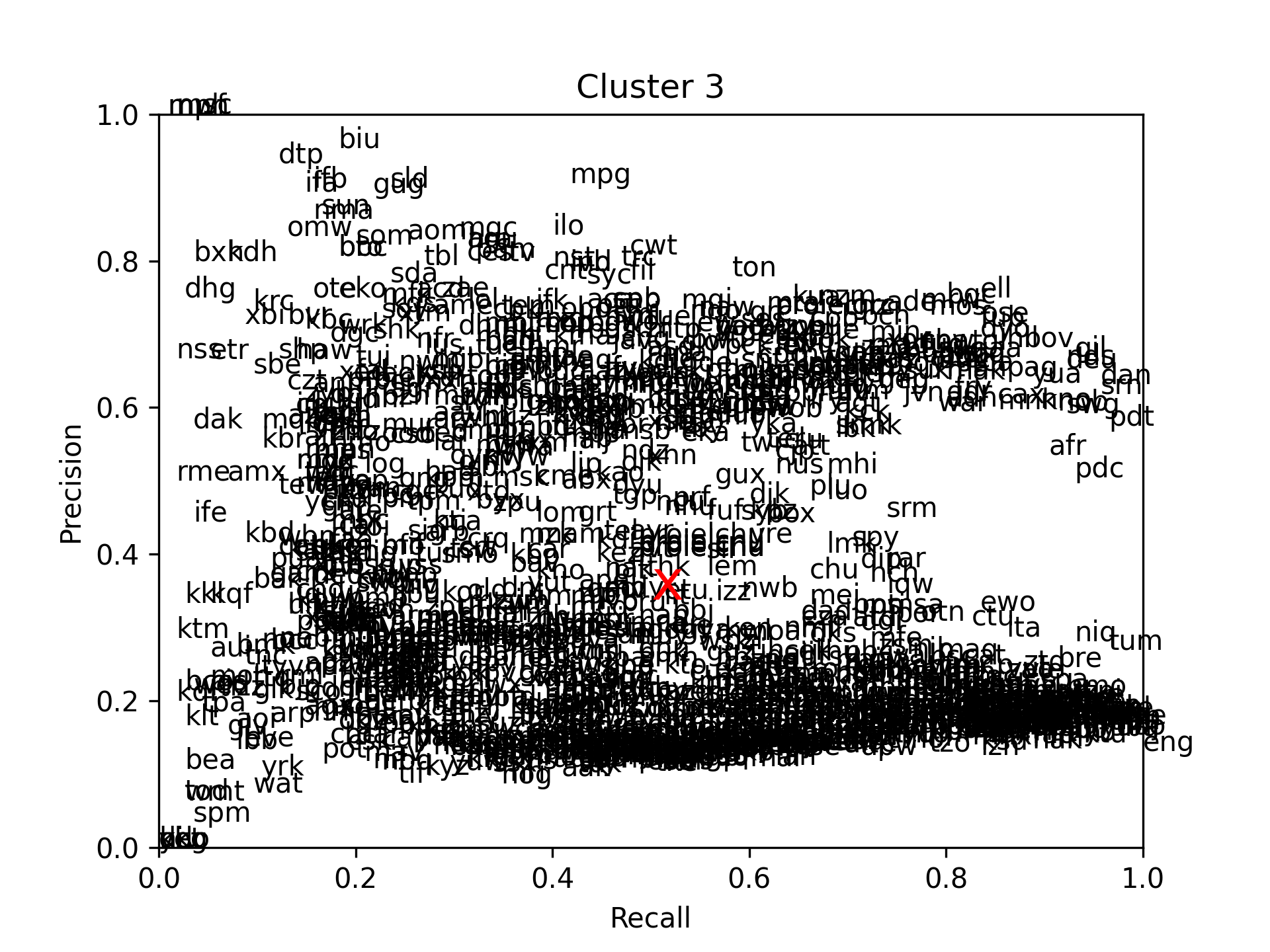} &   \includegraphics[width=72mm]{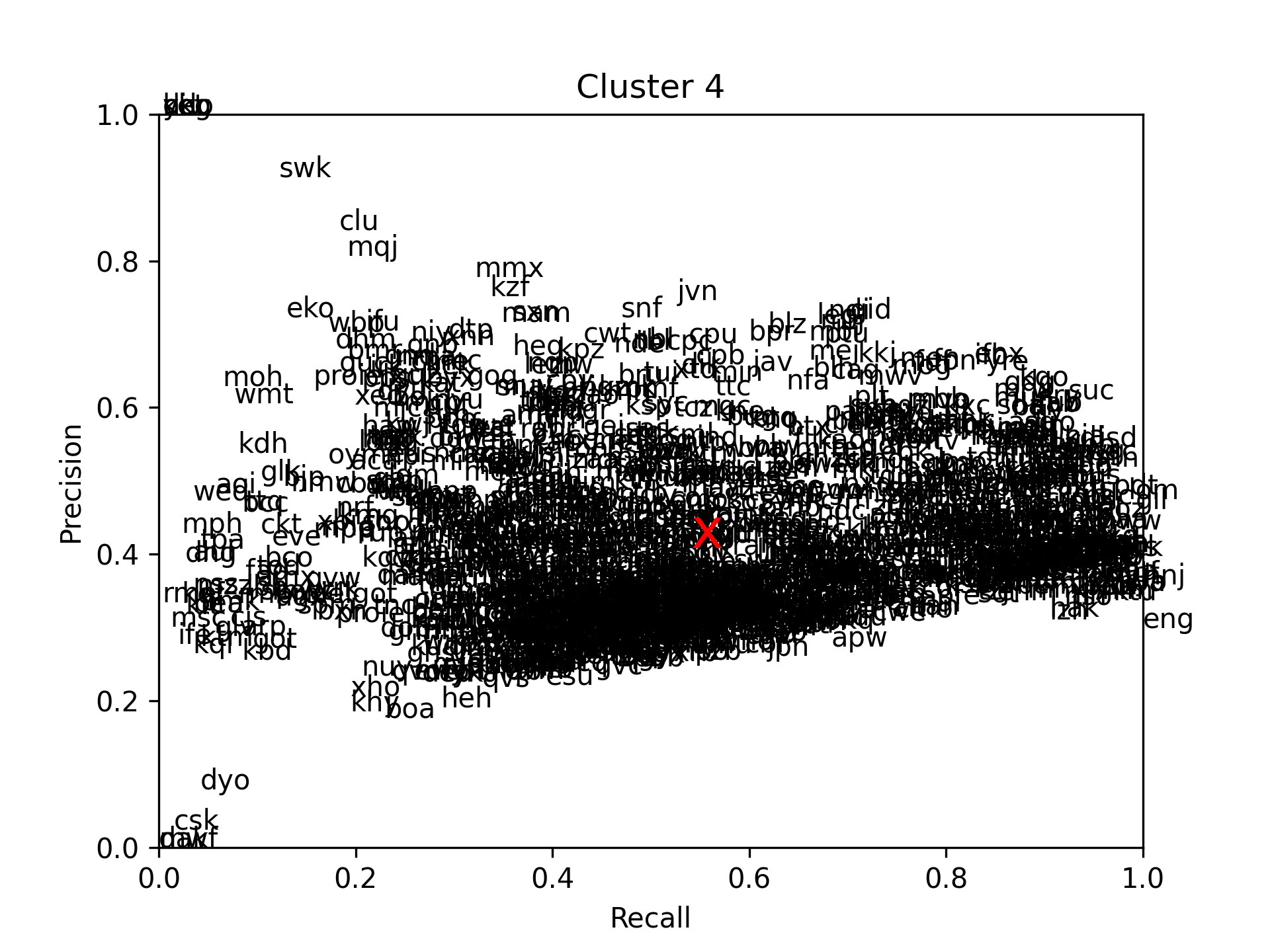} \\
  \includegraphics[width=72mm]{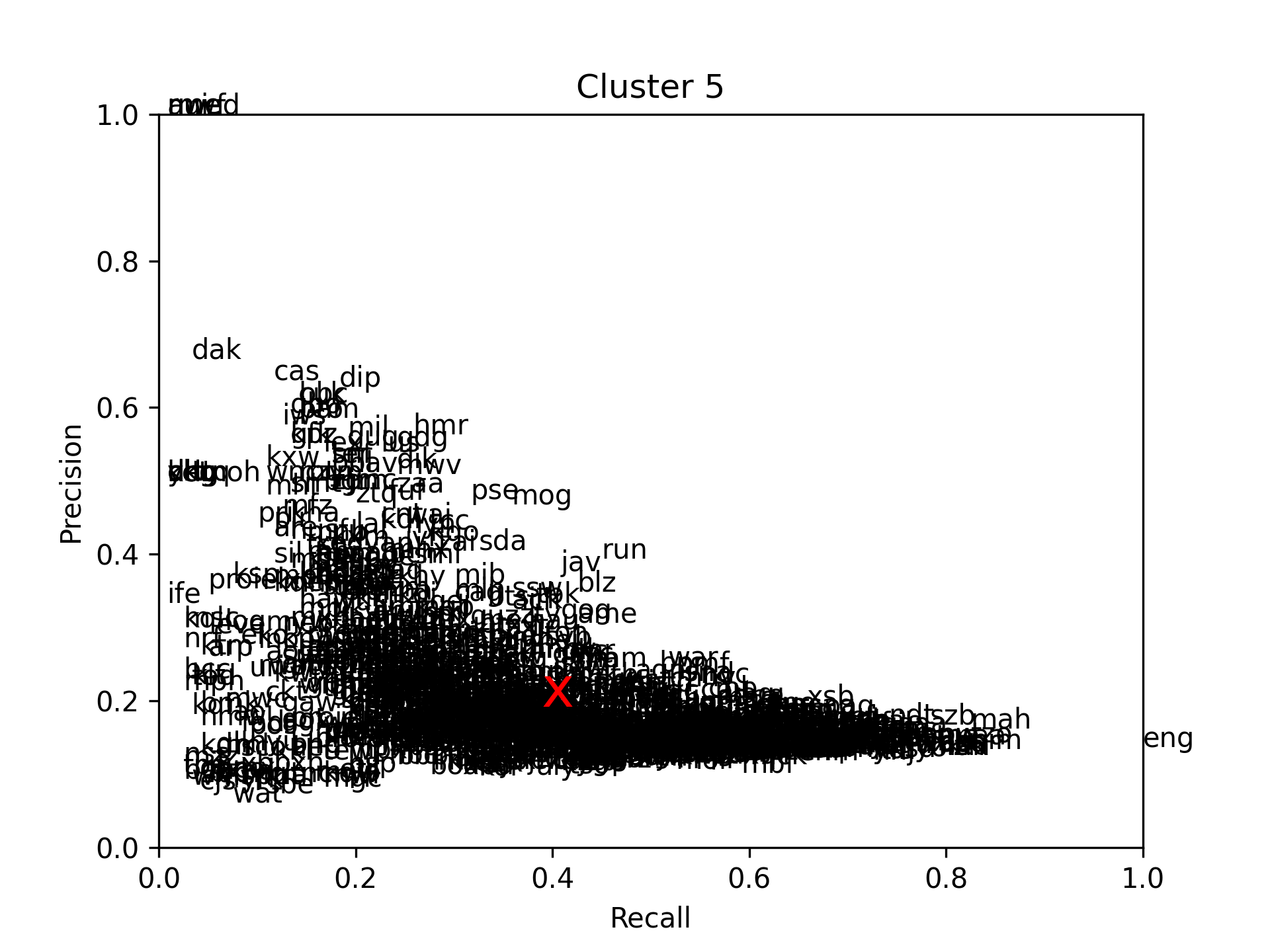} &   \includegraphics[width=72mm]{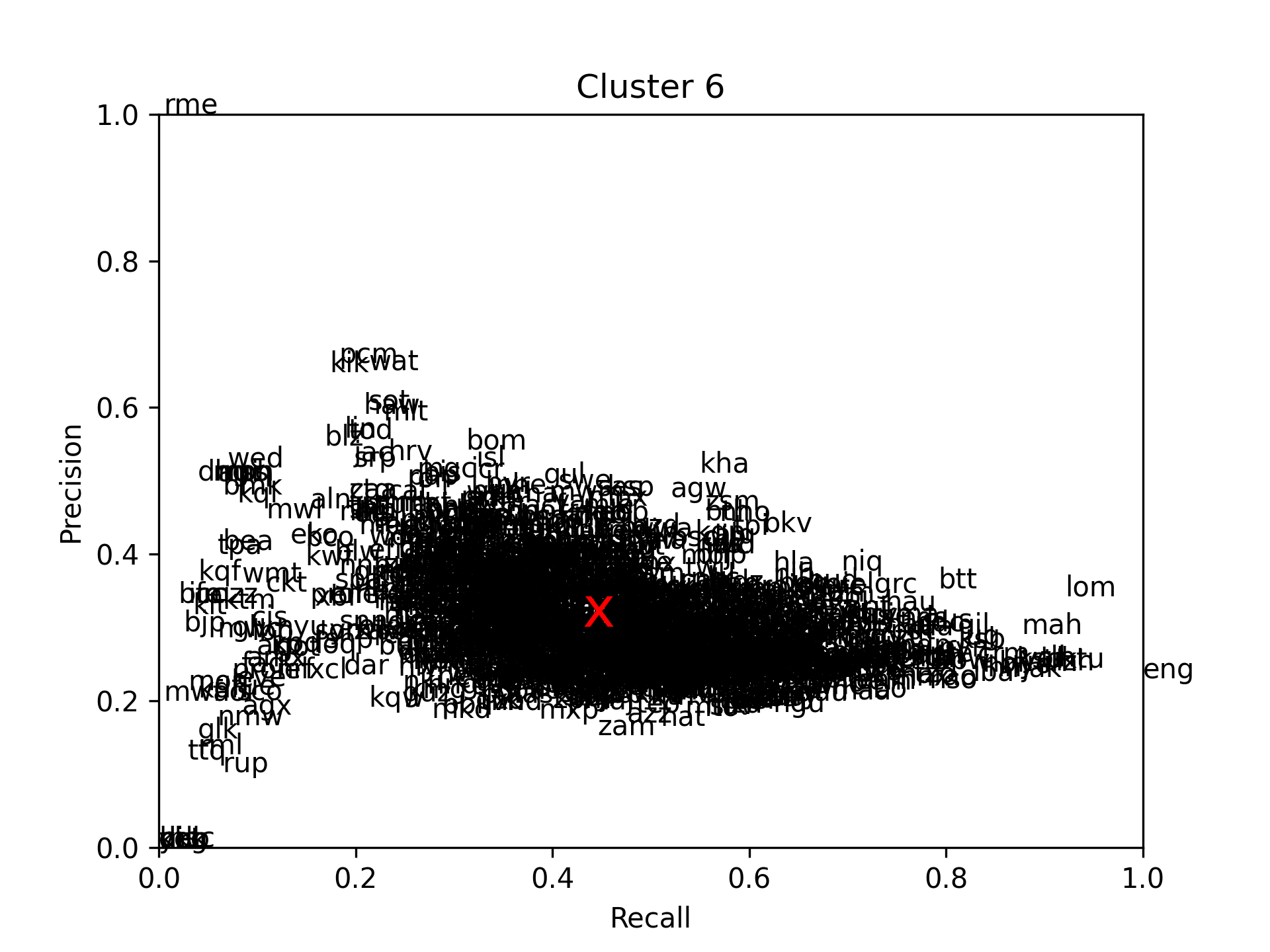} \\
\end{tabular}
\caption[Precision and recall plots for the six GMM clusters]{Precision and recall plots for the six GMM clusters shown in Figure \ref{gmm6}. The $x$ and $y$ values (recall and precision, respectively) for each language in each of the subplots correspond to the recall and precision of the item with the highest F1 score in that language for that cluster (compared to all other items in that language occurring at least once within that cluster). The red marker \textit{x} indicates the mean precision and recall value.}
    \label{fig:gmm-corres}
\end{figure}

As explained in the Methods section, we can look for potential cross-linguistic lexification patterns by checking how well each Kriging area in each language fit each of the GMM cluster. Recall that, to do so, we count how many observations within a Kriging area are found inside a given GMM cluster, how many outside of the same GMM cluster, as well as how many observations not from other Kriging areas (or not belonging to any Kriging area) are found inside that same GMM cluster. These numbers are then used as true positives, true negatives and false positives, respectively, which we can then use to calculate precision and recall for each Kriging area. The highest-scoring Kriging area is then considered the best match for that GMM cluster in the language. That is, if a language has one Kriging area yielding a high harmonic mean (F1 score) with respect to a particular GMM cluster, that will indicate that it has a particular means which is relatively specialized for whichever meaning or function of \textsc{when} that GMM cluster corresponds to. The higher the F1 score, the more specialized the means. \\
\indent The results of this method are shown in Figure~\ref{fig:gmm-corres}. Note that a high precision score will not necessarily correspond to a likely lexification of that GMM cluster if the recall is very low (i.e. close to 0)---most likely, these are rare items (possibly due to misalignment by the model), most or all of which happen to be distributed inside the GMM cluster under consideration. High recall with low precision is instead likely to indicate that the scope of the means in question is broader than the GMM cluster under consideration. English \textit{when}, for example, will show a recall of 1 for all clusters, since the \textit{whole} semantic map is under its scope. What we are looking for is, therefore, both high precision and high recall (i.e. a high F1 score).

\begin{figure}[!h]
\begin{subfigure}{0.50\textwidth}
\includegraphics[width=0.9\linewidth]{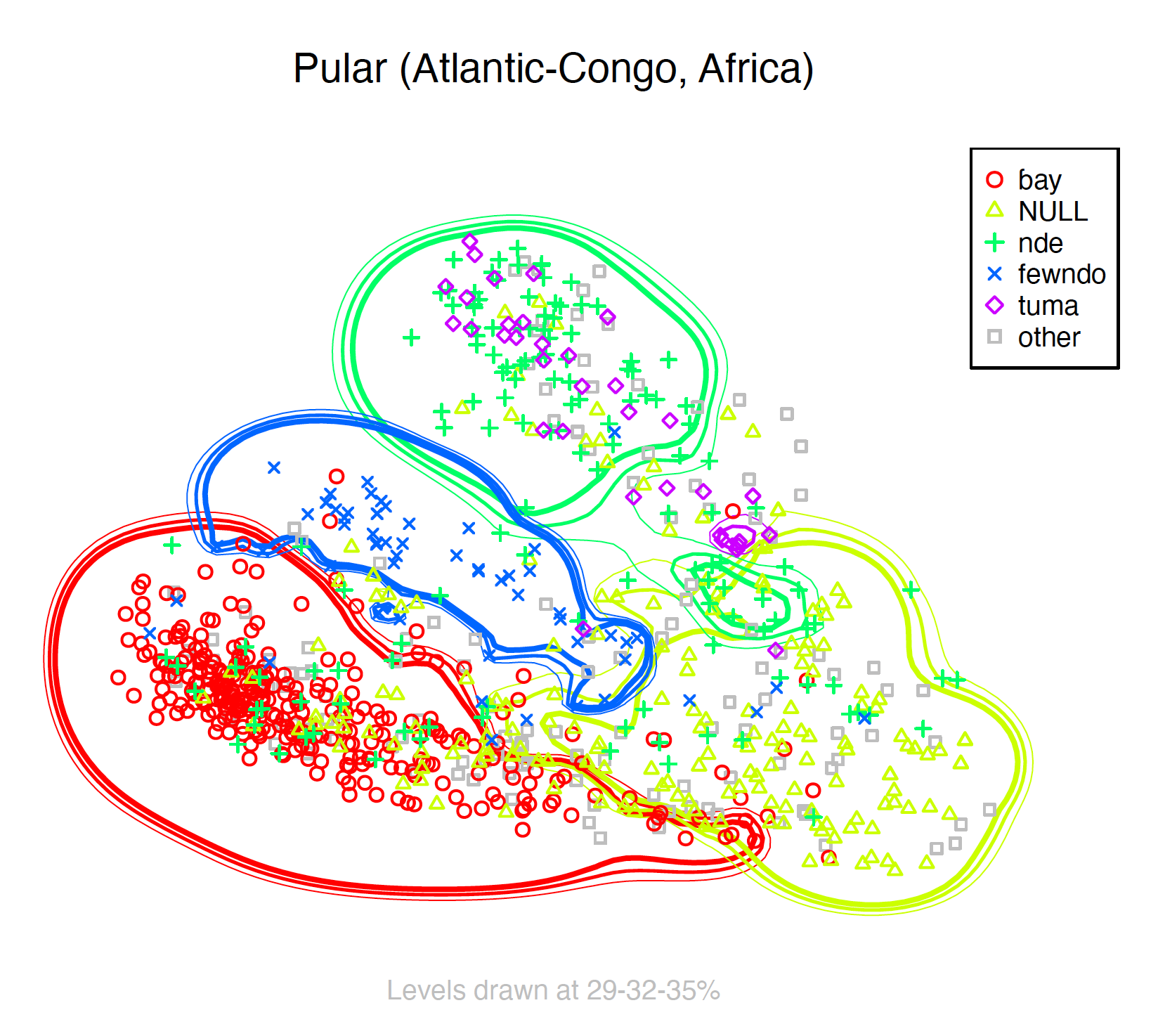}
\caption[Kriging map for Pular (Atlantic-Congo, Africa)]{}
\label{pular}
\end{subfigure}
\begin{subfigure}{0.50\textwidth}
\includegraphics[width=0.9\linewidth]{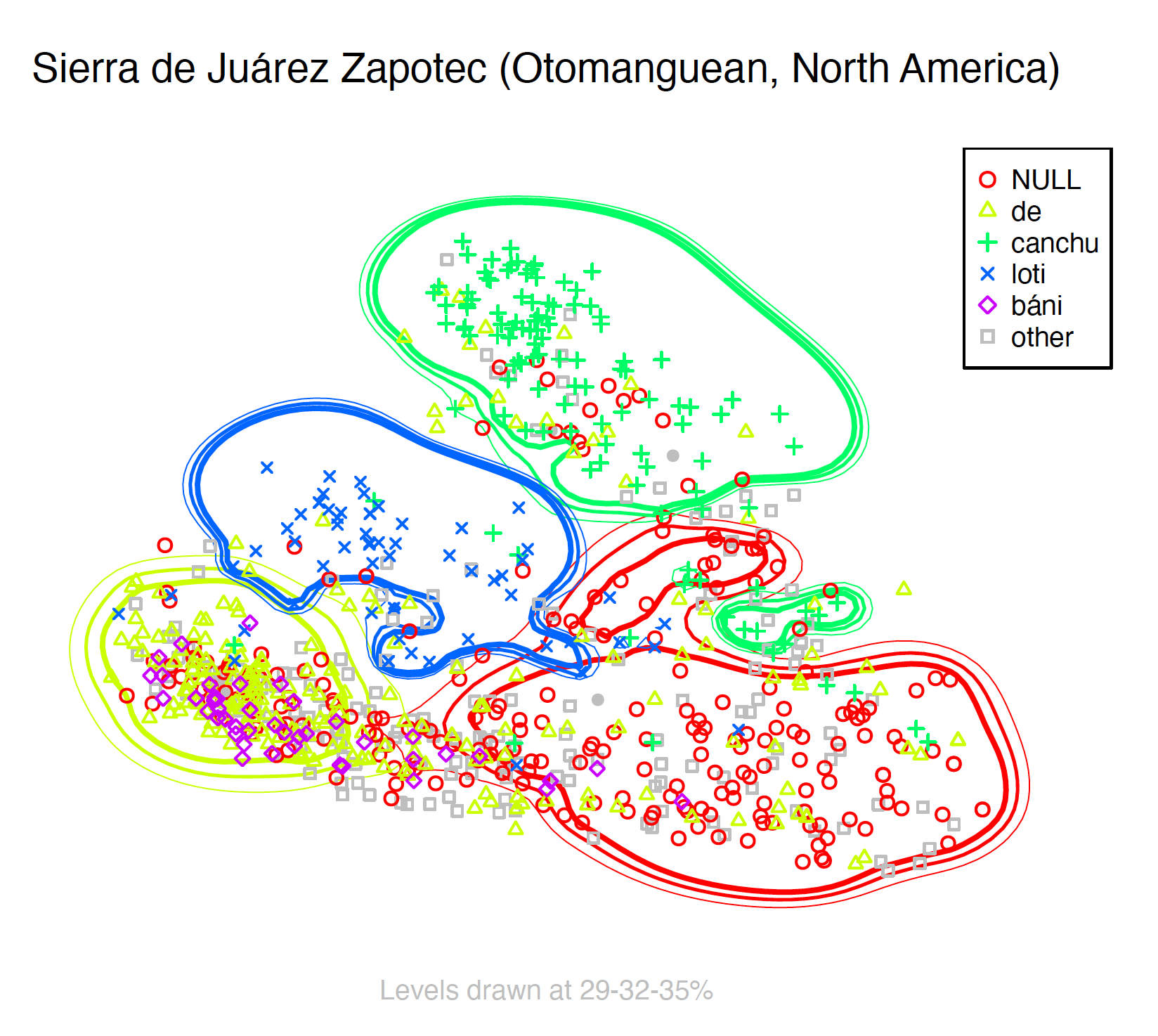}
\caption[Kriging map for Sierra de Juárez Zapotec (Otomanguean, North America)]{}
\label{zapotec}
\end{subfigure}
\caption[]{Kriging maps for Pular (Atlantic-Congo, Africa) and Sierra de Juárez Zapotec (Otomanguean, North America)}
\end{figure}

The potential generalization regarding common patterns of usage of overtly subordinated constructions as opposed to null constructions finds some degree of confirmation in the different F1 scores for the different GMM clusters.\\ 
\indent The results for GMM clusters 1, 5 and 6 do not immediately suggest any pattern. Languages in these plots cluster quite densely, with little variation between languages, and both mean precision and recall, indicated in the plots by a red marker \textit{x}, are low. Both the lack of visible differences between languages and the very low average precision and recall suggest that our method is not capturing cross-linguistic differences nor a clearly well-defined function from the observations in this cluster.\\
\indent Cluster 2, which corresponds to the left extremity of the ML cluster commented on in the previous section (cf. Figure \ref{threeway}), also has both low mean precision and recall, but this time we see a handful of languages with high recall ($\ge$ 0.6) and slightly higher precision ($\ge$ 0.4) than most other languages with the same recall. A closer look at some of them confirms that these are mostly languages with a relatively well-defined Kriging area for the ML cluster, clearer than languages such as Ancient Greek (Figure \ref{proielgrc0}) or Chumburung (Figure\ref{chumburung}), for example, which also have a separate Kriging area for the ML cluster but perhaps more intense competition with other means, or a wider scope extending to the BL area as well (e.g. as we saw in the Ancient Greek map, among others). Compare the clearer boundaries of the Kriging area for the ML cluster in Figure \ref{pular} and \ref{zapotec}, for Pular and Sierra de Juárez Zapotec, with those in Figures \ref{proielgrc0}-\ref{tii} seen earlier on.

\begin{figure}[!h]
    \centering
    \includegraphics[scale=0.2]{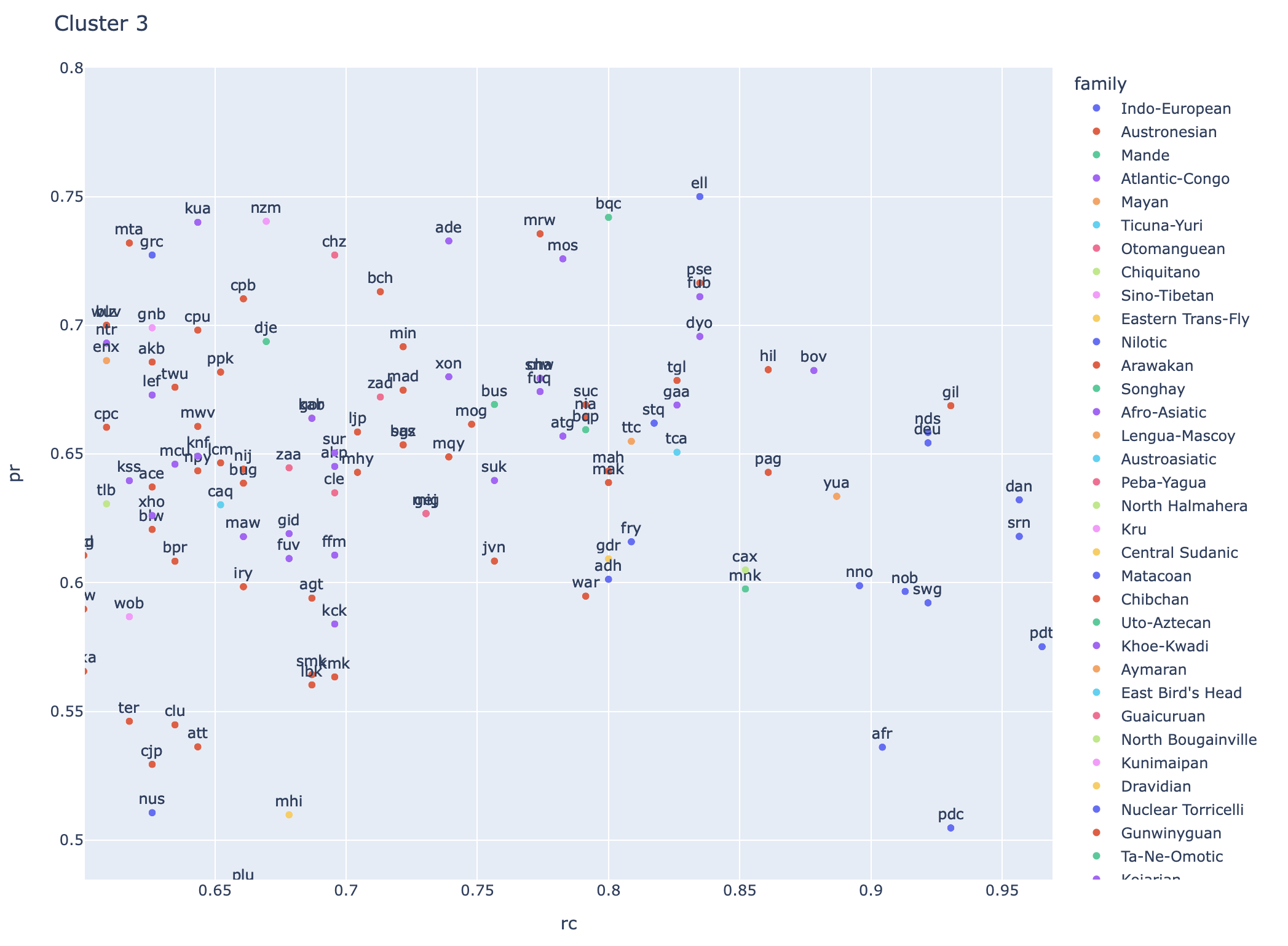}
    \caption{The top-scoring languages (by precision and recall) for GMM cluster 3}
    \label{fig:bestmatches3}
\end{figure}

The clearest pattern comes from the plot for GMM cluster 3, corresponding to the TL area commented on in the previous section. Languages divide into two bands, one with lower precision (between 0.1 and 0.3) and one with higher precision ($\ge$ 0.5), both stretching across the whole range of recall from 0 to 1. The clear pattern for this cluster suggests that GMM cluster 3 is likely to approximate a real gram type and that it is very common for this gram type to be expressed by some specialized means cross-linguistically.\\
\indent Figure \ref{fig:bestmatches3} shows the best matches to cluster 3, with a precision ranging between 0.5 to 0.75, and a recall between 0.6 all the way to nearly 1.

We see that some of the best matches are found, among others, across a range of Germanic languages, in particular Danish [dan] and Norwegian [nob/nno] \textit{når}, and German [deu] \textit{wenn} (also cf. Kiribati [gil], represented in Figure \ref{kiribati} as well). As we mentioned above, these languages have lexified the distinction between universal and existential readings in the past, with one subordinator generally serving for universal readings in any tense, as well as nonpast existential meanings, and another subordinator for past existential readings only.\\
\indent Another interesting pattern is the one obtained from GMM cluster 4 (i.e. the BL cluster). Here we do not get any clear division into bands among the world's languages. Instead, most languages show overall relatively high precision, but few as high as the upper band or as low as the lower band observed for cluster 3. Both mean precision and recall are, in fact, slightly higher than the one for cluster 3. On the one hand, this suggests that, with this method, we are not capturing cross-linguistic differences in the way languages express the situations in cluster 4, as is the case with cluster 3. On the other hand, however, it also points to the fact that it is very common for individual languages to express the situations in this area with one main means, but even in the presence of such means, there might be more competition with other constructions. Given what we have seen above regarding the distribution of null constructions across the world's languages, namely that they are more and more likely to occur the more we look at the situations to the bottom right of the map, a possibility is that cluster 4 is more likely to have null constructions as competitors to some other means. 

\begin{figure}[!h]
    \centering
    \includegraphics[scale=0.2]{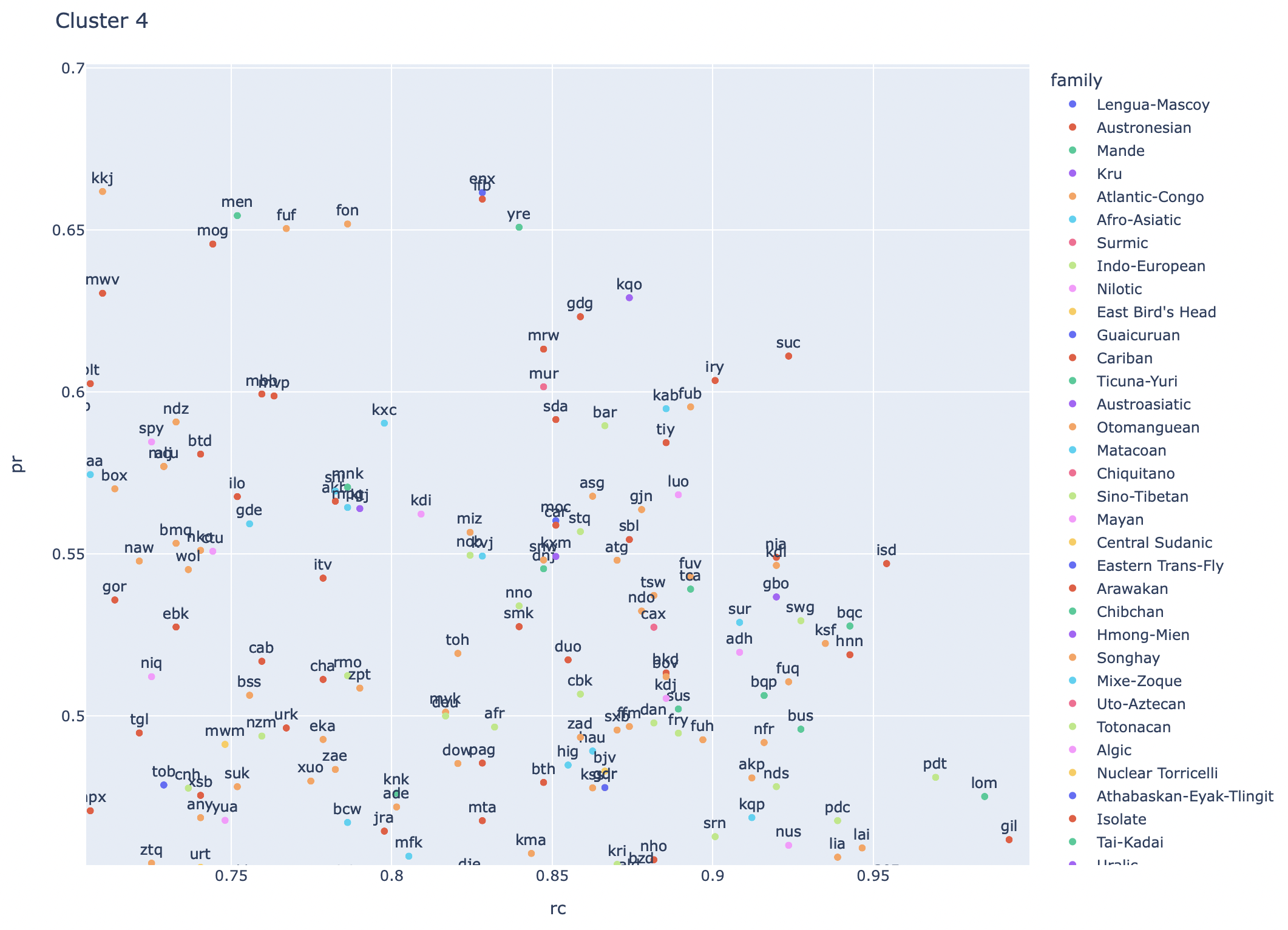}
    \caption{The top-scoring languages (by precision and recall) for GMM cluster 4}
    \label{fig:bestmatches4}
\end{figure}

\begin{figure}[!h]
\begin{subfigure}{0.50\textwidth}
\includegraphics[width=0.9\linewidth]{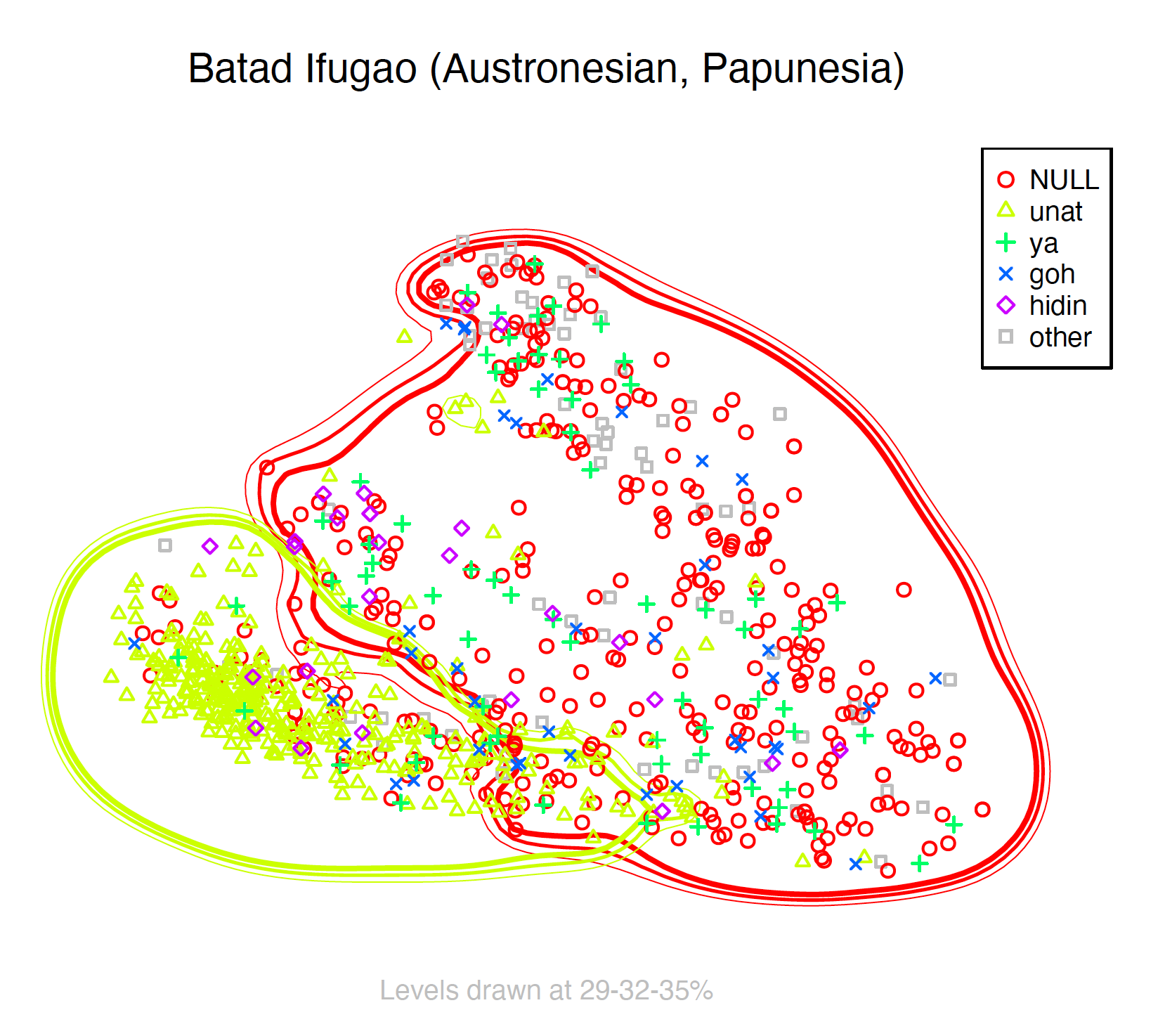}
\caption[Kriging map for Batad Ifugao (Austronesian, Papunesia)]{}
\label{batadifugao}
\end{subfigure}
\begin{subfigure}{0.50\textwidth}
\includegraphics[width=0.9\linewidth]{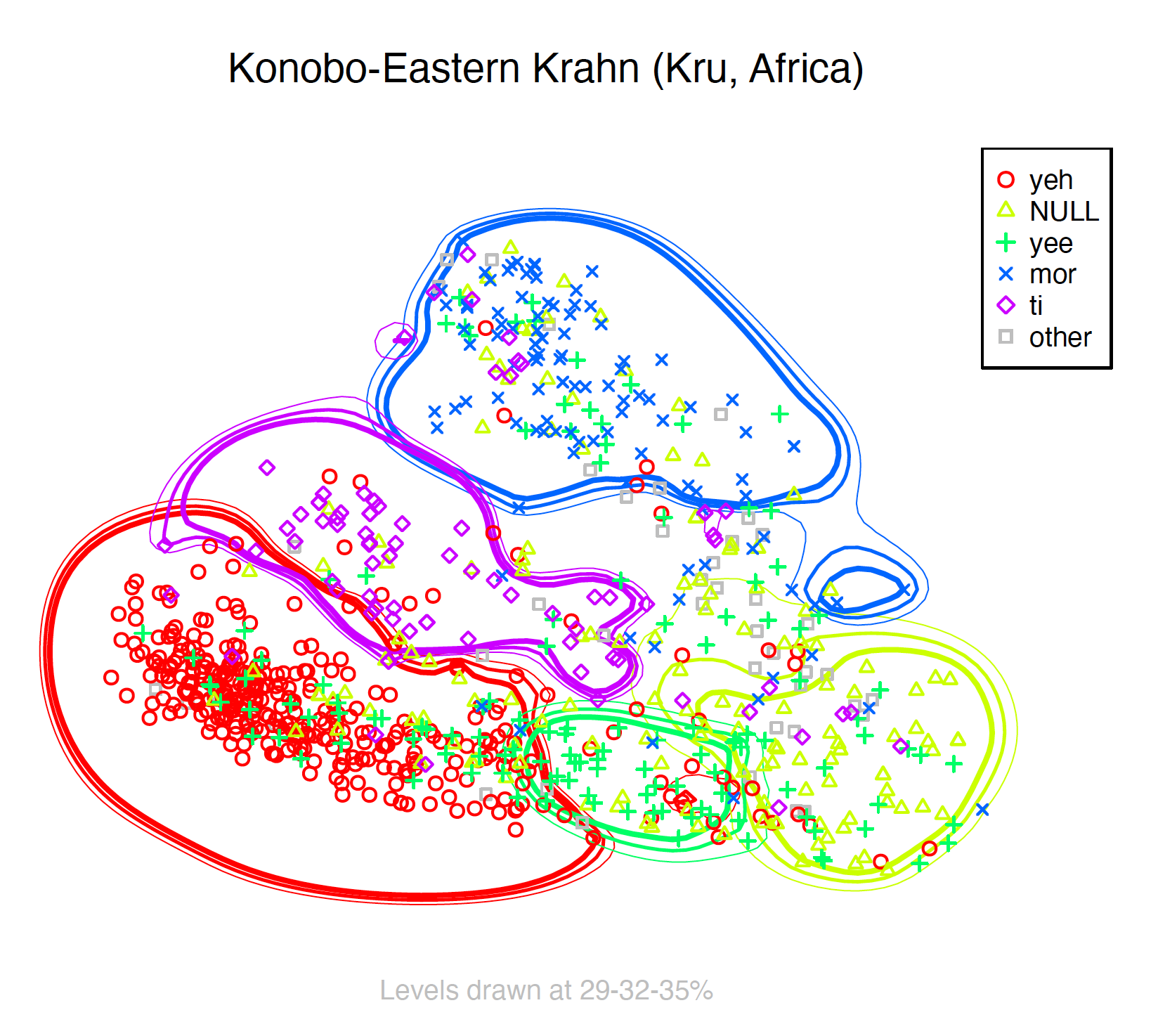}
\caption[Kriging map for Konobo-Eastern Krahn (Kru, Africa)]{}
\label{konobokrahn}
\end{subfigure}
\caption[]{Kriging maps for Batad Ifugao (Austronesian, Papunesia) and Konobo-Eastern Krahn (Kru, Africa)}
\end{figure}

Figure \ref{fig:bestmatches4} shows the best matches to cluster 4, with a precision ranging between 0.45 to 0.7, and a recall between 0.7 all the way to nearly 1. Languages at the higher range of precision have quite well-defined Kriging areas for the BL area, with relatively limited competition, considering some of the more extreme cases seen earlier on (cf. Sierra de Juárez Zapotek, Figure \ref{zapotec}, as well as all the maps in Figures \ref{proielgrc0}-\ref{tii} and Figures \ref{proielchu0}-\ref{goankonkani}). Some of these examples are, once again, Pular (Figure \ref{pular}), as well as, for example, Batad Ifugao (Figure \ref{batadifugao}) and Konobo-Eastern Krahn (Figure \ref{konobokrahn}).

\begin{figure}[!h]
\begin{subfigure}{0.50\textwidth}
\includegraphics[width=0.9\linewidth]{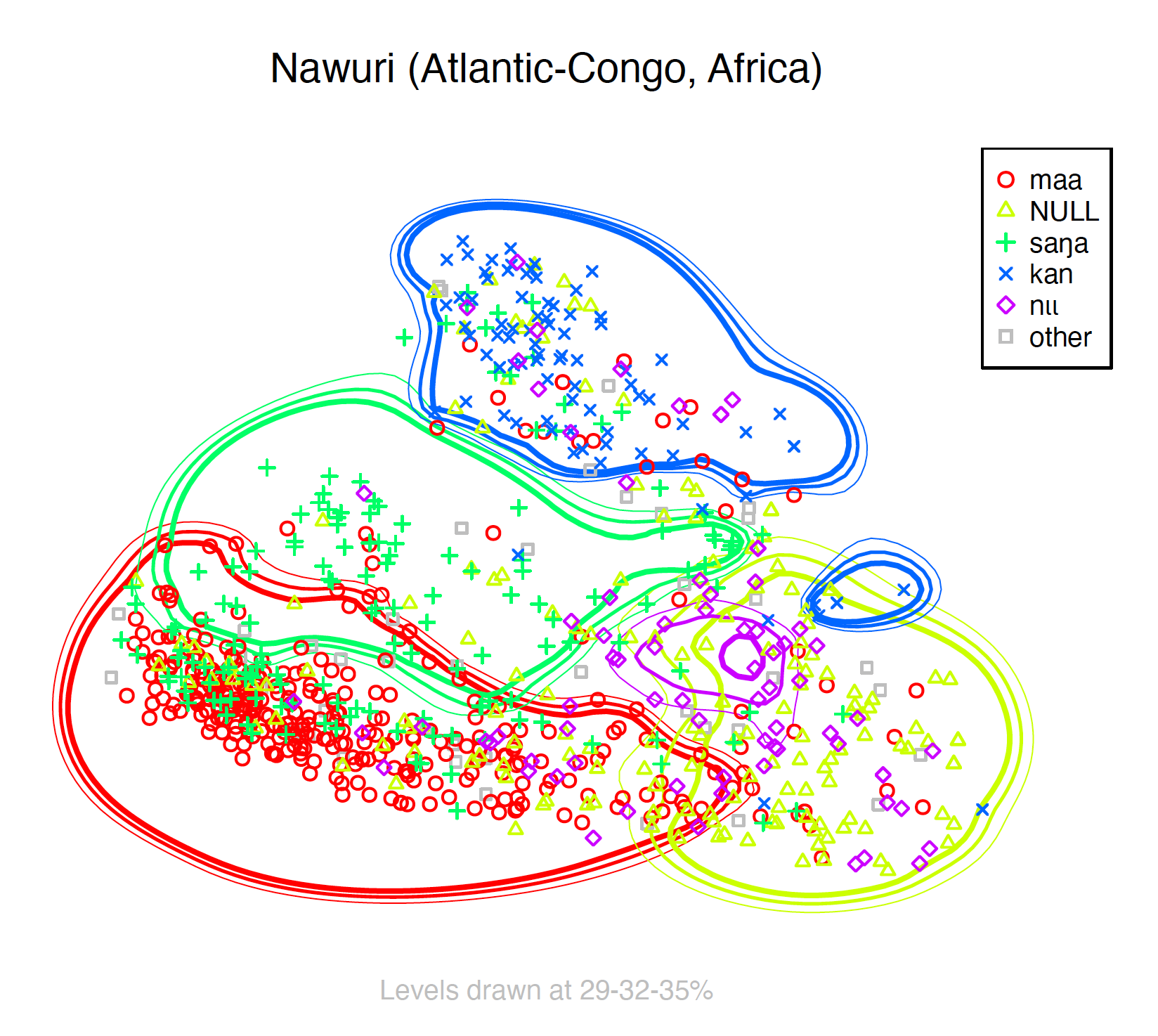}
\caption[Kriging map for Nawuri (Atlantic-Congo, Africa)]{}
\label{nawuri}
\end{subfigure}
\begin{subfigure}{0.50\textwidth}
\includegraphics[width=0.9\linewidth]{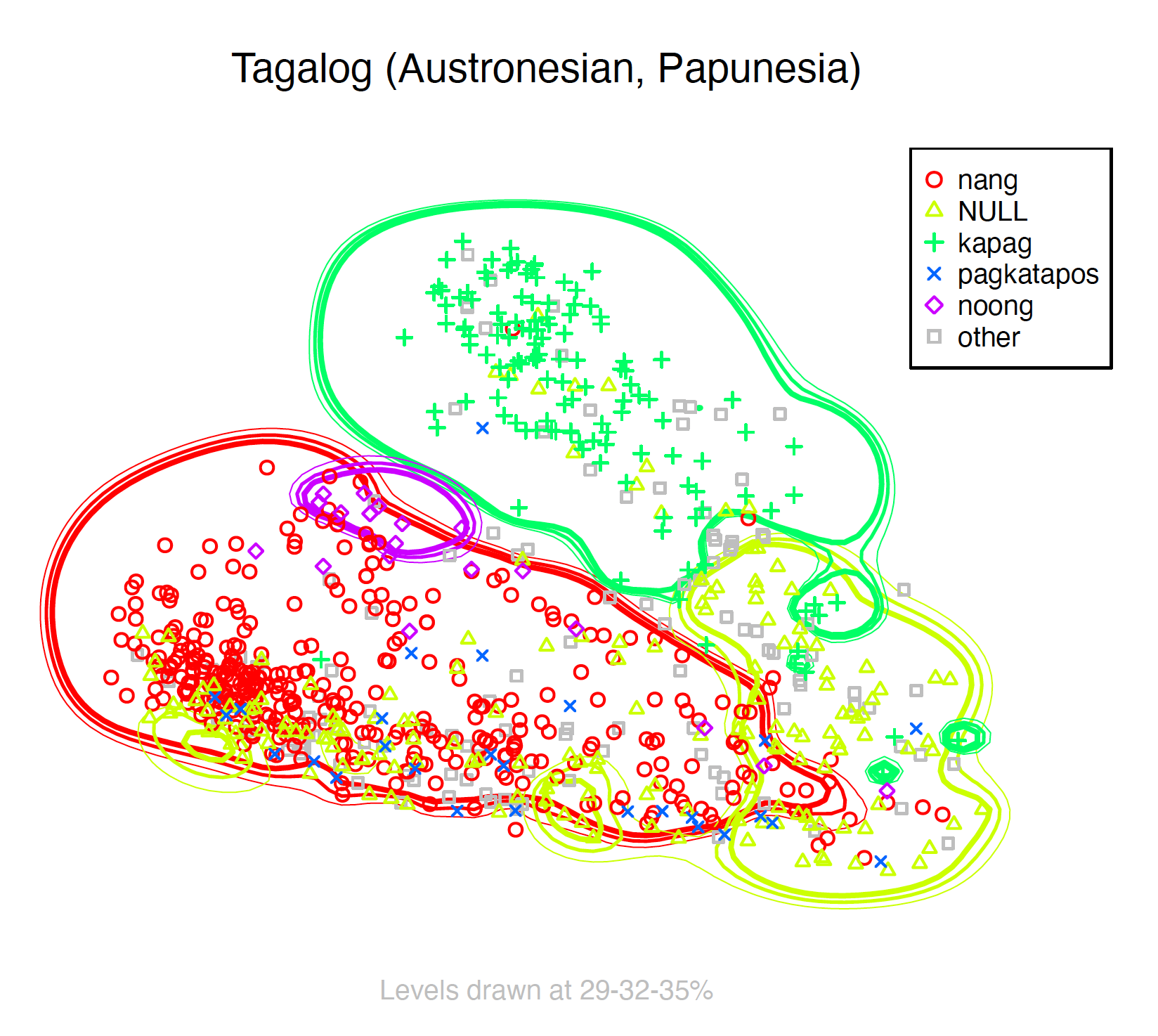}
\caption[Kriging map for Tagalog (Austronesian, Papunesia)]{}
\label{tagalog}
\end{subfigure}
\caption[]{Kriging maps for Nawuri (Atlantic-Congo, Africa) and Tagalog (Austronesian, Papunesia)}
\end{figure}

At the lower precision range (0.45-0.5) from Figure \ref{fig:bestmatches4}, we find languages that have some relatively well-defined Kriging area for cluster 4, but much more intense competition with other constructions. Some examples are Nawuri (Figure \ref{nawuri}) and Tagalog (Figure \ref{tagalog}).

\begin{figure}[!h]
\begin{subfigure}{0.50\textwidth}
\includegraphics[width=0.9\linewidth]{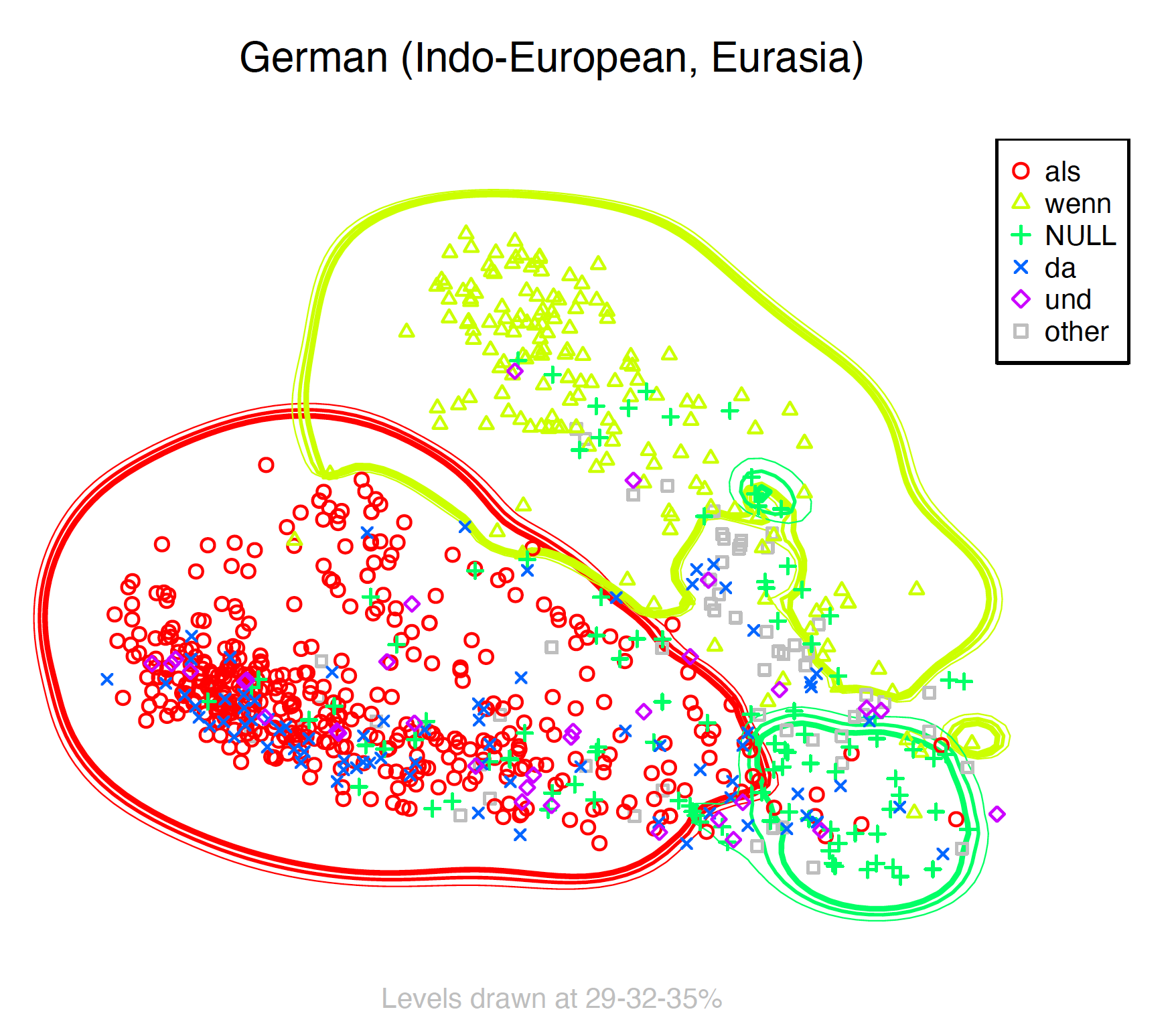}
\caption[Kriging map for German (Indo-European, Eurasia)]{}
\label{german}
\end{subfigure}
\begin{subfigure}{0.50\textwidth}
\includegraphics[width=0.9\linewidth]{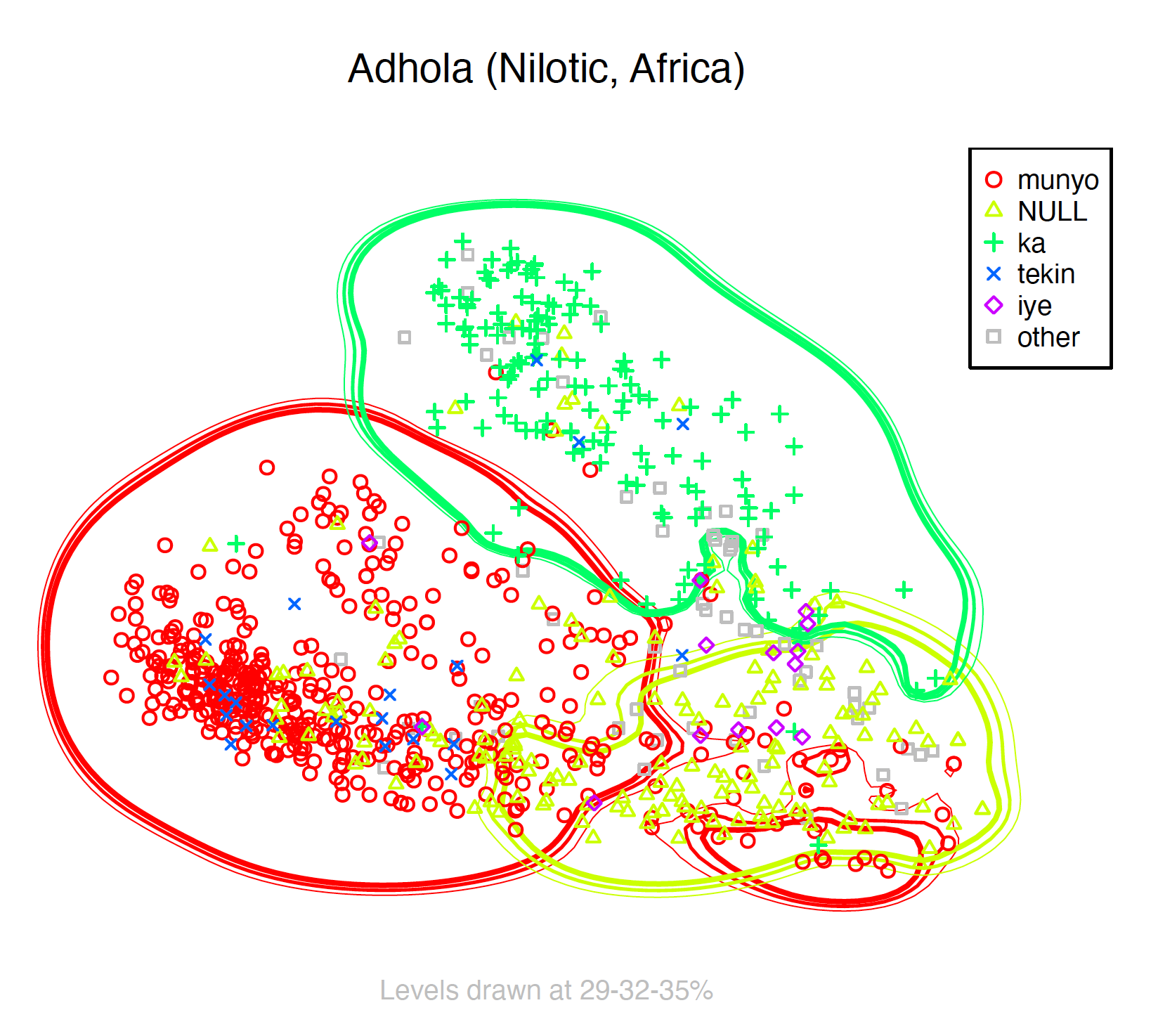}
\caption[Kriging map for Adhola (Nilotic, Africa)]{}
\label{adhola}
\end{subfigure}
\caption[]{Kriging maps for German (Indo-European, Eurasia) and Adhola (Nilotic, Africa)}
\end{figure}

At the higher recall range from Figure \ref{fig:bestmatches4}, we find several languages that have a relatively good match for cluster 4, but the same Kriging area covers more ground, typically including cluster 2 as well (i.e. the left extremity of the ML cluster). Most Germanic languages are one such example, like Afrikaans (Figure \ref{afrikaans}; $recall=0.83$) and German (Figure \ref{german}; $recall=0.81$), or Adhola [adh] (Figure \ref{adhola}; $recall=0.90$).

In Section \ref{sec:classif}, we saw that the observations in the TL area (identified on a close-reading basis, as shown in Figure \ref{threeway}) in the Old Church Slavonic and Ancient Greek maps appeared to be more neatly under the domain of the \textit{when}-equivalents available in the language (i.e. \textit{jegda} and \textit{hótan}) than those in the ML area and even less so in the BL area, where finite subordinators are instead clearly present but more heavily competing with null constructions, as the Kriging results also indicated. The temporal usage of \textit{wenn}, then, as exemplified in (\ref{deuexistvsuniv}b), may approximate some of the functions of \textsc{when} that are, overall, the \textit{least} likely to be rendered with null constructions in the two languages. 
% abs (67 tot):
% 4 (28) > 6 (22) > 5 (10) > 3 (6) > 1 (1) > 2 (0)
% 41.79 > 32.83 > 14.92 > 8.95 > 1.49 > 0

% xadv
% 197 tot
% 4 (87) > 6 (76) > 1 (20) > 5 (7) > 3 (7) > 2 (0)
% 44.16 >  38.57 > 10.15 > 3.55 > 3.55 >0

% egda
% 141 tot
% 3 (56) > 4 (34) > 5 (21) > 1 (21) > 2 (6) > 6 (3)
% 39.71 > 24.11 > 14.89 > 14.89 > 4.25 > 2.12

% In other words, competition with null constructions here is less likely? Do we have like 80/20 cluster 3, cluster 2 50/50, then cluster 3 smth like 20/80? Idk
Since we do not have access to a comparative concept \citep{haspelmath2010} that could tell us whether to include a particular usage point in a gram type, a possible route to finding a core or prototypical meaning for a particular GMM cluster could be to count how many languages express each observation in that cluster using their top-ranking lexical item (by F1 score, as described above) for that cluster. For instance, the top-ranking item for GMM cluster 3 for German, Old Church Slavonic, and Italian are \textit{wenn}, \textit{jegda}, and \textit{quando}. If an observation from inside cluster 3 is expressed by \textit{wenn}, \textit{jegda}, and \textit{quando}, its `prototypicality' score would be 3; if an observation from cluster 3 is expressed by \textit{wenn} and \textit{jegda} but not \textit{quando}, then the score would be 2; and so on for each sentence from cluster 3. After scoring each data point inside cluster 3 for prototypicality, the highest-ranking data points may then best approximate the prototypical usage of the gram type, if any, corresponding to that cluster. \\
\indent Among the prototypical examples for cluster 3, we find both universal \textit{when} in the present tense \ref{ex:corpus-universal} and existential \textit{when} in the future tense \ref{ex:corpus-existential}.

\begin{example}
    \label{ex:corpus-universal} And he said to them, “\textbf{When you pray}, say: “Father, hallowed be your name. Your kingdom come. (Luke 11:2)
\end{example}

\begin{example}
   \label{ex:corpus-existential} And he said, “Jesus, remember me \textbf{when you come} into your kingdom.” (Luke 23:42)
\end{example}

This shows that the colexification of existential \textit{when} in the future with the universal \textit{when}, which could appear to be an idiosyncrasy of German, is actually found across the languages that make some distinction between existential and universal \textbf{when}. Both these examples are expressed by \textit{jegda}-clauses in Old Church Slavonic and \textit{hótan}-clauses in Ancient Greek and are quite representative of the overall type of temporal relations found among the sentences in this cluster.\\
\indent When we extract the most prototypical sentences from cluster 2, we see that the top-ranking sentences are all stative when-clauses of the type `when I was a child', `when you were young', `when we were with you', `when you were slaves', `when you were pagans', and similar. Note that these stative clauses are in fact \textit{existential}, not \textit{universal}, when-clauses, since there is effectively only one eventuality of the type, for example, \textit{I be child} or \textit{you be pagans}. The matrix clause of these \textit{when}-clauses, however, seem to be mostly repeated occurrences of an eventuality, as in (\ref{protogmm3}) or (\ref{protogmm3b}). 

\begin{example}
    \textbf{When you were} slaves to sin, you were free from the control of righteousness (Romans 6:20)
    \label{protogmm3}
\end{example}

\begin{example}
   You know that \textbf{when you were} pagans you were led astray to mute idols, however you were led (1 Corinthians 12:2)
   \label{protogmm3b}
\end{example}

When the matrix clause is stative instead, then it seems more common for \textit{when}-clauses in this cluster to be eventive, as in (\ref{protogmm3c}). 

\begin{example}
    Now Thomas, one of the twelve, called the Twin, was not with them \textbf{when Jesus came} (John 20:24)
    \label{protogmm3c}
\end{example}

Note that this example is not fundamentally at odds with those in (\ref{protogmm3}) and (\ref{protogmm3b}) from the temporal-relation perspective. In both cases, the \textit{when}-clause works as an existential determiner over times. Since one of the two eventualities is stative, the temporal relation between the \textit{when}-clause eventuality and the matrix eventuality is symmetrical, in the sense that, even if the order in which they are presented is swapped, their temporal interpretation remains very similar and what may change is rather their information-structural status and discourse relation, as observed, among others, by \pgcitet{sabo2011}{3} and as we discussed in Chapter 1. \\
\indent Finally, a prototypical example from cluster 3 involves two eventive clauses, as in (\ref{protogmm4}a)-(\ref{protogmm4}c)

\begin{example}
    \begin{itemize}
        \item[a.] But \textbf{when Jesus heard} it he said, “This illness does not lead to death. It is for the glory of God, so that the Son of God may be glorified through it.” (John 11:4)
        \item[b.] \textbf{When Jesus saw} her, he called her over and said to her, “Woman, you are freed from your disability.” (Luke 13:12)
        \item[c.] So \textbf{when Martha heard} that Jesus was coming, she went and met him, but Mary remained seated in the house (John 11:20)
    \end{itemize}
    \label{protogmm4}
\end{example}

Old Church Slavonic translates (\ref{protogmm4}a)-(\ref{protogmm4}b) with a conjunct participle and (\ref{protogmm4}c) with a \textit{jegda}-clause. Intuitively, all three sentences involve the same temporal relation between clauses, namely one of temporal \textit{abutment}, whereby the matrix event, in this case, immediately follows the event of the adverbial clause. What seems to differ between (\ref{protogmm4}c) and the other two examples is the fact that the matrix clause in (\ref{protogmm4}c) continues with an (adversatively) coordinated clause with a different subject. 
% Also say something about the absence of repeated past events?
% Maybe add section around here?
Among the languages in which null constructions are predominant in GMM cluster 4, while clusters 2 and 3 are lexified (i.e. pattern-B or -D languages in our classification), we find that languages with converbs (or converb-like forms) or known for allowing serial constructions are particularly frequent.\footnote{see what reviewer said about this} Among pattern-D languages, for instance, we find numerous West African languages, where extensive use of serial verb constructions is a well-known prominent feature (cf. \citealt{givonserial2015, stahlkeserial1970, awoyale, bamgbose, lordserial}), as well as Yabem (cf. \citealt{bisangserial95}) and several other Austronesian and Papuan languages (cf. \citealt{senftserial, conradwogigaserial}), also oft-cited for their use of verb serialization. Among pattern-B languages, we find several North and South American language families, such as Arawakan, Aymaran, Chibchan and Tupian, all of which have also been studied with respect to their use of serial constructions (cf. \citealt{aikhenvald}). Languages that make extensive use of converbs, such as Korean and Kumyk, or predicative participles functionally very similar to converbs, such as Ancient Greek, are also among pattern-B or -D languages in which null constructions are predominant in GMM cluster 4. \\
\indent The intuition is that the situations found at the bottom half of the semantic map are more likely to be found as part of sequential events, which can be expressed with serial verb constructions or clause chaining by the languages where these are possible. \\
\indent As already discussed, the way in which the data was sampled (i.e. starting from a single, albeit relatively underspecified, lexified means, namely English \textit{when}) does not allow us to say much about cross-linguistic correspondences between different types of null constructions using the semantic map without further layers of annotation. What is important to highlight here, however, is that both the analysis of the Kriging maps and the precision and recall analysis of the GMM clusters indicate that there are clear cross-linguistic patterns as to which situations are more likely to be encoded by overtly subordinated, finite \textit{when}-clauses and which are more likely expressed with juxtaposed constructions. This means that the division of labour between \textit{jegda}-clauses and participle constructions in Early Slavic is likely to follow broader principles followed by many languages cross-linguistically.
% However, we can formulate some hypothesis on the basis of languages for which more granular information on null constructions is available, which is the case for the historical Indo-European languages in the PROIEL Treebank. 

% [Where I'm getting is: clause-chaining, serial constructions, converbs, etc. all have in common that they are in-between subordinates and main predications from the discourse perspective. What's important here is that the relation between these and the main verb is even more underspecified than with \textit{when}-clauses, which might make a language prefer them in contexts where what is stressed is the sequencing of events?]

% But what is the difference between clusters 2 and 4? Our general analysis of the left-hand side clusters 

% Notice that this holds true even if the competition in the lower half of the plot is just a quirk of German. But the more interesting question is whether there is a general tendency, across languages, to have more variation of expression in the lower half of the semantic map. CAN WE ANSWER THIS WITH HOMOGENEITY DATA?

% IF SO, also give a brief possible explanation in terms of structure, quantificational subordination etc.
% HERE ARE SOME MAPS WE COULD USE: Do we find the same tendency in completely unrelated languages? gil- does not have it (both areas are homogeneous), mnk- does, tgl- too (though in a different way, one words vs. null alternating with a lot of other things). pbi- would be an exception where both clusters are heterogeneous.

\section{Summary}

In this chapter, the semantic space of the temporal connective \textit{when} was explored by leveraging a massively parallel corpus of Bible translations and starting from the semantically underspecified subordinator \textit{when} in English. The methods presented in the chapter provided tentative, but promising results, particularly thanks to clear evidence for a cross-linguistic gram type expressing \textit{universal} \textsc{when}.

Another clear finding is that null constructions, including Early Slavic participle constructions, cluster in particular regions of the semantic map. This means that they are not equally viable as alternatives to any use of \textsc{when}, but carry particular meanings that make them less suitable for some functions of \textsc{when}. The way the data was sampled so far, though ideal for the exploration of lexified \textsc{when}-clauses, was not well suited to further investigations in this area, but rather to single out the broad domains where overtly subordinated \textit{when}-clauses and null constructions are respectively predominant. What we observed is that the former have the lowest competition with null constructions in correspondence to \textit{universal} \textsc{when}-clauses, as well as \textit{existential} \textsc{when}-clauses in the future, followed by existential, but stative \textsc{when}-clauses in the past tense (i.e. GMM cluster 2 in our analysis). Competition seems stronger (both options are viable) with two eventive eventualities, but the Kriging analysis suggested that this is most typically the domain of null constructions. \\
\indent The next chapter will add layers of annotation to the semantic map for Old Church Slavonic and Ancient Greek, and look for potential finer-grained observations at the level of null constructions. Similarly, I will leverage the extensive literature and data on languages which showed similar patterns to Old Church Slavonic to corroborate the intuitions we had on the latter regarding the usage of null constructions in contexts where the sequentiality of eventualities is stressed.
% However, preliminary results obtained by superimposing linguistic annotation from PROIEL seem to suggest that the results from Chapter \ref{chapter2} correspond to some well-attested cross-linguistic pattern: absolute constructions are likely to be specialized \textit{frames}, whereas foregrounded material is likely to be expressed by conjunct participles. 
% The next chapter will focus on corroborating these results and model NULLs at finer-grained level, by building a semantic map that starts from different NULL constructions and taking languages for which annotation at the dependency level is available as source data for new GMM and alignment models. 

\chapter{Early Slavic participle clauses and their typological parallels}

\section{Participle constructions in the Old Church Slavonic WHEN-map}\label{ocsmaps}
\begin{figure}[!h]
\begin{subfigure}{0.50\textwidth}
\includegraphics[width=0.9\linewidth]{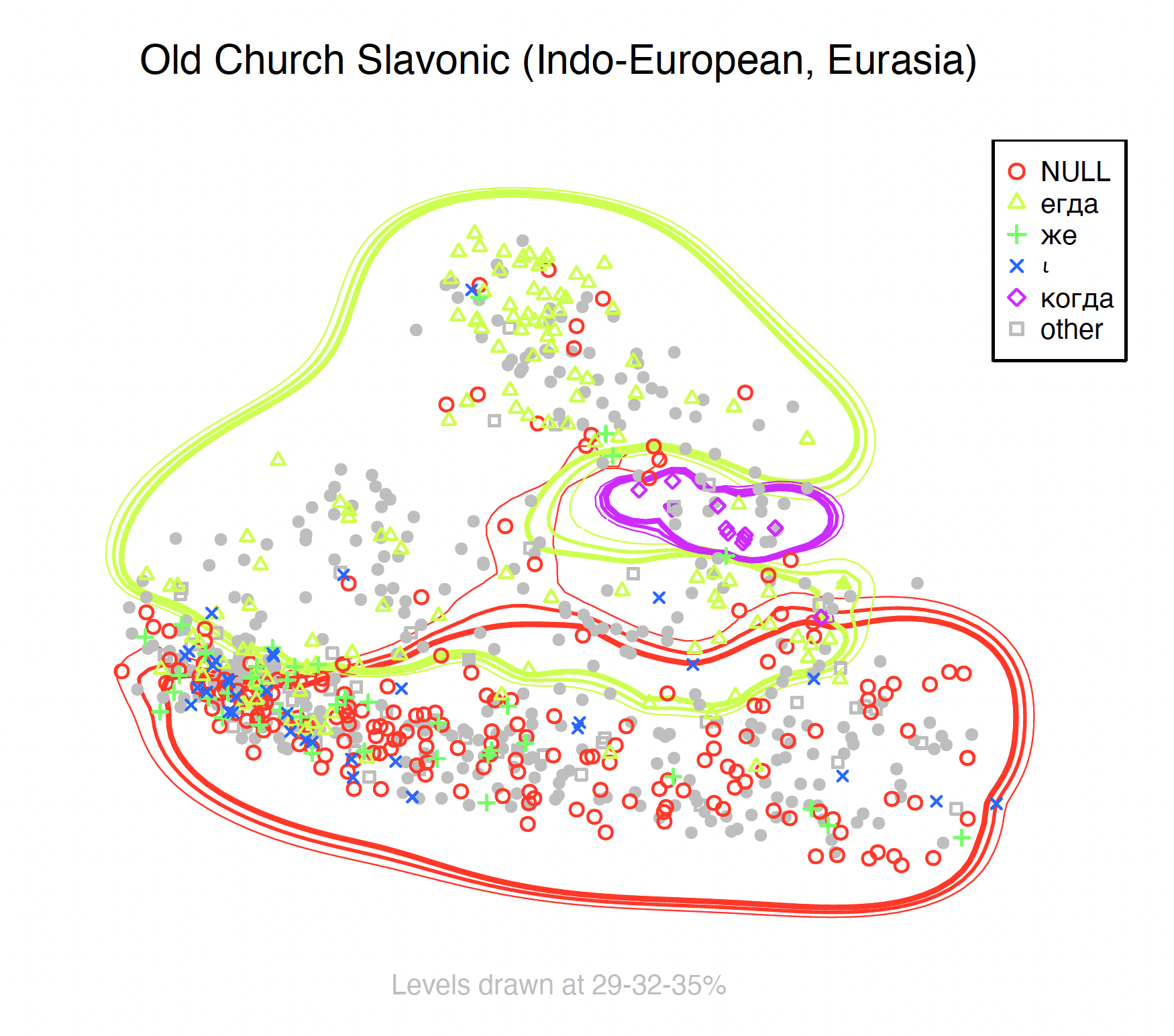} 
\caption[Kriging map for Old Church Slavonic (Indo-European, Eurasia), repeated]{}
\label{proielchu0rep}
\end{subfigure}
\begin{subfigure}{0.50\textwidth}
\includegraphics[width=0.9\linewidth]{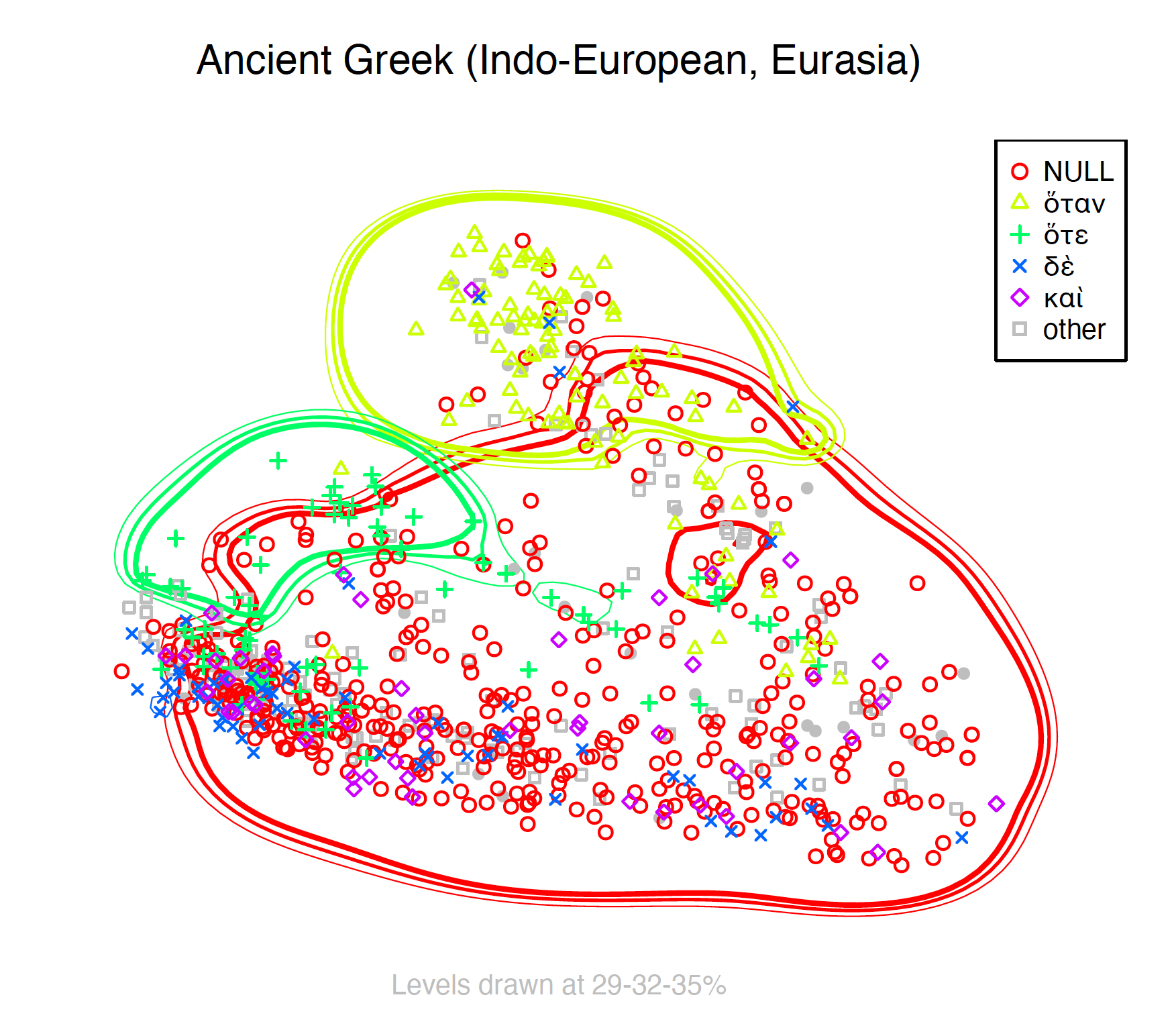} 
\caption[Kriging map for Ancient Greek (Indo-European, Eurasia), repeated]{}
\label{proielgrc0rep}
\end{subfigure}
\caption{Kriging maps for Old Church Slavonic (Indo-European, Eurasia) and Ancient Greek (Indo-European, Eurasia), repeated}
\end{figure}

A clear finding of the previous chapter was that null constructions cluster, cross-lin\-guis\-ti\-cal\-ly, in particular regions of the semantic map of \textsc{when}, indicating that they are not equally viable as alternatives to any use of \textsc{when}, but carry particular meanings that make them less suitable for some of its functions. We also observed that the Old Church Slavonic map for \textsc{when} (here repeated in Figure \ref{proielchu0rep}) belongs to a configuration (which we called pattern B) in which the top-left (TL) and middle-left (ML) areas are colexified, while area the bottom-left (BL) area forms a separate cluster where null constructions are predominant. In comparison, Ancient Greek (map repeated in Figure \ref{proielgrc0rep}), which possesses two main generic \textit{when}-counterparts, belongs to a slightly different configuration, in which the TL and ML areas are encoded by two separate means (\textit{hótan} and \textit{hóte}, respectively), while the BL area is under the scope of the Kriging cluster for null constructions, as in Old Church Slavonic.

The Old Church Slavonic map contains multiple unattested observations (all the dots in grey), due to the several gaps in the Gospels from the Codex Marianus. \posscitet{mayer-cysouw} corpus contains a more recent Church Slavonic translation of the Bible, possibly a version of the \textit{Elizavetinskaja Biblija} (Elizabeth Bible), published in 1751.\footnote{The source of the text indicated by \citet{mayer-cysouw} does not report which translation it is based on.} Since the syntax of this version seems to be faithfully close to the oldest attestations,\footnote{All occurrences of the dative absolute and of \textit{jegda}-clauses found in the Codex Marianus were also found in this version after a manual check.} we can check whether the missing parallels in the Codex Marianus might have caused MDS and Kriging to skew the results. The map for the more recent Church Slavonic version is, in fact, very close to the Old Church Slavonic one from the Codex Marianus:

\begin{figure}[!h]
\centering
\includegraphics[width=0.6\textwidth]{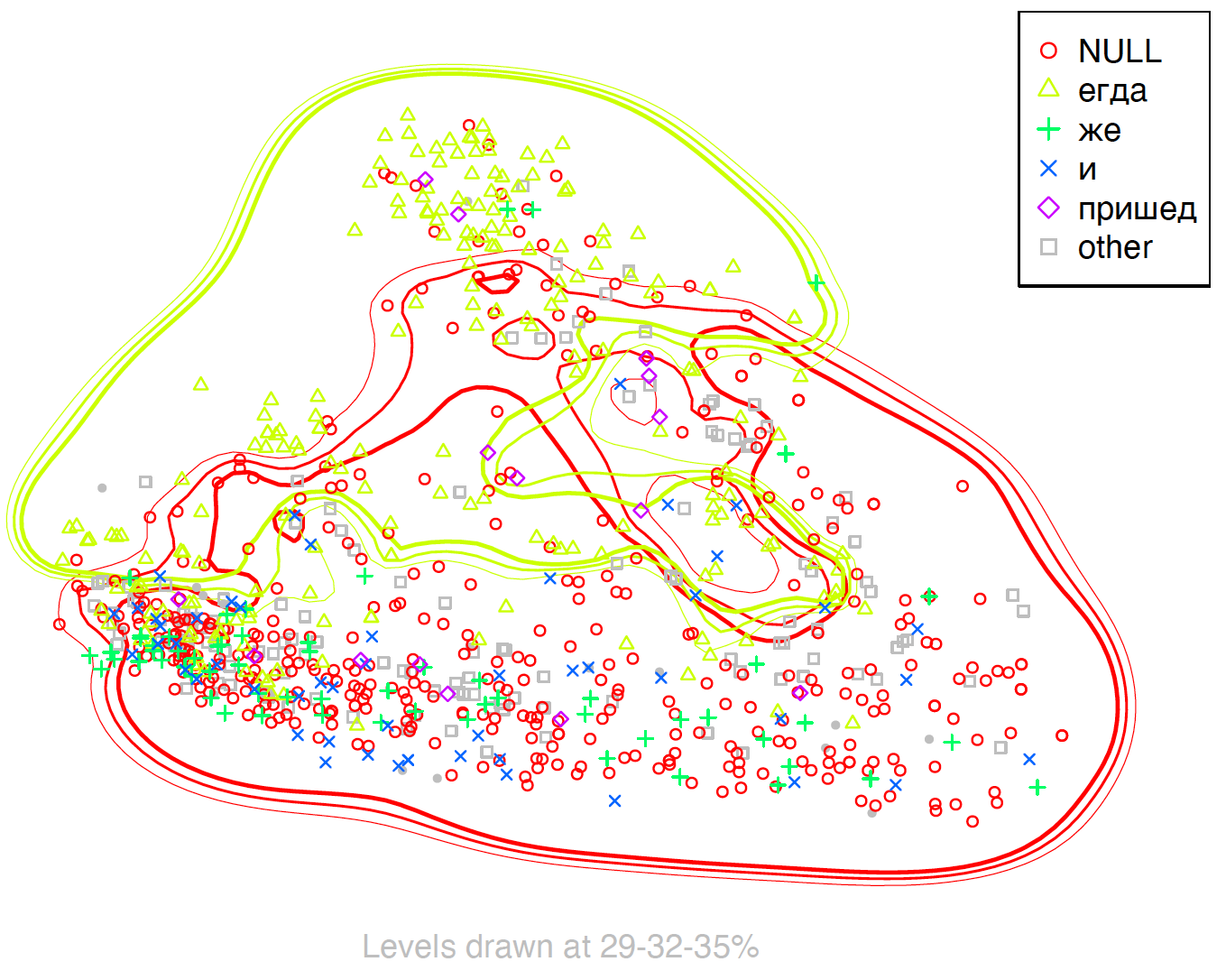}
\caption{\label{elizchu-when}Kriging map for Later Church Slavonic}
\end{figure}

The grouping of the TL and ML areas together, as well as the predominance of null constructions in the BL area, are the same in the two maps, which suggests that, despite the missing occurrences, the results from Kriging on Old Church Slavonic are quite reliable. \\
\indent As we can see from Figure \ref{proielchu0rep} (and equally in Figure \ref{elizchu-when}), the alignment model from English to Old Church Slavonic incorrectly identified some tokens (i.e. \textit{že} \textsc{ptc} and \textit{i} `and') among the most-common \textit{when}-parallels because they occur frequently and consistently just before certain null constructions, particularly participles. The same is true of \textit{prišed\foreignlanguage{russian}{ъ}} `having arrived, when (he) arrived', which is a very common conjunct participle throughout the Gospels. Moreover, there are occurrences of \textit{jegda} appearing in the less-frequent Old Church Slavonic spelling \textit{jegda} (which is, in fact, the spelling of the dictionary form used in TOROT), which are thus not counted together with the more frequent \textit{jegda}. \\
\indent As explained in the previous chapter, the maps for \textsc{when} were expected to be informative as far as \textit{when} equivalents in the target languages are concerned, as well as regarding the overall split between \textit{when}-situations (i.e. constructions with an overt temporal subordinator) and null-situations (i.e. generally either non-finite constructions or specific multi-word expressions not directly a counterpart of \textit{when}), but not when it comes to distinguishing participles from other null constructions (or between conjunct participles and dative absolutes). We can thus improve the Old Church Slavonic map by re-processing the source text in the following way:
\begin{itemize}
    \item[1.] lemmatized all forms of \textit{jegda} to \textit{jegda} (out of consistency with TOROT);
    \item[2.] remove \textit{že} and \textit{i} as stopwords;
    \item[3.] insert a placeholder before the relevant participle constructions (\textit{xadv} for conjunct participles and \textit{absoluteadv} for absolutes), as a way of levelling out the different surface realizations of conjunct participles and dative absolutes, thus allowing the models to capture the placeholders as dummy subordinating conjunctions.
\end{itemize}

\begin{figure}[!h]
\begin{subfigure}{0.50\textwidth}
\includegraphics[width=0.9\linewidth]{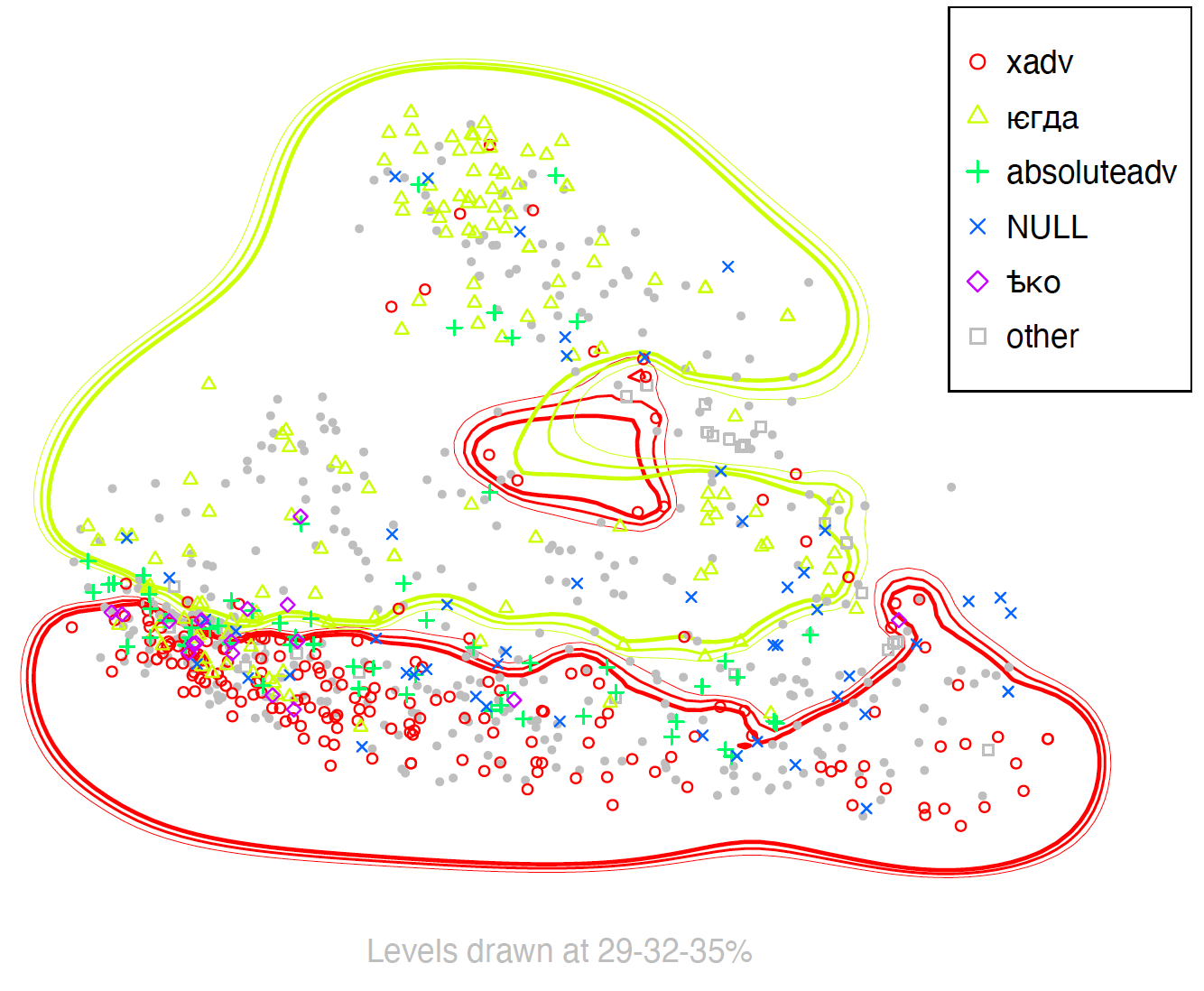} 
\caption[Kriging maps for Old Church Slavonic after preprocessing and with syntactic placeholders for absolute constructions and conjunct participles]{}
\label{proielchu4-when}
\end{subfigure}
\begin{subfigure}{0.50\textwidth}
\includegraphics[width=0.9\linewidth]{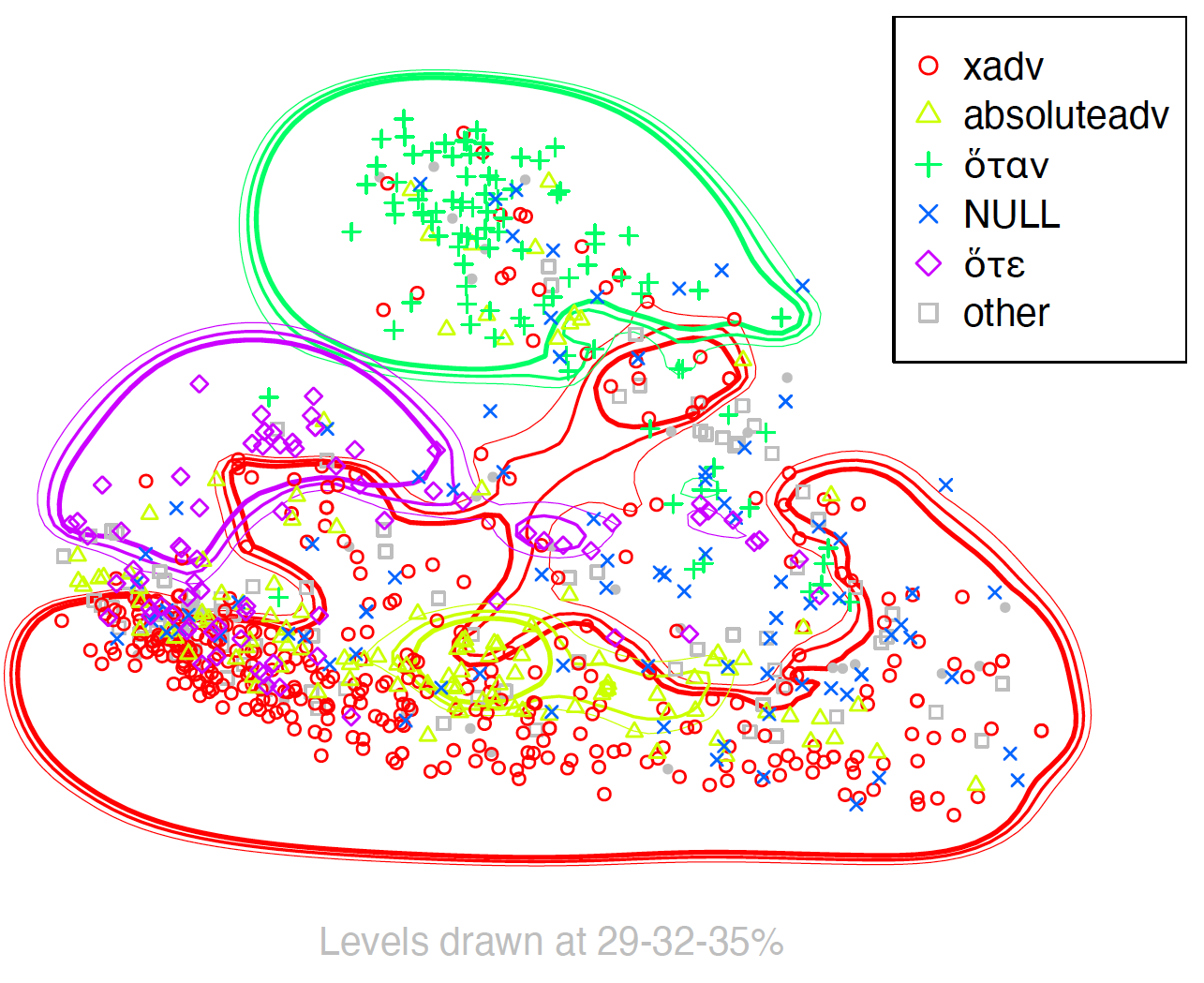}
\caption[Kriging maps for Ancient Greek after preprocessing and with syntactic placeholders for absolute constructions and conjunct participles]{}
\label{proielgrc2-when}
\end{subfigure}
\caption[]{Kriging maps for (a.) Old Church Slavonic and (b) Ancient Greek after preprocessing and with syntactic placeholders for absolute constructions and conjunct participles}
\end{figure}

After realigning the English version to the processed Old Church Slavonic version, we are able to capture some finer-grained differences in distribution between participles and \textit{jegda}-clauses, and between conjunct participles and absolute constructions, as Figure \ref{proielchu4-when} shows. Despite the lemmatization of \textit{jegda}, the area of the latter did not change substantially. The former null/\textit{jegda} division is now a much more clear-cut conjunct participles-\textit{jegda} split, with a smaller island of the same conjunct-participle cluster roughly between the TL and ML areas. Although dative absolutes are all mostly concentrated between conjunct participles and \textit{jegda}-clauses, they were not detected by Kriging as a separate area in the map. This is unlike Ancient Greek, after adding the same placeholders and preprocessing the text by removing \textit{dé} \textsc{ptc} and \textit{kaí} `and', as Figure \ref{proielgrc2-when} shows.

\begin{figure}[!h]
\begin{subfigure}{0.50\textwidth}
\includegraphics[width=0.9\linewidth]{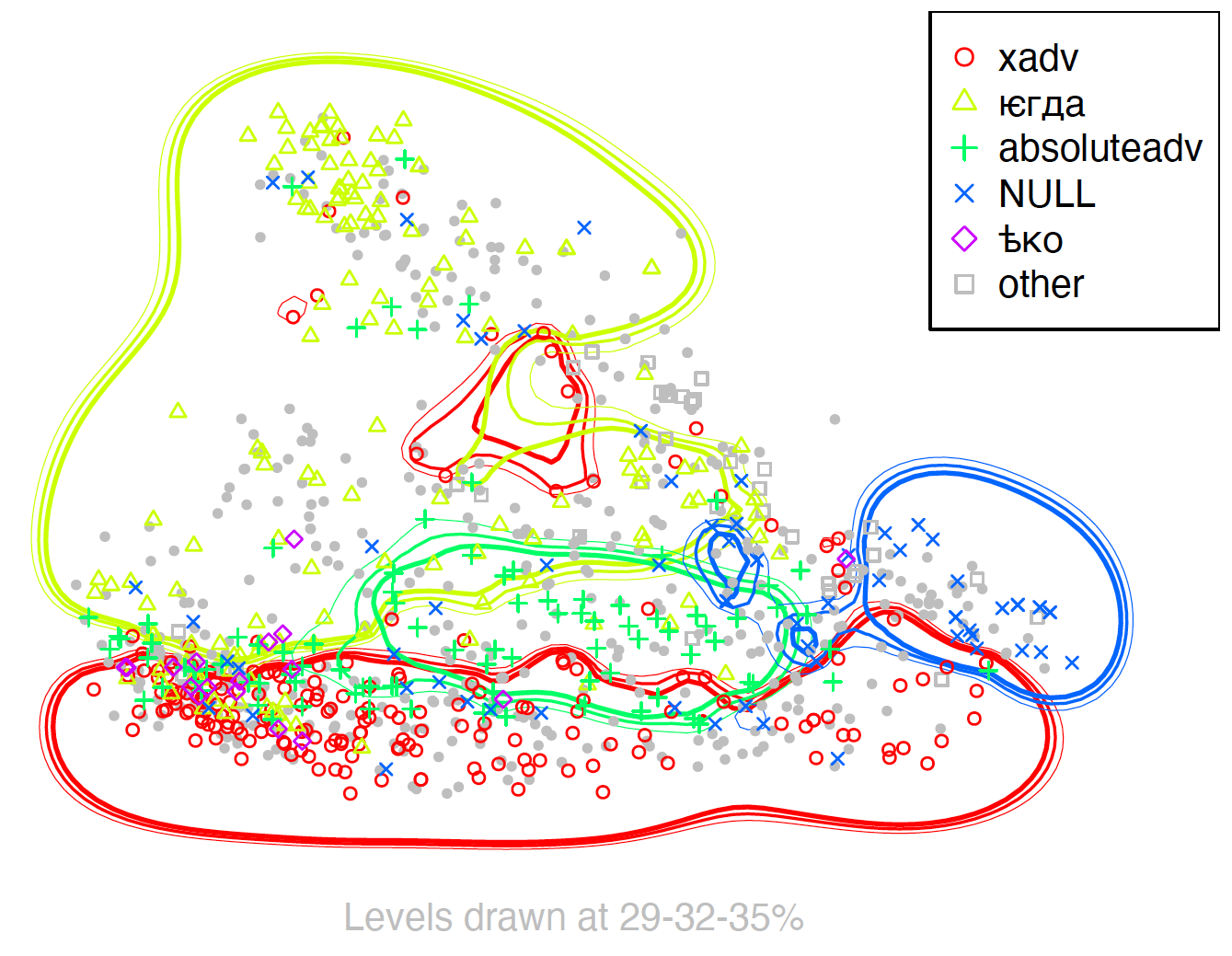} 
\caption[Kriging map for Old Church Slavonic after adding the parallels to English \textit{while} (in addition to \textit{when}) and syntactic placeholders for absolute constructions and conjunct participles]{}
\label{proielchu-whenwhile}
\end{subfigure}
\begin{subfigure}{0.50\textwidth}
\includegraphics[width=0.9\linewidth]{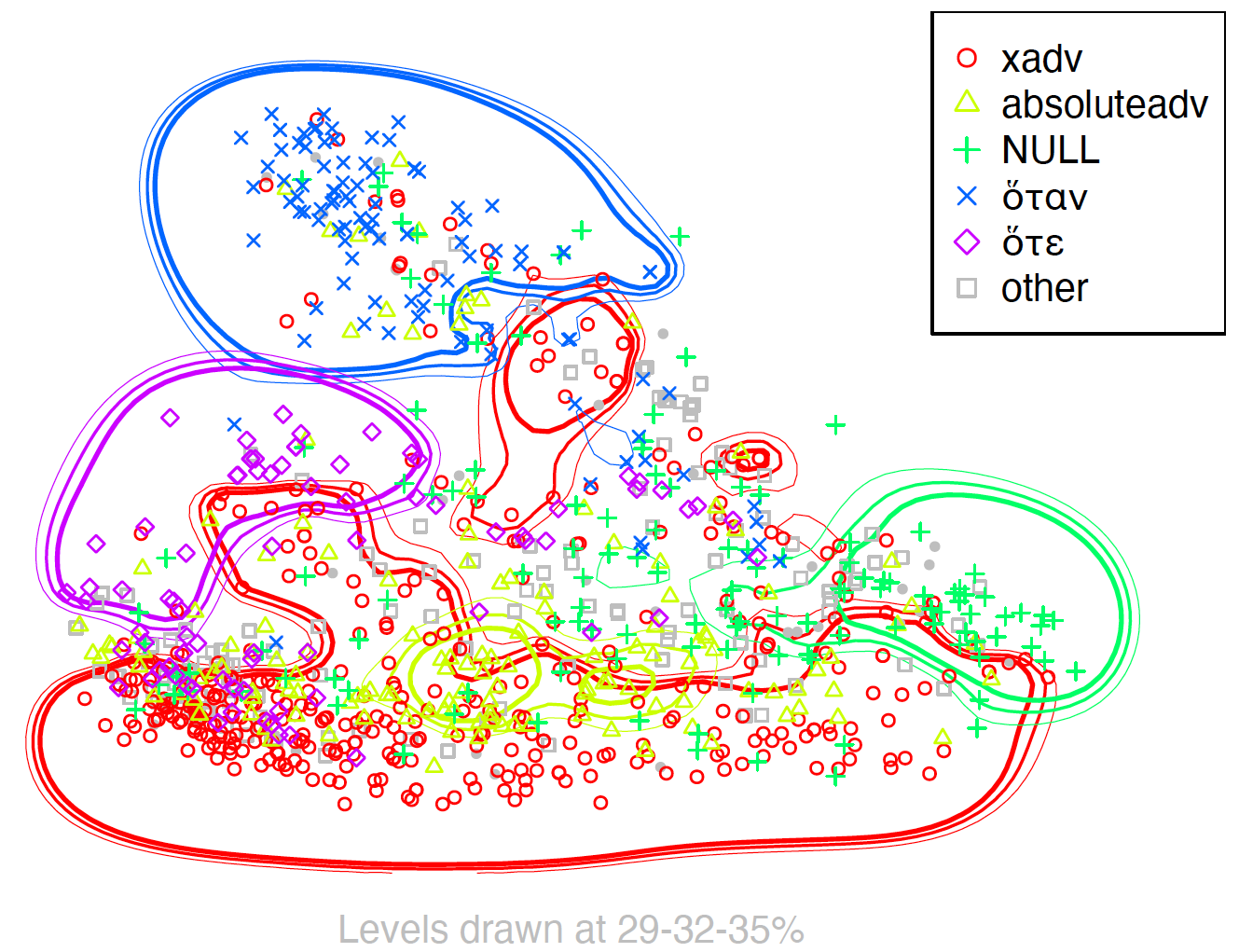}
\caption[Kriging map for Ancient Greek after adding the parallels to English \textit{while} (in addition to \textit{when}) and syntactic placeholders for absolute constructions and conjunct participles]{}
\label{proielgrc-whenwhile}
\end{subfigure}
\caption[]{Kriging map for (a.) Old Church Slavonic and (b) Ancient Greek after adding the parallels to English \textit{while} (in addition to \textit{when}) and syntactic placeholders for absolute constructions and conjunct participles}
\end{figure}

Absolutes in Greek form a Kriging area of their own, which seems to overlap almost entirely with the conjunct participle area. We should still remain wary of not over-interpreting the differences in boundaries between Old Church Slavonic and Ancient Greek, since the latter has virtually complete coverage of the New Testament, which may affect the boundaries of the Kriging areas. Nevertheless, the differences in the overall number and position of the clusters can be considered relatively reliable despite the missing occurrences in the Old Church Slavonic dataset, as Figure \ref{elizchu-when} from the later Church Slavonic version suggested. \\
\indent In Chapter 1, we saw that, although not in a statistically significant way, dative absolutes are more frequent in the imperfective, which means that `while \textit{x}-ing' may be a more likely English translation than `when \textit{x}-ing' for the 108 imperfective dative absolutes in the dataset.\\
\indent Adding \textit{while}-clauses to the \textit{when}-maps affects the Old Church Slavonic map in a useful way, since this time dative absolutes occupy a clearly defined Kriging area (Figure \ref{proielchu-whenwhile}). 

\begin{figure}[!h]
\centering
\includegraphics[width=0.6\textwidth]{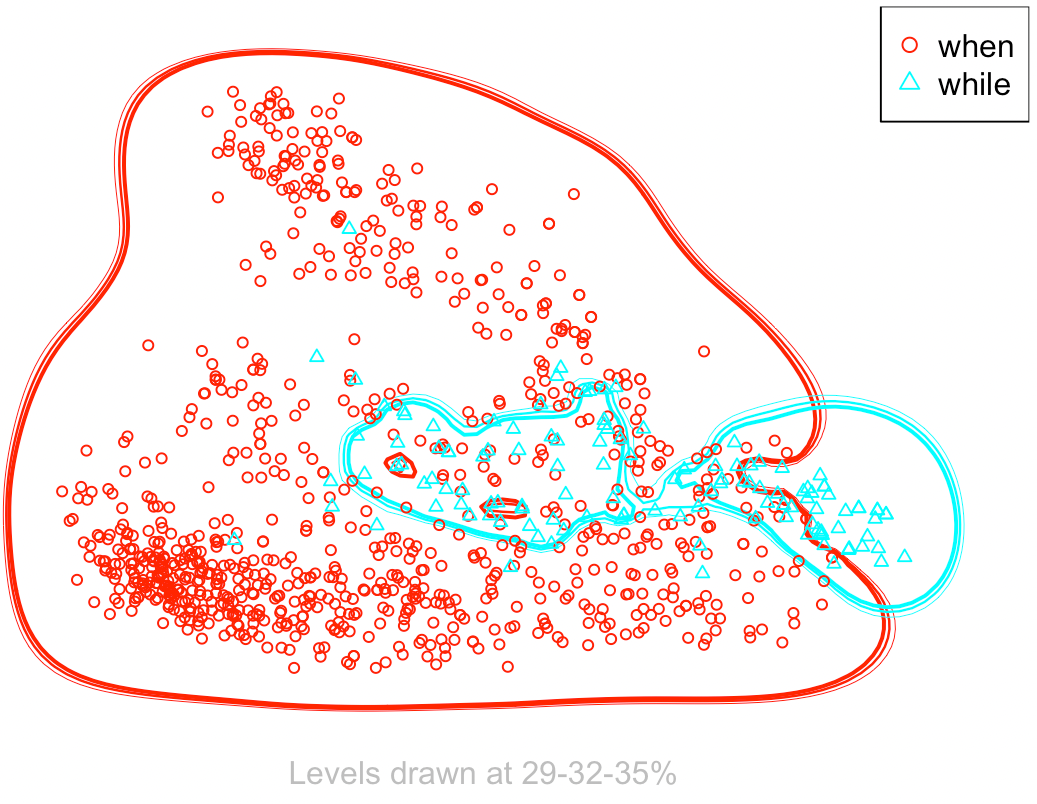}
\caption{\label{eng29-whenwhile}Kriging map for English (\textit{when} and \textit{while} situation).}
\end{figure}

If we compare these maps with the English map for \textit{when} and \textit{while} (Figure \ref{eng29-whenwhile}), we clearly see that Old Church Slavonic absolute constructions largely overlap with the \textsc{while} area, suggesting that English \textit{while} and dative absolutes are at least partially used in very similar contexts. Since a cluster is ultimately the result of the cross-linguistic comparison which was carried out via Hamming distance, the fact that the data points corresponding to absolutes in Old Church Slavonic are close together in an MDS map in itself indicates that those points are also likely often encoded by the same means cross-linguistically.

\begin{figure}[!h]
\begin{subfigure}{0.50\textwidth}
\includegraphics[width=0.9\linewidth]{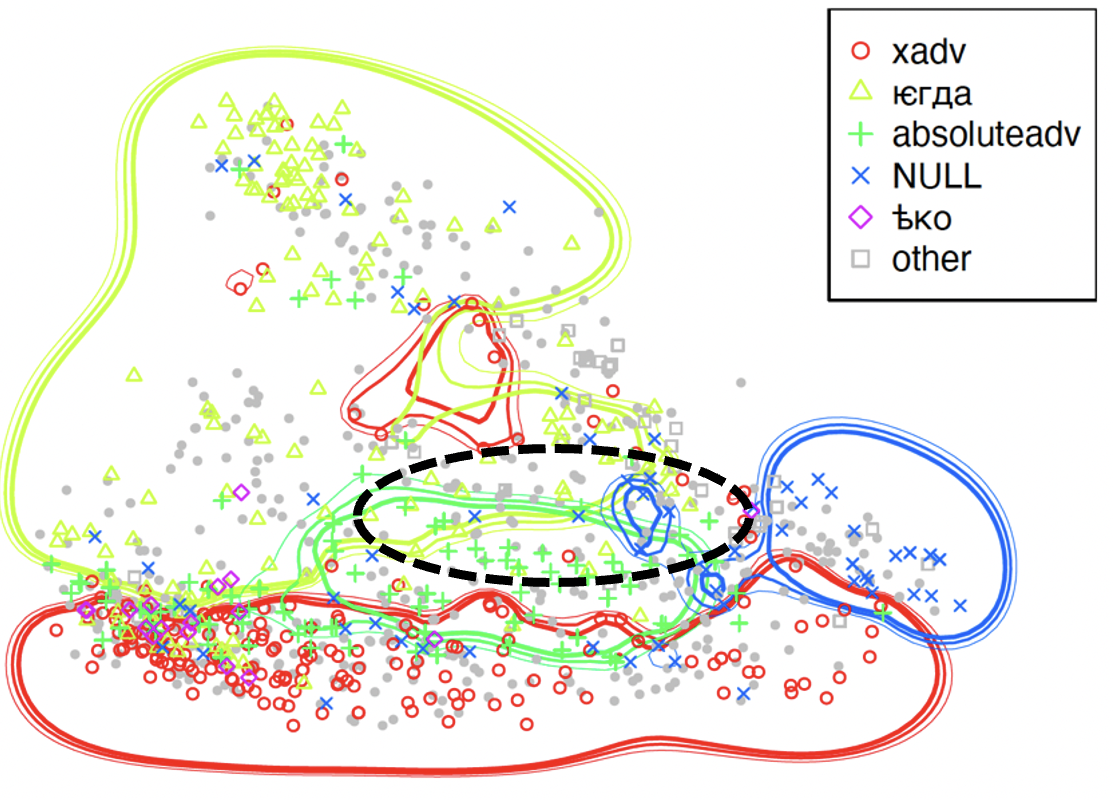} 
\caption[Overlap between the Kriging area for Old Church Slavonic absolute constructions and the Kriging area for English \textit{while}]{}
\label{proeilchuoverlap-whenwhile}
\end{subfigure}
\begin{subfigure}{0.50\textwidth}
\includegraphics[width=0.9\linewidth]{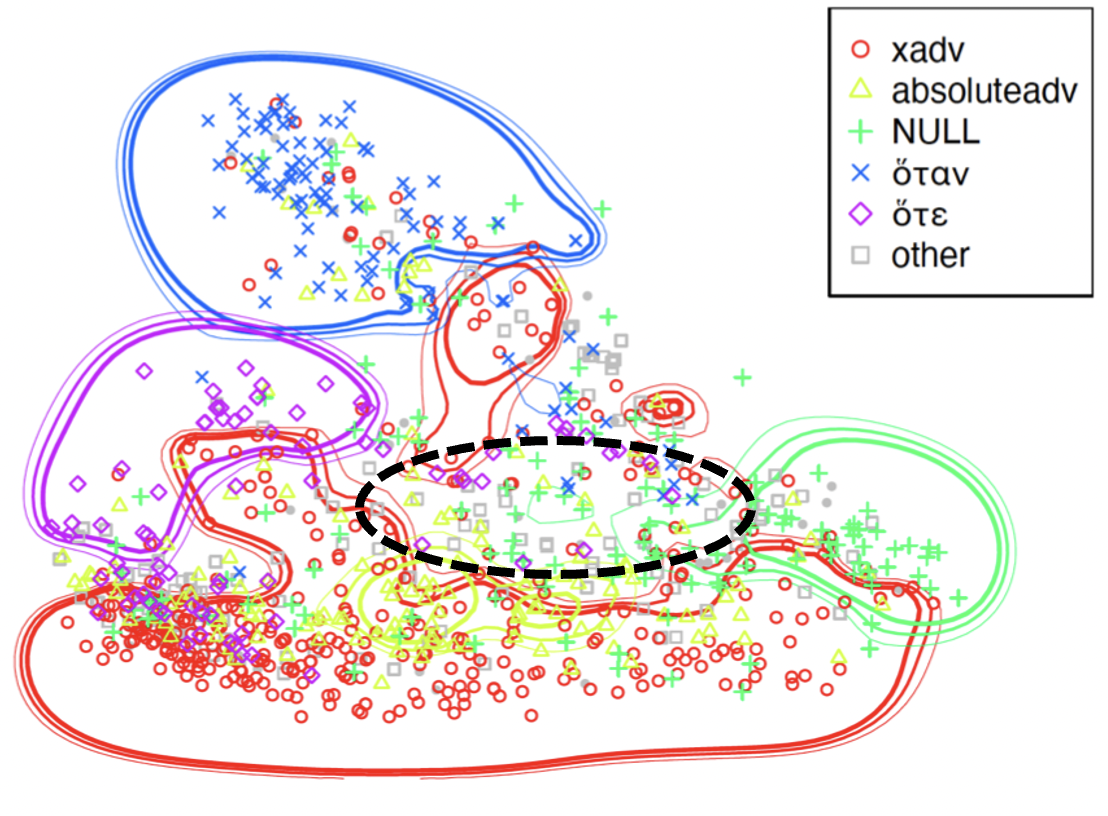}
\caption[Overlap between the Kriging area for Ancient Greek absolute constructions and the Kriging area for English \textit{while}]{}
\label{proeilgrcoverlap-whenwhile}
\end{subfigure}
\caption[]{Kriging map for (a.) Old Church Slavonic and (b) Ancient Greek \textsc{when} and \textsc{while} situations. The black dashed circle corresponds to part of the English \textit{while} area overlapping with absolutes in the two languages.}
\end{figure}

The Ancient Greek map (Figure \ref{proielgrc-whenwhile}) remains similar to the previous one, with absolutes occupying the same overall portion of the map as Old Church Slavonic dative absolutes. The Kriging areas for the construction in the two languages are, however, not quite identical. In both languages, they can be seen as forming a continuum with conjunct participles and are in particularly intense competition with both \textit{when}-equivalents and conjunct participles in the BL area of the map, as already noted for other languages in the previous chapter. In Old Church Slavonic, the Kriging area overlaps with part of the \textit{while} area in English (Figure \ref{proeilchuoverlap-whenwhile}), whereas this is not as neatly the case in Ancient Greek (Figure \ref{proeilgrcoverlap-whenwhile}).

This mismatch between Old Church Slavonic and Greek may reflect the observation made in Chapter 2 that, although in the majority of cases Old Church Slavonic absolutes correspond to an Ancient Greek absolute, there is a small but consistent subset of Old Church Slavonic occurrences that do not correspond to an absolute in Greek, notably Greek \textit{en tō} with infinitive (`during, while \textit{x}-ing'). Overall, however, the fact that several \textit{while}-clauses in English cluster to the very right corner of the semantic map, where all languages have some null constructions, suggests that they \textit{may} not correspond to lexified counterparts as frequently as \textit{when}-clauses. Whether this reflects some cross-linguistic regularity (i.e. \textsc{while}-clauses are less likely to be separately lexified) or a lexical feature of the corpus is an open question which, evidently, cannot be investigated at scale with New Testament data. 

\section{Lexified counterparts to Early Slavic participle constructions in the world's languages}\label{sec:pular}
\begin{figure}[!h]
\centering
\includegraphics[width=0.8\textwidth]{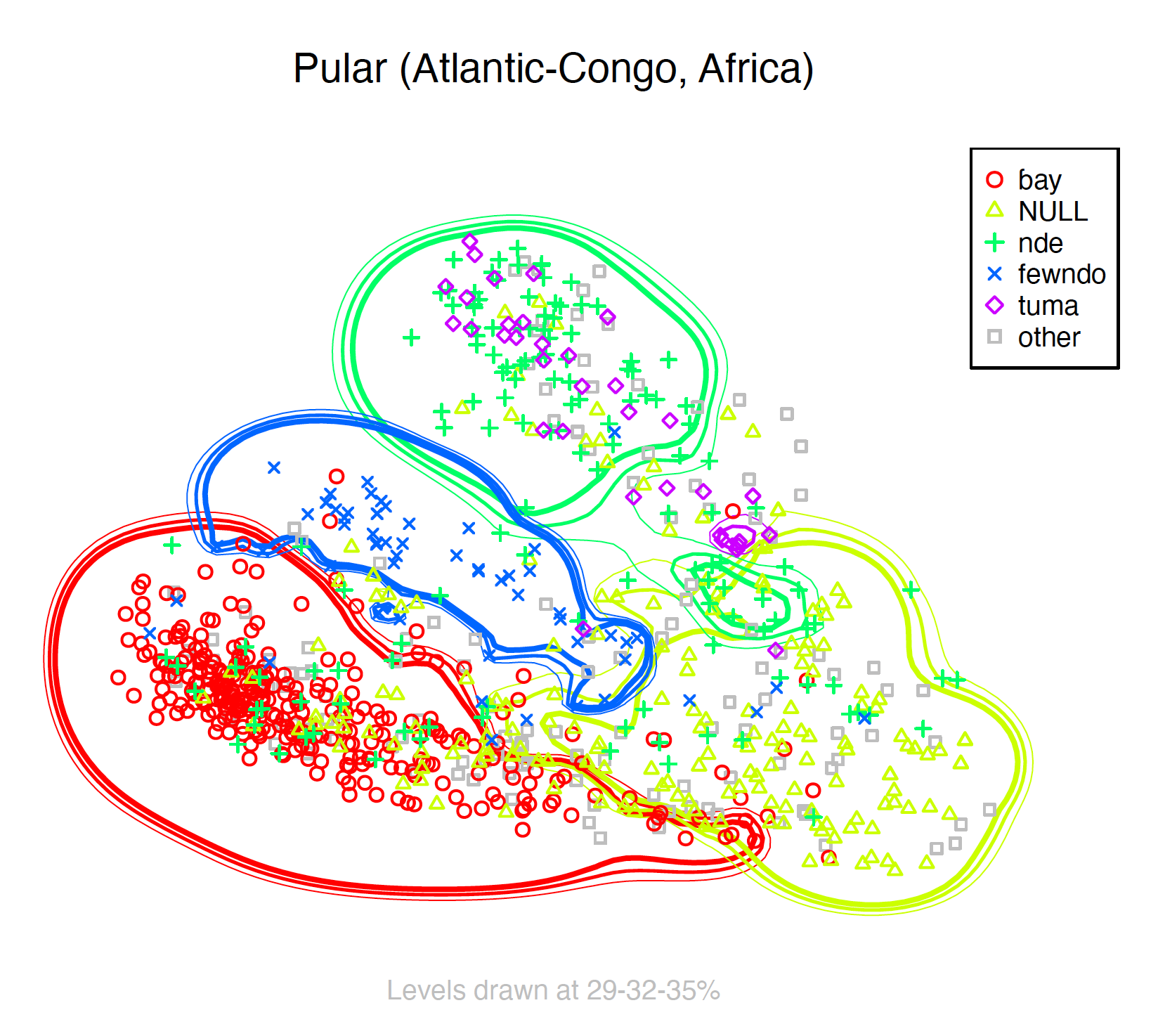}
\caption{\label{pularrep}Kriging map for Pular (Atlantic-Congo, Africa), repeated}
\end{figure}

In the analysis in Chapter 5, we observed that few languages have relatively clear boundaries for \textit{both} the middle-left (ML) cluster (GMM clusters 2) and the bottom-left (BL) cluster (GMM cluster 4), since in most cases the competition between different constructions is quite evident in one or both of those areas. Among these languages, we saw Pular, which had lexicalized means for both the ML and the BL clusters, as well as relatively low competition compared to what we observed in several other languages. In our precision and recall analysis in the previous chapter, this language always figured among those with a relatively high F1 score for the different clusters. An interesting characteristic of Pular is that it possesses several counterparts to \textit{when}, several more, in fact, than the ones suggested by the Kriging areas in the Pular map (Figure \ref{pularrep}, repeated from Chapter 5). 

\begin{figure}[!h]
\centering
\includegraphics[width=0.8\textwidth]{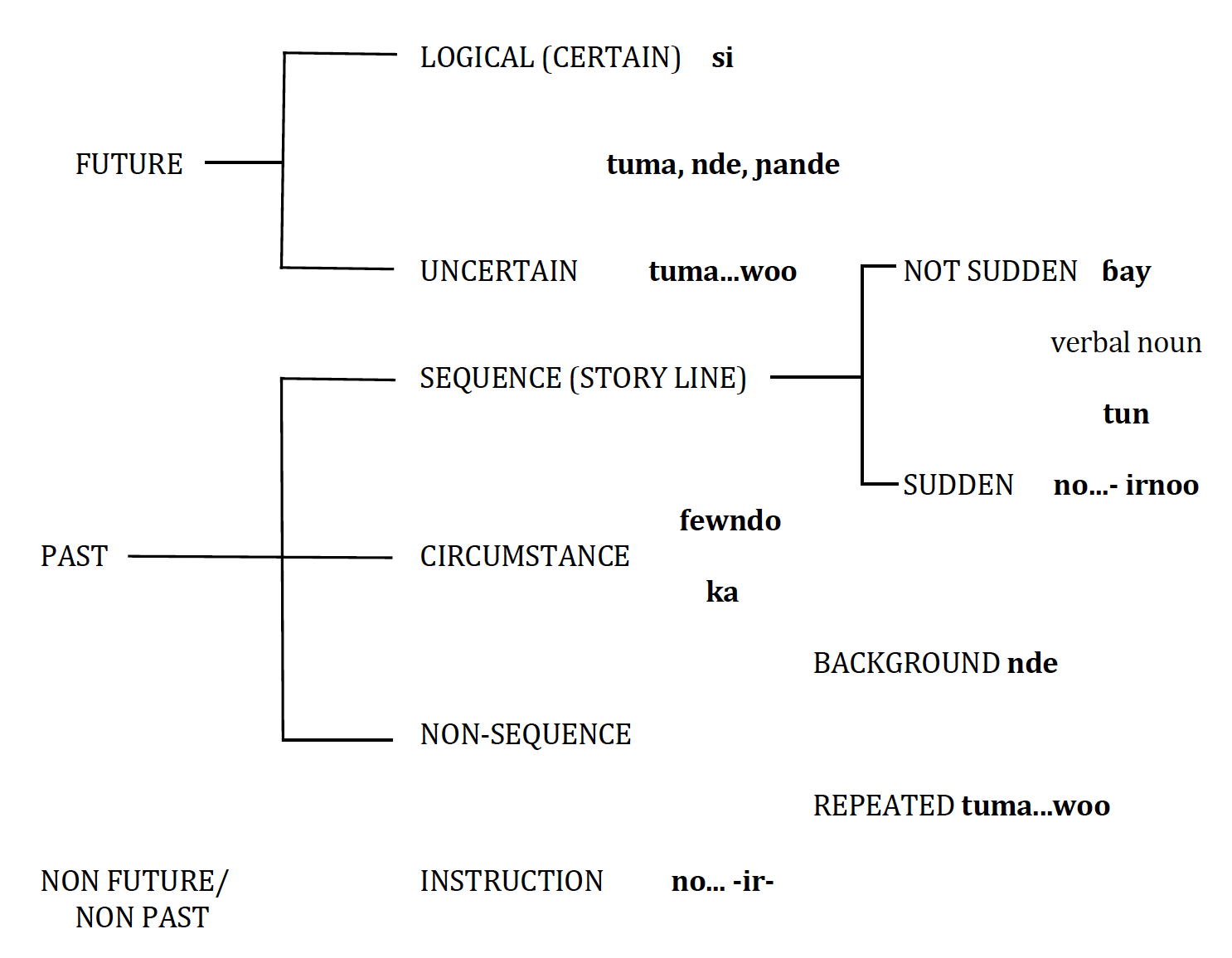}
\caption{\label{pularconnectors}Temporal connectives meaning ‘when’ in Pular according to \pgcitet{pulargrammar}{61}}
\end{figure}

We can attempt to use existing descriptions of Pular to understand the meanings of each subordinator, potentially using them as a means of confirming (or refining) the observations made so far about the semantics of the different clusters in the semantic map of \textsc{when}\\
\indent According to \citet{pulargrammar}, each connective conveys a particular temporal meaning and can generally only be found in combination with specific temporal-aspectual morphology on the clause it introduces. Figure \ref{pularconnectors} schematizes the range of their functions.

By crossing \posscitet{pulargrammar} overview with the Kriging map in Figure \ref{pularrep}, we see that four of the several means in Figure \ref{pularconnectors} are captured as Kriging areas. The TL cluster is mostly under the domain of \textit{nde} but competes with \textit{tuma}, which, in fact, has a small Kriging area in between the Kriging area for \textit{nde} and the one for null constructions. \textit{Nde}, according to the overview of \citet{pulargrammar}, is used to introduce both future and past eventualities. However, a number of occurrences of \textit{nde} are found in the bottom area, so it is possible that the latter reflect the use of this connective for past-tense clauses, whereas those under the Kriging area at the top-left of the map are, presumably, those introducing future-tense clauses, similarly to this usage of German \textit{wenn}, Norwegian/Danish \textit{når} or Afrikaans \textit{wanneer} seen in the previous chapter. This seems to be confirmed by its competition with \textit{tuma}, which, according to the overview, has a similar function. \\
\indent The ML cluster is under the scope of \textit{fewndo}, which the overview labels as \textsc{past circumstance}. According to \pgcitet{pulargrammar}{62}, when introducing past events, the main difference among all the connectives is between those that are used to express sequentiality (or `main event line') and those that introduce background information. The \textsc{circumstance} category is considered as an intermediate category since `it can be used in the story line, but can also be used to provide a setting for what follows' (\pgcitealt{pulargrammar}{65}). \\
\indent The BL cluster is instead under the domain of \textit{\textipa{\!b}ay}. This connective is one of several that are used to express sequentiality between the events they connect, differing in the level of `suddenness' they convey. With \textit{\textipa{\!b}ay}, `there is no sense of suddenness; rather one of expected development' (\pgcitealt{pulargrammar}{63}), whereas the verbal noun (-\textit{ugol}), \textit{tun}, and \textit{no... -irnoo}, convey a `greater sense of suddenness', `a fair degree of suddenness', and `the greatest sense of urgency', respectively. (\ref{pularseq}a), (\ref{pularseq}b) and (\ref{pularseq}c) are examples containing \textit{\textipa{\!b}ay}, \textit{tun}, and \textit{no...-irnoo}, respectively, where the difference in `suddenness' or `urgency' is reflected, to some extent, in the English translations.

\begin{example}
    \begin{itemize}
        \item[a.] \textit{\AS{\textbf{\m{B}ay}} \textit{o} hewtii ka suudu, bum\AS{\m{b}}e \AS{\m{b}}en \AS{\m{b}}adii mo. Iisaa wi'i \AS{\m{b}}e: ``Hi\AS{\m{d}}on sikka mi\AS{\m{d}}o waawi wa\AS{\m{d}}ude \AS{\m{d}}un?'' \AS{\m{B}}e jaabii mo: ``Hiiyii, yaa an Koohoojo.''}\\
        `When he entered the house, the blind men came to him, and Jesus said to them, “Do you believe that I am able to do this?” They said to him, “Yes, Lord.”' (Matthew 9:28)
        \item[b.] \textit{\textbf{Tun} o hewti, o \AS{\m{b}}adii Iisaa, e house himo wi'a: ``Karamoko'en!'' Onsay o hirbii mo}.\\
        `And when he came, he went up to him at once and said, “Rabbi!” And he kissed him' (Mark 14:45)
        \item[c.] \textit{\textbf{No} Elisabaatu nan\textbf{irnoo} salminaango Mariyama ngon, \AS{\m{b}}i\AS{\m{d}}\AS{\m{d}}o makko on memminii ka nder reedu makko, kanko Elisabaatu o heewi Ruuhu Senii\AS{\m{d}}o on}\\
        `And when Elizabeth heard the greeting of Mary, the baby leaped in her womb. And Elizabeth was filled with the Holy Spirit' (Luke 1:41)
    \end{itemize}
    \label{pularseq}
\end{example}

\textit{Tun} and \textit{no...-irnoo}, along with several other connectives from the overview in Figure \ref{pularconnectors}, are not visible from the Kriging map but are represented, albeit to a much lesser extent, in the dataset of \textit{when}-parallels. We can abstract from the individual forms and use their meanings as provided by \citet{pulargrammar} and run Kriging on the functional labels instead of the forms themselves. We can then look at where in the semantic map the different functions of \textsc{when}, as encoded by Pular connectives, cluster. To do so, unambiguous connectives with their functional labels are first annotated automatically. These include the following: 
\begin{itemize}
    \item \textit{si}, which is used `to refer to a proposed or potential action in the future by relating it to another action which should precede it' (\pgcitealt{pulargrammar}{61}). However, a closer look at all the occurrences of \textit{si} suggested that what is meant by \textsc{Future:Certain} in the overview are, in fact, non-past clauses with a universal (as opposed to existential) interpretation, as in (\ref{pularsi}).

    \begin{example}
    \textit{Kono \textbf{si} fewta e Joomiraa\AS{\m{d}}o on, taway tiggaare nden ittaama ka \AS{\m{b}}er\AS{\m{d}}e ma\AS{\m{b}}\AS{\m{b}}e}.\\
    But when one turns to the Lord, the veil is removed. (2 Corinthians 3:16)
        \label{pularsi}
    \end{example}
    
    The `logical' or `certainty' relation referred to be \citet{pulargrammar}, therefore, may have to do with the observation that what is described by the main eventuality occurs \textit{whenever} the eventuality described by the \textsc{when}-clause occurs. Recall that this is precisely the use of universal \textsc{when} commented on in the previous chapter. All these occurrences were therefore tagged as \textsc{NonPast:Universal}.
    \item \textit{\textipa{\textltailn}ande} and \textit{tuma} (on its own, i.e. without \textit{woo} following): \pgcitet{pulargrammar}{62} explains that `uncertainty has basically to do with the time the event will happen, not whether the event will happen (although inevitably there is also some implication that way as well)'. This time, a closer look at the relevant examples indicates that these are instead non-past existential \textsc{when}-clauses, as in (\ref{pulartuma}).

    \begin{example}
    \textit{\textbf{Tuma} o aroyi, o wee\AS{\m{b}}itanay aduna on fii junuubu e peewal e \textipa{\textltailn}aawoore}\\
    And when he comes, he will convict the world concerning sin and righteousness and judgment (John 16:8)
        \label{pulartuma}
    \end{example}

    These examples are therefore tagged as \textsc{NonPast:Existential}.
    \item \textit{\AS{\m{b}}ay}, verbal nouns (-\textit{ugol}) or \textit{tun}, and \textit{no...-irnoo}: as explained above, these all introduce sequential events in the past with different degrees of `suddenness' or `urgency'. Among these connectives, \textit{\textipa{\!b}ay} is the only frequent one, the others occurring relatively rarely in our dataset. All of them are therefore tagged as \textsc{Past:Sequence}.
    \item \textit{fewndo}-clauses were all tagged as \textsc{Past:Circumstance}. A closer look at all the occurrences indicates that this connective is used predominantly in existential, but stative/durative past clauses, as in (\ref{pularfewndo}).

    \begin{example}
        \textit{\textbf{Fewndo} o joo\AS{\m{d}}odi e ma\AS{\m{b}}\AS{\m{b}}e fii \textipa{\textltailn}aamugol, o \AS{\m{y}}etti bireedi on, o du'ii, o ta\AS{\m{y}}iti, o jonni \AS{\m{b}}e}.\\
        When he was at table with them, he took the bread and blessed and broke it and gave it to them. (Luke 24:30)
        \label{pularfewndo}
    \end{example}

\end{itemize}

Connectives with more than one meaning were then manually disambiguated depending on the context. These are \textit{tuma}...\textit{woo} and \textit{nde}, which can be used, according to \citet{pulargrammar}, in clauses referring to the future or to the past. However, \textit{tuma}...\textit{woo} \textit{can} convey repetition (i.e. a universal reading), not only in past but also in nonpast sentences, as (\ref{tumawoo}) clearly shows. In fact, no example of this construction in past sentences was found in the dataset, so all its occurrences were tagged as \textsc{NonPast:Universal}.

\begin{example}
    \textit{\textbf{Tuma} men hoynaa \textbf{woo}, men du'anoo. \textbf{Tuma} men cukkaa \textbf{woo}, men wakkiloo. \textbf{Tuma} men \textipa{\textltailn}o'aa \textbf{woo}, men jaabora no newori. Men woni tuundi oo aduna, men hawkaa e \AS{\m{d}}i fow, haa weetaango hande ngoo.}\\
    When reviled, we bless; when persecuted, we endure; when slandered, we entreat. We have become, and are still, like the scum of the world, the refuse of all things. (1 Corinthians 4:11-13)
    \label{tumawoo}
\end{example}

\textit{Nde}, which is widely represented in the dataset, was tagged as \textsc{Past:Background} when in past clauses, as indicated by \citet{pulargrammar}; however, it was also often found in pre-matrix position in past sentences, unlike what is stated by the author, as in (\ref{pularnde}c). Regardless of position, \textit{nde} in past sentences was assumed to introduce \textit{Past:Background}. A close look at the of \textit{nde} in non-past sentences showed that it is used both existentially, as in (\ref{pularnde}a) and universally (\ref{pularnde}b). These were tagged as \textsc{NonPast:Existential} and \textsc{NonPast:Universal}, respectively.

\begin{example}
    \begin{itemize}
        \item[a.] \textit{\AS{\m{d}}un, \textbf{nde} mi aroyi, mi wattiday e \AS{\m{b}}e su\AS{\m{b}}i\AS{\m{d}}on \AS{\m{b}}en dokke mon \AS{\m{d}}en, wondude e \AS{\m{b}}ataake, \AS{\m{b}}e na\AS{\m{b}}a Yerusalaam}\\
        And when I arrive, I will send those whom you accredit by letter to carry your gift to Jerusalem. (1 Corinthians 16:3)
        \item[b.] \textit{\textbf{Nde} ne\AS{\m{d}}\AS{\m{d}}o dolnu\AS{\m{d}}o, jom aalaaji ayni galle mun woo, jawle mun \AS{\m{d}}en da\AS{\m{d}}ay.}\\
        When a strong man, fully armed, guards his own palace, his goods are safe (Luke 11:21)
        \item[c.] \textit{\textbf{Nde} Iisaa hewtunoo e ley kin \AS{\m{y}}ibbehi, o tiggitii, o wi'i}\\
        And when Jesus came to the place, he looked up and said to him (Luke 19:5)
    \end{itemize}
    \label{pularnde}
\end{example}

\begin{figure}[!h]
\centering
\includegraphics[width=0.8\textwidth]{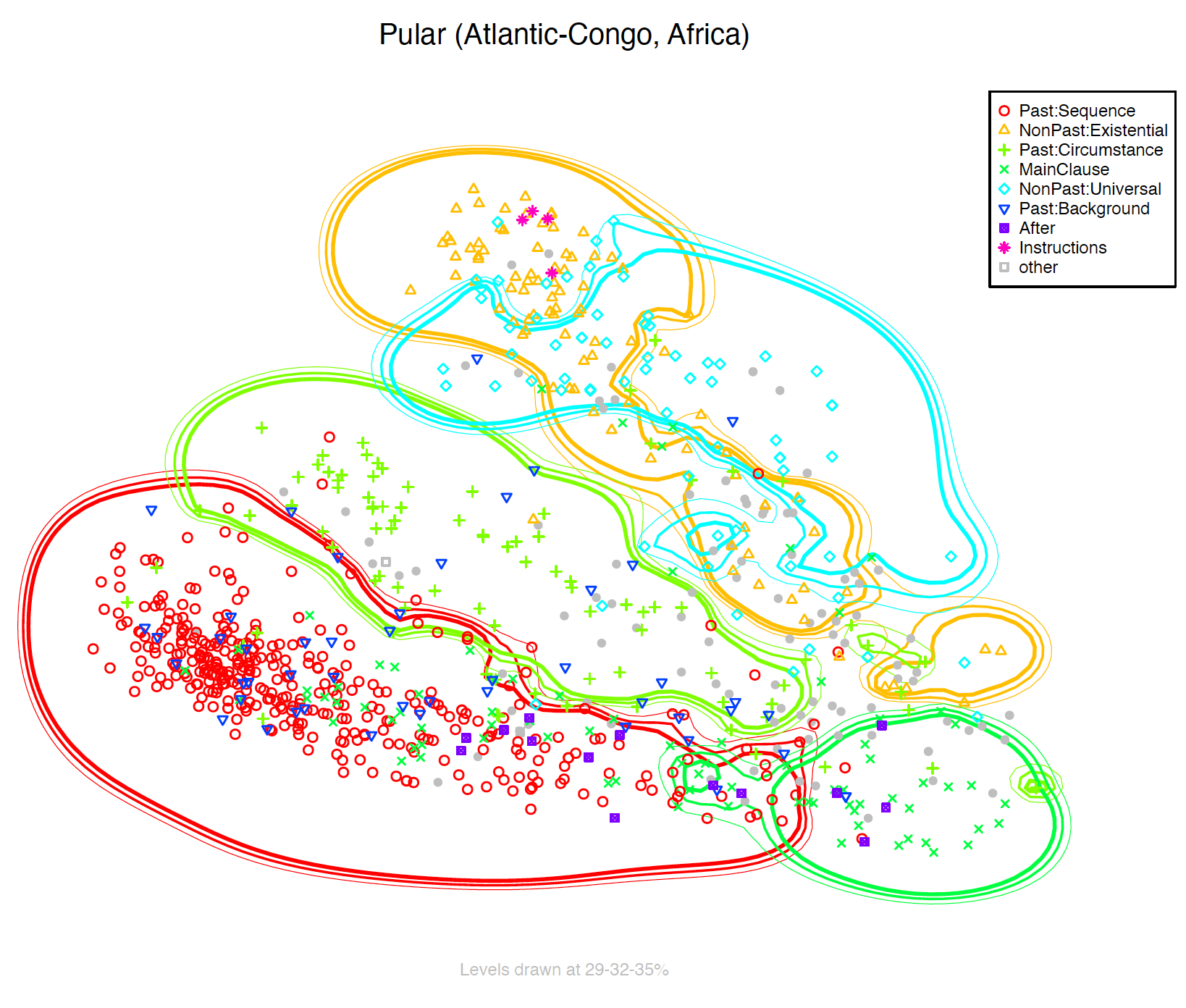}
\caption{\label{pular_withannot}Pular \textsc{when}-map with Kriging based on the functional annotation of Pular connectives}
\end{figure}

\textit{Ka}, expressing \textsc{Past:Circumstance} was left out because of a high degree of homophony with other instances of \textit{ka} and/or multifunctionality, which made it difficult to annotate it with confidence.\\
\indent Additionally, \textit{\textipa{\!b}awto} was annotated as \textsc{After} (see \citealt{pulardictio}), while \textit{ontuma} and \textit{onsay}, both meaning `at that time' (\pgcitealt{pulargrammar}{64}), were tagged as \textsc{MainClause} if no other connective was found, else as \textit{other}.\footnote{The annotated dataset is available in the project repository at \url{https://doi.org/10.6084/m9.figshare.24166254}.} \\
\indent The result of running Kriging on the new labels is shown in Figure \ref{pular_withannot}.

The new Kriging map shows better-defined areas for the functional scopes of the various Pular connectives. Besides confirming that both universal and existential non-past clauses cluster at the top of the map, it shows a more clear-cut separation between the top area and the bottom area by means of connectives conveying what \citet{pulargrammar} calls \textsc{Past:Circumstance}. The intermediate status of these constructions between sequential and non-sequential uses of \textsc{when}-clauses referred to by \citet{pulargrammar} thus finds clear confirmation in the intermediate position of this functional cluster between the \textsc{when}-clauses at the top of the map and those at the bottom in the semantic map. It also provides further confirmation that one of the main differences between the top and bottom area, the latter corresponding to the vast majority of absolute constructions and conjunct participles in the Old Church Slavonic map, is the stress on sequentiality conveyed by the situations corresponding to the data points at the bottom. We also observe that main clauses dominate the very bottom right corner, as expected, where Old Church Slavonic and Ancient Greek main clauses corresponding to English \textit{when} also cluster (cf. Figure \ref{averagevector}).\\
\indent It is important, however, to stress that cases such as Pular simply confirm that the bottom area of the semantic map, within which most participle constructions fall in Early Slavic, corresponds to a relatively well-defined function from the perspective of temporal relations, namely that between existential-eventive clauses, which translates into the sequentiality between events being stressed. What it does \textit{not} help clarify is the competing motivations governing the choice between \textit{when}-counterparts that also occur in the same area and participle clauses. In all likelihood, this is a matter of information structure and discourse organization, which the semantic map does not capture well. \\
\indent In Chapters 1-4, the analysis of the information-structural properties of subjects in \textit{jegda}-clauses and participle clauses suggested that \textit{jegda}-clause may be a better choice when there is a topic continuation from the immediately preceding discourse, as opposed to a topic shift, more likely to be signalled by some sentence-initial framing participle. This observation was prompted, among other things, by the fact that several \textit{jegda}-clauses have a null subject despite more frequently having a different subject from that of their matrix clause. As we will see in the next section, the topic-shift function of framing participles finds some cross-linguistic parallels in other types of null constructions, particularly those used with a clause-linkage function. However, other factors related to discourse organization and Time-Space-Participant-connectedness (Rothstein 2003, Fabricius-Hansen 2006) are at play, which different phenomena that have fallen so far under the generic notion of `null constructions' in the world's languages can help clarify.

\section{Null constructions in a cross-linguistics perspective}
As discussed in previous chapters, the discourse functions of conjunct participles in Ancient Greek and Early Slavic can partly be inferred compositionally from the relative order and tense-aspect of participle and matrix clause (\citealt{haug2012a}), allowing us to single out their usage as \textsc{independent rhemes}, which are discourse-coordinated and can be found stacked up as sequences of foregrounded clauses controlled by an argument of the matrix clause, as in the Old Church Slavonic and Ancient Greek example in (\ref{grcchain}a) and (\ref{grcchain}b).

\begin{example}
\begin{itemize}
\item[a.]
\gll i abie \textbf{tek\foreignlanguage{russian}{ъ}} edin\foreignlanguage{russian}{ъ} ot\foreignlanguage{russian}{ъ} nich\foreignlanguage{russian}{ъ}. i \textbf{priem\foreignlanguage{russian}{ъ}} gǫbǫ. \textbf{ispl\foreignlanguage{russian}{ь}n\foreignlanguage{russian}{ь}} oc\foreignlanguage{russian}{ь}ta. i \textbf{v\foreignlanguage{russian}{ь}znez\foreignlanguage{russian}{ъ}} na tr\foreignlanguage{russian}{ь}st\foreignlanguage{russian}{ь}. napaěše i
and immediately run.\textsc{ptcp.pfv.m.nom.sg} one.\textsc{m.nom.sg} from he.\textsc{gen.pl} and take.\textsc{ptcp.pfv.m.nom.sg} sponge.\textsc{acc.sg} fill.\textsc{ptcp.pfv.m.nom.sg} vinegar.\textsc{gen.sg} and put.\textsc{ptcp.pfv.m.nom.sg} on reed.\textsc{acc.sg} give.to.drink.\textsc{impf.3.sg} he.\textsc{acc.sg}
\glt
\glend
\item[b.]
\gll kai eutheōs \textbf{dramōn} heis ex autōn kai \textbf{labōn} spongon \textbf{plēsas} te oxous kai \textbf{peritheis} kalamō epotizen auton
and immediately run.\textsc{ptcp.pfv.m.nom.sg} one.\textsc{nom.sg} from he.\textsc{gen.pl} and take.\textsc{ptcp.pfv.m.nom.sg} sponge.\textsc{acc.sg} fill.\textsc{ptcp.pfv.m.nom.sg} with vinegar.\textsc{gen.sg} and put.\textsc{ptcp.pfv.m.nom.sg} reed.\textsc{dat.sg} give.to.drink.\textsc{impf.3.sg} he.\textsc{acc.sg}
\glt ‘Immediately one of them ran and took a sponge, filled it with sour wine and put it on a reed, and offered it to him to drink’ (Matthew 27:48)
\glend
\label{grcchain}
\end{itemize}
\end{example}

Pre-matrix perfective conjunct participles in New Testament Greek and Old Church Slavonic were singled out as most typically \textsc{independent rhemes}, although an (often ambiguous) \textsc{frame} reading can be available for sentence-initial conjunct participles. We have seen in Chapters 1 and 2 that absolutes regularly occur sentence-initially, often introducing a series of foregrounded clauses, and can instead be considered typical background material (i.e. \textsc{frames}) regardless of tense-aspect (cf. also \citealt{pedrazzinijhs}). 

\begin{figure}[!h]
\begin{subfigure}{0.50\textwidth}
\includegraphics[width=0.9\linewidth]{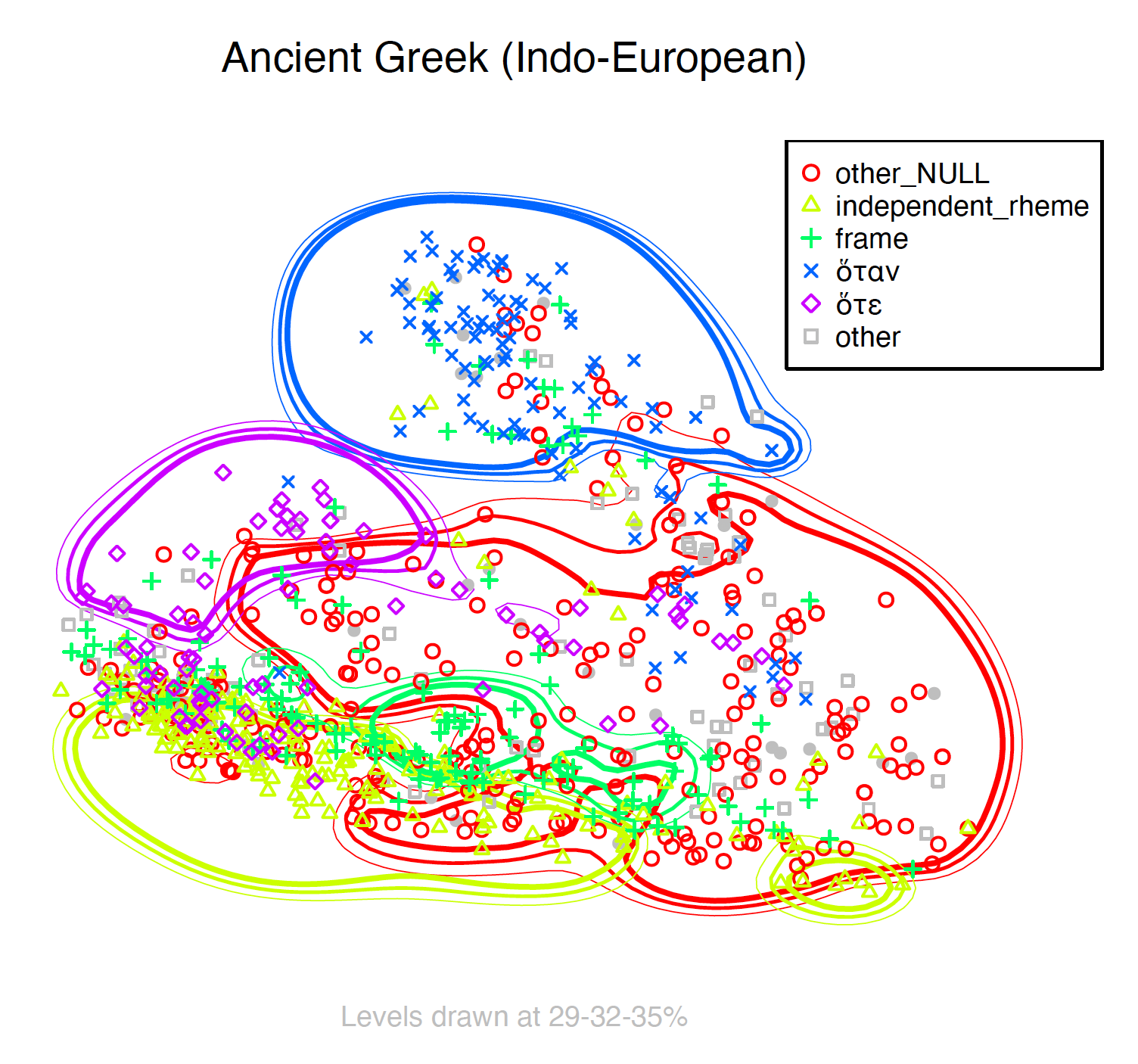} 
\caption[Ancient Greek \textsc{when}-map with Kriging based on functional labels, namely typical \textsc{independent rhemes} and typical \textsc{frames}]{}
\label{fig:proielgrc-withextralabels}
\end{subfigure}
\begin{subfigure}{0.50\textwidth}
\includegraphics[width=0.9\linewidth]{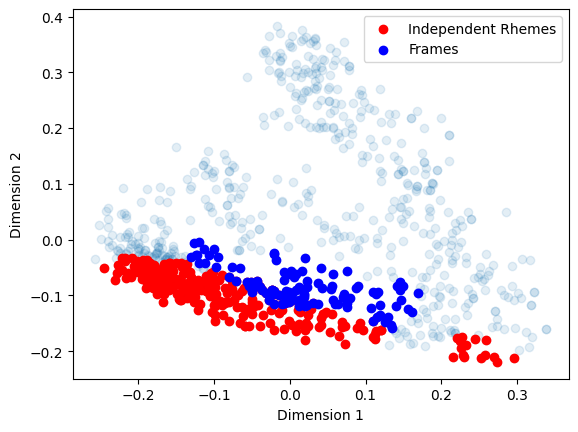}
\caption[Highlights of any point found inside the Kriging areas for \textsc{frames} and \textsc{independent rhemes} (red) in the Kriging map for Ancient Greek with added functional labels]{}
\label{fig:proielgrc-highlights}
\end{subfigure}
\caption[]{(a) Ancient Greek \textsc{when}-map with Kriging based on functional labels, namely typical \textsc{independent rhemes} (i.e. pre-matrix, perfective, with an aorist main verb) and typical \textsc{frames} (i.e. sentence-initial absolute constructions); (b) highlights of any point found inside the Kriging areas for \textsc{frames} (blue) and \textsc{independent rhemes} (red).}
\end{figure}

We can test this claim by identifying typical \textsc{independent rhemes} and \textsc{frames} among null alignments using the linguistic annotation in PROIEL, which may allow us to look at differences in where they cluster in the semantic map. Using Ancient Greek, because of its greater coverage of the New Testament, all pre-matrix aorist conjunct participles with an aorist main verb were tagged as \textsc{independent rhemes} and all sentence-initial absolute constructions as \textsc{frames}. This is reminiscent of what we did in Figure \ref{proielgrc-whenwhile}; this time, however, \textit{while}-situations are not included, and a specific configuration of conjunct participles (pre-matrix, aorist) is isolated as a construction of its own. After running Kriging on the newly labelled data points, we obtain the map in Figure \ref{fig:proielgrc-withextralabels} (remaining null alignments are labelled as `other\textunderscore NULL'). For ease of comparison, Figure \ref{fig:proielgrc-highlights} highlights all and only the null observations belonging to the Kriging areas for the identified typical \textsc{frames} and \textsc{independent rhemes}.

\begin{figure}[!h]
\begin{subfigure}{0.50\textwidth}
\includegraphics[width=0.9\linewidth]{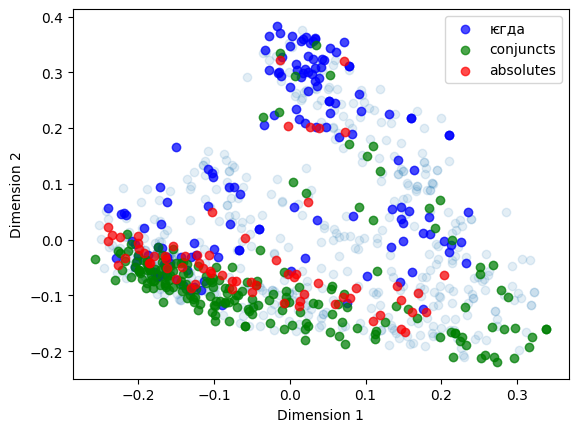} 
\caption[MDS maps of \textsc{when}, highlighting all \textit{jegda}-clauses, absolute constructions and conjunct participles in Old Church Slavonic]{}
\label{chucontinuum}
\end{subfigure}
\begin{subfigure}{0.50\textwidth}
\includegraphics[width=0.9\linewidth]{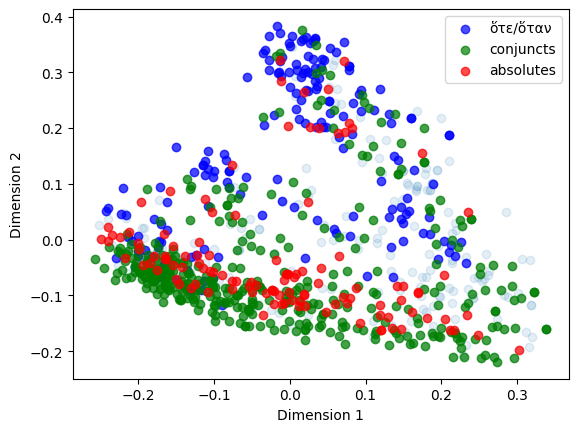}
\caption[MDS maps of \textsc{when}, highlighting all \textit{hote}/\textit{hotan}-clauses, absolute constructions and conjunct participles in Ancient Greek]{}
\label{grccontinuum}
\end{subfigure}
\caption[]{MDS maps of \textsc{when}, highlighting all finite subordinates (\textit{jegda} and \textit{hóte}/\textit{hótan}), absolute constructions and conjunct participles in (a) Old Church Slavonic and (b) Ancient Greek}
\end{figure}

As the figures show, typical \textsc{independent rhemes} and \textsc{frames} (or, in other terms, foreground and background matter, respectively) are predominant in a dedicated Kriging area at the bottom half of the map, stretching out from the area corresponding to GMM cluster 4 (i.e. the BL area) towards the right side of the map, where other, non-further-defined NULL constructions are found. \textsc{Independent rhemes} and \textsc{frames} each receive a contiguous, but relatively well-defined Kriging area. \\
\indent It is interesting to notice that typical \textsc{frames} are found above \textsc{independent rhemes} in the map, that is, closer to lexified \textsc{when}-clauses. \\
\indent As discussed in Chapter 5, the connective \textit{when} (and similarly \textit{lorsque} and \textit{quand} in the literature on French, as well as English \textit{while}) have been widely recognized as `triggers' or `clues' for backgrounding rhetorical relations in formal frameworks of discourse representation (\citealt{reeseetal, Asher2004,prevot2004,asher2007a}), namely as introducers of a background frame for a foregrounded event(uality). If an equivalence be made in discourse-structural terms between \textit{hóte}/\textit{hótan} and \textit{when} as \textit{Background}-triggers, then the relative greater closeness of \textsc{frames} to \textit{hóte} and \textit{hótan} in the map in Figure \ref{fig:proielgrc-withextralabels} adds a further layer of distinctions within the \textsc{when}-map---that between backgrounds and foregrounds (or \textsc{frames} and \textsc{independent rhemes}). If this were the case, then the semantic map for Ancient Greek is close to the idea of a continuum between the various constructions, including null ones---from typical foregrounded clauses at the bottom to typical backgrounded clauses at the top of the map. \\
\indent As Figures \ref{chucontinuum} and \ref{grccontinuum} show, even by simply highlighting finite \textit{when}-counterparts, absolute constructions and conjunct participles on the MDS map, we clearly see that absolute constructions occupy an intermediate position along the $y$-axis between the bulk of conjunct participles (at the very bottom of the map) and \textit{when}-counterparts (\textit{jegda}, \textit{hóte}/\textit{hótan}, at the top of the map).

\begin{figure}[!h]
\includegraphics[width=0.9\linewidth]{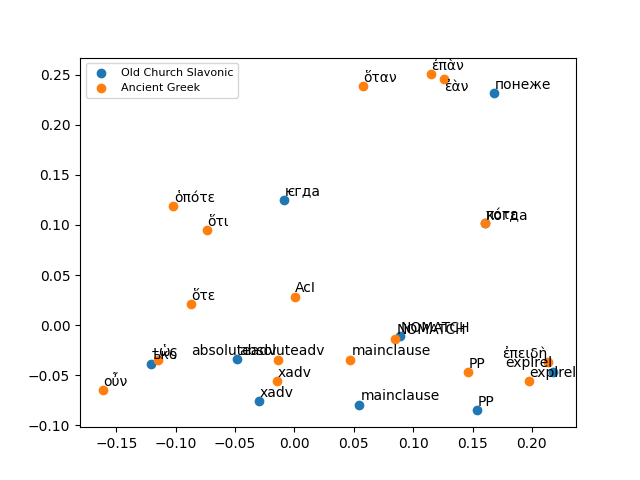} 
\caption[Location of the average observation for each different construction in Old Church Slavonic and Ancient Greek]{Location of the average observation for each different construction in Old Church Slavonic and Ancient Greek. \textit{AcI} = accusative with infinitive, \textit{absoluteadv} = absolute construction, \textit{xadv} = conjunct participle, \textit{PP} = prepositional phrase, \textit{explrel} = explicit relative.}
\label{averagevector}
\end{figure}

In a way, this continuum also reflects the intuition that the closer we get to the conjunct participles at the bottom of the map, the less subordinate-like the constructions are from the pragmatic perspective---from \textit{when}-counterparts at the top (most subordinate-like) to the conjunct participles at the bottom (least subordinate-like). At the same time, the constructions should be less and less subordinate-like moving rightwards along the $x$-axis, since the bottom right corner is where null constructions are found in \textit{all} languages in the dataset, albeit to very different extents. This is, in fact, reflected in the location of Ancient Greek and Old Church Slavonic independent clauses corresponding to \textit{when}-clauses in the semantic map. Figure \ref{averagevector} shows the average location of different constructions corresponding to \textit{when} in Ancient Greek and Old Church Slavonic. To obtain the map, the Ancient Greek and Old Church Slavonic counterparts to \textit{when} were first divided into groups, corresponding to different constructions in the two languages. Each construction was then represented on the map by averaging all the observations for that particular group. In this way, we can then check where in the map the average observation for a particular construction is found.

As expected, independent clauses are found, on average, towards the bottom right of the semantic map, similarly to prepositional phrases (e.g. \textit{v\foreignlanguage{russian}{ъ} syrě drěvě}, corresponding to English `when the tree is green', in Old Church Slavonic literally `in the moist [as opposed to dry] tree'), among others.\\
\indent The blurred boundaries between coordination and subordination when it comes to participle clauses had already been stressed by \citet{baryhaug2011} in their account of Ancient Greek conjunct participles. As we could already gather from the cross-linguistic pattern of concentration of null observations to the bottom half of the map (cf. the heatmap in Figure \ref{fig:nulls} in Chapter 5), this is by no means unique to Old Church Slavonic and Ancient Greek. The dependence of clauses on context for their interpretation as coordinate or subordinate structures is, in fact, a major feature of languages that make extensive use of converbal forms and/or that allow clause chaining and has thus been an important topic in the relevant literature (see, among others, \citealt{finer84,finer85,roberts1988,bickel91,bisang95,konig95,broadwell,longacre_2007,dooleychains,nonatophd}). The rest of this section touches on a number of (often interconnected) phenomena, observed in languages that appear to be typologically similar to Old Church Slavonic and Ancient Greek (i.e. with a similar split between lexified \textit{when}-counterparts and null constructions), or whose Kriging map only contains null constructions. Section \ref{sec:clausechaining} looks at clause chaining, which has already been mentioned in passing in previous sections, before looking at examples of bridging (or clause-linkage) (Section \ref{sec:bridging}), switch-reference (Section \ref{sec:switch}), and insubordination (Section \ref{sec:insub}). As we will see, the clear parallels to these phenomena, attested in genetically and areally unrelated languages, provide some independent evidence in support of the observations made so far on the functional scope of null constructions in the \textsc{when}-map, but will also provide some clarifications regarding the division of labour between absolute constructions and conjunct participles.

\subsection{Clause chaining}\label{sec:clausechaining}
\textit{Clause chaining} is the `possibility of long sequences of foreground clauses with operator dependence, typically within the sentence' (\pgcitealt{dooleychains}{3}). \textit{Operator dependence} refers to the fact that the clauses in clause chains are typically \textit{deranked} verb forms (\citealt{Stassen-1985,Croft-1990,Cristofaro-1998,cristofarosubordination}), that is, lacking marking of one or more tense, aspect, or mood distinctions (`operators') compared to independent clauses in the same language. They can be additionally marked by specific means signalling their status as `medial' clauses or `converbs', as the Avar and Korean examples in the previous chapter, for instance, showed. However, as is clear from \posscitet{dooleychains} definition, the presence of converbal morphology in a language (e.g. Italian or Ukrainian, as we saw in the previous chapter) does not necessarily entail the possibility of clause chaining, which specifically refers to the possibility of having sequences of \textit{foregrounded} clauses, namely clauses that function as independent predications from the discourse perspective. The Avar sentence in (\ref{avarex}) is a prototypical example of clause chaining.\footnote{All glossing in this chapter is my own unless otherwise stated.}

\begin{example}
 \label{avarex} Avar (North Caucasian)
    \gll \textit{Ładał} \textit{\textbf{ččun}} \textit{\textbf{vaqun}} \textit{hebsaġat} \textit{$\dot{q}$vaţive} \textit{\textbf{łuhun}} \textit{va\v{ç}ana} \textit{Ġisa}. \textit{Hebmexał} \textit{zobgi} \textit{\textbf{ķibiļizabun}}, \textit{mikkidul} \textit{suratalda} \textit{\textbf{boržun}} \textit{ba\v{ç}un}, \textit{Allahasul} \textit{Ru$\hbar$} \textit{Ġisaqe} \textit{reššţuneb} \textit{bixana}.
    water dip.\textsc{pfv.cvb} rise.\textsc{pfv.cvb} immediately out go.\textsc{pfv.cvb} behold Jesus then sky split.\textsc{pfv.cvb} dove image fly.\textsc{pfv.cvb} approach.\textsc{pfv.cvb} God Spirit Jesus rest see.\textsc{aor}
    \glt `And when Jesus was baptized, immediately he went up from the water, and behold, the heavens were opened to him, and he saw the Spirit of God descending like a dove and coming to rest on him' (Matthew 3:16).
    \glend
\end{example}

As \citet{dooleychains} observes, a clause chaining sentence may, and often does, \textit{also} contain backgrounded clauses, which can have the same type of operator dependence as the foregrounded clauses, that is, they can be morphologically identical to foregrounded clauses, but their content is presuppositional and serves to limit the scope of the main predication, similarly to how we have been defining \textsc{frames} in the previous chapters. The Old East Slavic and Mbyá Guaraní sentences in (\ref{chuchain}) and (\ref{guaranichain}), respectively, from our corpus exemplify this.

\begin{example}
    \gll i \textbf{priš\foreignlanguage{russian}{ь}d\foreignlanguage{russian}{ъ}} k\foreignlanguage{russian}{ъ} prpdb\foreignlanguage{russian}{ь}numu antoniju jego že \textbf{viděv\foreignlanguage{russian}{ъ}} i \textbf{pad\foreignlanguage{russian}{ъ}} pokloni sja jemu s\foreignlanguage{russian}{ъ} sl\foreignlanguage{russian}{ь}zami molja sja jemu da by u nego byl\foreignlanguage{russian}{ъ}
    {and} {come.{\sc ptcp.pfv.m.nom.sg}} {to} {saint.{\sc dat.sg}} {Antonij.{\sc dat}} {\sc 3.sg.m.gen} {\sc ptc} {see.{\sc ptcp.pfv.m.nom.sg}} {and} {fall.{\sc ptcp.pfv.m.nom.sg}} {bow.{\sc aor.3.sg}} {\sc refl} {\sc 3.sg.m.dat} {with} {tear.{\sc inst.pl}} {beg.{\sc ptcp.ipfv.m.nom.sg}} {\sc refl} {\sc 3.sg.m.dat} {so that} {be. {\sc aor.3.sg}} {at} {\sc 3.sg.m.gen} {be.{\sc ptcp.result.act.m.nom.strong}}
    \glt `When he came to St. Antonij and saw him, he prostrated himself and bowed, begging him with tears to let him stay with him' (\textit{Life of Feodosij Pečerskij}, Uspenskij Sbornik f. 31b) %278039
    \glend
    \label{chuchain}
\end{example}

\begin{example}
Mbyá Guaraní (Tupian)
    \gll {Ha'e nunga} ha'e kuery \textbf{oendu} \textbf{vy} ko'e\~mba'i ramove templo py \textbf{oike} \textbf{vy} nhombo'e
    {such things} they \textsc{coll} hear \textsc{ss} daybreak {right after} temple in enter \textsc{ss} teach
    \glt `And when they heard that, they entered the temple early in the morning and taught' (Acts 5:21)
    \glend
    \label{guaranichain}
\end{example}

In (\ref{chuchain}), the conjunct participles \textit{priš\foreignlanguage{russian}{ь}d\foreignlanguage{russian}{ъ}} (lit.) `having come' and \textit{viděv\foreignlanguage{russian}{ъ}} (lit.) `having seen' can be considered as background material (i.e. \textsc{frames}), unlike the morphologically identical participle \textit{pad\foreignlanguage{russian}{ъ}} (lit.) `having fallen', which belongs instead to the foreground, as also reflected in the English translation. Similarly, the first verb in the Mbyá Guaraní chain sentence in (\ref{guaranichain}) belongs to the background, while the following is part of the foreground.\\
\indent As \pgcitet{dooleychains}{12} observes, a clause in a chaining structure can be ambiguous between foreground and background, similarly to the ambiguity we have seen between \textsc{independent rheme} and \textsc{frame} participles sentence-initially, as the example in (\ref{ambig1}) shows.

\begin{example}
\begin{itemize}
    \item[a.]
    \gll i \textbf{povelěv\foreignlanguage{russian}{ъ}} narodom\foreignlanguage{russian}{ъ} v\foreignlanguage{russian}{ь}zlešti na trěvě i \textbf{priem\foreignlanguage{russian}{ъ}} pęt\foreignlanguage{russian}{ь} chlěb\foreignlanguage{russian}{ъ} i d\foreignlanguage{russian}{ь}vě rybě \textbf{v\foreignlanguage{russian}{ь}z\foreignlanguage{russian}{ь}rěv\foreignlanguage{russian}{ъ}} na nbo blgsvi. i \textbf{prělom\foreignlanguage{russian}{ъ}} chlěby dast\foreignlanguage{russian}{ъ} oučenikom\foreignlanguage{russian}{ъ}. oučenici že narodom\foreignlanguage{russian}{ъ}.
    {and} {command.{\sc ptcp.pfv.m.nom.sg}} {crowd.{\sc dat.pl}} {lay down.{\sc inf}} {on} {grass.{\sc loc}} {and} {take.{\sc sptcp.pfv.m.nom.sg}} {five.{\sc acc}} {bread.{\sc gen.pl}} {and} {two.{\sc acc}} {fish.{\sc acc}} {look up.{\sc ptcp.pfv.m.nom.sg}} {to} {heaven.{\sc acc}} {bless.{\sc aor.3.sg}} {and} {break.{\sc ptcp.pfv.m.nom.sg}} {bread.{\sc acc}} {give.{\sc aor.3.sg}} {disciple.{\sc dat.pl}} {disciple.{\sc nom.pl}} {\sc ptc} {crowd.{\sc dat.pl}}
    \glt
    \glend
    \item[b.]
    \gll {kai} {\textbf{keleusas}} {tous} {okhlous} {anaklithenai} {epi} {tou} {khortou} {\textbf{labōn}} {tous} {pente} {artous} {kai} {tous} {duo} {ikhthuas} {\textbf{anablepsas}} {eis} {ton} {ouranon} {eulogesen} {kai} {\textbf{klasas}} {edōken} {tois} {mathetais} {tous} {artous} {hoi} {de} {mathetai} {tois} {okhlois}
    {and} {command.{\sc ptcp.pfv.m.nom.sg}} {the.{\sc acc.pl}} {crowd.{\sc acc.pl}} {lay down.{\sc inf.pas.aor}} {on} {the.{\sc gen}} {grass.{\sc gen}} {take.{\sc ptcp.pfv.m.nom.sg}} {the.{\sc acc.pl}} {five} {bread.{\sc acc.pl}} {and} {the.{\sc acc.pl}} {two} {fish.{\sc acc.pl}} {look up.{\sc ptcp.pfv.m.nom.sg}} {to} {the.{\sc acc}} {heaven.{\sc acc}} {bless.{\sc aor.3.sg}} {and} {break.{\sc ptcp.pfv.m.nom.sg}} {give.{\sc aor.3.sg}} {the.{\sc dat.pl}} {disciple.{\sc dat.pl}} {the.{\sc acc.pl}} {bread.{\sc acc.pl}} {the.{\sc nom.pl}} {\sc ptc} {disciple.{\sc nom.pl}} {the.{\sc dat.pl}} {crowd.{\sc dat.pl}}
    \glt `Then he ordered the crowds to sit down on the grass, and taking the five loaves and the two fish, he looked up to heaven and said a blessing. Then he broke the loaves and gave them to the disciples, and the disciples gave them to the crowds' (Matthew 14:19) %38887
    \glend
\end{itemize}
\label{ambig1}
\end{example}

In (\ref{ambig1}), the series of conjunct participles could be seen as ambiguous between background and foreground material. This is also reflected in the different English translations---the English Standard Version (given under the example), for instance, interprets the clause corresponding to the first conjunct participle as an independent clause (\textit{he ordered the crowds}), the second as a subordinate (\textit{taking the five loaves}), and all the others as independent clauses (\textit{he looked up... said a blessing ... broke the loves ... gave them to the disciples}). Other versions show different interpretations. For example, the King James Version (and similarly NKJV) translates the first two clauses as main clauses (\textit{he commanded... and took the five loaves}), the third as subordinate (\textit{looking up}), then all the remaining ones as main clauses (\textit{he blessed, and brake, and gave}).

\begin{figure}[!h]
\begin{subfigure}{0.50\textwidth}
\includegraphics[width=0.9\linewidth]{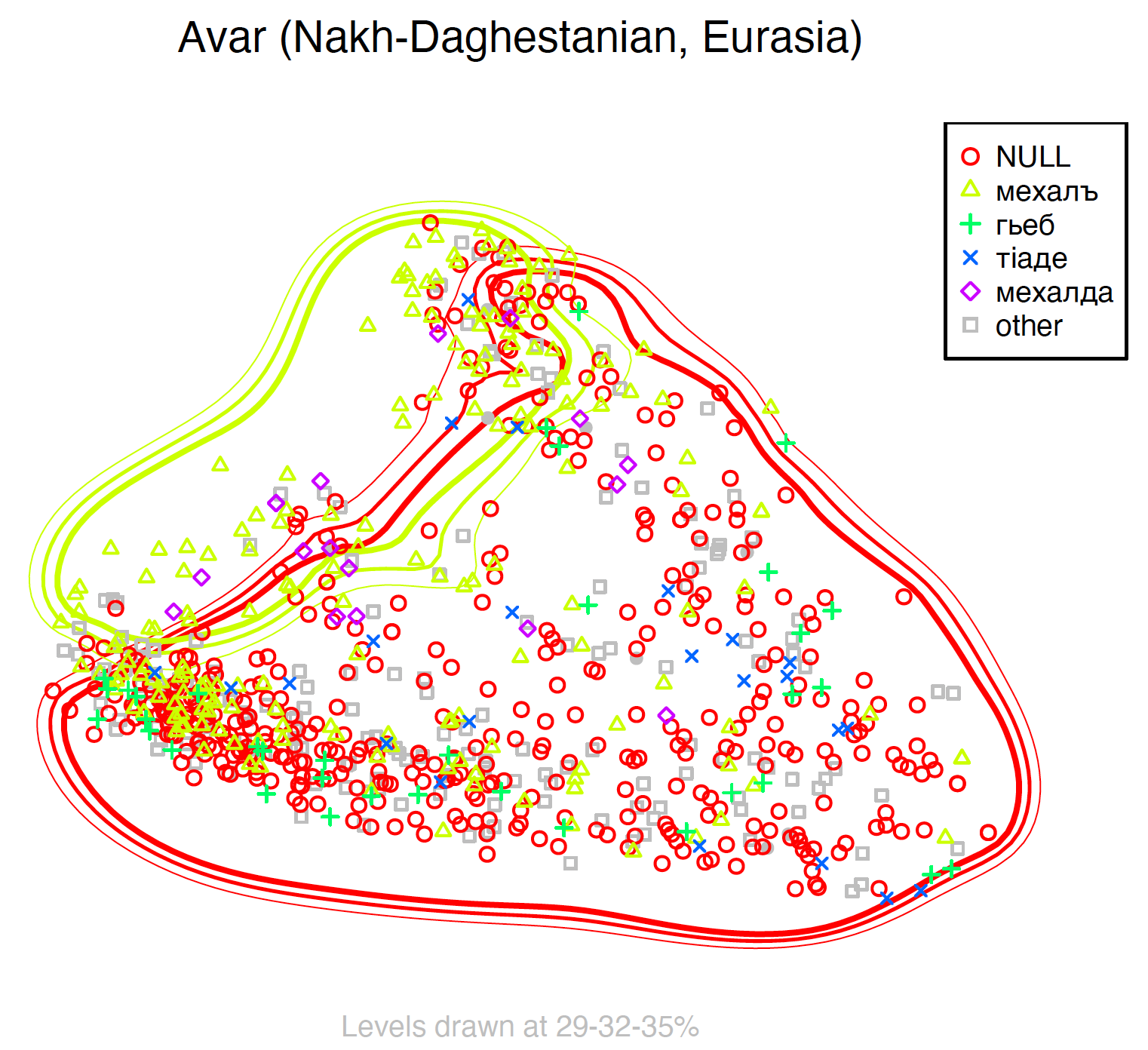} 
\caption[Kriging map for Avar (Nakh-Daghestanian, Eurasia)]{}
\label{avar}
\end{subfigure}
\begin{subfigure}{0.50\textwidth}
\includegraphics[width=0.9\linewidth]{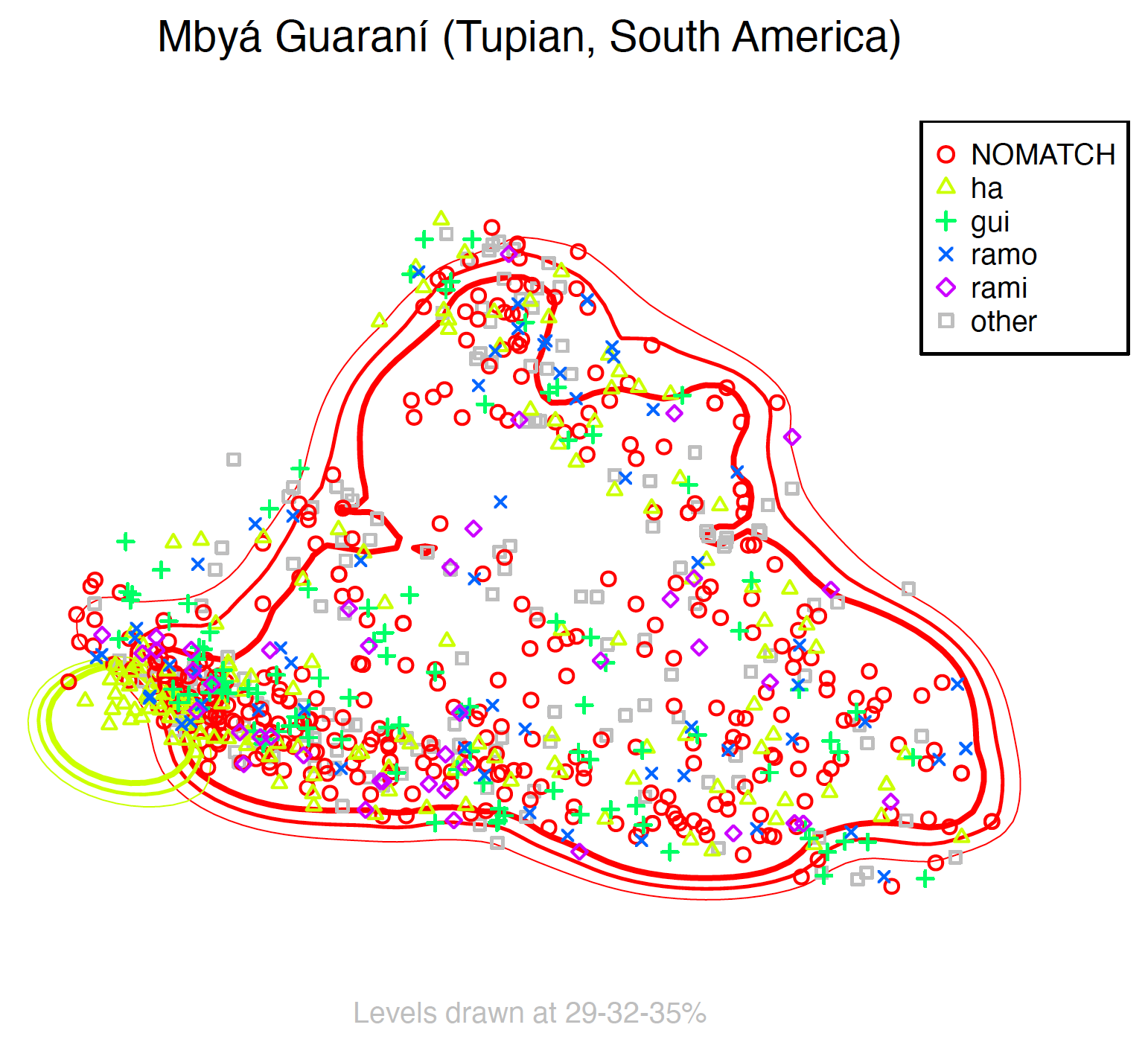}
\caption[Kriging map for Mbyá Guaraní (Tupian, South America)]{}
\label{guarani}
\end{subfigure}
\caption[]{Kriging maps for Avar (Nakh-Daghestanian, Eurasia) and Mbyá Guaraní (Tupian, South America)}
\end{figure}

As pointed out by \pgcitet{dooleychains}{13}, not every language with a chaining construction is necessarily a clause-chaining language. Whichever the case for Early Slavic, there are clear similarities in the use of chaining constructions in clause-chaining languages such as Avar and Mbyá Guaraní and series of conjunct participles in Old Church Slavonic. Even among clause-chaining languages, there are major differences in the functions of chaining constructions, as is also reflected in the different maps for Mbyá Guaraní (Figure \ref{guarani}) and Avar (Figure \ref{avar}), despite both being clause-chaining languages.\\
\indent Several authors who have proposed explanations for the use of clause chaining as a narrative strategy have put forward criteria that could be subsumed under a general principle of \textit{event grouping}, as is referred to by \pgcitet{dooleychains}{14}. In a nutshell, the assumption is that the \textit{syntactical} integration of clauses in narrative chaining reflects some level of \textit{conceptual} conflation as well (\pgcitealt{dooleychains}{14}; cf. also \pgcitealt{chafe94}{145}; \citealt{givon2001}; \citealt{talmy91}). A chaining construction may thus constitute a complete discourse or `developmental' unit (Levinsohn 2009). In \pgposscitet{thurmanchuave}{344} words:

\begin{quote}
Primarily final verbs [i.e. the finite verb at the end of a chaining construction, A/N] are used to alter the perspective on the topic at hand by terminating one chain so that another may begin. This may mean a change in participant orientation so that another character in the narrative becomes prominent; or it may involve reorientation only in time or space.
\end{quote} 

Similarly, \citet{fabhansen2007} and \citet{sabo2011} argued that absolute constructions, and similarly co-predicative participles, provide a guarantee of tense-space-participant connectedness and `serve to build groups of events or states, expressing that the host and supplement eventualities form interesting sums of eventualities' whereby `the core event and co-eventualities all add up to one super-group eventuality' (\pgcitealt{sabo2011}{1438}).

\subsection{Bridging}\label{sec:bridging}
Background clauses in clause chains, just like \textsc{frames}, are presuppositional (in the sense of \pgcitealt{lambrecht1994a}{36-73}), that is, they contain reference to given or accessible information and, because they can link back to the previous discourse, they often work as bridging devices. It is not uncommon for languages with clause chaining to use bridging systematically at the beginning of most chaining constructions by repeating the content of the previous foregrounded clause and re-purposing it, as it were, as the background for a new clause chain. This is a phenomenon variously referred to, among others, as \textit{bridging constructions} (e.g. \citealt{guerinbridge}), \textit{tail-head linkage} (e.g. \citealt{devrieslinkage}), \textit{recapitulation clauses} (e.g. \citealt{stirling_1993}), summary-head linkage (\citealt{thompsonadverbialclauses}), or simply \textit{clause linkage} (\citealt{anderboisaltshuler22}). While bridging is not used as systematically in Early Slavic as in some of these languages, we have seen how we frequently find \textsc{frames} using either previously foregrounded eventualities or event participants as part of the background for a new portion of foregrounded discourse to come (cf. Chapter 2, Section \ref{secbdinski}).
Bridging as a systematic discourse-organization device follows precisely this overall principle, as is clear from its definition in \pgcitet{guerinbridge}{2--3}:

\begin{quote}
    A bridging construction is a linkage of three clauses. The first clause of the construction (i.e., the reference clause) is the final clause in a unit of discourse. The second clause (i.e., the bridging clause) recapitulates the reference clause. It usually immediately follows the reference clause but it acts as the initial (albeit non-main) clause of a new discourse unit. The primary discourse function of a bridging construction is to add structure and cohesion: recapitulation backgrounds the proposition of the reference clause and foregrounds the clause following the bridging clause. This third clause is discourse-new and typically sequentially ordered.
\end{quote}

\begin{figure}[!h]
\centering
\includegraphics[width=0.6\linewidth]{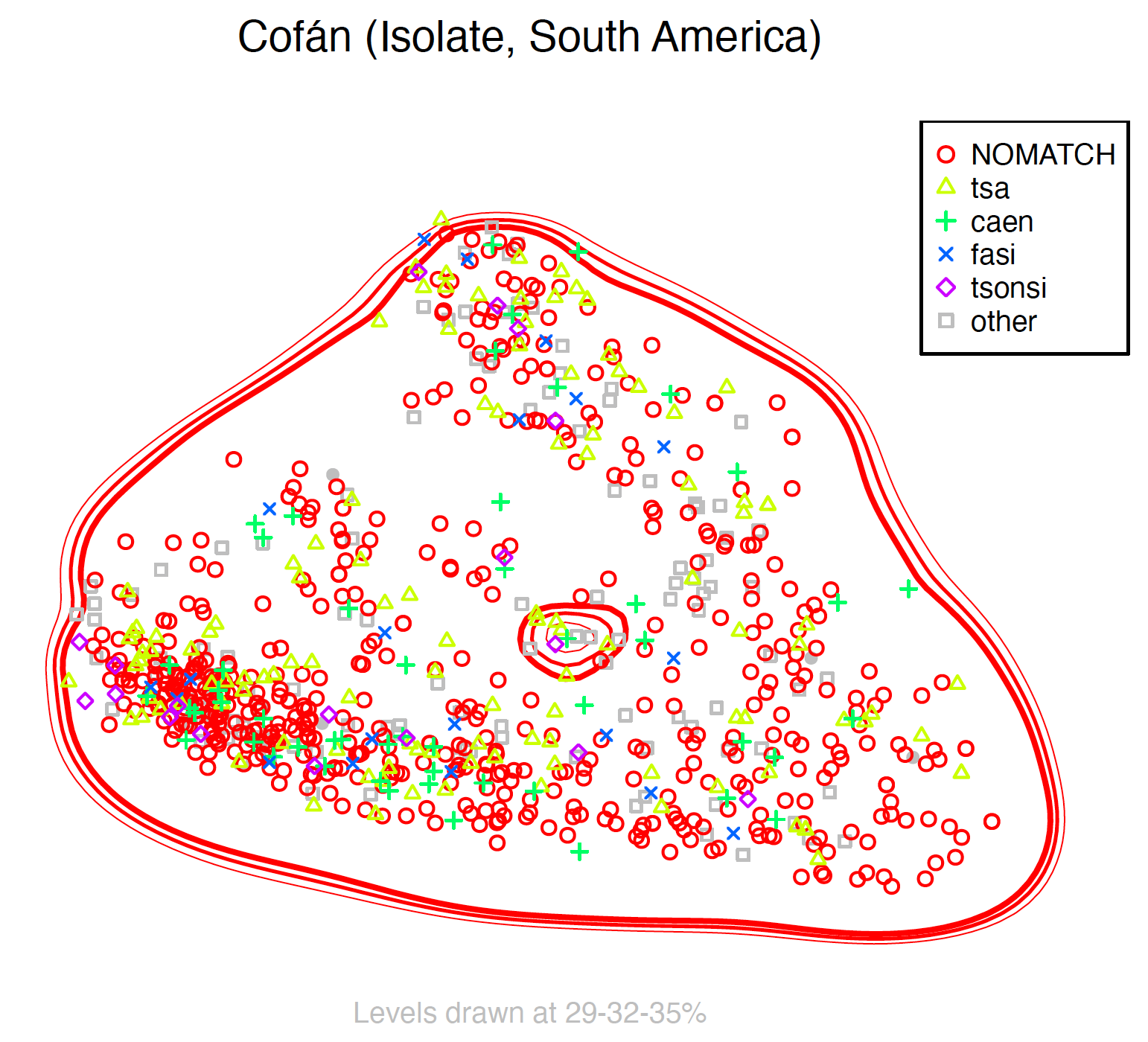}
\caption{Kriging map for Cofán (Isolate, South America)}
\label{cofanfig}
\end{figure}

Or, in \pgposscitet{anderboisaltshuler22}{802} schematization:

\begin{quote}
    [ \textsc{Reference} ]$_{\textsc{R}}$. [[ \textsc{Bridge} ]$_{\textsc{B}}$ \textsc{Continuation} ]$_{\textsc{C}}$
\end{quote}

where \textbf{R} is the reference clause as per the definition by \citet{guerinbridge} above, \textbf{B} is the bridging clause which, together with the discourse unit following the bridging clause (the third, foregrounded clause of \possciteauthor{guerinbridge} definition), constitute the continuation clause (\textbf{C}).\\
\indent \citet{anderboisaltshuler22} studied the phenomenon in A’ingae (Cofán; see the map in Figure \ref{cofanfig}, where only \textit{null} constructions have a Kriging area) and gave an SDRT account of the rhetorical relations systematically triggered by bridging constructions. 

According to them, bridging requires the relations \textit{Narration}(R,[B,C]) and Back\-ground\-$_{Forward}$(B,C). In other words, the reference clause attaches to the complex discourse unit made of the bridging clause and continuation clause via \textit{Narration}, whereas the bridging clause attaches to the continuation clause via \textit{Background$_{Forward}$}. This is the same principle that we observed in the case study on strategically annotated treebanks in Chapter 2, specifically when we observed that dative absolutes appeared to systematically repeat \textit{some} portion of a previously foregrounded unit, either the immediately preceding clause or some clause not far behind in the discourse. This is what we see in (\ref{bridge1}), repeated from Chapter 2, and (\ref{bridge2})-(\ref{bridge3}).

\begin{example}
\gll {\normalfont[}{\textbf{prizovi}} mi ju{\normalfont]}$_{{\normalfont \textbf{R} }}$ da se poveselju dn\foreignlanguage{russian}{ь}s s njeju. zělo bo jako slyšal\foreignlanguage{russian}{ь} jesm o njei godě mi jest\foreignlanguage{russian}{ь}. {\normalfont[}{\normalfont[}{\textbf{prizvaně}} že {\textbf{byvši}} ei{\normalfont]}$_{{\normalfont \textbf{B} }}$. {pride} k njemu{\normalfont]}$_{{\normalfont \textbf{C} }}$
summon.\textsc{imp} \textsc{1.sg.dat} \textsc{3.sg.f.acc} {so that} \textsc{refl} enjoy.\textsc{prs.1.sg} today with \textsc{3.sg.f.inst} much for that hear.\textsc{ptcp.result.m.sg} be.\textsc{prs.1.sg} about \textsc{3.sg.f.loc} pleasant.\textsc{adv} \textsc{1.sg.dat} be.\textsc{prs.3.sg} summon.\textsc{ptcp.pfv.pass.f.dat.sg} \textsc{ptc} be.\textsc{ptcp.pfv.f.dat.sg} \textsc{3.sg.dat} come.\textsc{aor.3.sg} to \textsc{3.sg.dat}
\glt ‘``Summon her, so that I can enjoy myself with her today. From what I've heard of her, I am much attracted by her”. After being summoned, she came to him.’ (\textit{Life of Mary, Abraham's Niece}, Bdinski Sbornik f. 7v)
\glend
\label{bridge1}
\end{example}

\begin{example}
\gll n\foreignlanguage{russian}{ъ} sice s\foreignlanguage{russian}{ъ}tvorivě da {\normalfont[}az\foreignlanguage{russian}{ъ} ti \textbf{ljagu} \textbf{na} {\textbf{vozě}{\normalfont]}$_{{\normalfont \textbf{R} }}$} ty že mogyi na koni ěchati. to že blaženyi s\foreignlanguage{russian}{ъ} v\foreignlanguage{russian}{ь}sjakyim\foreignlanguage{russian}{ь} s\foreignlanguage{russian}{ъ}měrenijem\foreignlanguage{russian}{ь} v\foreignlanguage{russian}{ъ}stav\foreignlanguage{russian}{ъ} sěde na koni. {\normalfont[}{\normalfont[}a onomu že \textbf{leg\foreignlanguage{russian}{ъ}šju} {\textbf{na}} {\textbf{vozě}{\normalfont.]}$_{{\normalfont \textbf{B} }}$} i {idjaše} put\foreignlanguage{russian}{ь}m\foreignlanguage{russian}{ь} raduja sę i slavja ba.{\normalfont]}$_{{\normalfont \textbf{C} }}$ 
{but} {thus} {do.{\sc imp.1.du}} {that} {\sc 1.sg.m.nom} {\sc ptc} {lie down.{\sc prs.1.sg}} {on} {chariot.{\sc loc.sg}} {\sc 2.sg.nom} {\sc ptc} {be able.{\sc ptcp.ipfv.m.nom}} {on} {horse.{\sc loc.sg}} {ride.{\sc inf}} {so} {\sc ptc} {blessed.{\sc m.nom.sg}} {with} {all.{\sc n.inst.sg}} {humility.{\sc n.inst.sg}} {stand up.{\sc ptcp.pfv.m.nom.sg}} {sit.{\sc aor.3.sg}} {on} {horse.{\sc loc.sg}} {and} {that.{\sc dat.sg}} {\sc ptc} {lie down.{\sc ptcp.pfv.m.dat.sg}} {on} {chariot.{\sc loc.sg}} {and} {go.{\sc impf.3.sg}} {way.{\sc inst.sg}} {rejoicing.{\sc ptcp.ipfv.m.nom.sg}} {\sc refl} {and} {glorify.{\sc ptcp.ipfv.m.nom.sg}} {god.{\sc gen}}
\glt `[Izyaslav:] ``But let's do this: I'll lie down in the chariot, and you can ride the horse". The Blessed Theodosius humbly got up and mounted his horse. And as he [Izyaslav] lay down in the chariot, he [Theodosius] continued on his way, rejoicing and glorifying God.' (\textit{Life of Feodosij Pečerskij}, Uspenskij Sbornik f. 43b) % (Usp.sbor. 26) %278505
\glend
\label{bridge2}
\end{example}

\begin{example}
\gll i došed\foreignlanguage{russian}{ъ} velika dne. vskrsn\foreignlanguage{russian}{ь}ja po obyčaju. prazdnova světlo. {\normalfont[}\textbf{vpade} \textbf{v} {\textbf{bolězn\foreignlanguage{russian}{ь}}{\normalfont.]}$_{{\normalfont \textbf{R} }}$} {\normalfont[}{\normalfont[}\textbf{razbolěvšju} \textbf{bo} \textbf{sę} {\textbf{jemu}{\normalfont.]}$_{{\normalfont \textbf{B} }}$} i bolěvšju. dnii .e. po sem\foreignlanguage{russian}{ь} byvšju večeru. {povelě} iznesti sę na dvor\foreignlanguage{russian}{ъ}{\normalfont.]}$_{{\normalfont \textbf{C} }}$
{and} {reach.{\sc ptcp.pfv.m.nom.sg}} {great.{\sc m.gen.sg}} {day.{\sc m.gen.sg}} {resurrection.{\sc n.gen.sg}} {according to} {custom.{\sc dat.sg}} {celebrate.{\sc aor.3.sg}} {joyfully} {fall.{\sc aor.3.sg}} {in} {illness.{\sc acc}} {fall ill.{\sc ptcp.pfv.m.nom.sg}} {\sc ptc} {\sc refl} {\sc 3.sg.m.dat} {and} {be ill.{\sc ptcp.pfv.m.dat.sg}} {day.{\sc gen.pl}} {five} {after} {this.{\sc loc.sg}} {be.{\sc ptcp.pfv.m.dat.sg}} {evening.{\sc dat.sg}} {command.{\sc aor.3.sg}} {carry.{\sc inf}} {\sc refl} {in} {yard.{\sc acc}}
\glt `But when the great day of Easter arrived [lit. when he reached the great day of Easter], according to custom, he celebrated it joyfully. He fell ill. Having fallen ill and having been ill for five days, once in the evening he ordered to be carried out to the yard' (PVL, 62d) % (Lav. 186.5–8) %266811
\glend
\label{bridge3}
\end{example}
% И дошед\foreignlanguage{russian}{ъ} велика дн҃е. вскрсн\foreignlanguage{russian}{ь}я по ѡбычаю. 186.5празднова свѣтло. |53впаде в болѣзн\foreignlanguage{russian}{ь}. |54разболѣвшю 186.6бо сѧ ѥму. и болѣвшю. дн҃ии .е҃. 186.7По сем\foreignlanguage{russian}{ь} бывшю вечеру. повелѣ изнести сѧ на 186.8двор\foreignlanguage{russian}{ъ}

In (\ref{bridge1}), the dative absolute \textit{prizvaně byvši} `being summoned' repeats the main verb of the previous sentence (\textit{prizovi} `summon'), repurposing it as the background for the new portion of the discourse (\textit{pride} `she came'), sequentially located \textit{after} the summoning event. This could be analysed following the structure described by \citet{guerinbridge} and \citet{anderboisaltshuler22}, where \textit{prizovi} is part of the reference clause, \textit{prizvaně byvši} is the bridging clause and \textit{pride} is part of the continuation clause. Very similarly, in (\ref{bridge2}), \textit{ljagu} `I will lie down' can be seen as the reference clause, \textit{onomu že leg\foreignlanguage{russian}{ъ}šju} `having lain down' as the bridging clause, and \textit{idjaše} `he went' as the continuation clause. Likewise, in (\ref{bridge3}), \textit{vpade v bolězn\foreignlanguage{russian}{ь}} `he got sick' would be the reference clause, the immediately following \textit{razbolěvšju bo sę jemu} `having gotten sick' the bridging clause, and \textit{povelě} `he commanded' as the continuation clause.\\
\indent As the simple example in (\ref{cofanbridge}) shows, bridging clauses in A'ingae are introduced by subordinators with switch-reference morphology, indicating whether the subject of the bridging clause is the same or different from that of the continuation clause (cf. \citealt{cofansubord,anderboisaltshulersilva23}).\footnote{As we will see in more detail in Section \ref{sec:switch}, `switch-reference' does not specifically only refer to different-subject marking, as the term has sometimes been used in the literature on Early Slavic dative absolutes (e.g. \citealt{collins2011a}), but it is the general term used to describe the phenomenon whereby clauses are marked according to whether their subject is co-referential or not with that of the matrix (i.e. it includes both different-subject and same-subject markers).}

\begin{example}
\gll Tse'faei'ccuyi {\normalfont[}aje \textbf{chattian.}{\normalfont]}$_{{\normalfont \textbf{R} }}$ {\normalfont[}{\normalfont[}Aje {\textbf{chattian-si}{\normalfont]}$_{{\normalfont \textbf{B} }}$} {tsa} pushesuja poiyi'cconga {cui'ña}{\normalfont]}$_{{\normalfont \textbf{C} }}$.
immediately fever alleviate fever alleviate-\textsc{ds} \textsc{ana} woman all serve
\glt `and the fever left her, and she began to serve them' (Matthew 8:15)
\glend
\label{cofanbridge}
\end{example}

Marking bridging clauses for switch reference is very common among languages where such marking is available (cf. \citealt{stirling_1993,devrieslinkage,guerinbridge}). In fact, we find many similarities between the use of participle constructions in Early Slavic and switch reference systems of several language families, most notably many Papuan (\citealt{robertsswitch97}) and South American (\citealt{gijnetalsouthame,vangijn2012southame,vangijn2016srsouthame,overall2014,overall2016}) families, beyond bridging constructions. We now therefore turn to looking at some of these parallels in more detail.

% - On Kotiria (concerning intersection between backgrounding, clause linkage and SR): "Longacre (1985), as well as Thompson, Longacre and Hwang (2007) propose that in many languages, series of sentences in discourse can be viewed as comprised of nuclei – main clauses that express the principal events in the storyline – with other structural units, such as adverbial subordinate clauses, surrounding, or ‘draped around’ their margins. Some of the principal functions of these surrounding or peripheral clauses are to promote discourse cohesion and organize competing perspectives of events (Thompson \& Longacre 1985: 206). In such discourse, one of the frequently observed patterns of use of adverbial clauses is in ‘tail-head linkage’ (THL), and we will see that it is exactly in the context of THL that overt SR marking in Kotiria discourse is most frequently observed." (Stenzel in Gijn, R. V., \& Hammond, J. (Eds.). (2016). Switch reference 2.0, p. 436)

% - [An interesting observation is made in Bickel (1999) who hypothesizes that many synchronic SR systems have evolved from absolute constructions. In fact, classical absolute constructions as in Ancient Greek (as well as other languages) come close to SR systems because they are in “pragmatic competition with conjunct participles (participia coniuncta) that show case agreement with a coreferential argument of the matrix, occupy roughly the same adsentential position as absolutes, and fulfill a similar discourse function” (ibid.: 46).] (p. 41 gijn)

\subsection{Switch reference}\label{sec:switch}
In a nutshell, \textit{switch reference} is a morphological system for tracking referents in an ongoing discourse (\pgcitealt{roberts_2017}{538}). In a `canonical' switch-reference system (cf. \pgcitealt{haimanmunro}{ix}), a clause is marked to signal whether its subject is co-referential or not with the subject of another, usually adjacent, clause. With subject co-reference, a same-subject marker is used (\textsc{ss}), else a different-subject marker is employed (\textsc{ds}). Switch reference is overwhelmingly present in languages that allow and use clause chaining extensively, in which case switch reference marking would typically occur on each medial clause leading up to the final verb. (\ref{canonsr}) is an example of `canonical' switch reference in Huichol-Wixárika from our corpus.\footnote{In the Huichol example, I kept the spelling of the Bible translation in \posscitet{mayer-cysouw} corpus. Note, however, that this is not the most common orthography found in most studies on Huichol today.} In (\ref{canonsr}a), \textit{me-'u'-axüa-cu} has the \textsc{ds} marker \textit{-cu}, indicating that its subject referent differs from that of its matrix clause. Conversely, \textit{me-'u'-axüa-ca} (\ref{canonsr}b) has the \textsc{ss} marker \textit{-ca}, indicating co-referentiality with the matrix subject.

\begin{example}
Huichol-Wixárika (Uto-Aztecan)
\begin{itemize}
\item[a.]
\gll Hesüana \textbf{me-'u'-axüa-cu} müpaü ti-ni-va-ru-ta-hüave 
to.him \textsc{3.pl.sbj-vis-}arrive\textsc{.pl-ds} thus \textsc{distr-narr-3.pl.nsbj-pl-sg-}say
\glt `When they came to him he said to them' (Acts 20:18)
\glend
\item[b.]
\gll Hesüana \textbf{me-'u'-axüa-ca} müme, müpaü me-te-ni-ta-hüave
to.him \textsc{3.pl.sbj-vis-}arrive\textsc{.pl-ss} men thus \textsc{3.pl.sbj-distr-narr-sg-}say
\glt ‘When the men had come to him they said’ (Luke 7:20)
\glend
\label{canonsr}
\end{itemize}
\end{example}

However, definitions of switch reference based on canonical switch reference and on purely morphosyntactic criteria (most notably \citealt{finer84,finer85}) have now long been challenged and revisited to account for the fact that several languages use switch reference markers regardless of subject co-reference, to indicate some other type of (dis)continuity. This discussion is extraordinarily reminiscent of previous literature on the functions and origin of Early Slavic participle constructions,\footnote{\pgcitet{corin1995a}{268}, for example, argued that `exceptions to the subject non-identity condition imply neutralization of the sole syntactic feature which differentiates the function of absolute and non-absolute participial constructions', and similarly \pgcitet{ve1996}{190}; \pgcitet{bauer}{280} considers the lack of co-reference with the matrix subject in Old Church Slavonic as evidence that `absolute constructions were fully understood' and therefore not a Greek imitation. \citet{collins2004a,collins2011a} instead challenges views based on allegedly `canonical' constraints on subject non-coreference, providing instead possible explanations based on discourse-structuring criteria.} and the parallels between the use of participles in Early Slavic and switch-reference systems in the world's languages are, in fact, several. \\
\indent In her seminal account of switch reference and discourse representation, \citet{stirling_1993} defines switch reference more broadly as a `clause level function which does not deal with the reference of NPs as such but with degrees and types of cohesion between eventualities' (\pgcitealt{stirling_1993}{136}; similarly \citealt{mithun} and more recently \citealt{mckenzie2012,mckenzie2015a,mckenzie2015b} and \citealt{keine}) so that while referential (dis-)continuity is the basic meaning marked by all switch-reference systems, these may be seen as `types of clause linkage rather than as mere devices of referential tracking' (\pgcitealt{stirling_1993}{151}). She identifies at least six main factors, or `pivots', playing a role in the use of switch-reference markers among the world's languages, which provide a useful basis for comparison with the different contexts in which Early Slavic dative absolutes as opposed to conjunct participles are used:

\begin{itemize}
    \item[(1)] Referential (non-)identity between one or more NPs. This largely corresponds to \posscitet{haimanmunro} definition of canonical switch reference, although Stirling's approach to referential continuity also encompasses cases in which what is tracked are syntactic roles other than subjects (e.g. objects) or semantic roles (e.g. agents). 
    \item[(2)] The agentivity of important event participants: `the SS/DS markers signal not simply continuity of reference, but more complexly, continuity of reference and of agentivity value: SS always implies both same referent and same agentivity or control value (whether high or low), while DS marks a discontinuity in either reference or agentivity value (high to low or low to high)' (\pgcitealt{stirling_1993}{150}; see also \citealt{foleyvalin84}).
    \item[(3)] The time of the event: temporal shifts between situations or `from one situation or event complex to another' (\pgcitealt{stirling_1993}{220}) can trigger \textsc{ds} marking (similarly also \pgcitealt{roberts1988}{98–100}). Conversely, temporal continuity may license \textsc{ss} marking even without referential identity (see \pgcitealt{mithun}{126}; \pgcitealt{vangijn2016}{35}).
    \item[(4)] The location of the event: a shift in location can trigger \textsc{ds} marking. This seems to occur most frequently with verbs of motion (cf. also \pgcitealt{roberts1988}{60}).
    \item[(5)] Mood of the clause, whereby \textsc{ds} marking may be used to indicate a shift between realis and irrealis mood. Related to this are shifts in what is referred to as `actuality' (cf. \pgcitealt{kaplandemon}{19}), which a change between current and reported speaker can indicate. 
    \item[(6)] `Continuance or shift out of a cohesive sequence of events'. This includes \textsc{ds} marking used for signalling an unexpected, `surprise' change in the course of events (see also \pgcitealt{roberts1988}{98–100}), as well as a change in \textit{grounding} (\citealt{grimes75}; \citealt{hopper1979}) or \textit{transitivity} (in the sense of \citealt{hopperthompson}).
\end{itemize}

It is interesting to see that all of these factors, to different extents, also explain several of the occurrences of both `canonical' and `non-canonical' participle constructions, particularly dative absolutes whose subject is co-referential with the one of the matrix clause. \\
\indent Point (2), namely a change of agentivity as a possible trigger for \textsc{ds} marking, has also been pinpointed as one of the most common explanations for co-referential dative absolutes in (Old) Church Slavonic. \pgcitet{collins2011a}{110-112}, for example, argues that a change in semantic role or shift in the degree of agentivity can warrant the use of an absolute construction, as (\ref{ocsagentchange1}) and (\ref{ocsagentchange2}) exemplify. Similarly, \pgcitet{kure2006a}{105} observes that some of the several occurrences of co-referential dative absolutes in medieval Serbian/Serbian Church Slavonic are motivated by the different semantic roles of the subjects in the absolute and in the matrix clause. In languages with switch reference, it is not uncommon to find comparable instances. In Eastern Pomo, for example, changes in agentivity regularly license the use of \textsc{ds} markers, despite subject co-reference, as in (\ref{pomoagentchange}).\footnote{The opposite also occurs, i.e. continuity in agentivity licenses the use of \textsc{ss} even if the subjects are not coreferential).}

\begin{example}
\gll i byst\foreignlanguage{russian}{ъ} \textbf{idǫštem\foreignlanguage{russian}{ъ}} im\foreignlanguage{russian}{ь}. ištistišę sę
and happen.\textsc{aor.3.sg} go.\textsc{ptcp.ipfv.dat.pl} \textsc{3.pl.dat} cleanse.\textsc{aor.3.pl} \textsc{refl}
\glt ‘And it came to pass that, as they went, they were cleansed’ (Luke 17.14)
\glend
\label{ocsagentchange1}
\end{example}

\begin{example}
\gll i \textbf{prišedšju} vyšegorodu razbolě sę velmi.
{and} {arrive.{\sc ptcp.pfv.m.dat.sg}} {Vyšegorod.{\sc dat}} {get sick.{\sc aor.3.sg}} {\sc refl} {very}
\glt ‘And when he arrived in Vyšegorod, he got very sick’ (\textit{Primary Chronicle}, Codex Laurentianus f. 54d) %266179
\glend
\label{ocsagentchange2}
\end{example}

\begin{example}
Eastern Pomo (Pomoan, North America; from \citealt{mclendon}, cited in \pgcitealt{roberts_2017}{549})
\begin{itemize}
\item[a.]
\gll Há: \textbf{káluhu-y} si:má:mérqaki:hi 
\textsc{1.sg.a} go.home-\textsc{ss} {went to bed}
\glt `I went home and then went to bed'
\glend
\item[b.]
\gll Há: \textbf{káluhu-qan} mí:\'{p} mérqaki:hi
\textsc{1.sg.a} go.home-\textsc{ds} \textsc{3.sg.a} {went to bed}
\glt ‘I went home and he went to bed’
\glend
\item[c.]
\gll Há: xá: \textbf{qákki-qan} wi \'{q}a:lál ṭá:la
\textsc{1.sg.a} water bathe-\textsc{ds} \textsc{1.sg.u} sick become
\glt ‘I took a bath and got sick’
\glend
\label{pomoagentchange}
\end{itemize}
\end{example}

The Eastern Pomo examples show some of the contexts in which \textsc{ds} and \textsc{ss} markers are used. (\ref{pomoagentchange}a) is a typical, `canonical' example of \textsc{ss} marking, indicating referential identity between the subjects of the two clauses. (\ref{pomoagentchange}b), conversely, is a typical example of \textsc{ds} marking, indicating referential non-identity between the subjects of the two clauses. (\ref{pomoagentchange}c) is an example of `non-canonical' switch-reference marking: \textsc{ds} is used even if the subjects of the two clauses have the same referent, which can be explained with a shift in agentivity or semantic role of the subject referent, from agent/actor to experiencer/undertaker (\pgcitealt{roberts_2017}{549}). \\
\indent A similar explanation can be given to both the Old Church Slavonic example (\ref{ocsagentchange1}) and the Old East Slavic example (\ref{ocsagentchange2}), where the subjects of the absolute construction and of the matrix clause are co-referential, but a shift in the agentivity of the subject can be observed. Note how, incidentally, the matrix clause in the Old East Slavic example in (\ref{ocsagentchange2}) and that in the Easter Pomo example (\ref{pomoagentchange}c) have verbs with the same meaning in the two languages (i.e. `get sick', a typically non-agentive predicate).\\
\indent Point (3) and (4), namely temporal (dis-)continuity between situations and shifts in the location of an event as factors in the use of a switch-reference marker, also find some parallels in the use of absolute constructions as opposed to conjunct participles in Early Slavic. Among the world's languages, this is the case, for instance, in Amele, which may use \textsc{ds} marking even with subject co-reference if a new temporal setting or location is introduced, as the examples in (\ref{amelechangeintimeandplace}) exemplify. 

\begin{example}
Amele (Nuclear Trans New Guinea; from \pgcitealt{stirling_1993}{114})\footnote{These examples are originally from \pgcitet{roberts1988}{61}, but I am here using the more explicit glossing by \citet{stirling_1993}, which, for example, also adds glosses for zero-marking on third persons singular.}
\begin{itemize}
    \item[a.] 
    \gll Eu 1977 jagel November na \textbf{odo-co-b} cul-ig-$\emptyset$-en
    that 1977 month November in do-\textsc{ds-3.sg} leave-\textsc{1.pl-3.sg-remp}
    \glt `That was in November 1977 that he did that and then he left it for us'
    \glend
    \item[b.]
    \gll Age ceta \textbf{gul-do-co-bil} l-i bahim na tac-ein.
    \textsc{3.pl} yam carry-\textsc{3.sg-ds-3.pl} go-\textsc{pred.ss} floor on fill-\textsc{3.pl.remp}
    \glt `They carried the yams on their shoulders and went and filled up the yam store'
    \glend
\end{itemize}
\label{amelechangeintimeandplace}
\end{example}

According to \citet{roberts1987,roberts1988,roberts1988b}, the \textsc{ds} marker on \textit{odo-co-b} `did' in (\ref{amelechangeintimeandplace}a) can be explained with the fact that there is a temporal shift between the first and the second clause, as also signalled by the time adverbial `in November 1977'. Similarly, the \textsc{ds} marking in \textit{gul-do-co-bil} `carried' in (\ref{amelechangeintimeandplace}b), despite its subject being coreferential with the one of the matrix clause, can be motivated by the change in place from the \textsc{ds}-marked clause to the following. In the words of \pgcitet{roberts1988}{60}:

\begin{quote}
    The explanation given by native speakers for such instances is that ``something has changed" or this is ``a new situation", and often it is obvious that the change being indicated is deictic rather than syntactic and that these deictic changes are in the area of world, time, or place reference points. For example, a change of time marked by the SR system is often backed up by a temporal expression; a change of place marked by the SR system occurs most frequently with verbs of motion, and a change of location can also be indicated by a locative expression; a change of world marked by the SR system is normally a switch from an intended or proposed action to the real action itself or vice versa
\end{quote}

Many co-referential dative absolutes in the corpus, particularly in Old East Slavic, fit \posscitet{roberts1988} above characterization, whereby a change in place (often accompanied by a change in time setting) can motivate the use of \textsc{ds} marking. (\ref{ocschangeplace1})-(\ref{ocschangeplace3}) are all co-referential dative absolutes from Old East Slavic texts where movement of some kind from one setting to another is involved.

\begin{example}
\gll i \textbf{všedšim} v gorod\foreignlanguage{russian}{ъ} utěšista volodimercě
{and} {enter.{\sc ptcp.pfv.dat.pl}} {in} {city.{\sc acc.sg}} {console.{\sc aor.3.du}} {people of Vladimir.{\sc acc.pl}}
\glt ‘And when they entered the city, they consoled the people of Vladimir’ (\textit{Suzdal Chronicle}, Codex Laurentianus f. 126r) %275330
\glend
\label{ocschangeplace1}
\end{example}

\begin{example}
\gll i v\foreignlanguage{russian}{ъ} jedin\foreignlanguage{russian}{ъ} ot dnii \textbf{š\foreignlanguage{russian}{ь}d\foreignlanguage{russian}{ъ}šju} k\foreignlanguage{russian}{ъ} tomu blagomu i bonos\foreignlanguage{russian}{ь}nomu ocju našemu ḟeodosiju. i jako v\foreignlanguage{russian}{ъ}nide v\foreignlanguage{russian}{ъ} chram\foreignlanguage{russian}{ъ} ideže bě knęz\foreignlanguage{russian}{ь} sědę. i se vidě mnogyja igrajušta prěd\foreignlanguage{russian}{ъ} nim\foreignlanguage{russian}{ь}
{and} {in} {one.{\sc m.acc}} {from} {day.{\sc m.gen.pl}} {go.{\sc ptcp.pfv.m.dat.sg}} {to} {that.{\sc m.dat.sg}} {good.{\sc m.dat.sg}} {and} {God-bearing.{\sc m.dat.sg}} {father.{\sc m.dat.sg}} {our.{\sc m.dat.sg}} {Feodosij.{\sc dat}} {and} {as} {enter.{\sc aor.3.sg}} {in} {palace.{\sc acc.sg}} {where} {be.{\sc impf.3.sg}} {prince.{\sc nom.sg}} {sit.{\sc ptcp.ipfv.m.nom.sg}} {and} {behold} {see.{\sc aor.3.sg}} {much.{\sc m.acc.pl}} {play.{\sc ptcp.ipfv.acc.pl}} {before} {\sc 3.sg.m.inst}
\glt ‘Once our good and God-bearing father Theodosius came to the prince and as he entered the palace where the prince was, he saw many musicians playing in front of him’ (\textit{Life of Feodosij Pečerskij}, Uspenskij Sbornik f. 59g) %279072
\glend
\label{ocschangeplace2}
\end{example}

\begin{example}
\gll polovci že gjurgevi ni po strělě pustiv\foreignlanguage{russian}{ъ}še togda poběgoša. a potom olgoviči. a potom gjurgi z dětmi. i \textbf{běžaštim} im\foreignlanguage{russian}{ъ} čres\foreignlanguage{russian}{ъ} rut\foreignlanguage{russian}{ъ}. mnogo družiny potope v rutu. bě bo gręzok\foreignlanguage{russian}{ъ}. i \textbf{běžaštim\foreignlanguage{russian}{ъ}} im\foreignlanguage{russian}{ъ} ověch\foreignlanguage{russian}{ъ} izbiša. a drugyja iz\foreignlanguage{russian}{ъ}imaša. i poludne priběgoša ko dněpru.
{Polovtsy.{\sc nom.pl}} {\sc ptc} {Yuri's.{\sc nom.pl}} {\sc neg} {with} {arrow.{\sc dat.sg}} {release.{\sc ptcp.pfv.nom.pl}} {then} {run.{\sc aor.3.pl}} {but} {after} {Olgoviči.{\sc nom.pl}} {but} {after} {Yuri.{\sc nom}} {with} {child.{\sc inst.pl}} {and} {run.{\sc ptcp.ipfv.dat.pl}} {\sc 3.pl.m.dat} {through} {Rut'.{\sc acc}} {much.{\sc n.nom.sg}} {retinue.{\sc gen.sg}} {drown.{\sc aor.3.sg}} {in} {Rut'.{\sc loc}} {be.{\sc impf.3.sg}} {because} {swampy.{\sc nom.sg}} {and} {run.{\sc ptcp.ipfv.dat.pl}} {\sc 3.pl.m.dat} {this.{\sc m.gen.pl}} {kill.{\sc aor.3.pl}} {but} {other.{\sc m.acc.pl}} {capture.{\sc aor.3.pl}} {and} {noon.{\sc loc}} {come running.{\sc aor.3.pl}} {to} {Dnieper.{\sc dat}} 
\glt ‘Yuri's Polovtsy, without shooting arrows at that time, ran, and then the Olgoviči, and then Yuri with his children. And when they fled through the [river] Rut', several of the retinues drowned in Rut' because it was swampy. And as they were fleeing, they killed some and captured others. And by noon, they had run to the Dnieper’ (\textit{Suzdal Chronicle}, Codex Laurentianus f. 111v) %279072
\glend
\label{ocschangeplace3}
\end{example}

(\ref{ocschangeplace1})-(\ref{ocschangeplace3}) are only some of the several examples of coreferential dative absolutes introducing a shift in setting. The absolute in (\ref{ocschangeplace1}) introduces the arrival of Yaropolk Rostislavich in Vladimir, where he becomes Grand Prince, after travelling from Moscow and, before that, Chernihiv. In (\ref{ocschangeplace2}), Feodosij Pečerskij (Theodosius of Kyiv) arrives to visit his brother, the Grand Prince Iziaslav Yaroslavich of Kyiv. The example is from a series of passages in which the narrative goes back and forth from the prince's palace to Feodosij's monastery. In (\ref{ocschangeplace3}), where the broader context is also provided, the rapid shifts in location are clear from the repeated verb (\textit{po})\textit{bežati} `run, flee', following the movement of Yuri and Yuri's Polovtsy and of the Olgoviči through the river Rut' till the Dnieper.\\
\indent Point (5) from \posscitet{stirling_1993} pivots, namely shifts in the mood or actuality from the marked clause to the next, also finds a potential parallel in Early Slavic. This is specifically in regard to Stirling's observation that a shift in actuality can be seen in correspondence to changes between current and reported speakers, a change which can be marked by \textsc{ds}. In Early Slavic it is relatively common to see dative absolutes introducing reported speech, even if the subject of its syntactic matrix is coreferential with the subject of the absolute. (\ref{ocsdirectspeech1})-(\ref{ocsdirectspeech4}) are some of such examples.

\begin{example}
\gll onomu že \textbf{rekšu} něst\foreignlanguage{russian}{ъ} doš\foreignlanguage{russian}{ъ}la zapad\foreignlanguage{russian}{ъ}nyich\foreignlanguage{russian}{ъ} stran\foreignlanguage{russian}{ъ} arijan\foreignlanguage{russian}{ъ}skaja chula
{that.{\sc m.dat.sg}} {\sc ptc} {say.{\sc ptcp.pfv.m.dat.sg}} {not be.{\sc prs.3.sg}} {reach.{\sc ptcp.result.f.nom.sg}} {Western.{\sc gen.pl}} {land.{\sc gen.pl}} {Arian.{\sc nom.sg}} {controversy.{\sc nom.sg}} 
\glt ‘He said that the Arian controversy had not reached the Western lands’ (\textit{Vita of Isaakios of Dalmatos}, Codex Suprasliensis f.  99r) %247879
\glend
\label{ocsdirectspeech1}
\end{example}

\begin{example}
\gll prišedšju že igumenu ko mně. i \textbf{rekšju} mi poidevě v pečeru k feodos\foreignlanguage{russian}{ь}jevi [...] onomu \textbf{glštju} ko mně oudariša v bilo. mně že \textbf{rekuštju} prokopach\foreignlanguage{russian}{ъ} ouže
{arrive.{\sc ptcp.pfv.m.dat.sg}} {\sc ptc} {abbot.{\sc m.dat.sg}} {to} {\sc 1.sg.m.dat} {and} {say.{\sc ptcp.pfv.m.dat.sg}} {\sc 1.sg.m.dat} {go.{\sc imp.1.du}} {in} {cave.{\sc acc}} {to} {Feodosij.{\sc sg.m.dat}} [...] {that.{\sc m.dat.sg}} {say.{\sc ptcp.ipfv.m.dat.sg}} {to} {\sc 1.sg.m.dat} {hit.{\sc aor.3.pl}} {in} {semantron.{\sc acc}} {\sc 1.sg.m.dat} {\sc ptc} {say.{\sc ptcp.ipfv.m.dat.sg}} {dig.{\sc aor.1.sg}} {already}
\glt `So, the abbot came to me and said: ``Let's go to the cave to Feodosij" [...] And when he told me: ``They hit the semantron," I said: ``I finished digging"'
(\textit{Primary Chronicle}, Codex Laurentianus f. 70b) %267388
\glend
\label{ocsdirectspeech2}
\end{example}

\begin{example}
\gll oněm\foreignlanguage{russian}{ъ} \textbf{mnjaštem\foreignlanguage{russian}{ъ}} jako bratii polunošt\foreignlanguage{russian}{ь}noje pěnije s\foreignlanguage{russian}{ъ}v\foreignlanguage{russian}{ь}r\foreignlanguage{russian}{ь}šajuštem\foreignlanguage{russian}{ъ}. i tako paky otidoša. čajušte don\foreignlanguage{russian}{ъ}deže sii s\foreignlanguage{russian}{ъ}kon\foreignlanguage{russian}{ь}čajut\foreignlanguage{russian}{ь} pěnije. i t\foreignlanguage{russian}{ъ}gda v\foreignlanguage{russian}{ъ}š\foreignlanguage{russian}{ь}d\foreignlanguage{russian}{ъ}še v\foreignlanguage{russian}{ъ} crk\foreignlanguage{russian}{ъ}v\foreignlanguage{russian}{ь} pojemljut\foreignlanguage{russian}{ь} v\foreignlanguage{russian}{ь}sę suštaja v\foreignlanguage{russian}{ъ} i
{that.{\sc dat.pl}} {think.{\sc ptcp.ipfv.dat.pl}} {that} {brethren.{\sc dat}} {midnight.{\sc acc.sg}} {prayer.{\sc acc.sg}} {finish.{\sc ptcp.ipfv.dat.pl}} {and} {so} {again} {leave.{\sc aor.3.pl}} {wait.{\sc ptcp.ipfv.nom.pl}} {until} {this.{\sc m.nom.pl}} {finish.{\sc prs.3.pl}} {singing.{\sc acc}} {and} {then} {enter.{\sc ptcp.pfv.nom.pl}} {in} {church.{\sc acc}} {take.{\sc prs.3.pl}} {all.{\sc n.acc.pl}} {be.{\sc ptcp.ipfv.n.acc.pl}} {in} {\sc 3.sg.f.acc}
\glt `They thought that the brethren was [still] finishing singing the midnight prayers, so they [the robbers] left again, waiting until they finished singing, when they could then enter the church and take everything that was in it' (\textit{Life of Feodosij Pečerskij}, Uspenskij Sbornik f. 46g) %278646
\glend
\label{ocsdirectspeech3}
\end{example}

\begin{example}
\gll otc\foreignlanguage{russian}{ъ} že naš\foreignlanguage{russian}{ъ} sergii poklon\foreignlanguage{russian}{ь} sę i reče jako gvi godě tako i budi blsven\foreignlanguage{russian}{ъ} g\foreignlanguage{russian}{ь} vo věki. i vsěm\foreignlanguage{russian}{ъ} \textbf{rekšim\foreignlanguage{russian}{ъ}} amin\foreignlanguage{russian}{ь}
{father.{\sc nom.sg}} {\sc ptc} {our.{\sc m.nom.sg}} {Sergij.{\sc nom}} {bow.{\sc ptcp.pfv.m.nom.sg}} {\sc refl} {and} {say.{\sc aor.3.sg}} {that} {lord.{\sc dat}} {pleasant} {so} {then} {be.{\sc imp.3.sg}} {blessed.{\sc m.nom.sg}} {lord.{\sc m.nom.sg}} {for} {ever} {and} {all.{\sc dat.pl}} {say.{\sc ptcp.pfv.dat.pl}} {Amen}
\glt `Our father Sergij bowed and answered: “As the Lord pleases, so be it; blessed be the Lord forever!”, and they all said “Amen!"' (\textit{Life of Sergij of Radonezh} f. 102v)
\glend
\label{ocsdirectspeech4}
\end{example}

In (\ref{ocsdirectspeech1}) the dative absolute \textit{rekšu} `(he) said' has no apparent matrix clause and, in light of similar non-canonical usages of switch-reference markers in the world's languages, its use could then be explained by the shift to reported speech that the absolute introduces. \\
\indent The absolutes in (\ref{ocsdirectspeech2}) introduce the reported speech of two different protagonists (one of them in the first person) and are part of a longer dialogue chain.\\
\indent In (\ref{ocsdirectspeech3}), the absolute introduces the thoughts of the subject. In this case, the reported thought is introduced as an indirect speech, as \textit{jako} signals, and this time the absolute clause is overtly coordinated with the matrix clause, as is also stressed by the discourse-coordinating connective \textit{tako} `so, thus'. \\
\indent Finally, in (\ref{ocsdirectspeech4}), the post-matrix, overtly coordinated absolute \textit{rekšim\foreignlanguage{russian}{ъ}} `(they) said' does not seem to have any other apparent function but introducing a reported speech. The speaker introduced by the absolute \textit{is} different from the previously reported speaker, which was, in fact, introduced by the finite clause \textit{reče} `(he) said'. 

\indent Turning now to the final point from \posscitet{stirling_1993} pivots, the use of switch reference to signal a change in \textit{grounding}, namely from \textit{foregrounding} to \textit{backgrounding} or vice-versa, is, as we have repeatedly seen, the main constant function of dative absolutes and conjunct participles emerged from the previous chapters. In Chapter 2, we have also seen that even in post-matrix position the temporal relation of dative absolutes to their matrix clause is most typically consistent with the temporal relation they have with the matrix when they are fronted, i.e. as typical \textsc{frames}. Given the attested, albeit much less frequent, use of post-matrix absolute constructions as \textsc{elaborations} and the fact that these are all non-coreferential with the subject of their matrix clause, we can consider referential (non-)identity as the basic criterion, as it were, whereby an absolute is chosen over a conjunct participle and vice-versa. That is, given a discourse function, if both conjunct participles and absolute constructions are equally viable means for fulfilling that function, then the former is used when the subject referents are the same, the latter if they differ. This is in many ways similar to what has been claimed about switch reference systems cross-linguistically, namely that `all other things being equal, the same referent marker [...] appears in situations of referential identity between the subject participants of two clauses, and the different referent marker [...] when there is no such referential identity. However, other factors, to do with discourse cohesion may overrule this basic set-up'\footnote{I have masked the reference to the specific markers in Tsafiki, namely \textsc{ss} -\textit{to} and \textsc{ds} -\textit{sa}, from the source since it is clear from \posscitet{vangijn2016} overview that this statement is generally applicable to switch-reference systems cross-linguistically, where participant (non-)identity is the basic factor even when several other discourse variables can supersede those factors.} (\pgcitealt{vangijn2016}{3}). Regarding referential (non-)identity as the baseline of all switch reference systems, \pgcitet{stirling_1993}{151}, after listing the six `pivots' for switch-reference marking reported above, claims that all switch-reference systems mark at least point (1), namely referential (non-)identity. However, the common cross-linguistic coverage of all six pivots indicates that switch reference may be regarded as a type of clause linkage and not simply a referential-tracking device.\\
\indent At an event-ontological level, it has been observed that there are clear ties between disjoint reference and backgrounding and between coreference and foregrounding. Building on \posscitet{hopperthompson} notion of `transitivity' as a value assignable to clauses on the basis of properties such as the number of participants (i.e. the basic notion of transitivity), the Aktionsart (e.g. `kinesis', i.e. states versus non-states; `punctuality', i.e. punctual versus durative, or `nonpunctual') or the individuation of the event participants, \pgcitet{stirling_1993}{147--150} highlights the apparent correlation between backgrounding and low transitivity, and between foregrounding and high transitivity, adding that

\begin{quote}
    chains of foregrounded clauses (within one episode) typically continue to talk about the same participants rather than introducing new ones, thereby maintaining same subject. Furthermore, chains of foregrounded clauses tend to present sequential events, while simultaneous events (along with states and on-going or repeated events) appear in background clauses.
\end{quote}

\begin{figure}[!h]
\begin{subfigure}{0.50\textwidth}
\includegraphics[width=0.9\linewidth]{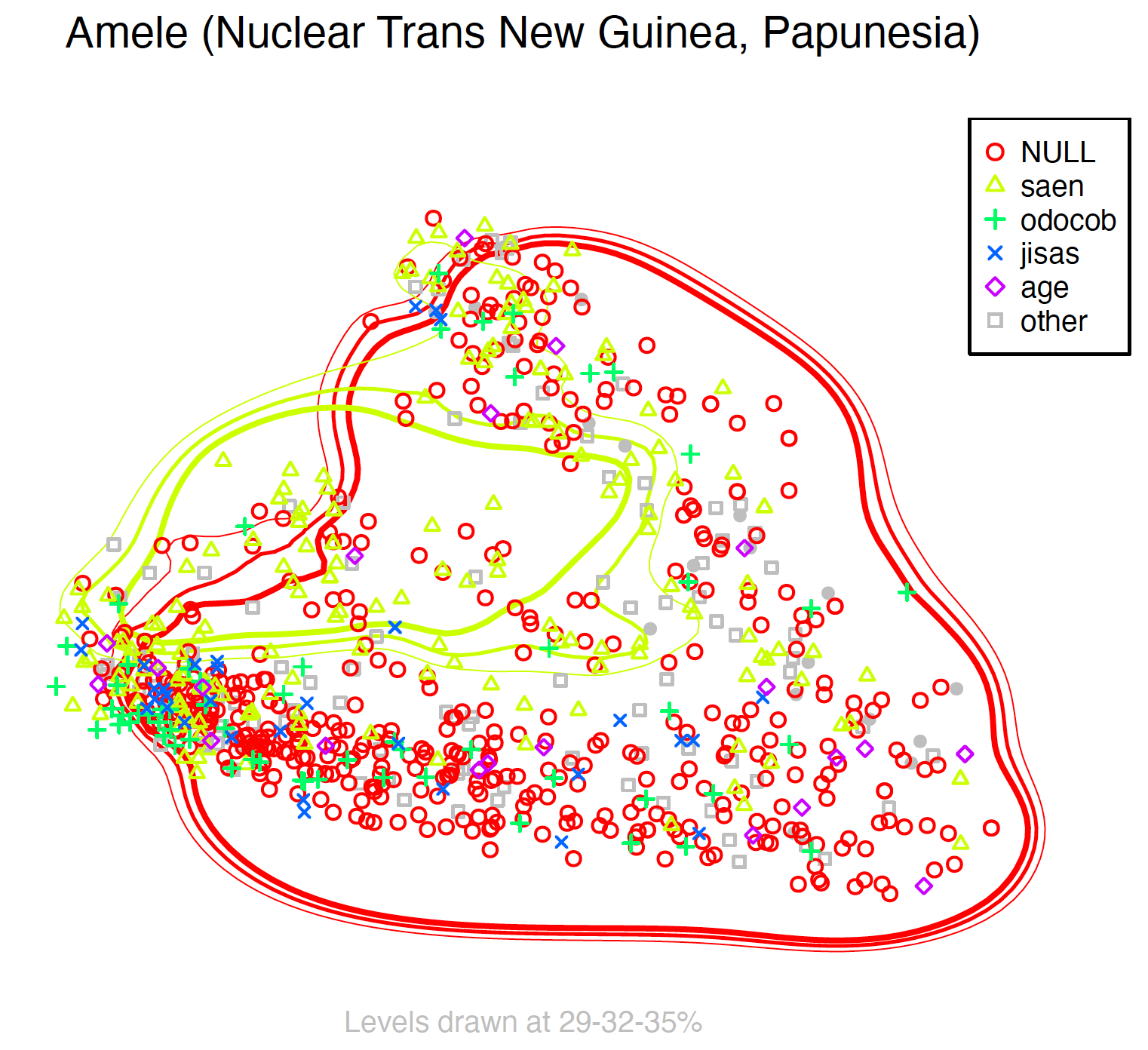} 
\caption[Kriging map for Amele (Nuclear Trans New Guinea, Papunesia)]{}
\label{amele-krig}
\end{subfigure}
\begin{subfigure}{0.50\textwidth}
\includegraphics[width=0.9\linewidth]{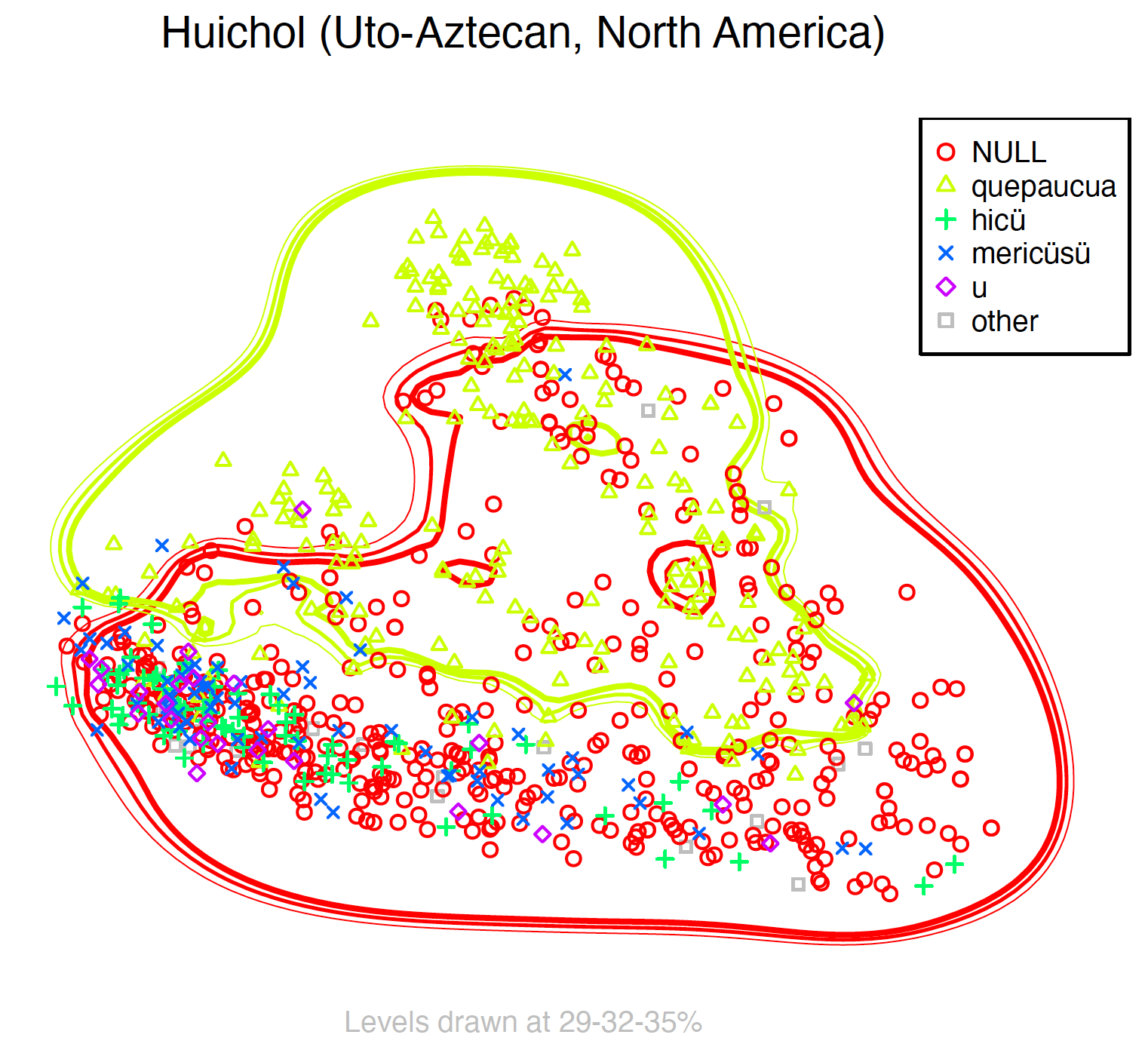}
\caption[Kriging map for Huichol (Uto-Aztecan, North America)]{}
\label{huichol-krig}
\end{subfigure}
\caption[]{Kriging maps for Amele (Nuclear Trans New Guinea, Papunesia) and Huichol (Uto-Aztecan, North America)}
\end{figure}

This is an observation that can be easily applied to \textsc{independent rhemes} (foregrounded clauses), typically encoded by conjunct participles and typically involving the same participants as their matrix (and any other conjunct participle dependent on it) and presenting sequential events, as opposed to \textsc{frames} (backgrounded clauses), more typically encoded by absolute constructions and normally involving overlap/simultaneity/inclusion. The similarity between absolute constructions and switch-reference markers had, in fact, been observed before by \citet{bickel99}, who argued that many formal synchronic switch-reference systems have originated from absolute constructions `in pragmatic competition with conjunct participles (\textit{participia coniuncta}) that show case agreement with a coreferential argument of the matrix, occupy roughly the same adsentential position as absolutes, and fulfill a similar discourse function' (\pgcitealt{bickel99}{46}), which, Bickel argues, is the grammaticalization paths followed by the switch-reference systems of languages in the Muskogean, Yuman and Uto-Aztecan families.\\
\indent There are thus clearly stronger connections between \textsc{ds} markers in switch reference systems and absolute constructions, and between \textsc{ss} markers and conjunct participles, than the mere marking of referential (non-)identity. The languages mentioned in this section that are also in our dataset, namely Amele and Huichol, had already been assigned to the same pattern (B) as Old Church Slavonic in the classification in the previous Chapter. Their Kriging maps, particularly the one for Huichol are, in fact, quite similar to the Old Church Slavonic one (Figure \ref{proielchu0rep} above), as Figures \ref{amele-krig} and \ref{huichol-krig} show.

\begin{figure}[!h]
\begin{subfigure}{0.50\textwidth}
\includegraphics[width=0.9\linewidth]{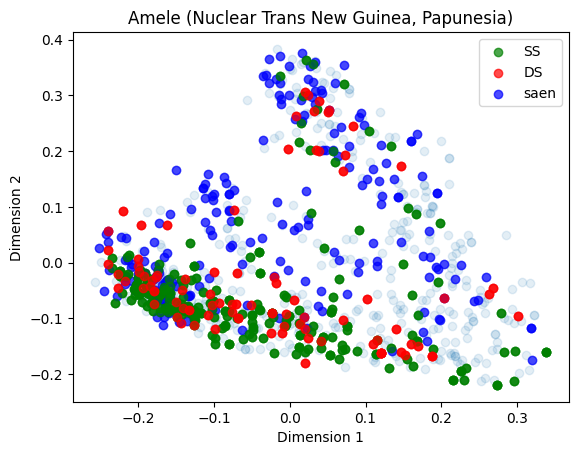} 
\caption[MDS map of \textsc{when} highlighting overt subordinators and switch reference markers in Amele (Nuclear Trans New Guinea, Papunesia)]{}
\label{amelecontinuum}
\end{subfigure}
\begin{subfigure}{0.50\textwidth}
\includegraphics[width=0.9\linewidth]{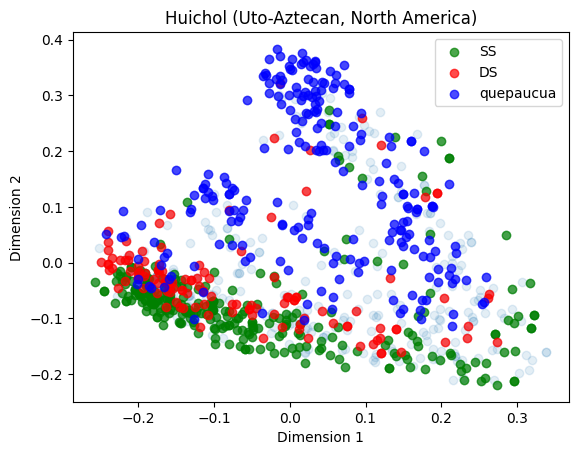}
\caption[MDS map of \textsc{when} highlighting overt subordinators and switch reference markers in Huichol (Uto-Aztecan, North America)]{}
\label{huicholcontinuum}
\end{subfigure}
\begin{subfigure}{0.9\textwidth}
\centering
\includegraphics[width=0.5\linewidth]{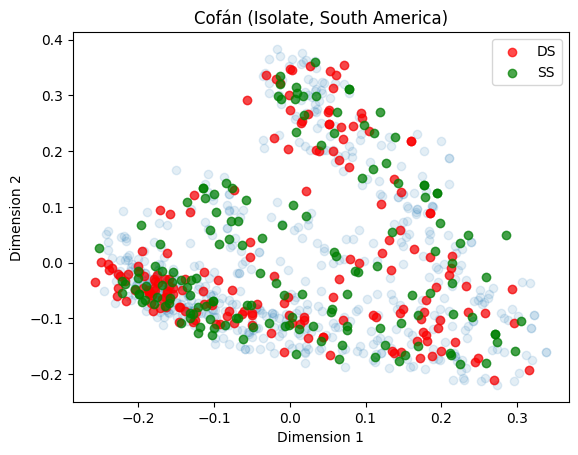}
\caption[MDS map of \textsc{when} highlighting overt subordinators and switch reference markers in Cofán (Isolate, South America)]{}
\label{cofancontinuum}
\end{subfigure}
\caption[]{MDS map of \textsc{when} highlighting overt subordinators and switch reference markers in (a) Amele (Nuclear Trans New Guinea, Papunesia), (b) Huichol (Uto-Aztecan, North America) and (c) Cofán (Isolate, South America).}
\end{figure}

Since both Amele and Huichol, as well as Cofán (A'ingae) from the previous section, also use converbal/medial forms marked for switch reference as temporal clauses, in an ideal scenario, a complete semantic map for \textsc{when} would also include different switch-reference markers. This would allow us to capture the competition not only between potential subordinating conjunctions (if present at all) and null constructions, but also between different null constructions. Doing this at scale would require detailed annotation, at least, at the morphosyntactic level, as the one we currently have for Old Church Slavonic and Ancient Greek in the PROIEL treebanks, which did, in fact, allow us to add a further layer of distinctions in their semantic map. Gathering such information for all the languages in the dataset would be hardly feasible, but a preliminary, automatic test on Amele, Huichol, and Cofán indicates that such direction could lead to very informative maps and may thus be worth pursuing for more languages. All three languages have relatively isolable switch-reference morphemes. Amele has -\textit{co}-/-\textit{ce}- + personal endings for \textsc{ds} and -\textit{me}- + personal endings for \textsc{ss}.\footnote{Following \pgcitet{roberts1988}{48} summary of endings, this gives the following possible markers for \textsc{ds}: \textit{-comin}, \textit{-com}, \textit{-cob}, \textit{-cohul}, \textit{-cobil}, \textit{-comun}, \textit{-cemin}, \textit{-cem}, \textit{-ceb}, \textit{-cehul}, \textit{-cebil}, \textit{-cemun}; and these for \textsc{ss}: \textit{-meig}, \textit{-meg}, \textit{-meu}, \textit{-meb}, \textit{-mesi}, \textit{-mei}. Note that these are only the `sequential' switch-reference endings in Amele, namely those that are used to express the sequentiality of events. Those expressing simultaneity are not isolable as they involve stem reduplication.} Huichol has \textit{-ku} and -\textit{ka} (spelled as \textit{-cu} and \textit{ca} in our dataset) for \textsc{ds} and and \textsc{ss} respectively (\citealt{comriehuichol}). Cofán has -\textit{si} for \textsc{ds} and -\textit{pa}/-\textit{mba} for \textsc{ss} (\citealt{anderboisaltshulersilva23}). Similarly to what did for Old Church Slavonic and Ancient Greek participles in Figures \ref{chucontinuum} and \ref{grccontinuum}, the placeholders `DS' and `SS' were inserted before any word in the corpus ending with the respective markers in each of these languages, thus allowing the model to capture the placeholders as dummy subordinators. After realigning the processed texts to the English version and extracting \textit{when} and its parallels in the three languages, we can superimpose the occurrences of \textit{when} aligned to the dummy switch-reference subordinators on the raw MDS map. Figures \ref{amelecontinuum}-\ref{cofancontinuum} show the results for the three languages. Based on the previous Kriging analysis, Amele and Huichol also have lexified \textit{when}-counterparts, whereas Cofán does not. These are, therefore, also superimposed on the map for the former two to allow for easier comparison with the maps for Old Church Slavonic and Greek in Figures \ref{chucontinuum} and \ref{grccontinuum}.

The result for Cofán indicates that switch-reference markers in this language have a much wider scope and are likely to be used with different functions than Old Church Slavonic absolutes and conjunct participles. Judging from the Kriging map shown in the previous section, Cofán does not appear to have a lexified counterpart to \textit{when} but instead uses switch-reference markers for the whole semantic space in the \textsc{when}-map.\\
\indent On the other hand, the maps for Amele and Huichol, but particularly the latter, bear great similarities with the ones we have seen for Old Church Slavonic and Ancient Greek in Figures \ref{chucontinuum} and \ref{grccontinuum}. At least in Huichol, we observe that same continuum from \textit{when}-counterparts (\textit{quepaucua}) at the top of the map to \textsc{ss} markers at the very bottom (corresponding to what the highlighted conjunct participles in the Old Church Slavonic and Ancient Greek maps), with \textsc{ds} markers also clustering in the bottom area, but in a relatively defined band above \textsc{ss} markers, similarly to the position occupied by absolute constructions in Figures \ref{chucontinuum} and \ref{grccontinuum}. In the Amele map, this continuum is less neat but does overall reflect a very similar distribution between switch reference markers and \textit{when}-counterpart (\textit{saen}). Besides, the Amele map only includes switch-reference markers indicating sequentiality, not those conveying simultaneity, for reasons of isolability of the morphemes, as explained earlier. Further annotation may thus improve the map considerably.
% We have seen in the previous chapters, particularly in Chapters 3 and 4, that one of the clearest differences between perfective \textsc{frames} and perfective \textsc{independent rhemes} is that with the former the temporal relation to their matrix clause is never strictly one of abutment (i.e. \textit{Strong Narration}), since they can be seen as introducing a (resultative) state within which the events of the matrix clause unfold, whereas an abutment relation, by definition, is one in which the post-state of the first event is pragmatically incompatible with the pre-state of the second event, as is the case with many perfective conjunct participles. Whether this makes absolute clauses always simultaneous to the matrix event is debatable, especially considering how simultaneity is not a clear-cut concept and can encompass event identity and inclusion, and we have seen that these fine-grained temporal distinctions are ultimately what helps define some of the differences between participle constructions and \textit{jegda}-clauses, as we saw in Chapter 3. Nevertheless, there are once again obvious parallels between what is claimed in the literature on switch reference and what we observe in Early Slavic, and it is true that (pre-matrix) conjunct participles and dative absolutes can be broadly conceived as being respectively broadly correlated with sequentiality and overlap. \\

\subsection{Insubordination}\label{sec:insub}
\indent As we saw in Chapter 2, seemingly independent dative absolutes often become clearly backgrounding devices for some portion of discourse to come when units larger than clauses and sentences are taken into consideration (cf. \citealt{collins2011a}), and this may well be the case for all the examples of seemingly independent absolutes seen above, as in (\ref{ocsdirectspeech1})-(\ref{ocsdirectspeech4}).\footnote{As with coreferential dative absolutes, syntactically independent dative absolutes have also been specifically discussed in previous literature as `non-canonical' occurrences of the construction. On the topic of independent/coordinated absolutes, see, for example, \pgcitet{corin1995a}{262-276}, \pgcitet{ve1961a}{50–51}, \citet{alekseev87}, \pgcitet{gebert}{568}, \pgcitet{remneva}{35-39}, \pgcitet{collins2011a}{113-126}.} However, what is interesting to notice there is that, even at the inter-clausal level, such `non-canonical' instances of dative absolutes could be motivated as devices to indicate a shift in actuality, as \citet{stirling_1993} puts it, and that these instances have cross-linguistic counterparts in the way switch reference systems may use \textsc{ds} markers for goals other than subject co-reference. Yet, both functions, namely the use of seemingly independent absolutes as background for discourse segments larger than sentences and as markers of shifts in actuality, are by no means mutually exclusive. Similar functions have been observed cross-linguistically on so-called `insubordinate' constructions (\citealt{evans2007,evanswatanabeintro}).\\
\indent \textit{Insubordination} is widely attested among the world's languages independently of switch-reference systems and can be defined, from the synchronic perspective, as `the independent use of constructions exhibiting prima facie characteristics of subordinate clauses' (\pgcitealt{evanswatanabeintro}{2}), representing a `redeployment of linkages from intra-clausal to general discourse links' (\pgcitealt{evans2007}{370}). \\
\indent Within switch-reference systems, insubordination has been observed by \pgcitet{wilkins}{154--156}, for example, in Mparntwe Arrernte (Pama–Nyungan, Australia), where \textsc{ss}- or \textsc{ds}-marked clauses may occur on their own in a sentence without any apparent matrix clause (a phenomenon referred to as \textit{trans-sentential switch-reference} by \citeauthor{wilkins}), which, in most occurrences, is explained with the fact that `a main clause in a text or discourse can, later in the text or discourse, be referred to by an anaphor or can be ellipsed, as long as it is contextually recoverable' (\pgcitealt{wilkins}{154}). \\
\indent In the switch reference system of Yuman languages (Cochimí-Yuman), the \textsc{ss} marker -\textit{k} is typically used, at the inter-clausal level, to represent closely related events as belonging to a higher level event (i.e. the typical stacking of clauses in chains to form a sentence), while the \textsc{ds} marker -m is used to link clauses `representing less closely related event or states' (\pgcitealt{mithun2008}{100}; see also \citealt{winter76,slater77,langdonmunro79,hardy82,gordon83}). However, some languages within the family (in the Hualapai branch) have extended the use of the markers to `marking relations of full sentences to each other in larger discourse contexts, providing higher-level chunking of experience into events' (\pgcitealt{mithun2008}{100}), which is very similar to what has been claimed about non-canonical dative absolutes in studies such as \citet{collins2011a}. \\
\indent Insubordination with functions similar to those observed on non-canonical absolutes is well-attested also beyond switch-reference markers. The adverbializer =\textit{go} in Navajo (Athabaskan-Eyan-Tlingit, North America), for example, which marks the subordinate relations of adverbial clauses within sentences often also `specifies relations of sentences to larger stretches of discourse' (\pgcitealt{mithun2008}{72}). \\
\indent In Navajo narrative discourse, we can find a series of clauses, seemingly independent, some marked with \textit{=go}, some unmarked (i.e. formally independent). According to Mithun, the latter advance the storyline, whereas the former `provide background, incidental information, explanation, and emotional evaluation' and, as observed by Mithun, can thus be reconducted within models of narrative structure (e.g. \citealt{hopper1979,chafe94}) to foreground and background, respectively. \\
\indent Similarly, in Central Alaskan Yup'ik (Eskimo-Aleut), participles can be used insubordinately to `provide background information, parenthetical comments off the event line, elaboration, and evaluation' and are `often used when speakers open narratives or episodes by setting the scene' (\pgcitealt{mithun2008}{90}). As pointed out by \pgcitet{heineetalinsub}{51}, the functions of insubordinate constructions observed by Mithun are described in the literature on parentheticals as `meta-discursive' (e.g. \pgcitealt{mosegaardhansen}{236}), `meta-textual' (e.g. \pgcitealt{traugottdasher}{155}), and `meta-communicative' (e.g. \pgcitealt{grenobletheticals}{1953}). This is precisely how \citet{collins2011a} describes one of the main functions of `independent' participle constructions. (\ref{ocsparenth1})-(\ref{ocsparenth3}) are examples of independent participle clauses which, as \pgcitet{collins2011a}{120--122} observes, function as `meta-discursive', `meta-comment', or `meta-textual' devices.

\begin{example}
\gll troje jest\foreignlanguage{russian}{ь} nemošt\foreignlanguage{russian}{ь}no mi razouměti a četvora ne razoumějǫ slěda or\foreignlanguage{russian}{ь}lou lětęštou xristosovo v\foreignlanguage{russian}{ъ}š\foreignlanguage{russian}{ь}stije i potii zmiin\foreignlanguage{russian}{ъ} po kameni dijavol\foreignlanguage{russian}{ъ} ne obrěte bo slěda grěxov\foreignlanguage{russian}{ь}naago na tělesi xsvě i pouti lodija po vodě plovoušti \textbf{cr\foreignlanguage{russian}{ь}k\foreignlanguage{russian}{ъ}vi} aky v\foreignlanguage{russian}{ъ} poučině žit\foreignlanguage{russian}{ь}ja sego naděždejǫ jaže v\foreignlanguage{russian}{ъ} xsa krs\foreignlanguage{russian}{ь}m\foreignlanguage{russian}{ь} \textbf{pravimě} i poutii moužę v\foreignlanguage{russian}{ъ} jǫnosti roždenaago ot\foreignlanguage{russian}{ъ} svętaago dxa i ot\foreignlanguage{russian}{ъ} dvca
three be\textsc{.prs.3.sg} impossible \textsc{1.sg.dat} understand\textsc{.inf} and fourth \textsc{neg} understand\textsc{.prs.1.sg} track\textsc{.gen.sg} eagle.\textsc{dat.sg} fly.\textsc{ptcp.ipfv.m.dat.sg} Christ's.\textsc{nom.sg} ascension.\textsc{nom.sg} and path\textsc{.gen.pl} serpent's.\textsc{gen.sg} over rock.\textsc{loc.sg} devil.\textsc{nom} \textsc{neg} find.\textsc{aor.3.sg} because trace.\textsc{gen.sg} sin's.\textsc{gen.sg} on body.\textsc{loc.sg} Christ's.\textsc{loc.sg} and path.\textsc{gen.sg} boat\textsc{.f.gen.sg} through water.\textsc{loc} float.\textsc{ptcp.ipfv.f.gen.sg} Church.\textsc{dat} like in gulf.\textsc{loc.sg} life.\textsc{gen.sg} this.\textsc{gen.sg} hope.\textsc{f.inst.sg} which.\textsc{f.nom.sg} in Christ.\textsc{gen} cross.\textsc{inst.sg} govern.\textsc{ptcp.pass.ipfv.f.dat.sg} and path.\textsc{gen.pl} man.\textsc{m.gen.sg} in youth.\textsc{loc} born\textsc{.m.gen.sg} from holy\textsc{.gen} spirit.\textsc{gen} and from virgin.\textsc{gen}
\glt `Three things it is impossible for me to understand, and a fourth I do not understand: the track of an eagle flying (Christ’s Ascension); and a serpent’s paths over a rock (the Devil; for he did not find a trace of sin on Christ’s body); and the path of a boat sailing through the water (like the Church being governed in the gulf of this life by hope in Christ, by the Cross); and the paths of a man in youth (the One born of the Holy Spirit and the Virgin)’ (\textit{Izbornik of 1073}, f. 156c1–20)
\glend
\label{ocsparenth1}
\end{example}

\begin{example}
\begin{itemize}
    \item[a.]
    \gll mol \textbf{lěžjǫšte} na loži
    {prayer.{\sc nom.sg}} {lie.{\sc ptcp.ipfv.nom.pl}} {in} {bed.{\sc loc}}
    \glt 'A prayer [when people are] lying in bed' (\textit{Euchologium Sinaiticum} f. 37a)
    \glend
    \item[b.]
    \gll mol egda \textbf{chotęšte} vinograd\foreignlanguage{russian}{ъ} saditi
    {prayer.{\sc nom.sg}} {when} {want.{\sc ptcp.ipfv.nom.pl}} {vineyard.{\sc acc.sg}} {plant.{\sc inf}}
    \glt `A prayer when [people are] wanting to plant a vineyard' (\textit{Euchologium Sinaiticum} f. 13b) %293476 %293442
    \glend
\end{itemize}
\label{ocsparenth2}
\end{example}

\begin{example}
    \gll ot {mě:stę:t\foreignlanguage{russian}{ь} :} ko {ga:vo:š\foreignlanguage{russian}{ь} :} {i :} {ko :} {so:di:l\foreignlanguage{russian}{ь} :} po:py:ta:i:=ta {mi :} {konę :} {a :} \textbf{{mě:stę:ta :}} \textbf{{sę :}} \textbf{va:ma} \textbf{{poklanę :}} {a:že :} {va :} {c\foreignlanguage{russian}{ь}:to :} {na:do:b\foreignlanguage{russian}{ь} :} a {soli=ta :} ko {mon\foreignlanguage{russian}{ь} :} a {gramotuo :} vodai=ta a uo {pavla :} skota poprosi:=ta a m\foreignlanguage{russian}{ь}:stę
    from Mestjata\textsc{.gen.sg} to Gavša\textsc{.dat.sg} and to Sdila\textsc{.dat.sg} find\textsc{.imp=ptc} \textsc{1.dat.sg} horse\textsc{.gen.sg} and Mestjata\textsc{.nom.sg} \textsc{refl} \textsc{2.dat.du} bow.{\sc ptcp.ipfv.m.nom.sg} if \textsc{2.dat.du} something needed and send\textsc{.imp=ptc} to \textsc{1.dat.sg} and letter\textsc{.acc.sg} give\textsc{.imp=ptc} and from Pavel\textsc{.gen.sg} money\textsc{.acc.sg} ask\textsc{.imp=ptc} and Mestja[ta]
    \glt `From Mestjata to Gavša and to Sdila. Get a horse for me. And Mestjata bows to you (both). If you need anything, send to me, and give (the messenger) a letter. And ask for money from Pavel, and Mestja[ta] (bows to you).' (\textit{Birchbark letter} N422, translation by \pgcitealt{schaeken}{146})
    \glend
\label{ocsparenth3}
\end{example}

(\ref{ocsparenth1}), from the \textit{Izbornik of 1073} (also known as \textit{Izbornik of Svjatoslav})\footnote{The \textit{Izbornik of 1073} is not part of the datasets used in this thesis. The passage and translation are from \pgcitet{collins2011a}{120}, but transcription and glossing are mine and follow the photographic edition in \citet{izbornik}.}, presents a passage from the Book of Proverbs (30:18–19) interpolated with exegetical commentaries (rendered in \posscitet{collins2011a} translation with parentheses). These comments, including the third (the dative absolute), are clearly `syntactic parentheticals functioning as metadiscourse rather than as part of the basic text' (\pgcitealt{collins2011a}{121}).\\
\indent (\ref{ocsparenth2}a-b) are standard examples of the rubrics with which \textit{euchologia} introduce prayers to be recited aloud by the clerical readers in the specified occasion. In the \textit{Euchologium Sinaiticum}, we also find several examples of the meta-comment \textit{imę rek\foreignlanguage{russian}{ъ}} `having said the name' in correspondence of names to be said out loud throughout the prayer (ten occurrences, according to \pgcitealt{nagtigal}{xxxix}, as reported in \pgcitealt{collins2011a}{121}). As \pgcitet{collins2011a}{121} points out, `the indirect command conveyed by \textit{imę rekŭ} can be understood as a continuation of this same implicit speech act [the one conveyed by the rubrics]: Here say the beneficiary’s name, and then continue the prayer'.\\
\indent (\ref{ocsparenth3}) contains the nominative absolute \textit{a městęta sę vama poklanę} `and Mestjata bows to you (both)', where \textit{you} refers to addressees of the letter (Gavša and Sdila) and the clause is an instruction to the messenger to bow to the addressees on Mestjata's behalf, which reflects the common practice of reading the letters aloud to the addresses (\pgcitealt{zaliznyak04}{297}; \pgcitealt{schaeken}{11-12}). The metatextual function of the participle construction in this example is backed up by several other occurrences of instructions with similar functions (cf., for example, birchbark letters N354 and N358).\\
\indent All these examples indicate that the `non-canonical', independent use of participle constructions corresponds, in fact, to a rather well-attested and well-understood pattern.

\section{Summary}
This chapter addressed the issue of understanding the functions of null constructions in the semantic map of \textsc{when}, started in the previous chapter, not only at the level of their competition with \textit{when}-counterparts, but also with respect to the competition between different null constructions, which the semantic map of \textsc{when}, without further annotation, could not address. \\
\indent Section \ref{ocsmaps} looked at the distribution of absolute constructions and conjunct participles in the map for Old Church Slavonic by adding targeted annotation, allowing to distinguish the two constructions among otherwise unspecified null alignments to \textit{when}. The analysis showed that, as anticipated in the previous chapter, both absolutes and conjunct participles cluster at the bottom of the map and occupy rather contiguous than fully overlapping areas, with absolutes relatively closer to the main \textit{jegda}-cluster at the top of the map.  Adding \textit{while} to the Kriging maps indicated that absolutes in Old Church Slavonic cover much of the same ground as English \textit{while}, unlike Greek, whose Kriging for area for absolute constructions, both in the \textit{when}-only and in the when-while-map, corresponds largely to uses of English \textit{when}. This difference corresponds to the observation made in Chapter 1 on Old Church Slavonic-Greek mismatches, which showed that there is a small but consistent group of constructions in Greek (\textit{en tōi} + infinitive) that are translated into absolutes in Old Church Slavonic and convey temporal inclusion of the main eventuality in the eventuality introduced by the adverbial clause, similarly to the function of English \textit{while}.\\
\indent Section \ref{sec:pular} looked into an example of language (Pular) with relatively well-defined lexified counterparts to \textit{when} not only for the top and middle cluster as in several other languages (including Old Church Slavonic), but also for the bottom left cluster. Following grammatical descriptions of this language, functional labels were substituted for the particular connectives, with the aim of pinpointing functional differences between the top half and bottom half of the semantic map of \textsc{when}, as well as specifically clarifying differences in the meaning of the middle-left cluster (corresponding to the Kriging area for Greek \textit{hóte} and part of the Kriging area for Old Church Slavonic \textit{jegda}) and the bottom-left cluster. The analysis confirmed, as was suspected in the previous chapter, that the bottom area of the semantic map, within which most participle construction fall in Early Slavic, corresponds to a relatively well-defined function from the perspective of temporal relations, namely that between existential-eventive clauses, which translates into the sequentiality between events being stressed. \\
\indent Finally, Section \ref{sec:insub} looked at various well-studied phenomena from languages that displayed similar patterns to Old Church Slavonic in the previous Chapter or that seemingly \textit{only} use null constructions across the semantic map of \textsc{when}, specifically clause chaining, bridging, switch reference and insubordination. The goal was to clarify the competing motivations governing the choice between different types of null constructions, as well as between null constructions and \textsc{when}-counterparts that appear to be in intense competition with null constructions in the bottom region of the semantic map. \\
\indent First, we substituted functional labels (\textsc{independent rhemes} and \textsc{frames}) in the Ancient Greek map (because of its greater coverage of the New Testament) for specific configurations of participle constructions known from previous chapters to be correlated with those functions. The maps generated with these functional labels indicated more clearly that typical framing participles (sentence-initial absolutes) and \textsc{independent rhemes} (pre-matrix, aorist participles with an aorist matrix) dominate in contiguous, but not fully overlapping areas in the map, with \textsc{frames} being relatively closer to finite competitors at the top of the map than \textsc{independent rhemes}. The same pattern was observed by simply overlaying the observation for the three constructions to the raw MDS map. \\
\indent The overview on clause chaining suggested that similar criteria of discourse organization govern the use of chaining constructions in clause-chaining languages and participle constructions in Old Church Slavonic, including the possibility of \textit{backgrounded} clauses alternating with \textit{foregrounded} clauses, as well as overarching principles of tense-space-participant connectedness and event conflation/integration. \\
\indent We then looked at how the `topic-shift' function of dative absolutes mentioned in previous chapters, whereby elements of a foregrounded eventuality in the previous discourse get repurposed as background for a new portion of foregrounded discourse to come, has, in fact, clear parallels in bridging constructions as a systematic, formalizable phenomenon. In particular, we saw that formal accounts of bridging based on the SDRT rhetorical relations \textit{Background} and Narration, which we already partially implemented in Chapter 5, are clearly applicable to several similar examples of absolute constructions in Old Church Slavonic.\\
\indent The clearest parallels to the use of conjunct participles as opposed to absolute constructions, as well as the common thread between all the phenomena touched upon in this section, were observed in synchronic switch-reference systems across the world's languages. Especially considering the areal and genetic unrelatedness of the languages mentioned in this overview to Old Church Slavonic, the similarities between the former and the latter were found to be quite striking. In particular, the automatic addition of switch reference markers to the semantic map of \textsc{when} for Amele, Huichol and Cofán suggested that languages with a switch reference system \textit{and} a lexicalized counterpart to \textit{when} (i.e. in our case Amele and Huichol, not Cofán) show an extremely similar distribution of \textit{when}-counterparts, \textsc{ds}-markers and \textsc{ss}-markers, approximately placed in a continuum comparable to \textit{jegda}-clauses, absolutes, and conjunct participles in Old Church Slavonic. Overall, they provided some independent evidence that the use of absolute constructions as opposed to conjunct participles is regulated by very similar principles to the `pivots' that play a role in the use of switch-reference markers cross-linguistically, as described in detail, among others, by \citet{stirling_1993}. \\
\indent Finally, we saw how insubordination, namely the `independent use of constructions exhibiting \textit{prima facie} characteristics of subordinate clause' (\pgcitealt{evanswatanabeintro}{2}), and the independent uses of dative absolutes analysed by \citet{collins2011a} also bear great similarities, particularly in light of the fact that insubordination and switch-reference are often related in the same languages.\\
\indent Crucially, all these principles are also able to motivate several of the uses of `non-canonical' participle constructions, particularly dative absolutes, which had otherwise often been written off as `aberrations' or as signs of `decay' by the earlier scholarship on Early Slavic (e.g., among many others, \pgcitealt{ve1961a}{49}; \pgcitealt{ve1996}{190}; \pgcitealt{corin1995a}{268}).

\chapter{Conclusions}

In the Introduction, the goal of this thesis was set up to be the study of `the semantic and pragmatic properties of Early Slavic conjunct participles and absolute constructions to understand what principles motivate their selection over one another and over \textit{jegda}-clauses at the synchronic level'. The investigation, I believe, led to some genuine progress on several fronts and raised some new questions on others.

The stated goal was addressed by adopting two broadly different approaches: a quantitative, corpus-based one, exploiting data from Early Slavic treebanks, and a typological one, leveraging massively parallel corpora consisting of translations of the New Testament in over 1400 linguistic varieties, including Old Church Slavonic. 

Since very early on in the investigation, it became clear that the competition between conjunct participles, absolute constructions, and \textit{jegda}-clauses occurs at the level of discourse organization, where the major common denominator is that all of them may function as event-framing devices. The concept of `framing device' was meant as a discourse segment that is used `to limit the applicability of the main predication to a certain restricted domain', which is \pgposscitet{chafe1976a}{50} definition of \textit{frame setters} and was consistently associated with the \textit{background} content of a (complex or elementary) discourse unit. The main contrast throughout the thesis, then, was between \textit{background} clauses and discourse segments corresponding to the \textit{foreground} content of a predication. The background-foreground contrast informed much of the discussion, encompassing all theoretical frameworks followed in this thesis.

From the investigation, the following findings can be highlighted.
% \begin{itemize}
%     \item Conjunct participles are typical foreground material
%     \item Dative absolutes are typical background material
%     \item Perfective \textit{jegda}-events can include perfective matrix events
%     \item \textit{Independent rhemes} trigger Strong Narration, \textit{frames} trigger Background
%     \item Sequentiality and event conflation are defining principles
%     \item `Non-canonical' participle constructions are canonical after all

% \end{itemize}

\section{Overview of the main findings}

\begin{itemize}

\item[(a.)] \textbf{Conjunct participles are typical foreground material}\\
Early Slavic conjunct participles can be both background and foreground clauses, but in their most typical functions they are part of the foreground content of a discourse unit. Like their Ancient Greek counterparts, following \citet{baryhaug2011} and \citet{haug2012a}, they are most typically either \textsc{independent rhemes} or \textsc{elaborations}, which are both foregrounded material in the sense that they encode new information and are part of the main line of events, but with very different semantic properties. As \textsc{independent rhemes}, they are discourse-coordinated and are typically interpreted as independent clauses, as the second and third conjunct participle in (\ref{indrheme_conc}). As \textsc{elaborations}, they add granularity to the matrix event and are most often interpreted as manner or means adverbials, as in (\ref{elab_conc}). 

\begin{example}
\gll borisa že {\normalfont [}oubivše{\normalfont ]}$_{frame}$ okan\foreignlanguage{russian}{ь}nii {\normalfont [}\textbf{ouvertěvše}{\normalfont ]}$_{indrheme}$ v šater\foreignlanguage{russian}{ъ} {\normalfont [}\textbf{v\foreignlanguage{russian}{ъ}zloživše}{\normalfont ]}$_{indrheme}$ na kola povezoša i
{Boris.{\sc sg.m.gen}} {\sc ptc} {kill.{\sc ptcp.pfv.m.nom.pl}} {cursed.{\sc m.nom.pl}} {wrap.{\sc ptcp.pfv.m.nom.pl}} {in} {tent.{\sc sg.m.acc}} {load.{\sc ptcp.pfv.m.nom.pl}} {on} {cart.{\sc sg.n.acc}} {take.{\sc aor.3.pl}} {\sc 3.sg.m.acc}
\glt ‘After killing Boris, the raiders wrapped him in a canvas, loaded him on a cart, and took him off’ (\textit{Primary Chronicle}, Codex Laurentianus 46b)
\glend
\label{indrheme_conc}
\end{example}

\begin{example}
\gll stoěchu \textbf{plačjušte} se i \textbf{divešte} se
stand.\textsc{impf.3.pl} cry.\textsc{ptcp.ipfv.m.nom.pl} \textsc{refl} and marvel.\textsc{ptcp.ipfv.m.nom.pl} \textsc{refl}
\glt ‘They stood crying and marvelling’ (\textit{Life of Mary, Abraham's Niece}, Bdinski Sbornik f. 16r)
\glend
\label{elab_conc}
\end{example}

Their temporal effects are also very different. \textsc{Independent rhemes} can (and normally do) introduce new temporal referents, and when perfective (which is most of the time in narrative discourse), they always induce narrative progression. \textsc{Elaborations} depend on the matrix event for their temporal reference and always temporally coincide with the matrix event. 

When occurring in chaining constructions, the leftmost conjunct participles can be ambiguous between \textsc{independent rhemes} and, in \posscitet{baryhaug2011} terminology, \textsc{frames}, which set the stage for and provide temporal anchoring for the matrix event, as in (\ref{bdinskiambigconc}). In specific configurations, they are instead most clearly \textsc{frames}, most notably in sentence-initial position with the subject following the participle, as in the first conjunct participle in (\ref{indrheme_conc}) above, or as part of two or more pre-matrix coordinated conjunct participles, as in (\ref{framecoordin}), where the coordination suggests that they are found in the specifier of the matrix I' (as with \textsc{frames}) rather than adjoined to I' (as with \textsc{independent rheme}) (\citealt{haug2012a}).

\begin{example}
\gll i {\normalfont [}\textbf{rastr\foreignlanguage{russian}{ь}zavši}{\normalfont ]}$_{indrheme/frame}$ ryzi svoe biěše se po licju i chotěše se sama udaviti ot pečaly
and tear.\textsc{part.prf.m.nom.sg} robe.\textsc{acc.pl} {her} beat.\textsc{impf.3.sg} \textsc{refl} on face.\textsc{dat} and want.\textsc{impf.3.sg} \textsc{refl} self suffocate.\textsc{inf} from sorrow.\textsc{gen.sg}
\glt ‘And {after tearing her own robes/she tore her own and} she started striking her own face and wanted to suffocate herself from sorrow’ (\textit{Life of Mary, Abraham's Niece}, Bdinski Sbornik f. 3v)
\glend
\label{bdinskiambigconc}
\end{example}

\begin{example}
\gll {\normalfont [}Sia že v\foreignlanguage{russian}{ъ}si \textbf{slyšavše} i \textbf{viděvše}{\normalfont ]} množaišǫ věrǫ i usr\foreignlanguage{russian}{ъ}die k\foreignlanguage{russian}{ъ} stmu pokazovaachǫ 
this.\textsc{n.acc.pl} \textsc{ptc} this.\textsc{n.acc.pl} hear.\textsc{ptcp.pfv.m.nom.sg} and see.\textsc{ptcp.pfv.m.nom.sg} more.\textsc{f.acc} faith.\textsc{f.acc} and zeal.\textsc{n.acc} to saint.\textsc{m.dat.sg} show.\textsc{impf.3.pl}
\glt `After hearing and seeing all these things, they started showing greater faith and zeal towards the Saint' \textit{Life of Ivan of Rila}, Zografski Sbornik f. 102r)
\glend
\label{framecoordin}
\end{example}

\item[(b.)] \textbf{Dative absolutes are typical background material}

Early Slavic dative absolutes are instead typically \textsc{frames}, i.e. background clauses. As such, they depend on the context for their own temporal reference, which is given by previously mentioned or easily inferable events. When occurring after the matrix clause, where they are much less frequent, they are most often interpreted as clause-final restrictive adjuncts (\citealt{fabricius-hansen2012b}). These are very similar to \textsc{frames} from the temporal and event-structuring perspective, since they restrict the domain of the matrix clause by introducing a \textit{state} within which the main eventuality holds, as in (\ref{postdasclearframe_conc}). Like clause-final restrictive adjuncts, post-matrix absolutes may also receive a contrastive reading, as in (\ref{postdasclearcontrast_conc}). 

\begin{example}
\gll on\foreignlanguage{russian}{ъ} že šed\foreignlanguage{russian}{ъ} sěde na stolě černigově jaroslavu \textbf{suštju} nověgorodě togda
{that.{\sc m.nom.sg}} {\sc ptc} {go.{\sc ptcp.pfv.m.nom.sg}} {sit.{\sc aor.3.sg}} {on} {throne.{\sc sg.m.loc}} {Chernigov.{\sc sg.m.loc}} {Yaroslav.{\sc m.dat.sg}} {be.{\sc ptcp.ipfv.m.dat.sg}} {Novgorod.{\sc sg.m.loc}} {then}
\glt ‘He thus departed thence and established himself upon the throne of Chernigov, while Yaroslav was in Novgorod.’ (\textit{Primary Chronicle}, Codex Laurentianus f. 50b)
\glend
\label{postdasclearframe_conc}
\end{example}

\begin{example}
\gll i povelě prinesti svěštę goręštę. i prižagati lice paule. propovědniku \textbf{v\foreignlanguage{russian}{ь}pijǫštu} i glagolǫštu
{and} {command.{\sc aor.3.sg}} {bring.{\sc inf}} {candle.{\sc pl.f.acc}} {burn.{\sc ptcp.ipfv.f.acc.pl}} {and} {burn.{\sc inf}} {face.{\sc sg.n.acc}} {Paul's.{\sc sg.n.acc.strong}} {preacher.{\sc m.dat.sg}} {cry out.{\sc ptcp.ipfv.m.dat.sg}} {and} {say.{\sc ptcp.ipfv.m.dat.sg}}
\glt ‘And he commanded to bring burning candles and burn Paul’s face, while the preacher was crying out, saying’ (\textit{Vita of Paul and Juliana}, Codex Suprasliensis f. 7r)
\glend
\label{postdasclearcontrast_conc}
\end{example}
  
A small number of post-matrix occurrences can function as \textsc{elaborations}, similarly to post-matrix conjunct participles, as in (\ref{postdasclearelab_conc}).
  
\begin{example}
\gll i to rek\foreignlanguage{russian}{ъ} paki po pervoe dr\foreignlanguage{russian}{ъ}žaše sę dobroe ustroenie. \textbf{bogu} \textbf{pomagajuštu} \textbf{emu} na blagoe proizvolenie
{and} {that.{\sc n.acc.sg}} {say.{\sc ptcp.pfv.m.nom.sg}} {more} {on} {righteous.{\sc n.acc.sg}} {stay.{\sc impf.3.sg}} {\sc refl} {good.{\sc n.acc.sg}} {behaviour.{\sc sg.n.acc}} {god.{\sc sg.m.dat}} {help.{\sc ptcp.ipfv.m.dat.sg}} {\sc 3.sg.m.dat} {on} {good.{\sc n.acc.sg}} {intention.{\sc n.acc.sg}}
\glt `After saying this, he stayed on the righteous path more than before, with God helping him in his good intentions' (\textit{Life of Sergij of Radonezh} f. 46v)
\glend
\label{postdasclearelab_conc}
\end{example}

However, several elaborating absolutes occur in rather predictable, perhaps formulaic, verb-subject combinations (as in (\ref{postdasclearelab_conc}) above), suggesting that the role of absolutes as \textsc{elaborations} is quite peripheral. 

\item[(c.)] \textbf{Perfective \textit{jegda}-events can include perfective matrix events}

\textit{Jegda}-clauses can be analysed as background clauses by default because of the explicit subordinator \textit{jegda} `when' (following SDRT's account of \textit{when}-clauses, see \citealt{asher2007a}), but their precise interpretation depends on several distinctions at the level of event structure (e.g. events versus states, durativity versus punctuality, telicity versus atelicity). Despite clear overlaps with participle \textsc{frames} in their distribution and discourse functions, at least one clear semantic difference emerged between \textit{jegda}-clauses and participle constructions. Namely, participle clauses, when perfective, always induce narrative progression, so that a perfective matrix clause is always interpreted as occurring (\textit{just}) \textit{after} the participle event. This led to the observation that perfective participle \textsc{frames} always introduce a consequent state within which the matrix clause occurs. While this is also possible with perfective \textit{jegda}-clauses, perfective \textit{jegda}-clauses can introduce a durative but \textit{bounded} referent \textit{within} which the matrix event occurs, which is a temporal reading that does not seem to be available to perfective participle constructions. (\ref{egdaoccasion_conc}) is one such example.

\begin{example}
\gll Byst\foreignlanguage{russian}{ъ} že \textbf{egda} \textbf{kr\foreignlanguage{russian}{ъ}stišę} \textbf{sę} v\foreignlanguage{russian}{ь}si ljud\foreignlanguage{russian}{ь}e. isou kr\foreignlanguage{russian}{ь}št\foreignlanguage{russian}{ь}šju sę i molęštju sę. otvr\foreignlanguage{russian}{ъ}ze sę nebo
{happen.{\sc aor.3.sg}} {\sc ptc} {when} {baptize.{\sc aor.3.pl}} {\sc refl} {all.{\sc pl.m.nom}} {people.{\sc pl.m.nom}} {Jesus.{\sc dat}} {baptize.{\sc ptcp.pfv.m.sg.dat}} {\sc refl} {and} {pray.{\sc ptcp.ipfv.m.sg.dat}} {\sc refl} {open.{\sc aor.3.sg}} {\sc refl} {heaven.{\sc sg.n.nom}}
\glt `It came to pass that, when all the people were baptized, as Jesus was baptized and as he was praying, the heaven was opened' (Luke 3:21)
\glend
\label{egdaoccasion_conc}
\end{example}

\item[(d.)] \textbf{{\small INDEPENDENT RHEMES} trigger Strong Narration, {\small FRAMES} trigger Background}

A formalization was proposed under an SDRT framework whereby conjunct participles functioning as \textsc{independent rhemes} trigger a Strong Narration rhetorical relation (\citealt{brasetal2001}) to the next event in a sequence. Strong Narration involves a strict discourse topic requirement and a temporal relation of \textit{abutement} (strict sequentiality) to the following event in a series (i.e. either another \textsc{independent rheme} or its matrix clause). Pre-matrix \textsc{frames} are instead formalized as triggers of a Background$_{forward}$ relation (\citealt{asher2007a}), which does not attach directly to the foreground (e.g. a series of \textsc{independent rhemes}), but introduces a more complex structure consisting of a Framing Topic, linked to the background clause(s) (e.g. dative absolutes or framing conjunct participles) via Background and to the foreground clause(s) via Elaboration. The discourse topic requirement of \textsc{independent rhemes} therefore nicely fits the possible interpretation of sentence-initial conjunct participles as \textsc{frames}, as well as the predominance of sentence-initial dative absolutes over those in any other positions in the sentence. In other words, while \textsc{independent rhemes} involve a strict discourse topic requirement, \textsc{frames} introduce one explicitly. 

\item[(e.)] \textbf{Sequentiality and event conflation as defining principles?}

The competition between participle adjuncts and \textit{jegda}-clauses cannot be captured in terms of discrete, categorical variables but must be modelled as a continuum, allowing a large degree of overlap. Yet, there are very clear cross-linguistic patterns in the way this continuum is divided between overtly subordinated (finite) constructions and juxtaposed (non-finite or `unbalanced', cf. \citealt{cristofarowals2,cristofarowals,cristofarosubordination}) constructions. A semantic map of \textsc{when} built from around 1400 linguistic varieties, including Old Church Slavonic, showed that `null' constructions (e.g. participles, converbs, simple main clause juxtaposition, that is, non-overtly subordinated constructions) cluster in particular regions, indicating that they are not equally viable as alternatives to any use of \textsc{when}, but carry particular meanings that make them less suitable for some of its functions. Comparisons with genealogically and areally unrelated but typologically similar languages showed that the majority of participle constructions in the semantic map correspond to cross-linguistic means used to stress the sequentiality of events, whereas \textit{jegda}-clauses show virtually no competition when it comes to \textit{universal} uses of \textsc{when} in any tense, as well as \textit{existential} uses \textit{in non-past tenses}. 

The stronger suitability of participle constructions to express sequentiality may be connected to the different temporal semantic properties mentioned above. These are the Strong Narration relation seemingly triggered by most pre-matrix conjunct participles (since these are most typically \textsc{independent rhemes}) and the possibility for perfective \textit{jegda}-clauses to be interpreted as \textit{including} the following event in line (matrix or adjunct), which, as we saw, is instead never a possible reading for perfective participle constructions (which always induce narrative progression). It is possible that participle \textsc{frames} contribute to stressing the sequentiality of events by triggering a Background relation, which entails the explicit introduction of a Framing Topic. The Framing Topic, then, ensures discourse cohesion between the foreground events and satisfies the strict discourse topic requirement of \textsc{independent rhemes}. 

Similar criteria of discourse organization govern the use of chaining constructions in clause-chaining languages and participle constructions in Old Church Slavonic, both of which are regulated, behind the specific principles and constraints defined in each language, by the discourse distinction between foreground and background (\citealt{dooleychains}). Crucially, both clause chaining in clause-chaining languages and participle constructions (whether as \textsc{frames}, \textsc{independent rhemes} or \textsc{elaborations}) in Early Slavic may be seen as a broader strategy for event grouping (or `conflation') and as a guarantee of tense-space-participant connectedness (\citealt{dooleychains}; Saebo 2011).

\item[(e.)] \textbf{`Non-canonical' participle constructions are canonical after all}

Virtually all `non-canonical' usages of participle constructions mentioned or discussed in the previous literature on Early Slavic participles (i.e. co-referential dative absolutes, syntactically independent absolutes and conjunct participles, and participle constructions with no apparent matrix clause; cf. \citealt{collins2011a}) are very well-attested in cross-linguistically phenomena that are functionally very similar to Early Slavic participles. 

In particular, a shift in agentivity of a subject, a shift in time, location or grounding (from background to foreground) can all warrant the use of absolute constructions even when their subject is co-referential with the one of their matrix clause (i.e. the most common `non-canonical' example of absolute constructions), similarly to how these same factors can explain the use of different-subject markers in several switch-reference systems across the world's languages. 

A shift in mood or actuality (\citealt{stirling_1993}), for example when introducing direct speech, as well as parenthetical or metadiscursive functions, may instead warrant insubordination (i.e. the use as independent clauses) of conjunct participles and dative absolutes. 

Given how both subject co-referentiality in dative absolutes and insubordination of participle constructions are found particularly frequently in original (non-translated) sources and outside of sacred and liturgical Old Church Slavonic texts, these clear cross-linguistic parallels incidentally also support the `nativeness' of the constructions and their independent use from their Greek counterparts.

\end{itemize}

\section{Where to from here?}
The cross-linguistic perspective on Early Slavic participle constructions has shown, I believe, that token-based typological approaches can be very fruitful when adopted to study constructions in historical languages, for which elicitation of new data is not an option, for obvious reasons. The final chapter, in particular, has highlighted the great similarities between the functions of phenomena attested cross-linguistically in languages areally and genealogically unrelated to Old Church Sla\-vo\-nic, particularly switch-reference systems, and provided Early Slavic examples that clearly support such similarities. These examples were found upon close reading the occurrences of the constructions and checking their surrounding discourse. However, some of these phenomena lend themselves well to larger-scale, quantitative analyses. 

In particular, clause bridging, presented in Section \ref{sec:bridging} of Chapter 6, has a regular, schematizable structure (cf. \citealt{guerinbridge} and \citealt{anderboisaltshuler22}), which is worthwhile attempting to capture quantitatively in Early Slavic, by leveraging the recurrence of lemmas in adjacent portions of discourse. We have preliminarily shown (particularly in Section \ref{secbdinski} of Chapter 2, besides Section \ref{sec:bridging} of Chapter 6) that clause bridging can be considered one of the functions of dative absolutes, but the close reading of examples from the corpus indicated that it likely deserves a separate, dedicated study investigating how pervasive it is, whereas in this thesis it only received relatively sporadic mentions.

Similar observations hold for the correspondences to switch-reference systems in the world's languages. As pointed out in Section \ref{sec:switch} of Chapter 6, a large-scale computational comparison would require annotation at the morphosyntactic level for more languages than current computational tools can achieve, especially considering that several switch-reference languages are extremely low-resourced. However, we have seen that it is possible to obtain some clear patterns by automatically adding annotation for switch-reference markers in languages where these are relatively isolable. A promising next step, therefore, is to annotate several more languages for switch reference marking, following the plethora of existing descriptions of such systems. A larger parallel dataset with switch-reference annotation may then allow for semantic maps capturing finer-grained semantic and discourse distinctions in the usage of switch-reference markers and make better-informed comparisons with the distributions of participle constructions in Early Slavic.\footnote{See an ongoing experiment in `subtoken-based typology' in \citet{pedrazziniamericasnlp}.}

An aspect that was not taken into account in the semantic map of \textsc{when}, nor in this thesis as a whole, is the discourse functions of much more polyfunctional subordinators, most notably Old Church Slavonic \textit{jako}, whose meanings span `when', `as', `if', `because', `how', `since', among others, and similarly Ancient Greek \textit{hōs}, which is very similar in meaning to Old Church Slavonic \textit{jako}. `When' used as a question word (\textit{kogda} in Old Church Slavonic) has also emerged from the semantic maps of Old Church Slavonic and is likely to deserve separate investigation. The semantic map of \textsc{when} pointed to potential cross-linguistic patterns of coexpression, whereby the question word \textit{when}, when not separately lexicalized, is colexified with universal \textsc{when}. However, this issue was not addressed directly, and our study hardly allows for proper generalizations on this matter.

From a theoretical perspective, I only touched upon the idea that clause chaining in Early Slavic and clause-chaining languages may convey a higher degree of event integration than what is conveyed by \textit{jegda}-clauses and finite temporal subordinates in general. However, event conflation and the structure of macro-events are theoretically well-established, so a comprehensive account of finite and non-finite competitors in Early Slavic may benefit from fully modelling the competition within an event conflation theory as conceived and progressively refined mainly by Leonard Talmy since the 1970s (\citealt{talmy72,talmy75,talmy85,talmy91,talmy00}).

Finally, the formulaic tendencies observed on dative absolutes across the corpus certainly deserve a dedicated study. We have seen (particularly throughout Chapter 2, but also as generally transpired from the low lexical variation) that dative absolutes tend to occur in very predictable participle lemma-subject lemma combinations and, not infrequently, in what seems like fixed or semi-fixed expressions. We have also noticed how some of these expressions are attested across different texts and genres, for example in generic temporal \textsc{frames} such as \textit{pozdě byv\foreignlanguage{russian}{ъ}šu} `when it got late' and variations of the basic phrase \textit{ne věduštju nikomuže} `without anyone knowing', as well as seemingly semi-fixed expressions of the type `God <\textsc{help}> <\textsc{person}>', where \textsc{help} is generally one of a set of verbs semantically similar to \textit{help} (e.g. \textit{s\foreignlanguage{russian}{ъ}bljudati} `guard, watch over', \textit{pomagati} `help', or \textit{ukrěpljati} `give strength'). Other expressions, however, seem predictable only contextually to a specific text or genre, for instance in `reigning <\textsc{ruler}> in <\textsc{place}>' (e.g. \textit{crstvujuštju ol\foreignlanguage{russian}{ь}kse v\foreignlanguage{russian}{ъ} csrigradě} `when Oleksa was reigning in Constantinople'), which are predominantly found in the chronicles. The `predictability' of dative absolutes, then, seems to be well-suited for a study of the occurrences of the construction as potential instantiations of a limited set of \textit{constructions} in the Construction Grammar sense of the term (i.e. `learned pairings of form with semantic or discourse functions'; \pgcitealt{goldberg2005}{5}). The question may be fruitfully addressed under a distributional semantic framework and using language models trained on Early Slavic to capture the productivity of absolute constructions at different levels of abstraction (i.e. a lower level of abstraction may consider position and tense-aspect combinations as a construction, e.g. <absolute lemma-\textsc{pfv} + matrix lemma-\textsc{aor}>; a higher level of abstraction may only consider aspect; the highest level may consider any configuration, and so on), in the spirit of what studies such as \citet{perek}, \citet{bozzonephd}, \citet{rodda-etal-2019-vector} and \citet{roddaphd} have done on different phenomena in English and Ancient Greek. Despite the high diachronic and diatopic variation of historical languages, historical language modelling has been advancing at a rapid pace, which makes distributional semantic studies of variation in participle constructions much more likely to be achievable than had been previously conceived.\footnote{Ancient Greek word embeddings I have developed with colleagues, for example, are described in \citet{stopponietalalp2023} and \citet{ichlspecial} and available in \citet{stopponi_2023_8369516}. Language modelling of Early Slavic, unsurprisingly, has so far received much less (and virtually no) attention, except for preliminary attempts by \citet{pedrazziniocsembeddings}.} 

\appendix

\chapter{Massively parallel dataset: corpus representativeness} 
% \minitoc

\section{Language families}

Table \ref{tab:completelangdfamilies} lists all the world's language families according to the Glottolog database (\url{https://glottolog.org}; \citealt{nordhoffhammarstrom}, \citealt{glottolog2021}), how many languages belong to each of them, always following Glottolog, compared to how many languages from each family are represented in the massively parallel Bible corpus used in Chapters 5 and 6. In the table, \textit{bible\textunderscore } refers to the massively parallel corpus and \textit{world\textunderscore } to the languages family from Glottolog. \textit{raw} are the raw number of languages belonging to the relevant family, \textit{rel} is the relative frequency of these in relation to the total number of languages in the respective dataset (the parallel Bible dataset for \textit{bible\textunderscore }, the whole Glottolog language database for \textit{world\textunderscore }).

\begin{longtable}{|l|l|l|l|l|}
        \hline
        \textbf{family} & \textbf{world\textunderscore raw} & \textbf{world\textunderscore rel} & \textbf{bible\textunderscore raw} & \textbf{bible\textunderscore rel}\\ \hline
        \textbf{Atlantic-Congo} & 1380 & 18.1\% & 249.0 & 17.2\% \\ \hline
        \textbf{Austronesian} & 1289 & 16.9\% & 246.0 & 17.0\% \\ \hline
        \textbf{Indo-European} & 595 & 7.8\% & 110.0 & 7.6\% \\ \hline
        \textbf{Sino-Tibetan} & 441 & 5.8\% & 90.0 & 6.2\% \\ \hline
        \textbf{Afro-Asiatic} & 371 & 4.9\% & 47.0 & 3.3\% \\ \hline
        \textbf{Nuclear Trans New Guinea} & 313 & 4.1\% & 94.0 & 6.5\% \\ \hline
        \textbf{Pama-Nyungan} & 268 & 3.5\% & 13.0 & 0.9\% \\ \hline
        \textbf{Otomanguean} & 180 & 2.4\% & 79.0 & 5.5\% \\ \hline
        \textbf{Isolate} & 167 & 2.2\% & 16.0 & 1.1\% \\ \hline
        \textbf{Austroasiatic} & 160 & 2.1\% & 10.0 & 0.7\% \\ \hline
        \textbf{Sign Language} & 156 & 2.1\% & Absent & Absent \\ \hline
        \textbf{Tai-Kadai} & 91 & 1.2\% & 3.0 & 0.2\% \\ \hline
        \textbf{Dravidian} & 84 & 1.1\% & 6.0 & 0.4\% \\ \hline
        \textbf{Mande} & 77 & 1.0\% & 24.0 & 1.7\% \\ \hline
        \textbf{Tupian} & 73 & 1.0\% & 17.0 & 1.2\% \\ \hline
        \textbf{Central Sudanic} & 65 & 0.9\% & 15.0 & 1.0\% \\ \hline
        \textbf{Uto-Aztecan} & 64 & 0.8\% & 26.0 & 1.8\% \\ \hline
        \textbf{Nilotic} & 57 & 0.7\% & 21.0 & 1.5\% \\ \hline
        \textbf{Arawakan} & 56 & 0.7\% & 23.0 & 1.6\% \\ \hline
        \textbf{Nuclear Torricelli} & 55 & 0.7\% & 9.0 & 0.6\% \\ \hline
        \textbf{Algic} & 48 & 0.6\% & 8.0 & 0.6\% \\ \hline
        \textbf{Athabaskan-Eyak-Tlingit} & 46 & 0.6\% & 7.0 & 0.5\% \\ \hline
        \textbf{Quechuan} & 45 & 0.6\% & 27.0 & 1.9\% \\ \hline
        \textbf{Turkic} & 45 & 0.6\% & 21.0 & 1.5\% \\ \hline
        \textbf{Uralic} & 41 & 0.5\% & 8.0 & 0.6\% \\ \hline
        \textbf{Kru} & 38 & 0.5\% & 9.0 & 0.6\% \\ \hline
        \textbf{Hmong-Mien} & 37 & 0.5\% & 3.0 & 0.2\% \\ \hline
        \textbf{Mayan} & 35 & 0.5\% & 25.0 & 1.7\% \\ \hline
        \textbf{Sepik} & 34 & 0.4\% & 8.0 & 0.6\% \\ \hline
        \textbf{Pano-Tacanan} & 33 & 0.4\% & 11.0 & 0.8\% \\ \hline
        \textbf{Nakh-Daghestanian} & 32 & 0.4\% & 6.0 & 0.4\% \\ \hline
        \textbf{Lower Sepik-Ramu} & 30 & 0.4\% & 4.0 & 0.3\% \\ \hline
        \textbf{Cariban} & 29 & 0.4\% & 8.0 & 0.6\% \\ \hline
        \textbf{Salishan} & 27 & 0.4\% & Absent & Absent \\ \hline
        \textbf{Artificial} & 26 & 0.3\% & 2.0 & 0.1\% \\ \hline
        \textbf{Nuclear-Macro-Je} & 25 & 0.3\% & 7.0 & 0.5\% \\ \hline
        \textbf{Tucanoan} & 23 & 0.3\% & 16.0 & 1.1\% \\ \hline
        \textbf{Timor-Alor-Pantar} & 23 & 0.3\% & Absent & Absent \\ \hline
        \textbf{Chibchan} & 21 & 0.3\% & 9.0 & 0.6\% \\ \hline
        \textbf{Lakes Plain} & 20 & 0.3\% & Absent & Absent \\ \hline
        \textbf{Ta-Ne-Omotic} & 19 & 0.2\% & 7.0 & 0.5\% \\ \hline
        \textbf{Dogon} & 19 & 0.2\% & 1.0 & 0.1\% \\ \hline
        \textbf{Siouan} & 18 & 0.2\% & 1.0 & 0.1\% \\ \hline
        \textbf{Mixe-Zoque} & 17 & 0.2\% & 8.0 & 0.6\% \\ \hline
        \textbf{Yam} & 17 & 0.2\% & Absent & Absent \\ \hline
        \textbf{Anim} & 16 & 0.2\% & Absent & Absent \\ \hline
        \textbf{Mongolic-Khitan} & 16 & 0.2\% & 3.0 & 0.2\% \\ \hline
        \textbf{Pidgin} & 16 & 0.2\% & 1.0 & 0.1\% \\ \hline
        \textbf{Border} & 15 & 0.2\% & 2.0 & 0.1\% \\ \hline
        \textbf{North Halmahera} & 15 & 0.2\% & 3.0 & 0.2\% \\ \hline
        \textbf{Nubian} & 14 & 0.2\% & Absent & Absent \\ \hline
        \textbf{Angan} & 13 & 0.2\% & 4.0 & 0.3\% \\ \hline
        \textbf{Ndu} & 13 & 0.2\% & 3.0 & 0.2\% \\ \hline
        \textbf{Japonic} & 13 & 0.2\% & 1.0 & 0.1\% \\ \hline
        \textbf{Tor-Orya} & 13 & 0.2\% & Absent & Absent \\ \hline
        \textbf{Khoe-Kwadi} & 13 & 0.2\% & 2.0 & 0.1\% \\ \hline
        \textbf{Iroquoian} & 13 & 0.2\% & 2.0 & 0.1\% \\ \hline
        \textbf{Tungusic} & 12 & 0.2\% & 1.0 & 0.1\% \\ \hline
        \textbf{Totonacan} & 12 & 0.2\% & 8.0 & 0.6\% \\ \hline
        \textbf{Eskimo-Aleut} & 12 & 0.2\% & 4.0 & 0.3\% \\ \hline
        \textbf{Worrorran} & 12 & 0.2\% & Absent & Absent \\ \hline
        \textbf{Western Daly} & 11 & 0.1\% & Absent & Absent \\ \hline
        \textbf{Great Andamanese} & 11 & 0.1\% & Absent & Absent \\ \hline
        \textbf{Geelvink Bay} & 10 & 0.1\% & 1.0 & 0.1\% \\ \hline
        \textbf{Miwok-Costanoan} & 10 & 0.1\% & Absent & Absent \\ \hline
        \textbf{Sko} & 10 & 0.1\% & Absent & Absent \\ \hline
        \textbf{Songhay} & 10 & 0.1\% & 2.0 & 0.1\% \\ \hline
        \textbf{Heibanic} & 10 & 0.1\% & 1.0 & 0.1\% \\ \hline
        \textbf{Nyulnyulan} & 10 & 0.1\% & Absent & Absent \\ \hline
        \textbf{Ijoid} & 10 & 0.1\% & Absent & Absent \\ \hline
        \textbf{Gunwinyguan} & 10 & 0.1\% & 2.0 & 0.1\% \\ \hline
        \textbf{Saharan} & 10 & 0.1\% & Absent & Absent \\ \hline
        \textbf{Surmic} & 10 & 0.1\% & 1.0 & 0.1\% \\ \hline
        \textbf{Dagan} & 9 & 0.1\% & 4.0 & 0.3\% \\ \hline
        \textbf{Chocoan} & 9 & 0.1\% & 3.0 & 0.2\% \\ \hline
        \textbf{South Bougainville} & 9 & 0.1\% & 1.0 & 0.1\% \\ \hline
        \textbf{Maban} & 9 & 0.1\% & Absent & Absent \\ \hline
        \textbf{Cochimi-Yuman} & 9 & 0.1\% & Absent & Absent \\ \hline
        \textbf{Koiarian} & 8 & 0.1\% & 5.0 & 0.3\% \\ \hline
        \textbf{Greater Kwerba} & 8 & 0.1\% & Absent & Absent \\ \hline
        \textbf{Narrow Talodi} & 8 & 0.1\% & Absent & Absent \\ \hline
        \textbf{Dajuic} & 7 & 0.1\% & Absent & Absent \\ \hline
        \textbf{Muskogean} & 7 & 0.1\% & Absent & Absent \\ \hline
        \textbf{Matacoan} & 7 & 0.1\% & 6.0 & 0.4\% \\ \hline
        \textbf{Pomoan} & 7 & 0.1\% & Absent & Absent \\ \hline
        \textbf{Bosavi} & 7 & 0.1\% & 3.0 & 0.2\% \\ \hline
        \textbf{Tuu} & 7 & 0.1\% & Absent & Absent \\ \hline
        \textbf{Kiwaian} & 6 & 0.1\% & Absent & Absent \\ \hline
        \textbf{East Strickland} & 6 & 0.1\% & 2.0 & 0.1\% \\ \hline
        \textbf{Baining} & 6 & 0.1\% & 1.0 & 0.1\% \\ \hline
        \textbf{Kartvelian} & 6 & 0.1\% & 1.0 & 0.1\% \\ \hline
        \textbf{Nambiquaran} & 6 & 0.1\% & 1.0 & 0.1\% \\ \hline
        \textbf{Left May} & 6 & 0.1\% & 1.0 & 0.1\% \\ \hline
        \textbf{Lengua-Mascoy} & 6 & 0.1\% & 2.0 & 0.1\% \\ \hline
        \textbf{Chumashan} & 6 & 0.1\% & Absent & Absent \\ \hline
        \textbf{Mailuan} & 6 & 0.1\% & Absent & Absent \\ \hline
        \textbf{Kiowa-Tanoan} & 6 & 0.1\% & 1.0 & 0.1\% \\ \hline
        \textbf{Kadugli-Krongo} & 6 & 0.1\% & Absent & Absent \\ \hline
        \textbf{Zaparoan} & 6 & 0.1\% & 1.0 & 0.1\% \\ \hline
        \textbf{Wakashan} & 6 & 0.1\% & Absent & Absent \\ \hline
        \textbf{South Bird's Head Family} & 6 & 0.1\% & Absent & Absent \\ \hline
        \textbf{Arawan} & 6 & 0.1\% & 2.0 & 0.1\% \\ \hline
        \textbf{Yareban} & 5 & 0.1\% & 2.0 & 0.1\% \\ \hline
        \textbf{Caddoan} & 5 & 0.1\% & Absent & Absent \\ \hline
        \textbf{Abkhaz-Adyge} & 5 & 0.1\% & 1.0 & 0.1\% \\ \hline
        \textbf{South Omotic} & 5 & 0.1\% & 1.0 & 0.1\% \\ \hline
        \textbf{Yanomamic} & 5 & 0.1\% & 2.0 & 0.1\% \\ \hline
        \textbf{Pauwasi} & 5 & 0.1\% & 1.0 & 0.1\% \\ \hline
        \textbf{Kxa} & 5 & 0.1\% & Absent & Absent \\ \hline
        \textbf{Guaicuruan} & 5 & 0.1\% & 4.0 & 0.3\% \\ \hline
        \textbf{Yeniseian} & 5 & 0.1\% & 1.0 & 0.1\% \\ \hline
        \textbf{Chukotko-Kamchatkan} & 5 & 0.1\% & 1.0 & 0.1\% \\ \hline
        \textbf{Keram} & 5 & 0.1\% & Absent & Absent \\ \hline
        \textbf{Barbacoan} & 5 & 0.1\% & 4.0 & 0.3\% \\ \hline
        \textbf{Eleman} & 5 & 0.1\% & Absent & Absent \\ \hline
        \textbf{Mirndi} & 5 & 0.1\% & Absent & Absent \\ \hline
        \textbf{Guahiboan} & 5 & 0.1\% & 3.0 & 0.2\% \\ \hline
        \textbf{Misumalpan} & 5 & 0.1\% & 2.0 & 0.1\% \\ \hline
        \textbf{Sahaptian} & 5 & 0.1\% & Absent & Absent \\ \hline
        \textbf{Yuat} & 5 & 0.1\% & 1.0 & 0.1\% \\ \hline
        \textbf{Nimboranic} & 5 & 0.1\% & Absent & Absent \\ \hline
        \textbf{Huitotoan} & 5 & 0.1\% & 2.0 & 0.1\% \\ \hline
        \textbf{West Bird's Head} & 5 & 0.1\% & Absent & Absent \\ \hline
        \textbf{Naduhup} & 4 & 0.1\% & 1.0 & 0.1\% \\ \hline
        \textbf{Huavean} & 4 & 0.1\% & 1.0 & 0.1\% \\ \hline
        \textbf{Koman} & 4 & 0.1\% & 1.0 & 0.1\% \\ \hline
        \textbf{Tangkic} & 4 & 0.1\% & Absent & Absent \\ \hline
        \textbf{North Bougainville} & 4 & 0.1\% & 1.0 & 0.1\% \\ \hline
        \textbf{Suki-Gogodala} & 4 & 0.1\% & Absent & Absent \\ \hline
        \textbf{Sentanic} & 4 & 0.1\% & Absent & Absent \\ \hline
        \textbf{Chicham} & 4 & 0.1\% & 4.0 & 0.3\% \\ \hline
        \textbf{Iwaidjan Proper} & 4 & 0.1\% & 1.0 & 0.1\% \\ \hline
        \textbf{Walioic} & 4 & 0.1\% & Absent & Absent \\ \hline
        \textbf{Chapacuran} & 4 & 0.1\% & Absent & Absent \\ \hline
        \textbf{Yukaghir} & 4 & 0.1\% & 1.0 & 0.1\% \\ \hline
        \textbf{Eastern Jebel} & 4 & 0.1\% & Absent & Absent \\ \hline
        \textbf{Blue Nile Mao} & 4 & 0.1\% & Absent & Absent \\ \hline
        \textbf{Mangarrayi-Maran} & 4 & 0.1\% & Absent & Absent \\ \hline
        \textbf{Maiduan} & 4 & 0.1\% & Absent & Absent \\ \hline
        \textbf{Wintuan} & 4 & 0.1\% & Absent & Absent \\ \hline
        \textbf{Eastern Trans-Fly} & 4 & 0.1\% & 2.0 & 0.1\% \\ \hline
        \textbf{Turama-Kikori} & 4 & 0.1\% & Absent & Absent \\ \hline
        \textbf{Maningrida} & 4 & 0.1\% & 1.0 & 0.1\% \\ \hline
        \textbf{Chinookan} & 3 & 0.0\% & Absent & Absent \\ \hline
        \textbf{Lepki-Murkim-Kembra} & 3 & 0.0\% & Absent & Absent \\ \hline
        \textbf{Kwalean} & 3 & 0.0\% & 1.0 & 0.1\% \\ \hline
        \textbf{Giimbiyu} & 3 & 0.0\% & Absent & Absent \\ \hline
        \textbf{Kayagaric} & 3 & 0.0\% & Absent & Absent \\ \hline
        \textbf{Yangmanic} & 3 & 0.0\% & Absent & Absent \\ \hline
        \textbf{Tamaic} & 3 & 0.0\% & Absent & Absent \\ \hline
        \textbf{Dizoid} & 3 & 0.0\% & Absent & Absent \\ \hline
        \textbf{Kunimaipan} & 3 & 0.0\% & 3.0 & 0.2\% \\ \hline
        \textbf{ArafundiArafundi} & 3 & 0.0\% & Absent & Absent \\ \hline
        \textbf{East Bird's Head} & 3 & 0.0\% & 3.0 & 0.2\% \\ \hline
        \textbf{West Bomberai} & 3 & 0.0\% & Absent & Absent \\ \hline
        \textbf{Kalapuyan} & 3 & 0.0\% & Absent & Absent \\ \hline
        \textbf{Kolopom} & 3 & 0.0\% & Absent & Absent \\ \hline
        \textbf{Saliban} & 3 & 0.0\% & Absent & Absent \\ \hline
        \textbf{Tsimshian} & 3 & 0.0\% & Absent & Absent \\ \hline
        \textbf{Jarrakan} & 3 & 0.0\% & Absent & Absent \\ \hline
        \textbf{Aymaran} & 3 & 0.0\% & 1.0 & 0.1\% \\ \hline
        \textbf{Kamula-Elevala} & 3 & 0.0\% & 2.0 & 0.1\% \\ \hline
        \textbf{Mixed Language} & 3 & 0.0\% & Absent & Absent \\ \hline
        \textbf{Mairasic} & 3 & 0.0\% & Absent & Absent \\ \hline
        \textbf{Kuliak} & 3 & 0.0\% & Absent & Absent \\ \hline
        \textbf{Bororoan} & 3 & 0.0\% & Absent & Absent \\ \hline
        \textbf{Amto-Musan} & 2 & 0.0\% & Absent & Absent \\ \hline
        \textbf{Kresh-Aja} & 2 & 0.0\% & Absent & Absent \\ \hline
        \textbf{Gumuz} & 2 & 0.0\% & 1.0 & 0.1\% \\ \hline
        \textbf{Teberan} & 2 & 0.0\% & 2.0 & 0.1\% \\ \hline
        \textbf{Senagi} & 2 & 0.0\% & 1.0 & 0.1\% \\ \hline
        \textbf{Harakmbut} & 2 & 0.0\% & 1.0 & 0.1\% \\ \hline
        \textbf{Manubaran} & 2 & 0.0\% & 1.0 & 0.1\% \\ \hline
        \textbf{Doso-Turumsa} & 2 & 0.0\% & Absent & Absent \\ \hline
        \textbf{Inanwatan} & 2 & 0.0\% & Absent & Absent \\ \hline
        \textbf{Furan} & 2 & 0.0\% & Absent & Absent \\ \hline
        \textbf{Nyimang} & 2 & 0.0\% & Absent & Absent \\ \hline
        \textbf{GarrwanGarawa} & 2 & 0.0\% & 1.0 & 0.1\% \\ \hline
        \textbf{Limilngan-Wulna} & 2 & 0.0\% & Absent & Absent \\ \hline
        \textbf{Peba-Yagua} & 2 & 0.0\% & 1.0 & 0.1\% \\ \hline
        \textbf{Cahuapanan} & 2 & 0.0\% & 1.0 & 0.1\% \\ \hline
        \textbf{Eastern Daly} & 2 & 0.0\% & Absent & Absent \\ \hline
        \textbf{Kakua-Nukak} & 2 & 0.0\% & 1.0 & 0.1\% \\ \hline
        \textbf{Bulaka River} & 2 & 0.0\% & Absent & Absent \\ \hline
        \textbf{Keresan} & 2 & 0.0\% & Absent & Absent \\ \hline
        \textbf{Tarascan} & 2 & 0.0\% & 2.0 & 0.1\% \\ \hline
        \textbf{Haida} & 2 & 0.0\% & Absent & Absent \\ \hline
        \textbf{Zamucoan} & 2 & 0.0\% & 2.0 & 0.1\% \\ \hline
        \textbf{East Kutubu} & 2 & 0.0\% & Absent & Absent \\ \hline
        \textbf{Chonan} & 2 & 0.0\% & Absent & Absent \\ \hline
        \textbf{Jarawa-Onge} & 2 & 0.0\% & Absent & Absent \\ \hline
        \textbf{Palaihnihan} & 2 & 0.0\% & Absent & Absent \\ \hline
        \textbf{Konda-Yahadian} & 2 & 0.0\% & Absent & Absent \\ \hline
        \textbf{Yuki-Wappo} & 2 & 0.0\% & Absent & Absent \\ \hline
        \textbf{Mombum-Koneraw} & 2 & 0.0\% & Absent & Absent \\ \hline
        \textbf{Yawa-Saweru} & 2 & 0.0\% & 1.0 & 0.1\% \\ \hline
        \textbf{Baibai-Fas} & 2 & 0.0\% & Absent & Absent \\ \hline
        \textbf{Tequistlatecan} & 2 & 0.0\% & 1.0 & 0.1\% \\ \hline
        \textbf{Rashad} & 2 & 0.0\% & Absent & Absent \\ \hline
        \textbf{Temeinic} & 2 & 0.0\% & Absent & Absent \\ \hline
        \textbf{Hibito-Cholon} & 2 & 0.0\% & Absent & Absent \\ \hline
        \textbf{Piawi} & 2 & 0.0\% & 1.0 & 0.1\% \\ \hline
        \textbf{Koreanic} & 2 & 0.0\% & 1.0 & 0.1\% \\ \hline
        \textbf{Bunaban} & 2 & 0.0\% & Absent & Absent \\ \hline
        \textbf{Katukinan} & 2 & 0.0\% & Absent & Absent \\ \hline
        \textbf{Northern Daly} & 2 & 0.0\% & Absent & Absent \\ \hline
        \textbf{Kwomtari-Nai} & 2 & 0.0\% & Absent & Absent \\ \hline
        \textbf{Southern Daly} & 2 & 0.0\% & 1.0 & 0.1\% \\ \hline
        \textbf{Bogia} & 2 & 0.0\% & Absent & Absent \\ \hline
        \textbf{ChimakuanChimakum} & 2 & 0.0\% & Absent & Absent \\ \hline
        \textbf{Somahai} & 2 & 0.0\% & Absent & Absent \\ \hline
        \textbf{Boran} & 2 & 0.0\% & 2.0 & 0.1\% \\ \hline
        \textbf{Coosan} & 2 & 0.0\% & Absent & Absent \\ \hline
        \textbf{Marrku-Wurrugu} & 2 & 0.0\% & Absent & Absent \\ \hline
        \textbf{Pahoturi} & 2 & 0.0\% & Absent & Absent \\ \hline
        \textbf{Katla-Tima} & 2 & 0.0\% & Absent & Absent \\ \hline
        \textbf{Bayono-Awbono} & 2 & 0.0\% & Absent & Absent \\ \hline
        \textbf{Namla-Tofanma} & 2 & 0.0\% & Absent & Absent \\ \hline
        \textbf{Kaure-Kosare} & 2 & 0.0\% & Absent & Absent \\ \hline
        \textbf{Hurro-Urartian} & 2 & 0.0\% & Absent & Absent \\ \hline
        \textbf{Uru-Chipaya} & 2 & 0.0\% & 1.0 & 0.1\% \\ \hline
        \textbf{Araucanian} & 2 & 0.0\% & 1.0 & 0.1\% \\ \hline
        \textbf{Puri-Coroado} & 1 & 0.0\% & Absent & Absent \\ \hline
        \textbf{Kawesqar} & 1 & 0.0\% & Absent & Absent \\ \hline
        \textbf{Ainu} & 1 & 0.0\% & Absent & Absent \\ \hline
        \textbf{Kamakanan} & 1 & 0.0\% & Absent & Absent \\ \hline
        \textbf{Hatam-Mansim} & 1 & 0.0\% & Absent & Absent \\ \hline
        \textbf{Shastan} & 1 & 0.0\% & Absent & Absent \\ \hline
        \textbf{Yokutsan} & 1 & 0.0\% & Absent & Absent \\ \hline
        \textbf{Nivkh} & 1 & 0.0\% & 1.0 & 0.1\% \\ \hline
        \textbf{Chiquitano} & 1 & 0.0\% & 1.0 & 0.1\% \\ \hline
        \textbf{Jicaquean} & 1 & 0.0\% & 1.0 & 0.1\% \\ \hline
        \textbf{Ticuna-Yuri} & 1 & 0.0\% & 1.0 & 0.1\% \\ \hline
        \textbf{Taulil-Butam} & 1 & 0.0\% & Absent & Absent \\ \hline
\caption{Language families represented in the massively parallel Bible dataset compared with the world's language families according to Glottolog.}\label{tab:completelangdfamilies}
\end{longtable}

\chapter{Language classification by pattern of co-expression} 
% \minitoc

Table \ref{langpatternscomplete} contains all the languages in the massively parallel dataset used in Chapters 5 and 6 to generate the semantic map of \textsc{when}. In the table, languages are ordered by their ISO 639-3 code and classified by pattern and subpattern, which was assigned automatically as outlined below. Each language has an indication of its name, language family, and main macroarea in which it is attested. This metadata, as explained in Section \ref{sec:data} of Chapter 5, is obtained from the Glottolog database (\url{https://glottolog.org}; \citealt{nordhoffhammarstrom}, \citealt{glottolog2021}).

\section{Pattern and subpattern assignment}
The three basic patterns are defined on the basis of the top-left (TL), mid-left (ML) and bottom-left (BL) areas of the semantic map of when and assigned as shown in Table \ref{basicpatterns}. 

\begin{table}[H]
\centering
\begin{tabular}{|l|l|}
\hline
\textbf{pattern} & \textbf{areas coexpressed} \\\hline
\textbf{A} & TL = ML = BL \\
\textbf{B} & (TL = ML) $\ne$ BL \\
\textbf{C} & TL $\ne$ (ML = BL )\\
\textbf{D} & TL $\ne$ ML $\ne$ BL \\
\textbf{E} & (TL = BL) $\ne$ ML \\
\hline
\end{tabular}
\caption{Basic coexpression patterns}
\label{basicpatterns}
\end{table}

\noindent Subpatterns are assigned to languages that do not fit one of the five basic patterns above, due to one or more of the clusters being under the scope of more than one linguistic item. Depending on how many items there are in total for the three areas, there is a different number of total logical combinations, all of which are taken into consideration when assigning a subpattern. To take the possible combinations into account, we use a construction with the following template:

\begin{quote}
\centering
    \texttt{[  ]$_{TL}$ [  ]$_{ML}$ [  ]$_{BL}$}
\end{quote}

\noindent where each slot is filled by arbitrary letters standing for each different linguistic item having scope over the relevant cluster. To capture that two or more clusters are colexified, the same latter is repeated in all those clusters and, conversely, different letters indicate different linguistic items. In this way, we can schematically represent more complex patterns than those in Table \ref{basicpatterns}. For example, below, (a.) corresponds to pattern D, since the three cluster are dominated by three different linguistic items. Instead, (b.) represents a language in which the TL cluster is dominated by both a lexical item unique to that cluster (\texttt{X} in the template), but the same item that has scope on the ML cluster has also scope over the TL cluster (\texttt{Y} in the template). (c.) instead corresponds to a language in which the ML and BL clusters are colexified, while the TL cluster is under the scope of not one, but two lexical items.

\begin{quote}
\begin{itemize}
    \item[(a.)] \texttt{[X]$_{TL}$ [Y]$_{ML}$ [Z]$_{BL}$}
    \item[(b.)] \texttt{[XY]$_{TL}$ [Y]$_{ML}$ [Z]$_{BL}$}
    \item[(c.)] \texttt{[XY]$_{TL}$ [Z]$_{ML}$ [Z]$_{BL}$}
\end{itemize}
\end{quote}

In the overview in Table \ref{subpatternstab}, for ease of presentation, the three slots in the template are only separated by a space, but the order remains the same as in the template shown above. For example, (c.) above is represented as \texttt{XX Z Z}. \texttt{?} indicates `no lexical item', representing a language in which the relevant cluster has no Kriging area.
\newpage

\begin{table}[H]
\centering
\begin{tabular}{|l|l|}
\hline
\textbf{configuration} & \textbf{subpattern label} \\\hline
\texttt{XY Y X}  & \textbf{BxE} \\
\texttt{X XY Y}  & \textbf{BxC} \\
\texttt{X Y XY}  & \textbf{CxE} \\
\texttt{XW Y Z}  & \textbf{D2a} \\
\texttt{X YW Z}  & \textbf{D2b} \\
\texttt{X Y ZW}  & \textbf{D2c} \\
\texttt{XZ Y Y}  & \textbf{C3} \\
\texttt{X YZ X}  & \textbf{E3} \\
\texttt{X X YZ}  & \textbf{B3} \\
\texttt{XY X X}  & \textbf{CxA} \\
\texttt{X XY X}  & \textbf{ExA} \\
\texttt{X X XY}  & \textbf{BxA} \\
\texttt{X YX Z}  & \textbf{DxX} \\
\texttt{X Y ZX}  & \textbf{DxX} \\
\texttt{XY Y Z}  & \textbf{DxX} \\
\texttt{X Y ZY}  & \textbf{DxX} \\
\texttt{XZ Y Z}  & \textbf{DxX} \\
\texttt{X XY XY}  & \textbf{AxC} \\
\texttt{XY X XY}  & \textbf{AxE} \\
\texttt{XY XY X}  & \textbf{AxB} \\
\texttt{XYZ X X}  & \textbf{AxC3} \\
\texttt{X XYZ X}  & \textbf{AxE3} \\
\texttt{X X XYZ}  & \textbf{AxB3} \\
\texttt{XYZ X Y}  & \textbf{D-Other} \\
\texttt{X XYZ Y}  & \textbf{D-Other} \\
\texttt{X Y XYZ}  & \textbf{D-Other} \\
\texttt{XY XY XY}  & \textbf{A2} \\
\texttt{? X X}  & \textbf{A?C} \\
\texttt{? X Y}  & \textbf{B?D?E} \\
\texttt{X ? X}  & \textbf{A?E} \\
\texttt{X ? Y}  & \textbf{B?C?D} \\
\texttt{X X ?}  & \textbf{A?B} \\
\texttt{X Y ?}  & \textbf{C?D?E} \\
\hline
\end{tabular}
\caption{Subpatterns of coexpression}
\label{subpatternstab}
\end{table}

\setlength\tabcolsep{4pt}
\begin{longtable}{|p{1cm}|p{3cm}|p{3cm}|p{3cm}|p{1.5cm}|p{2.1cm}|}
\hline
\textbf{iso} & \textbf{name} & \textbf{family} & \textbf{macroarea} & \textbf{pattern} & \textbf{subpattern} \\
\hline
\endfirsthead
\multicolumn{6}{c}%
{\tablename\ \thetable\ -- \textit{Continued from previous page}} \\
\hline
\textbf{ISO} & \textbf{name} & \textbf{family} & \textbf{macroarea} & \textbf{pattern} & \textbf{subpattern} \\ \hline
\endhead
\hline \multicolumn{6}{r}{\textit{Continued on next page}} \\
\endfoot
\endlastfoot
        aai & Arifama-Miniafia & Austronesian & Papunesia & E & E \\ \hline
        aak & Ankave & Angan & Papunesia & A & A \\ \hline
        aau & Abau & Sepik & Papunesia & D & DxX \\ \hline
        aaz & Amarasi & Austronesian & Papunesia & C & C \\ \hline
        abt & Ambulas & Ndu & Papunesia & A & A \\ \hline
        abx & Inabaknon & Austronesian & Papunesia & D & D \\ \hline
        aby & Aneme Wake & Yareban & Papunesia & A & A \\ \hline
        aca & Achagua & Arawakan & South America & C & C \\
\end{longtable}

\bibliography{custom}
\addcontentsline{toc}{chapter}{References}

\end{document}